\documentclass[10pt,twocolumn,letterpaper]{article}

\usepackage{CVPR_style/cvpr}
\usepackage{times}
\usepackage{epsfig}
\usepackage{graphicx}
\usepackage{amsmath}
\usepackage{amssymb}
\usepackage{algorithm}
\usepackage[noend]{algpseudocode}
\usepackage{titletoc}

  \usepackage{color}
  \usepackage[font={small}]{subcaption}
  \usepackage[font={small}]{caption}
  \usepackage{tabularx}
  
\usepackage[pagebackref=true,breaklinks=true,letterpaper=true,colorlinks,bookmarks=false]{hyperref}

\cvprfinalcopy 

\def\cvprPaperID{25} 

\ifcvprfinal\fi
\begin{document}

\title{Fast Optical Flow using Dense Inverse Search}

\author{Till Kroeger\textsuperscript{1}  \qquad Radu Timofte\textsuperscript{1} \qquad Dengxin Dai\textsuperscript{1} \qquad Luc Van Gool\textsuperscript{1,2}\\
\textsuperscript{1}Computer Vision Laboratory, D-ITET, ETH Zurich\\
\textsuperscript{2}VISICS / iMinds, ESAT, K.U. Leuven\\
{\tt\small \{kroegert, timofter, dai, vangool\}@vision.ee.ethz.ch}
}

\maketitle

\begin{abstract}
Most recent works in optical flow extraction focus on the accuracy and neglect the time complexity.
However, in real-life visual applications, such as tracking, activity detection and recognition, the time complexity is critical.

We propose a solution with very low time complexity and competitive accuracy for the computation of dense optical flow.
It consists of three parts: 
1) inverse search for patch correspondences; 
2) dense displacement field creation through patch aggregation along multiple scales;
3) variational refinement. 
At the core of our \emph{Dense Inverse Search}-based method (DIS) is the efficient search of correspondences inspired by the inverse compositional image alignment proposed by Baker and Matthews~\cite{Baker-CVPR-2001,Baker-IJCV-2004} in 2001.

DIS is competitive on standard optical flow benchmarks with large displacements.
DIS runs at 300Hz up to 600Hz on a single CPU core\footnote{1024$\times$436 resolution. 42Hz / 46Hz when including preprocessing: disk access, image re-scaling gradient computation. More details in \S~\ref{ssc:implementation}}, reaching the temporal resolution of human's biological vision system~\cite{benosman2014event}.
It is order(s) of magnitude faster than state-of-the-art methods in the same range of accuracy, making DIS ideal for visual applications.
\vskip-7pt
\end{abstract}

\section{Introduction}

\label{sec:introduction}
Optical flow estimation is under constant pressure to increase both its quality and speed. Such progress allows for new applications. A higher speed enables its inclusion into larger systems with extensive subsequent processing (\eg reliable features for motion segmentation, tracking or action/activity recognition) and its deployment in computationally constrained scenarios (\eg embedded system, autonomous robots, large-scale data processing).

A robust optical flow algorithm should cope with discontinuities (outliers, occlusions, motion discontinuities), appearance changes (illumination, chromacity, blur, deformations), and large displacements.
Decades after the pioneering research of Horn and Schunck~\cite{Horn-1981} and Lucas and Kanade~\cite{Lucas-IJCAI-1981} we have solutions for the first two issues~\cite{Black-CVIU-1996,Papenberg-IJCV-2006} and recent endeavors lead to significant progress in handling large displacements~\cite{Steinbrucker-ICCV-2009, Brox-PAMI-2011, BrauxZin-ICCV-2013, Leordeanu-ICCV-2013, Timofte-WACV-2015,Kennedy-EMMCVPR-2015,Menze-PR-2015,Revaud-CVPR-2015,Bailer-CoRR-2015,Weinzaepfel-ICCV-2013,Wills-IJCV-2006,Wulff-CVPR-2015,Xu-PAMI-2012, Fischer-ICCV-2015}. 
This came at the cost of high run-times usually not acceptable in computationally constrained scenarios such as real-time applications.
Recently, only very few works aimed at balancing accuracy and run-time in favor of efficiency~\cite{Farneback-IA-2003,tao2012simpleflow,Wulff-CVPR-2015}, or employed massively parallelized dedicated hardware to achieve acceptable run-times~\cite{Bao-TIP-2014,Plyer-RTIP-2014,Fischer-ICCV-2015}.
In contrast to this, recently it has been noted for several computer vision tasks~\cite{handa2012real,Srinivasan2013High,benosman2014event,barranco2014contour}, that it is often desirable to trade-off powerful but complex algorithms for simple and efficients methods, and rely on high frame-rates and smaller search spaces for good accuracy.

\begin{figure}\setlength{\tabcolsep}{0.1pt}\renewcommand{\arraystretch}{0}
        \centering
        \resizebox{\columnwidth}{!}
        {
        \begin{tabular}{cc}
        \begin{tabular}{c}
        \includegraphics[width=0.65\columnwidth]{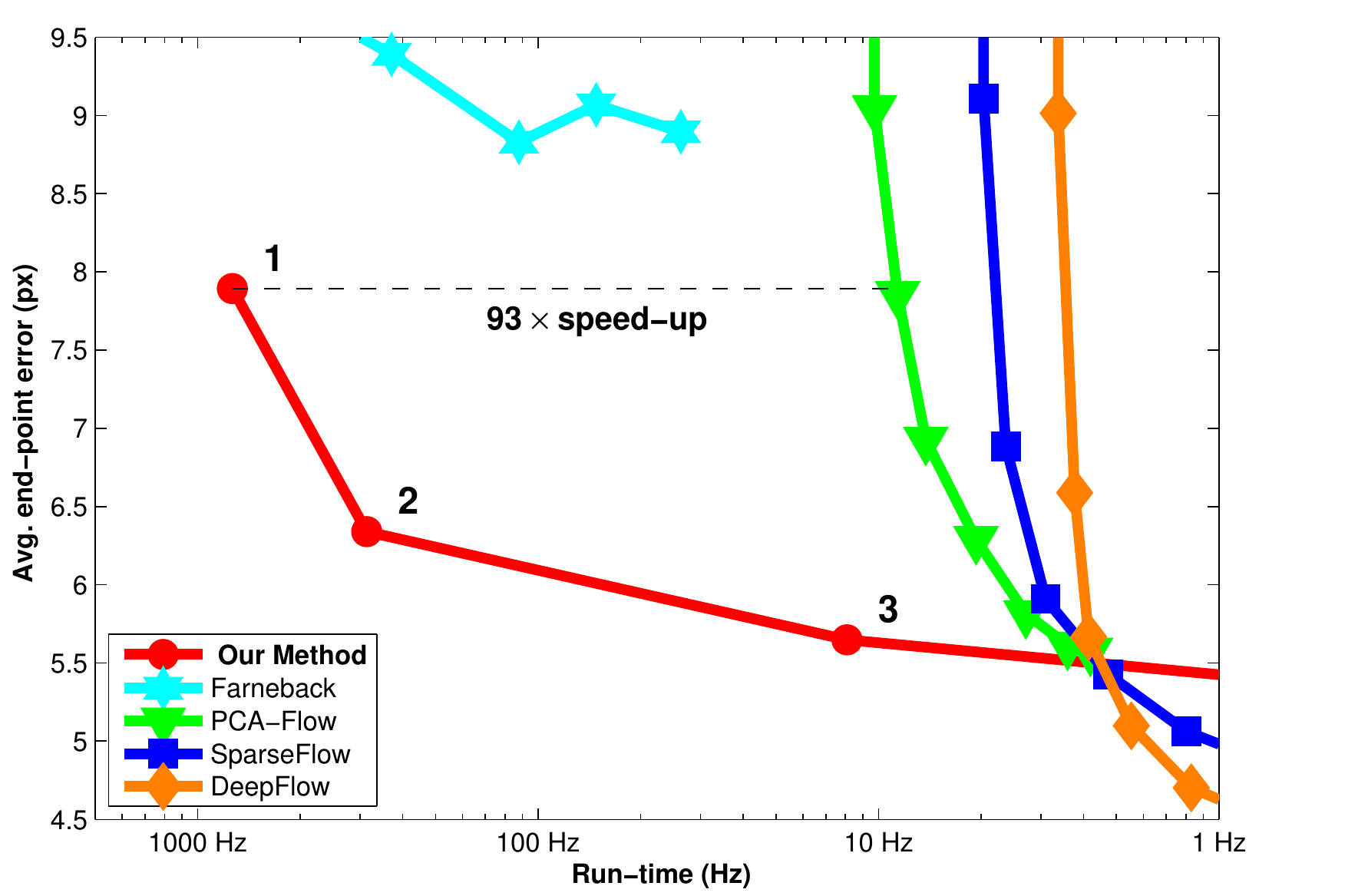} 
        \end{tabular}
        &
        \begin{tabular}{c}
        \includegraphics[width=0.2\columnwidth]{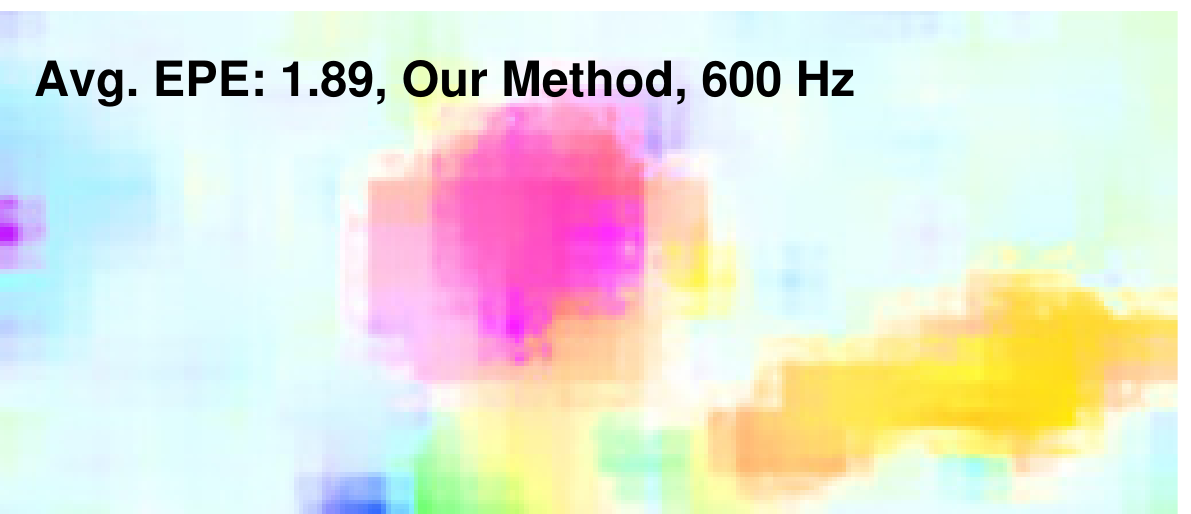}\\
        \tiny (1) DIS @ 600Hz \\
        \includegraphics[width=0.2\columnwidth]{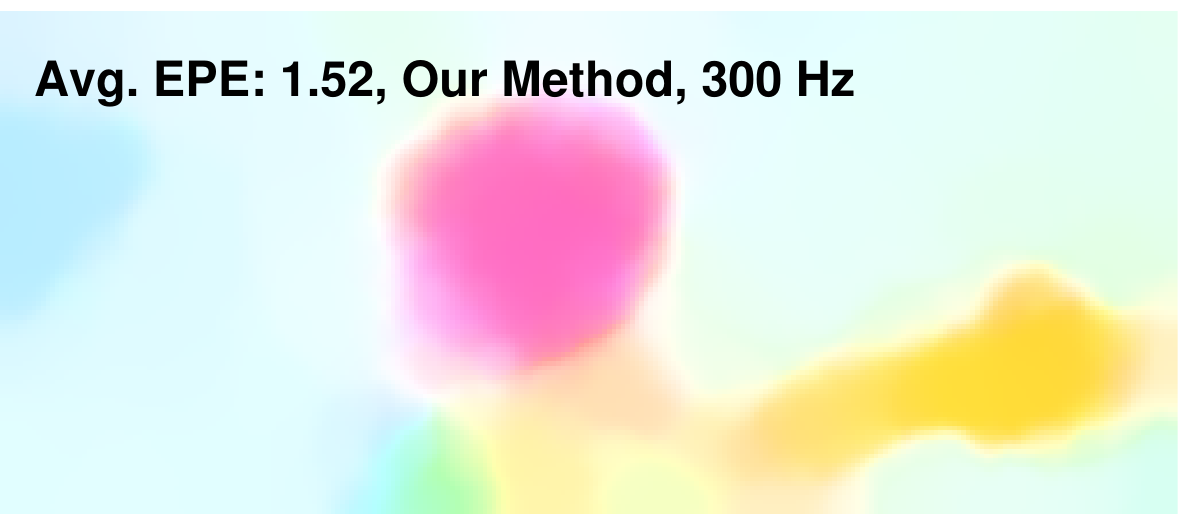}\\
        \tiny (2) DIS @ 300Hz\\
        \includegraphics[width=0.2\columnwidth]{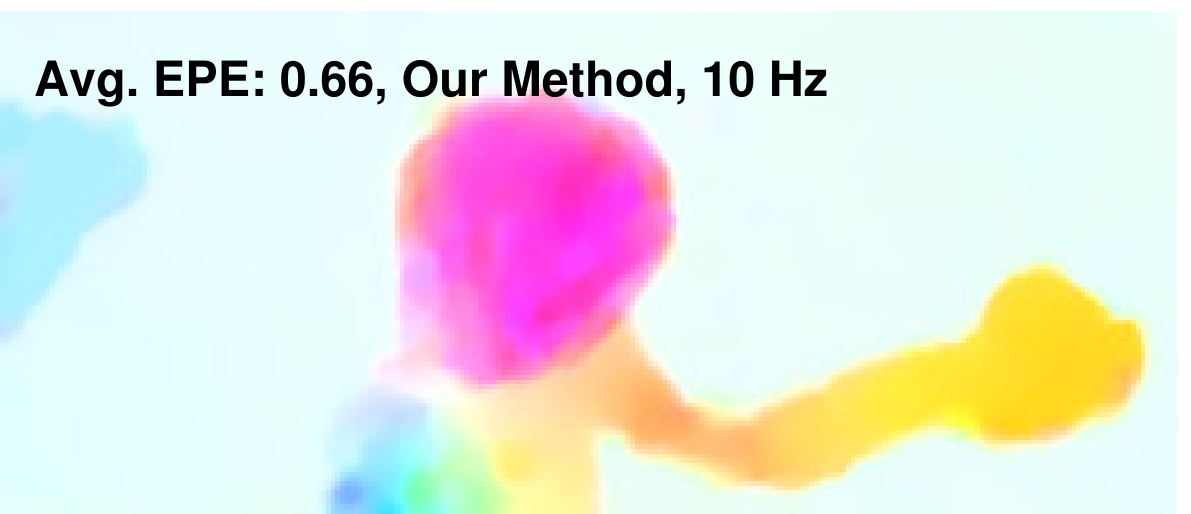}\\      
        \tiny (3) DIS @ 10Hz\\
        \includegraphics[width=0.2\columnwidth]{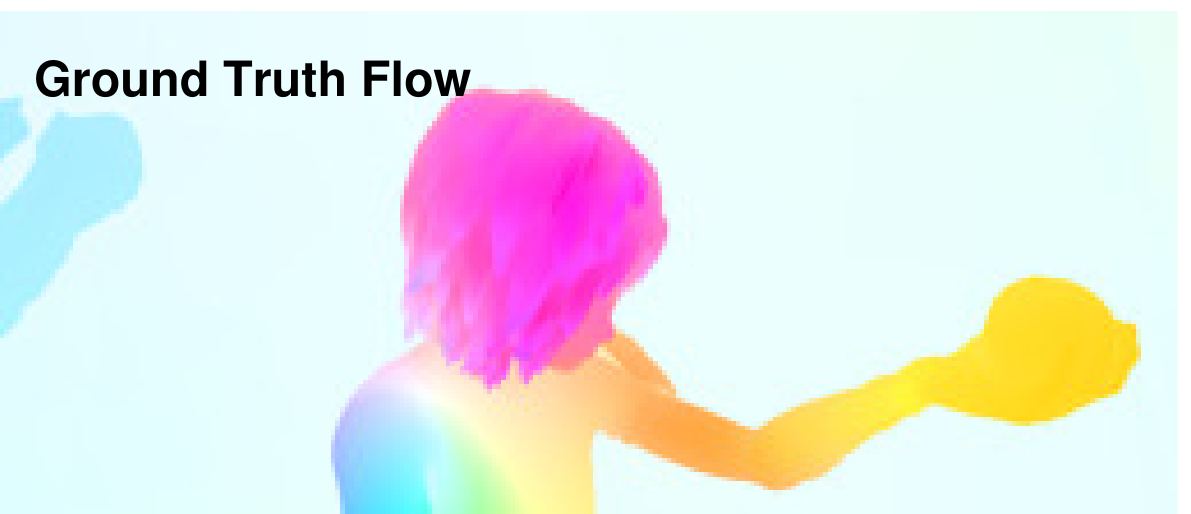}\\
        \tiny Ground Truth\\
        \end{tabular}
        \end{tabular}
        }
        \caption{Our DIS method runs at 10Hz up to 600Hz on a single core CPU for an average end-point pixel error smaller or similar to top optical flow methods at comparable speed. This plot excludes preprocessing time for \emph{all} methods. Details in \S~\ref{ssc:implementation},\ref{ssc:mpi_sintel}. }
        \label{fig:intro}
\vskip-7pt        
\end{figure} 

In this paper we focus on improving the speed of optical flow in general, non-domain-specific scenarios, while remaining close to the state-of-the-art flow quality. We propose two novel components with low time complexity, one using inverse search for fast patch correspondences, and one based on multi-scale aggregation for fast dense flow estimation. Additionally, a fast variational refinement step further improves the solution of our \emph{dense inverse search}-based method.
Altogether, we obtain speed-ups of 1-2 orders of magnitude over state-of-the-art methods at comparable flow quality operating points (Fig.~\ref{fig:intro}).
The run-times are in the range of 10-600 Hz on 1024$\times$436 resolution images by using a single CPU core on a common desktop PC, reaching the temporal resolution of human's biological vision system~\cite{benosman2014event}.
To the best of our knowledge, this is the first time that optical flow at several hundred frames-per-second has been reached with such high flow quality on any hardware.

\subsection{Related work}
\label{ssc:related_work}
Providing an exhaustive overview~\cite{Fortun-CVIU-2015} of optical flow estimation is beyond the scope of this paper.
Most of the work on improving the time complexity (without trading-off quality) combines some of the following ideas:

While, initially, the \textbf{feature descriptors} of choice were extracted sparsely, invariant under scaling or affine transformations~\cite{Mikolajczyk-IJCV-2005}, the recent trend in optical flow estimation is to densely extract rigid (square) descriptors from local frames~\cite{Tola-CVPR-2008,Brox-PAMI-2011,Liu-PAMI-2011}.
HOG~\cite{Dalal-CVPR-2005}, SIFT~\cite{Lowe-IJCV-2004}, and SURF~\cite{Bay-CVIU-2008} are among the most popular square patch support descriptors. 
In the context of scene correspondence, the SIFT-flow~\cite{Liu-PAMI-2011} and PatchMatch~\cite{Barnes-ECCV-2010} algorithms use descriptors or small patches.
The descriptors are invariant only to similarities which may be insufficient especially for large displacements and challenging deformations~\cite{Brox-PAMI-2011}.

The {\bf feature matching} usually employs a (reciprocal) nearest neighbor operation~\cite{Lowe-IJCV-2004,Barnes-ECCV-2010,Brox-PAMI-2011}.
Important exceptions are the recent works of Weinzaepfel~\etal~\cite{Weinzaepfel-ICCV-2013} (non-rigid matching inspired by deep convolutional nets), of Leordeanu~\etal~\cite{Leordeanu-ICCV-2013} (enforcing affine constraints), and of Timofte~\etal~\cite{Timofte-WACV-2015} (robust matching inspired by compressed sensing).
They follow Brox and Malik~\cite{Brox-PAMI-2011} and guide a variational optical flow estimation through (sparse) correspondences from the descriptor matcher and can thus handle arbitrarily large displacements.
Xu~\etal\cite{Xu-PAMI-2012} combine SIFT~\cite{Lowe-IJCV-2004} and PatchMatch~\cite{Barnes-ECCV-2010} matching for refined flow level initialization at the expense of computational costs.

An \textbf{optimization} problem is often at the core of the flow extraction methods. The flow is estimated by minimizing an energy that sums up matching errors and smoothness constraints.
While Horn and Schunck~\cite{Horn-1981} proposed a variational approach to globally optimize the flow, Lucas and Kanade~\cite{Lucas-IJCAI-1981} solve the correspondence problem for each image patch locally and independently. 
The local~\cite{Lucas-IJCAI-1981,tao2012simpleflow,senst2012robust} methods are usually faster but less accurate than the global ones.
Given location and smoothness priors over the image, MRF formulations have been proposed~\cite{Heitz-PAMI-1993,Szeliski-PAMI-2008}.

{\bf Parallel computation} is a natural way of improving the run-time of the optical flow methods by (re)designing them for parallelization.
The industry historically favored specialized hardware such as 
FPGAs~\cite{Pauwels-TOC-2012}, while the recent years brought the advance of GPUs~\cite{Plyer-RTIP-2014,Bao-TIP-2014,Fischer-ICCV-2015,Zach07aduality}.
Yet, multi-core design on the same machine is the most common parallelization. 
However, many complex flow methods are difficult to adapt for parallel processing.

\textbf{Learning. }
Most of the optical flow methods exploit training images for parameter tuning.
However, this is only a rough embedding of prior knowledge. 
Only recently methods were proposed that successfully learn specific models from such training material.
Wulff~\etal~\cite{Wulff-CVPR-2015} assume that any optical flow field can be approximated by a decomposition over a small learned basis of flow fields.
Fischer~\etal~\cite{Fischer-ICCV-2015} construct Convolutional Neural Networks (CNNs) capable to solve the optical flow estimation. 

{\bf Coarse-to-fine optimizations} have been applied frequently to flow estimation~\cite{Enkelmann1988,Brox-PAMI-2011} to avoid poor local minima, especially for large motions, and thus to improve the performance and to speed up the convergence.

\textbf{Branch and bound} and \textbf{Priority queues} have been used to find smart strategies to first explore the flow in the most favorable image regions and gradually refine it for the more ambiguous regions. 
This often leads to a reduction in computational costs.
The PatchMatch method of Barnes~\etal~\cite{Barnes-ECCV-2010} follows a branch and bound strategy, gradually fixing the most promising correspondences. Bao~\etal~\cite{Bao-TIP-2014} propose the EPPM method based on PatchMatch.

\textbf{Dynamic Vision Sensors}~\cite{lichtsteiner2008128}, asynchronously capturing illumination changes at microsecond latency, have recently been used to compute optical flow. Benosman~\cite{benosman2014event} and Barranco~\cite{barranco2014contour} note that realistic motion estimation tasks, even with large displacements, become simple when capturing image evidence with speed in the kilohertz-range.

\subsection{Contributions}
\label{ssc:contributions}
We present a novel optical flow method based on dense inverse search (DIS), which we demonstrate to provide high quality flow estimation at 10-600 Hz on a single CPU core. This method is 1-2 orders of magnitude times faster than previous results~\cite{Weinzaepfel-ICCV-2013,Timofte-WACV-2015,Wulff-CVPR-2015} on the Sintel~\cite{Butler-ECCV-2012} and KITTI~\cite{Geiger-IJRR-2013} datasets when considering all methods at comparable flow quality operating points.
At the same time it is significantly more accurate compared to existing methods running at \emph{equal speed}~\cite{Farneback-IA-2003,Lucas-IJCAI-1981}. This result is based on two main contributions:

\textbf{Fast inverse search for correspondences. }
Inspired by the inverse compositional image alignment of Baker and Matthews~\cite{Baker-CVPR-2001,Baker-IJCV-2004} we devise our inverse search procedure (explained in \S~\ref{ssc:inverse_search}) for fast mining of a grid of patch-based correspondences between two input images. While usually less robust than exhaustive feature matching, we can extract a uniform grid of correspondences in microseconds.

\textbf{Fast optical flow with multi-scale reasoning. }
Many methods assume sparse and outlier-free correspondences, and rely heavily on variational refinement to extract pixel-wise flow~\cite{Weinzaepfel-ICCV-2013,Timofte-WACV-2015}. This helps to smooth-out small errors, and cover regions with flat and ambigious textures, where exhaustive feature matching often fails.
Other methods rely directly on pixel-wise refinement~\cite{Plyer-RTIP-2014,Bao-TIP-2014}. 
We chose a middle ground and propose a very fast and robust patch-averaging-scheme, performed only once per scale, after grid-based correspondences have been extracted. This step gains robustness against outlier correspondences, and initializes a pixel-wise variational refinement, performed once per scale. 
We reach an optimal trade-off between accuracy and speed at 300Hz on a single CPU core, and reach 600Hz without variational refinement at the cost of accuracy. Both operating points are marked as {\bf (2)} and {\bf (1)} in Fig.~\ref{fig:intro}, \ref{fig:resultsintel}, \ref{fig:resultkitti}.

Related to our approach is \cite{Plyer-RTIP-2014}. Here, the inverse image warping idea~\cite{Baker-IJCV-2004} is used on \emph{all} the pixels, while our method optimizes patches independently. In contrast to our densification done once per scale, they rely on frequent flow interpolations, which requires a high-powered GPU, and still is significantly slower than our CPU-only approach. 

The structure of the paper is as follows:
In \S~\ref{sec:proposed_method} we introduce our DIS method. In \S~\ref{sec:experiments} we describe the experiments, separately evaluate the patch-based correspondence search, and analyse the complete DIS algorithm with and without the variational refinement. In \S~\ref{sec:conclusions} we conclude the paper.

\section{Proposed method}
\label{sec:proposed_method}
In the following, we introduce our dense inverse search-based method (DIS) by describing: how we extract single point correspondences between two images in \S~\ref{ssc:inverse_search}, how we merge a set of noisy point correspondences on each level $s$ of a scale-pyramid into a dense flow field ${\bf U}_s$ in \S~\ref{ssc:flow}, and how we refine ${\bf U}_s$ using variational refinement in \S~\ref{ssc:TV}.

\subsection{Fast inverse search for correspondences}
\label{ssc:inverse_search} 
The core component in our method to achieve high performance is the efficient search for patch correspondences. 
In the following we will detail how we extract one single point correspondence between two frames.

For a given template patch $T$ in the reference image $I_t$, with a size of $\theta_{ps} \times \theta_{ps}$ pixels, centered on location ${\bf x} = (x,y)^T$, we find the best-matching sub-window of $\theta_{ps} \times \theta_{ps}$ pixels in the query image $I_{t+1}$ using gradient descent. We are interested in finding a warping vector ${\bf u}=(u,v)$ such that we minimize the sum of squared differences over the sub-window between template and query location:
\begin{align}
{\bf u} = \text{argmin}_{ {\bf u'}}\; \sum_{x} \left[ I_{t+1}({\bf x}+{\bf u'}) - T({\bf x}) \right]^2 . \label{eq:minnorm}
\end{align}
Minimizing this quantity is non-linear and is optimized iteratively using the inverse Lukas-Kanade algorithm as proposed in~\cite{Baker-IJCV-2004}. For this method two steps are alternated for a number of iterations or until the quantity \eqref{eq:minnorm} converges. For the first step, the quantity \eqref{eq:minnorm2} is minimized around the current estimate ${\bf u}$ for an update vector $\Delta {\bf u}$ such that
\begin{align}
\Delta {\bf u} = \text{argmin}_{\Delta {\bf u'}} \sum_{x} \left[ I_{t+1}({\bf x}+{\bf u} + \Delta {\bf u'}) - T({\bf x}) \right]^2 \label{eq:minnorm2}
\end{align}
The first step requires extraction and bilinear interpolation of a sub-window $I_{t+1}({\bf x}+{\bf u})$ for sub-pixel accurate warp updates. The second step updates the warping ${\bf u} \leftarrow {\bf u}+  \Delta {\bf u}$.

The original Lukas-Kanade algorithm \cite{Lucas-IJCAI-1981} required expensive re-evaluation of the Hessian of the image warp at every iteration. 
As proposed in~\cite{Baker-IJCV-2004} the inverse objective function $\sum_{x} \left[T({\bf x}-\Delta {\bf u}) - I_{t+1}({\bf x}+{\bf u}) \right]^2$ can be optimized instead of \eqref{eq:minnorm2}, removing the need to extract the image gradients for $I_{t+1}({\bf x}+{\bf u})$ and to re-compute the Jacobian and Hessian at every iteration. 
Due to the large speed-up this inversion has successfully been used for point tracking in SLAM~\cite{Klein2007}, camera pose estimation \cite{Forster2014}, and is covered in detail in \cite{Baker-IJCV-2004} and our supplementary material.

In order to gain some robustness against absolute illumination changes, we mean-normalize each patch.

One challenge of finding sparse correspondences with this approach is that the true displacements cannot be larger than the patch size $\theta_{ps}$, since the gradient descent is dependent on similar image context in both patches. 
Often a coarse-to-fine approach with fixed window-size but changing image size is used~\cite{Klein2007,Forster2014}, firstly, to incorporate larger smoothed contexts at coarser scales and thereby lessen the problem of falling into local optima, secondly, to find larger displacements, and, thirdly, to ensure fast convergence.

\subsection{Fast optical flow with multi-scale reasoning}
\label{ssc:flow}
We follow this multi-scale approach, but, instead of optimizing patches independently, we compute an intermediate dense flow field and re-initialize patches at each level. This is because of two reasons: 1) the intermediate dense flow field smooths displacements and provides robustness, effectively filtering outliers and 2) it reduces the number of patches on coarser scales, thereby providing a speed-up.
We operate in a coarse-to-fine fashion from a first (coarsest) level $\theta_{ss}$ in a scale pyramid with a downscaling quotient of $\theta_{sd}$ to the last (finest) level $\theta_{sf}$. On each level our method consists of five steps, summarized in algorithm~\ref{alg:DISalgo}, yielding a dense flow field ${\bf U}_s$ in each iteration~$s$.

\begin{algorithm}[t!]
\caption{\quad {\bf D}ense {\bf I}nverse {\bf S}earch {\bf (DIS)}} \label{alg:DISalgo}
\begin{algorithmic}[1]
\State Set initial flow field ${\bf U}_{\theta_{ss}+1} \leftarrow {\bf 0}$
\For{$s=\theta_{ss}$ to $\theta_{sf}$}
\State ({\bf 1.}) Create uniform grid of $N_s$ patches
\State ({\bf 2.}) Initialize displacements from ${\bf U}_{s+1}$
\For{$i=1$ to $N_s$}
\State ({\bf 3.}) Inverse search for patch $i$
\EndFor
\State ({\bf 4.}) Densification: Compute dense flow field ${\bf U}_{s}$
\State ({\bf 5.}) Variational refinement of ${\bf U}_{s}$
\EndFor
\end{algorithmic}
\end{algorithm}

{\bf (1.) Creation of a grid: } We initialize patches in a uniform grid over the image domain. The grid density and number of patches $N_s$ is implicitly determined by the parameter $\theta_{ov} \in [0,1)$ which specifies the overlap of adjacent patches and is always floored to an integer overlap in pixels. A value of $\theta_{ov}=0$ denotes a patch adjacency with no overlap and $\theta_{ov}=1-\epsilon$ results in a dense grid with one patch centered on each pixel in the reference image.

{\bf (2.) Initialization: } For the first iteration ($s = \theta_{ss}$) we initialize all patches with the trivial zero flow. On each subsequent scale $s$ we initialize the displacement of each patch $i \in N_s$ at its location ${\bf x}$ with the flow from the previous (coarser) scale: ${\bf u}_{i,\text{init}} = {\bf U}_{s+1}({\bf x} / \theta_{sd}) \cdot \theta_{sd}$.  

{\bf (3.) Inverse search: } Optimal displacements are computed independently for all patches as detailed in \S~\ref{ssc:inverse_search}.

{\bf (4.) Densification: }
After step three we have updated displacement vectors ${\bf u}_i$. For more robustness against outliers, we reset all patches to their initial flow ${\bf u}_{i,\text{init}}$ for which the displacement update $\lVert {\bf u}_{i,\text{init}}-{\bf u}_i\rVert_2$ exceeds the patch size $\theta_{ps}$. We create a dense flow field ${\bf U}_s$ in each pixel ${\bf x}$ by applying weighted averaging to displacement estimates of all patches overlapping at ${\bf x}$ in the reference image:
\begin{align}
{\bf U}_s({\bf x}) = \frac{1}{Z} \; \sum_i^{N_s} \frac{\lambda_{i,{\bf x}}}{\max(1,\lVert d_i({\bf x}) \rVert_2)} \cdot {\bf u}_i \; \; \; , \label{eq:avg}
\end{align}
where the indicator $\lambda_{i,{\bf x}}=1$ iff patch $i$ overlaps with location ${\bf x}$ in the reference image, $d_i({\bf x}) = I_{t+1}({\bf x}+{\bf u}_i) - T({\bf x})$ denotes the intensity difference between template patch and warped image at this pixel, ${\bf u}_i$ denotes the estimated displacement of patch $i$, and normalization $Z =  \sum_i \lambda_{i,{\bf x}}/\max(1,\lVert d_i({\bf x}) \rVert_2)$.

{\bf (5.) Variational Refinement: } The flow field ${\bf U_s}$ is refined using energy minimization as detailed in \S~\ref{ssc:TV}.

\subsection{Fast Variational refinement}
\label{ssc:TV}
We use a simplified variant of the variational refinement of \cite{Weinzaepfel-ICCV-2013} without a feature matching term and intensity images only. We refine only on the current scale.
The energy is a weighted sum of intensity and gradient data terms ($E_{I}$, $E_{G}$) and a smoothness term ($E_S$) over the image domain $\Omega$:
\begin{align}
E({\bf U}) = \int_{\Omega} \sigma \, \Psi(E_I) + \gamma \, \Psi(E_G) + \alpha \, \Psi(E_S) \; d{\bf x}
\end{align}
We use a robust penalizer $\Psi(a^2)= \sqrt{a^2 + \epsilon^2}$, with $\epsilon=0.001$ for all terms as proposed in \cite{Sun-CVPR-2010}.

We use a separate penalization of intensity and gradient constancy assumption, with normalization as proposed in  \cite{Zimmer-IJCV-2011}: 
With the brightness constancy assumption $(\nabla_3^T I) {\bf u} = 0$, where $\nabla_3 = (\partial x, \partial y, \partial z)^T$ denotes the spatio-temporal gradient, we can model the intensity data term as $E_I={\bf u}^T \, \bar{{\bf J}}_0 \, {\bf u}$. We use the normalized tensor $\bar{{\bf J}}_0 = \beta_0 \, (\nabla_3 I)(\nabla_3^T I)$ to enforce brightness constancy, with normalization $\beta_0 = (\lVert \nabla_2 I\rVert^2 + 0.01)^{-1}$ by the spatial derivatives and a term to avoid division by zero as in \cite{Zimmer-IJCV-2011}.

Similarly, $E_G$  penalizes the gradient constancy: $E_G={\bf u}^T \, \bar{{\bf J}}_{xy} \, {\bf u}$ with $\bar{{\bf J}}_{xy} = \beta_x (\nabla_3 I_{dx})(\nabla_3^T I_{dx}) + \beta_y (\nabla_3 I_{dy})(\nabla_3^T I_{dy})$, and normalizations $\beta_x = (\lVert \nabla_2 I_{dx}\rVert^2 + 0.01)^{-1}$ and $\beta_y = (\lVert \nabla_2 I_{dy}\rVert^2 + 0.01)^{-1}$.

The smoothness term is a penalization over the norm of the gradient of displacements: $E_S = \lVert \nabla u \rVert^2 + \lVert \nabla v \rVert^2$.

The non-convex energy $E({\bf U})$ is minimized iteratively with $\theta_{vo}$ fixed point iterations and $\theta_{vi}$ iterations of Successive Over Relaxation (SOR) for the linear system, as in~\cite{Brox-ECCV-2004}. 

\subsection{Extensions}
\label{ssc:extensions}
\noindent Our method lends itself to five extensions as follows:

{\bf i. Parallelization} of all time-sensitive parts of our method (step 3, 5 in \S~\ref{ssc:flow}) is trivially achievable, 
since patch optimization operates independently, 
and in the variational refinement the linear systems per pixel are independent as well.
We parallelized our implementation with OpenMP and received an almost linear speed-up with number of cores. However, since for fast run-times the overhead of thread creation is significant, we used only a single core in our experiments. Parallelization will be useful in online-scenarios where thread creation is handled once at the start.

{\bf ii. Using RGB color} images, instead of intensity only, boosts the score in most top-performing optical flow methods. In our experiments, we found that using color is not worth the observed increase of the run-time.

{\bf iii. Enforcing forward-backward consistency} of flow can boost accuracy. Similarly to using color, we found that the boost is not worth the observed doubling of the run-time.


{\bf iv. Robust error norms}, such as L$1$ and the Huber-norm\cite{Werlberger2009a}, can be used instead of the L$2$-norm, implicit in the optimization of \eqref{eq:minnorm}. Experimentally, we found that the gained robustness is not worth the slower convergence.

{\bf v. Using DIS for stereo depth}, requires the estimation of the horizontal displacement of each pixel. Removing the vertical degree of freedom from our method is trivial. 

See the supplementary material for experiments on {\bf i.-v.}

\section{Experiments}
\label{sec:experiments}
In order to evaluate the performance of our method, we present three sets of experiments.
Firstly, we conduct an analysis of our parameter selection in \S~\ref{ssc:implementation}. Here, we also study the impact of variational refinement in our method. 
Secondly, we evaluate the inverse search (step 3 in algorithm~\ref{alg:DISalgo}) in \S~\ref{ssc:eval_inv_search} without densification (step 4). 
The complete pipeline for optical flow is evaluated in \S~\ref{ssc:mpi_sintel}, and \ref{ssc:kittiflow}. 
Thirdly, since the problem of recovering large displacements can also be handled by higher frame-rates combined with lower run-time per frame-pair,  we conduct an experiment in \S~\ref{ssc:framerate} to analyse the benefit of higher frame-rates.

\subsection{Implementation and Parameter Selection}
\label{ssc:implementation}

\begin{figure}
        \centering
        \begin{subfigure}[b]{0.25\textwidth}
        \centering
                \includegraphics[width=0.95\textwidth]{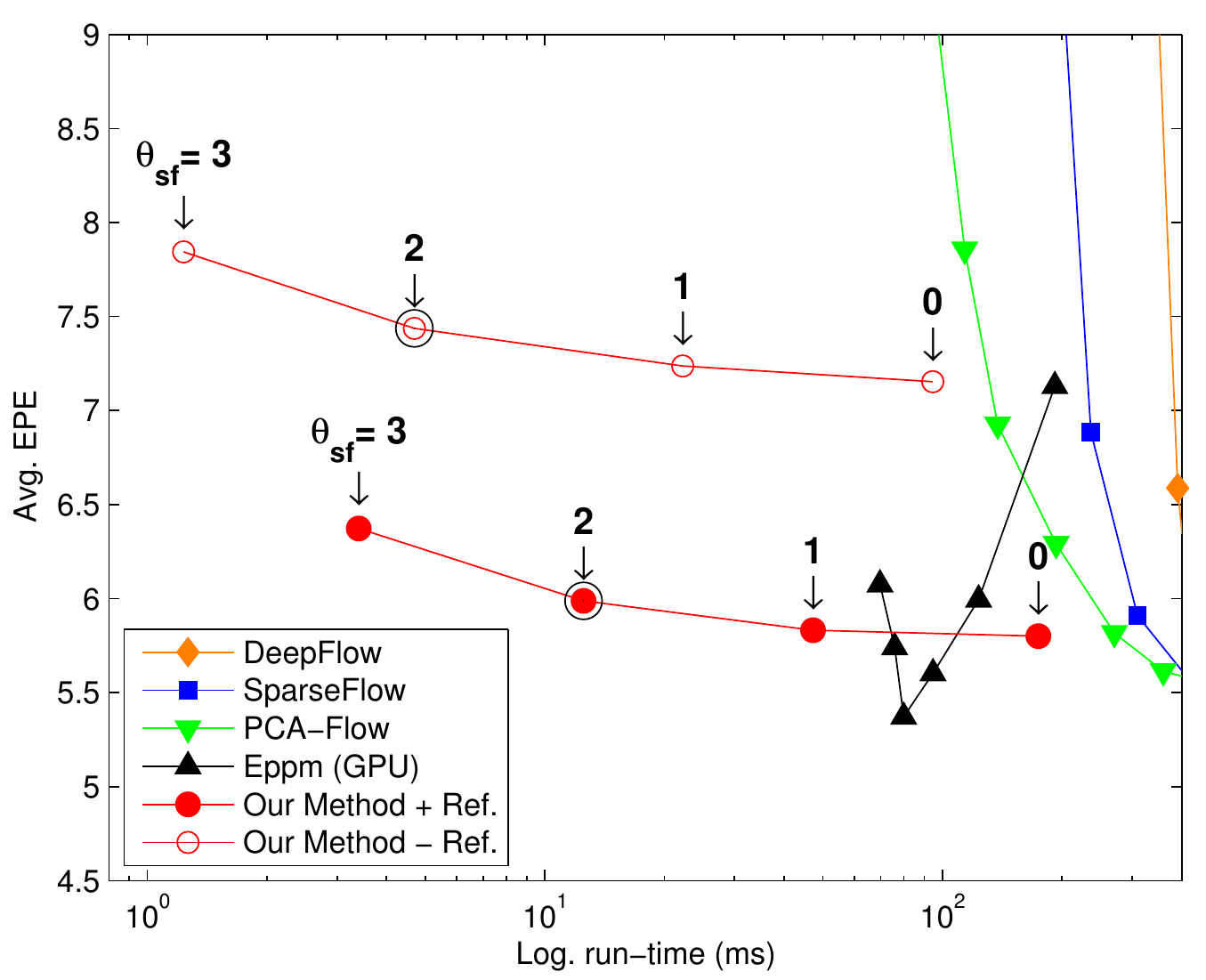}
        \end{subfigure}%
        \begin{subfigure}[b]{0.25\textwidth}
        \centering
                \includegraphics[width=0.95\textwidth]{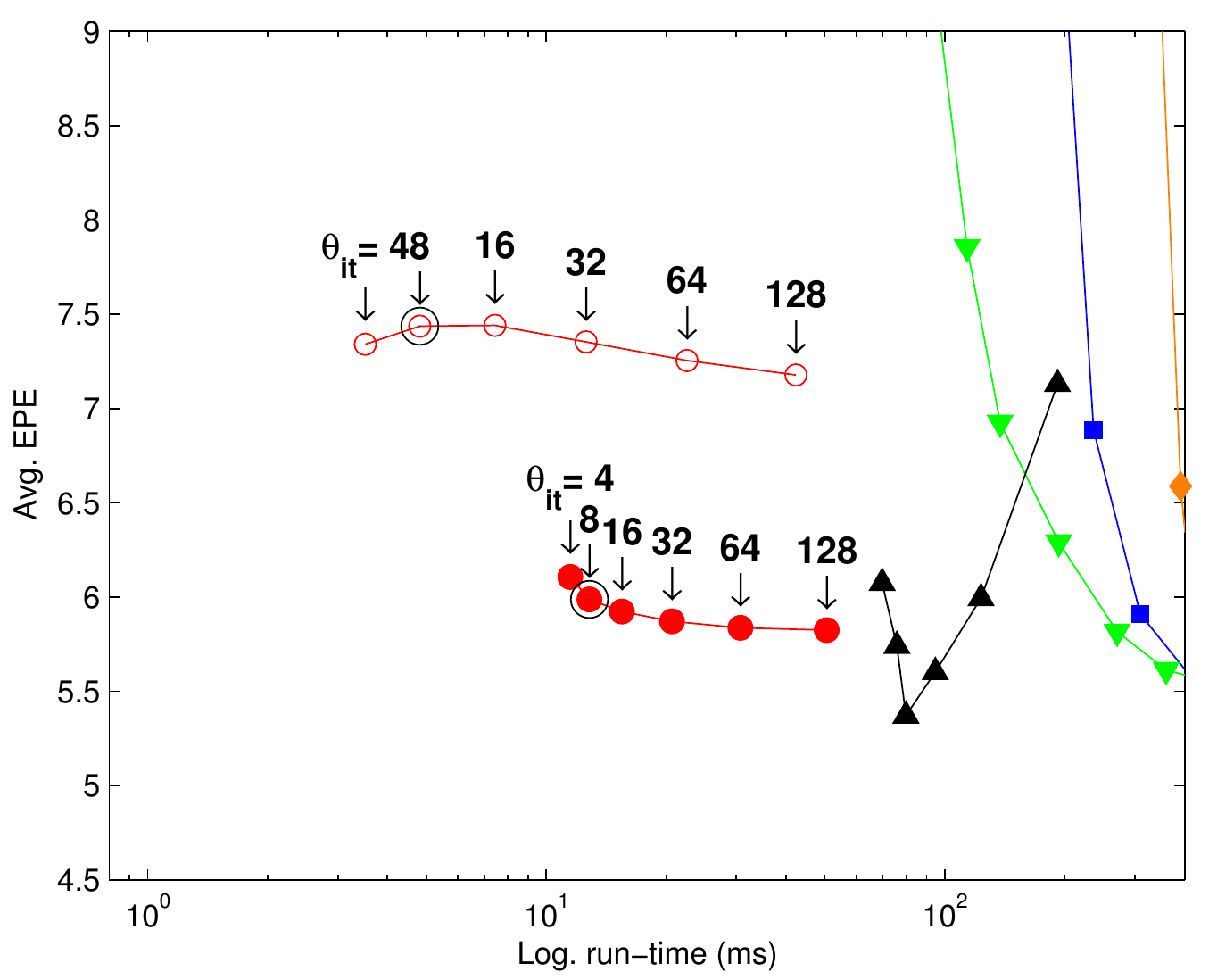}
        \end{subfigure} 
        \begin{subfigure}[b]{0.25\textwidth}
        \centering
                \includegraphics[width=0.95\textwidth]{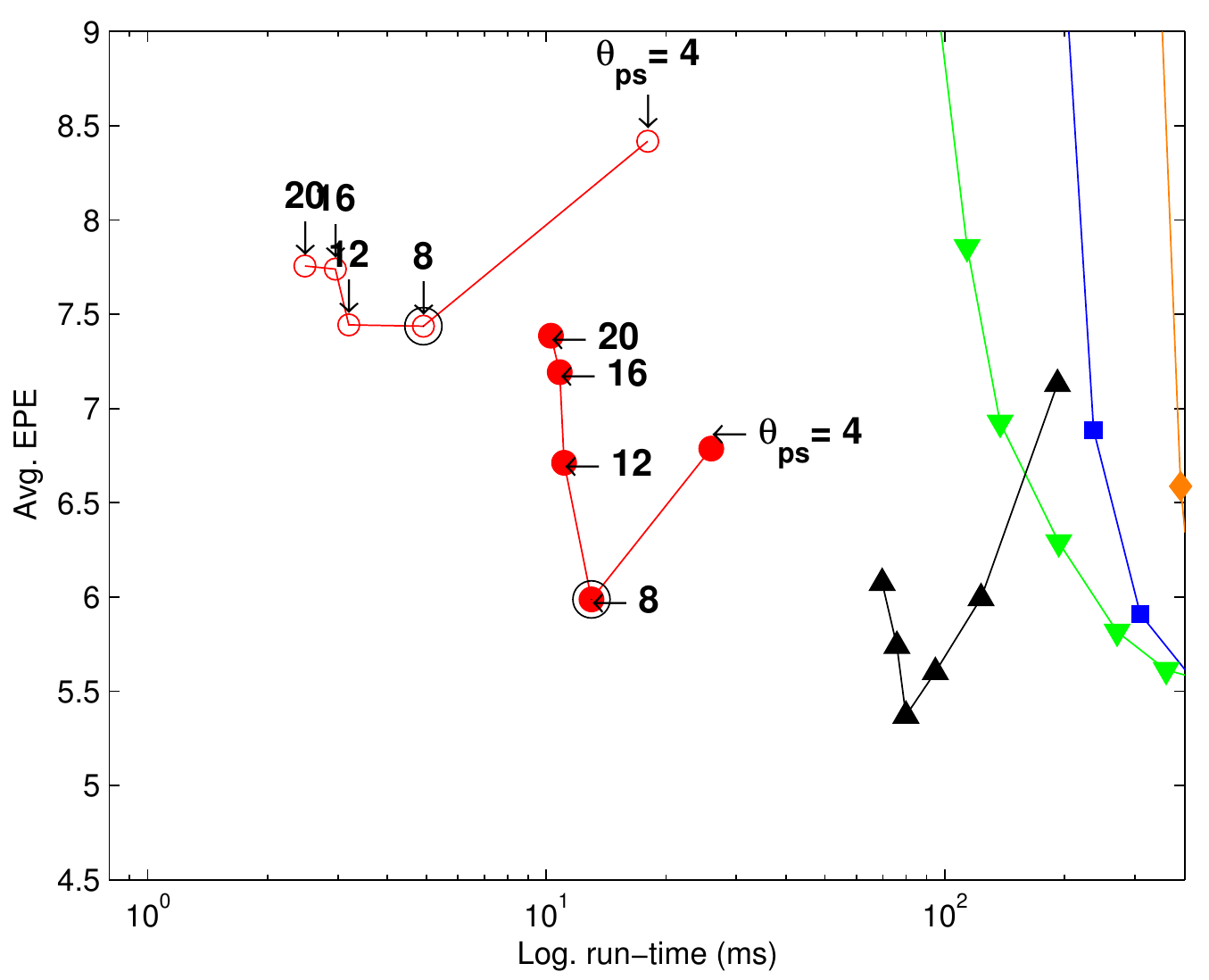}
        \end{subfigure}%
        \begin{subfigure}[b]{0.25\textwidth}
        \centering
                \includegraphics[width=0.95\textwidth]{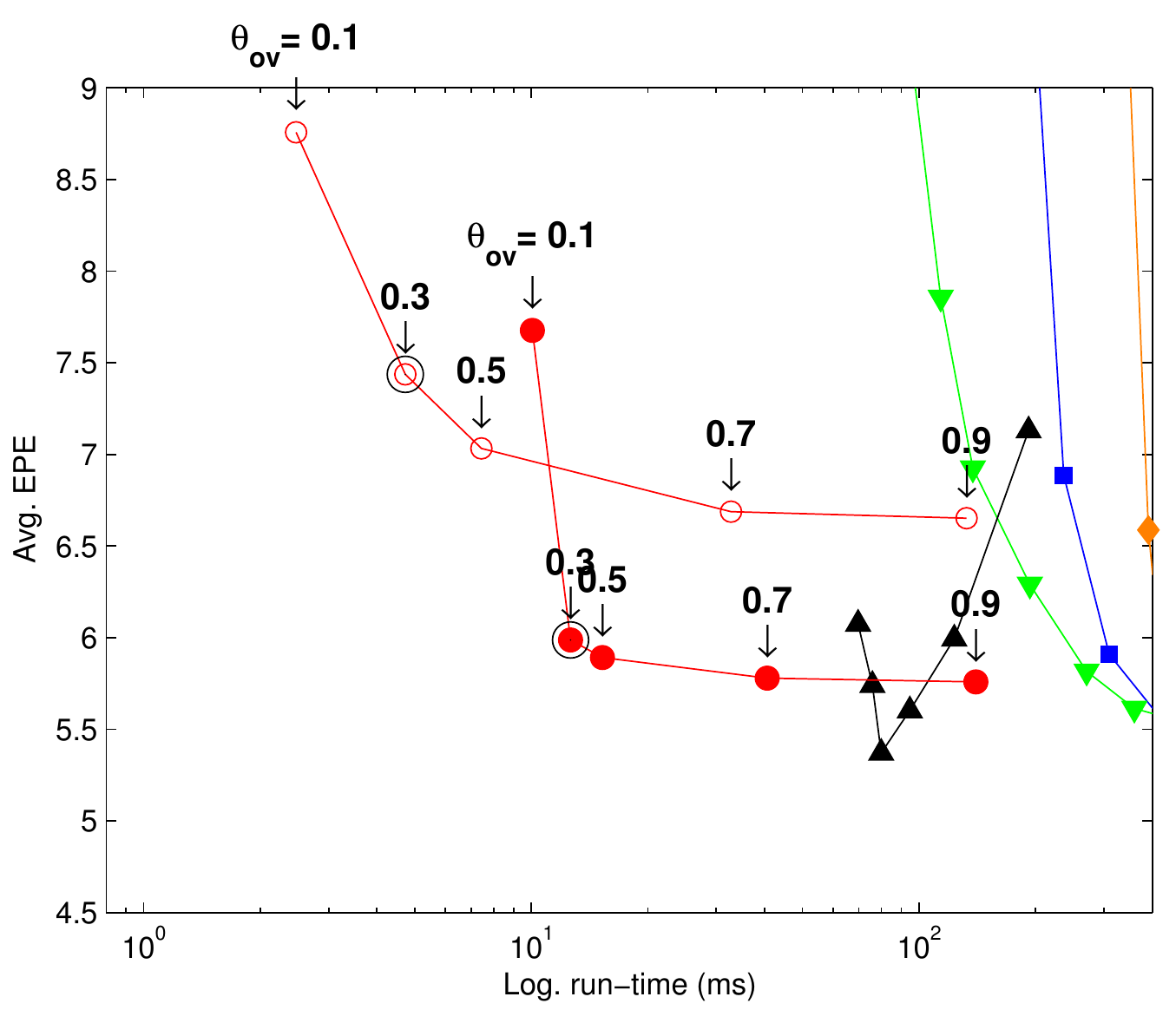}
        \end{subfigure}%
        \caption{Optical Flow result on Sintel with changing parameters. We set $\theta_{sf}=2$, $\theta_{it}=8$, $\theta_{ps}=8$, $\theta_{ov}=0.3$, marked with a black circle in all plots. From the top left to bottom right we vary the parameters $\theta_{sf}$, $\theta_{it}$, $\theta_{ps}$, and $\theta_{ov}$ independently in each plot.}\label{fig:paramcrossval}
        \vskip-5pt
\end{figure} 

We implemented\footnote{The code will be publicly available on the main author's website.} our method in C++ and run all experiments and baselines on a Core i7 CPU using a single core, and a GTX780 GPU for the EPPM~\cite{Bao-TIP-2014} baseline. For \emph{all} experiments on the Sintel and KITTI training datasets we report timings from which we exclude all operations which, in a typical robotics vision application, would be unnecessary, performed only once, or shared between multiple tasks: Disk access, creation of an image pyramid including image gradients with a downsampling quotient of $2$, all initializations of the flow algorithms. We do this for our method and \emph{all} baselines within their provided code. 
For EPPM, where only an executable was available, we subtracted the average overhead time of our method for fair comparison.
Please see the supplementary material for variants of these experiments where preprocessing times are included for all methods. Our method requires 20 ms of preprocessing, spent on disk access  (11~ms), image scaling and gradients (9~ms). 
For experiments on the Sintel and KITTI test datasets we include the preprocessing time to be comparable with reported timings in the online benchmarks.

{\bf Parameter selection.} Our method has four main parameters which affect speed and performance as explained in \S~\ref{sec:proposed_method}: $\theta_{ps}$ size of each rectangular patch, $\theta_{ov}$ patch overlap, $\theta_{it}$ number of iterations for the inverse search, $\theta_{sf}$ finest and final scale on which to compute the flow. 
We plot the change in the \emph{average end-point error} versus \emph{run-time} on the Sintel (\emph{training, final}) dataset~\cite{Butler-ECCV-2012} in Fig. \ref{fig:paramcrossval}. 
We draw three conclusions: 
Firstly, operating on finer scales (lower $\theta_{sf}$), more patch iterations (higher $\theta_{it}$), higher patch density (higher $\theta_{ov}$) generally lowers the error, but, depending on the time budget, may not be worth it. 
Secondly, the patch size $\theta_{ps}$ has a clear optimum at 8 and 12 pixels. This also did not change when varying $\theta_{ps}$ at lower $\theta_{sf}$ or higher $\theta_{it}$. Thirdly, using variational refinement always significantly reduced the error for a moderate increase in run-time.

More implementation details and timings of all parts of algorithm~\ref{alg:DISalgo} can be found in the supplementary material.

\begin{table}
\scriptsize
\centering
\setlength{\tabcolsep}{3pt}
\begin{tabular}{c|c}
 Parameter & Function \\
\hline
$\bf \theta_{sf}$ & {\bf finest scale in multi-scale pyramide} \\
$\bf \theta_{it}$ & {\bf number of gradient descent iterations per patch}\\
$\bf \theta_{ps}$ & {\bf rectangular patch size in (pixel)}\\
$\bf \theta_{ov}$ & {\bf patch overlap on each scale (percent) }\\
$\theta_{sd}$ & downscaling quotient in scale pyramid \\
$\theta_{ss}$ & coarsest scale in multi-scale pyramid \\
$\theta_{vo}, \theta_{vi}$ & number of outer and inner iterations for variational refinement \\
$\delta, \gamma, \alpha$ & intensity, gradient and smoothness weights for variational refinement
\end{tabular}
\caption{Parameters of our method. Parameters in {\bf bold} have a significant impact on performance and are cross-validated in \S~\ref{ssc:implementation}.}
\label{tab:parameters}
\end{table}

In addition we have several parameters of lower importance, which are fixed for all experiments. 
We set $\theta_{sd}=2$, \ie we use a standard image pyramid, where the resolution is halved with each downscaling.
We set the coarsest image scale $\theta_{ss}=5$ for \S~\ref{ssc:mpi_sintel} and $\theta_{ss}=6$ for \S~\ref{ssc:kittiflow} due to higher image resolutions. For different patch sizes and image pyramids the coarsest scale can be selected as $\theta_{ss} = \log_{\theta_{sd}}{ (2 \cdot width) / (f \cdot \theta_{ps})}$ and raised to the nearest integer, to capture motions of at least $1/f$ of the image width. 
For the variational refinement we fix intensity, gradient and smoothness weights as $\delta=5, \gamma=10, \alpha=10$ and keep iteration numbers fixed at $\theta_{vo}=1\cdot (s+1)$, where $s$ denotes the current scale and $\theta_{vi}=5$. In contrast to our comparison baselines \cite{Weinzaepfel-ICCV-2013, Timofte-WACV-2015, Wulff-CVPR-2015}, we do not fine-tune our method for a specific dataset in our experiments. 
We use a 20 percent subset of Sintel training to develop our method, and only the remaining training material is used for evaluation.
All parameters are summarized in Table~ \ref{tab:parameters}.

If the flow is not computed up to finest scale ($\theta_{sf}=0$), we scale-up the result  (linearly interpolated) to full resolution for comparison for all methods.

\subsection{Evaluation of Inverse Search}
\label{ssc:eval_inv_search}
\begin{figure}
        \centering
        \begin{subfigure}[b]{0.5\textwidth}
        \centering
                \includegraphics[width=0.60\textwidth]{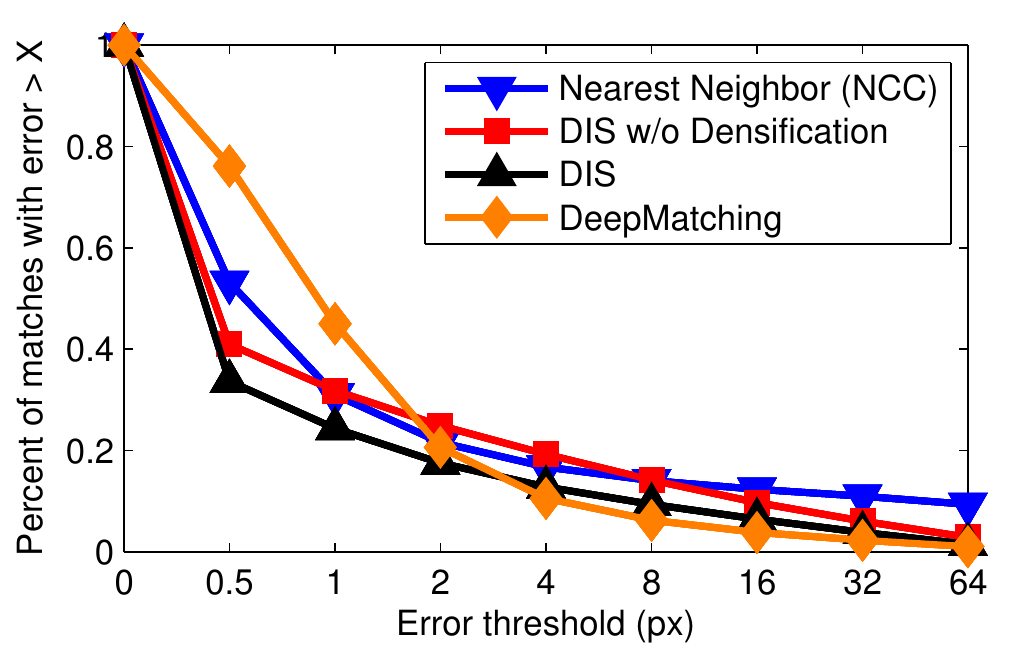}
        \end{subfigure}%
        \caption{Percent of sparse correspondences above error threshold.}\label{fig:diseval}
\end{figure}

{\tiny
\begin{center}
\begin{table}
\scriptsize
\centering
\begin{tabular}{l |c|c|c|c}
 &  EPE all & s0-10	& s10-40 & s40+\\
 \hline
NN &    32.06 &  13.64 &  53.77 & 101.00\\
DIS w/o Densification & 7.76  &  2.16  &  8.65  & 37.94\\
DIS & 4.16  &  {\bf 0.84} &    4.98 &  23.09 \\
DeepMatching \cite{Weinzaepfel-ICCV-2013} & {\bf 3.60} &   1.27 &    {\bf 3.91}  & {\bf 16.49}
\end{tabular}
\caption{Error of sparse correspondences (pixels). Columns left to right: i) average end-point error over complete flow field, ii) error in displacement range $<10$ px., iii) $10-40$ px., iv) $>40$ px.}
\label{tab:diseval}
\end{table}
\end{center}
}
In this section we evaluate the sparse point correspondences created by inverse search on the Sintel training dataset.
For each frame pair we initialized a sparse grid (given by Deep Matching~\cite{Weinzaepfel-ICCV-2013}) in the first image and computed point correspondences in the second image. The correspondences are computed by 
i) exhaustive \emph{Nearest Neighbor} search on normalized cross-correlation (\emph{NCC}),
ii) our method where we skip the densification step between each scale change (\emph{DIS w/o Densification}), iii) our method including the densification step (\emph{DIS}), and using
iv) \emph{DeepMatching}~\cite{Weinzaepfel-ICCV-2013}. 
The results are shown in Fig.~\ref{fig:diseval} and Table~\ref{tab:diseval}.

We have four observations: 
i) Nearest Neighbor search has a low number of incorrect matches and precise correspondences, but is very prone to outliers. 
ii) DeepMatching has a high percentage of erroneous correspondences (with small errors), but are  very good at large displacements.
iii) In contrast to this, our method (DIS w/o Densification) generally produces fewer correspondences with small errors, but is strongly affected by outliers. This is due to the fact that the implicit SSD (sum of squared differences) error minimization is not invariant to changes in orientation, contrast, and deformations. 
iv) Averaging all patches in each scale (DIS), taking into account their photometric error as described in eq.~\eqref{eq:avg}, introduces  robustness towards these outliers. It also decreases the error for approximately correct matches. Furthermore, it enables reducing the number of patches at coarser scales, leading to lower run-time.

\subsection{MPI Sintel optical flow results}
\label{ssc:mpi_sintel}

\begin{figure*}
\centering
\resizebox{\linewidth}{!}
{
\begin{tabular}{cccc}
\includegraphics[width=0.4\textwidth]{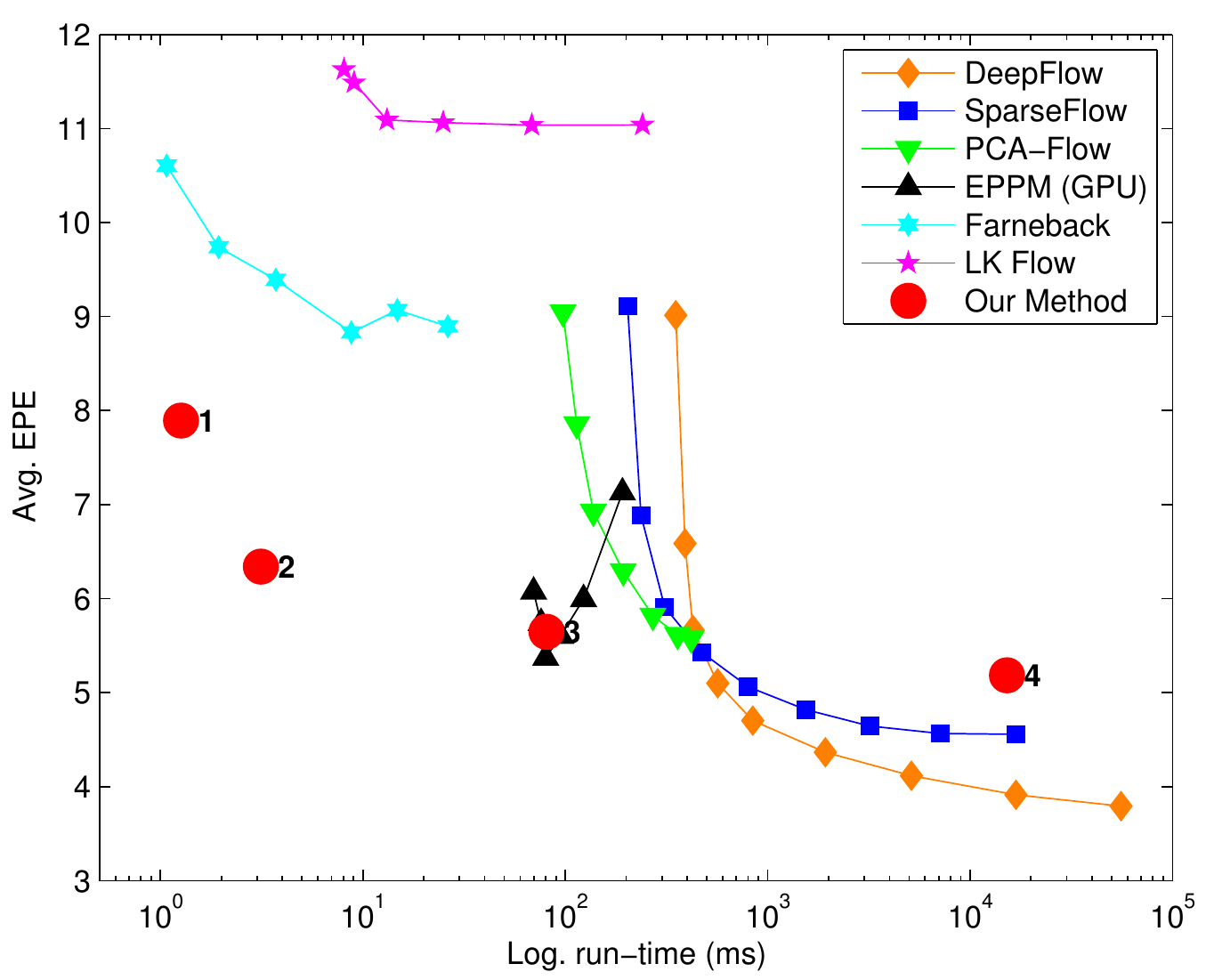}&
\includegraphics[width=0.4\textwidth]{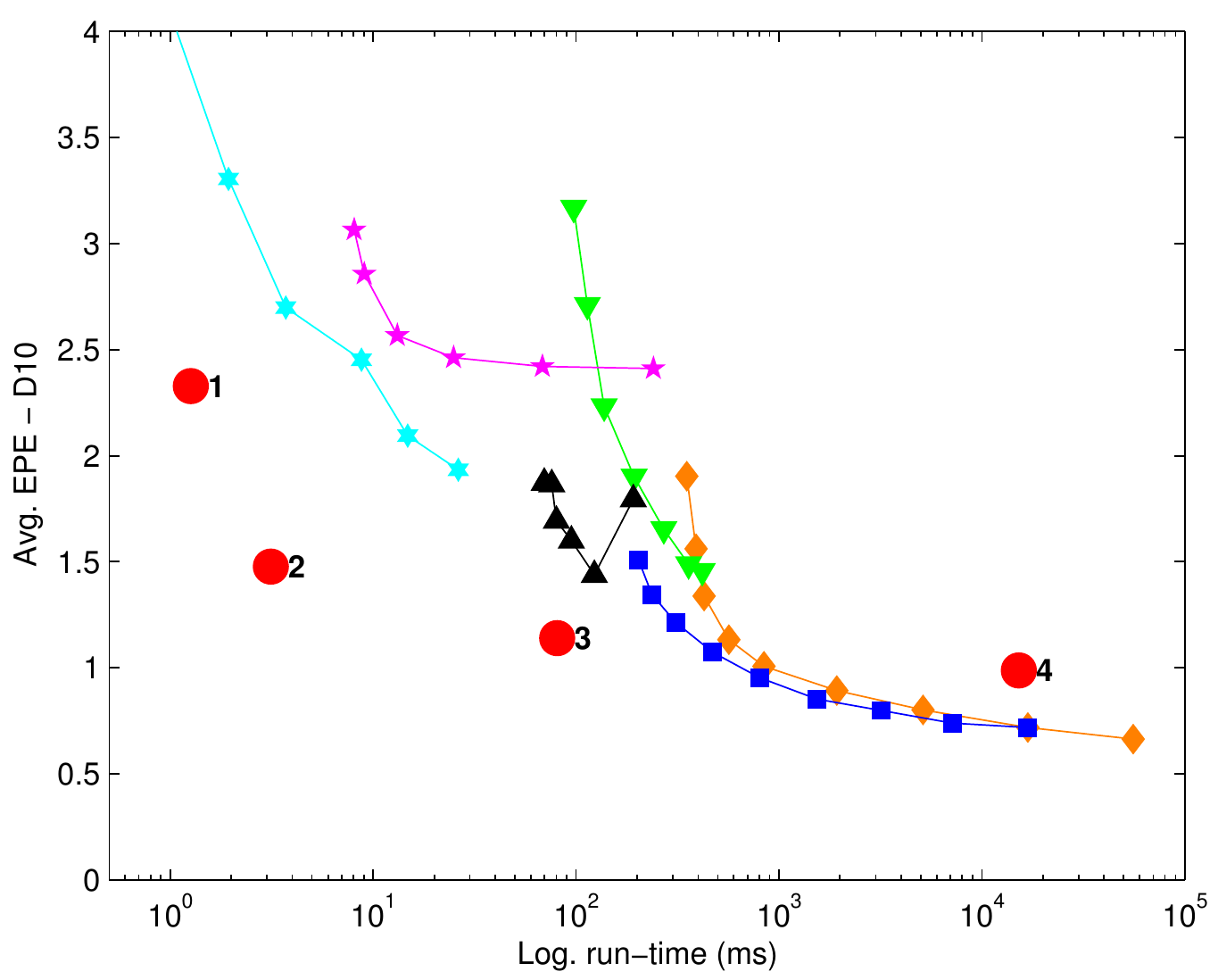}&
\includegraphics[width=0.4\textwidth]{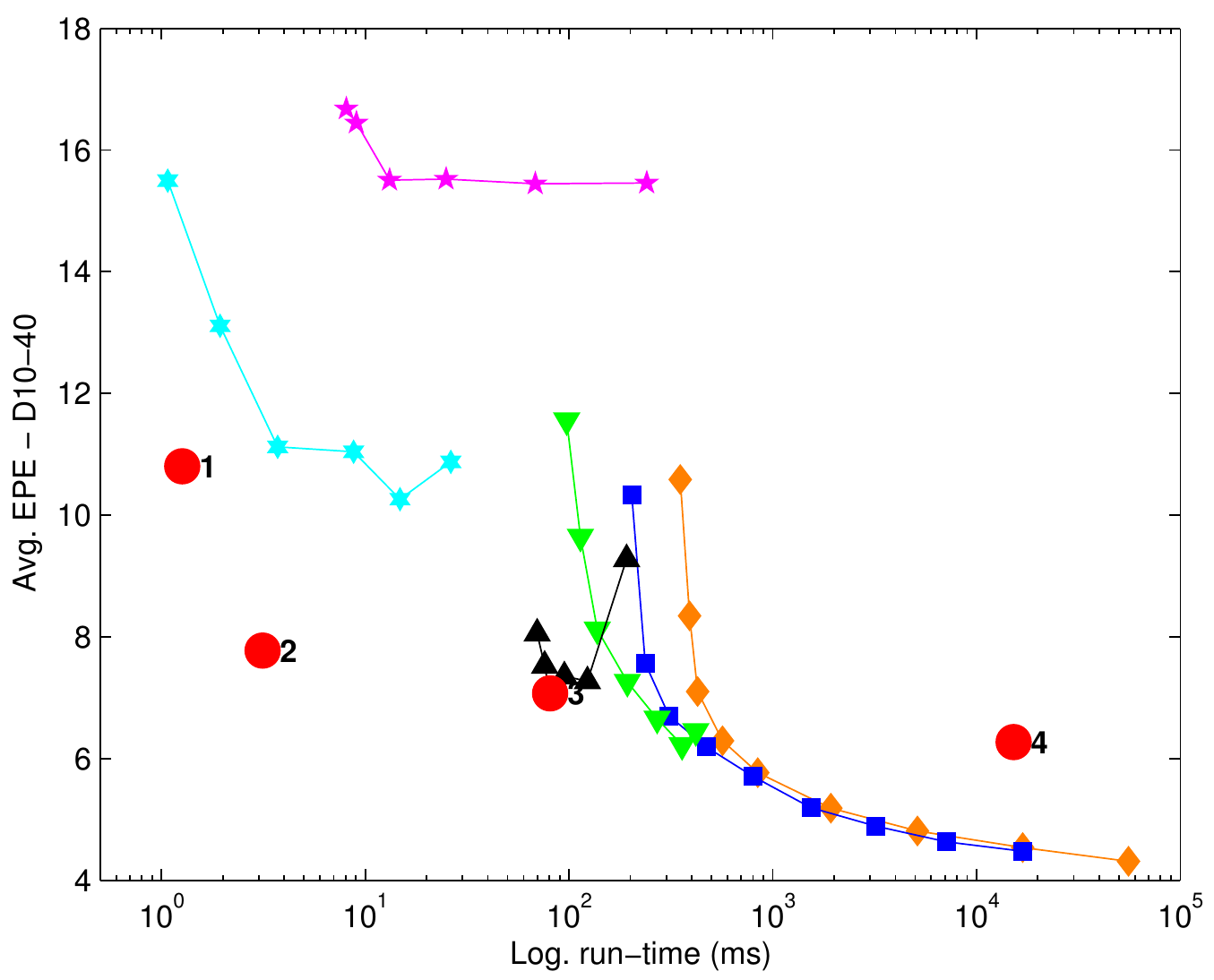}&
\includegraphics[width=0.4\textwidth]{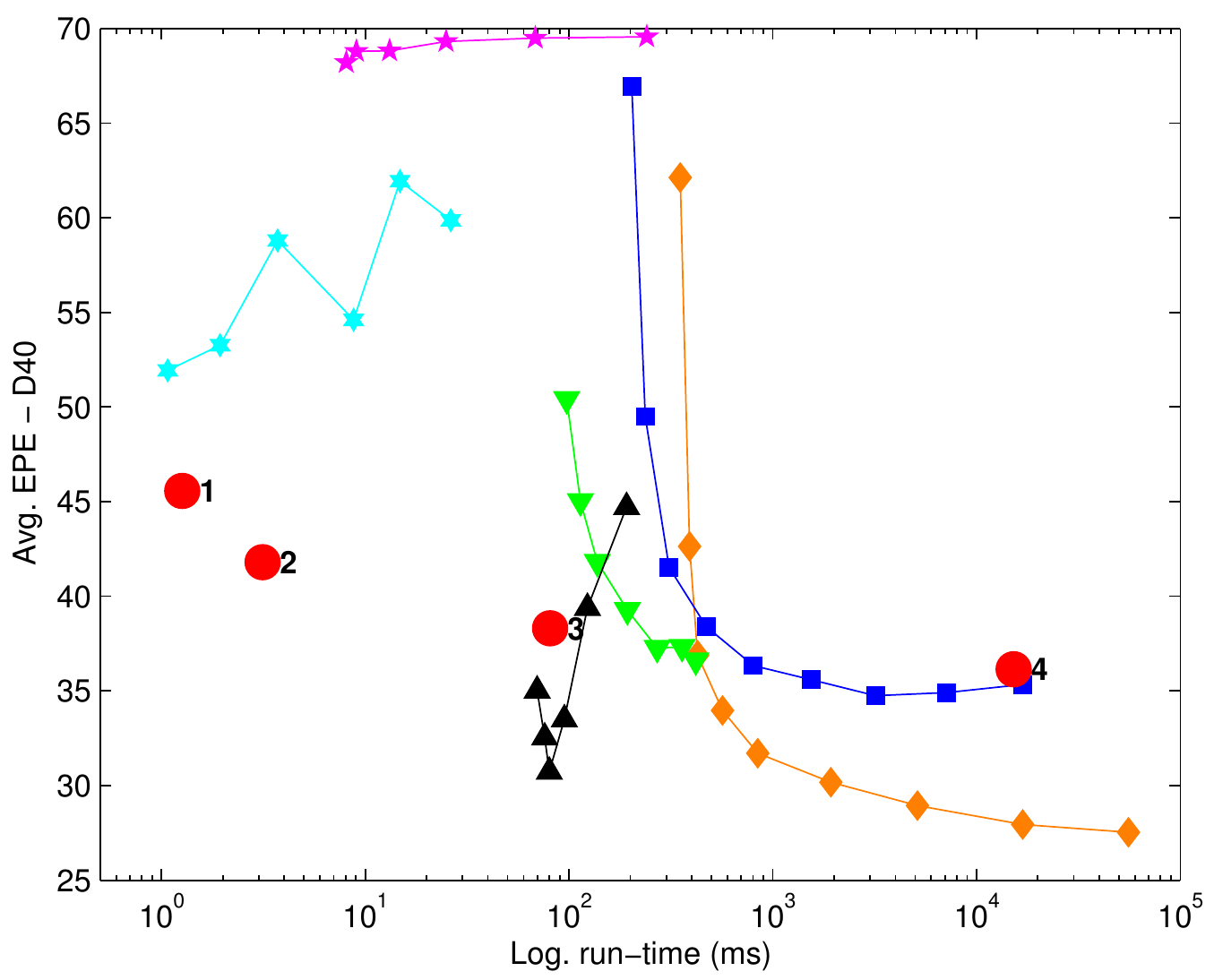}\\
 
                EPE, full range.&
                EPE for $<10$ px.&
                EPE for $10-40$ px.&
                EPE for $>40$ px.\\
\end{tabular}
}
\vspace{-0.2cm}
        \caption{Sintel (training) flow result. Average end-point error (pixel) versus run-time (millisecond) on various displacement ranges.}
        \label{fig:resultsintel}
\vspace{-0.5cm}
\end{figure*}

{\tiny
\begin{table}
\scriptsize
\centering
\setlength{\tabcolsep}{3pt}
\resizebox{\columnwidth}{!}{
\begin{tabular}{l|cccc|lcr}
 &  EPE all & s0-10	& s10-40 & s40+ & time (s) & CPU & GPU\\
\hline
FlowFields \cite{Bailer-CoRR-2015}	& {\bf 5.81} & 1.16 & {\bf 3.74} & {\bf 33.89} &  \textcolor{white}{00}18\textcolor{white}{000} \textsuperscript{\textdagger} & \checkmark &  \\
DeepFlow \cite{Weinzaepfel-ICCV-2013}        & 7.21	&   1.28 &	4.11 &	44.12 & \textcolor{white}{00}55 & \checkmark &  \\
SparseFlow \cite{Timofte-WACV-2015}      & 7.85	&   {\bf 1.07} &	3.77 &	51.35 & \textcolor{white}{00}16 & \checkmark &\\
EPPM \cite{Bao-TIP-2014}	        & 8.38	&   1.83 &	4.96 &	49.08 & \textcolor{white}{000}0.31& &\checkmark \\
PCA-Flow \cite{Wulff-CVPR-2015}        & 8.65	&   1.96 &	4.52 &	51.84 & \textcolor{white}{000}0.37 & \checkmark &\\
LDOF   \cite{Brox-PAMI-2011}          & 9.12	&   1.49 & 4.84 &	57.30 & \textcolor{white}{00}60\textcolor{white}{000} \textsuperscript{\textdagger}  \textsuperscript{\textdaggerdbl}& \checkmark &\\
Classic+NL-fast \cite{Sun-CVPR-2010} & 10.09 &  1.09 &	4.67 &	67.81 & \textcolor{white}{0}120\textcolor{white}{000} \textsuperscript{\textdagger} \textsuperscript{\textdaggerdbl} & \checkmark &\\
{\bf DIS-Fast}     & 10.13 &	2.17 &	5.93 &	59.70 & \textcolor{white}{000}{\bf 0.023}& \checkmark & \\
SimpleFlow \cite{tao2012simpleflow}      & 13.36 &	1.48 &	9.58 &	81.35 & \textcolor{white}{000}1.6\textcolor{white}{00} \textsuperscript{\textdagger} \textsuperscript{\textdaggerdbl} & &\checkmark \\
\end{tabular}
}
\caption{Sintel test errors in pixels (\url{http://sintel.is.tue.mpg.de/results}), retrieved on 31st Oct. 2015 for \emph{final} subset. Run-times are measured by us, except: \textsuperscript{\textdagger}self-reported, and \textsuperscript{\textdaggerdbl}on other datasets with same or smaller resolution.}
\label{tab:sintel_online}
\end{table}
}

Based on our observations on parameter selection in \S~\ref{ssc:implementation}, we selected four operating points, corresponding to
\begin{enumerate}
\itemsep0em
\item[{\bf (1)}] $\theta_{sf}=3,\theta_{it}=016,\theta_{ps}=08,\theta_{ov}=0.30$, {\bf at 600~Hz},
\item[{\bf (2)}] $\theta_{sf}=3,\theta_{it}=012,\theta_{ps}=08,\theta_{ov}=0.40$, {\bf at 300~Hz},
\item[{\bf (3)}] $\theta_{sf}=1,\theta_{it}=016,\theta_{ps}=12,\theta_{ov}=0.75$, {\bf at 10~Hz},
\item[{\bf (4)}] $\theta_{sf}=0,\theta_{it}=256,\theta_{ps}=12,\theta_{ov}=0.75$, {\bf at 0.5~Hz}.
\end{enumerate}
All operating points except {\bf (1)} use variational refinement.
We compare our method against a set of recently published baselines running on a single CPU core: DeepFlow~\cite{Weinzaepfel-ICCV-2013}, SparseFlow~\cite{Timofte-WACV-2015}, PCA-Flow~\cite{Wulff-CVPR-2015}; two older established methods: Pyramidal Lukas-Kanade Flow~\cite{bouguet2001pyramidal,Lucas-IJCAI-1981}, Farneback's method~\cite{Farneback-IA-2003}; and one recent GPU-based method: EPPM~\cite{Bao-TIP-2014}. 
Since run-times for optical flow methods are strongly linked to image resolution, we incrementally speed-up all baselines by downscaling the input images by factor of $2^n$, where $n$ starting at $n = 0$ is increased in increments of $0.5$.
We chose this \emph{non-intrusive} parameter of image resolution to analyse each method's trade-off between run-time and flow error.
We bilinearly interpolate the resulting flow field to the original resolution for evaluation.

We run all baselines and our method for all operating points on the Sintel~\cite{Butler-ECCV-2012} \emph{final} training (Fig.~\ref{fig:resultsintel}) and testing (Table~\ref{tab:sintel_online}) benchmark. 
As noted in \S~\ref{ssc:implementation}, run-times for all methods are reported without preprocessing for the training dataset to facilitate comparison of flow algorithms running in the same environment at high speed, and with preprocessing for the online testing benchmark to allow comparison with self-reported times.
From the experiments on the testing and training dataset, we draw several conclusions:
\emph{Operating point {\bf (2)}} points to the best trade-off between run-time and flow error.  
For the average end-point error of around 6 pixels, our method is approximately two orders of magnitude faster than the fastest CPU baseline (PCA-Flow~\cite{Wulff-CVPR-2015}) a still more than one order of magnitude faster than the fastest GPU baseline (EPPM~\cite{Bao-TIP-2014}). 
Our method can be further sped-up by removing the variational refinement at \emph{operating point {\bf (1)}} while maintaining reasonable flow quality (see Fig.~\ref{fig:res_sintel_qualitative}). 
\emph{Operating point {\bf (3)}} is comparable with the performance of EPPM, but slightly better for small displacements and worse for large displacements.
If we use all available scales, and increase the number of iterations, we obtain \emph{operating point {\bf (4)}}. 
At the run-time of several seconds per frame pair, more complex methods, such as DeepFlow, perform better, in particular for large displacements.

In the supplementary material we show variants of Fig.~\ref{fig:resultsintel}, and \ref{fig:resultkitti} where all preprocessing times are included. Furthermore, we provide flow error maps on Sintel, where typical failure cases of our method at motion discontinuities, large displacements, and frame boundaries are observable.

\subsection{KITTI optical flow results}
\label{ssc:kittiflow}

\begin{figure}
        \centering
        \resizebox{\linewidth}{!}
        {
\begin{tabular}{cc}
\includegraphics[width=0.4\textwidth]{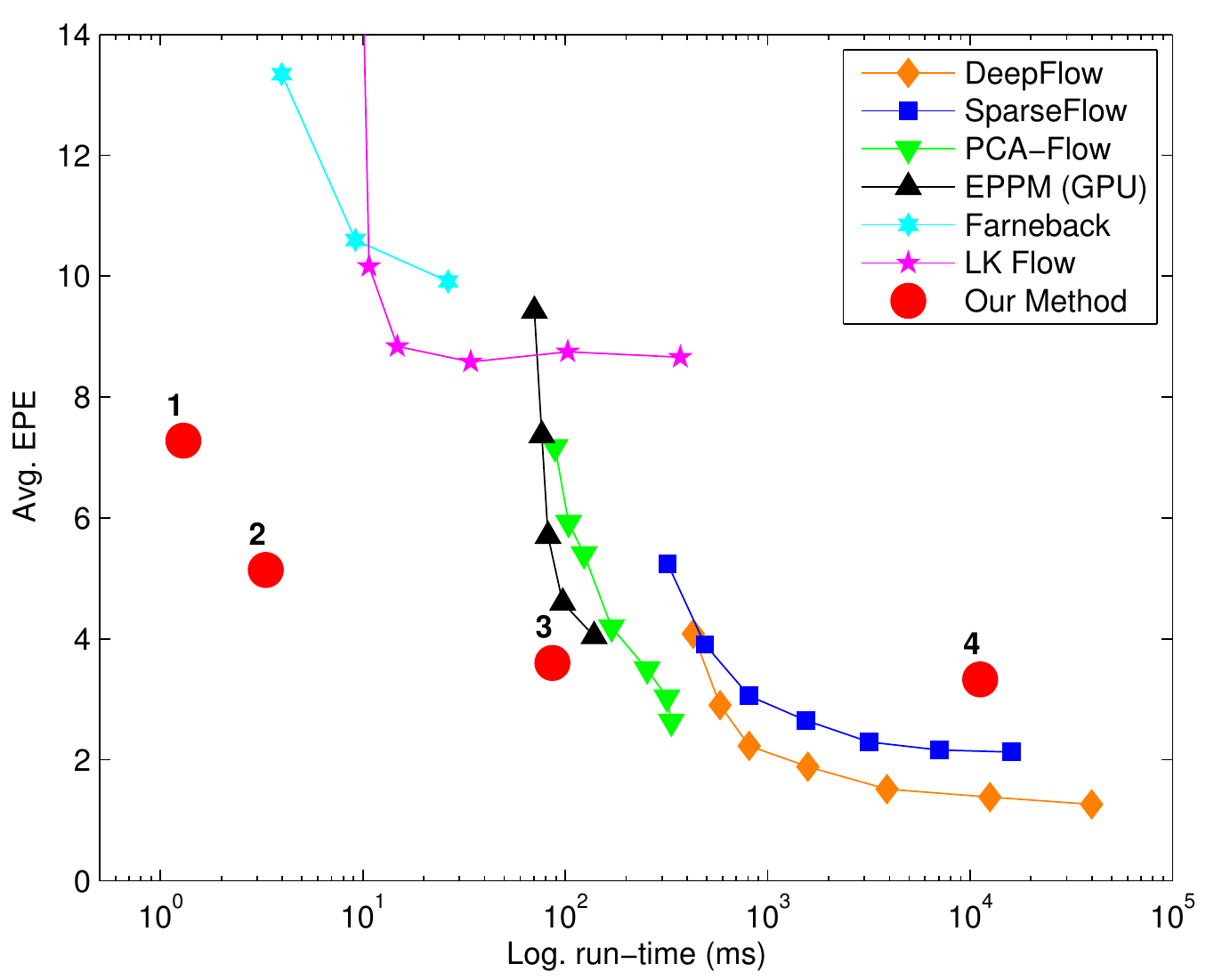}&  
\includegraphics[width=0.4\textwidth]{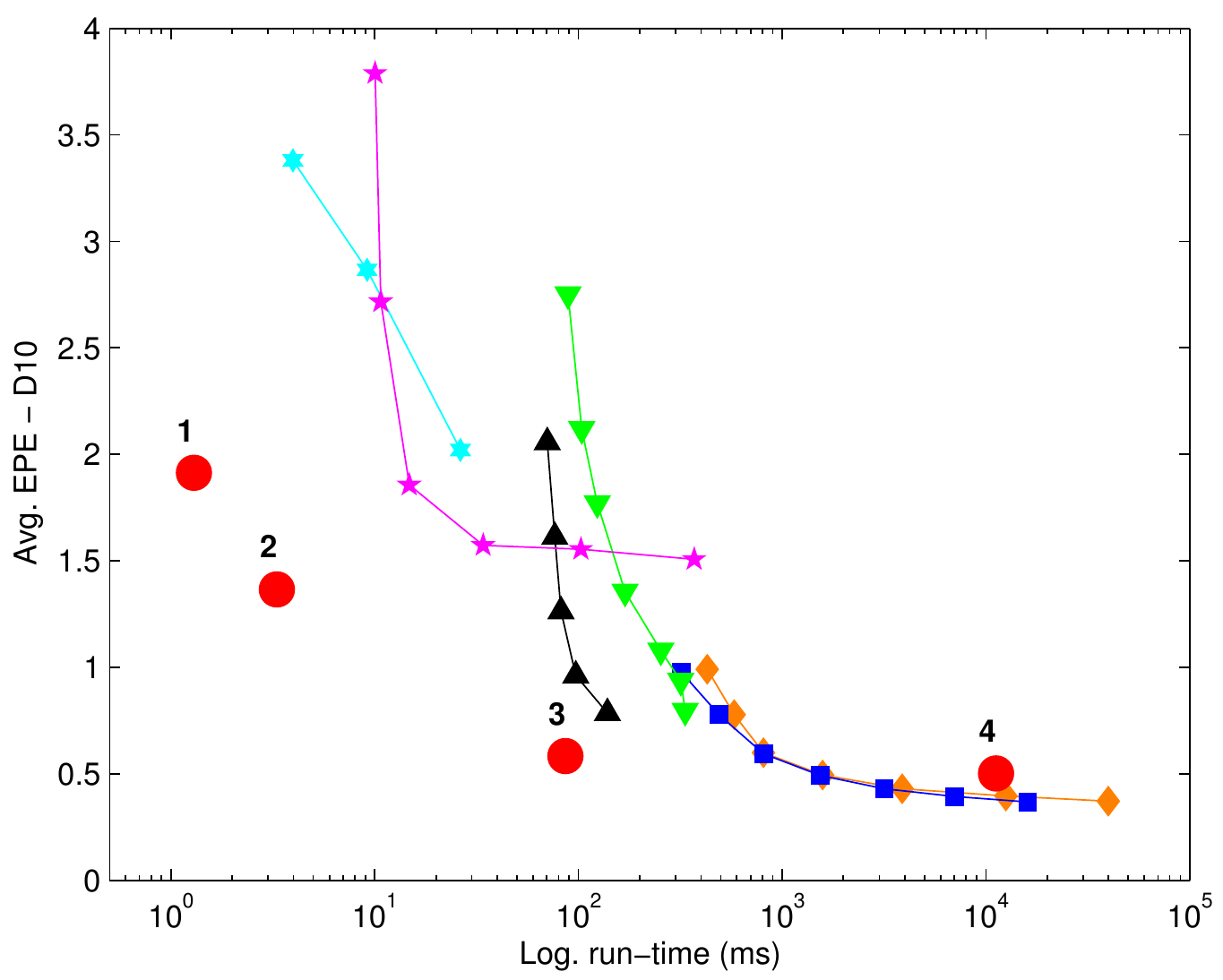}\\
EPE, full range. & EPE for $<10$ px.\\
\end{tabular}
}
\vspace{-0.2cm}
        \caption{KITTI (training) result. Average end-point error (px) versus run-time (ms) for average (left) and small displacements (right). See supplementary material for large displacement errors.}
        \label{fig:resultkitti}
\vspace{-0.2cm}
\end{figure} 

{\tiny
\begin{table}
\scriptsize
\centering
\setlength{\tabcolsep}{3pt}
\resizebox{\columnwidth}{!}{
\begin{tabular}{l |cccc|lcr}
& Out-Noc & Out-All	& Avg-Noc & Avg-All & time (s) & CPU& GPU\\
 \hline
PH-Flow \cite{yang2015dense}     & {\bf 5.76}  \% &	{\bf 10.57 \%} &	{\bf 1.3 px} &	{\bf 2.9 px}	& \textcolor{white}{0}800 & \checkmark &  \\
DeepFlow \cite{Weinzaepfel-ICCV-2013}    & 7.22  \% &	17.79 \% &	1.5 px &	5.8 px	& \textcolor{white}{00}17 & \checkmark &  \\
SparseFlow \cite{Timofte-WACV-2015}  & 9.09  \% &	19.32 \% &	2.6 px &	7.6 px	& \textcolor{white}{00}10 & \checkmark &  \\
EPPM \cite{Bao-TIP-2014}        & 12.75 \% &	23.55 \% &	2.5 px &	9.2 px	& \textcolor{white}{000}0.25 & & \checkmark \\
PCA-Flow \cite{Wulff-CVPR-2015}    & 15.67 \% &	24.59 \% &	2.7 px &	6.2 px	& \textcolor{white}{000}0.19  & \checkmark &  \\
eFolki \cite{Plyer-RTIP-2014}      & 19.31 \% &	28.79 \% &	5.2 px &	10.9 px	& \textcolor{white}{000}0.026 & & \checkmark \\
LDOF \cite{Brox-PAMI-2011}        & 21.93 \% &	31.39 \% &	5.6 px &	12.4 px	& \textcolor{white}{00}60 & \checkmark &  \\
FlowNetS+ft \cite{Fischer-ICCV-2015} & 37.05 \% &	44.49 \% &	5.0 px &	9.1 px	& \textcolor{white}{000}0.08 & & \checkmark \\
{\bf DIS-Fast} & 38.58 \% &	46.21 \% &	7.8 px &	14.4 px	& \textcolor{white}{000}{\bf 0.024} & \checkmark &  \\
RLOF \cite{senst2012robust}        & 38.60 \% &	46.13 \% &	8.7 px &	16.5 px	& \textcolor{white}{000}0.488 & & \checkmark
\end{tabular}
}
\caption{KITTI test results (\url{http://www.cvlibs.net/datasets/kitti/eval_flow.php}), retrieved on 31st Oct. 2015, for \emph{all pixels, at 3px threshold}.}
\label{tab:kitti_online}
\end{table}
\vskip-5pt
}

Complementary to the experiment on the synthetic Sintel dataset, we ran our method on the KITTI Optical Flow benchmark~\cite{Geiger-IJRR-2013} for realistic driving scenarios. The result is presented in fig.~\ref{fig:resultkitti}, \ref{fig:res_kitti_qualitative} (training) and Table~\ref{tab:kitti_online} (testing).
We use the same four operating points as in \S~\ref{ssc:mpi_sintel}.
Our conclusions from the Sintel dataset in \S~\ref{ssc:mpi_sintel} also apply for this dataset, suggesting a stable performance of our method, since we did not optimize any parameters for this dataset.
On the online test benchmark we are on par with RLOF~\cite{senst2012robust} and the recently published FlowNet~\cite{Fischer-ICCV-2015}. 
Even though both take advantage of a GPU, we are still faster by one order of magnitude at comparable performance.

We include a plot of more operating points on the training set of Sintel and KITTI in the supplementary material.

\begin{figure*} \centering\centering\setlength{\tabcolsep}{0.1pt}\renewcommand{\arraystretch}{0} 
\resizebox{\textwidth}{!}
{ 
\begin{tabular}{ccccc}
{\bf 600 Hz} & {\bf 300 Hz} & {\bf 10 Hz} & {\bf 0.5 Hz} & {\bf Ground truth}\\
\includegraphics[width=0.195\textwidth]{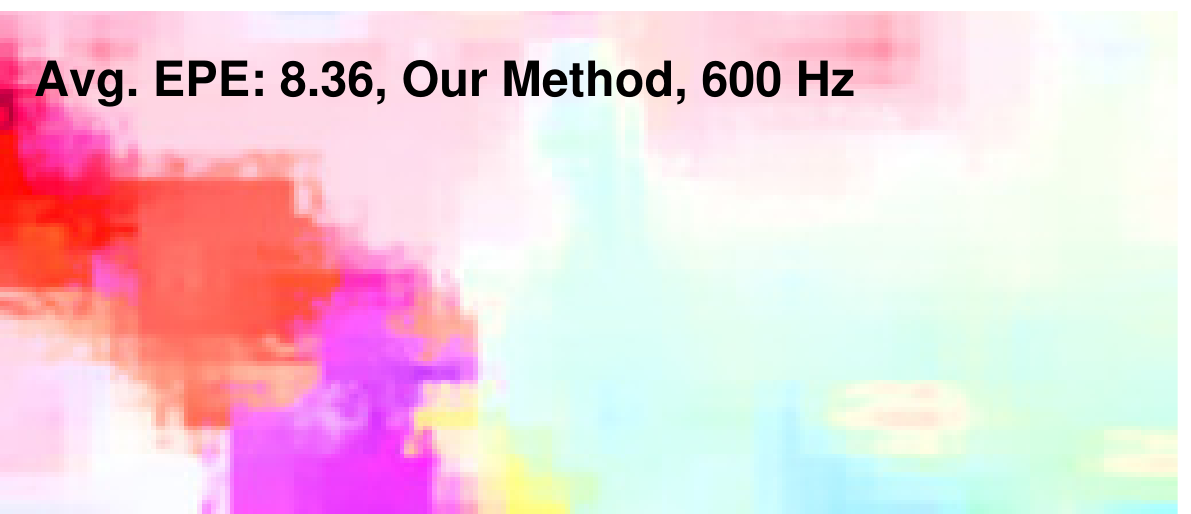}&
\includegraphics[width=0.195\textwidth]{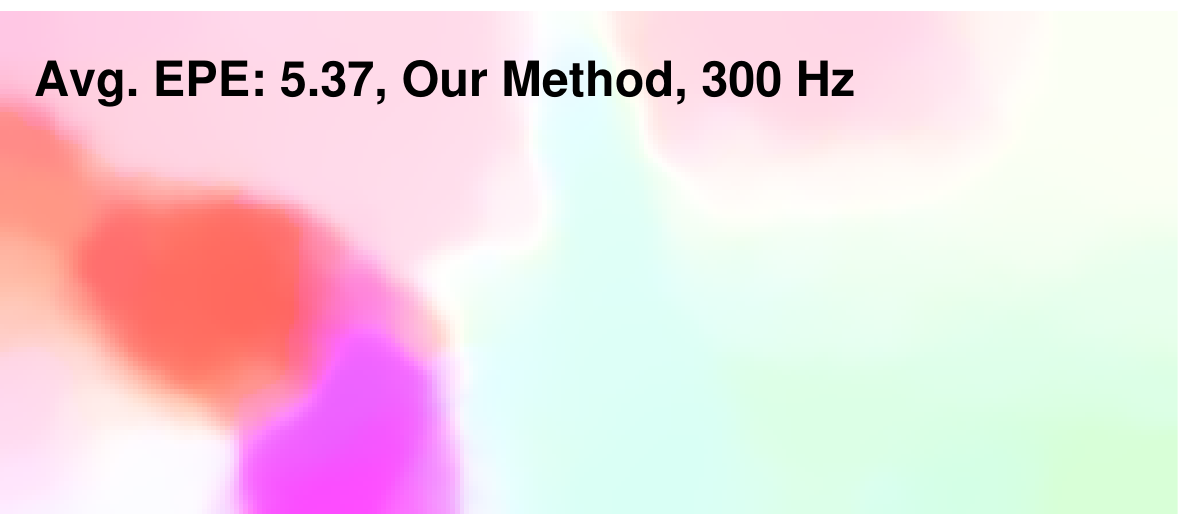}&
\includegraphics[width=0.195\textwidth]{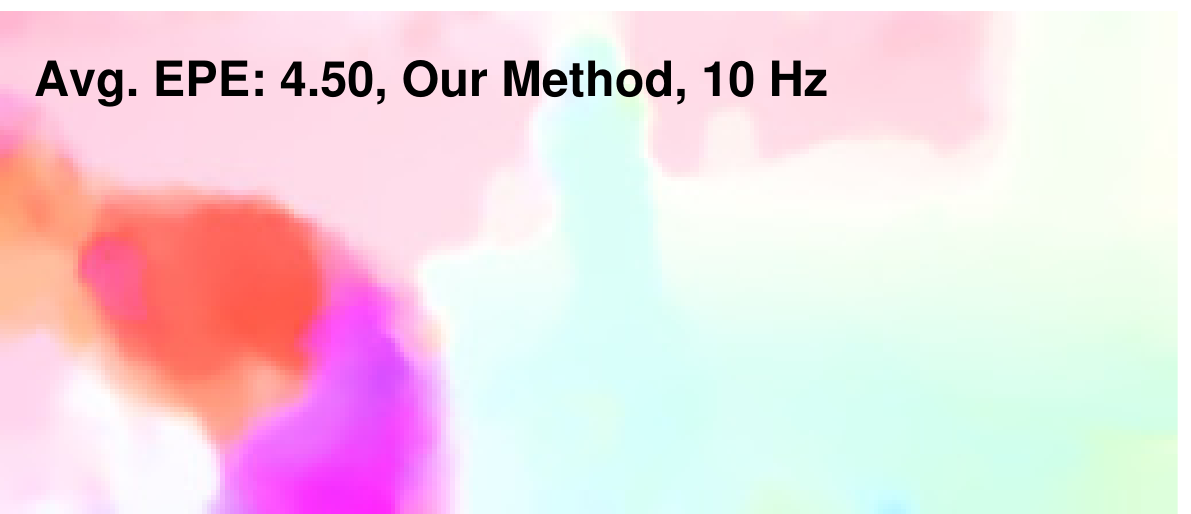}&
\includegraphics[width=0.195\textwidth]{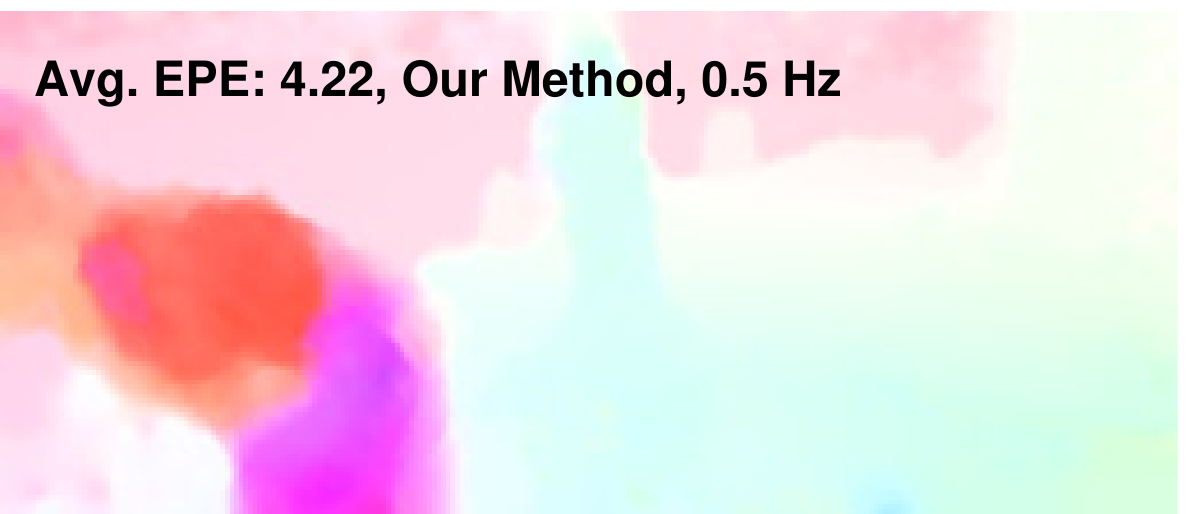}&
\includegraphics[width=0.195\textwidth]{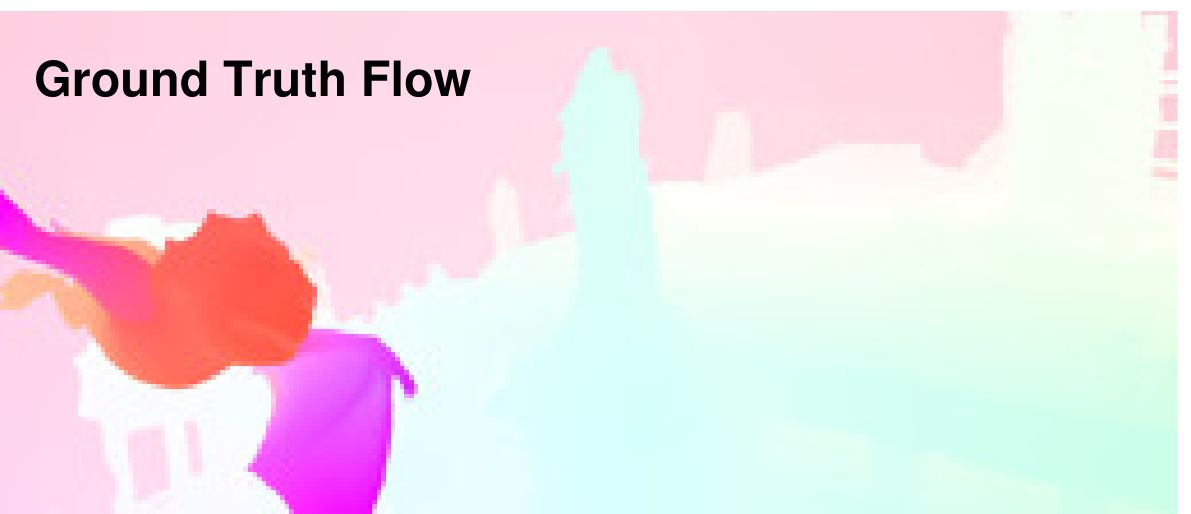}\\
\includegraphics[width=0.195\textwidth]{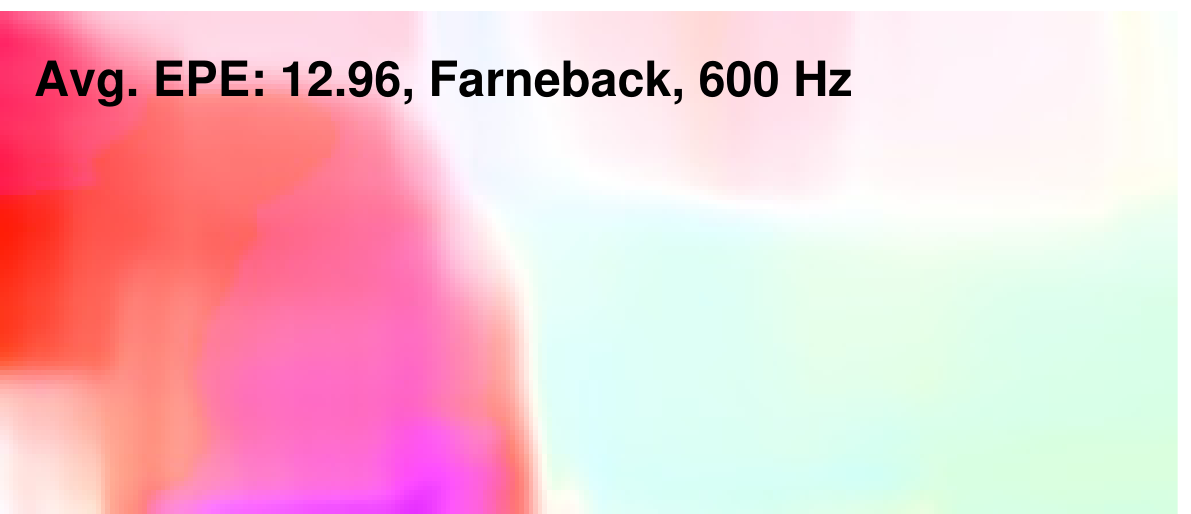}&
\includegraphics[width=0.195\textwidth]{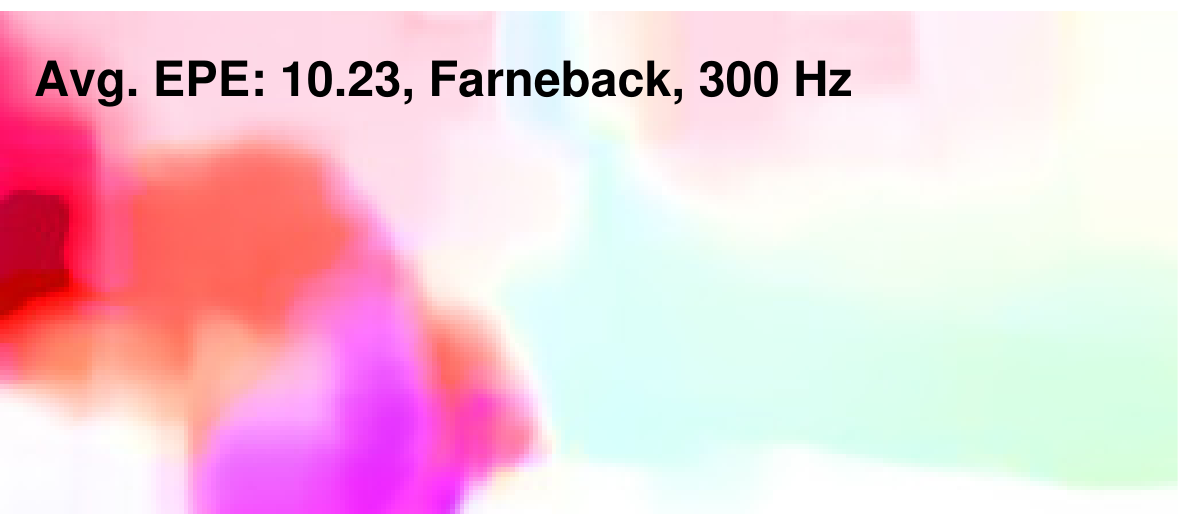}&
\includegraphics[width=0.195\textwidth]{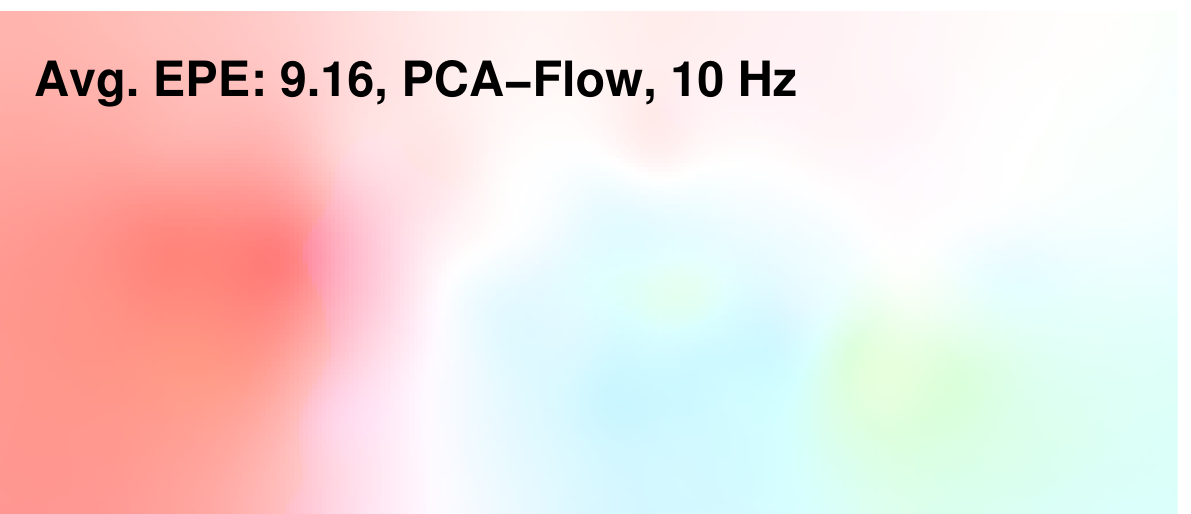}&
\includegraphics[width=0.195\textwidth]{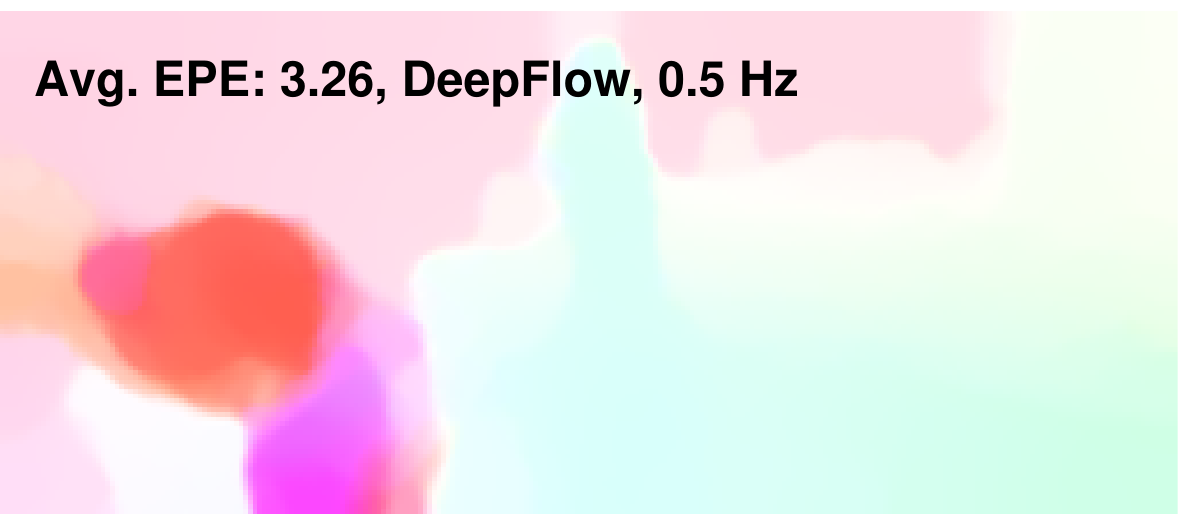}&
\includegraphics[width=0.195\textwidth]{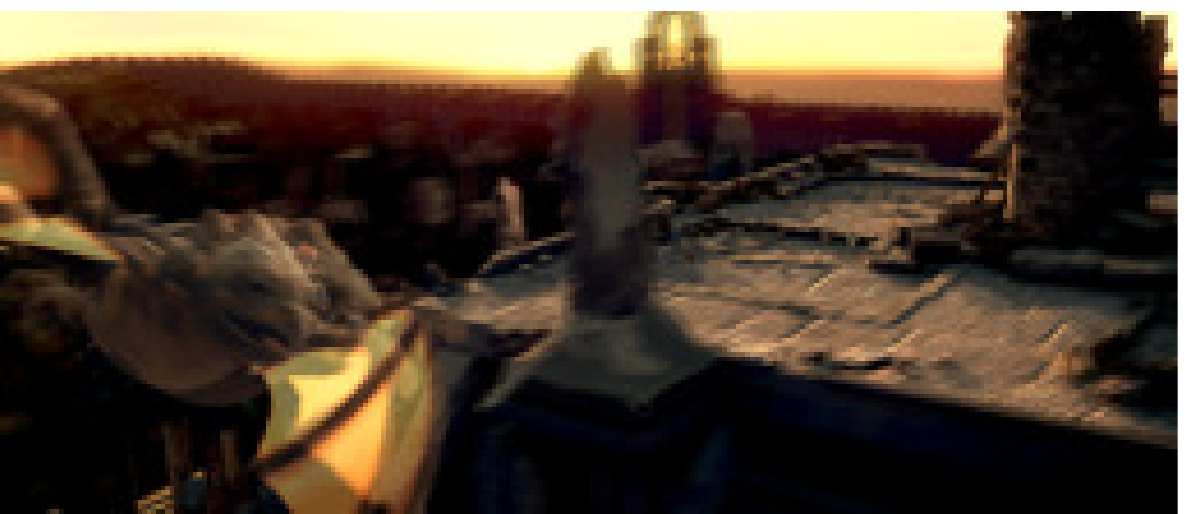}\\ [2pt]
\hline \\[2pt]
\includegraphics[width=0.195\textwidth]{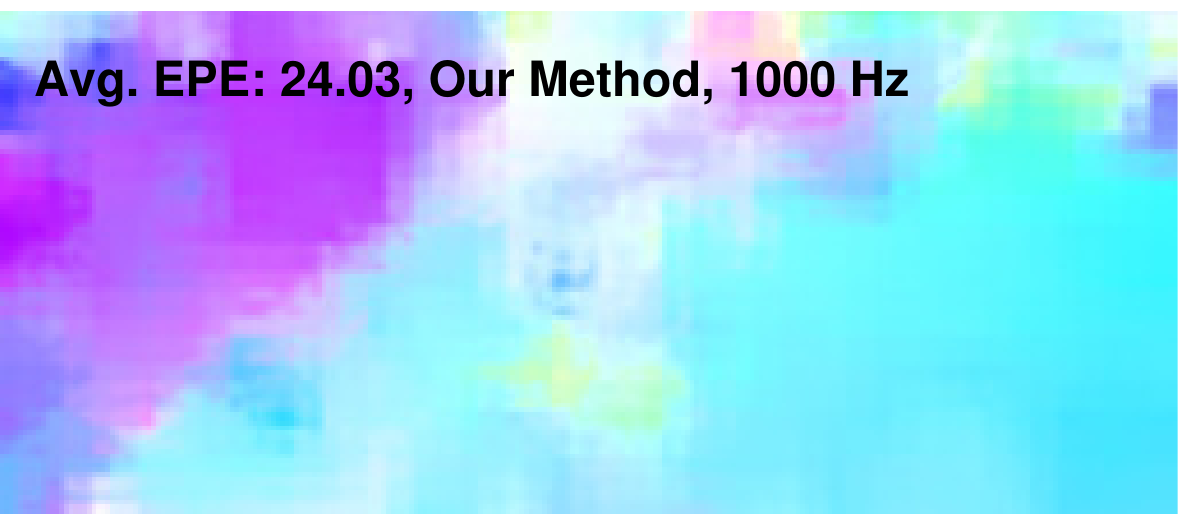}&
\includegraphics[width=0.195\textwidth]{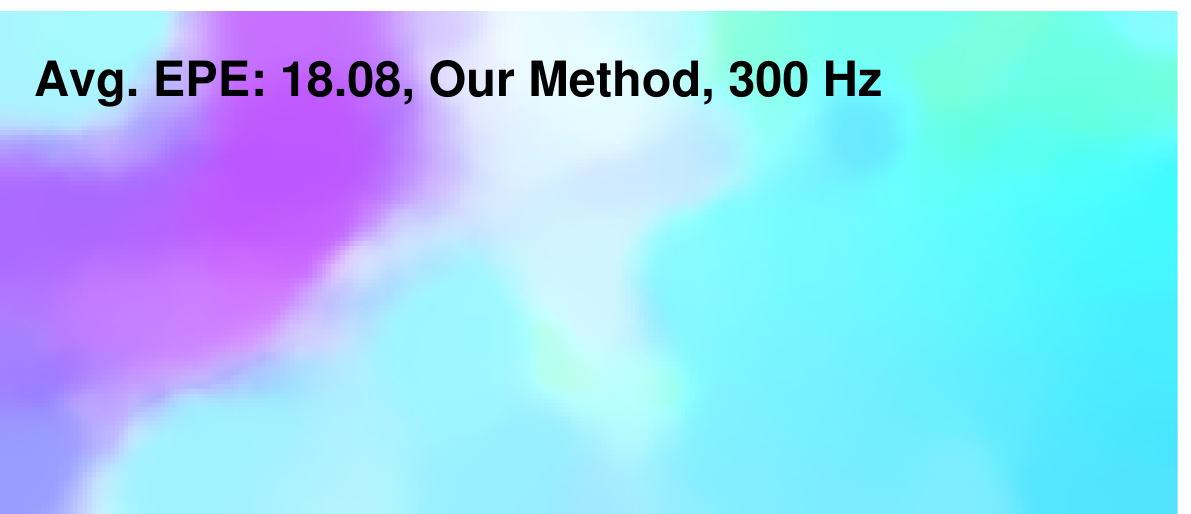}&
\includegraphics[width=0.195\textwidth]{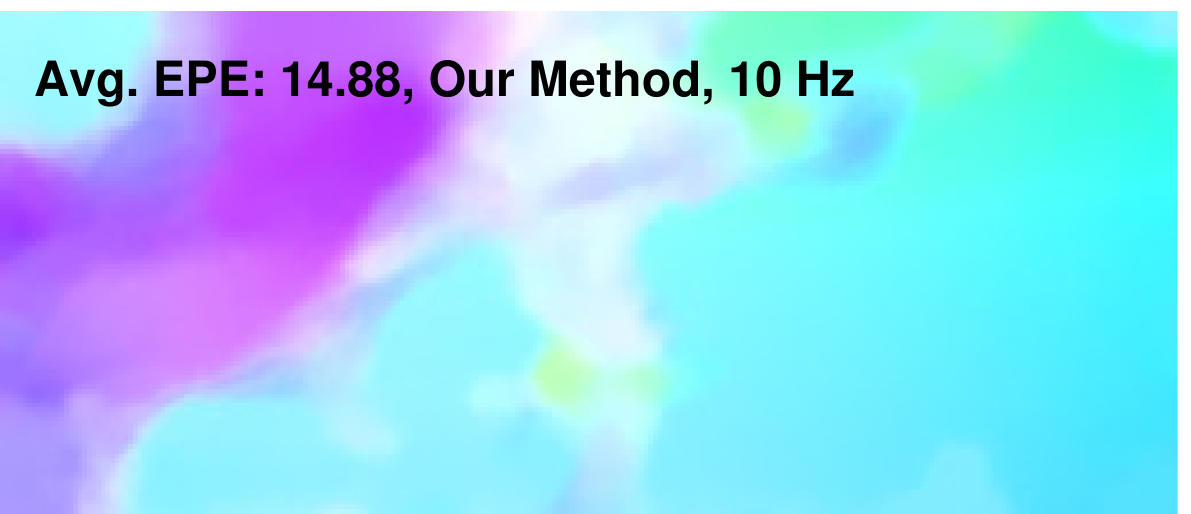}&
\includegraphics[width=0.195\textwidth]{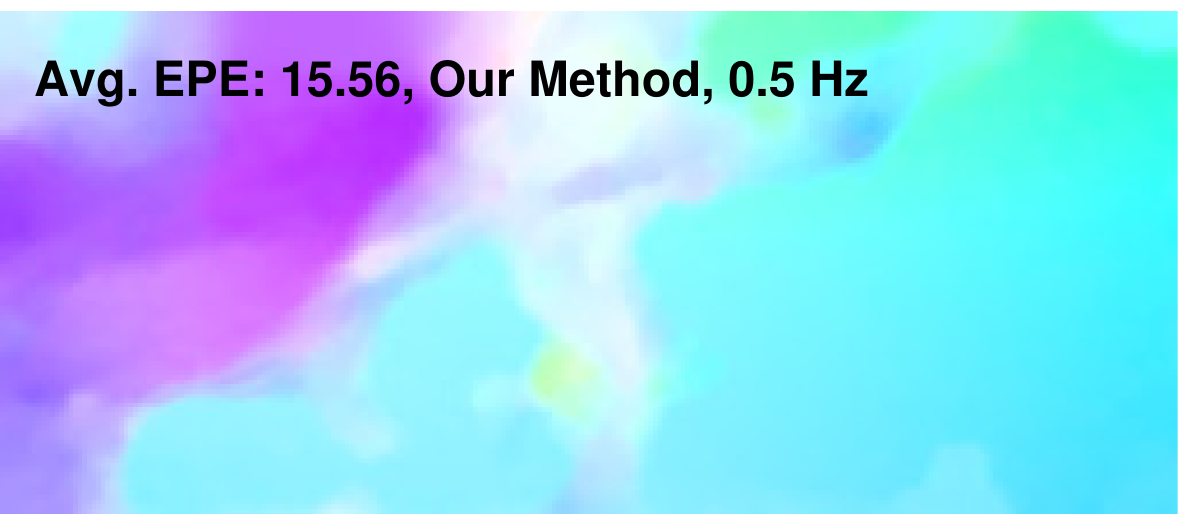}&
\includegraphics[width=0.195\textwidth]{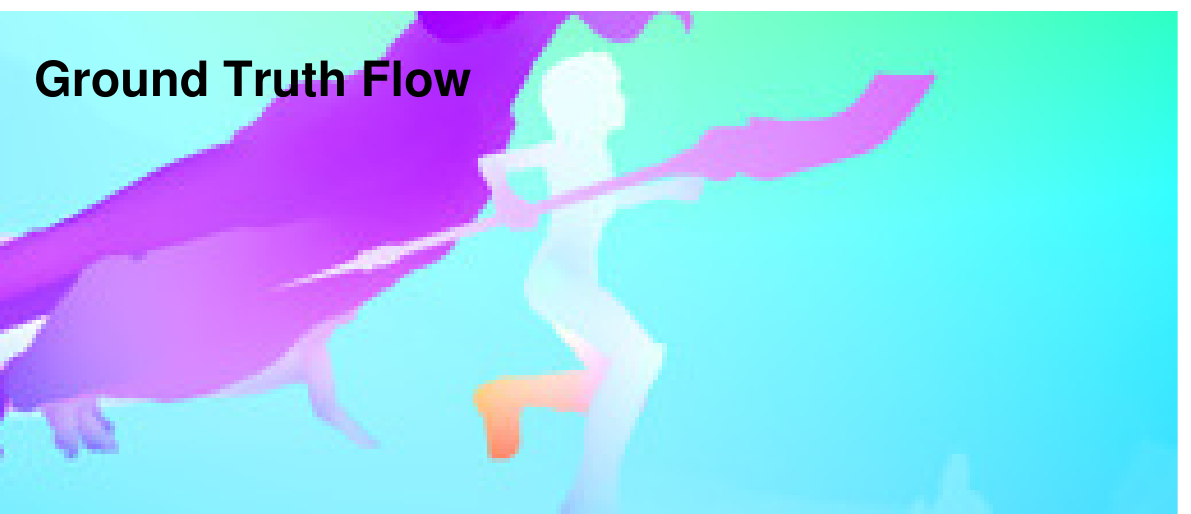}\\
\includegraphics[width=0.195\textwidth]{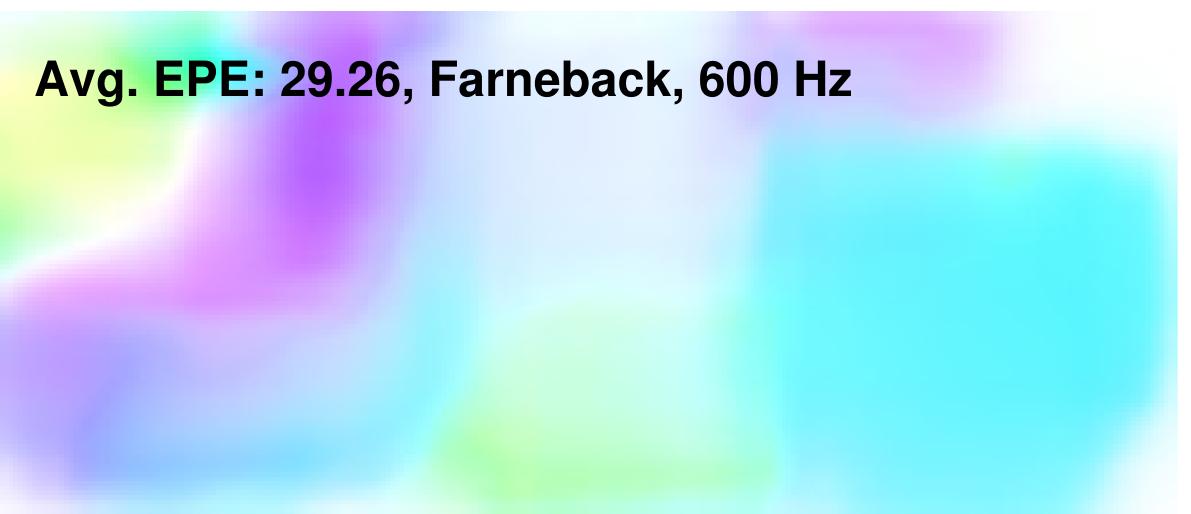}&
\includegraphics[width=0.195\textwidth]{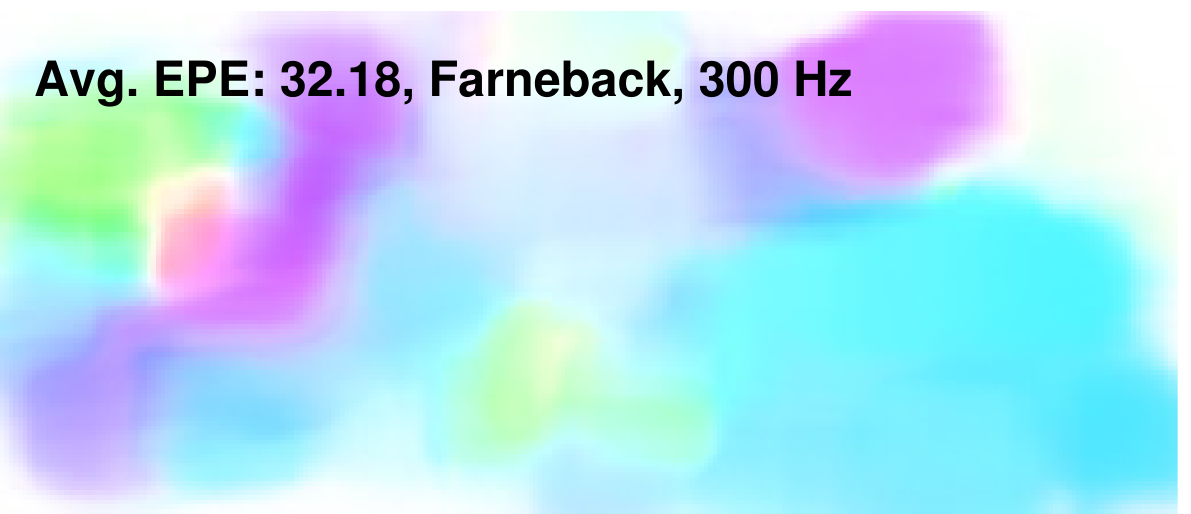}&
\includegraphics[width=0.195\textwidth]{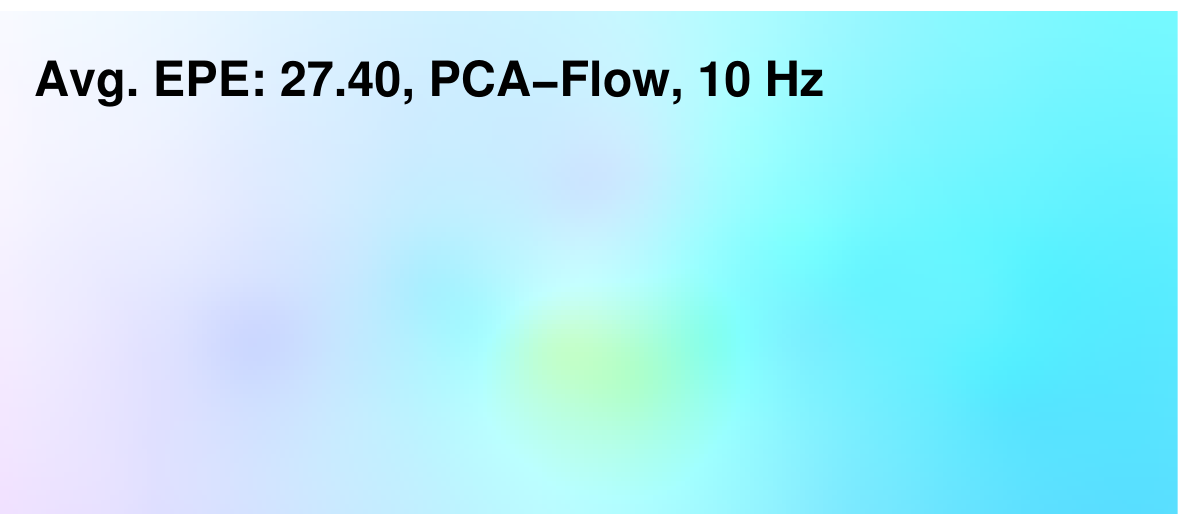}&
\includegraphics[width=0.195\textwidth]{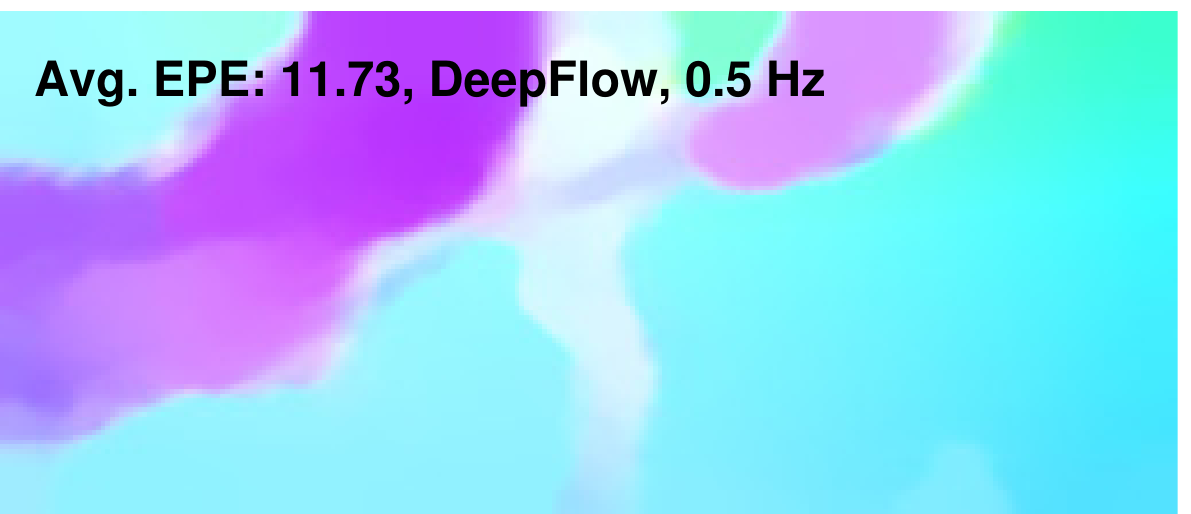}&
\includegraphics[width=0.195\textwidth]{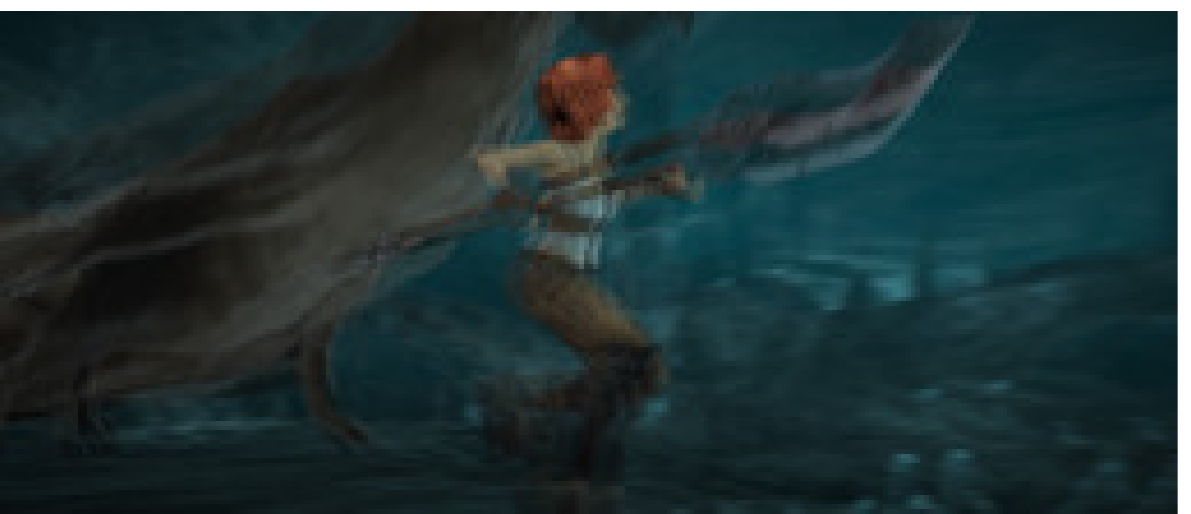}
\end{tabular}
}
 \vskip-11pt
 \caption{Examplary results on Sintel (training). In each block of $2 \times 6$ images.  Top row, left to right: Our method for operating points ({\bf 1})-({\bf 4}), Ground Truth. Bottom row: Farneback 600Hz, Farneback 300Hz, PCA-Flow 10Hz, DeepFlow 0.5Hz, Original Image.}\label{fig:res_sintel_qualitative} 
 \end{figure*} 
 
\begin{figure*} \centering\setlength{\tabcolsep}{0.1pt}\renewcommand{\arraystretch}{0} 
\resizebox{\textwidth}{!}
{ 
\begin{tabular}{ccccc}
{\bf 600 Hz} & {\bf 300 Hz} & {\bf 10 Hz} & {\bf 0.5 Hz} & {\bf Ground truth}\\
\includegraphics[width=0.195\textwidth]{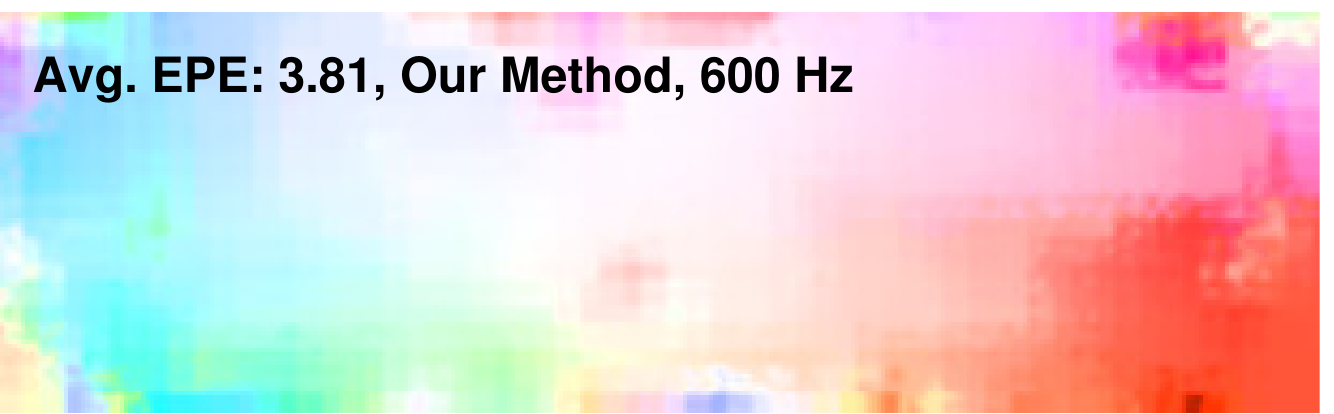}&
\includegraphics[width=0.195\textwidth]{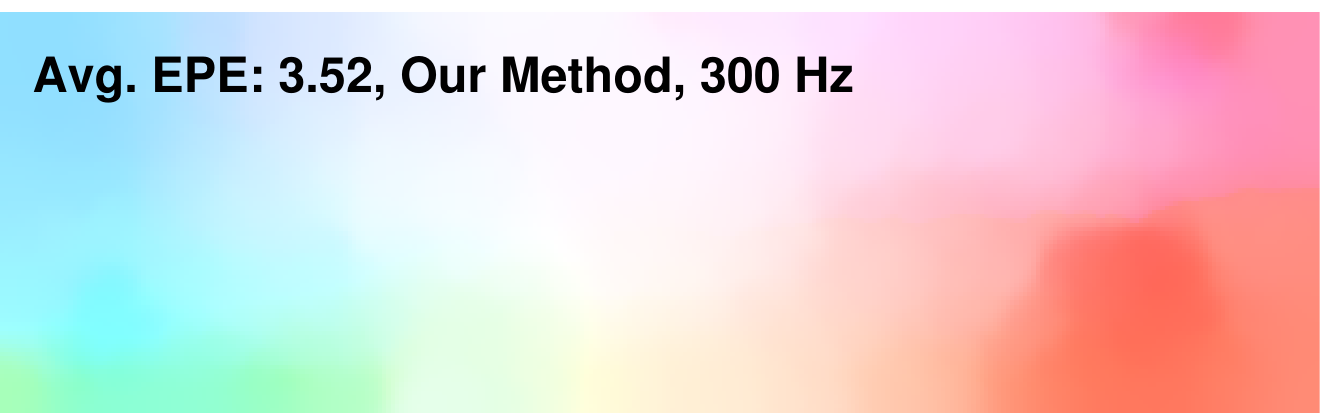}&
\includegraphics[width=0.195\textwidth]{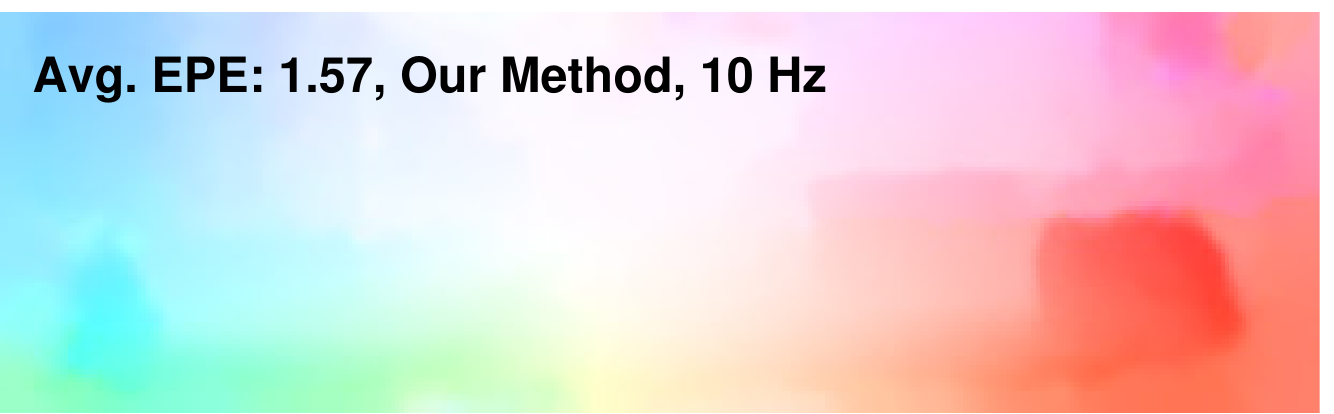}&
\includegraphics[width=0.195\textwidth]{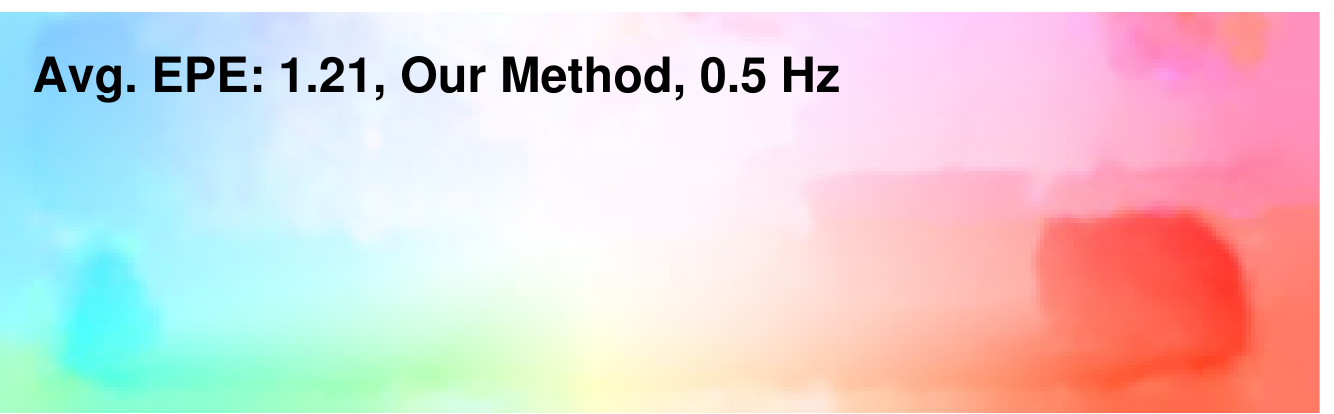}&
\includegraphics[width=0.195\textwidth]{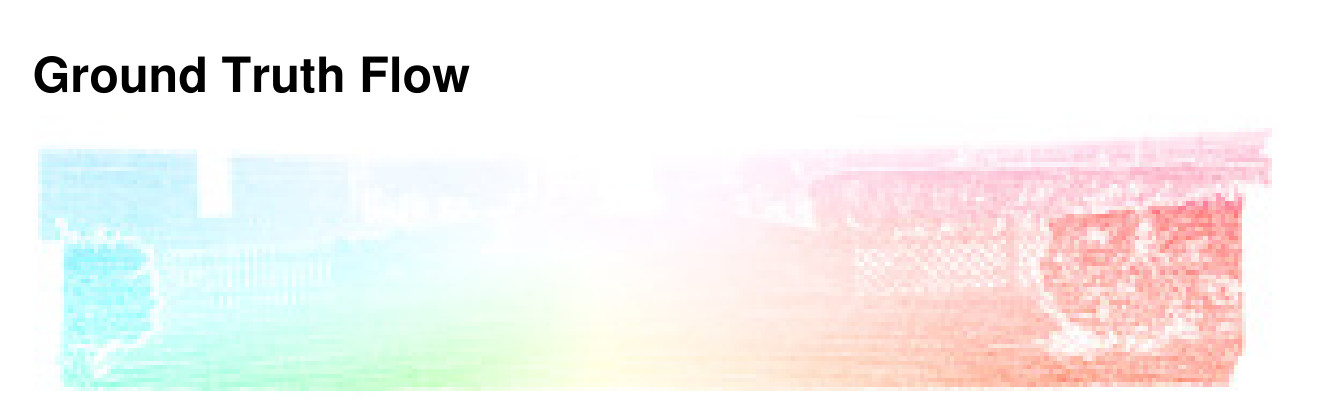}\\
\includegraphics[width=0.195\textwidth]{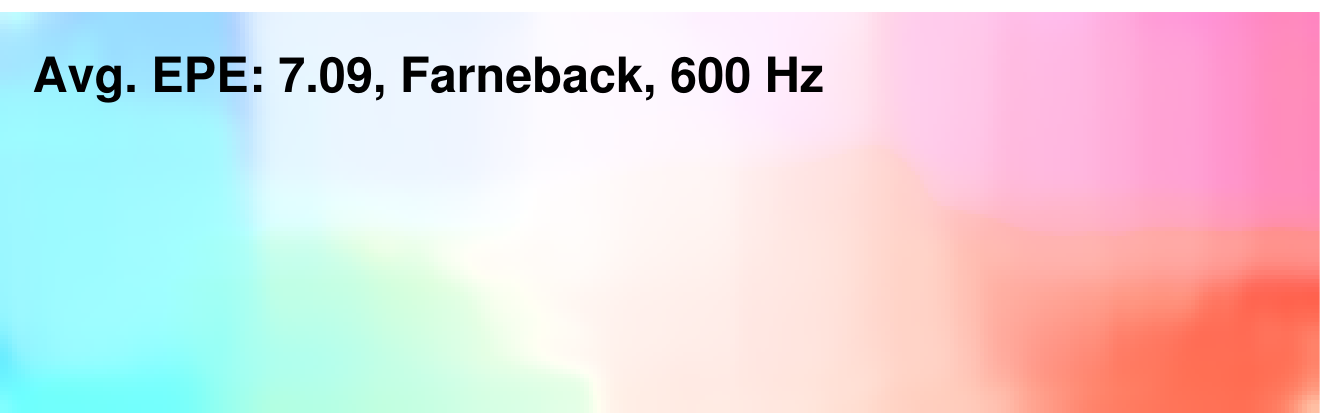}&
\includegraphics[width=0.195\textwidth]{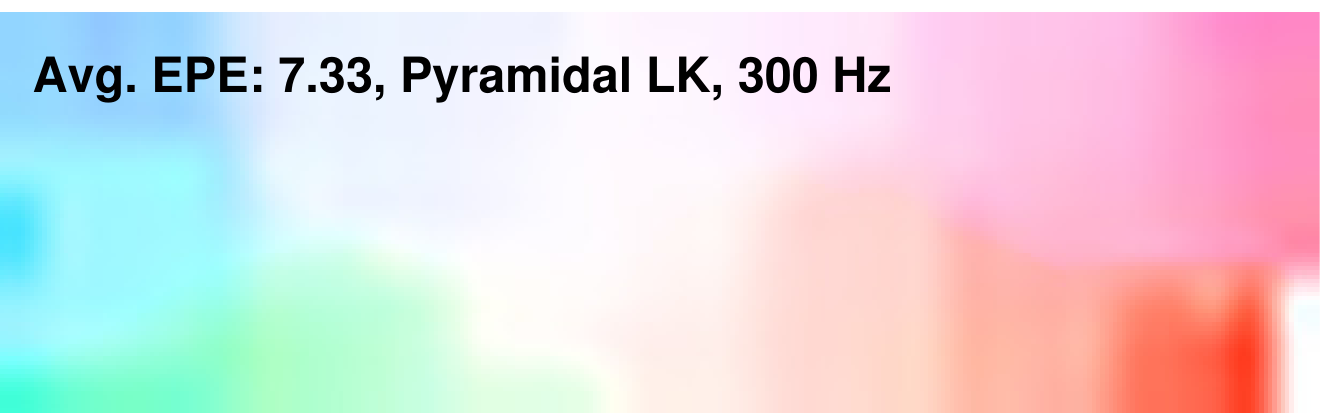}&
\includegraphics[width=0.195\textwidth]{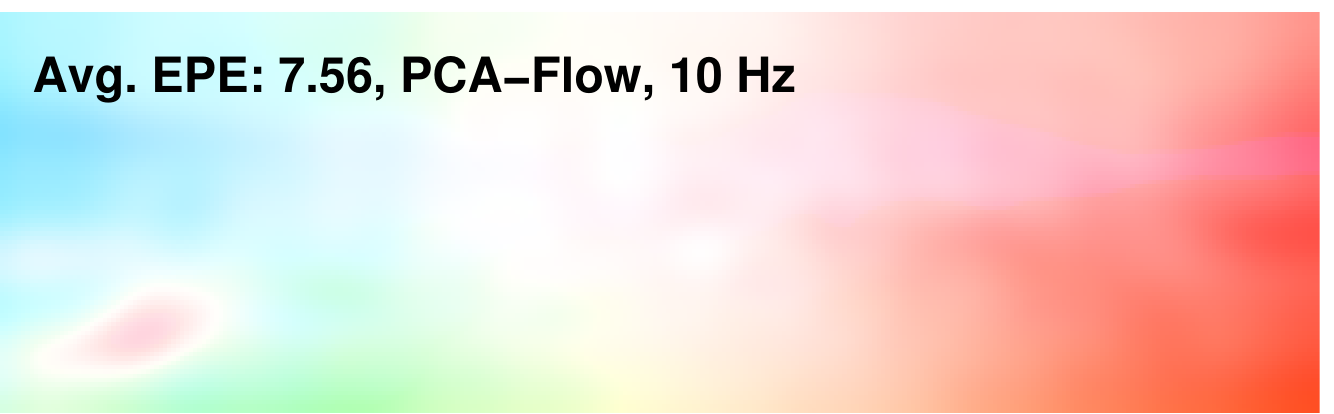}&
\includegraphics[width=0.195\textwidth]{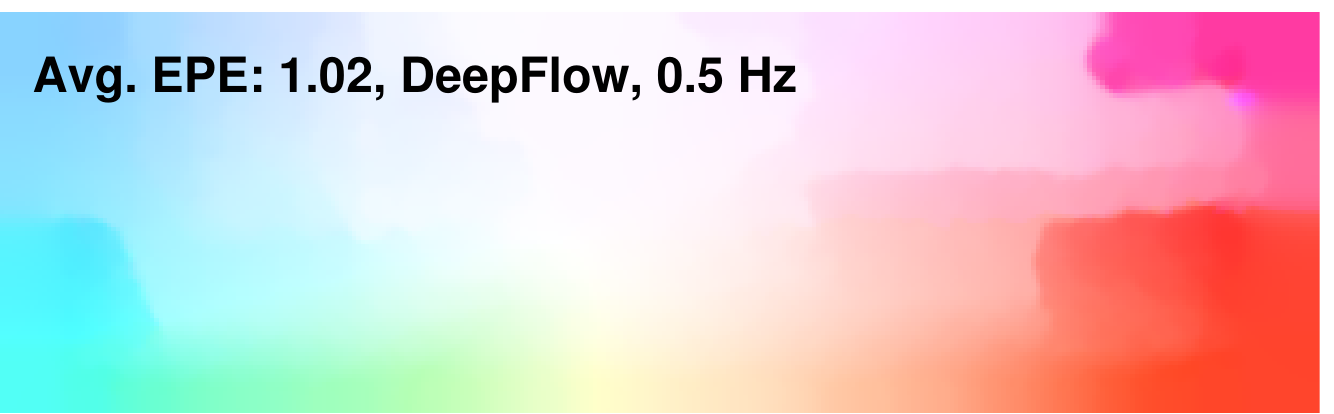}&
\includegraphics[width=0.195\textwidth]{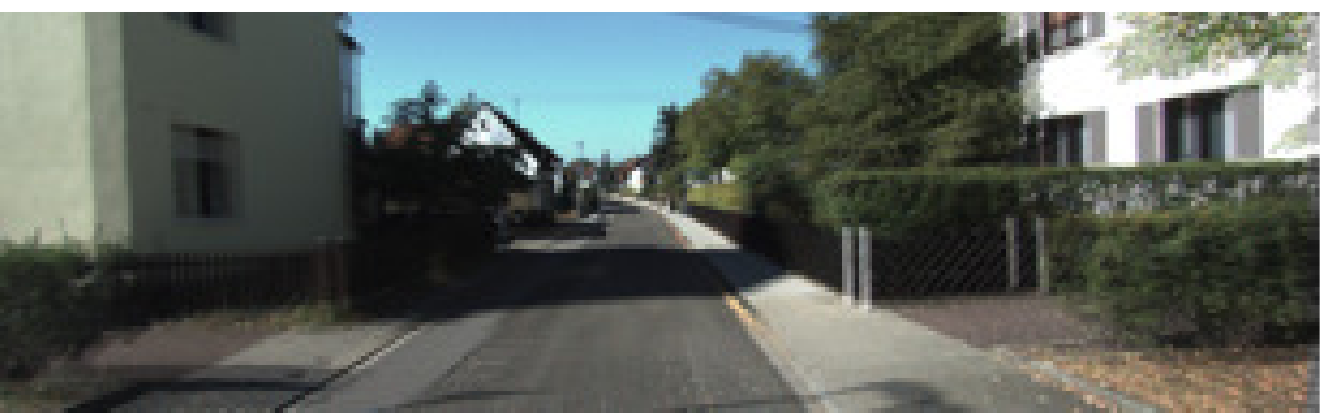}\\[2pt]
\hline \\[2pt]
\includegraphics[width=0.195\textwidth]{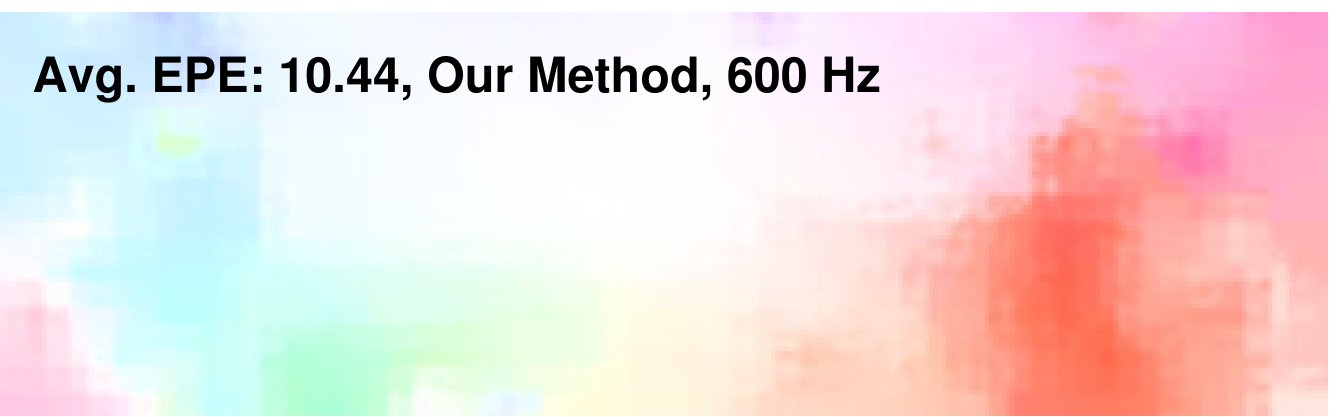}&
\includegraphics[width=0.195\textwidth]{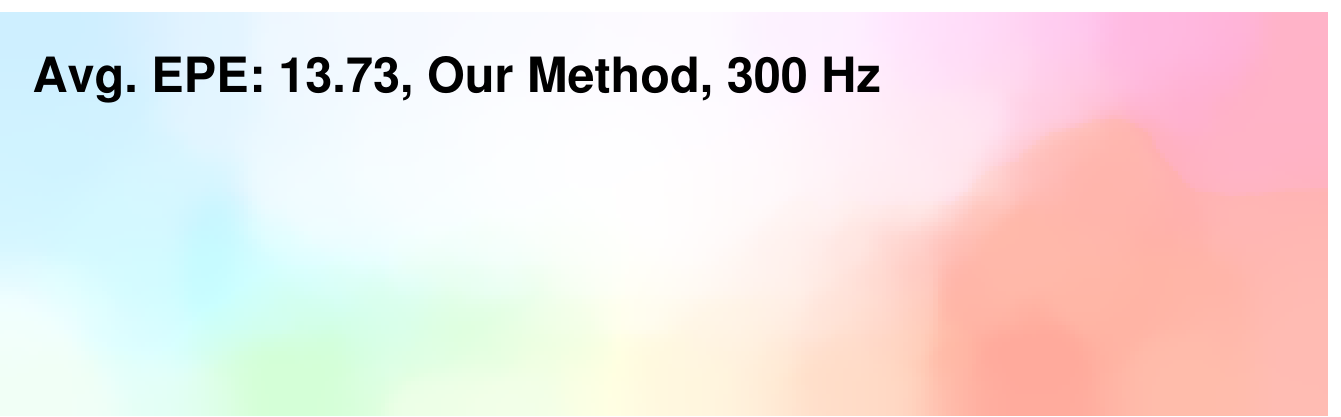}&
\includegraphics[width=0.195\textwidth]{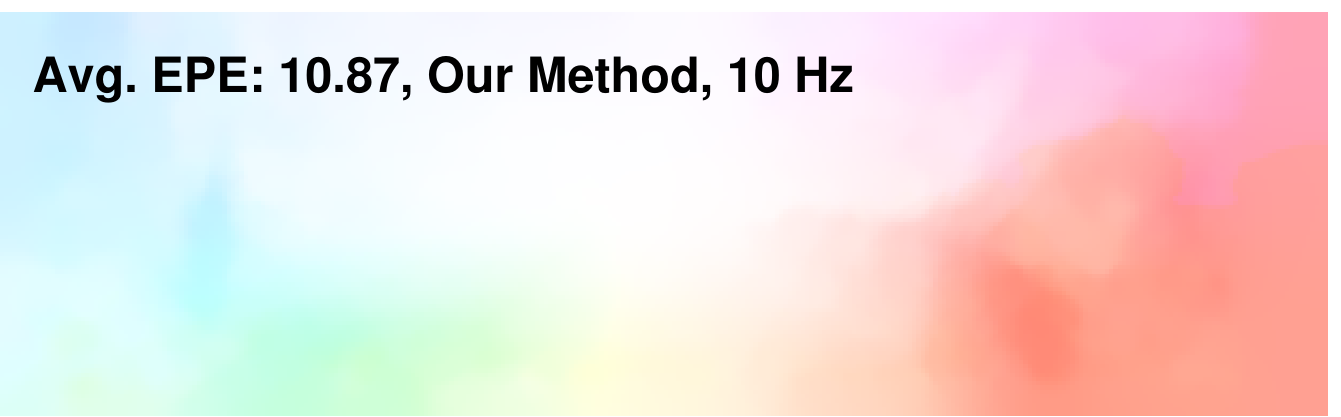}&
\includegraphics[width=0.195\textwidth]{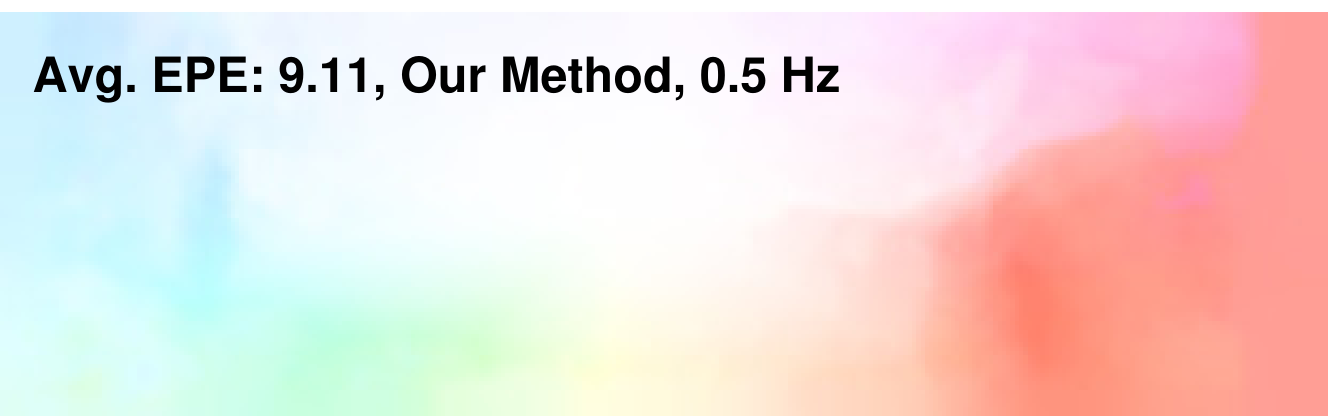}&
\includegraphics[width=0.195\textwidth]{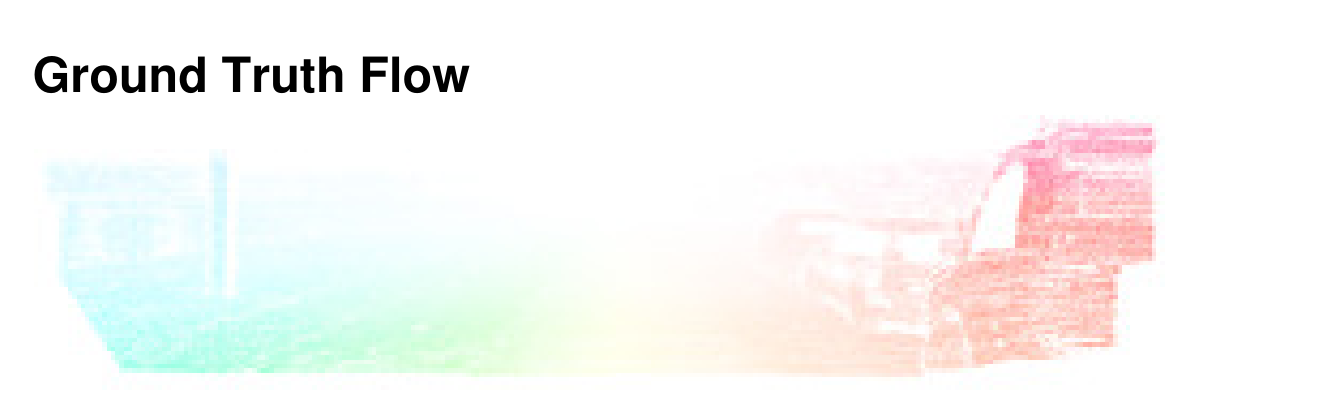}\\
\includegraphics[width=0.195\textwidth]{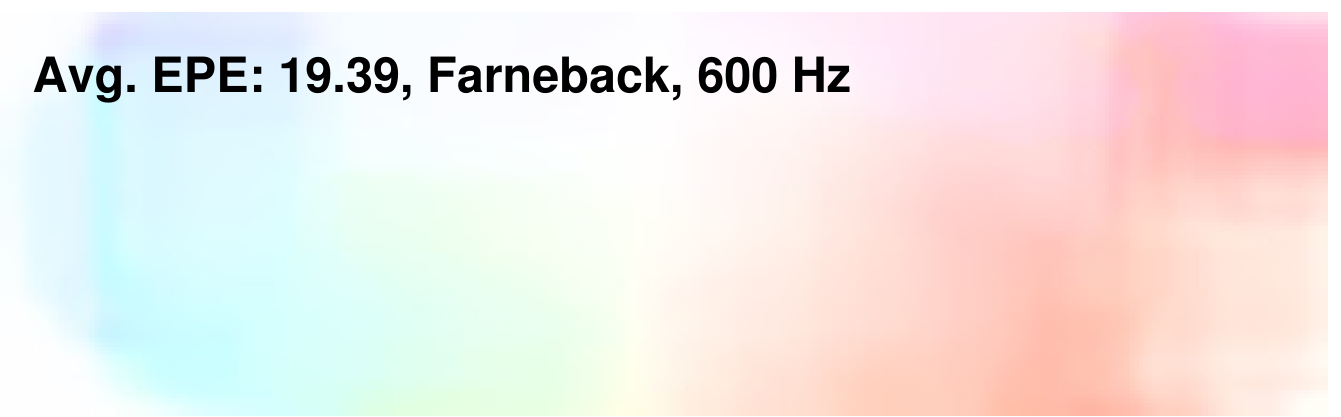}&
\includegraphics[width=0.195\textwidth]{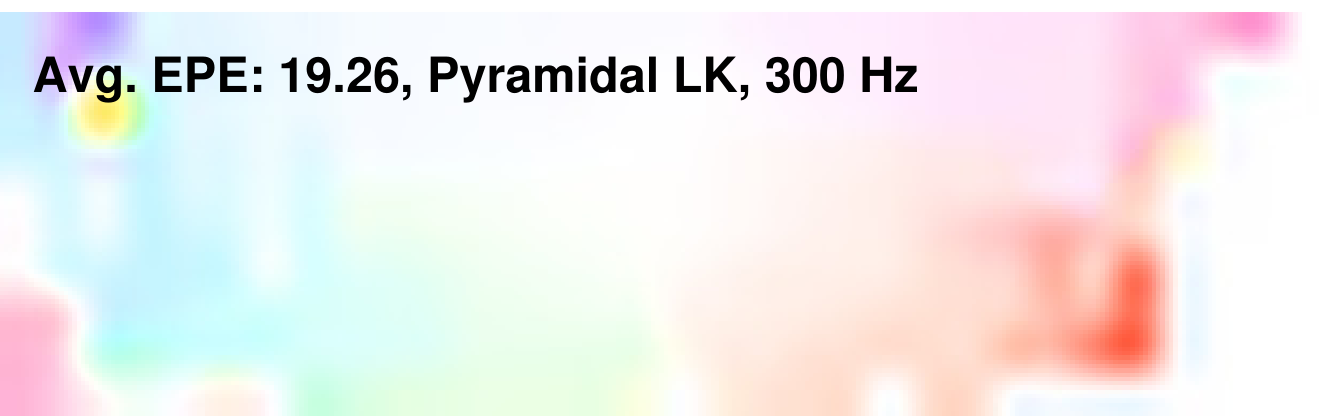}&
\includegraphics[width=0.195\textwidth]{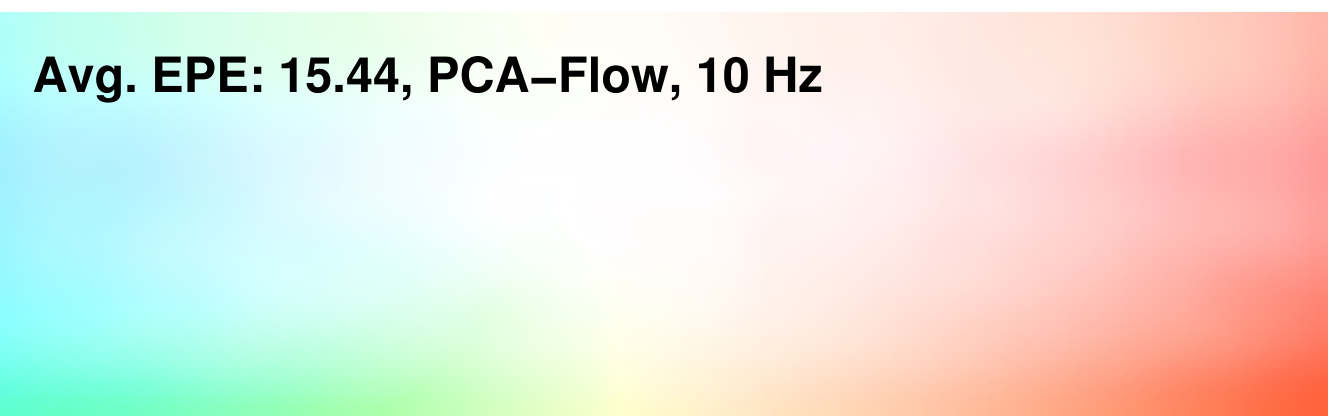}&
\includegraphics[width=0.195\textwidth]{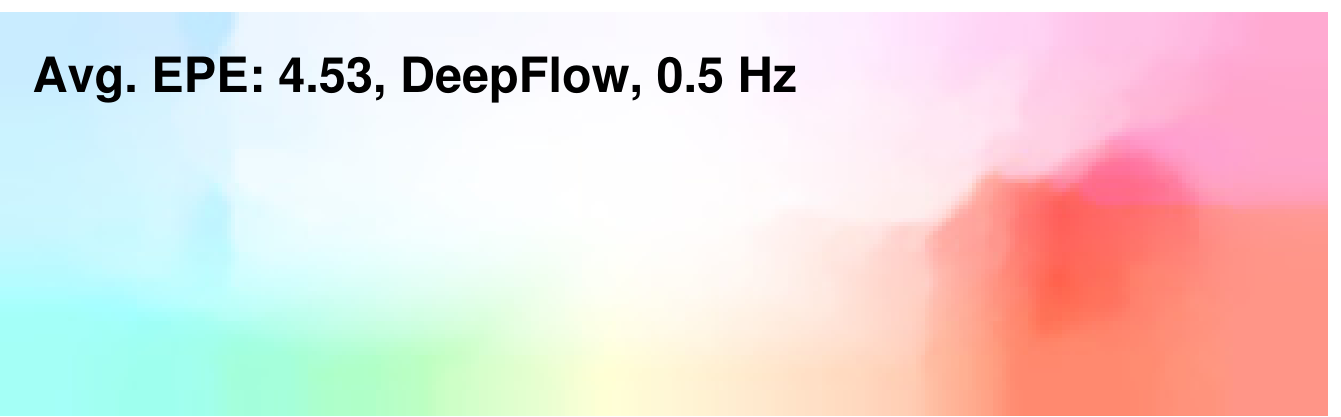}&
\includegraphics[width=0.195\textwidth]{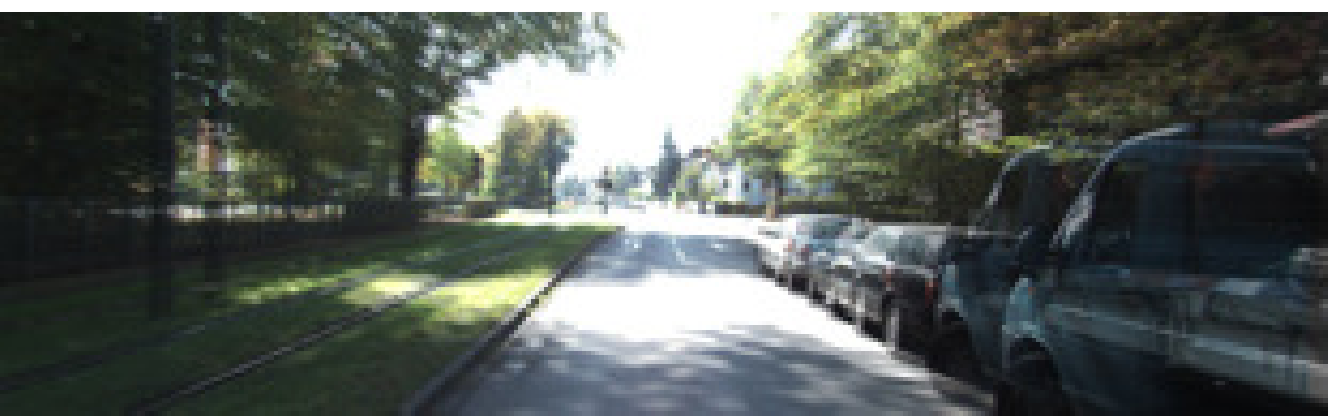}
\end{tabular}
}
 \vskip-9pt
\caption{Examplary results on KITTI (training). In each block of $2 \times 6$ images.  Top row, left to right: Our method for operating points ({\bf 1})-({\bf 4}), Ground Truth. Bottom row: Farneback, 600 Hz, Pyramidal LK 300Hz, PCA-Flow 10Hz, DeepFlow 0.5Hz, Original Image.}\label{fig:res_kitti_qualitative} 
 \vskip-7pt
\end{figure*} 

\begin{figure}
        \centering
        \includegraphics[width=0.43\textwidth]{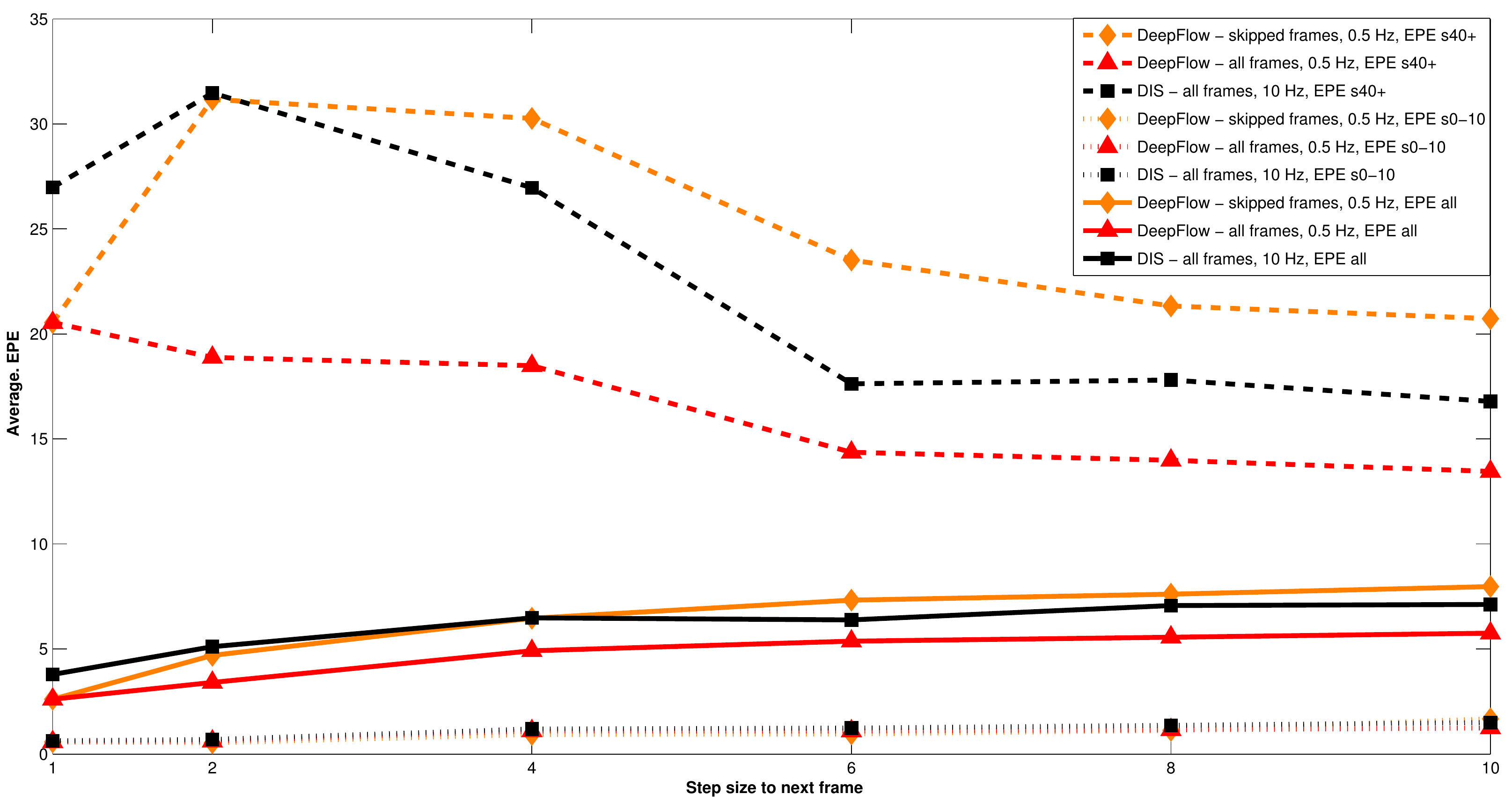}
        \caption{Flow result on Sintel with low temporal resolution. Accuracy of DeepFlow on large displacements versus DIS on small displacements, tracked through \emph{all intermediate} frames. As baseline we included the accuracy of DeepFlow for tracking small displacements. Note: While we use the same frame pairs to compute each vertical set of points, frame pairs differ over  stepsizes.}\label{fig:subsample}
\end{figure}

\begin{figure}\centering\setlength{\tabcolsep}{0.1pt}\renewcommand{\arraystretch}{0} 
        \resizebox{0.98\columnwidth}{!}
        {
        \begin{tabular}{c}
        \includegraphics[width=0.1\textwidth]{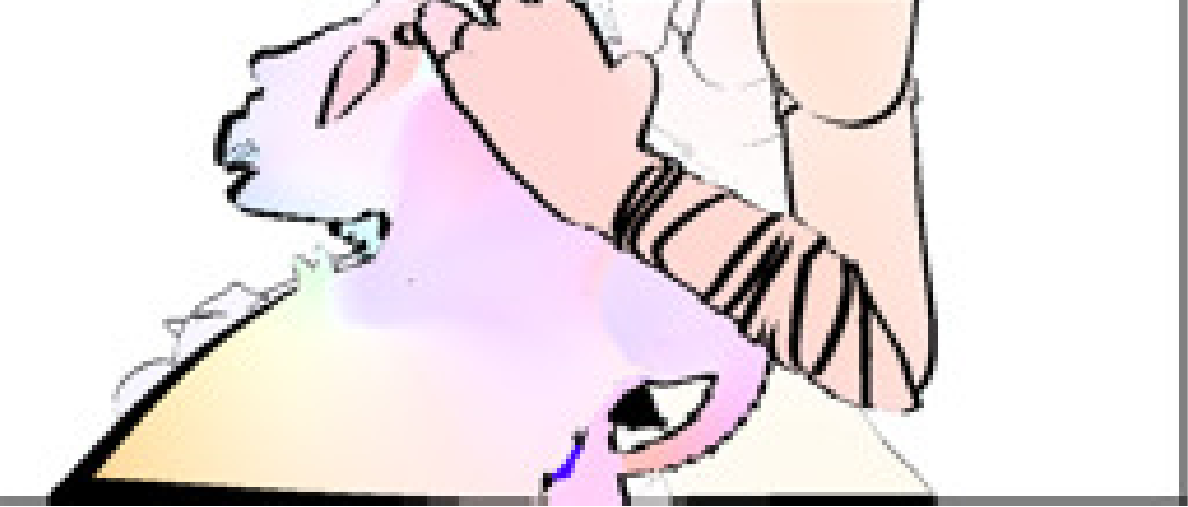}
        \includegraphics[width=0.1\textwidth]{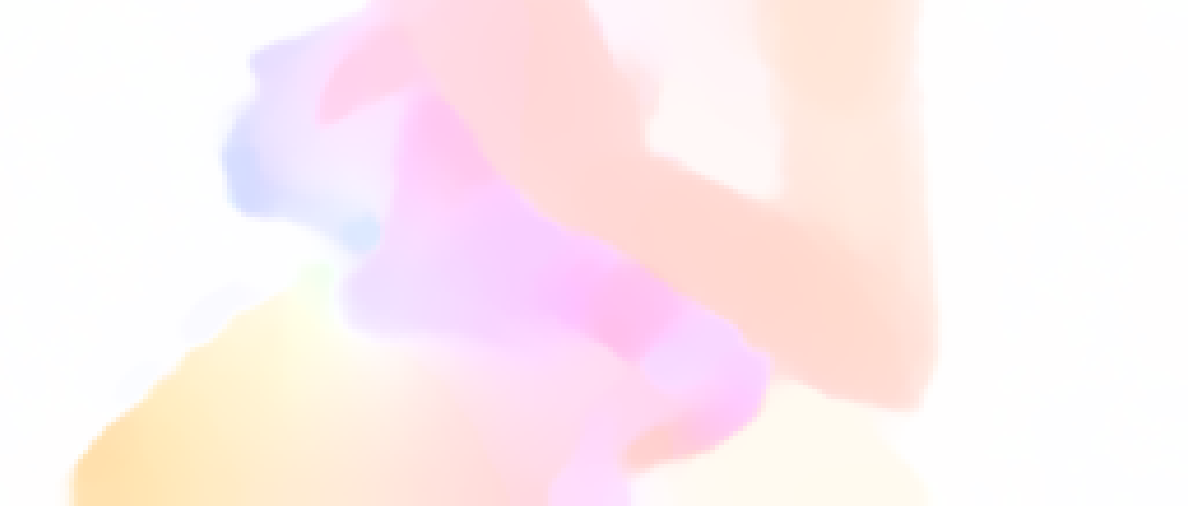}
        \includegraphics[width=0.1\textwidth]{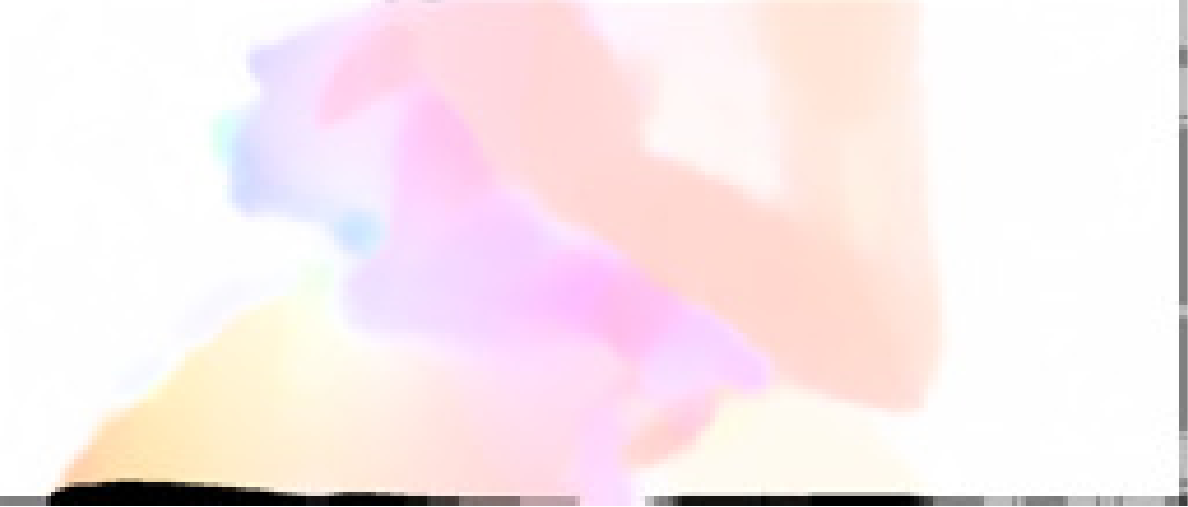}
        \includegraphics[width=0.1\textwidth]{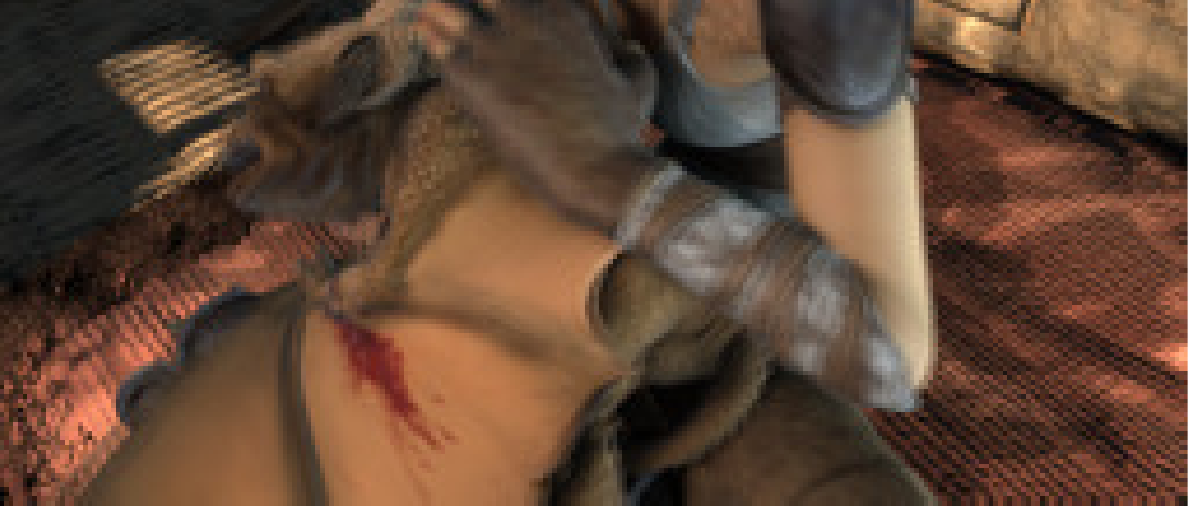} \\     
        \includegraphics[width=0.1\textwidth]{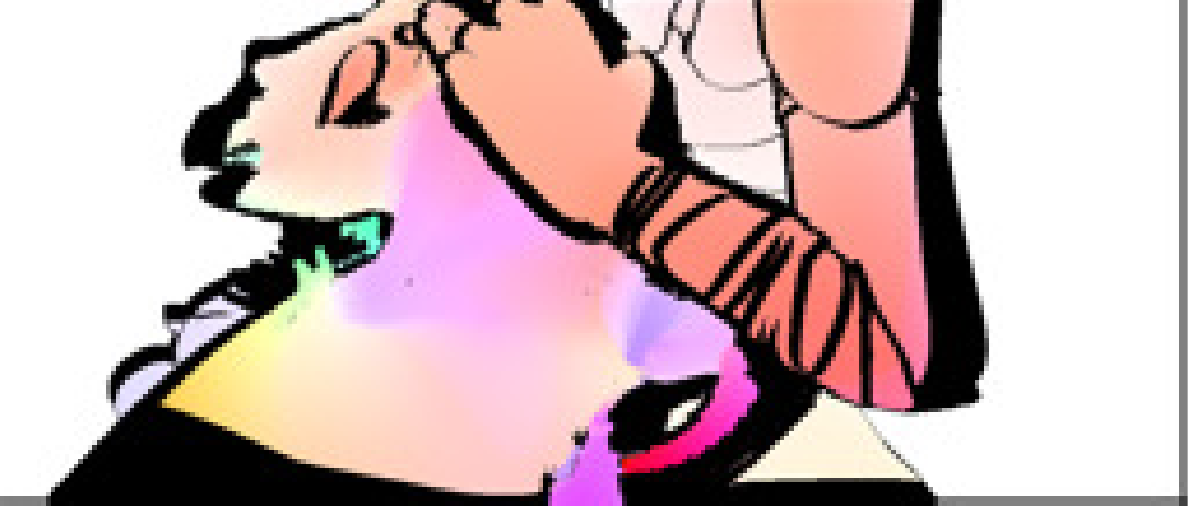}
        \includegraphics[width=0.1\textwidth]{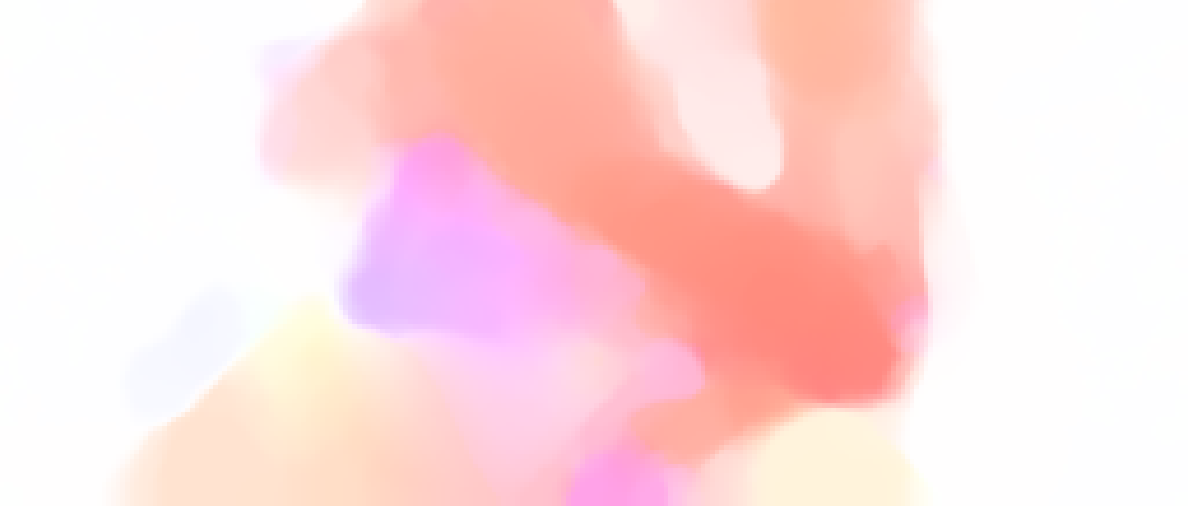}
        \includegraphics[width=0.1\textwidth]{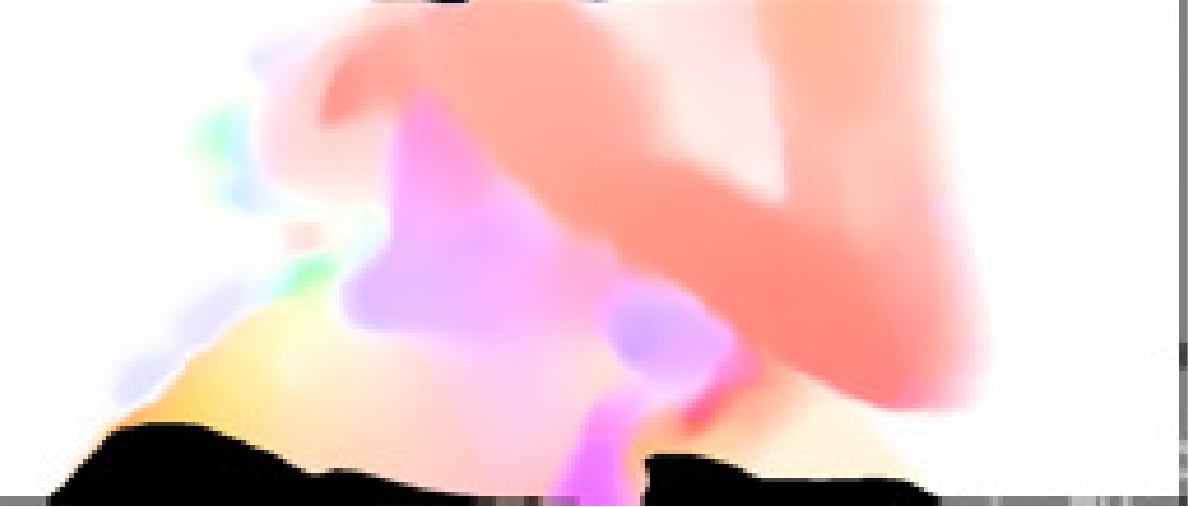}
        \includegraphics[width=0.1\textwidth]{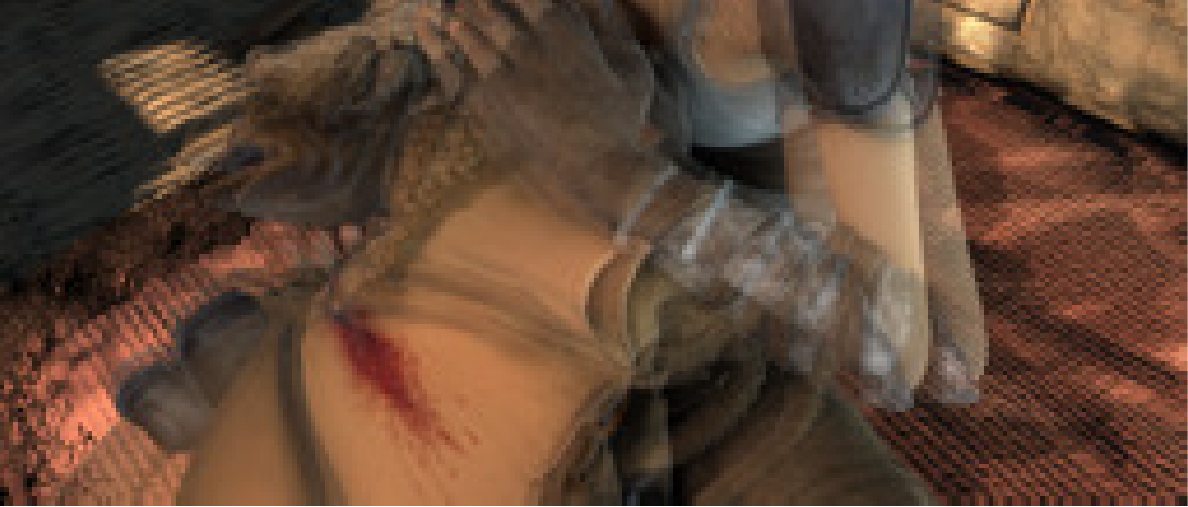} \\                     
        \includegraphics[width=0.1\textwidth]{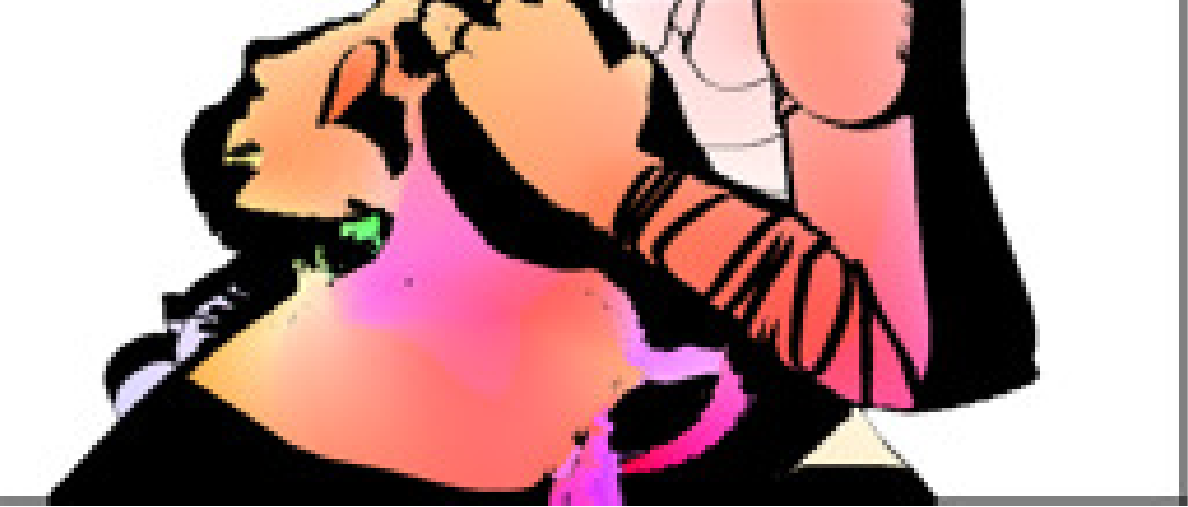}
        \includegraphics[width=0.1\textwidth]{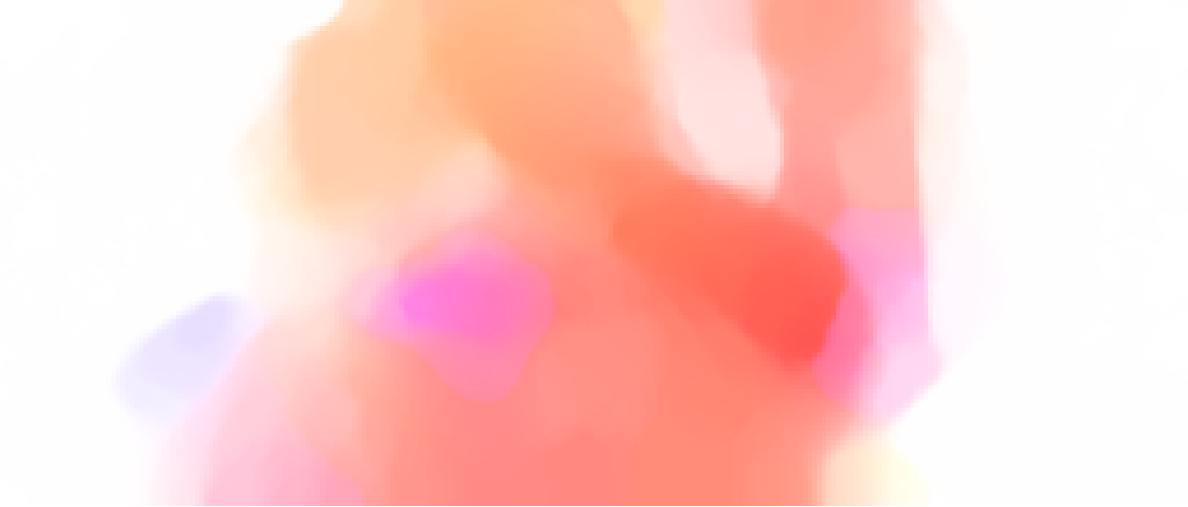}
        \includegraphics[width=0.1\textwidth]{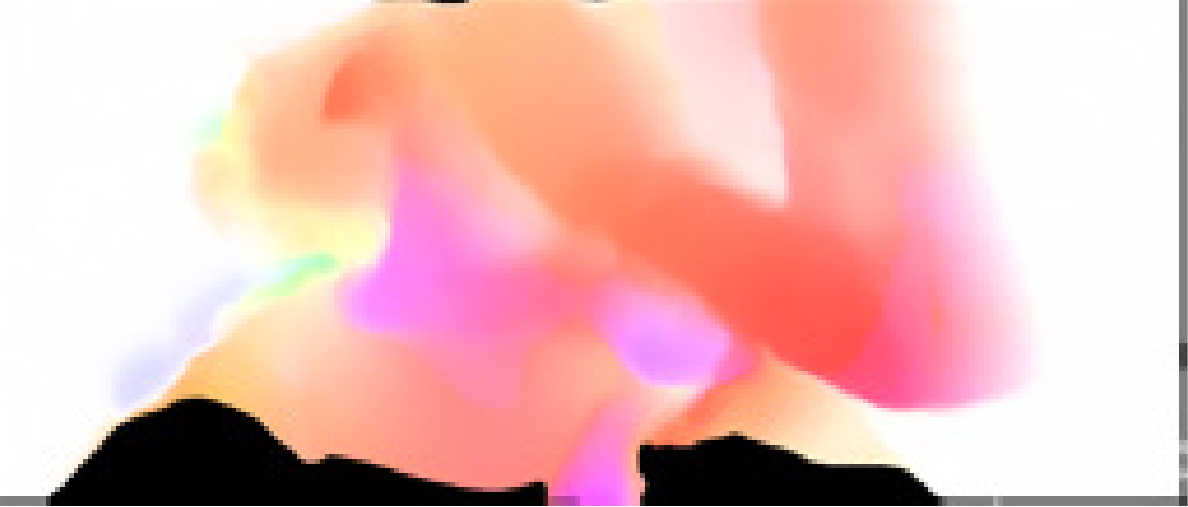}
        \includegraphics[width=0.1\textwidth]{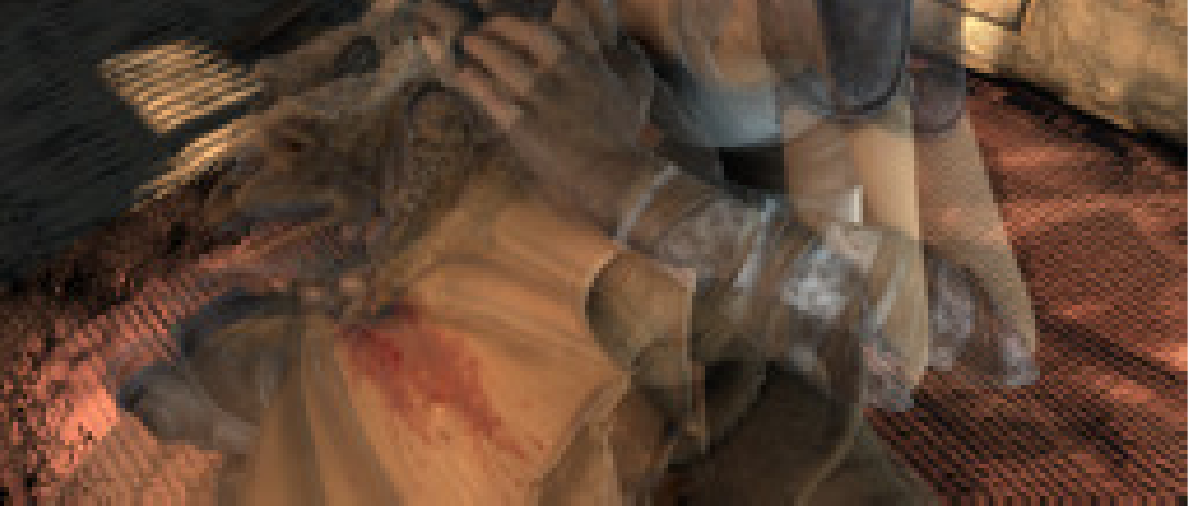}\\[2pt]
        \hline \\[2pt]
        \includegraphics[width=0.1\textwidth]{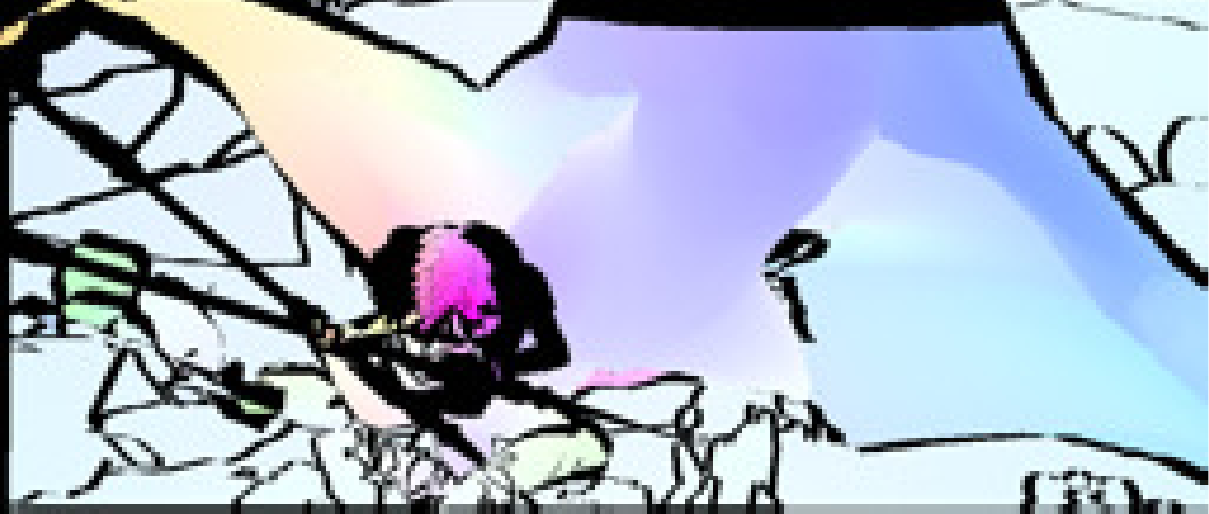}
        \includegraphics[width=0.1\textwidth]{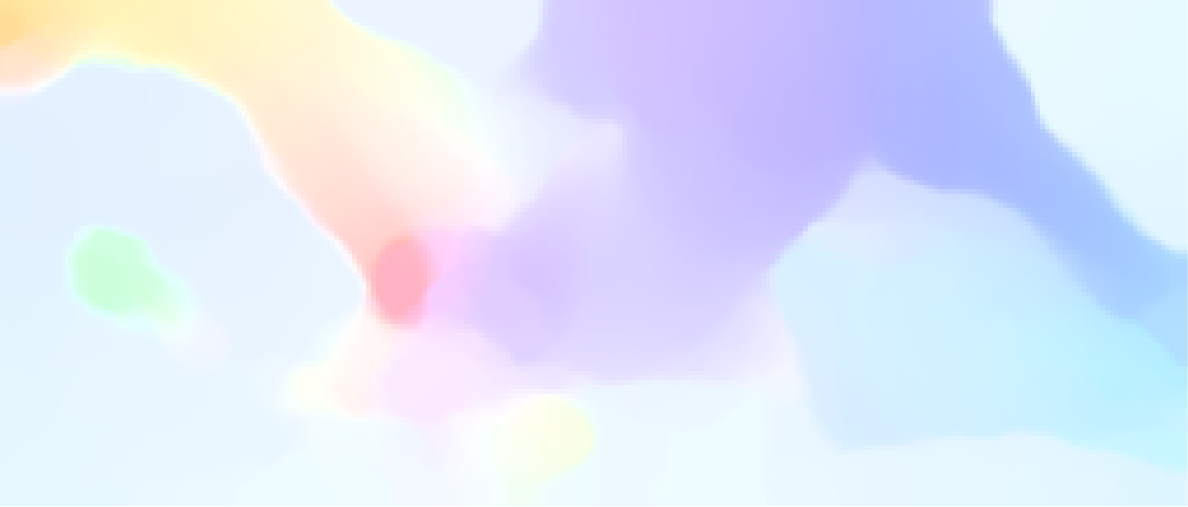}
        \includegraphics[width=0.1\textwidth]{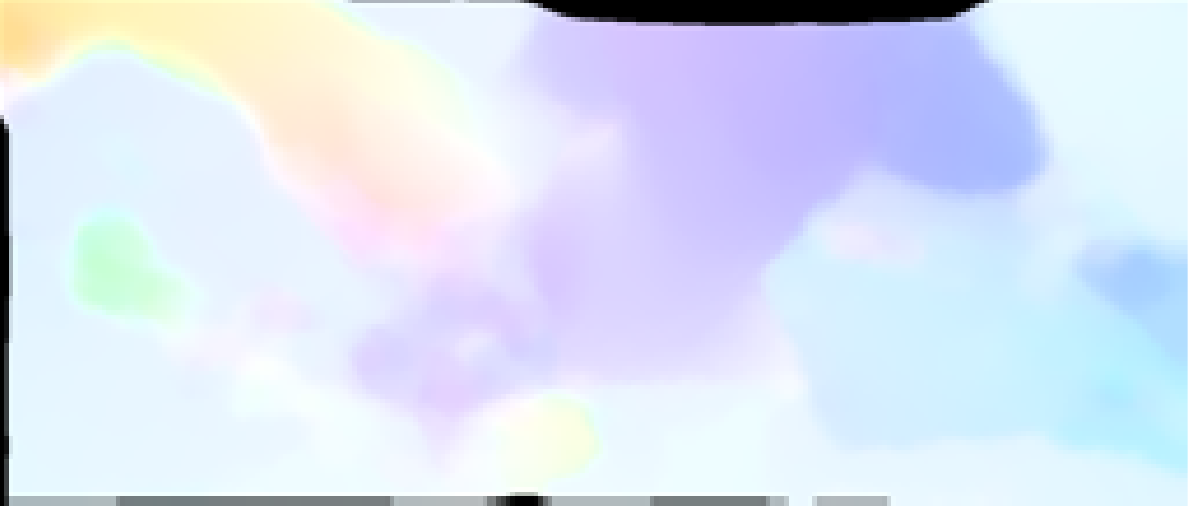}
        \includegraphics[width=0.1\textwidth]{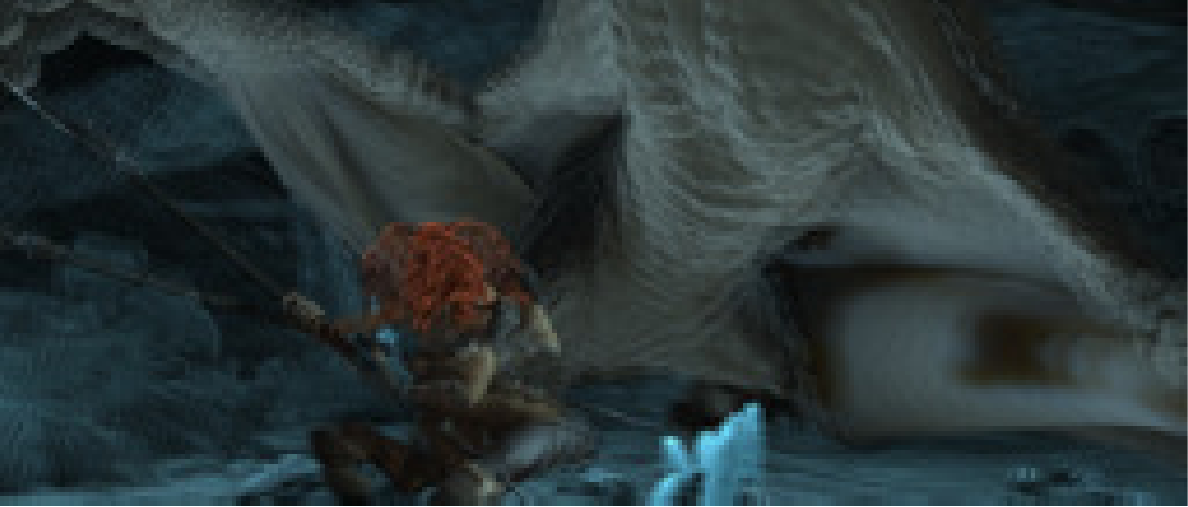} \\     
        \includegraphics[width=0.1\textwidth]{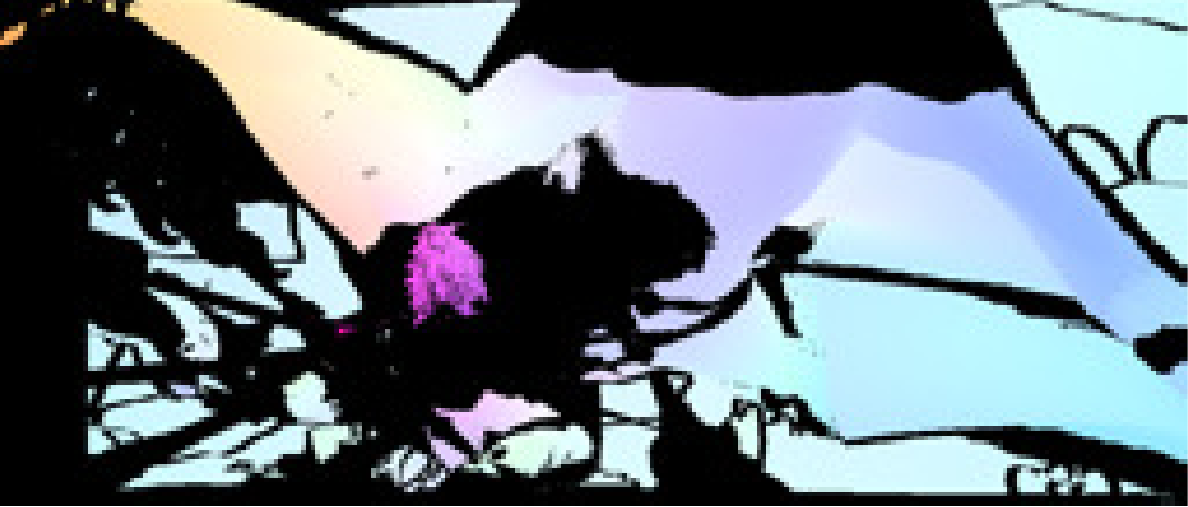}
        \includegraphics[width=0.1\textwidth]{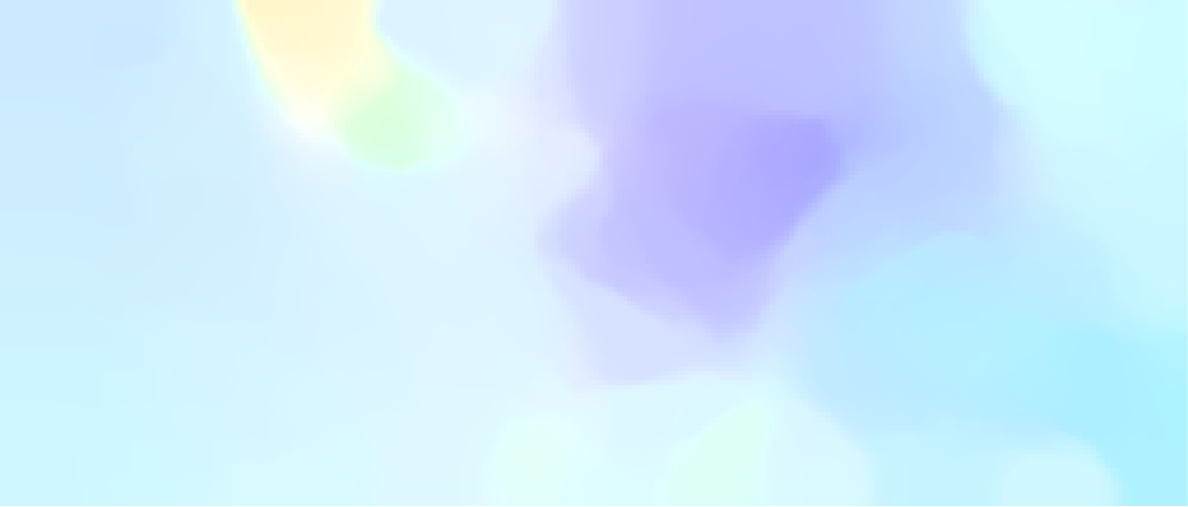}
        \includegraphics[width=0.1\textwidth]{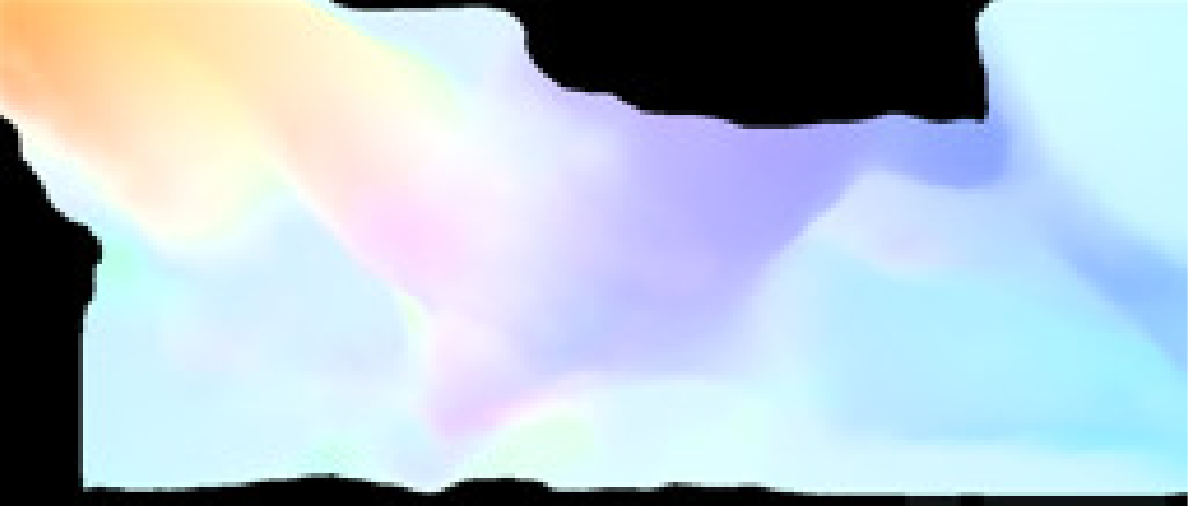}
        \includegraphics[width=0.1\textwidth]{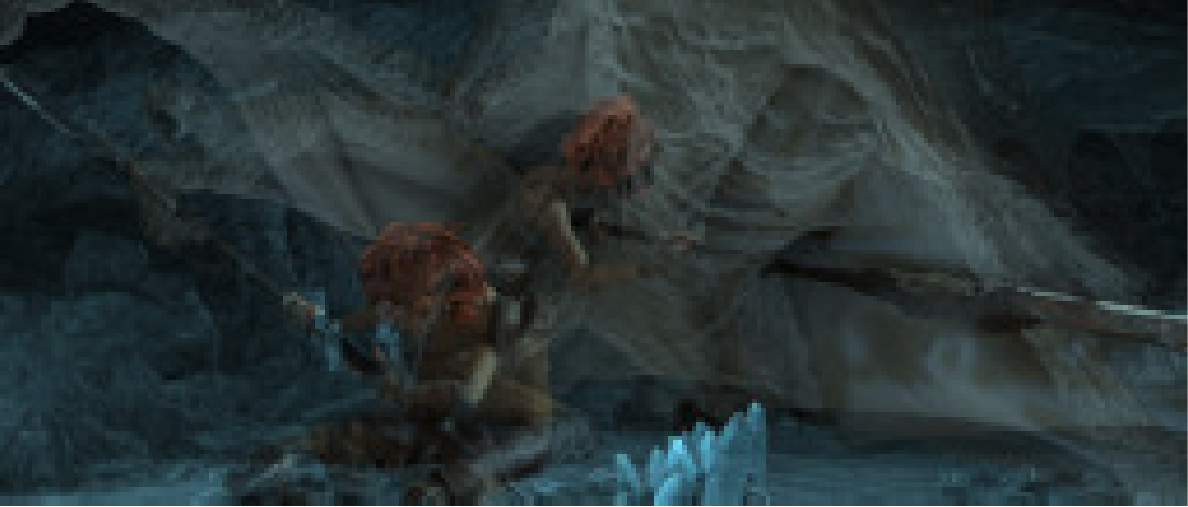} \\                     
        \includegraphics[width=0.1\textwidth]{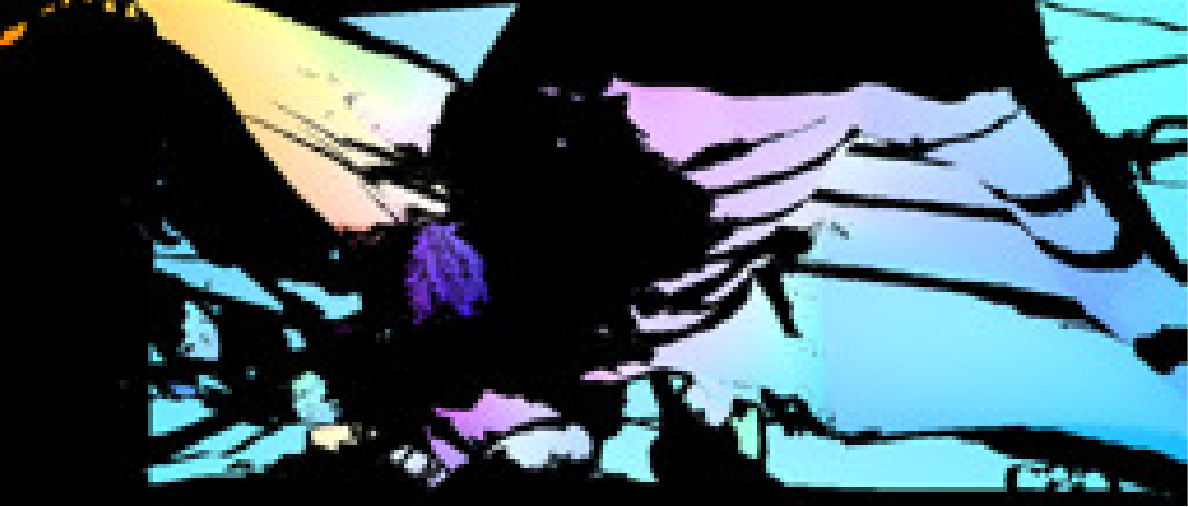}
        \includegraphics[width=0.1\textwidth]{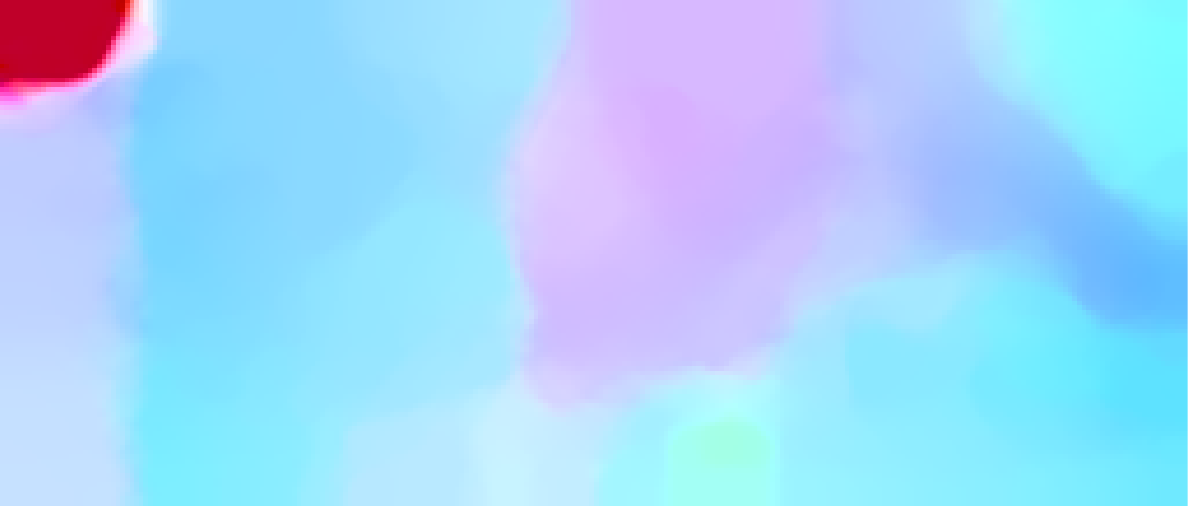}
        \includegraphics[width=0.1\textwidth]{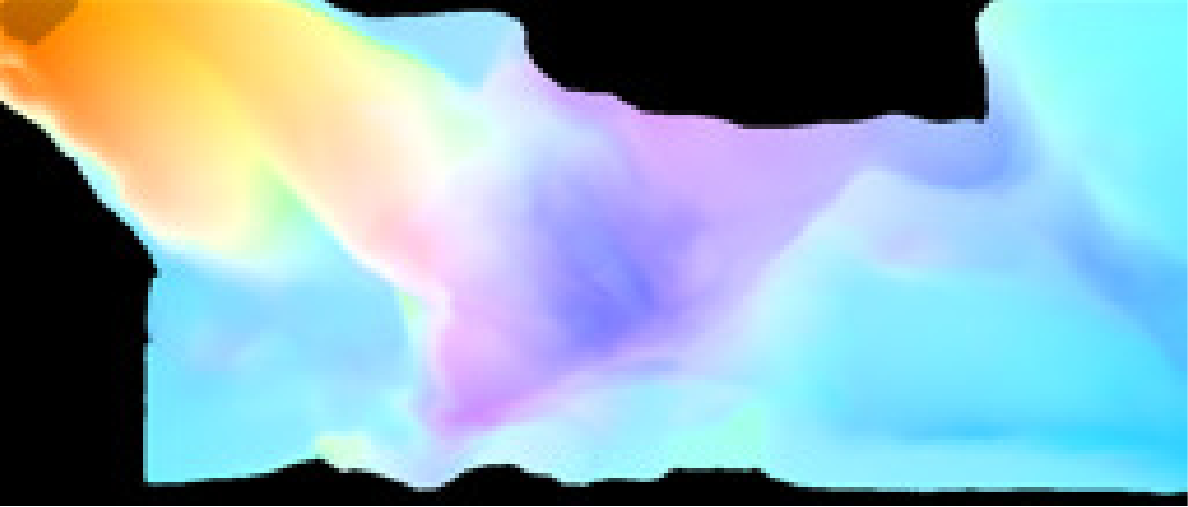}
        \includegraphics[width=0.1\textwidth]{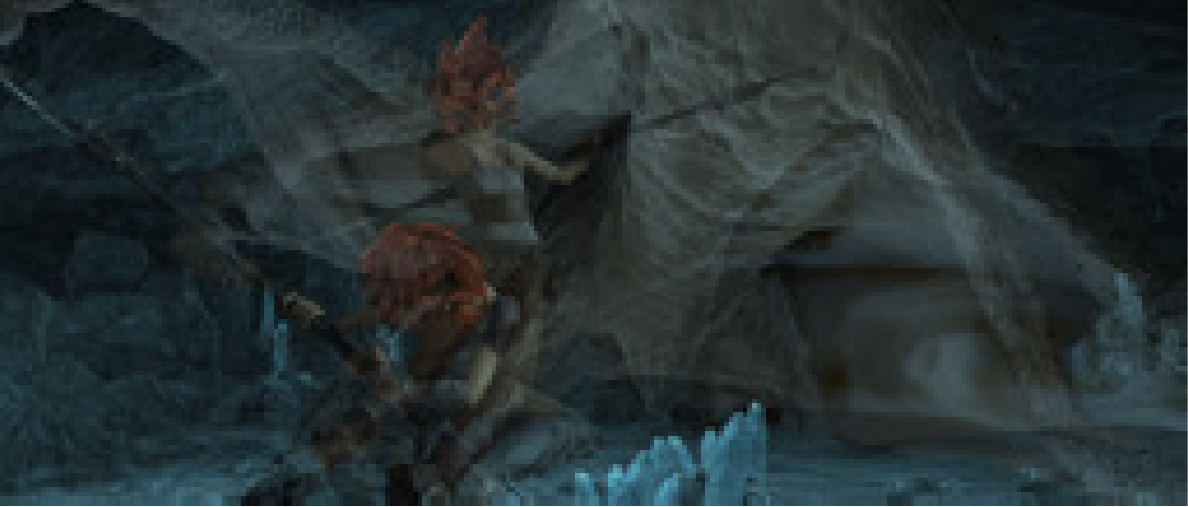}        
        \end{tabular}
        }
         \vskip-5pt
        \caption{Optical flow on Sintel with lower temporal resolution. In each block of 3x4: Rows, top to bottom, correspond to step sizes 1 (original frame-rate), 6, 10 frames. Columns, left to right, correspond to new ground truth, DeepFlow result, DIS result (through \emph{all intermediate frames}), original images. Large displacements are significantly better preserved by DIS through higher frame-rates.}\label{fig:subsample_exa}
        \vskip-7pt
\end{figure} 

\subsection{High frame-rate optical flow}
\label{ssc:framerate}
Often, a simpler and faster algorithm, combined with a higher temporal resolution in the data, can yield better accuracy than a more powerful algorithm, on lower temporal resolutions.
This has been analysed in detail in~\cite{handa2012real} for the task of visual odometry.
As noted in~\cite{benosman2014event,barranco2014contour} this is also the case for optical flow,  where large displacements, due to low-frame rate or strong motions are significantly more difficult to estimate than small displacements. 
In contrast to the recent focus on handling ever larger displacements~\cite{Brox-PAMI-2011, Xu-PAMI-2012, Weinzaepfel-ICCV-2013, Timofte-WACV-2015, Menze-PR-2015}, we want to analyse how \emph{decreasing} the run-time while \emph{increasing} the frame-rate affects our algorithm. For this experiment we selected a random subset of the Sintel training dataset, and synthesized new ground truth flow for lower frame-rates from the one provided in the dataset.
We create new ground truth for $1/2$ to $1/10$ of the source frame-rate from the original ground truth  and the additionally provided segmentation masks to invalidate occluded regions. 
We compare DeepFlow at a speed of 0.5Hz on this lower temporal resolution against our method (\emph{operating point} {\bf (3)}, 10 Hz), running through \emph{all intermediate frames} at the original, higher frame-rate. 
Thus, while DeepFlow has to handle larger displacements in one frame pair, our method has to handle smaller displacements, tracked through multiple frames and accumulates error drift.

We observe (Fig.~\ref{fig:subsample}) that DIS starts to outperform DeepFlow when running at twice the original frame-rate, notably for large displacements, while still being $10$ times faster. 
Fig.~\ref{fig:subsample_exa} shows examples of the new ground truth, results of DeepFlow and DIS.
We conclude, that it is advantageous to choose our method over DeepFlow, aimed at recovering large displacements, when the combination of video frame-rate and run-time per frame can be chosen freely.

\section{Conclusions}
\label{sec:conclusions}
In this paper we presented a novel and simple  way of computing dense optical flow. 
The presented approach trades off a lower flow estimation error for large decreases in run-time:
For the same level of error, the presented method is two orders of magnitude faster than current state-of-the-art approaches, as shown in experiments on synthetic (Sintel) and realistic (KITTI) optical flow benchmarks.

In the future we will address some open problems with our method:
Due to the coarse-to-fine approach small and fast motions can sometimes get lost beyond recovery. 
A sampling-based approach to recover over-smoothed object motions at finer scales may alleviate this problem.
The implicit minimization of the L2 matching error in our method is not invariant to many modes of change, such as in contrast, deformations, and occlusions. 
More robust error metrics may be helpful here.
Furthermore, a GPU implementation may yield another significant speed-up.

{\bf Acknowledgments:} This work was supported by the European Research Council project \emph{VarCity (\#273940)}. We thank Richard Hartley for his pertinent input on this work.

\newpage
{
\scriptsize
\bibliographystyle{CVPR_style/ieee}
\bibliography{egbib}

\begin{thebibliography}{10}\itemsep=-1pt

\bibitem{Baker-IJCV-2004}
Lucas-kanade 20 years on: A unifying framework.
\newblock {\em International Journal of Computer Vision}, 56(3):221--255, 2004.

\bibitem{Bailer-CoRR-2015}
C.~Bailer, B.~Taetz, and D.~Stricker.
\newblock Flow fields:dense correspondence fields for highly accurate large
  displacement optical flow estimation.
\newblock In {\em ICCV}, 2015.

\bibitem{Baker-CVPR-2001}
S.~Baker and I.~Matthews.
\newblock Equivalence and efficiency of image alignment algorithms.
\newblock In {\em CVPR}, 2001.

\bibitem{Bao-TIP-2014}
L.~Bao, Q.~Yang, and H.~Jin.
\newblock Fast edge-preserving patchmatch for large displacement optical flow.
\newblock {\em Image Processing, IEEE Transactions on}, 23(12):4996--5006, Dec
  2014.

\bibitem{Barnes-ECCV-2010}
C.~Barnes, E.~Shechtman, D.~B. Goldman, and A.~Finkelstein.
\newblock The generalized {PatchMatch} correspondence algorithm.
\newblock In {\em ECCV}, 2010.

\bibitem{barranco2014contour}
F.~Barranco, C.~Fermuller, and Y.~Aloimonos.
\newblock Contour motion estimation for asynchronous event-driven cameras.
\newblock {\em Proceedings of the IEEE}, 102(10):1537--1556, 2014.

\bibitem{Bay-CVIU-2008}
H.~Baya, A.~Essa, T.~Tuytelaarsb, and L.~{Van Gool}.
\newblock Speeded-up robust features (surf).
\newblock {\em CVIU}, 110(3):346--359, 2008.

\bibitem{benosman2014event}
R.~Benosman, C.~Clercq, X.~Lagorce, S.-H. Ieng, and C.~Bartolozzi.
\newblock Event-based visual flow.
\newblock {\em Neural Networks and Learning Systems, IEEE Transactions on},
  25(2):407--417, 2014.

\bibitem{Black-CVIU-1996}
M.~J. Black and P.~Anandan.
\newblock The robust estimation of multiple motions: parametric and
  piecewise-smooth flow fields.
\newblock {\em CVIU}, 1996.

\bibitem{bouguet2001pyramidal}
J.-Y. Bouguet.
\newblock Pyramidal implementation of the affine lucas kanade feature tracker
  description of the algorithm.
\newblock {\em Intel Corporation}, 5:1--10, 2001.

\bibitem{BrauxZin-ICCV-2013}
J.~Braux-Zin, R.~Dupont, and A.~Bartoli.
\newblock A general dense image matching framework combining direct and
  feature-based costs.
\newblock In {\em ICCV}, 2013.

\bibitem{Brox-ECCV-2004}
T.~Brox, A.~Bruhn, N.~Papenberg, and J.~Weickert.
\newblock High accuracy optical flow estimation based on a theory for warping.
\newblock In {\em ECCV}, 2004.

\bibitem{Brox-PAMI-2011}
T.~Brox and J.~Malik.
\newblock Large displacement optical flow: descriptor matching in variational
  motion estimation.
\newblock {\em IEEE Trans. PAMI}, 2011.

\bibitem{Butler-ECCV-2012}
D.~J. Butler, J.~Wulff, G.~B. Stanley, and M.~J. Black.
\newblock A naturalistic open source movie for optical flow evaluation.
\newblock In {\em ECCV}, 2012.

\bibitem{Dalal-CVPR-2005}
N.~Dalal and B.~Triggs.
\newblock Histograms of oriented gradients for human detection.
\newblock In {\em CVPR}, 2005.

\bibitem{Enkelmann1988}
W.~Enkelmann.
\newblock Investigations of multigrid algorithms for the estimation of optical
  flow fields in image sequences.
\newblock {\em Computer Vision, Graphics, and Image Processing}, 43:150--177,
  1988.

\bibitem{Farneback-IA-2003}
G.~Farneb\"ack.
\newblock Two-frame motion estimation based on polynomial expansion.
\newblock In {\em Image Analysis}, volume 2749 of {\em Lecture Notes in
  Computer Science}, pages 363--370. 2003.

\bibitem{Fischer-ICCV-2015}
P.~Fischer, A.~Dosovitskiy, E.~Ilg, P.~H{\"a}usser, C.~Haz{\i}rba{\c{s}},
  V.~Golkov, P.~van~der Smagt, D.~Cremers, and T.~Brox.
\newblock Flownet: Learning optical flow with convolutional networks.
\newblock In {\em ICCV}, 2015.

\bibitem{Forster2014}
C.~Forster, M.~Pizzoli, and D.~Scaramuzza.
\newblock {SVO: Fast semi-direct monocular visual odometry}.
\newblock {\em ICRA}, pages 15--22, may 2014.

\bibitem{Fortun-CVIU-2015}
D.~Fortun, P.~Bouthemy, and C.~Kervrann.
\newblock Optical flow modeling and computation: A survey.
\newblock {\em Computer Vision and Image Understanding}, 134:1 -- 21, 2015.
\newblock Image Understanding for Real-world Distributed Video Networks.

\bibitem{Geiger-IJRR-2013}
A.~Geiger, P.~Lenz, C.~Stiller, and R.~Urtasun.
\newblock Vision meets robotics: The {KITTI} dataset.
\newblock {\em IJRR}, 2013.

\bibitem{Geiger_2010_ACCV_ELAS}
A.~Geiger, M.~Roser, and R.~Urtasun.
\newblock Efficient large-scale stereo matching.
\newblock In {\em ACCV}, ACCV, 2011.

\bibitem{hager1998efficient}
G.~D. Hager and P.~N. Belhumeur.
\newblock Efficient region tracking with parametric models of geometry and
  illumination.
\newblock {\em TPAMI}, 20(10):1025--1039, 1998.

\bibitem{handa2012real}
A.~Handa, R.~A. Newcombe, A.~Angeli, and A.~J. Davison.
\newblock Real-time camera tracking: When is high frame-rate best?
\newblock In {\em ECCV}, pages 222--235. 2012.

\bibitem{Heitz-PAMI-1993}
F.~Heitz and P.~Bouthemy.
\newblock Multimodal estimation of discontinuous optical flow using markov
  random fields.
\newblock {\em TPAMI}, 15(12):1217--1232, Dec 1993.

\bibitem{hirschmuller2008stereo}
H.~Hirschm{\"u}ller.
\newblock Stereo processing by semiglobal matching and mutual information.
\newblock {\em TPAMI}, 30(2):328--341, 2008.

\bibitem{Horn-1981}
B.~K. Horn and B.~G. Schunck.
\newblock Determining optical flow.
\newblock {\em Proc. SPIE 0281, Techniques and Applications of Image
  Understanding}, 1981.

\bibitem{Kennedy-EMMCVPR-2015}
R.~Kennedy and C.~Taylor.
\newblock Optical flow with geometric occlusion estimation and fusion of
  multiple frames.
\newblock In {\em Energy Minimization Methods in Computer Vision and Pattern
  Recognition}, volume 8932 of {\em Lecture Notes in Computer Science}, pages
  364--377. 2015.

\bibitem{Klein2007}
G.~Klein and D.~Murray.
\newblock {Parallel Tracking and Mapping for Small AR Workspaces}.
\newblock {\em ISMAR}, 2007.

\bibitem{konolige1998small}
K.~Konolige.
\newblock Small vision systems: Hardware and implementation.
\newblock In {\em Robotics Research}, pages 203--212. Springer London, 1998.

\bibitem{Leordeanu-ICCV-2013}
M.~Leordeanu, A.~Zanfir, and C.~Sminchisescu.
\newblock Locally affine sparse-to-dense matching for motion and occlusion
  estimation.
\newblock In {\em ICCV}, 2013.

\bibitem{lichtsteiner2008128}
P.~Lichtsteiner, C.~Posch, and T.~Delbruck.
\newblock A 128$\times$ 128 120 db 15 $\mu$s latency asynchronous temporal
  contrast vision sensor.
\newblock {\em Solid-State Circuits, IEEE Journal of}, 43(2):566--576, 2008.

\bibitem{Liu-PAMI-2011}
C.~Liu, J.~Yuen, and A.~Torralba.
\newblock {SIFT} flow: Dense correspondence across scenes and its applications.
\newblock {\em TPAMI}, 2011.

\bibitem{Lowe-IJCV-2004}
D.~Lowe.
\newblock Distinctive image features from scale-invariant keypoints.
\newblock {\em IJCV}, 60(2):91--110, 2004.

\bibitem{Lucas-IJCAI-1981}
B.~D. Lucas and T.~Kanade.
\newblock An iterative image registration technique with an application to
  stereo vision.
\newblock In {\em IJCAI}, volume~81, pages 674--679, 1981.

\bibitem{Menze-PR-2015}
M.~Menze, C.~Heipke, and A.~Geiger.
\newblock In {\em Pattern Recognition}, volume 9358 of {\em Lecture Notes in
  Computer Science}, pages 16--28. 2015.

\bibitem{Mikolajczyk-IJCV-2005}
K.~Mikolajczyk, T.~Tuytelaars, C.~Schmid, A.~Zisserman, J.~Matas,
  F.~Schaffalitzky, T.~Kadir, and L.~{Van Gool}.
\newblock A comparison of affine region detectors.
\newblock {\em IJCV}, 2005.

\bibitem{Papenberg-IJCV-2006}
N.~Papenberg, A.~Bruhn, T.~Brox, S.~Didas, and J.~Weickert.
\newblock Highly accurate optic flow computation with theoretically justified
  warping.
\newblock {\em IJCV}, 2006.

\bibitem{Pauwels-TOC-2012}
K.~Pauwels, M.~Tomasi, J.~Diaz~Alonso, E.~Ros, and M.~Van~Hulle.
\newblock A comparison of fpga and gpu for real-time phase-based optical flow,
  stereo, and local image features.
\newblock {\em Computers, IEEE Transactions on}, 61(7):999--1012, 2012.

\bibitem{Plyer-RTIP-2014}
A.~Plyer, G.~Le~Besnerais, and F.~Champagnat.
\newblock Massively parallel lucas kanade optical flow for real-time video
  processing applications.
\newblock {\em Journal of Real-Time Image Processing}, pages 1--18, 2014.

\bibitem{Revaud-CVPR-2015}
J.~Revaud, P.~Weinzaepfel, Z.~Harchaoui, and C.~Schmid.
\newblock {EpicFlow:Edge-Preserving Interpolation of Correspondences for
  Optical Flow}.
\newblock In {\em CVPR}, 2015.

\bibitem{scharstein2014high}
D.~Scharstein, H.~Hirschm{\"u}ller, Y.~Kitajima, G.~Krathwohl,
  N.~Ne{\v{s}}i{\'c}, X.~Wang, and P.~Westling.
\newblock High-resolution stereo datasets with subpixel-accurate ground truth.
\newblock In {\em GCPR}. 2014.

\bibitem{senst2012robust}
T.~Senst, V.~Eiselein, and T.~Sikora.
\newblock Robust local optical flow for feature tracking.
\newblock {\em Circuits and Systems for Video Technology, IEEE Transactions
  on}, 22(9):1377--1387, 2012.

\bibitem{Srinivasan2013High}
N.~Srinivasan, R.~Roberts, and F.~Dellaert.
\newblock High frame rate egomotion estimation.
\newblock In {\em ICVS}, ICVS'13, pages 183--192, 2013.

\bibitem{Steinbrucker-ICCV-2009}
F.~Steinbrucker, T.~Pock, and D.~Cremers.
\newblock Large displacement optical flow computation without warping.
\newblock In {\em ICCV}, 2009.

\bibitem{Sun-CVPR-2010}
D.~Sun, S.~Roth, and M.~J. Black.
\newblock Secrets of optical flow estimation and their principles.
\newblock In {\em CVPR}, 2010.

\bibitem{Szeliski-PAMI-2008}
R.~Szeliski, R.~Zabih, D.~Scharstein, O.~Veksler, V.~Kolmogorov, A.~Agarwala,
  M.~Tappen, and C.~Rother.
\newblock A comparative study of energy minimization methods for markov random
  fields with smoothness-based priors.
\newblock {\em TPAMI}, 30(6):1068--1080, June 2008.

\bibitem{tao2012simpleflow}
M.~Tao, J.~Bai, P.~Kohli, and S.~Paris.
\newblock Simpleflow: A non-iterative, sublinear optical flow algorithm.
\newblock In {\em Computer Graphics Forum}, volume~31, pages 345--353. Wiley
  Online Library, 2012.

\bibitem{Timofte-WACV-2015}
R.~Timofte and L.~{Van Gool}.
\newblock Sparseflow: Sparse matching for small to large displacement optical
  flow.
\newblock In {\em WACV}, pages 1100--1106, Jan 2015.

\bibitem{Tola-CVPR-2008}
E.~Tola, V.~Lepetit, and P.~Fua.
\newblock A fast local descriptor for dense matching.
\newblock In {\em CVPR}, 2008.

\bibitem{Weinzaepfel-ICCV-2013}
P.~Weinzaepfel, J.~Revaud, Z.~Harchaoui, and C.~Schmid.
\newblock Deepflow:large displacement optical flow with deep matching.
\newblock In {\em ICCV}, 2013.

\bibitem{Werlberger2009a}
M.~Werlberger, W.~Trobin, T.~Pock, A.~Wedel, D.~Cremers, and H.~Bischof.
\newblock Anisotropic {Huber-L1} optical flow.
\newblock In {\em BMVC}, 2009.

\bibitem{Wills-IJCV-2006}
J.~Wills, S.~Agarwal, and S.~Belongie.
\newblock A feature-based approach for dense segmentation and estimation of
  large disparity motion.
\newblock {\em IJCV}, 2006.

\bibitem{Wulff-CVPR-2015}
J.~Wulff and M.~J. Black.
\newblock Efficient sparse-to-dense optical flow estimation using a learned
  basis and layers.
\newblock In {\em CVPR}, pages 120--130, 2015.

\bibitem{Xu-PAMI-2012}
L.~Xu, J.~Jia, and Y.~Matsushita.
\newblock Motion detail preserving optical flow estimation.
\newblock {\em IEEE Trans. PAMI}, 2012.

\bibitem{yamaguchi2014efficient}
K.~Yamaguchi, D.~McAllester, and R.~Urtasun.
\newblock Efficient joint segmentation, occlusion labeling, stereo and flow
  estimation.
\newblock In {\em ECCV}. 2014.

\bibitem{yang2015dense}
J.~Yang and H.~Li.
\newblock Dense, accurate optical flow estimation with piecewise parametric
  model.
\newblock In {\em CVPR}, pages 1019--1027, 2015.

\bibitem{Zach07aduality}
C.~Zach, T.~Pock, and H.~Bischof.
\newblock A duality based approach for realtime tv-l1 optical flow.
\newblock In {\em In Ann. Symp. German Association Patt. Recogn}, 2007.

\bibitem{Zimmer-IJCV-2011}
H.~Zimmer, A.~Bruhn, and J.~Weickert.
\newblock Optic flow in harmony.
\newblock {\em IJCV}, 2011.

\end{thebibliography}
}

\newpage
\appendix

\startcontents
\printcontents{atoc}{0}{\section*{Supplementary Material}}

\section{Derivation of the \emph{fast inverse search} in \S~2.1 of the paper}
We adopt the terminology of~\cite{Baker-CVPR-2001,Baker-IJCV-2004} and closely follow their derivation.
We consider ${\bf W(x; u)}$ a warp, parametrized by ${\bf u} = (u,v)^T$, on pixel ${\bf x}$ such that ${\bf W(x; u)} = (x + u, y+v)$.
The following derivation holds for other warps as well: See~\cite{Baker-IJCV-2004} for a discussion on the limits of its applicability.
The objective function for the inverse search, eq. (1) in the paper, then becomes
\begin{align}
\sum_{x} \left[ I_{t+1}({\bf W(x; u)}) - T({\bf x}) \right]^2  \; .
\end{align}
The warp parameter ${\bf u}$ is found by iteratively minimizing
\begin{align}
\sum_{x} \left[ I_{t+1}({\bf W(x; u+\Delta {\bf u})}) - T({\bf x}) \right]^2  \label{eq:eqggn_AP}
\end{align}
and updating the warp parameters as ${\bf u} \leftarrow {\bf u}+ {\bf \Delta u}$.

A Gauss-Newton gradient descent minimization is used in the following.
We use a first-order Taylor expansion of eq. \eqref{eq:eqggn_AP} on $I_{t+1}({\bf W(x; u+ {\bf \Delta u})})$:
\begin{align}
\sum_{x} \left[ I_{t+1}({\bf W(x; u)}) + \nabla I_{t+1}\frac{\partial{\bf W}}{\partial{\bf u}} {\bf \Delta u} - T({\bf x}) \right]^2  \label{eq:eqqtaylor_AP}
\end{align}
where $\nabla I_{t+1}$ is the image gradient at ${\bf W(x; u)}$. $\frac{\partial{\bf W}}{\partial{\bf u}}$ denotes the Jacobian of the warp. 
Writing the partial derivatives in $\frac{\partial{\bf W}}{\partial{\bf u}}$ with respect to a column vector as row vectors, this simply becomes the 2 $\times$ 2 identity matrix for the case of optical flow.

There is a closed-form solution for parameter update ${\bf \Delta u}$ using a least-squares formulation. Setting to zero the partial derivative of eq. \eqref{eq:eqqtaylor_AP} with respect to ${\bf \Delta u}$
\begin{align}
\sum_{x} S^{T} \cdot \left[ I_{t+1}({\bf W(x; u)}) + \nabla I_{t+1}\frac{\partial{\bf W}}{\partial{\bf u}} {\bf \Delta u} - T({\bf x}) \right]  \stackrel{!}{=} 0 \, ,\label{eq:lsqtaylor_AP}
\end{align}
where $S = \left[ \nabla I_{t+1}\frac{\partial{\bf W}}{\partial{\bf u}} \right]$, lets us solve for ${\bf \Delta u}$ as
\begin{align}
{\bf \Delta u} = H^{-1} \sum_{x} S^T \cdot \left[ T({\bf x}) - I_{t+1}({\bf W(x; u)}) \right]  \label{eq:solvereq_AP}
\end{align}
where $H =  \sum_{x} S^T S$ is the $n \times n$ approximation to the Hessian matrix.

Since $S$ depends on the image gradient of image $I_{t+1}$ at the displacement ${\bf u}$, $S$ and the Hessian $H$ have to be re-evaluated at each iteration. 
In order to avoid this costly re-evaluation it was proposed~\cite{Baker-CVPR-2001,Baker-IJCV-2004,hager1998efficient} to invert to roles of image and template.
As a result, the objective function becomes
\begin{align}
\sum_{x} \left[T({\bf W(x; \Delta {\bf u})}) - I_{t+1}({\bf W(x; u)}) \right]^2  \;  .  \label{eq:invobjfunc_AP}
\end{align}

\noindent The new warp is now composed using the \emph{inverse} updated warp ${\bf W(x;  {\bf u})} \leftarrow {\bf W(x;  {\bf u})} \circ {\bf W(x; \Delta {\bf u})}^{-1}$.
In the case of optical flow, this simply becomes: ${\bf u} \leftarrow {\bf u} - {\bf \Delta u}$.

The first order Taylor expansion of \eqref{eq:invobjfunc_AP} gives
\begin{align}
\sum_{x} \left[ T({\bf W(x; {\bf 0})})  + \nabla T\frac{\partial{\bf W}}{\partial{\bf u}} {\bf \Delta u} - I_{t+1}({\bf W(x; u)} \right]^2     \label{eq:invtaylor_AP}
\end{align}
where $W(x; {\bf 0})$ is the identity warp. Analogously to eq. \eqref{eq:solvereq_AP} we can solve \eqref{eq:invtaylor_AP} as a least squares problem.

\begin{align}
{\bf \Delta u} = H'^{-1} \sum_{x} S'^{T} \cdot \left[ I_{t+1}({\bf W(x; u)}) - T({\bf x}) \right]  \label{eq:solvereqinv_AP} 
\end{align}
where $S' = \left[ \nabla T\frac{\partial{\bf W}}{\partial{\bf u}} \right]$ and $H' =  \sum_{x} S'^T S'$. The Jacobian $\frac{\partial{\bf W}}{\partial{\bf u}}$ is evaluated at (${\bf x; 0}$).
Since neither $S'$, nor $H'$ depend in ${\bf u}$, they can be pre-computed. This also means that image gradients do not have to be updated in each iteration. 
In each iteration we only need, firstly, to compute the sum in \eqref{eq:solvereqinv_AP}, which includes the image difference, multiplication with the image gradients, and summation, secondly, to solve the linear system for ${\bf \Delta u}$, and, thirdly, update the warp parameter ${\bf u}$.

\section{Extensions: Color, forward-backward consistency, robust error norms}

We examined the benefit of, firstly, using RGB color images instead of intensity images, secondly, enforcing forward-backward consistency of optical flow, and, thirdly, using robust error norms.
\\

{\bf (i) Using RGB color images.} 
Instead of using a patch matching error over intensity images, we can use a multi-channel image for the dense inverse search and the variational refinement.
The objective function for dense the inverse search, eq. (1) in the paper, becomes
\begin{align}
\sum_{c \in [R,G,B]} \sum_{x} \left[ I_{t+1}^c({\bf x}+{\bf u}) - T^c({\bf x}) \right]^2 \; . \label{eq:minnorm_AP}
\end{align}
where $c$ iterates over all color channels.
Similarly, we can extend the variational refinement to operate on all color channels jointly, as detailed in~\cite{Weinzaepfel-ICCV-2013}.
\\

{\bf (ii) Enforcing forward-backward consistency.} 
In order to enforce forward-backward consistency, we run our algorithm in parallel from both directions: $I_t \rightarrow I_{t+1}$ and $I_{t+1} \rightarrow I_{t}$.
With the exception of the densification (step 4), all steps of the algorithm run independently for forward and backward flow computation.
In step 4 we merge both directions and create a dense forward flow field ${\bf U}^f_s$ in each pixel ${\bf x}$ by applying weighted averaging to displacement estimates of all patches \emph{from the forward and backward inverse search} overlapping at location ${\bf x}$ in the reference image $I_t$:

\begin{align}
{\bf U}^f_s({\bf x}) = \frac{1}{Z} \; \biggl[  &\sum_i^{N^f_s} \frac{\lambda^f_{i,{\bf x}}}{\max(1,\lVert d^f_i({\bf x}) \rVert_2)} \cdot {\bf u}^f_i - \\
&\sum_j^{N^b_s} \frac{\lambda^b_{j,{\bf x}}}{\max(1,\lVert d^b_j({\bf x}) \rVert_2)}  \cdot {\bf u}^b_j \biggl] \label{eq:averagingeq_AP}
\end{align}

where the indicator $\lambda^f_{i,{\bf x}}=1$ iff patch $i$ for the \emph{forward} displacement estimate overlaps with location ${\bf x}$ in the reference image, 
$\lambda^b_{j,{\bf x}}=1$ iff patch $j$ for the \emph{backward} displacement estimate overlaps with location ${\bf x}$ in the reference image after the displacement ${\bf u}^f_j$ was applied to it, 
$d^f_i({\bf x}) = I_{t+1}({\bf x}+{\bf u}^f_i) - T({\bf x})$ and $d^b_j({\bf x}) = I_{t}({\bf x}) - T({\bf x}-{\bf u}^b_j)$ denote the forward and backward warp intensity differences between template patches and warped images, 
${\bf u}^f_i$ and ${\bf u}^b_j$ denote the estimated forward and backward displacements of patches, and normalization 
\
\begin{align}
Z =  &\sum_i^{N^f_s} \lambda^f_{i,{\bf x}}/\max(1,\lVert d^f_i({\bf x}) \rVert_2) +\\ &\sum_j^{N^b_s} \lambda^b_{j,{\bf x}}/\max(1,\lVert d^b_j({\bf x}) \rVert_2).
\end{align}

Since displacement estimate ${\bf u}^b_j$ is generally not integer, we employ bilinear interpolation for the second term in equation \eqref{eq:averagingeq_AP}.

The densification for the dense backward flow field ${\bf U}^b_s$ is computed analogously. 
After the densification step on each scale, the variational refinement is again performed independently for each direction.
\\

{\bf (iii) Robust error norms.} 
Equation (1) in the paper minimizes an L2 norm which is strongly affected by outliers.
Since the L1 and the Huber norm are known to be more robust towards outliers, we examined their effect on our algorithm.
However, the objective function (1) cannot easily be changed to directly minimize a different error norm.
However, in each iteration we can transform the error $\varepsilon = I_{t+1}({\bf x}+{\bf u}) - T({\bf x})$ for each pixel, such that, implicitly after squaring the transformed error, eq. (1) minimizes a different norm.
We transform the error $\varepsilon$ on each pixel at location ${\bf x}$ as follows:
\begin{itemize}
 \item L1-Norm: $$\varepsilon \leftarrow \text{sign}(\varepsilon) \cdot \sqrt{\lvert \varepsilon \lvert}  $$
 \item Huber-Norm: $$\varepsilon \leftarrow \left\{
  \begin{array}{@{}ll@{}}
    \varepsilon  & \quad, \text{if}\ \varepsilon < b \\
    \text{sign}(\varepsilon) \cdot \sqrt{2b\lvert \varepsilon \lvert - b^2} & \quad,  \text{otherwise}
  \end{array}\right. $$
\end{itemize}
After the transformed error $\varepsilon$ is squared in the objection function the corresponding L1 or Huber norm is minimized in each iteration.

\begin{figure}[!ht]
        \centering
        \includegraphics[width=0.45\textwidth]{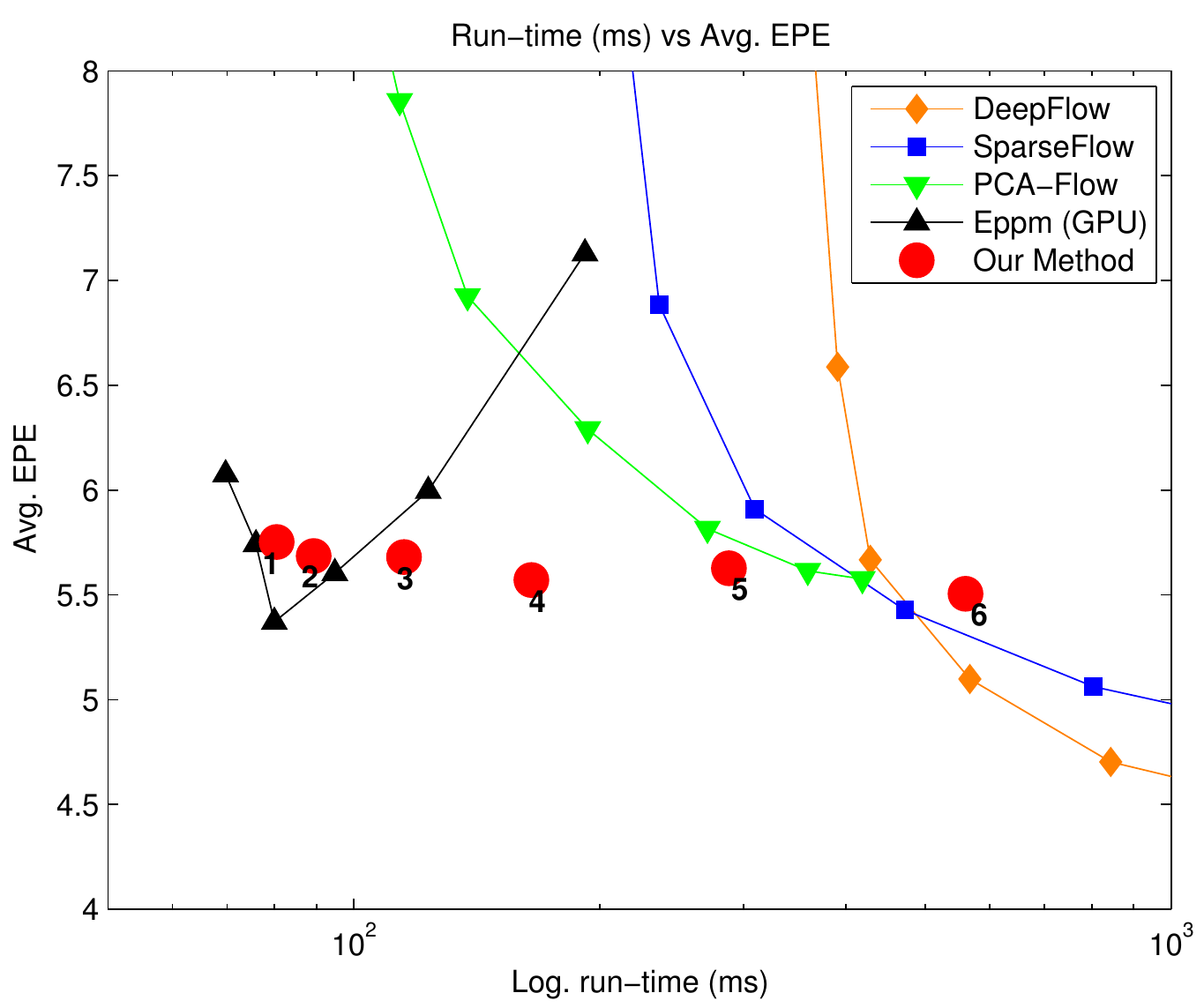}
        \caption{Evaluation of extensions. (1) Baseline, (2) baseline with color images, (3) baseline enforcing forward-backward consistency, (4) baseline using the L1 norm as objective function, (5) baseline using the Huber norm as objective function, (6) baseline using color images, enforcing forward-backward consistency and using the Huber norm as objective function.}\label{fig:extensions_AP}
\end{figure}

We plotted the result for all three extensions in Fig.~\ref{fig:extensions_AP}, where we compare all extensions on the Sintel training benchmark against our method without these extensions. As baseline we start from operating point {\bf (3)}, as described in the paper, and evaluate all extensions separately. The six numbered operating points correspond to: (1) Baseline, (2) baseline with color images, (3) baseline enforcing forward-backward consistency, (4) baseline using the L1 norm as objective function, (5) baseline using the Huber norm as objective function, (6) baseline using color images, enforcing forward-backward consistency and using the Huber norm as objective function.

We conclude that while all extensions decrease the resulting optical flow error, overall, the obtained error reduction remains too low to justify inclusion of these extensions in the method. 

\section{Implementation details, parallelization and memory consumption}
Our method is implemented in C++ and all run-times where computed on a Intel Core i7 3770K CPU at 3.50GHz. In this chapter we provide more engineering details and exact timings of the most important components, details on possible parallelization using OpenMP and overall memory consumption.

\subsection*{Implementation details}
For our method no special data structures are needed. What enables the high speed is a combination of a very fast flow initialization of DIS with a slow variational refinement per scale. Initialization and variational refinement constitute 32 \% and 57 \%, respectively, of the total run-time on each scale.
Care was taken to allocate all memory at once, use SSE instructions and inplace-operations whenever possible, and terminate iterations early for small residuals. 
Bottlenecks for further speed-ups are repeated (bilinearly interpolated) patch extraction (step 3 in {\bf Alg. 1} of the paper) and pixel-wise refinements (step 5).

In Table~\ref{tab:timing_AP} we show a direct run-time comparison on Sintel- and VGA-resolution images. The run-time scales linearly with the image area.
\begin{table}[h!]
\centering
\begin{tabular}{l|llll}
 &  DIS {\bf (1)} &  {\bf DIS (2)} &  DIS {\bf (3)} &  DIS {\bf (4)} \\
\hline
Sintel & 1.65 / 606 & 3.32 / {\bf 301} & 97.8 / 10.2 & 1925 / 0.52 \\
VGA & 0.93 / 1075 & 2.35 / {\bf 426} & 70.3 / 14.2 & 1280 / 0.78 \\
\end{tabular}
\caption{Sintel (1024$\times$436), VGA (640$\times$480) run-times in (ms/Hz).}
\label{tab:timing_AP}
\end{table}

We will break down the run-time of {\bf 3.32 ms} for {\bf DIS (2)} on Sintel-resolution images, running over 3 scales, for all components in table~\ref{tab:time1_AP}.

\begin{table}[!h]
\centering
\begin{tabular}{l|cc}
 Total run-time (3.32 ms) & ms &  \% of total run-time\\
\hline
Memory allocation & 0.17 & 5.27 \\
Processing scale $\theta_s=5$ & 0.23 & 6.91 \\
Processing scale $\theta_s=4$ & 0.65 & 19.6 \\
Processing scale $\theta_s=3$ & 2.26 & 68.1 \\
\end{tabular}
\caption{Break down of {\bf DIS (2)} total run-time on Sintel images}
\label{tab:time1_AP}
\end{table}

Run-time grows approximately by a factor of 4 for each finer scale, and is spend similarly on each scale. Representatively for the last scale ($\theta_s=3$), corresponding to a downscaling factor of 8 and 448 uniformly distributed 8x8 patches, the time of 2.26 ms spend on this layer breaks down following {\bf Alg. 1} in table~\ref{tab:time2_AP}.

\begin{table}[!h]
\centering
\begin{tabular}{l|cc}
 Run-time on scale $\theta_s=3$ (2.26 ms) & ms &  \% of run-time\\
\hline
Patch initialization, steps (1./2.) & 0.09 & 4.1 \\
Inverse Search, step (3.) & 0.74 & 32.8 \\
Densification, (4.) & 0.14 & 5.9 \\
Variational refinement, (5.) & 1.29 & 57.1 \\
\end{tabular}
\caption{Break down of {\bf DIS (2)} run-time on one scale}
\label{tab:time2_AP}
\end{table}

Step (3.) and (5.) are most time-consuming. Step (3.) breaks down in 1.66 $\mu$s for each of 448 patches. This breaks down into 0.08 $\mu$s for initialization (gradient/intensity extraction in reference image, computation of Hessian) and 0.13 $\mu$s for each of 12 optimization iterations (template intensity extraction, parameter update). Step (5.) breaks down for each of $\theta_{vo}=4$ iterations into 0.12 ms for computation of the data and smoothness terms and 0.2 ms for solving of the linear systems and updating the flow field per pixel.

\subsection*{Parallelization}
We examined the potential speed-up by parallelization using OpenMP on 4 cores.
We parallelized step 3 and 5 of our implementation, to operate independently on each patch and pixel, respectively.
The run-times for all for operating points are tabulated in Table~\ref{tab:parallel_AP}.
Since thread creation and management leads to an overhead of a few milliseconds, the speed-up of 4 cores only becomes significant for longer run-times.
For longer sequences, where threads are created only once, this overhead would be negligible.
Since each tread executes the same instructions and there is no need to communicate, a massive parallelization on a GPU will potentially yield an even larger speed-up.

\begin{table}
\centering
\begin{tabular}{c|cccc}
 DIS operating point & {\bf (1.)} &  {\bf (2.)} &  {\bf (3.)} &  {\bf (4.)}\\
\hline
Speed-up (x) & 1.29 & {\bf 1.75} & 2.15 & 3.70 \\
\end{tabular}
\caption{Speed-up factor from parallelization (OpenMP, 4 cores) for all four operating points chosen in the paper}
\label{tab:parallel_AP}
\end{table}

\subsection*{Memory Consumption}
\begin{table*}
\centering
\begin{tabular}{c|ccccccc}
  & DIS {\bf (1.)} &  {\bf DIS (2.)} &  DIS {\bf (3.)} &  DIS {\bf (4.)} & Farneback & PCAFlow & DeepFlow\\
\hline
Peak Mem. (MB) & 35.52 & {\bf 35.56} & 100.1 & 311.9 &  43.31 & 1216 & 6851\\
Speed (ms) & 1.65 & {\bf 3.32} & 97.8 & 1925 & 26.27 & 419.1 & 55752\\
Speed (Hz) & 606 & {\bf 301} & 10.2 & 0.52 & 38.06 & 2.38 & 0.018\\
\end{tabular}
\caption{Peak memory consumption (MB) on Sintel-resolution images for all four operating points chosen in the paper and for the baselines: Farneback method, PCAFlow and DeepFlow at full resolutions.}
\label{tab:memory_AP}
\end{table*}
We also examined the peak memory consumption of our algorithm on images from the Sintel benchmark. 
We tabulated the result in Table~\ref{tab:memory_AP}.
Roughly 15~MB are used by image data, and the rest by all patches and associated data, which is pre-allocated during initialization.
If parallelization is not used, and memory is a bottleneck, the memory footprint can be reduced by not pre-allocating all patch memory.

\section{Plots for more operating points on the KITTI and Sintel benchmarks}
\begin{figure*}[!ht]
        \centering
        \begin{tabular}{c}
        \includegraphics[width=0.38\textwidth]{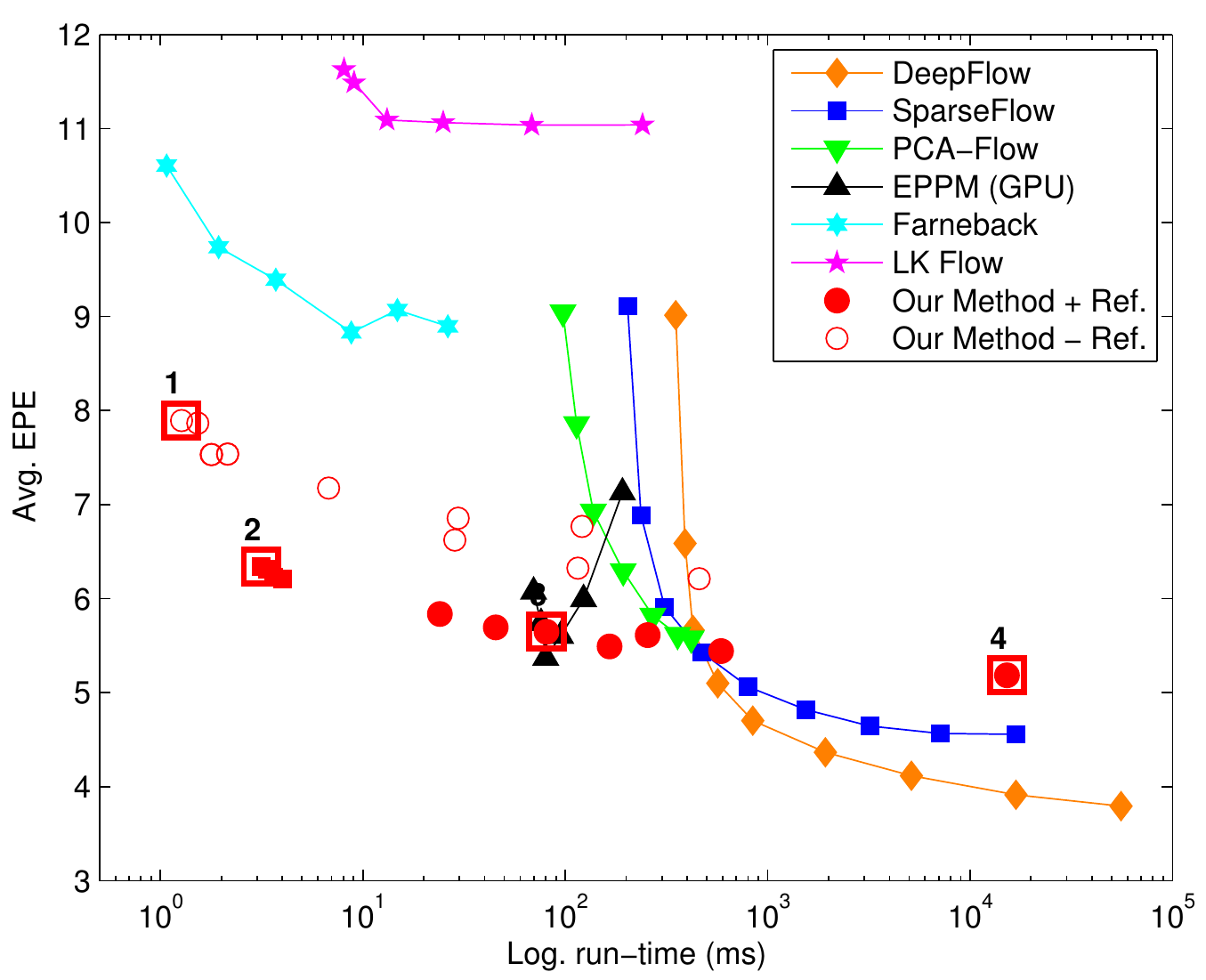}
        \includegraphics[width=0.38\textwidth]{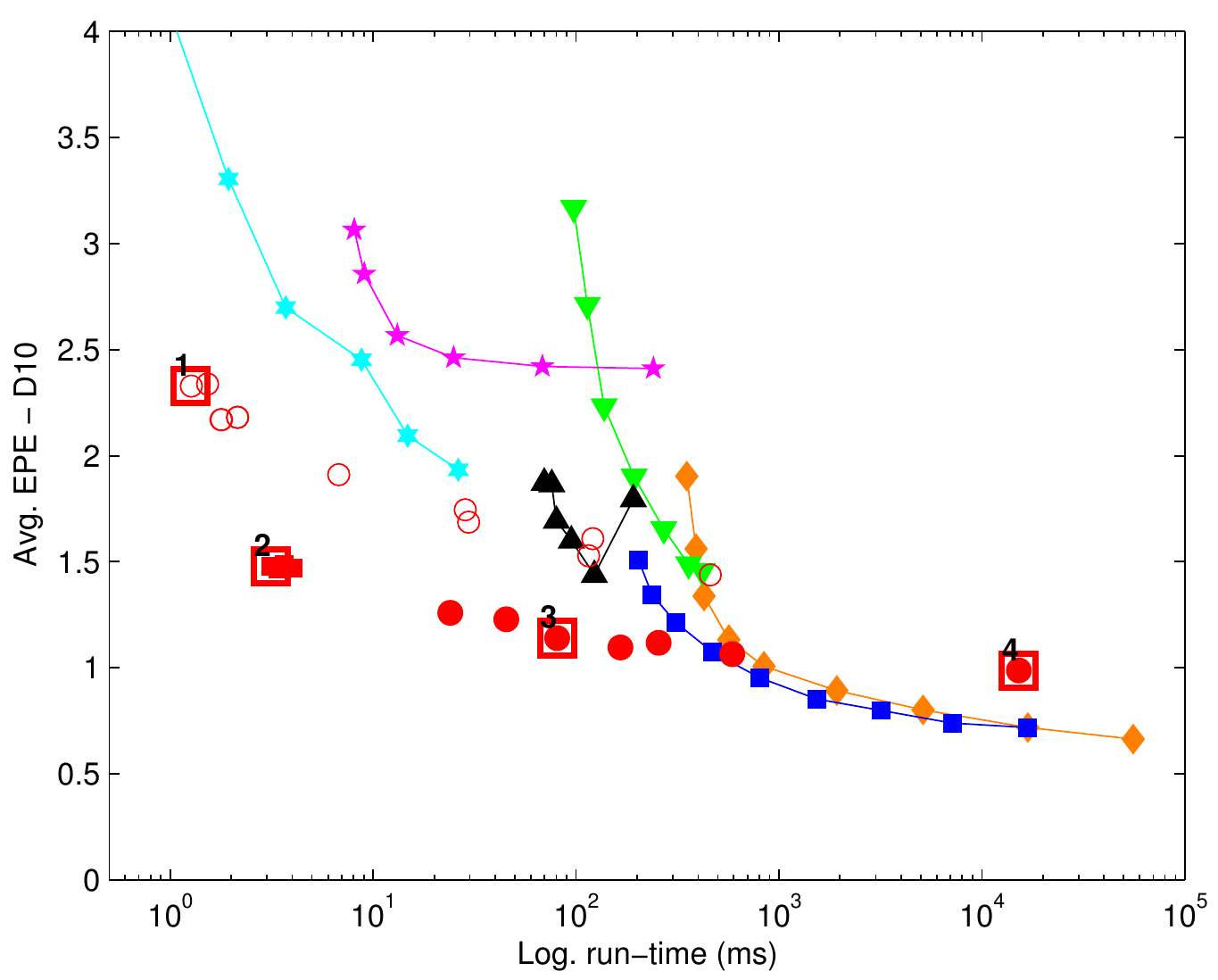} \\
        \includegraphics[width=0.38\textwidth]{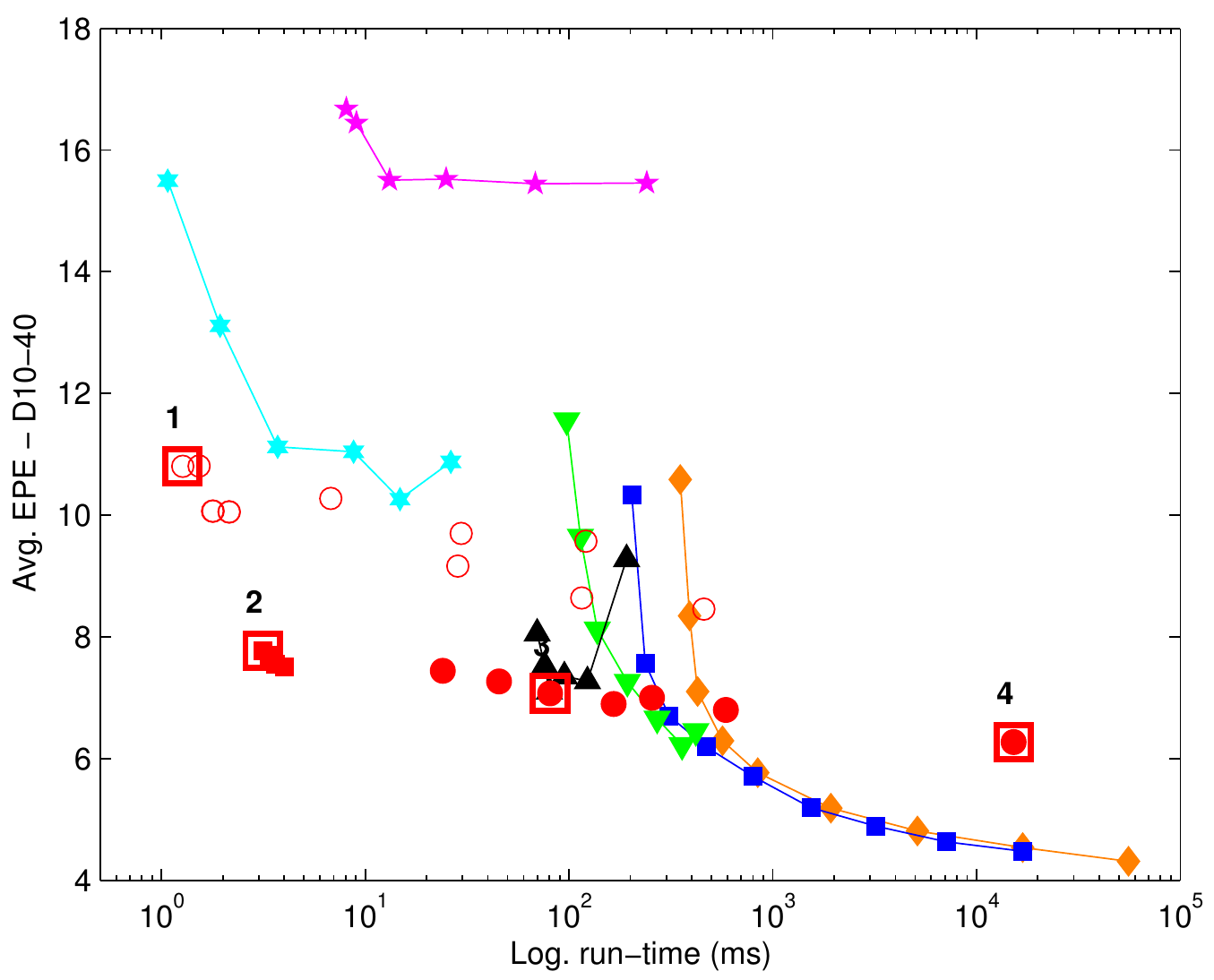}
        \includegraphics[width=0.38\textwidth]{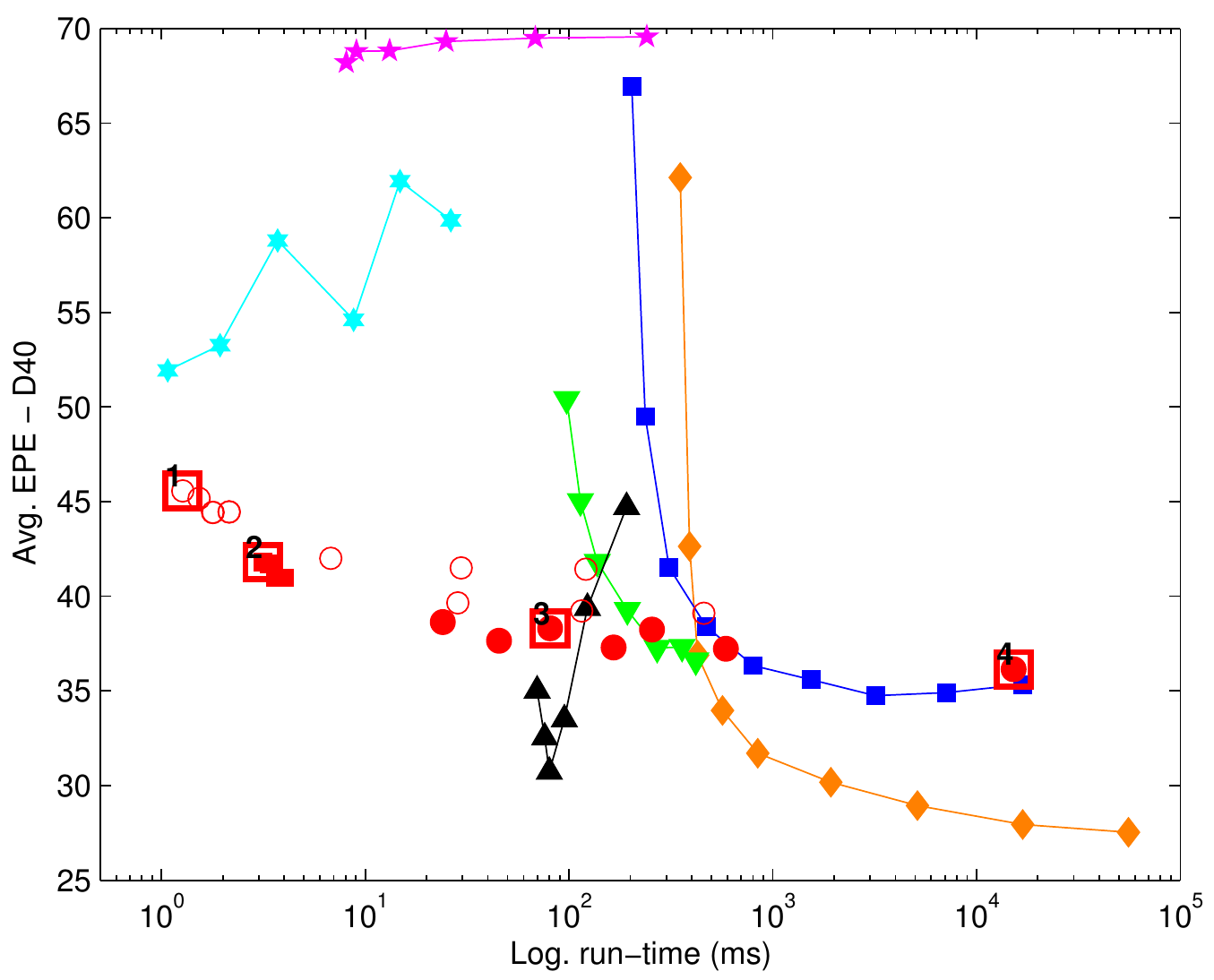} \\[2pt]
        \hline \\[2pt]
%
        \includegraphics[width=0.38\textwidth]{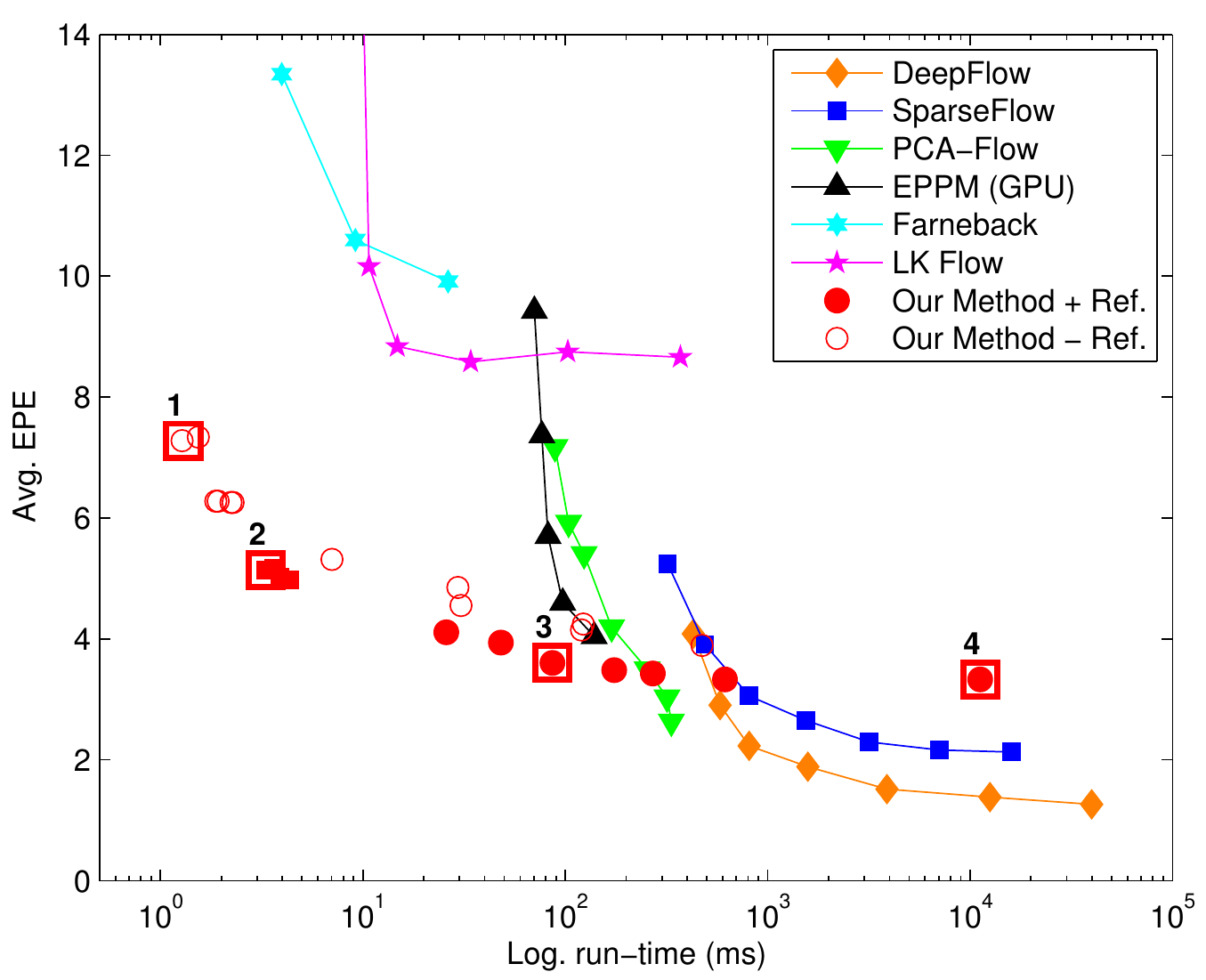}
        \includegraphics[width=0.38\textwidth]{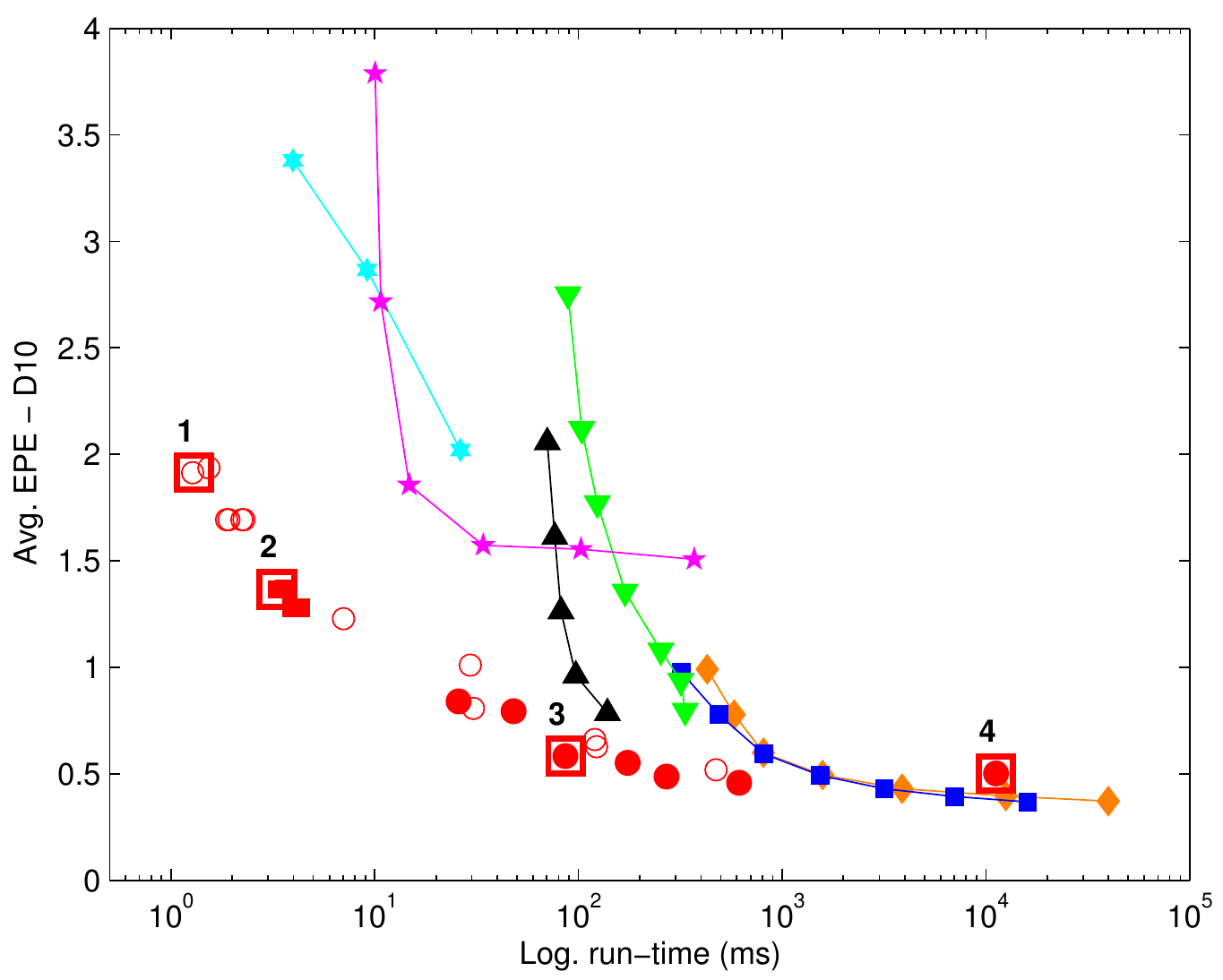} \\
        \includegraphics[width=0.38\textwidth]{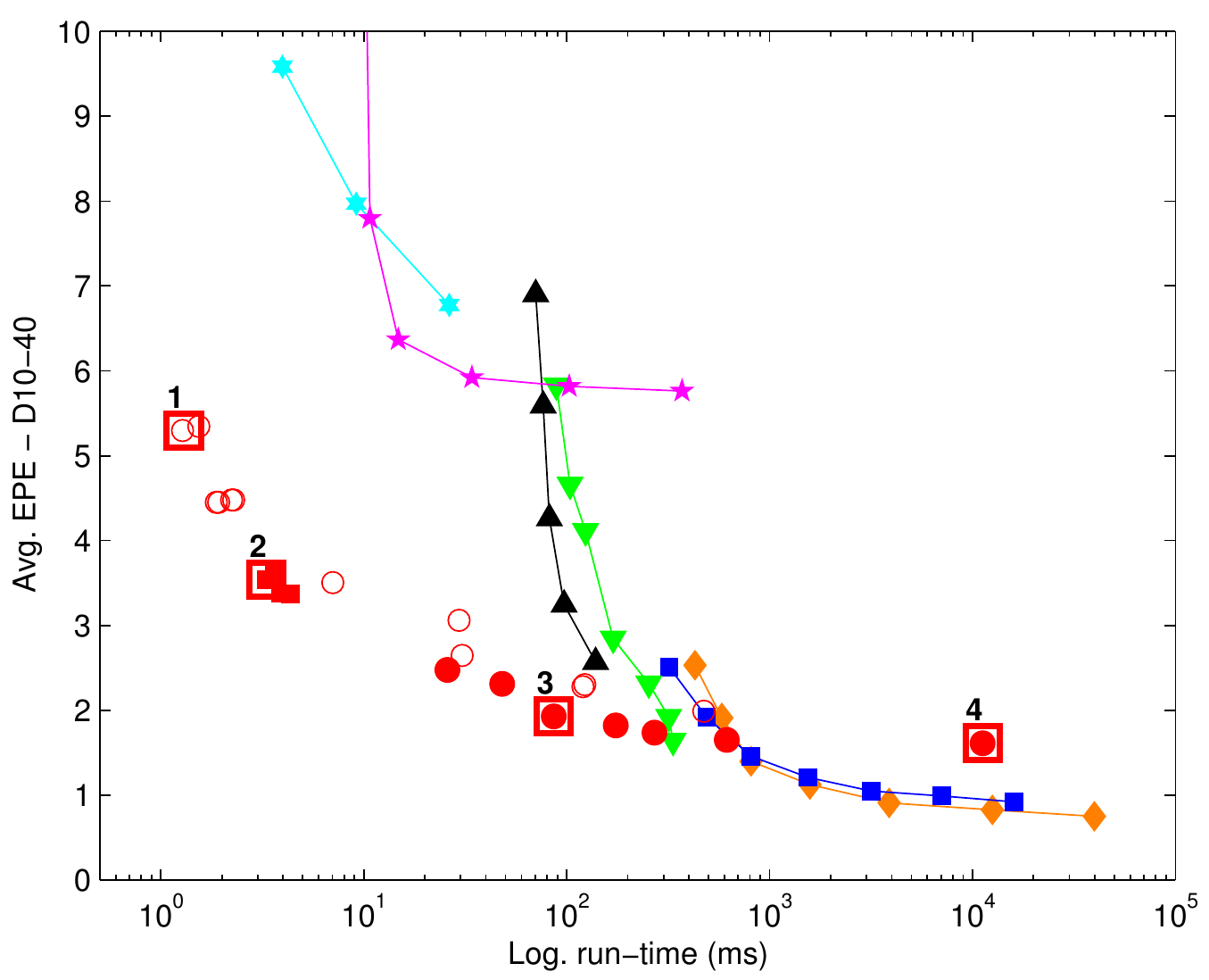}
        \includegraphics[width=0.38\textwidth]{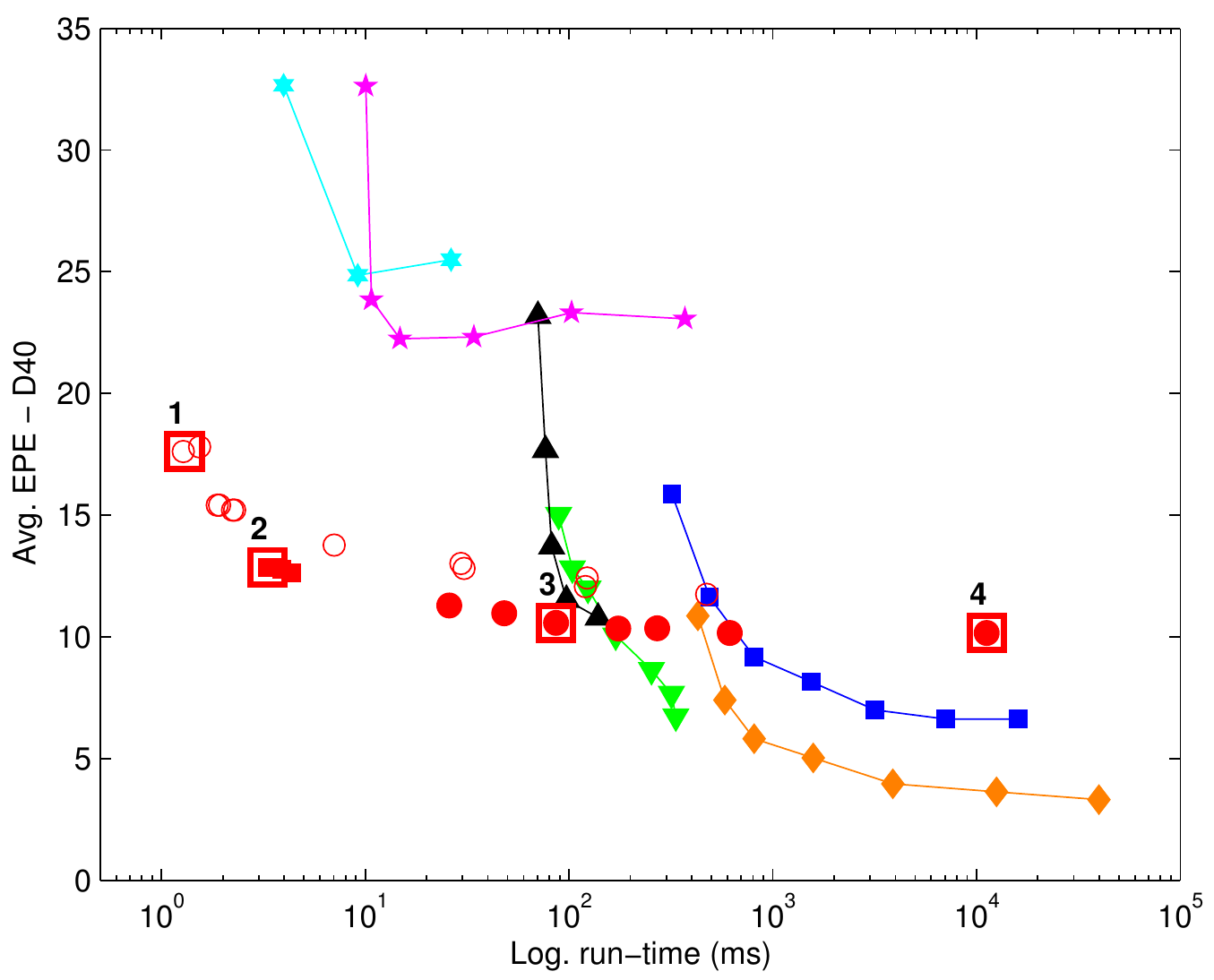}
        \end{tabular}
        \caption{Sintel (top 2 $\times$ 2 block) and KITTI (bottom 2 $\times$ 2 block)  flow result. In each block of 2 $\times$ 2, top to bottom, left to right: Average end-point error over image full domain, average end-point error  over pixels with ground truth displacement over $< 10$ pixels, average end-point error  over pixels with ground truth displacement between 10 and 40 pixels, average end-point error  over pixels with ground truth displacement over $< 40$ pixels. The four operating points used in the paper are marked with {\bf 1}-{\bf 4} in the plots.}\label{fig:sintelkittiopp_AP}
\end{figure*}

We plot the optical flow result on the Sintel and KITTI training benchmarks in Fig. \ref{fig:sintelkittiopp_AP}. 
Besides the four operating points chosen in the paper, we plot a variety of operating points for different parameter settings with and without variational refinement.
In addition to the observations detailed in the paper, we note that the variational refinement brings the strongest advantage in the small displacement range, whereas large displacement errors cannot be recovered easily by this refinement.

\section{Plots for experiments \S~3.3 and \S~3.4 including preprocessing time}
In our evaluation we excluded preprocessing (disk access, image rescaling, gradient computation) for all methods, by carefully timing time spend on these tasks within their provided code. For EPPM, where only binaries where provided, we subtracted the average preprocessing overhead of our method, since we were unable to measure overhead-time directly within their code.

The exclusion of the preprocessing time enables us to compare the actual time spend on flow computation, without evaluating the constant preprocessing overheads. This is particularly important when this preprocessing is shared between tasks or even unnecessary in robotics streaming applications, or when it heavily dominates the run-time and would therefore clutter the analysis.

\begin{figure*}[!ht]
        \centering
        \begin{tabular}{c}
        \includegraphics[width=0.38\textwidth]{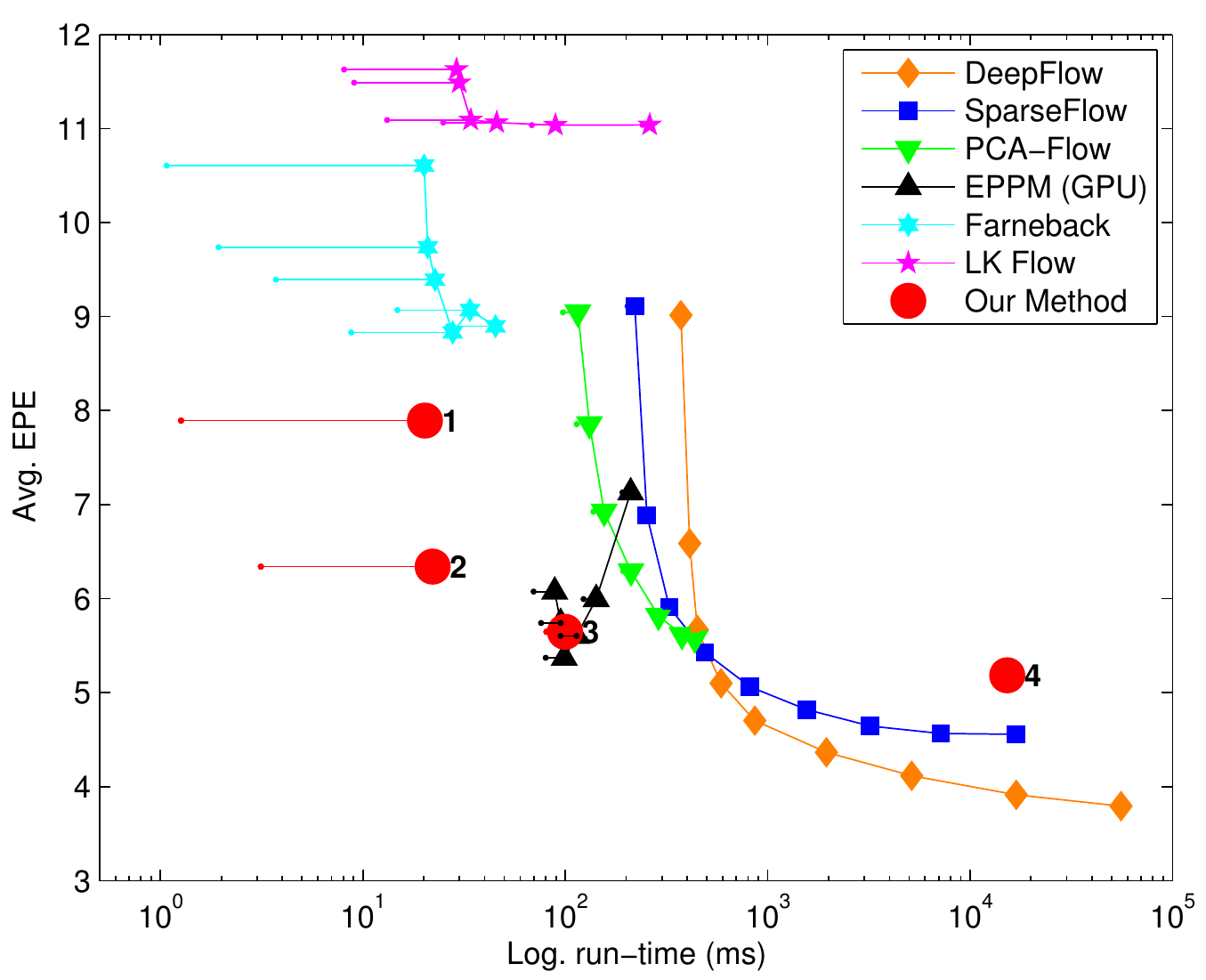}
        \includegraphics[width=0.38\textwidth]{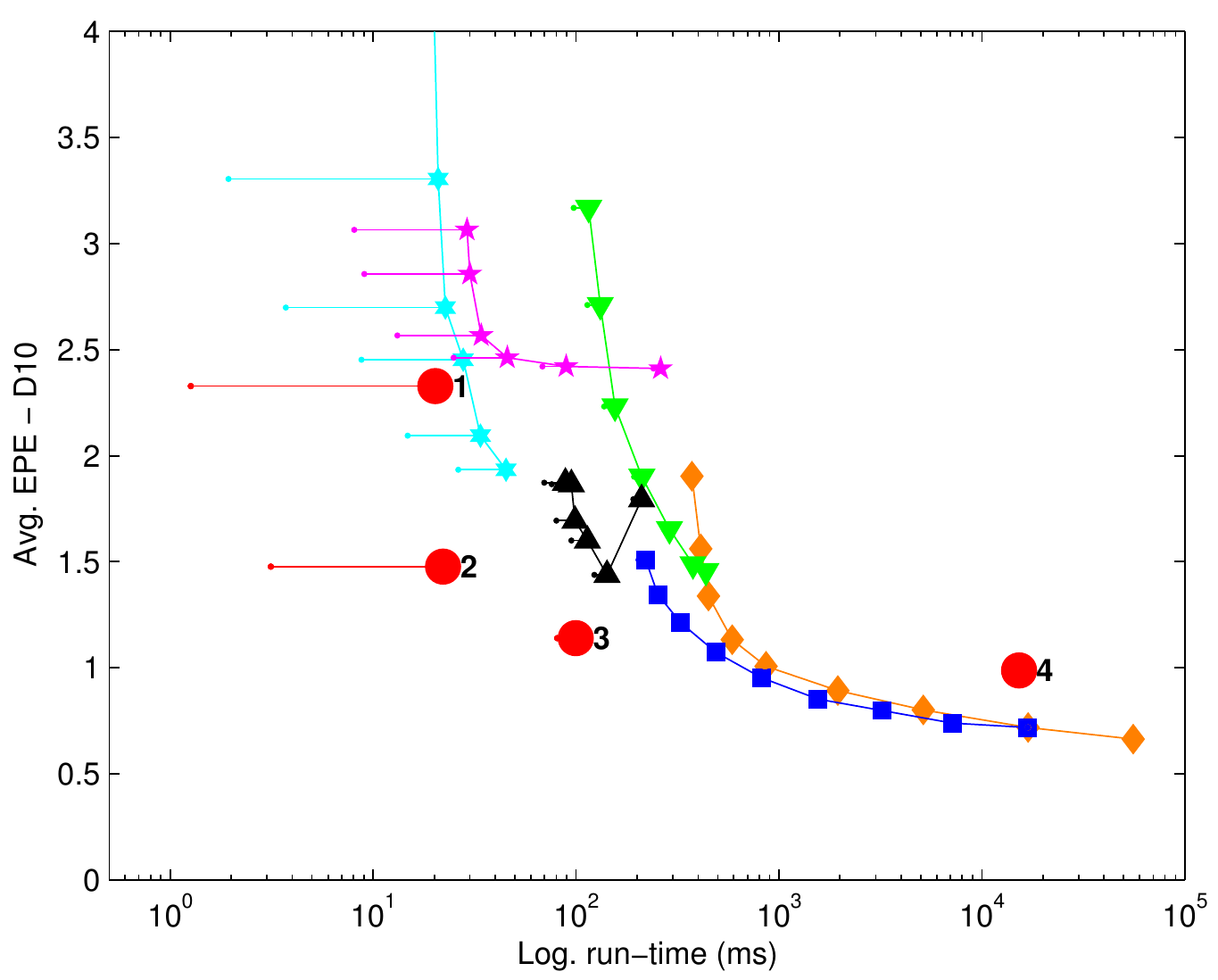} \\
        \includegraphics[width=0.38\textwidth]{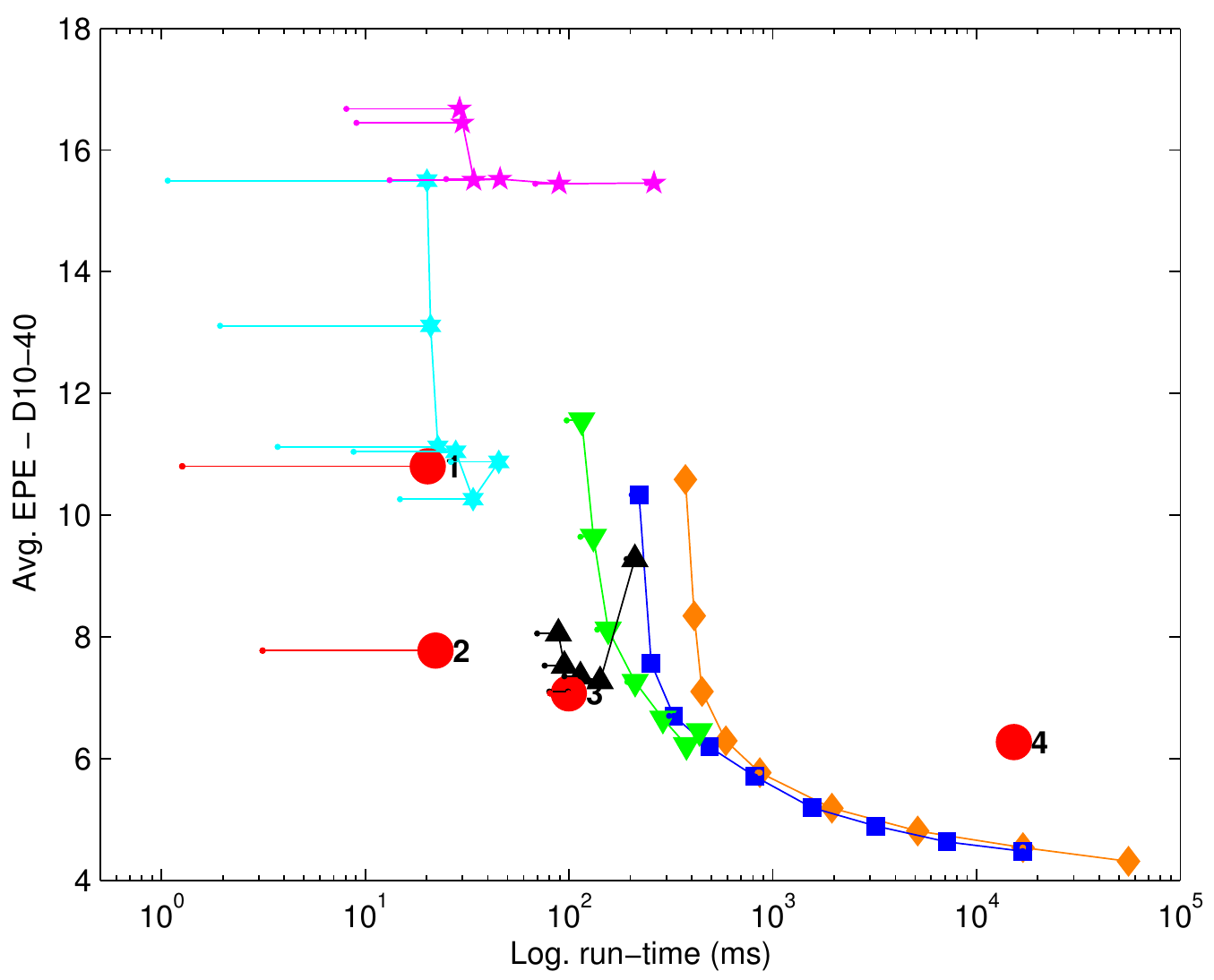}
        \includegraphics[width=0.38\textwidth]{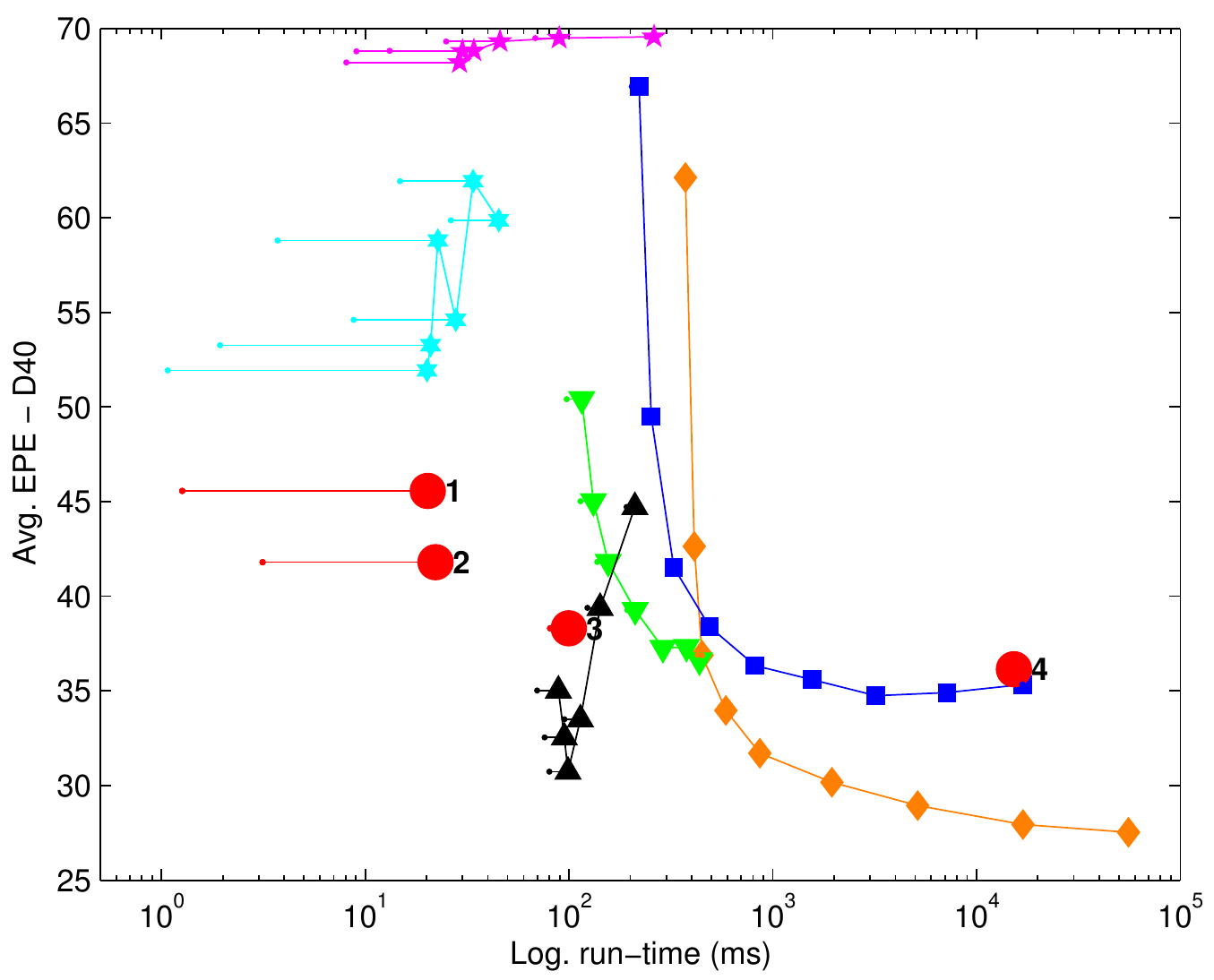} \\[2pt]
        \hline \\[2pt]
%
        \includegraphics[width=0.38\textwidth]{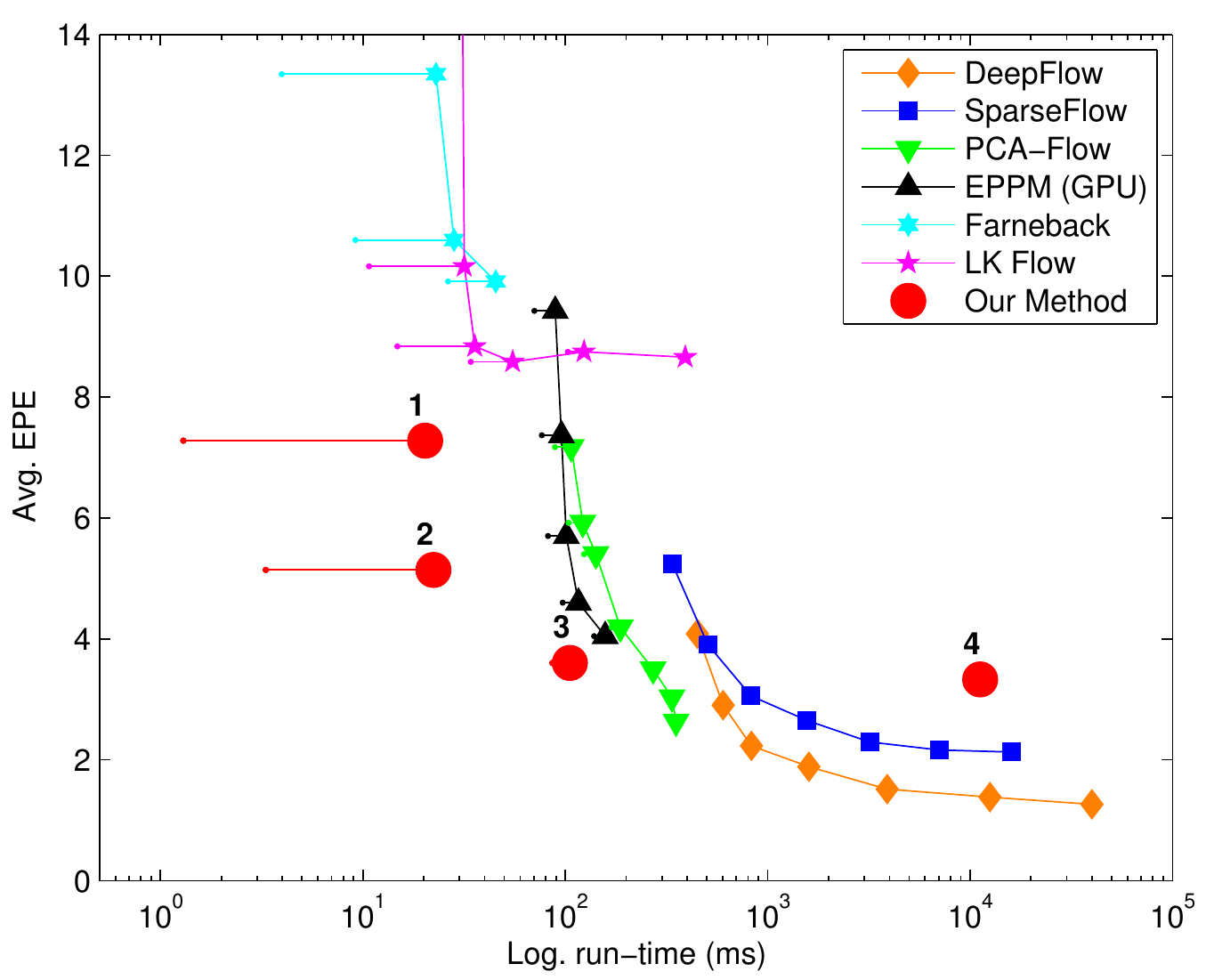}
        \includegraphics[width=0.38\textwidth]{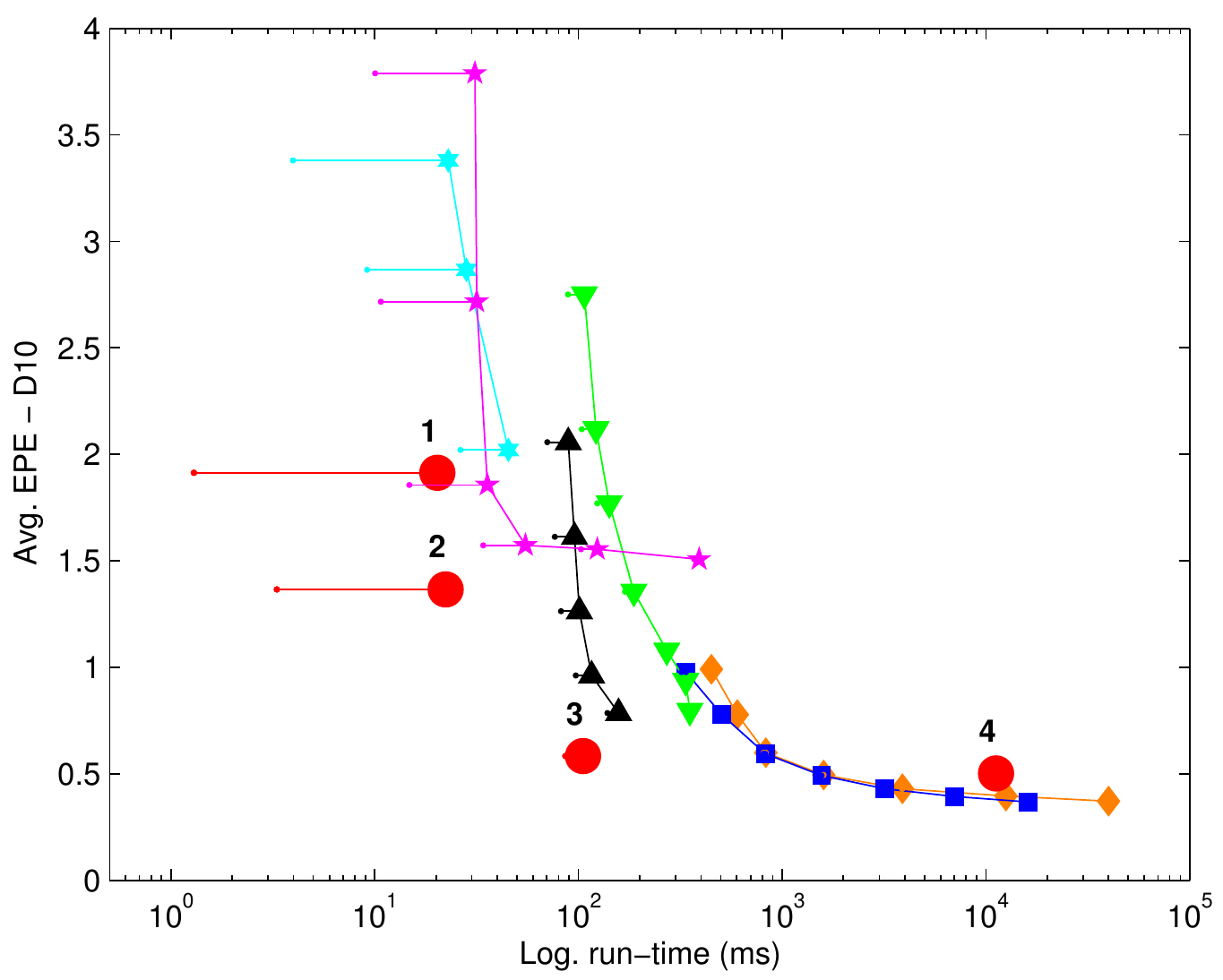} \\
        \includegraphics[width=0.38\textwidth]{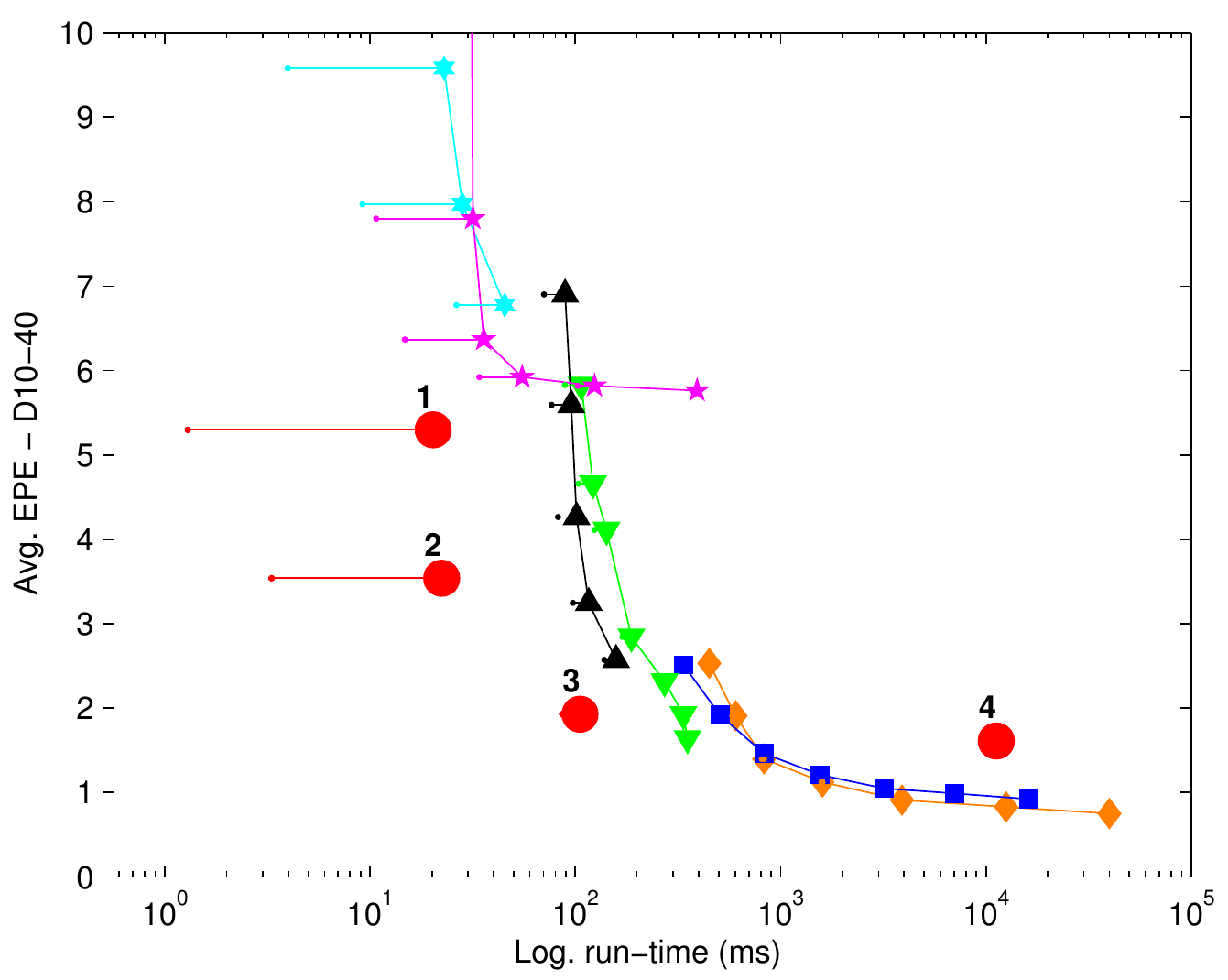}
        \includegraphics[width=0.38\textwidth]{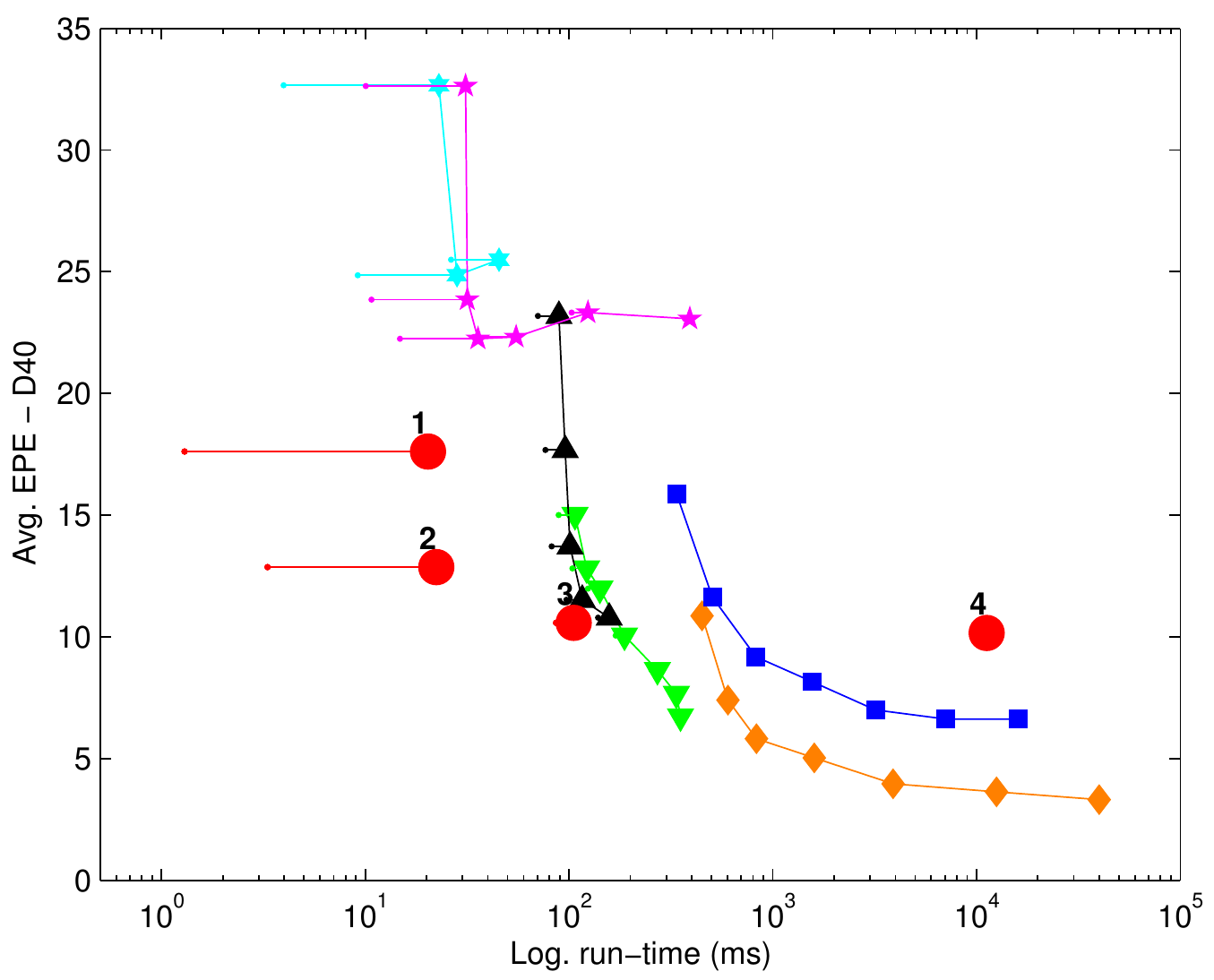}
        \end{tabular}
        \caption{Sintel (top 2 $\times$ 2 block) and KITTI (bottom 2 $\times$ 2 block)  flow result. Same plots as Figure 4 and 5 in the paper, but including preprocessing time for all methods. The horizontal bars for each method indicate portion of total run-time spend in preprocessing (disk access, image rescaling, gradient computation).}\label{fig:sintelkittiopp_IP_AP}
\end{figure*}

This is illustrated in Fig. \ref{fig:sintelkittiopp_IP_AP}, which shows the end-point error on the training sets of Sintel and Kitti (Figures 4 and 5 in the paper in \S~3.3, 3.4) versus total run-time of each method \emph{including} preprocessing. It is easy to observe, that in the range of several tens or hundreds of Hertz preprocessing dominates and makes an analysis diffult.

\section{More exemplary results for KITTI and Sintel (training)}
\begin{figure*} [!ht]
\centering\setlength{\tabcolsep}{0.1pt}\renewcommand{\arraystretch}{0} 
\begin{tabular}{ccccc}
 {\bf 600Hz} & {\bf 300Hz} & {\bf 10Hz} & {\bf 0.5Hz}& {\bf Ground Truth}\\
\includegraphics[width=0.195\textwidth]{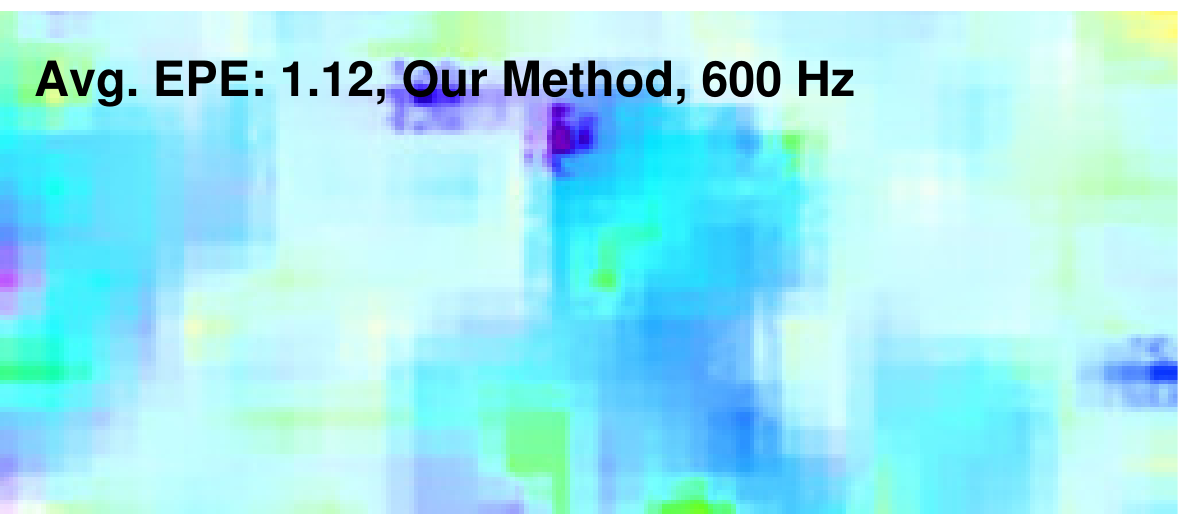}&
\includegraphics[width=0.195\textwidth]{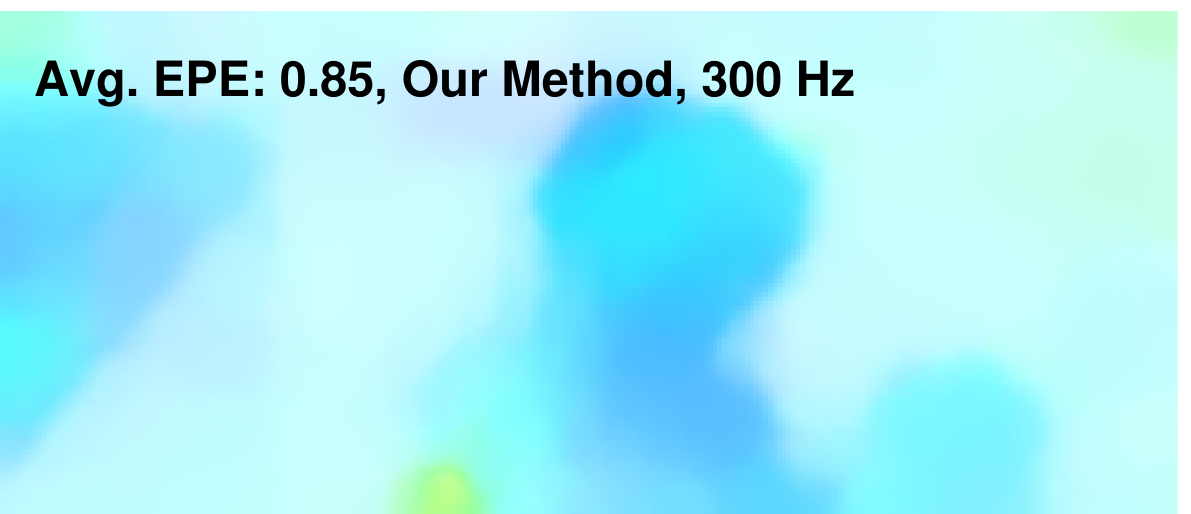}&
\includegraphics[width=0.195\textwidth]{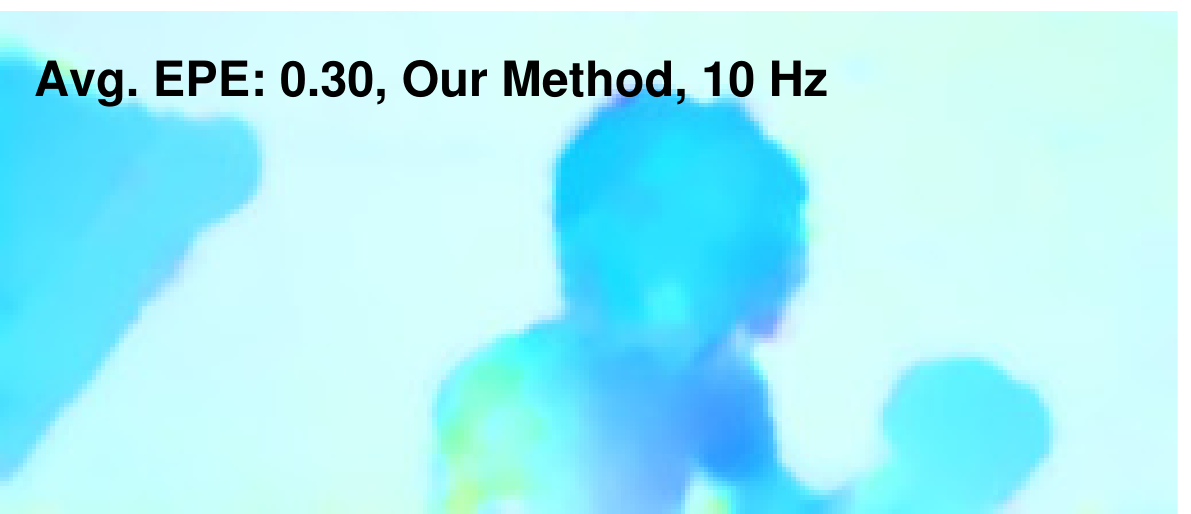}&
\includegraphics[width=0.195\textwidth]{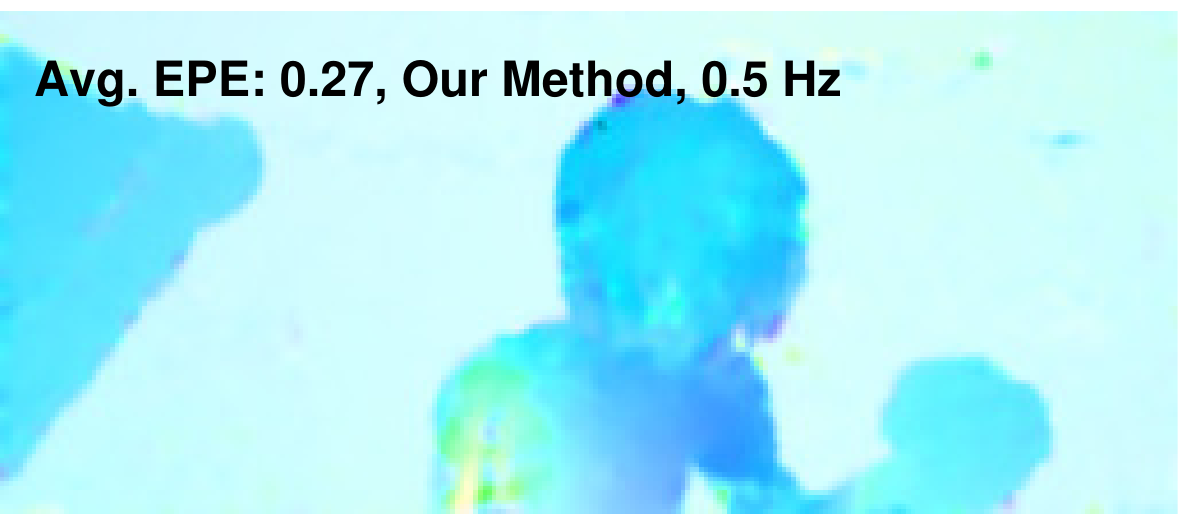}&
\includegraphics[width=0.195\textwidth]{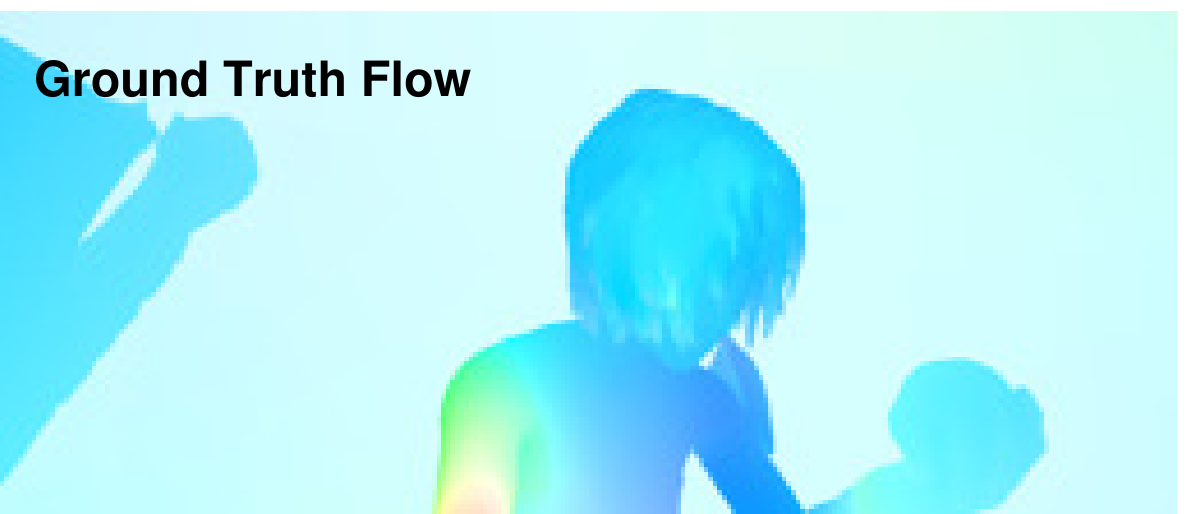}\\
\includegraphics[width=0.195\textwidth]{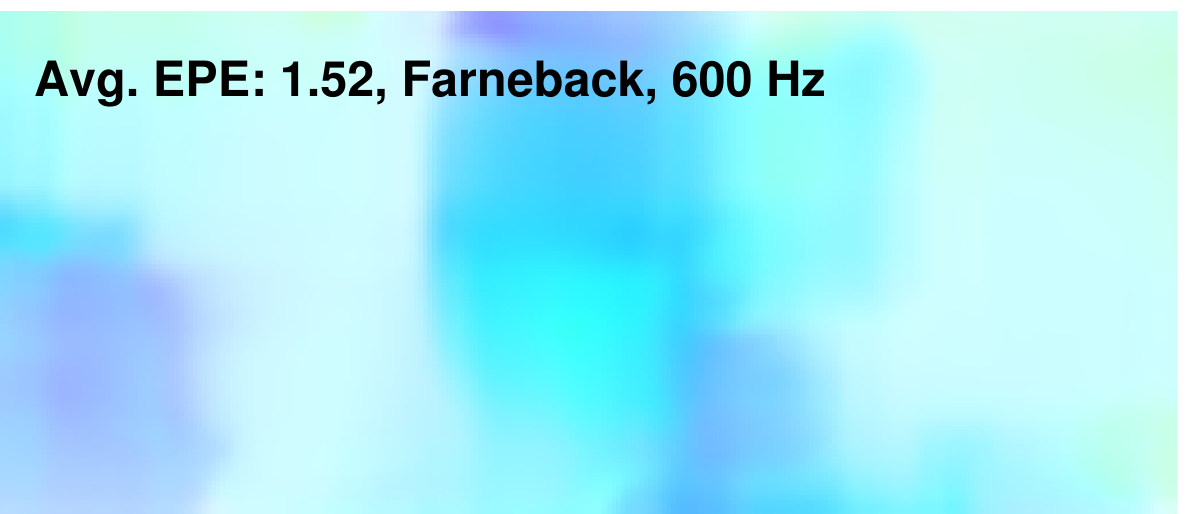}&
\includegraphics[width=0.195\textwidth]{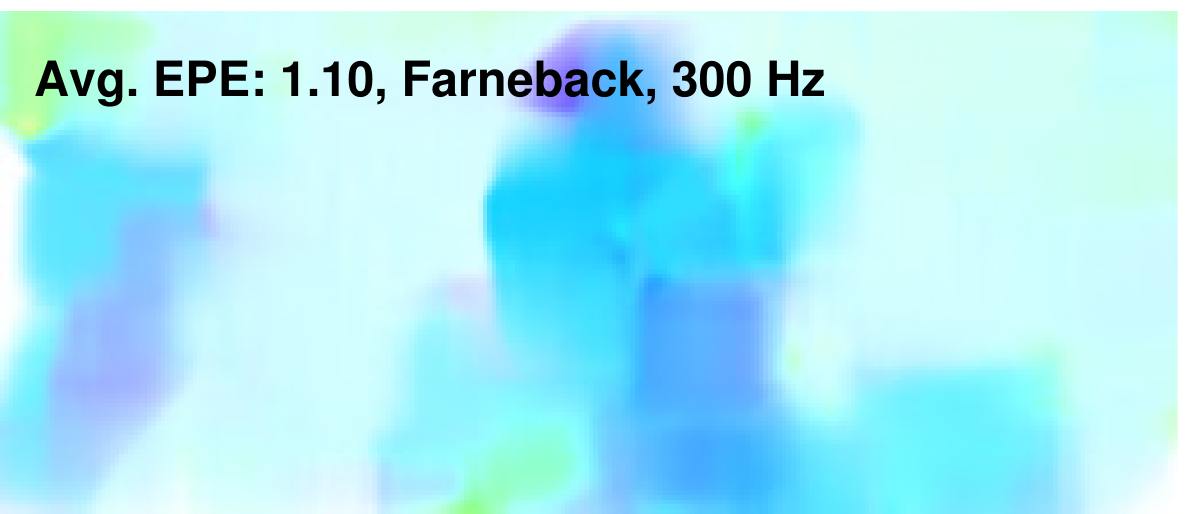}&
\includegraphics[width=0.195\textwidth]{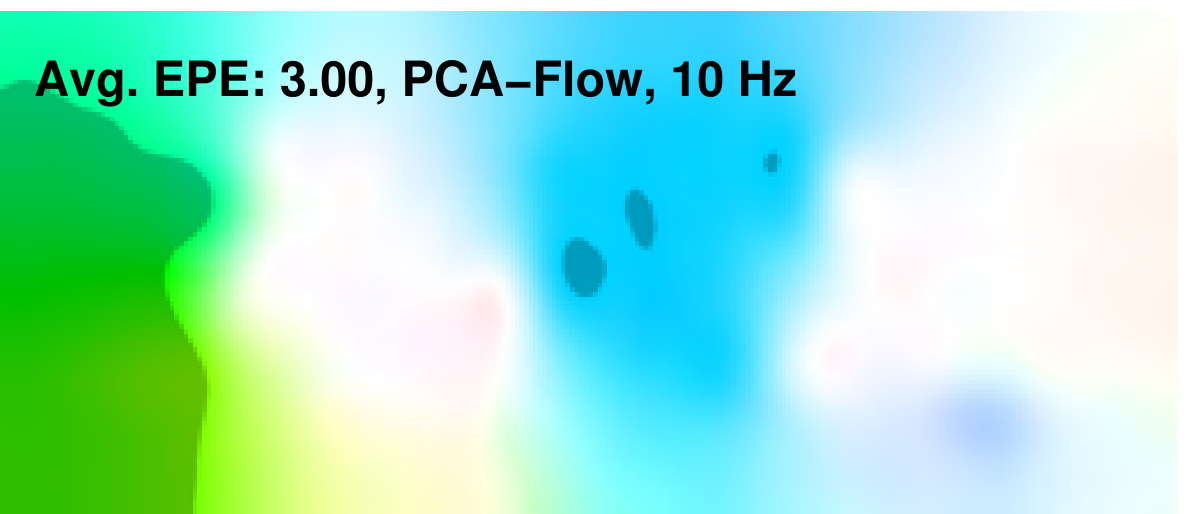}&
\includegraphics[width=0.195\textwidth]{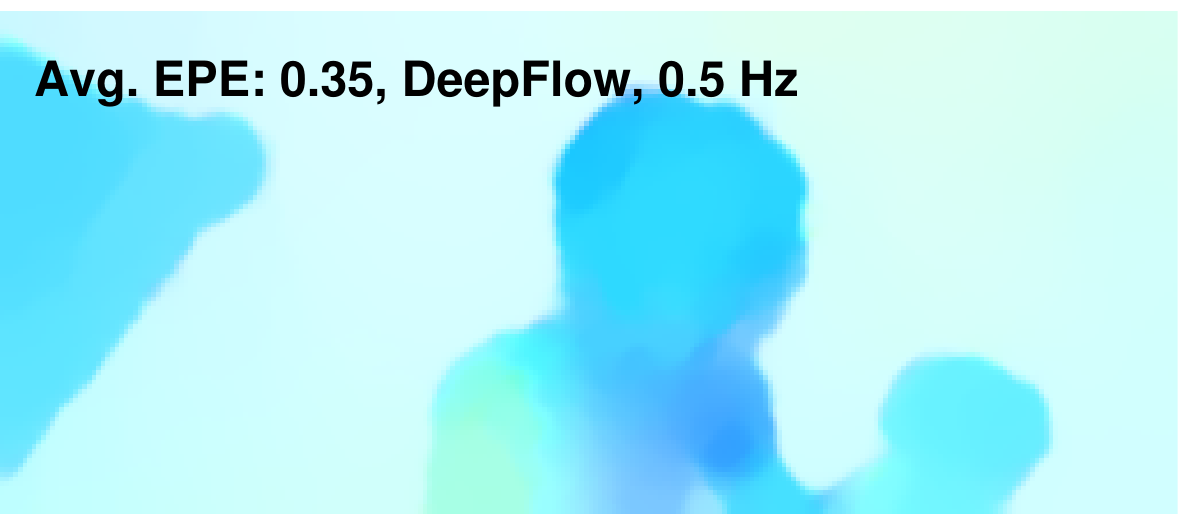}&
\includegraphics[width=0.195\textwidth]{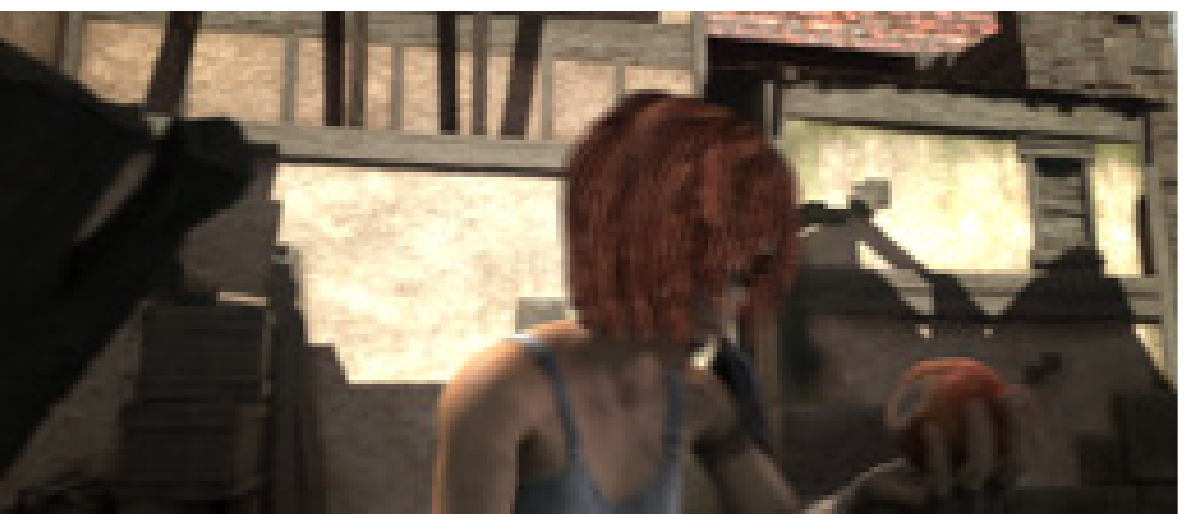}\\[6pt]
\includegraphics[width=0.195\textwidth]{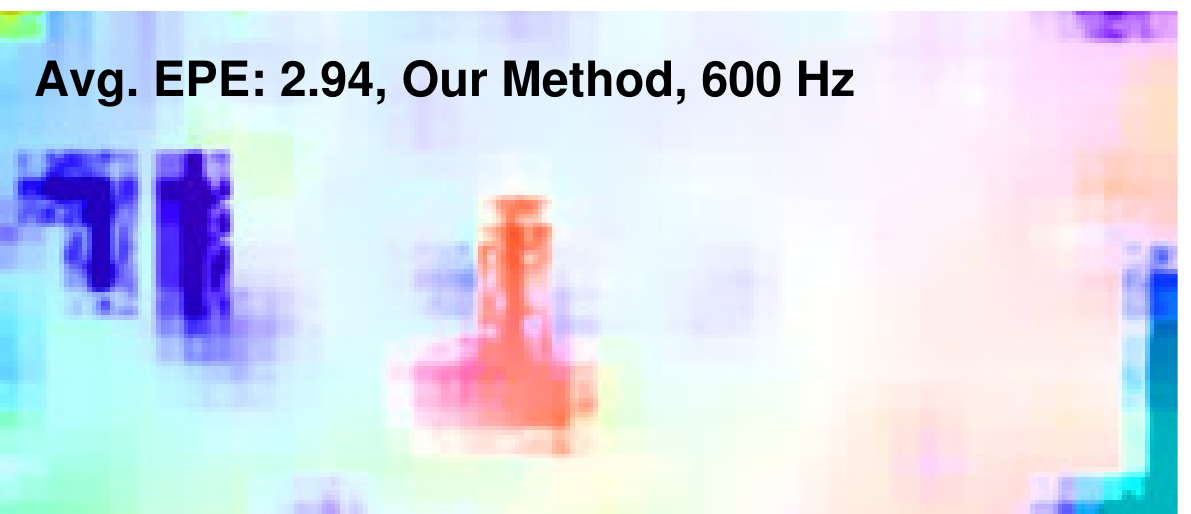}&
\includegraphics[width=0.195\textwidth]{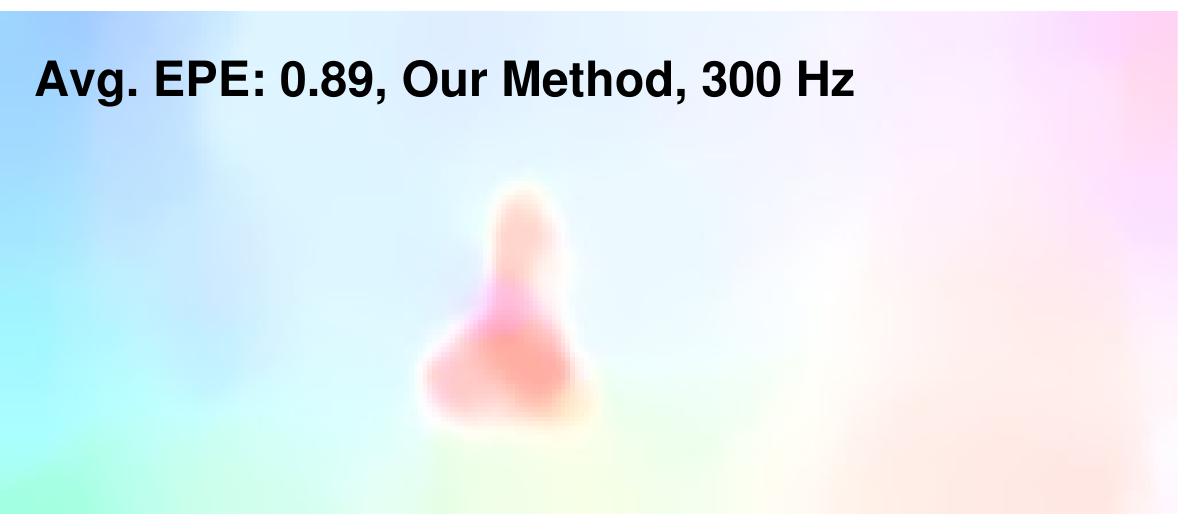}&
\includegraphics[width=0.195\textwidth]{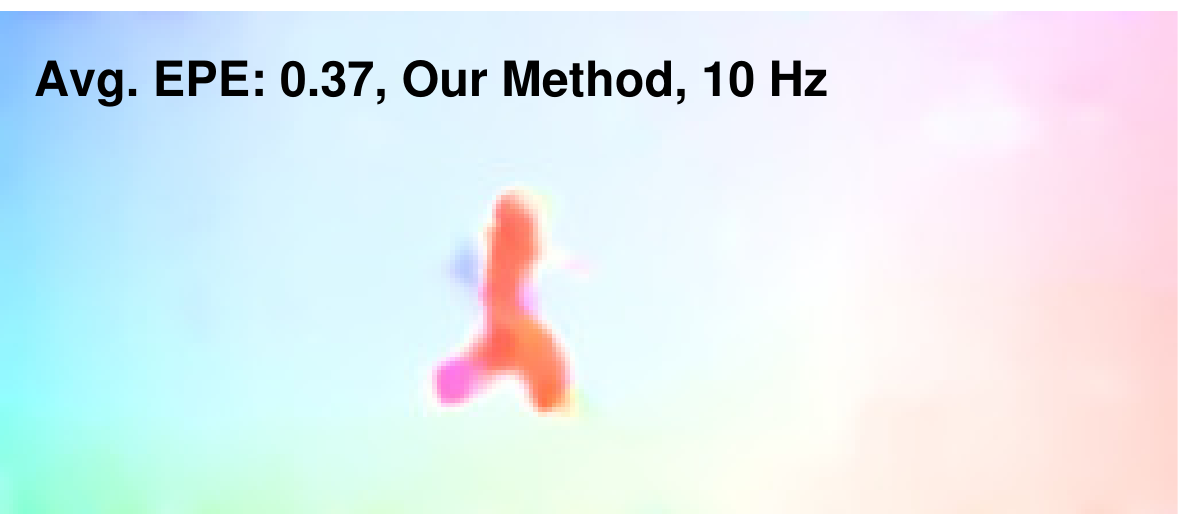}&
\includegraphics[width=0.195\textwidth]{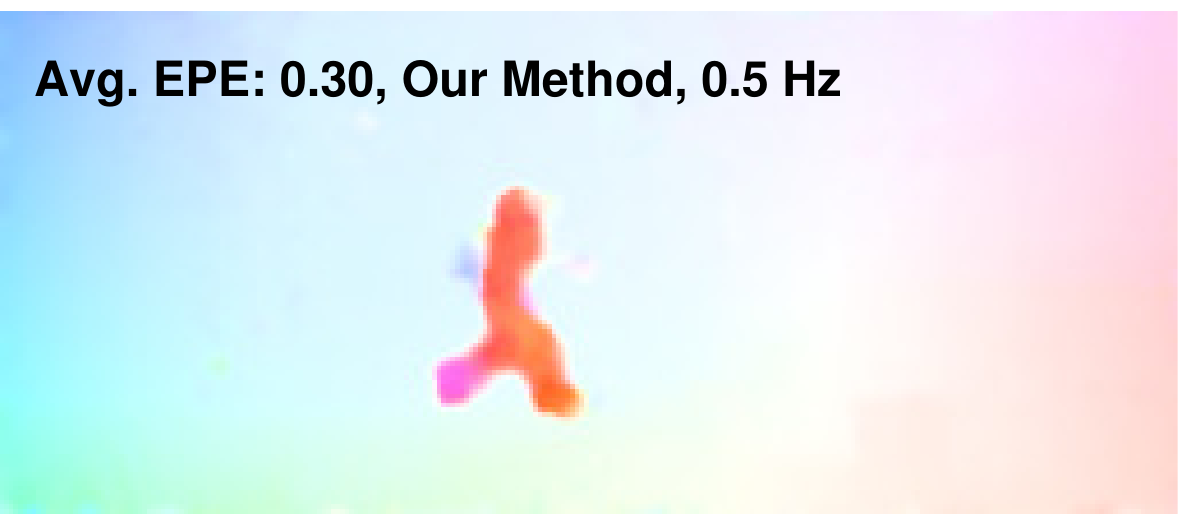}&
\includegraphics[width=0.195\textwidth]{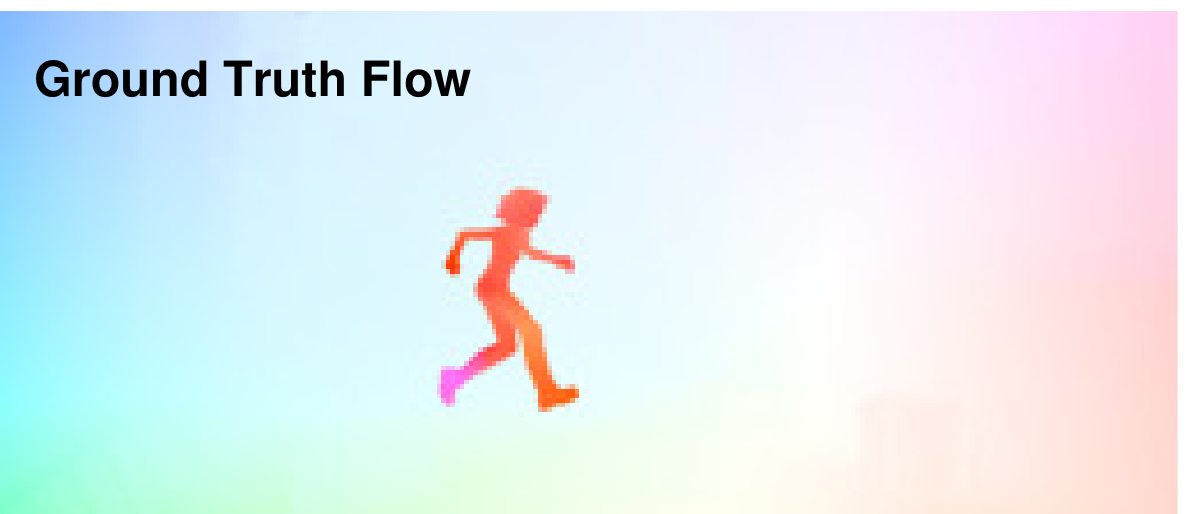}\\
\includegraphics[width=0.195\textwidth]{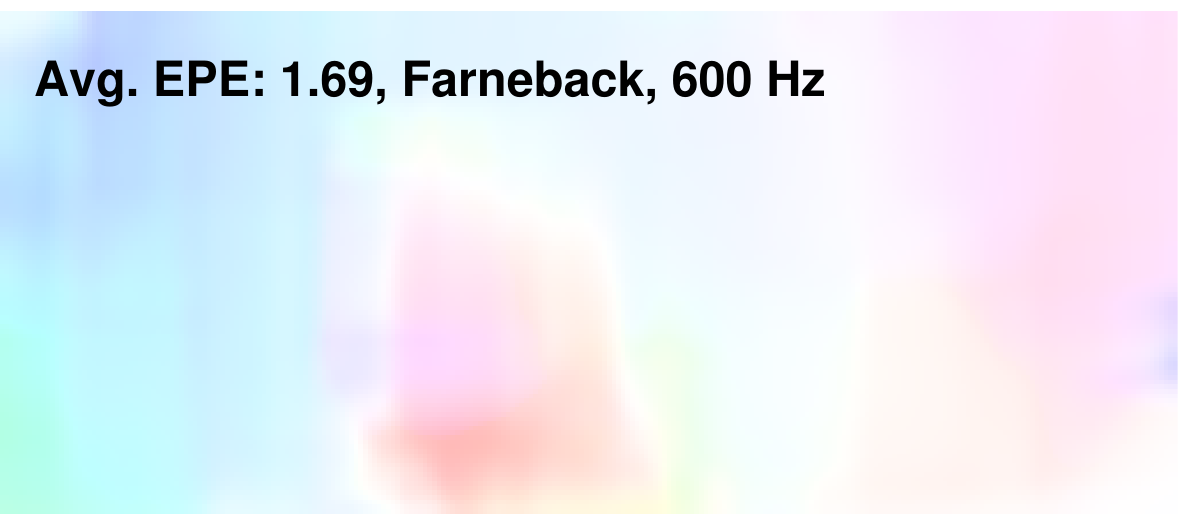}&
\includegraphics[width=0.195\textwidth]{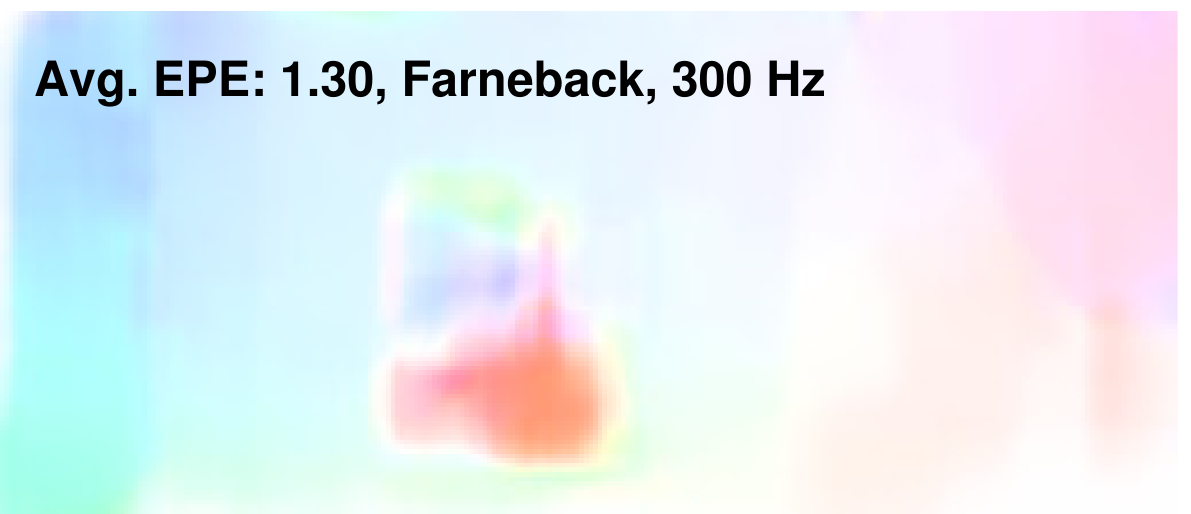}&
\includegraphics[width=0.195\textwidth]{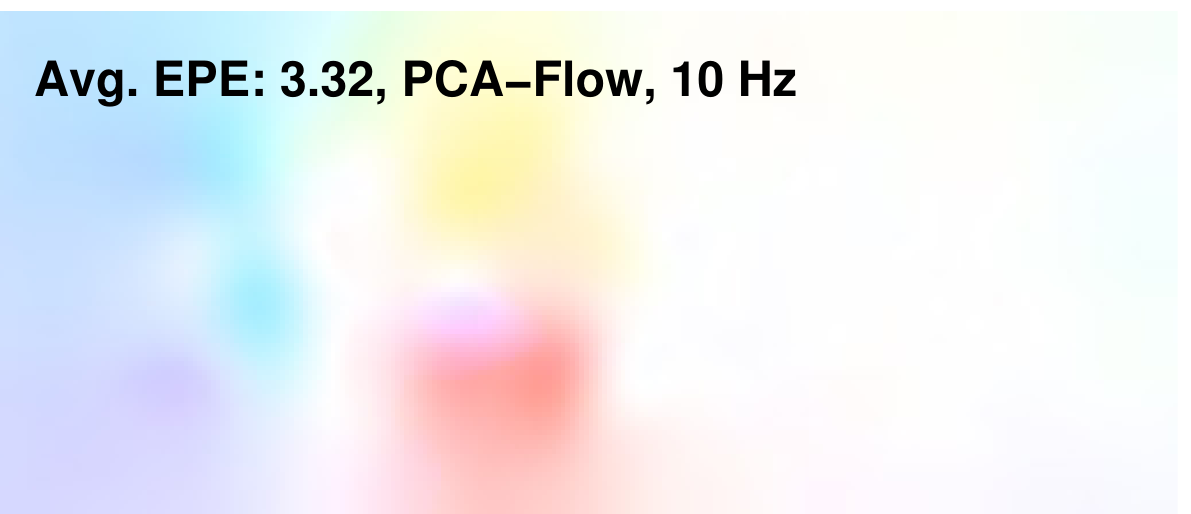}&
\includegraphics[width=0.195\textwidth]{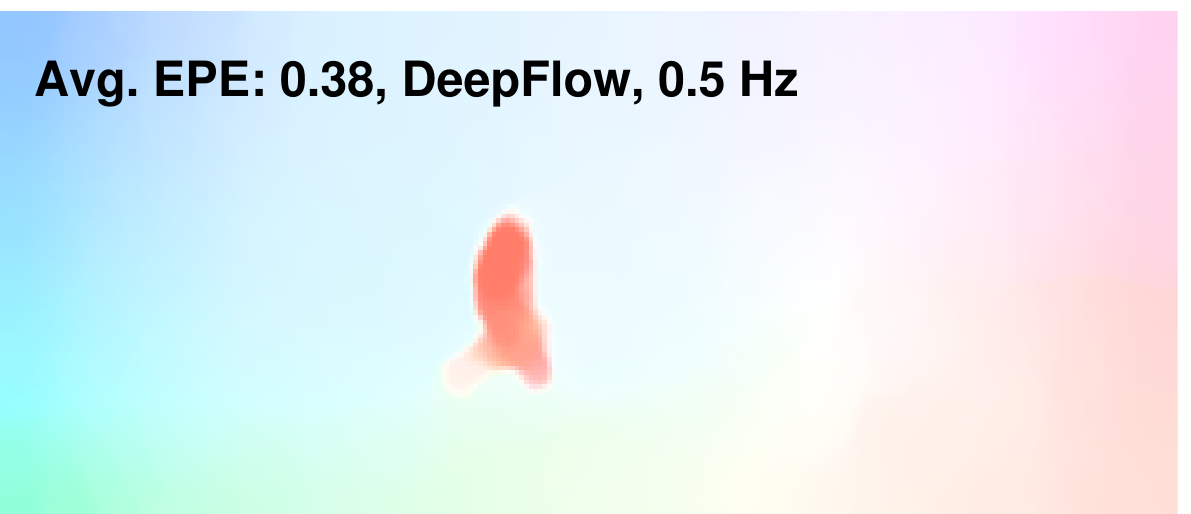}&
\includegraphics[width=0.195\textwidth]{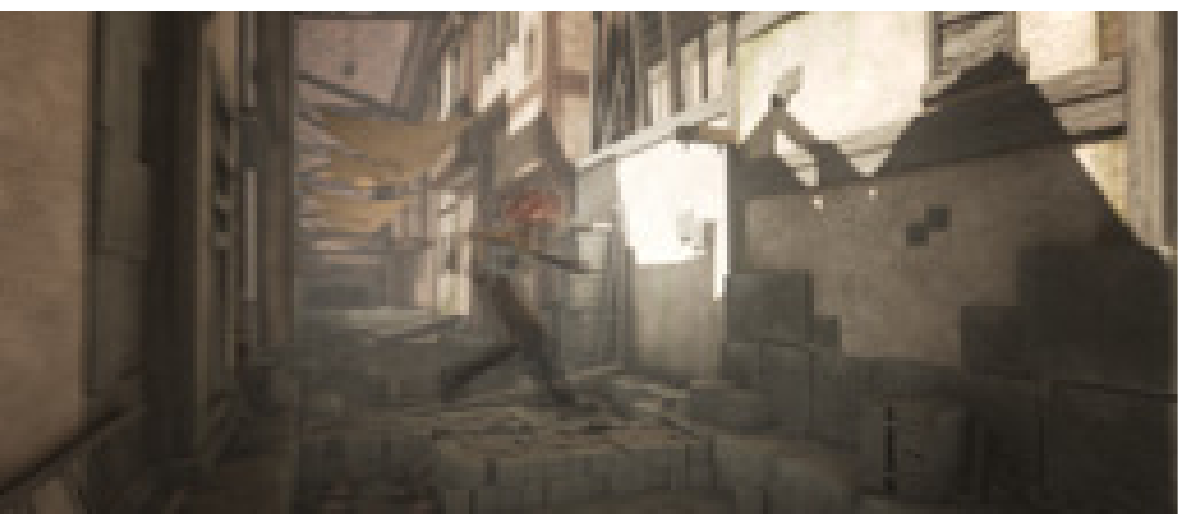}\\[6pt]
\includegraphics[width=0.195\textwidth]{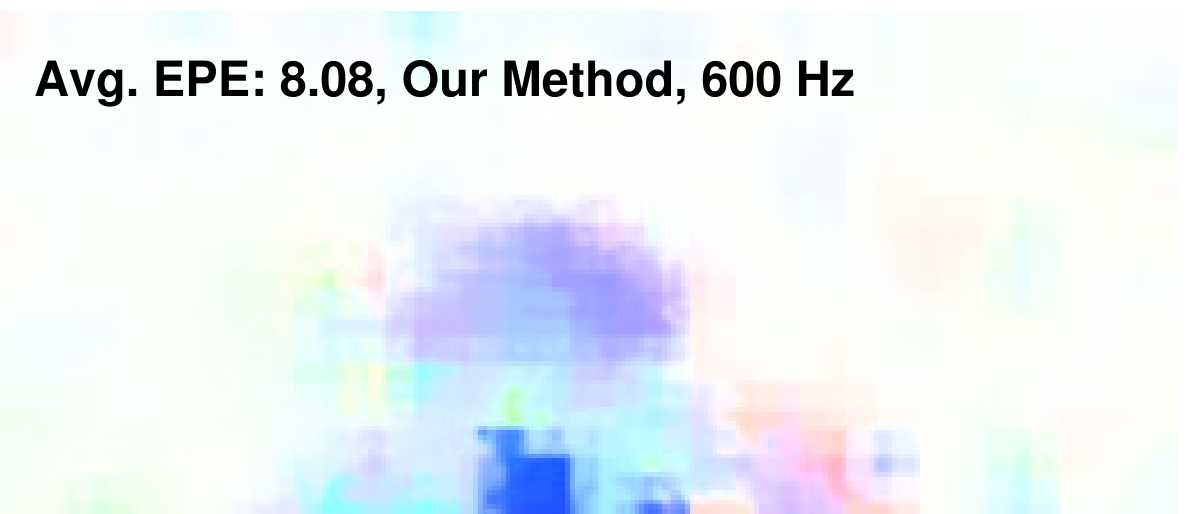}&
\includegraphics[width=0.195\textwidth]{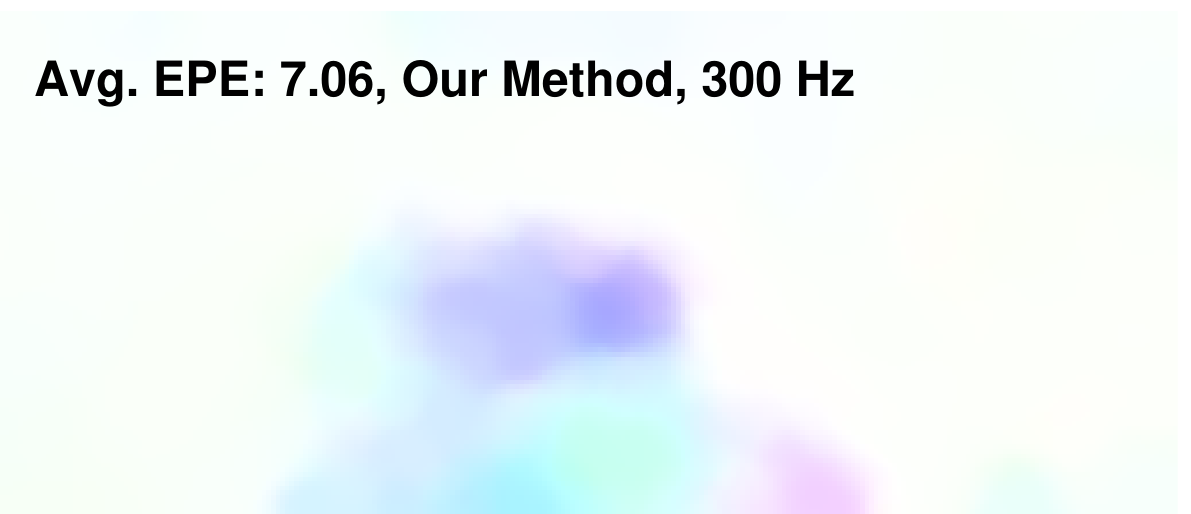}&
\includegraphics[width=0.195\textwidth]{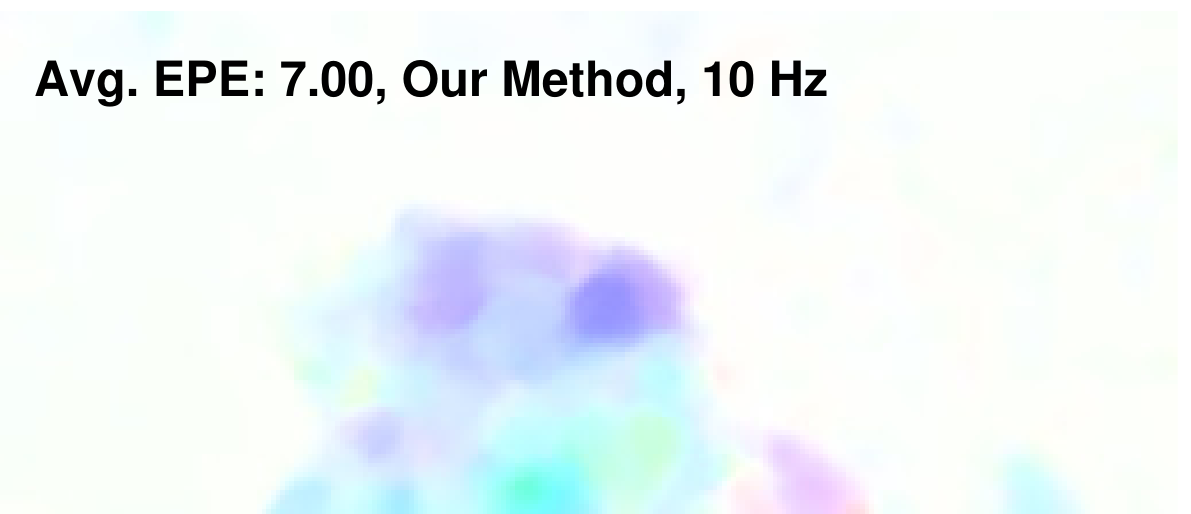}&
\includegraphics[width=0.195\textwidth]{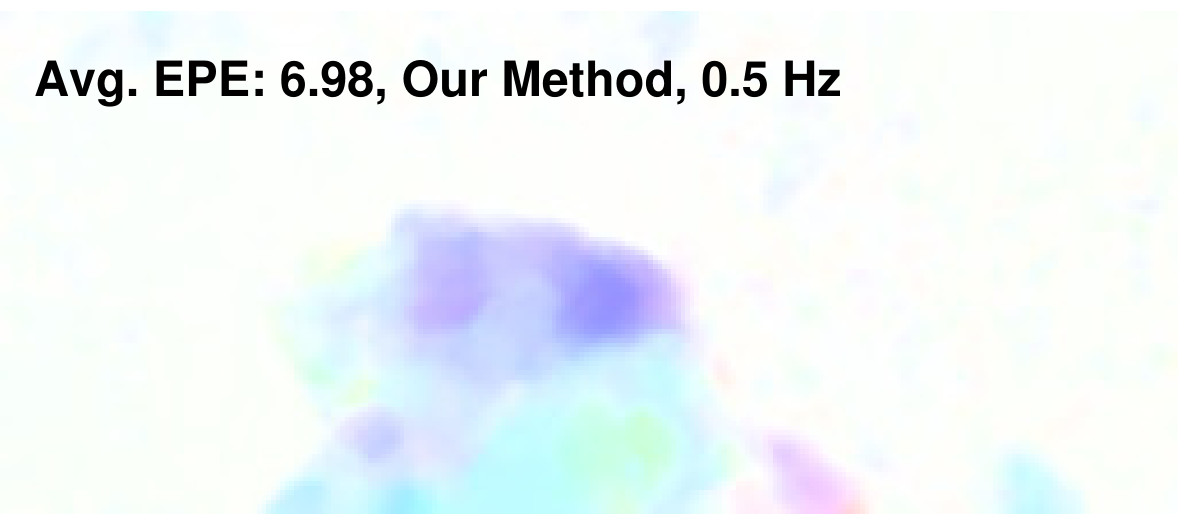}&
\includegraphics[width=0.195\textwidth]{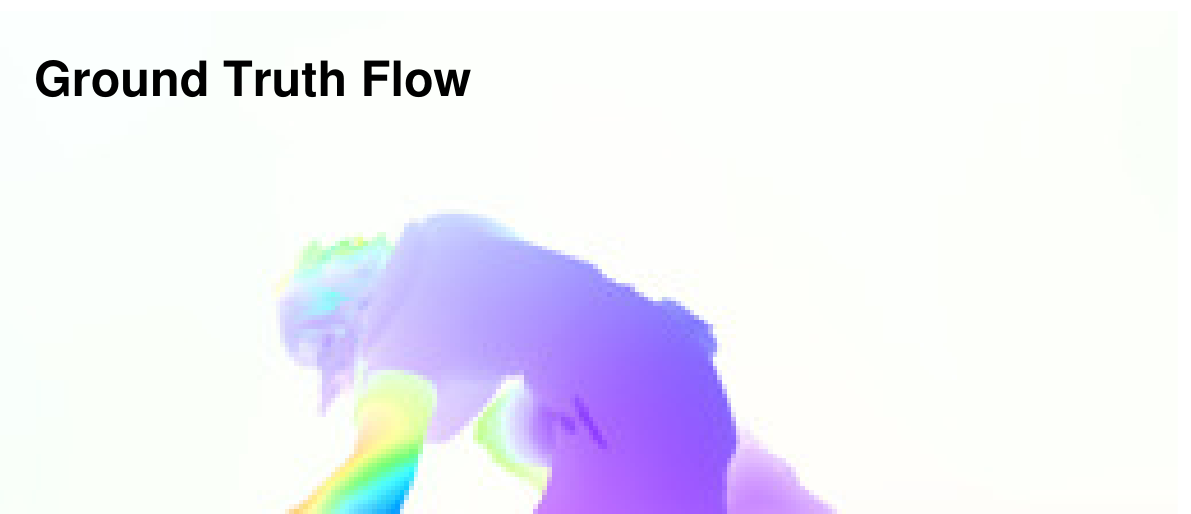}\\
\includegraphics[width=0.195\textwidth]{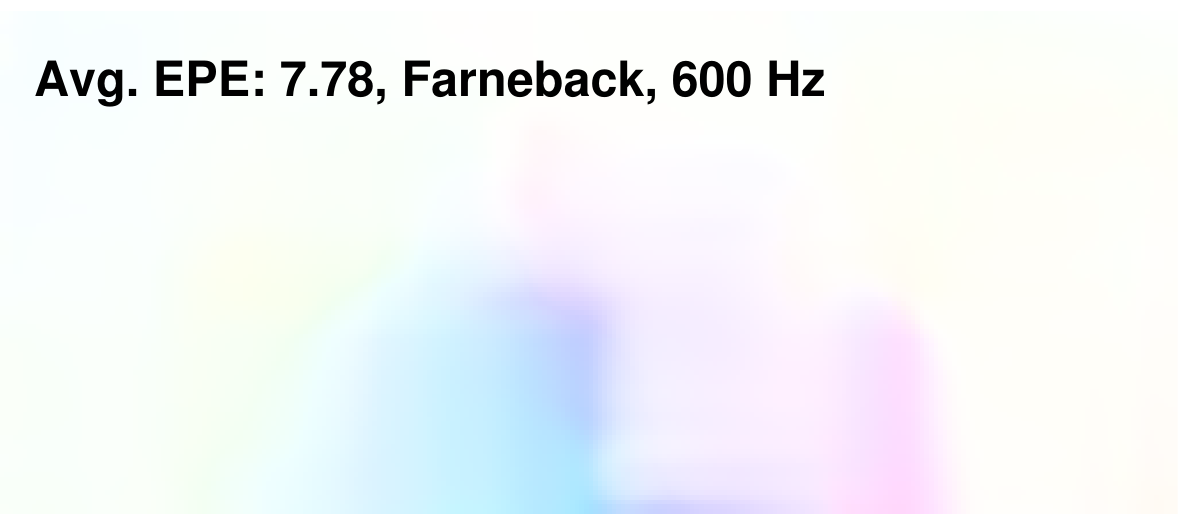}&
\includegraphics[width=0.195\textwidth]{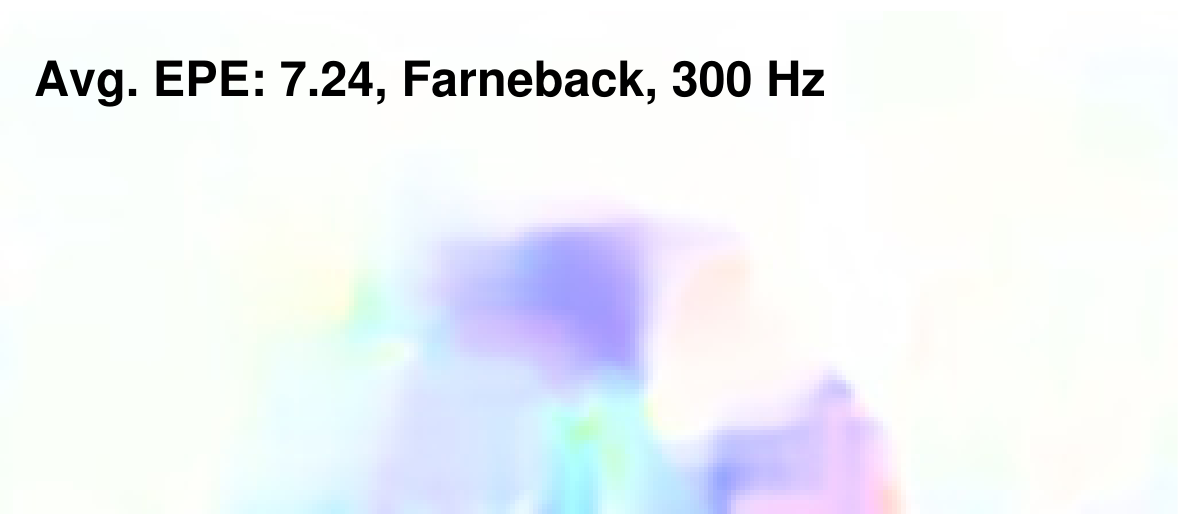}&
\includegraphics[width=0.195\textwidth]{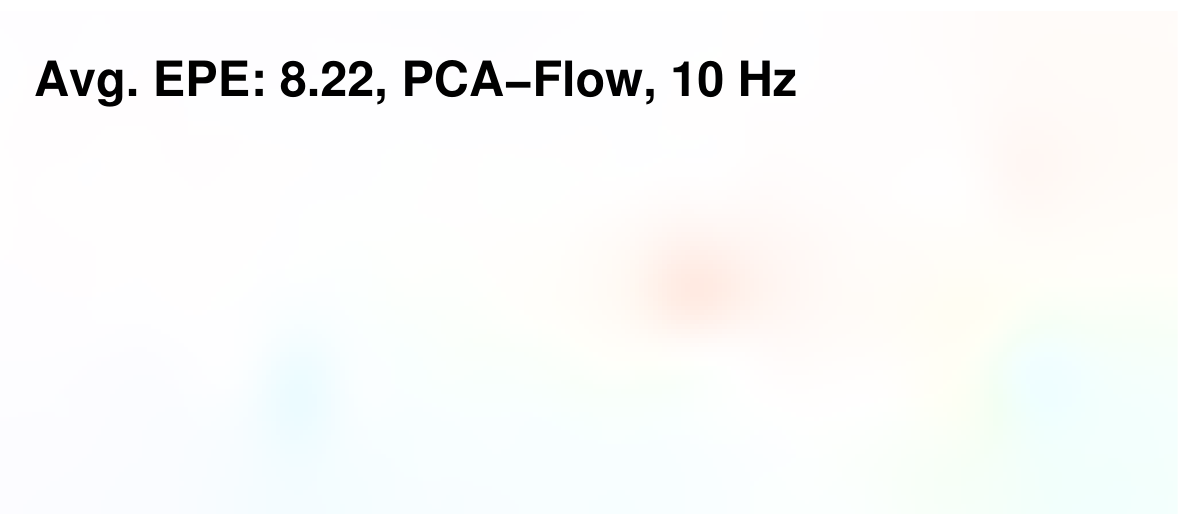}&
\includegraphics[width=0.195\textwidth]{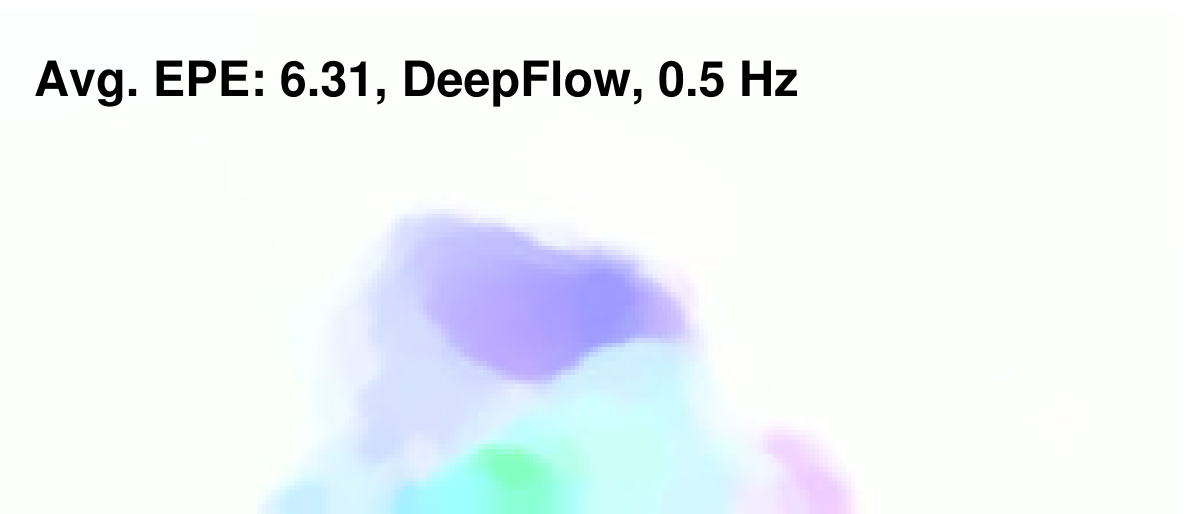}&
\includegraphics[width=0.195\textwidth]{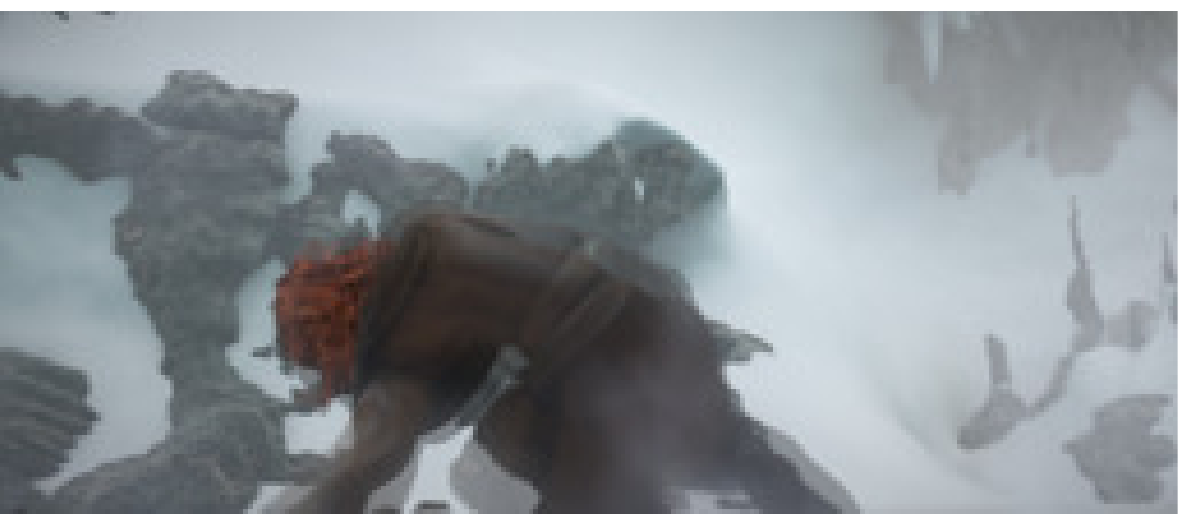}\\[6pt]
\includegraphics[width=0.195\textwidth]{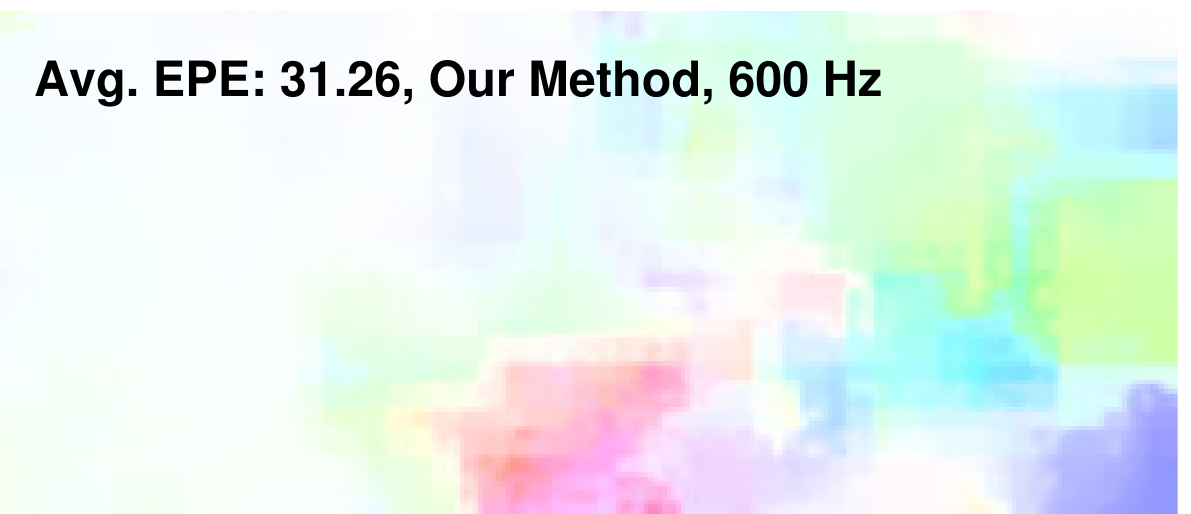}&
\includegraphics[width=0.195\textwidth]{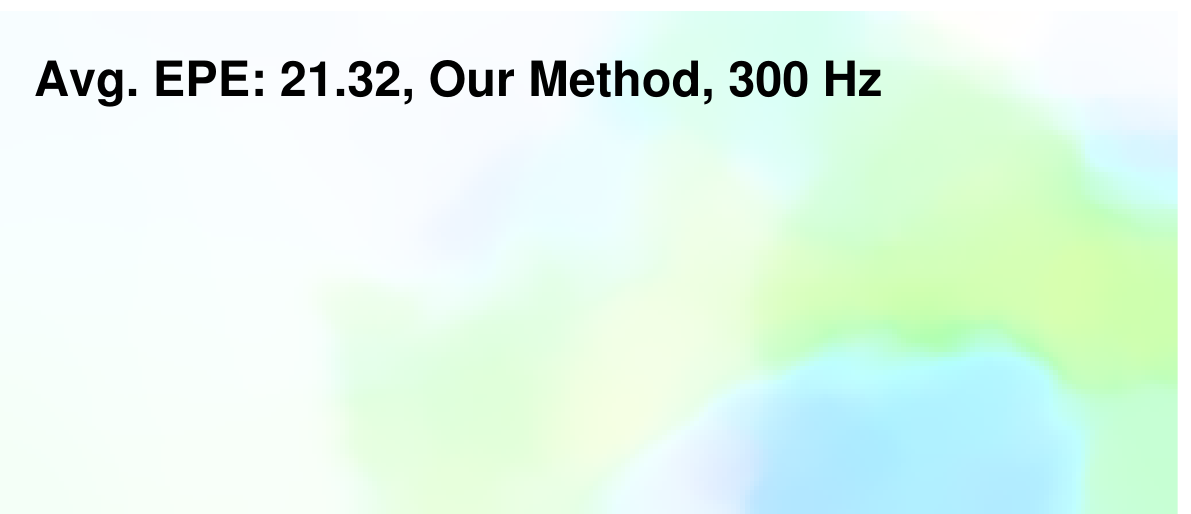}&
\includegraphics[width=0.195\textwidth]{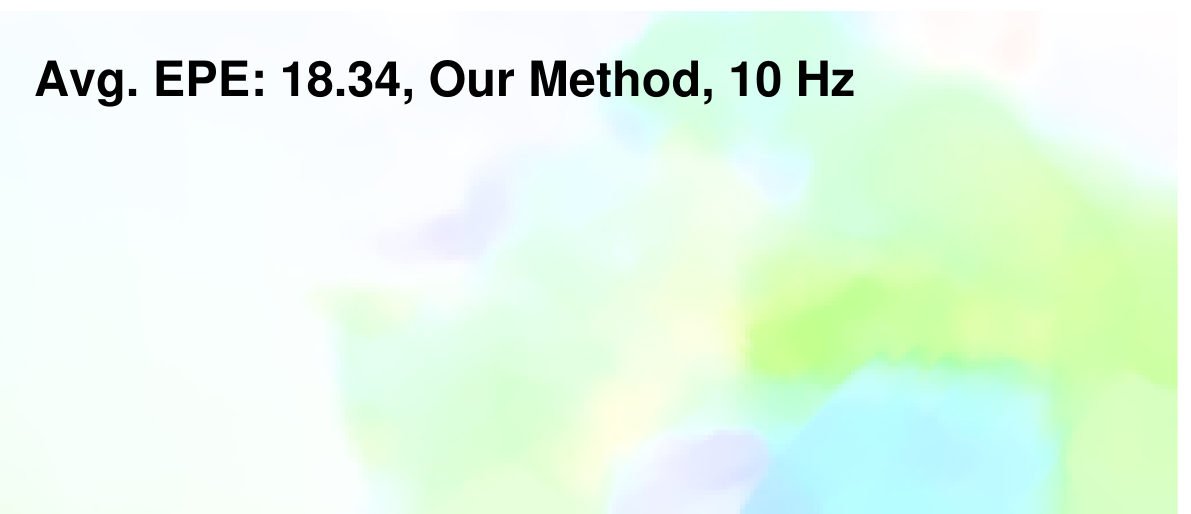}&
\includegraphics[width=0.195\textwidth]{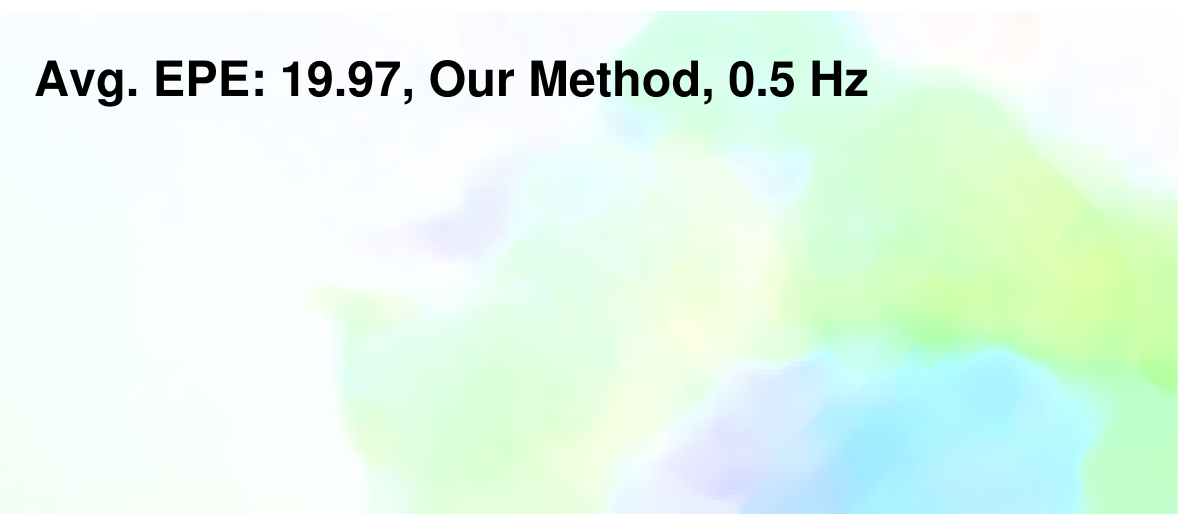}&
\includegraphics[width=0.195\textwidth]{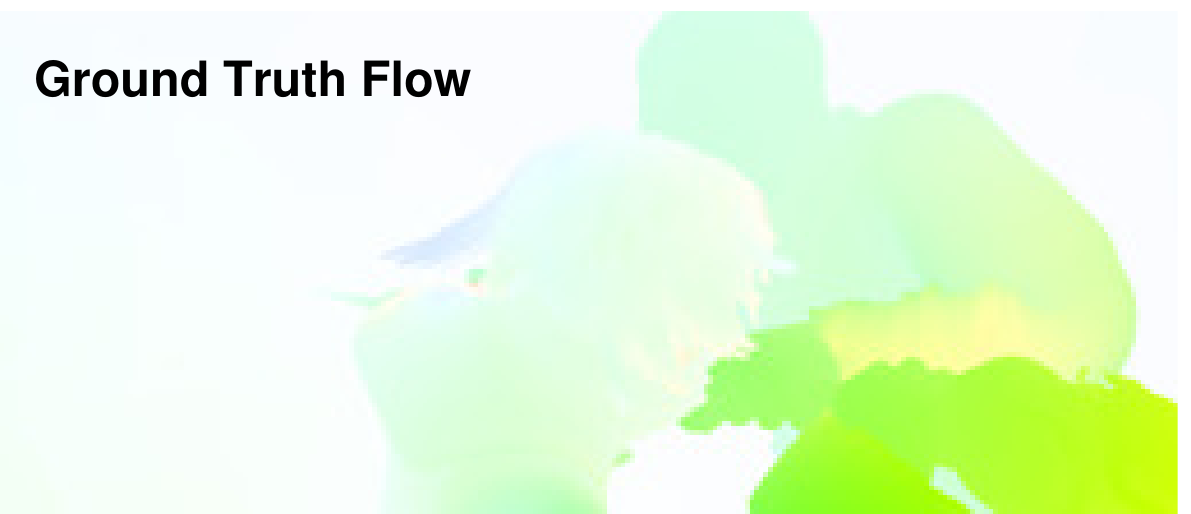}\\
\includegraphics[width=0.195\textwidth]{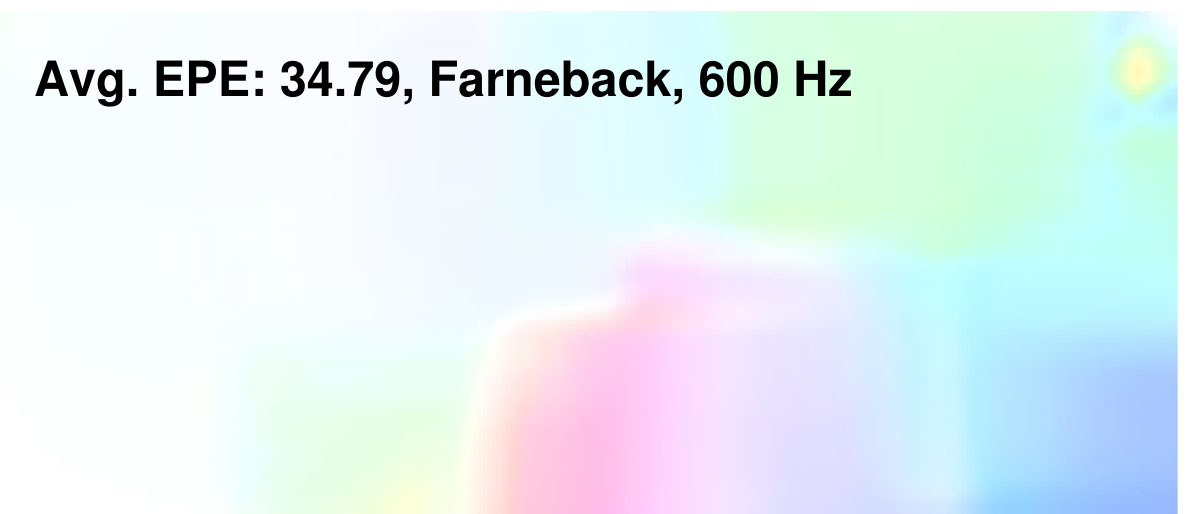}&
\includegraphics[width=0.195\textwidth]{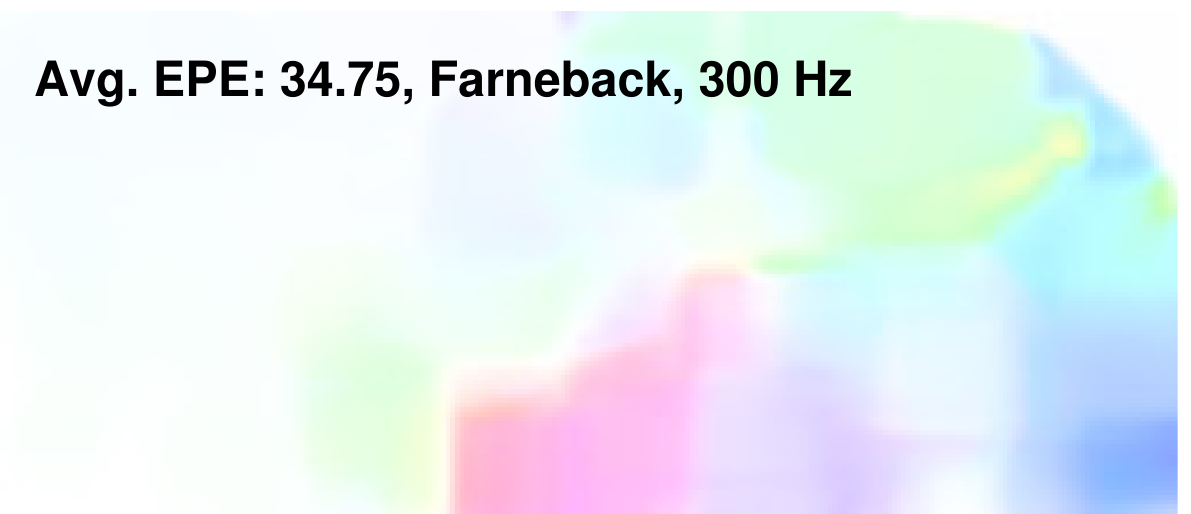}&
\includegraphics[width=0.195\textwidth]{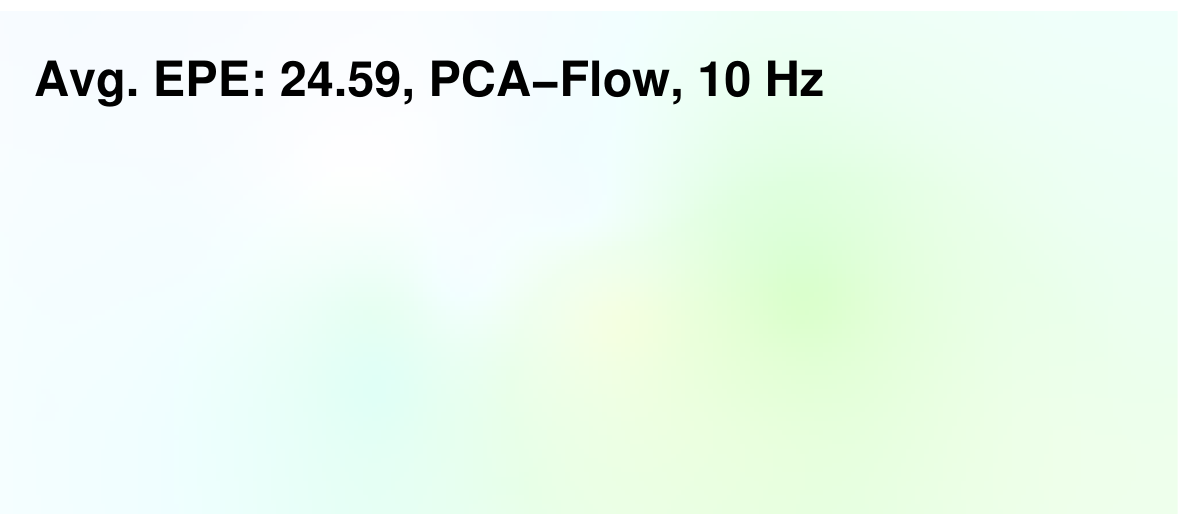}&
\includegraphics[width=0.195\textwidth]{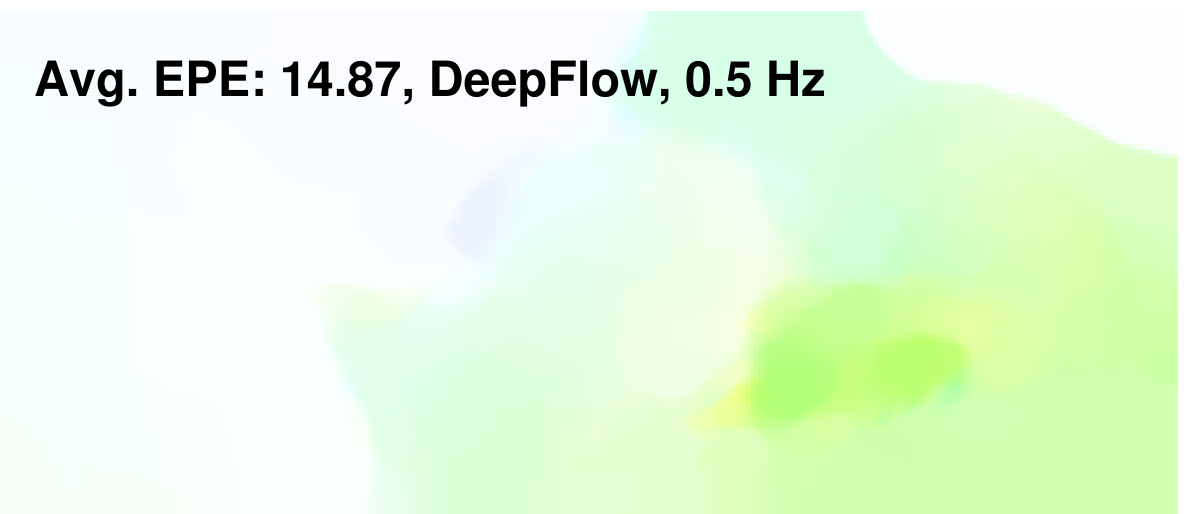}&
\includegraphics[width=0.195\textwidth]{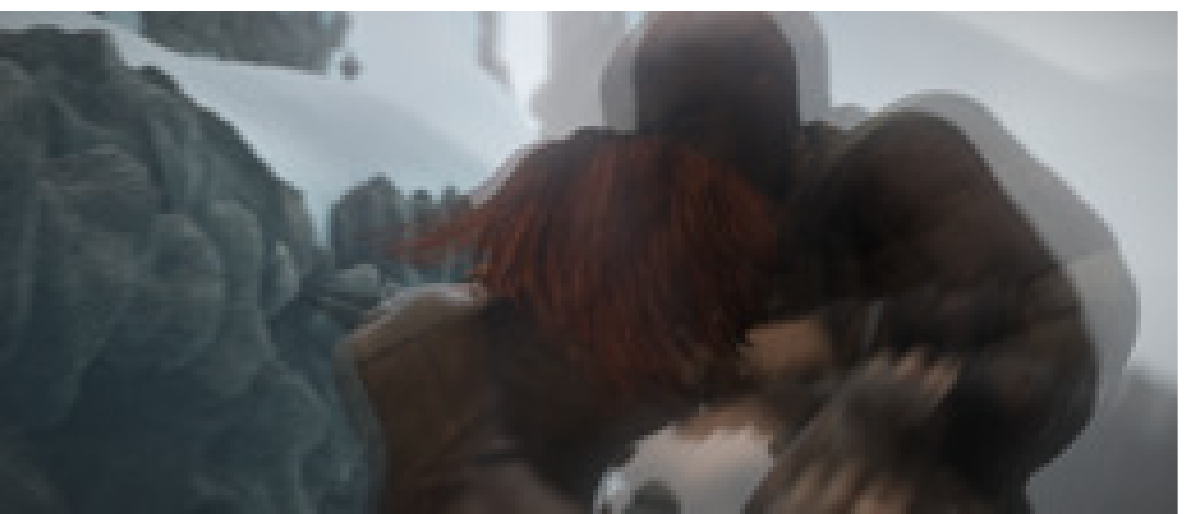}\\[6pt]
\includegraphics[width=0.195\textwidth]{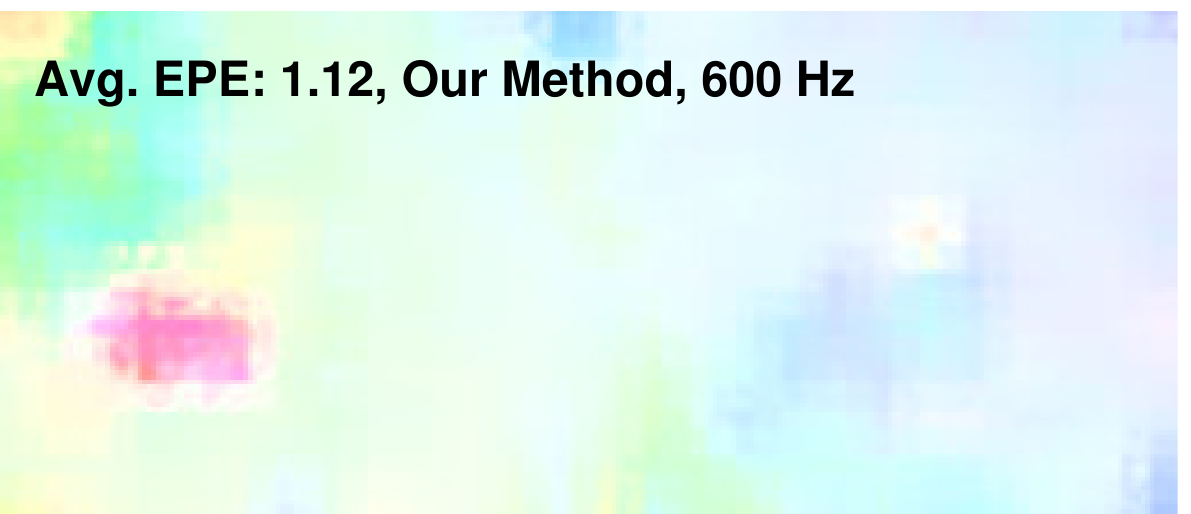}&
\includegraphics[width=0.195\textwidth]{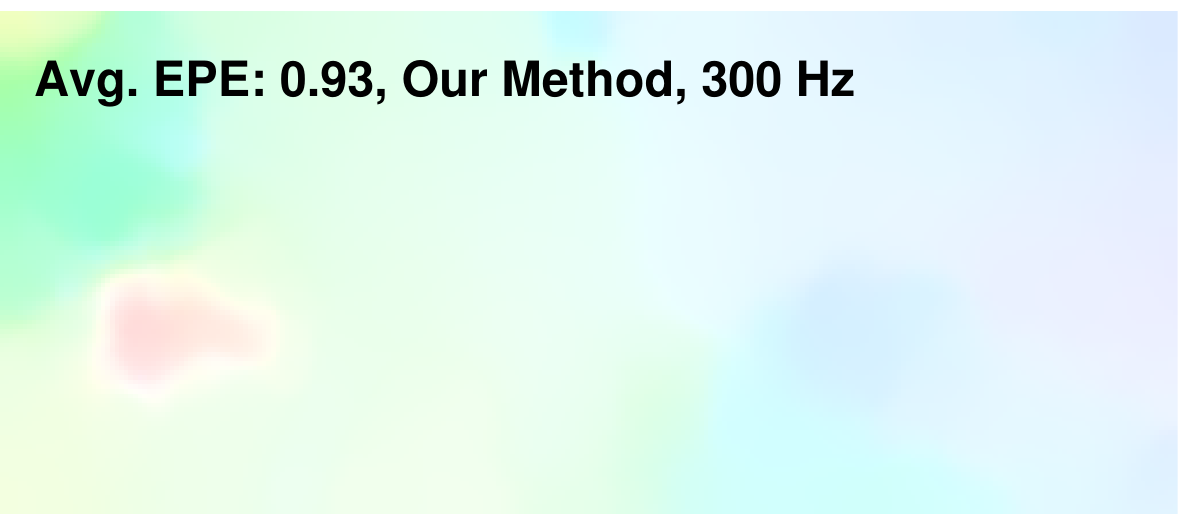}&
\includegraphics[width=0.195\textwidth]{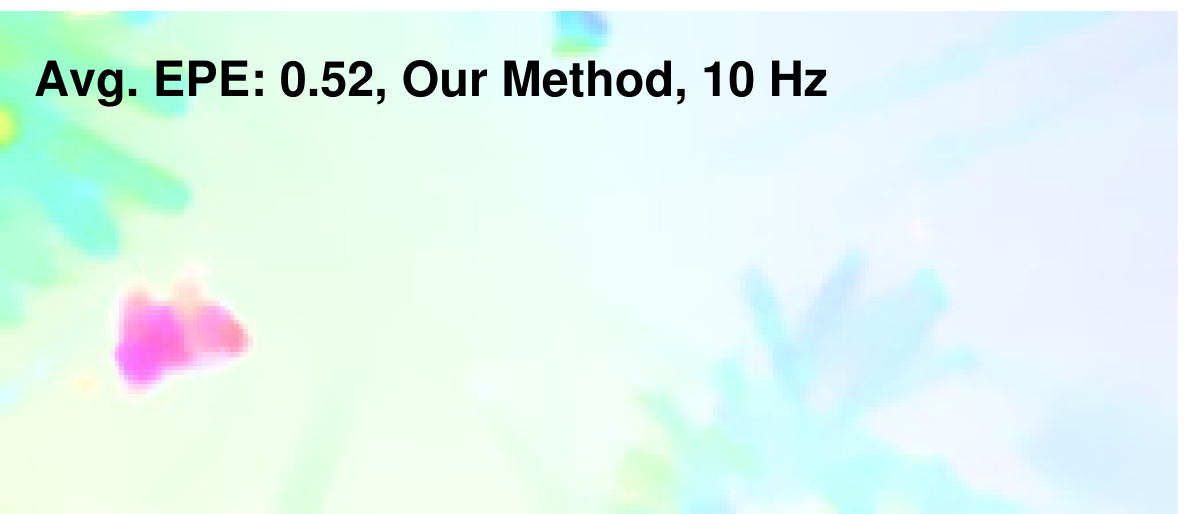}&
\includegraphics[width=0.195\textwidth]{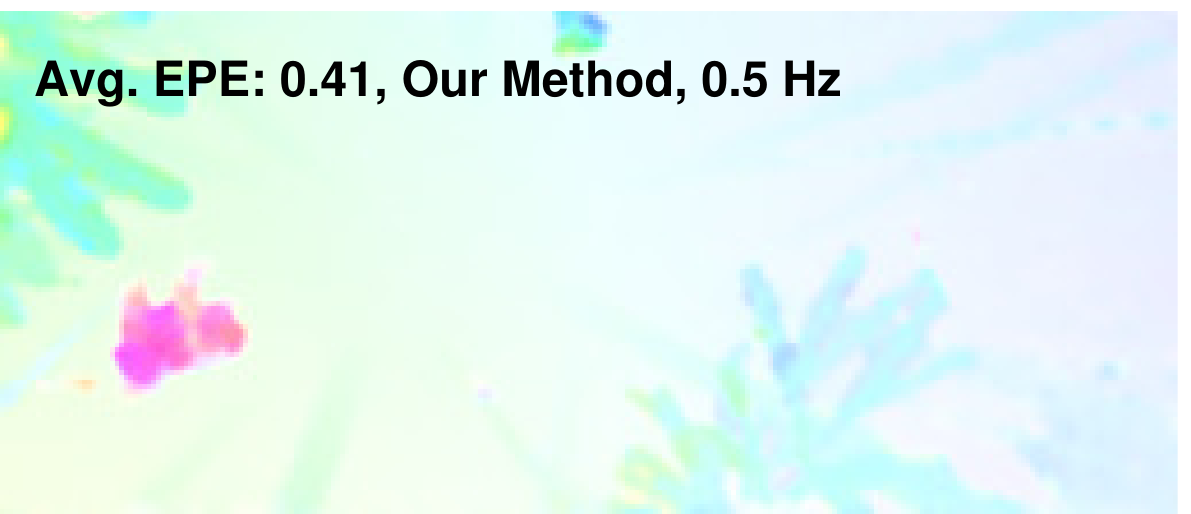}&
\includegraphics[width=0.195\textwidth]{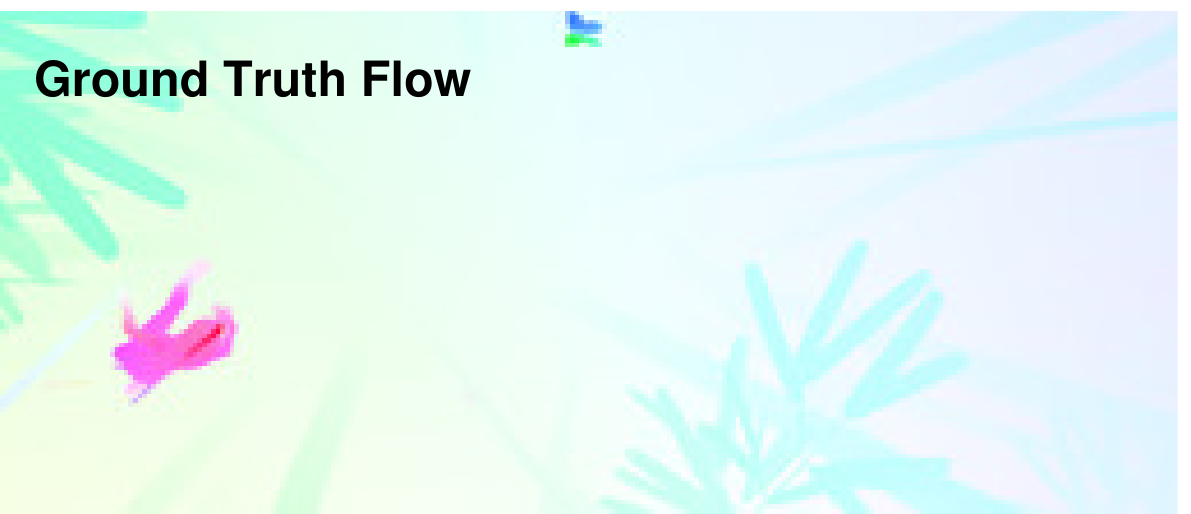}\\
\includegraphics[width=0.195\textwidth]{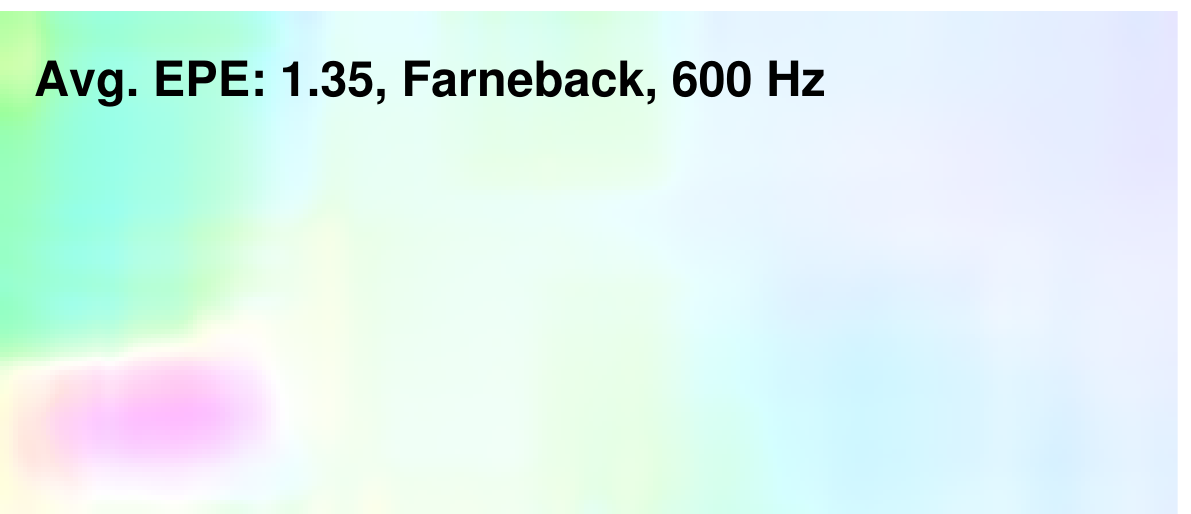}&
\includegraphics[width=0.195\textwidth]{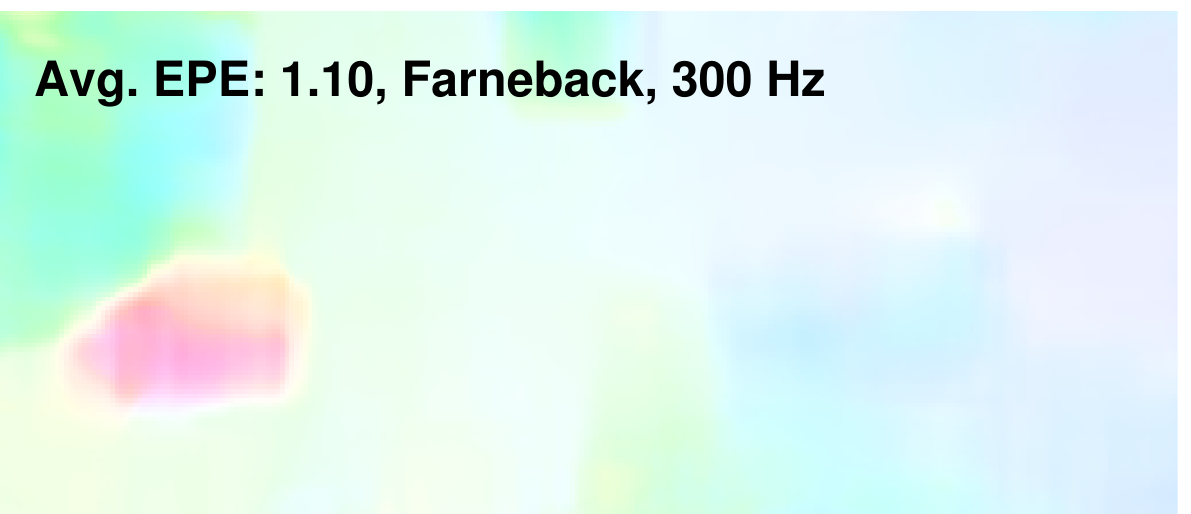}&
\includegraphics[width=0.195\textwidth]{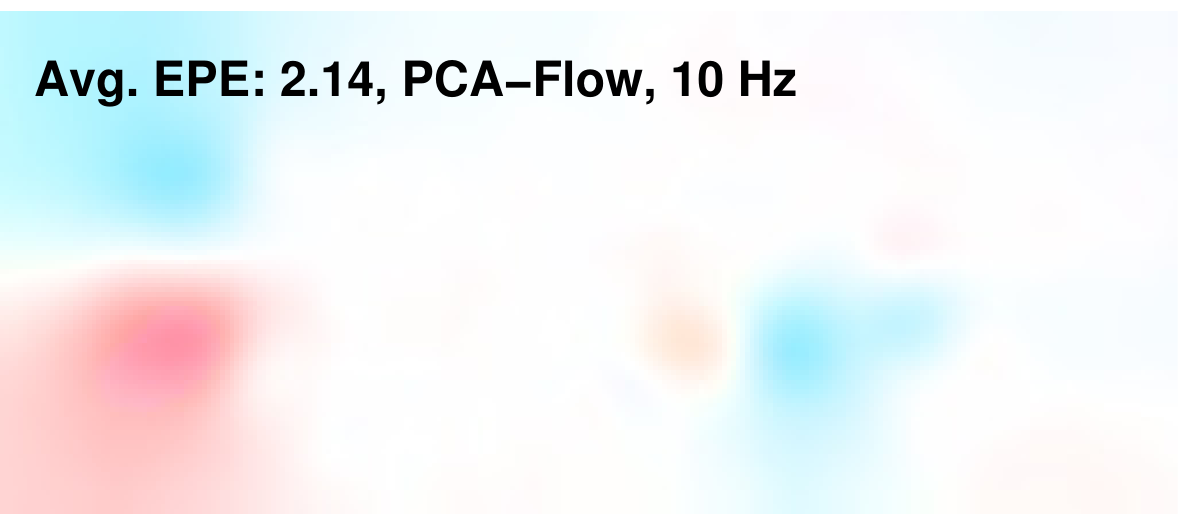}&
\includegraphics[width=0.195\textwidth]{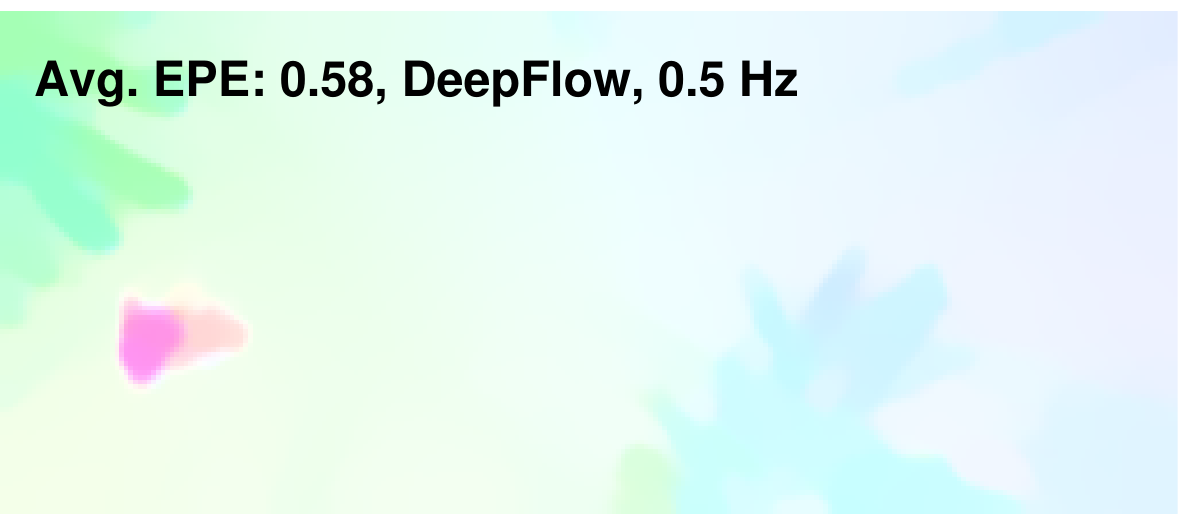}&
\includegraphics[width=0.195\textwidth]{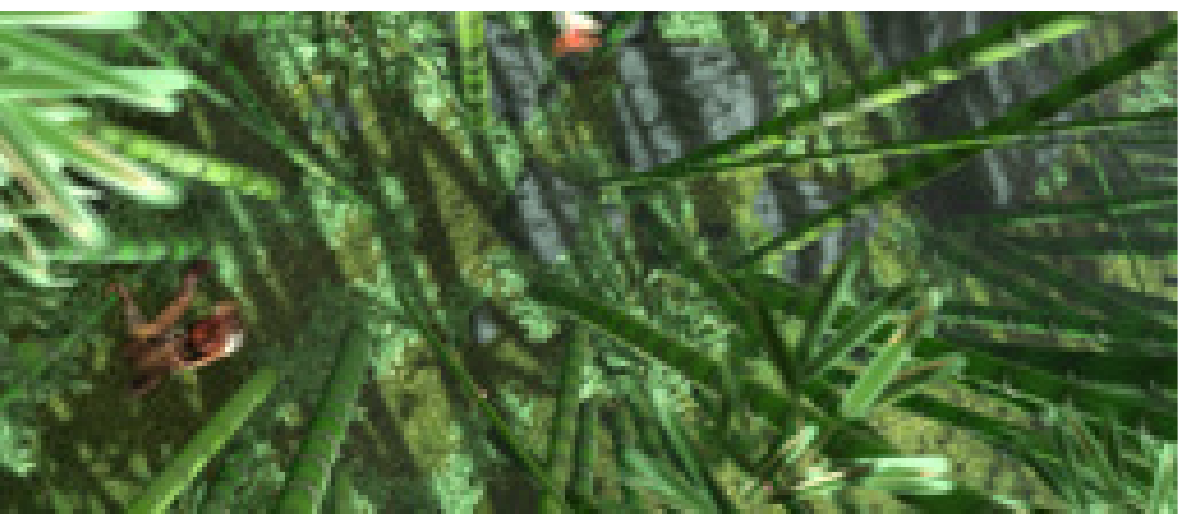}\\[6pt]
\includegraphics[width=0.195\textwidth]{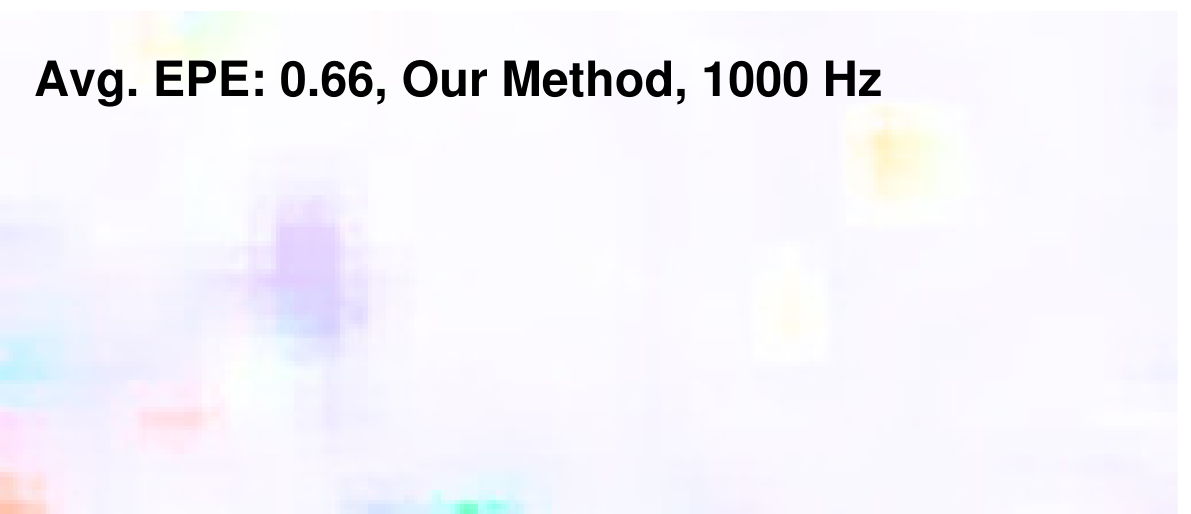}&
\includegraphics[width=0.195\textwidth]{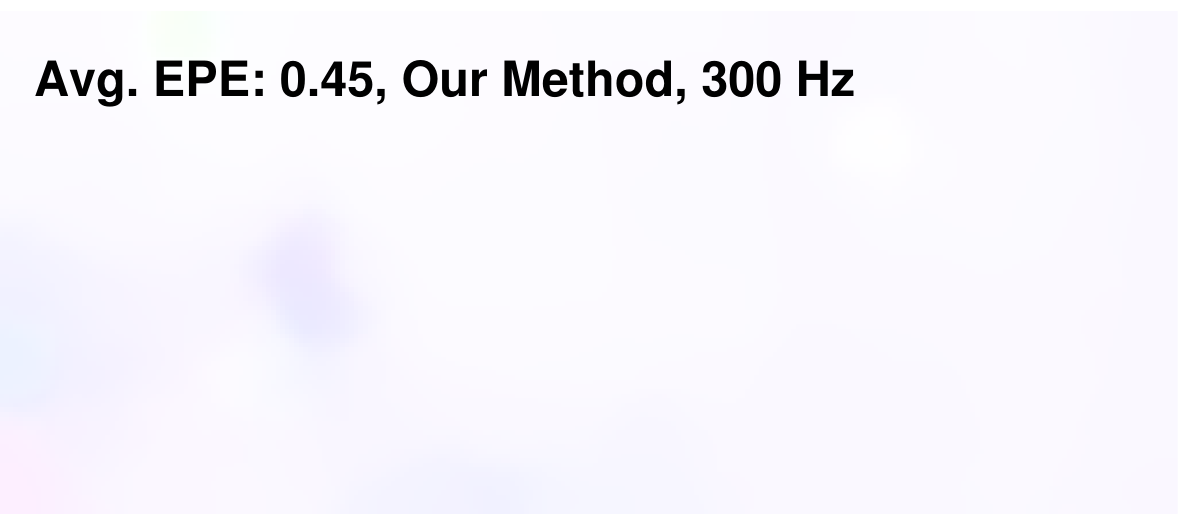}&
\includegraphics[width=0.195\textwidth]{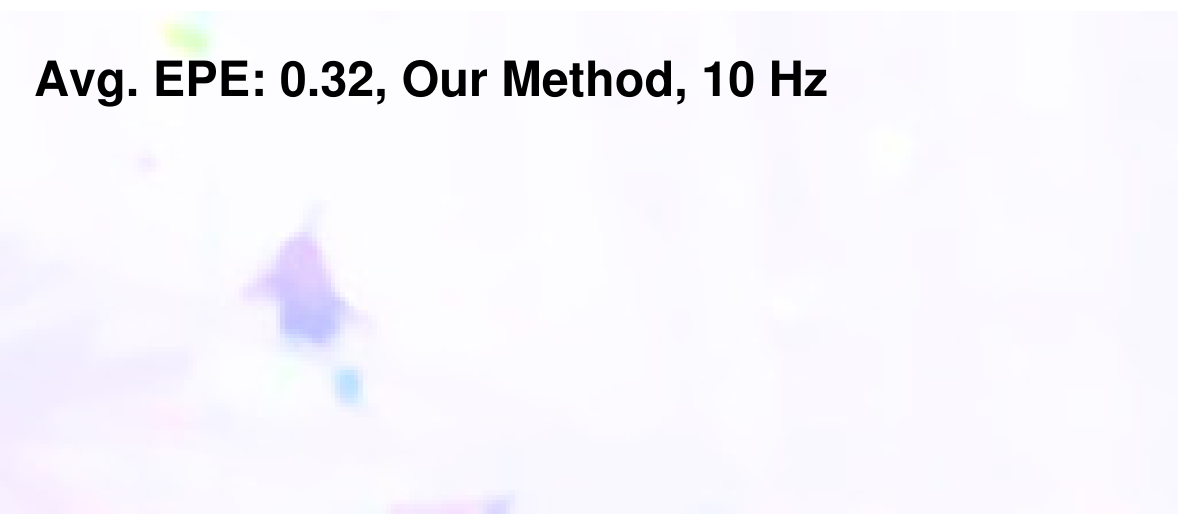}&
\includegraphics[width=0.195\textwidth]{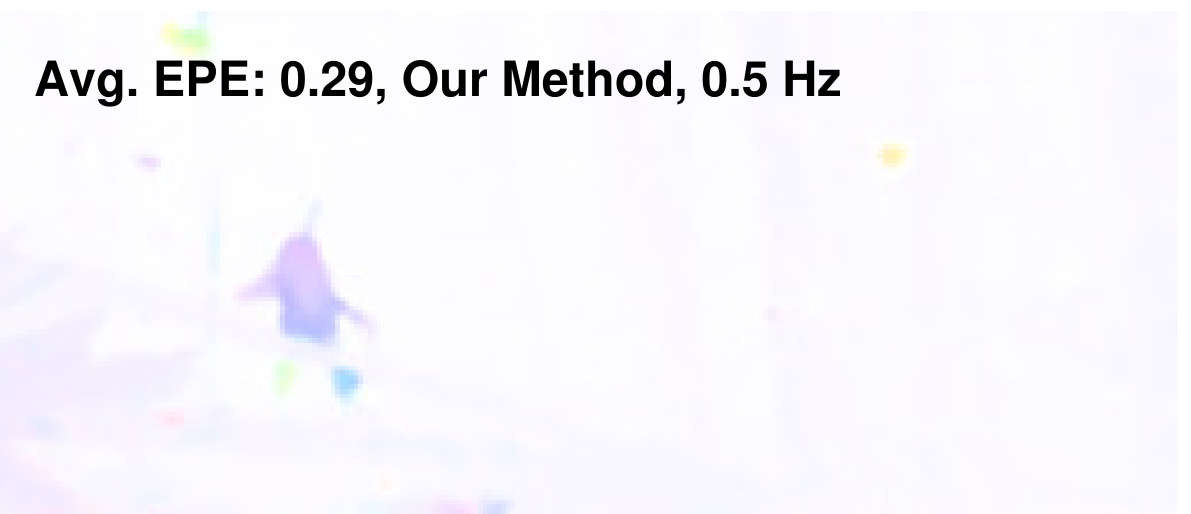}&
\includegraphics[width=0.195\textwidth]{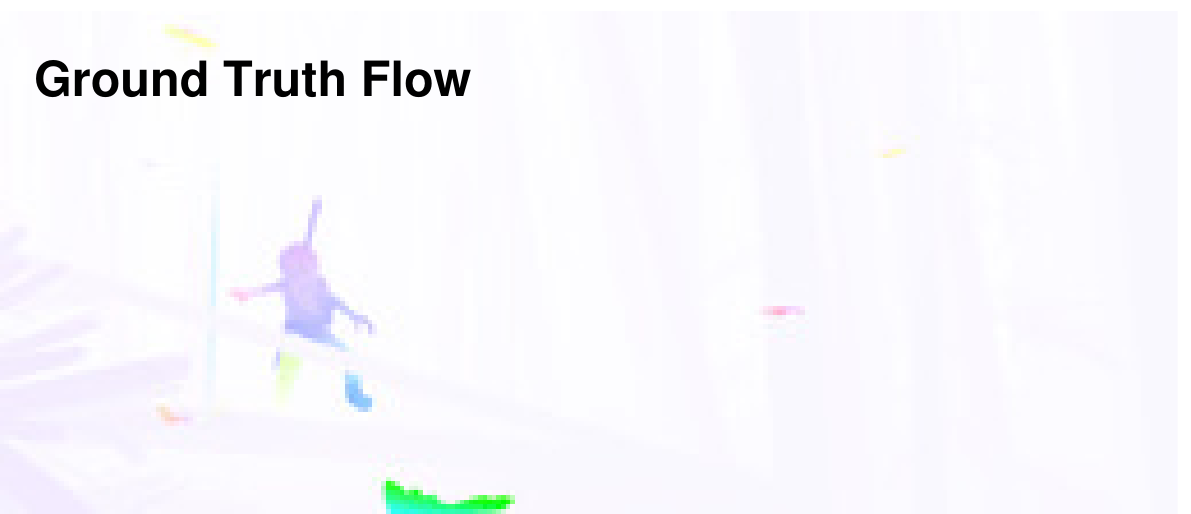}\\
\includegraphics[width=0.195\textwidth]{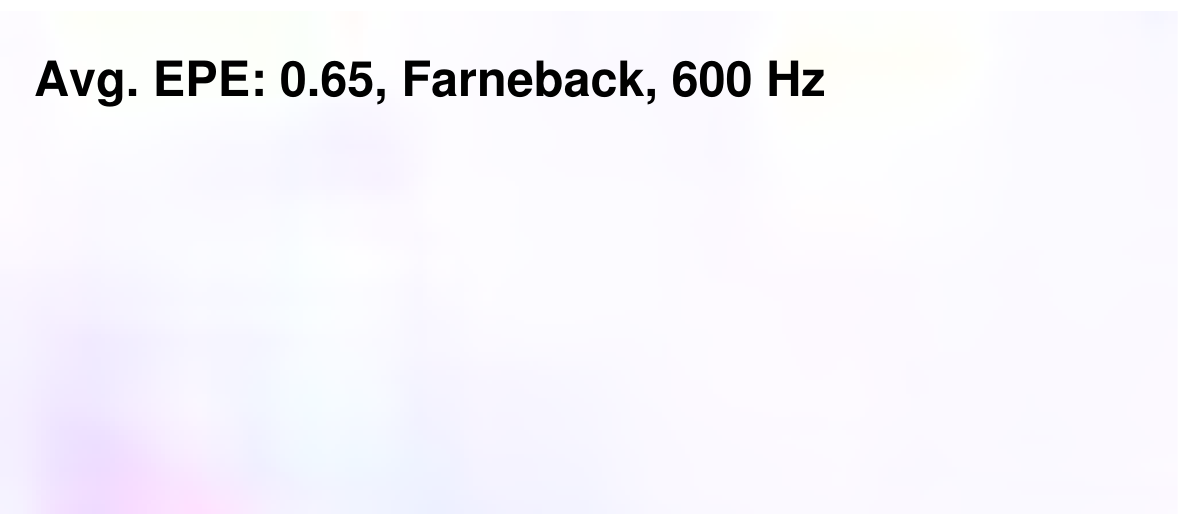}&
\includegraphics[width=0.195\textwidth]{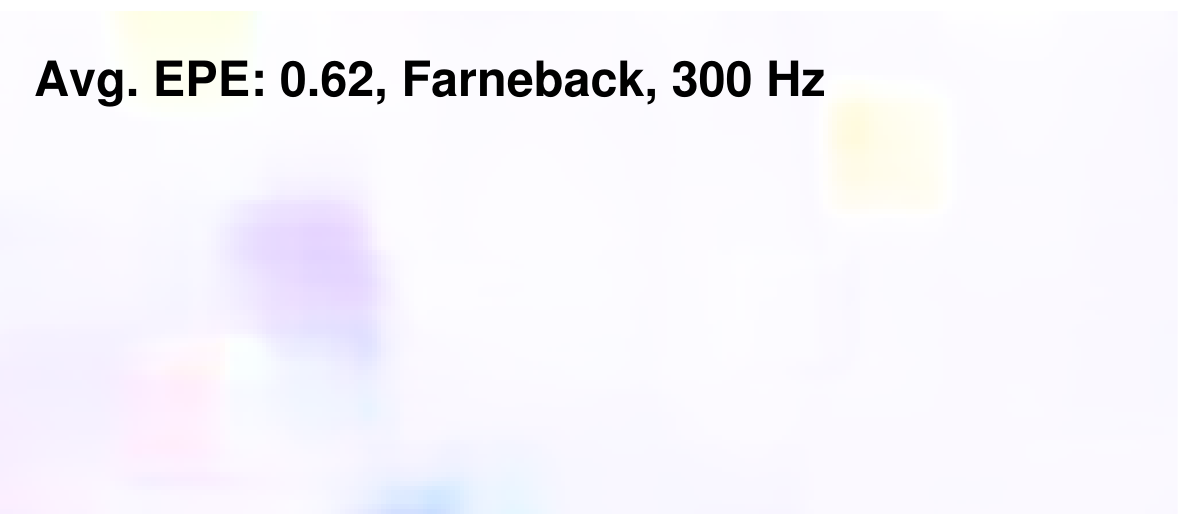}&
\includegraphics[width=0.195\textwidth]{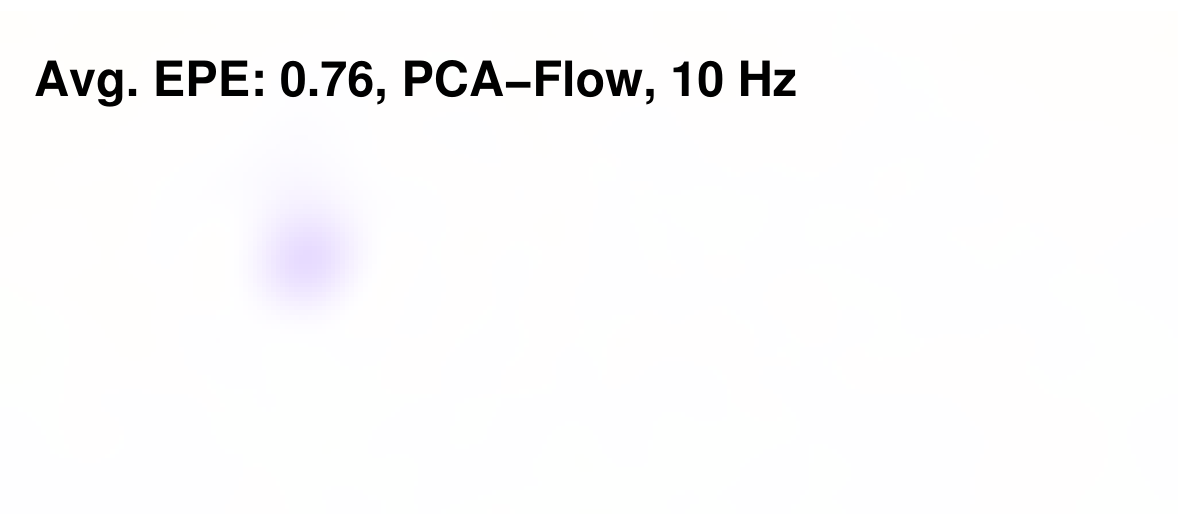}&
\includegraphics[width=0.195\textwidth]{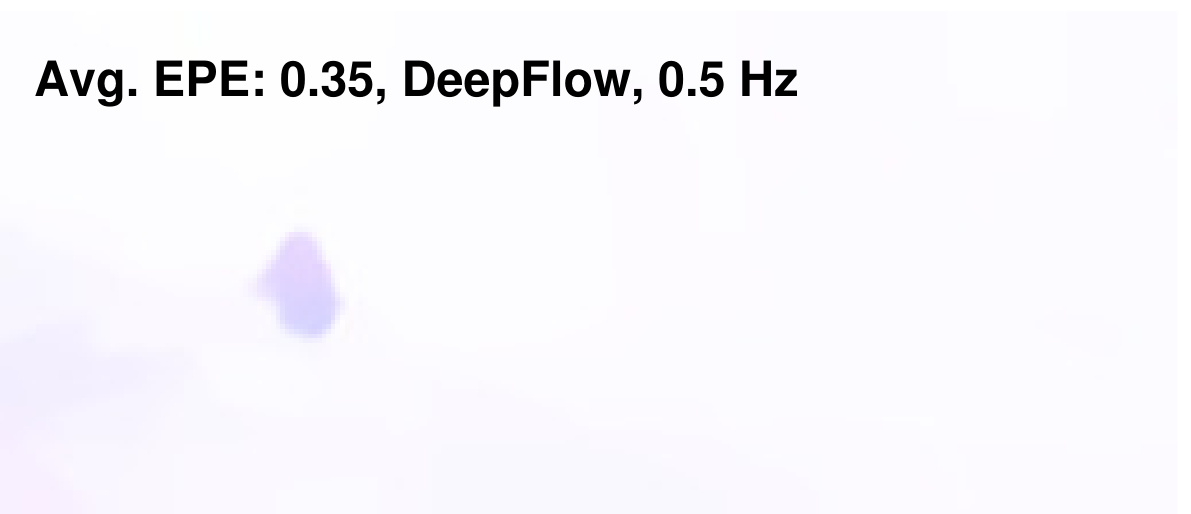}&
\includegraphics[width=0.195\textwidth]{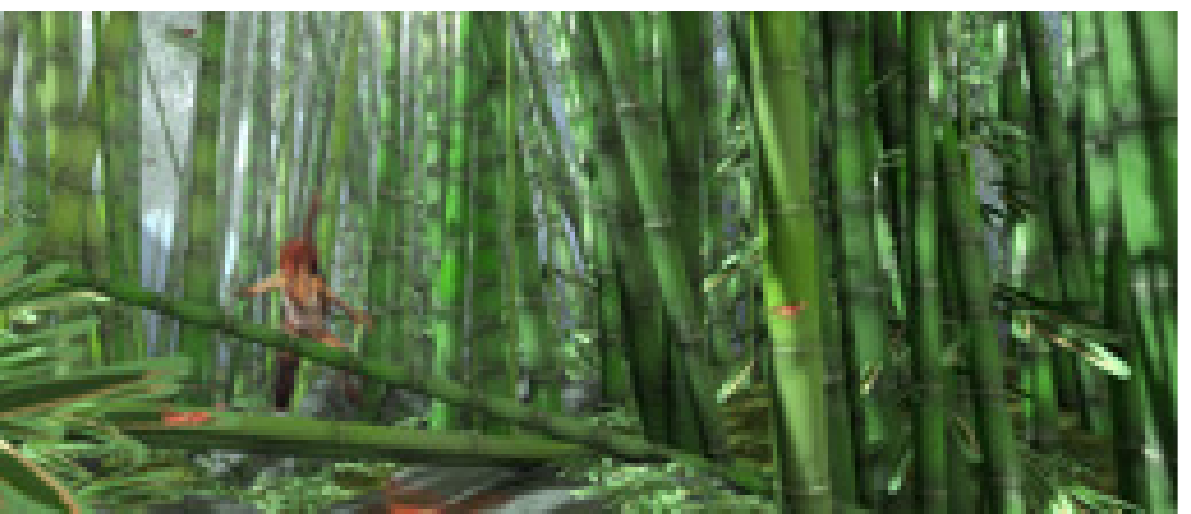}\\[6pt]
\includegraphics[width=0.195\textwidth]{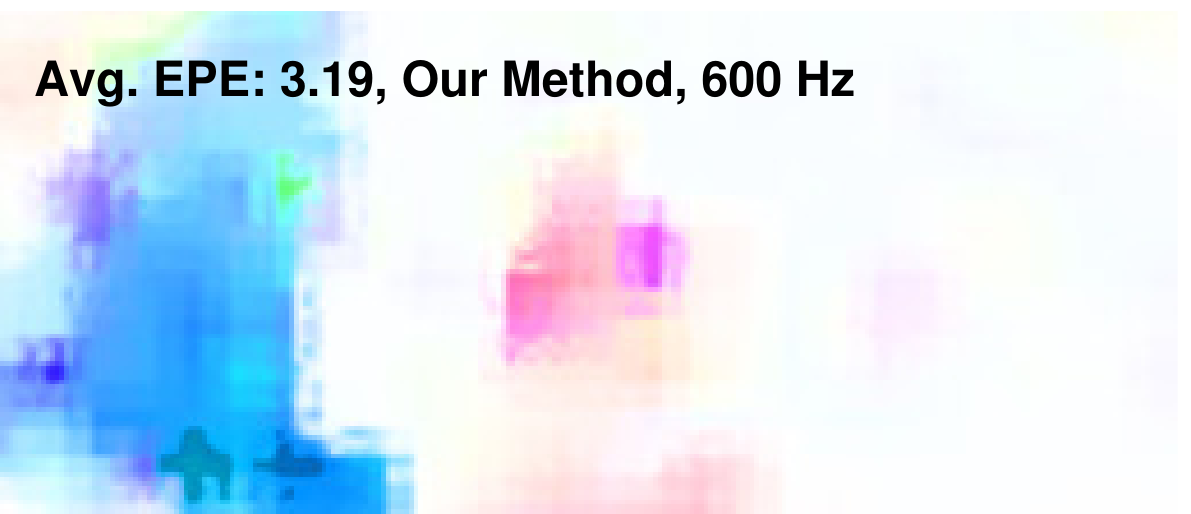}&
\includegraphics[width=0.195\textwidth]{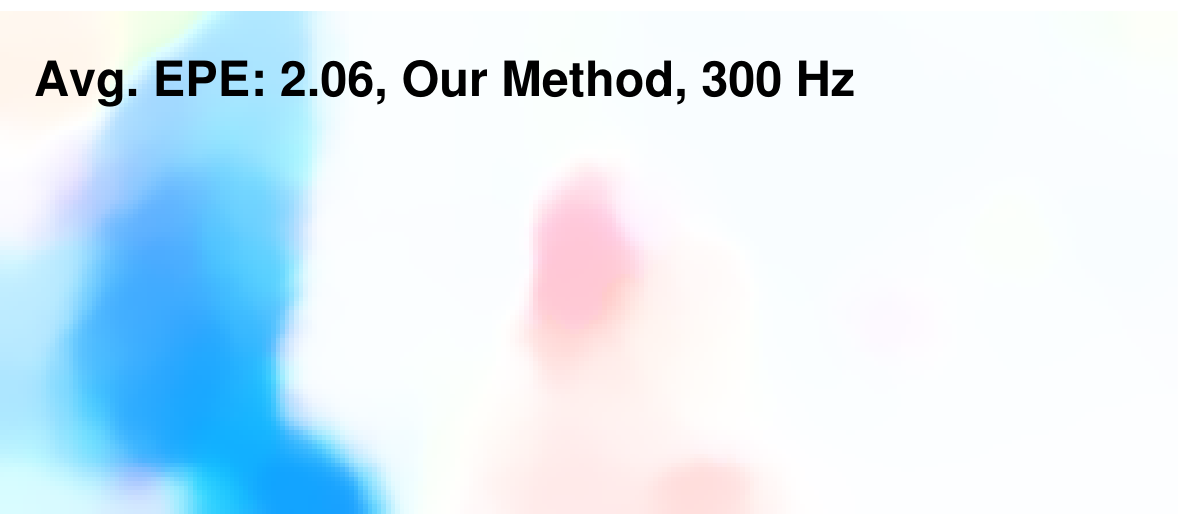}&
\includegraphics[width=0.195\textwidth]{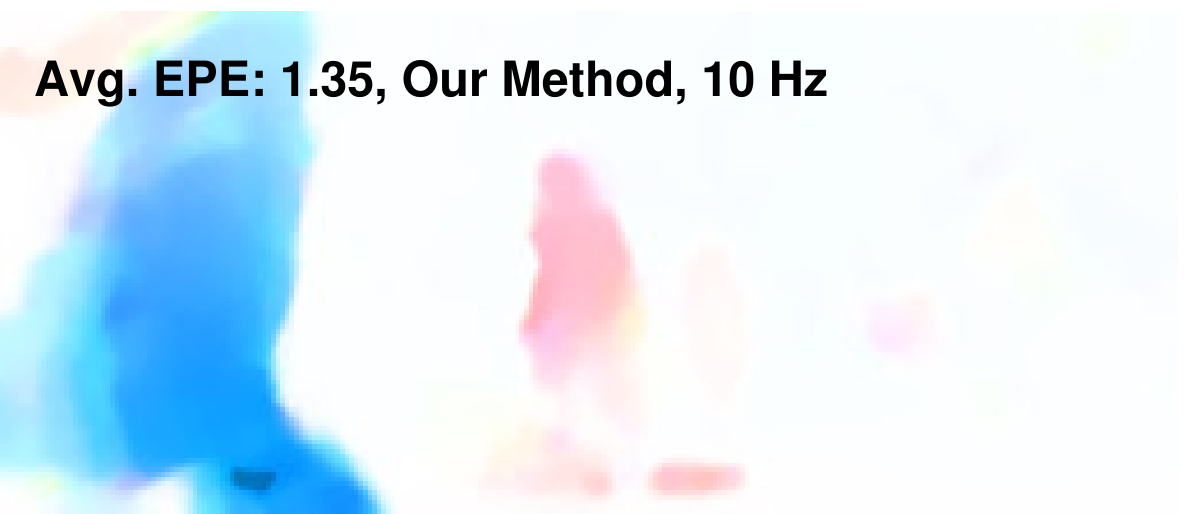}&
\includegraphics[width=0.195\textwidth]{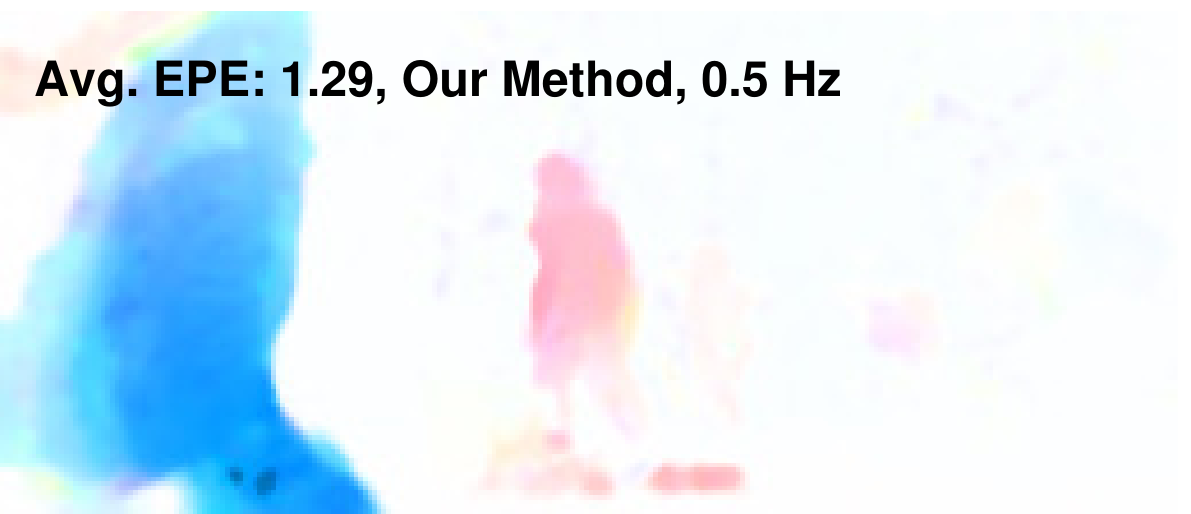}&
\includegraphics[width=0.195\textwidth]{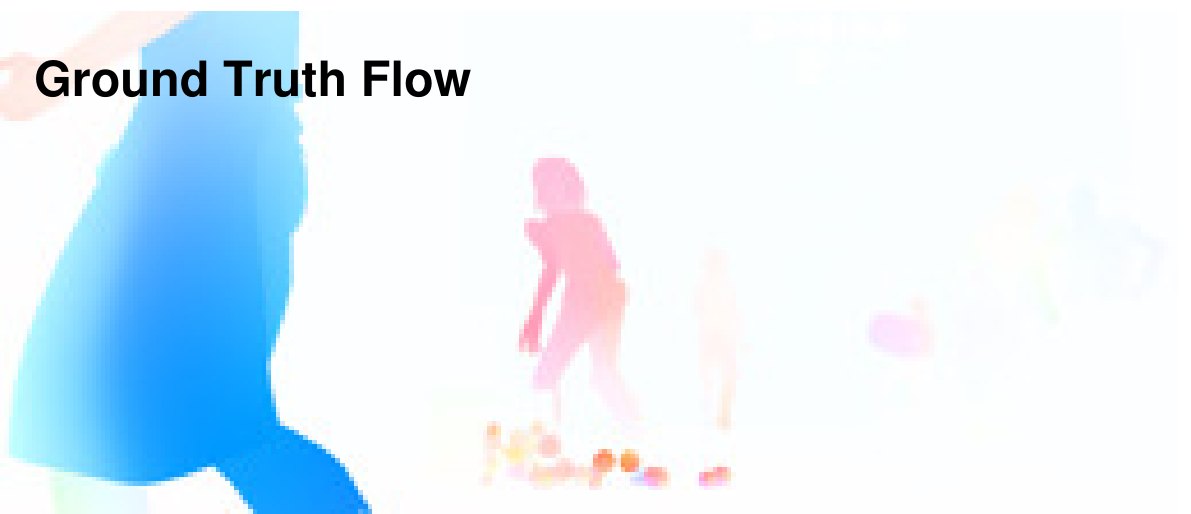}\\
\includegraphics[width=0.195\textwidth]{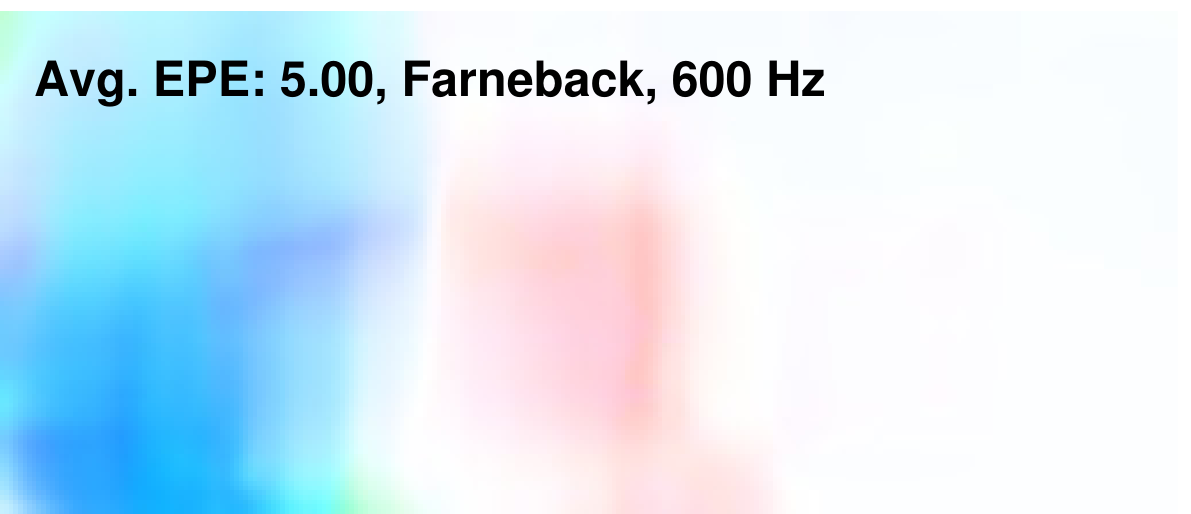}&
\includegraphics[width=0.195\textwidth]{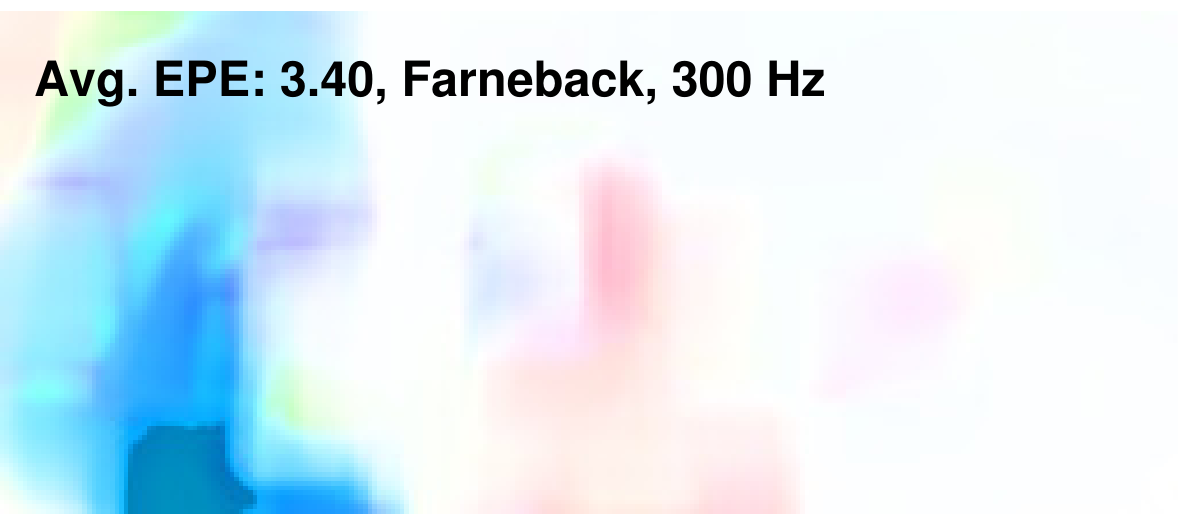}&
\includegraphics[width=0.195\textwidth]{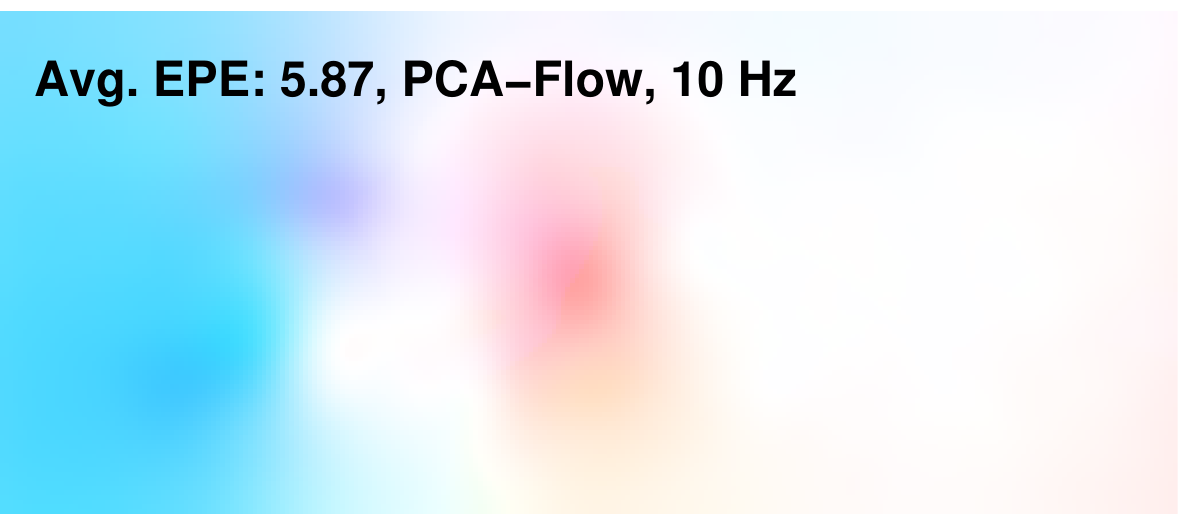}&
\includegraphics[width=0.195\textwidth]{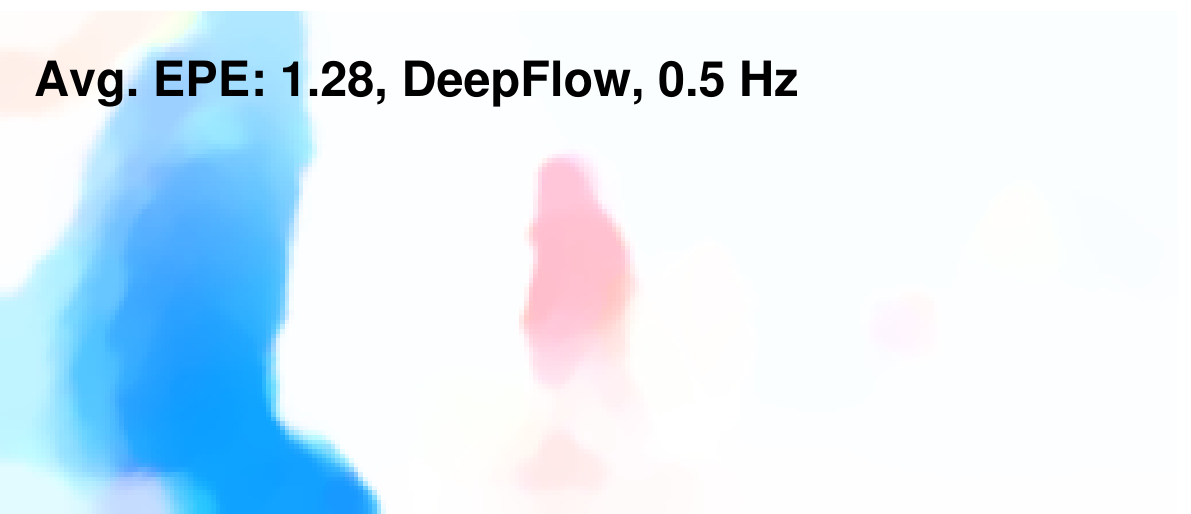}&
\includegraphics[width=0.195\textwidth]{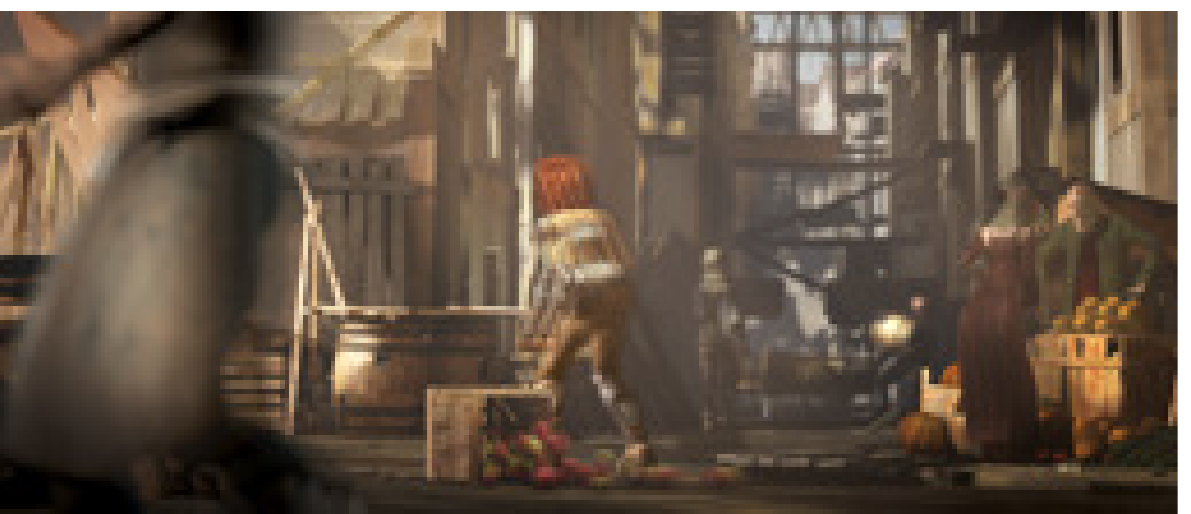}\\[6pt]
\end{tabular}
\caption{Exemplary results on Sintel (training). In each block of $2 \times 6$ images.  Top row, left to right: Our method for operating points ({\bf 1})-({\bf 4}), Ground Truth. Bottom row: Farneback 600Hz, Farneback 300Hz, PCA-Flow 10Hz, DeepFlow 0.5Hz, Original Image. See Fig. \ref{fig:sintel1res_errmap_AP} for error maps.}\label{fig:sintel1res_AP} \end{figure*}

\begin{figure*} [!ht]
\centering\setlength{\tabcolsep}{0.1pt}\renewcommand{\arraystretch}{0} 
{
\begin{tabular}{ccccc}
 {\bf 600Hz} & {\bf 300Hz} & {\bf 10Hz} & {\bf 0.5Hz}& {\bf Ground Truth}\\
\includegraphics[width=0.195\textwidth]{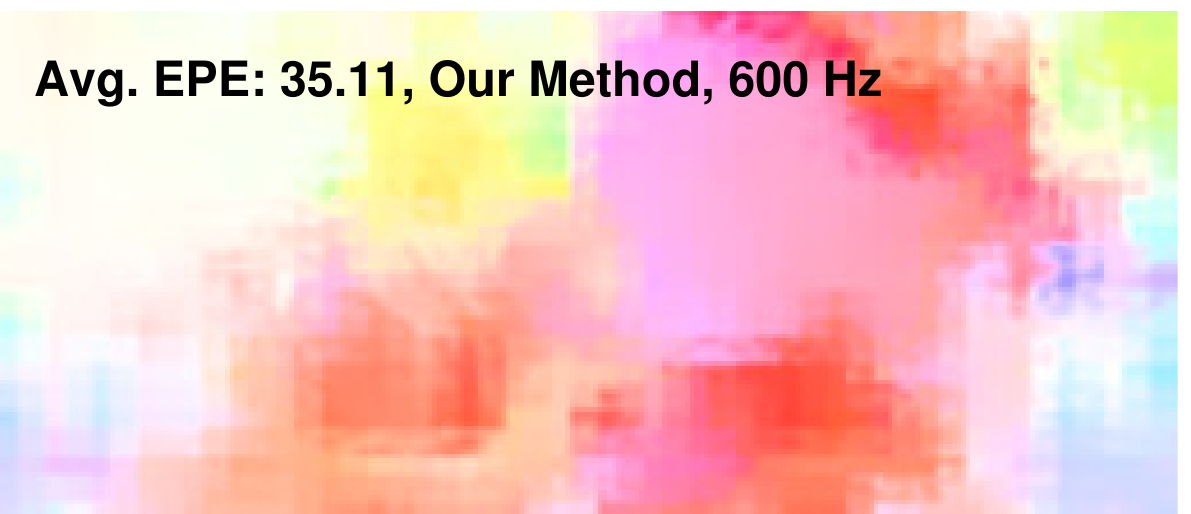}&
\includegraphics[width=0.195\textwidth]{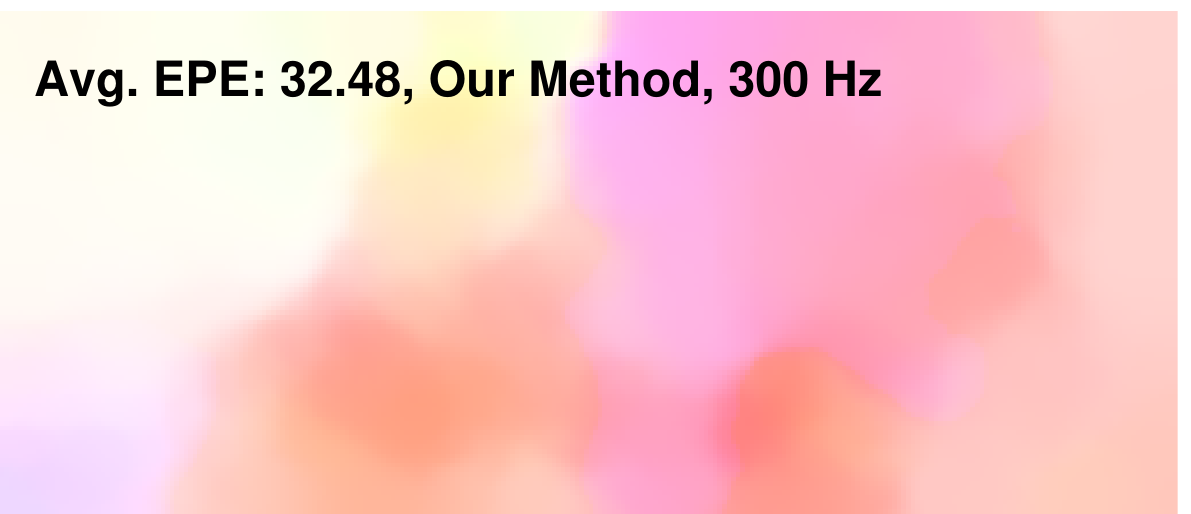}&
\includegraphics[width=0.195\textwidth]{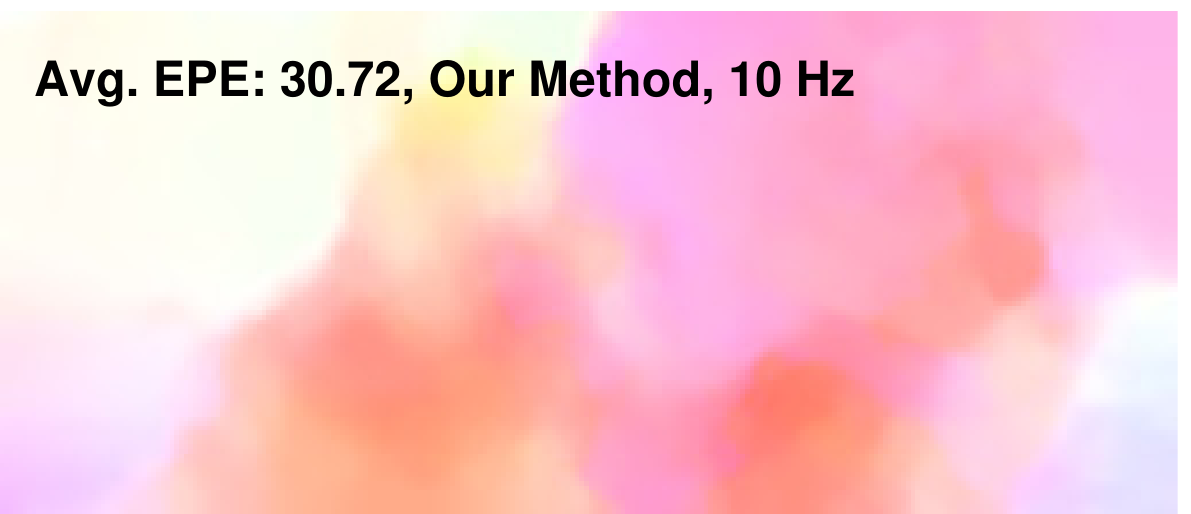}&
\includegraphics[width=0.195\textwidth]{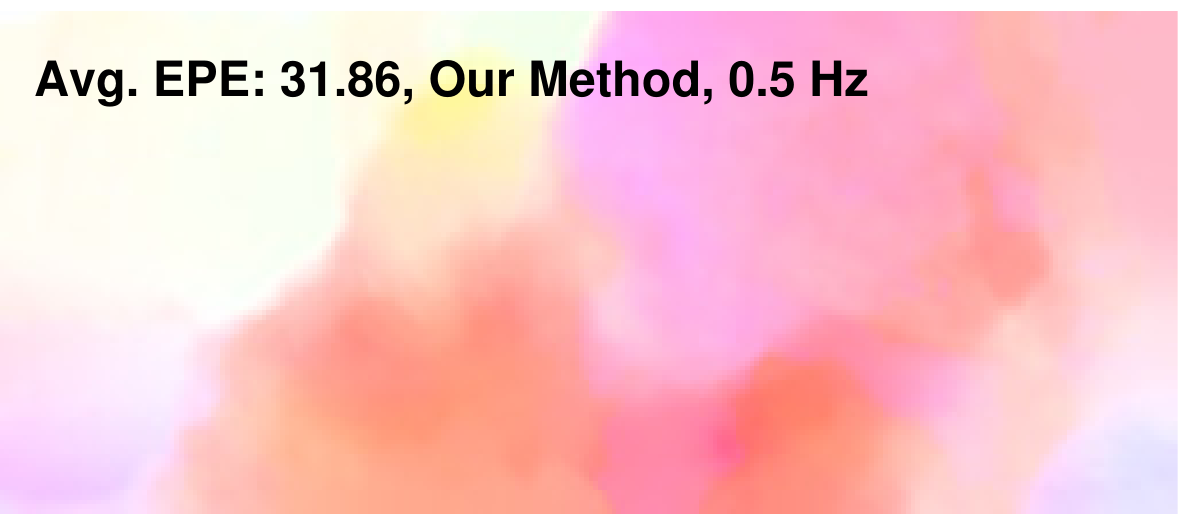}&
\includegraphics[width=0.195\textwidth]{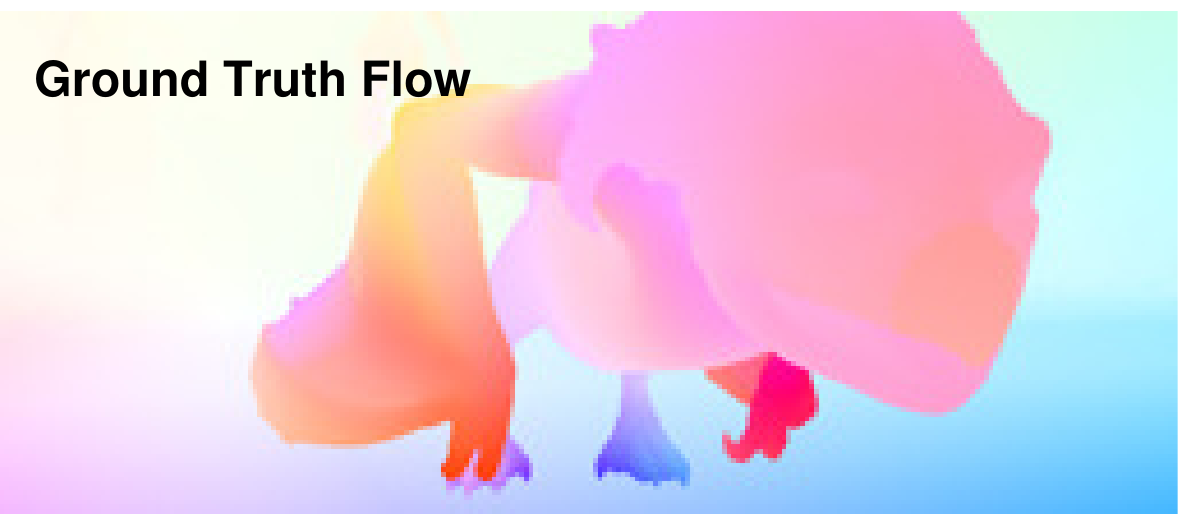}\\
\includegraphics[width=0.195\textwidth]{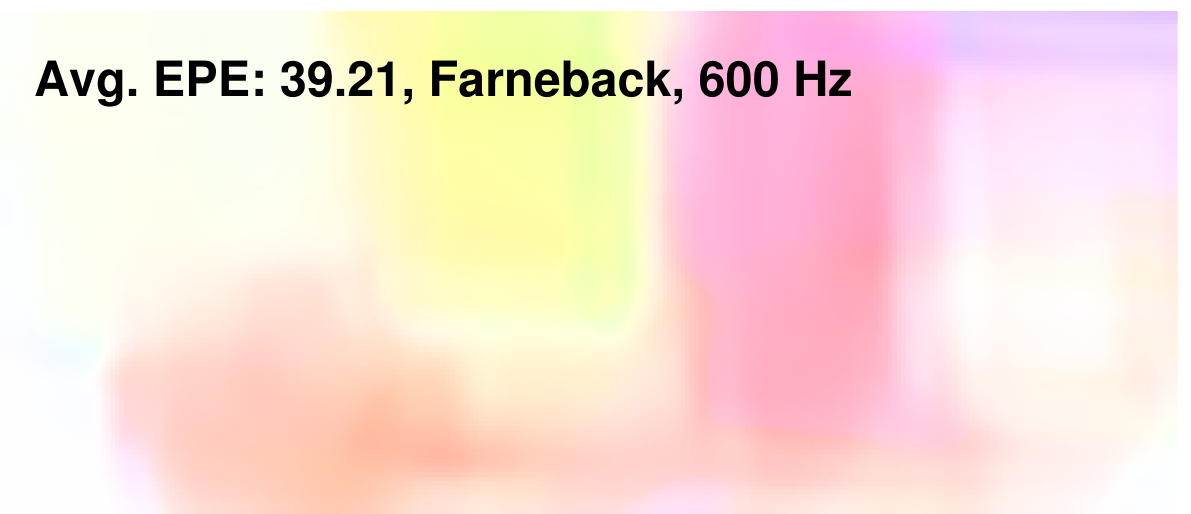}&
\includegraphics[width=0.195\textwidth]{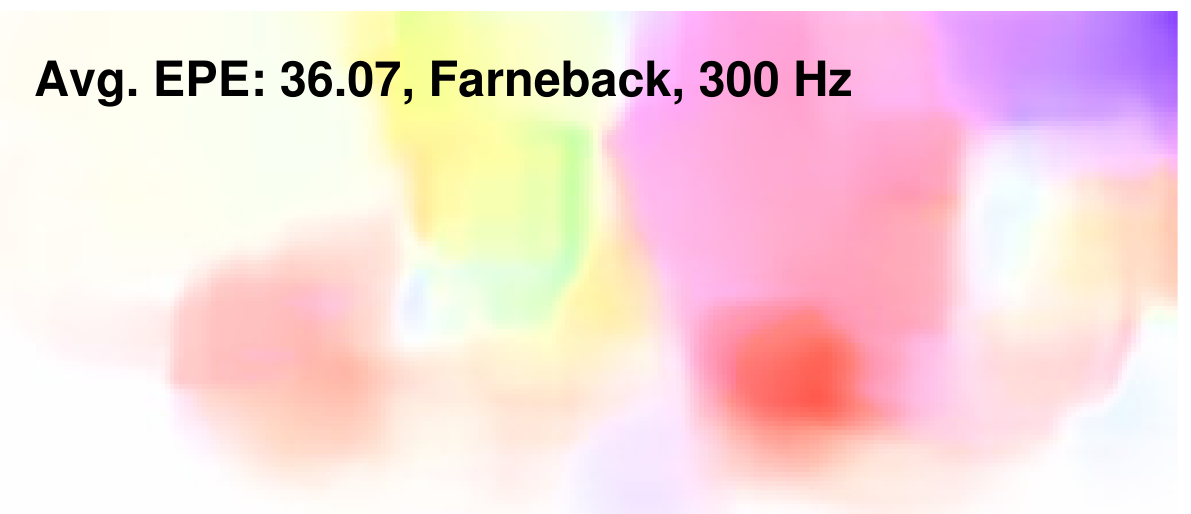}&
\includegraphics[width=0.195\textwidth]{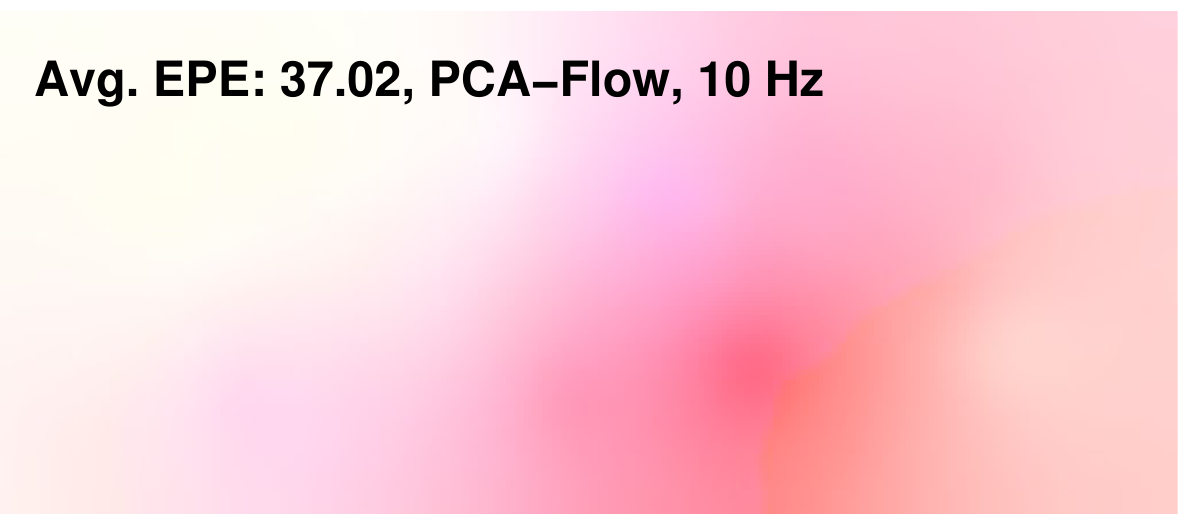}&
\includegraphics[width=0.195\textwidth]{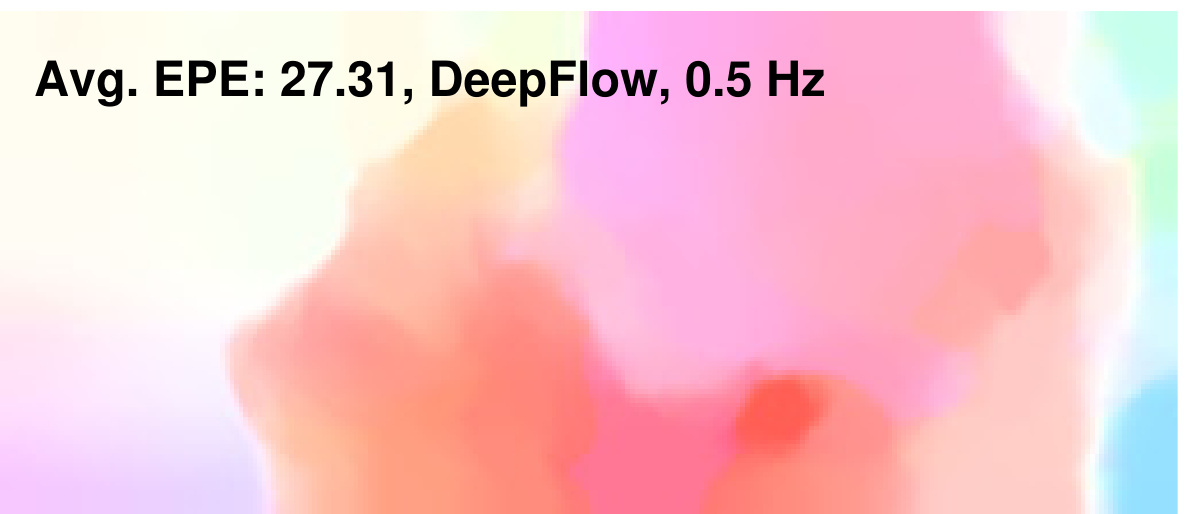}&
\includegraphics[width=0.195\textwidth]{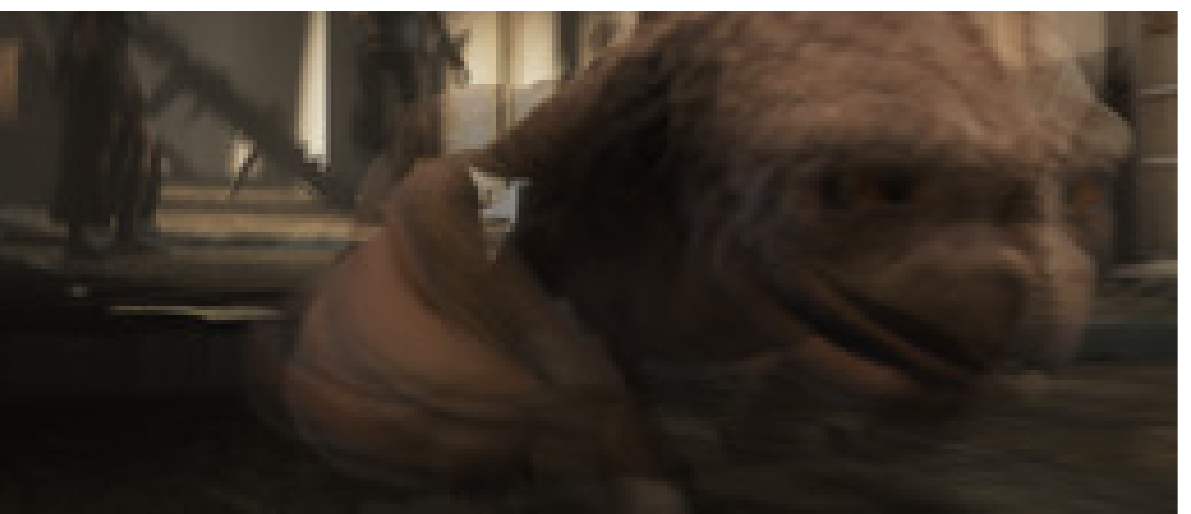}\\[6pt]
\includegraphics[width=0.195\textwidth]{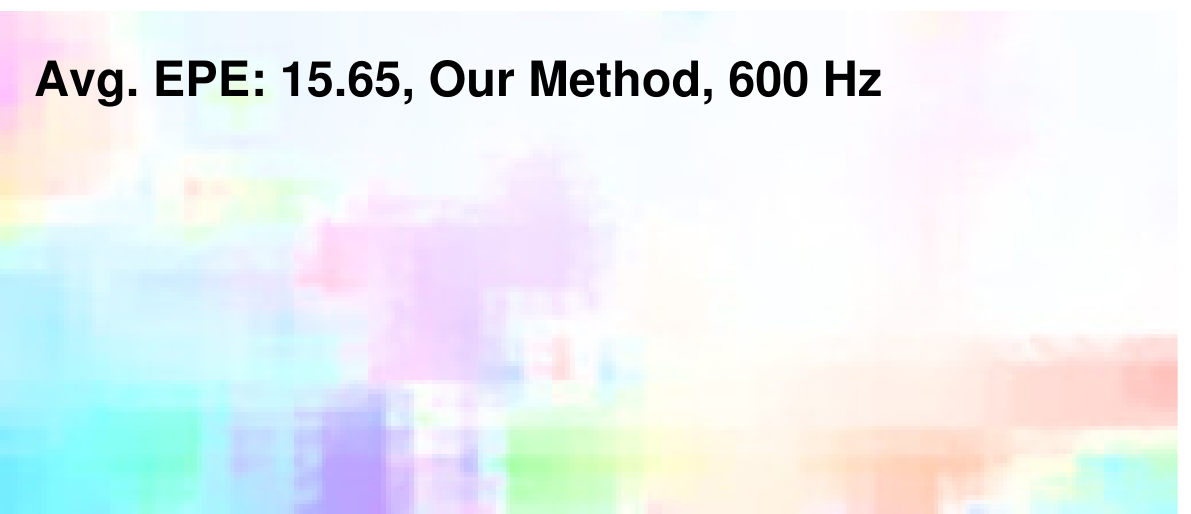}&
\includegraphics[width=0.195\textwidth]{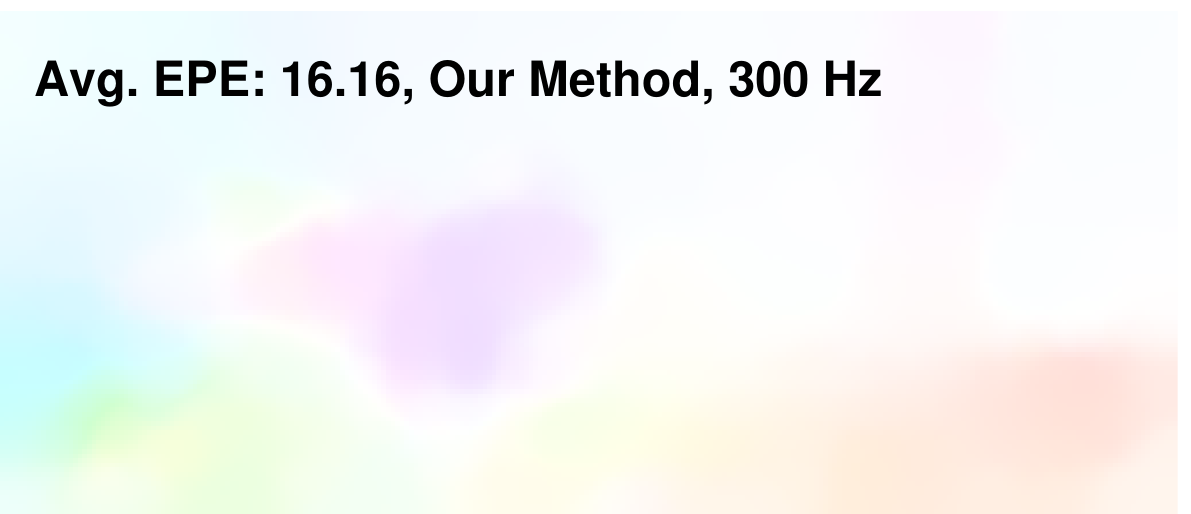}&
\includegraphics[width=0.195\textwidth]{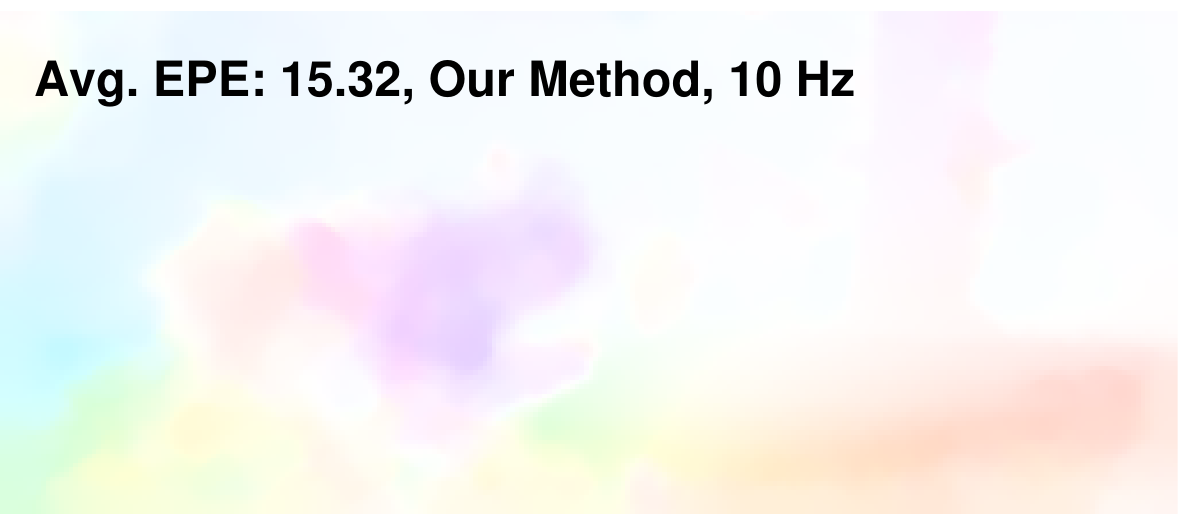}&
\includegraphics[width=0.195\textwidth]{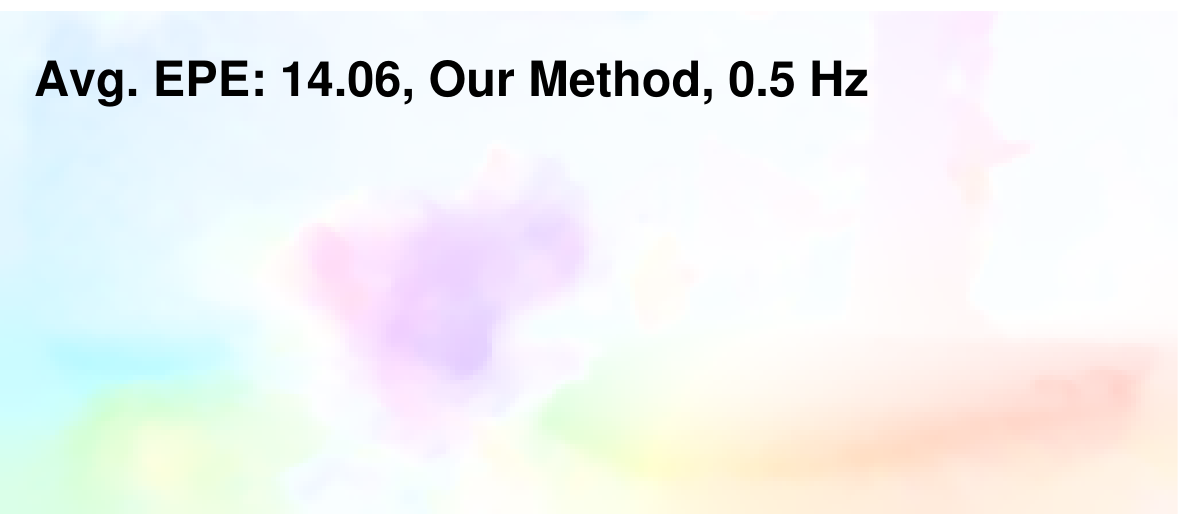}&
\includegraphics[width=0.195\textwidth]{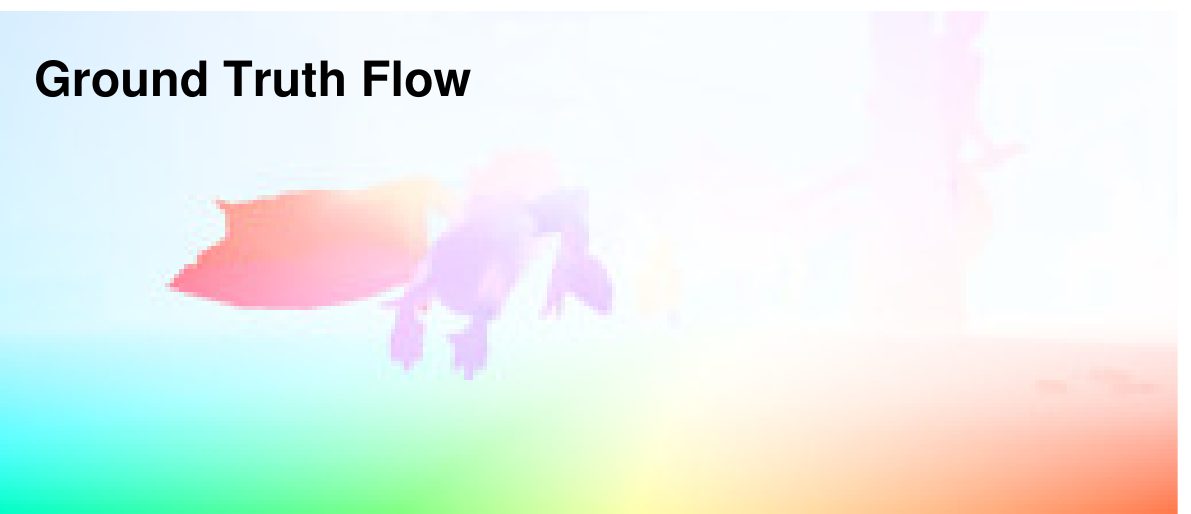}\\
\includegraphics[width=0.195\textwidth]{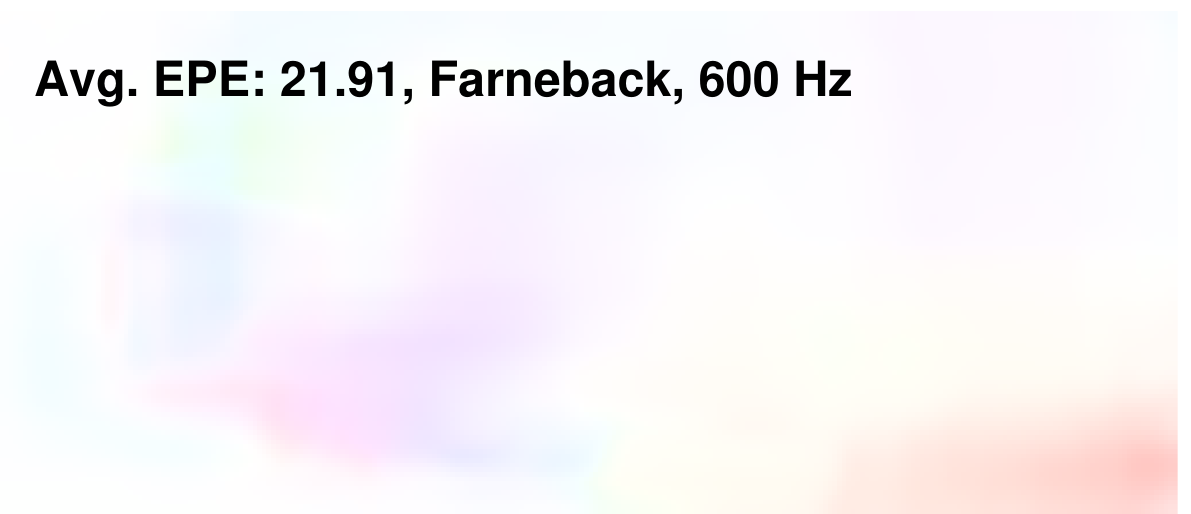}&
\includegraphics[width=0.195\textwidth]{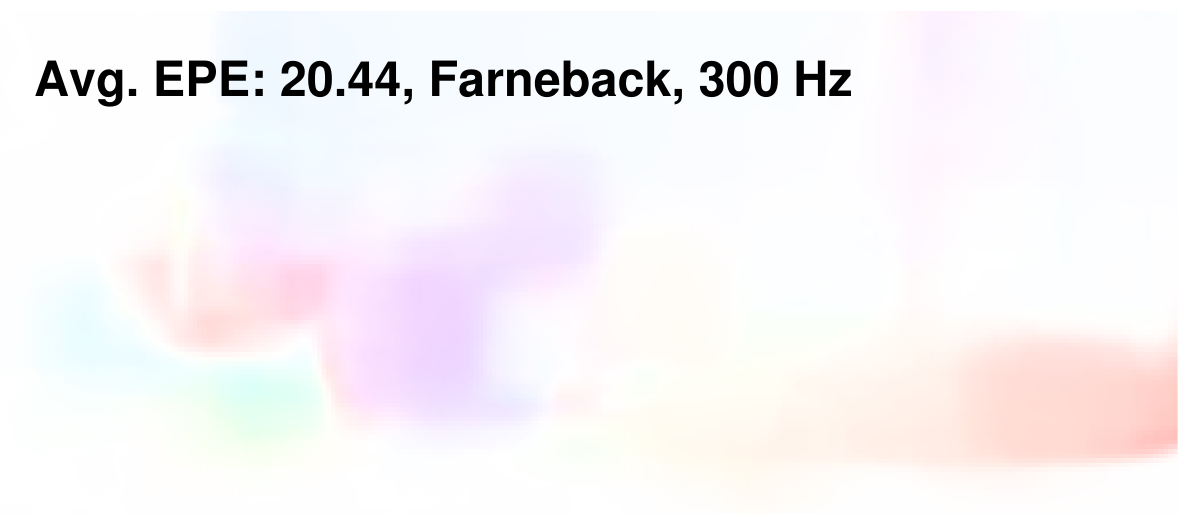}&
\includegraphics[width=0.195\textwidth]{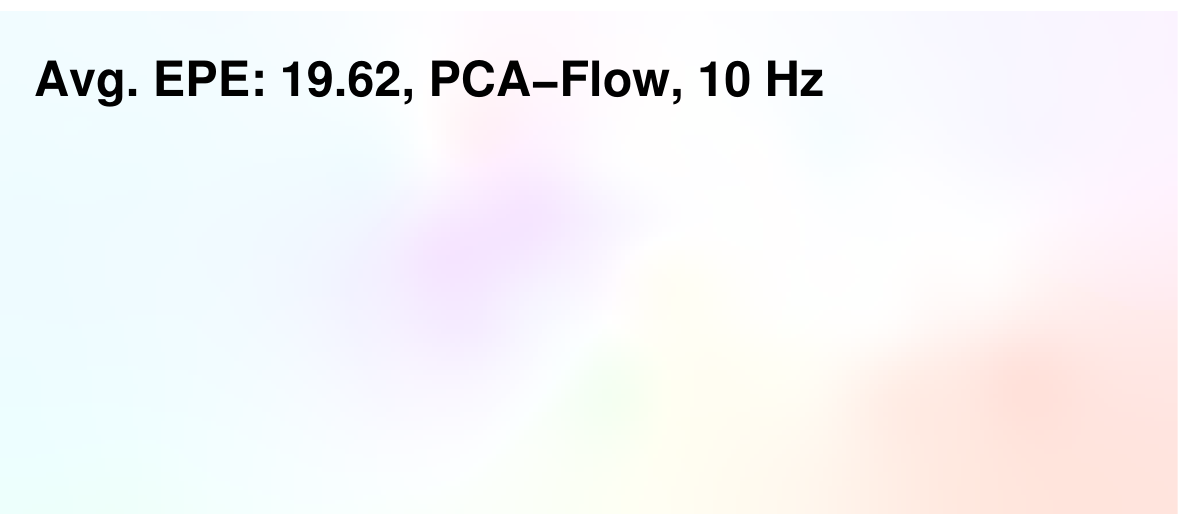}&
\includegraphics[width=0.195\textwidth]{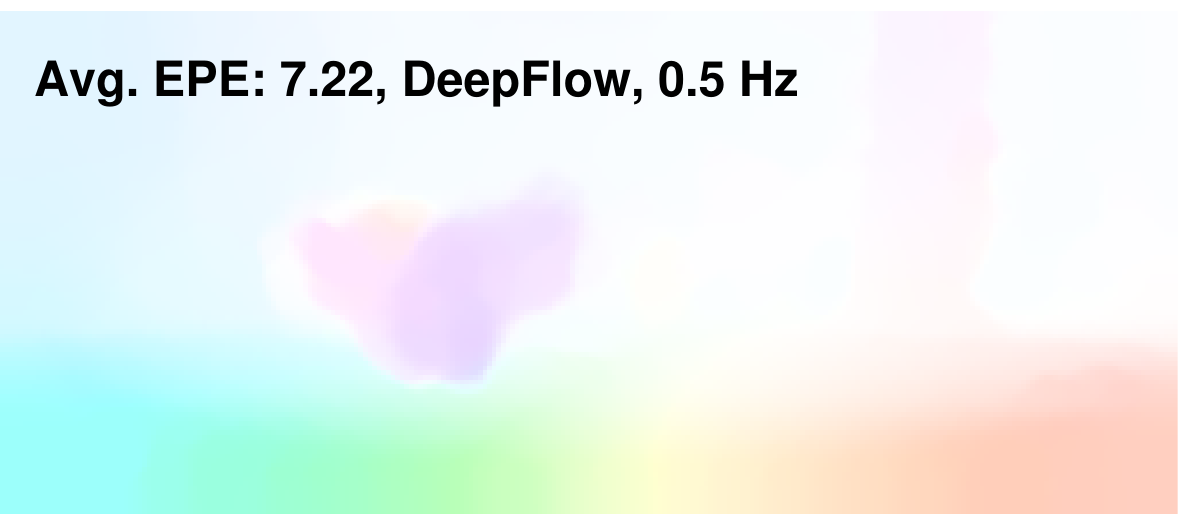}&
\includegraphics[width=0.195\textwidth]{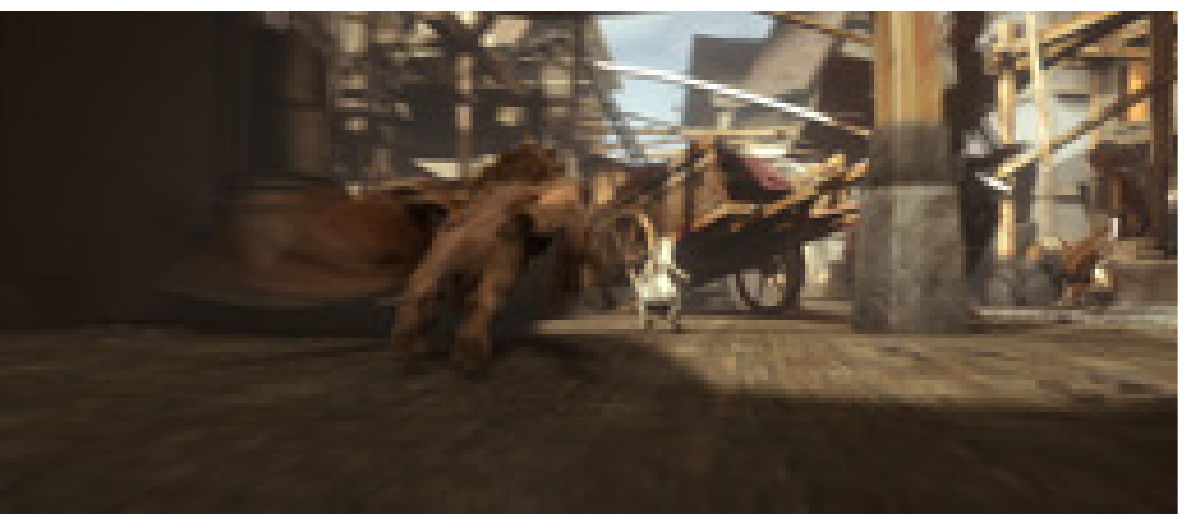}\\[6pt]
\includegraphics[width=0.195\textwidth]{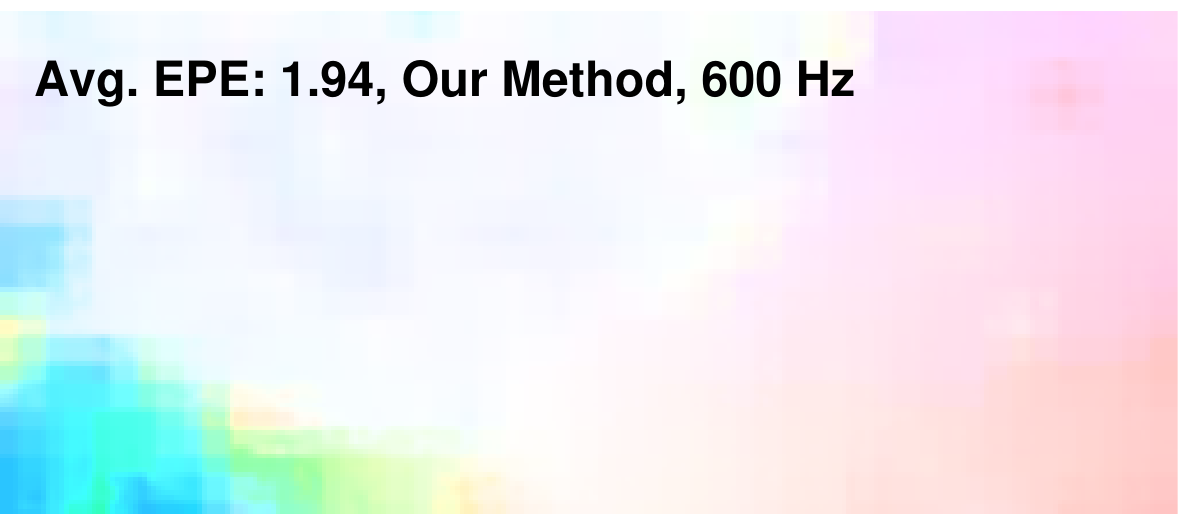}&
\includegraphics[width=0.195\textwidth]{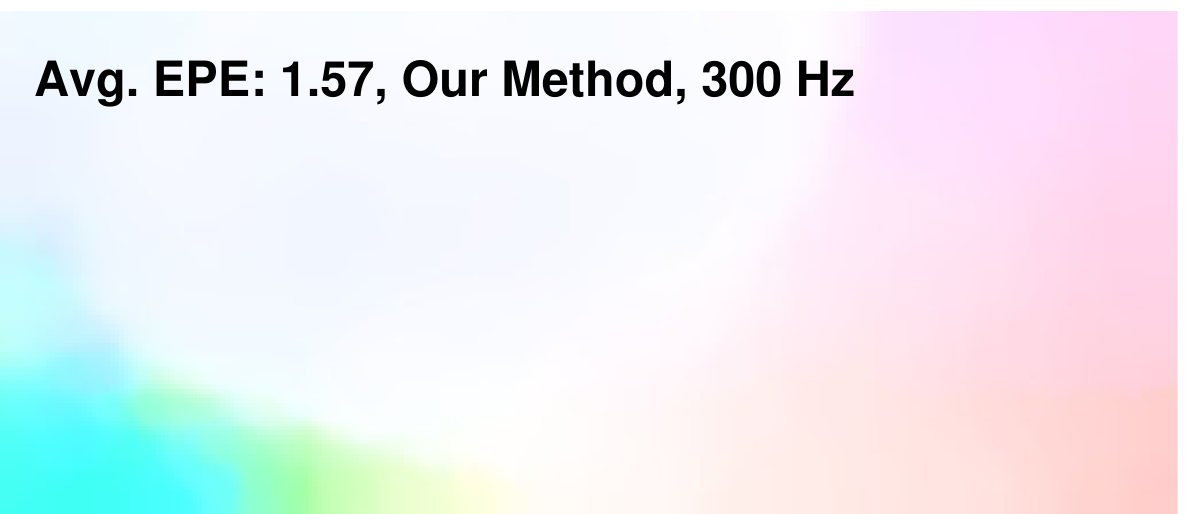}&
\includegraphics[width=0.195\textwidth]{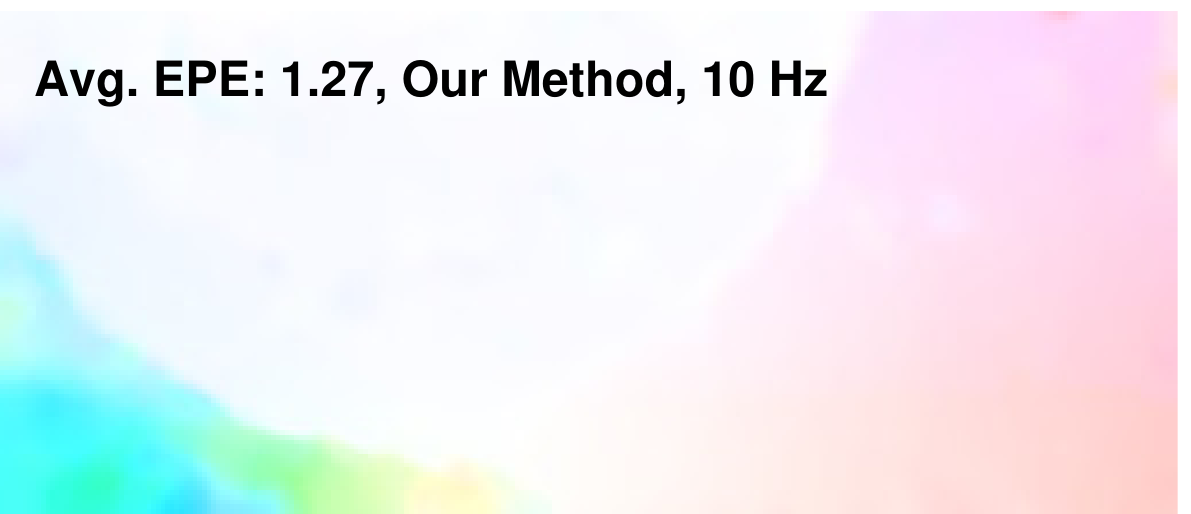}&
\includegraphics[width=0.195\textwidth]{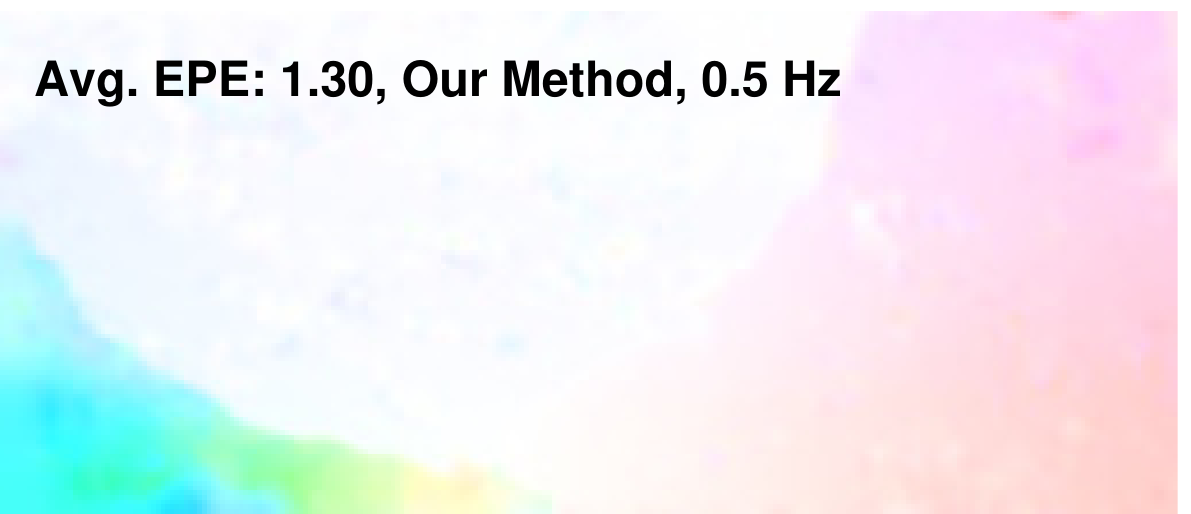}&
\includegraphics[width=0.195\textwidth]{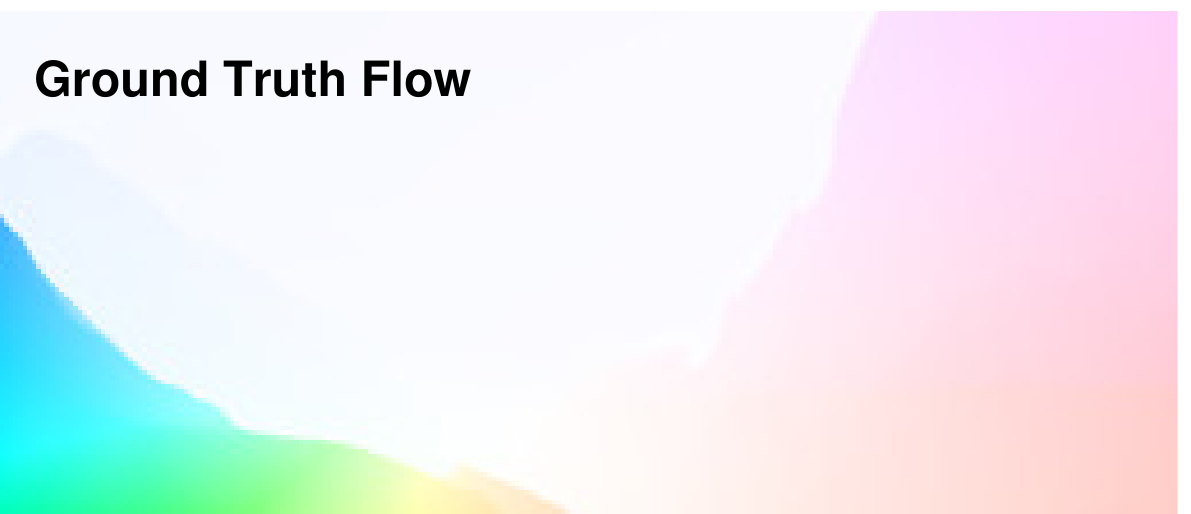}\\
\includegraphics[width=0.195\textwidth]{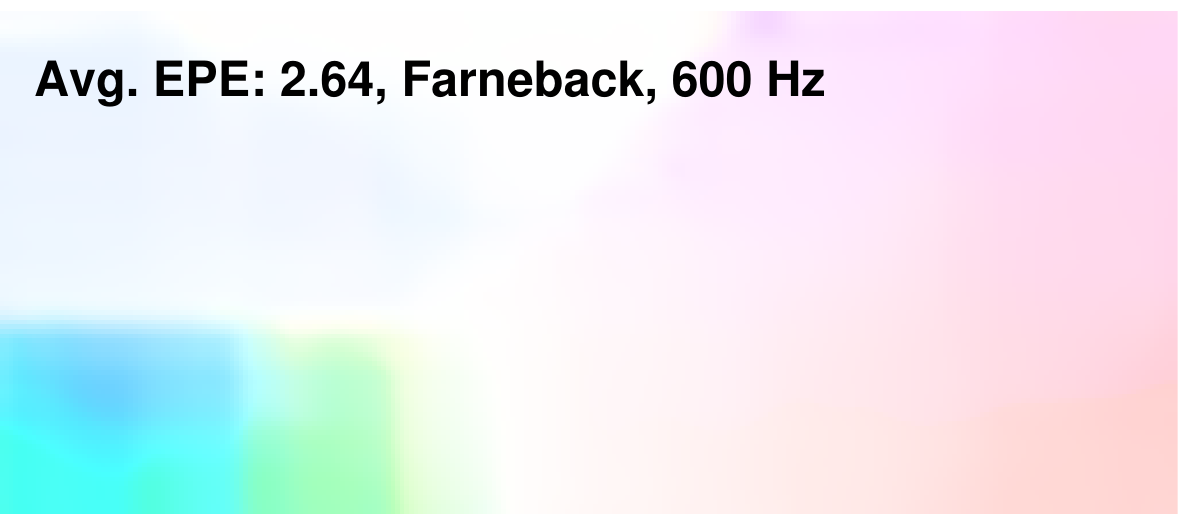}&
\includegraphics[width=0.195\textwidth]{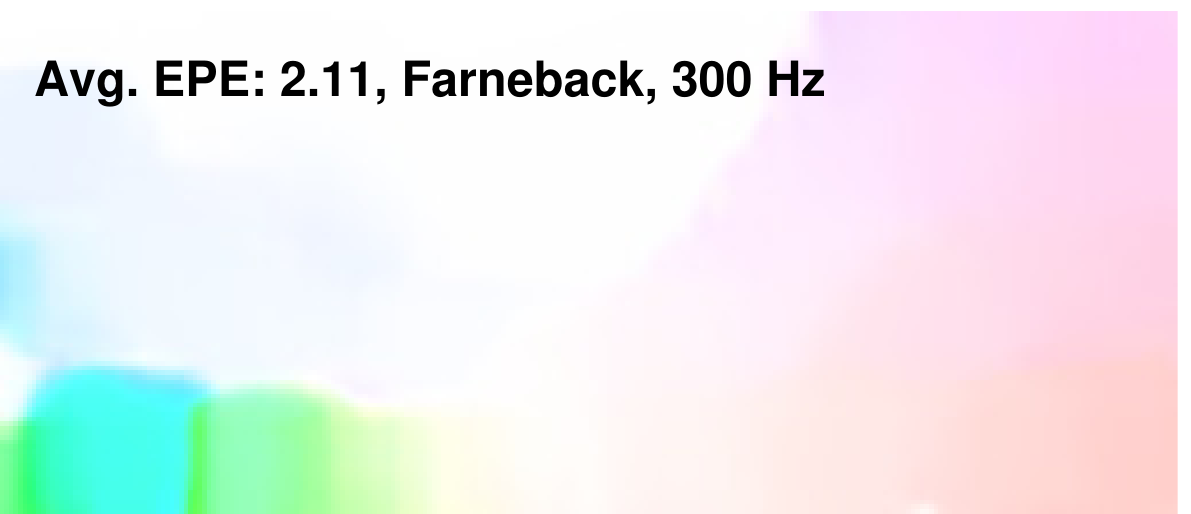}&
\includegraphics[width=0.195\textwidth]{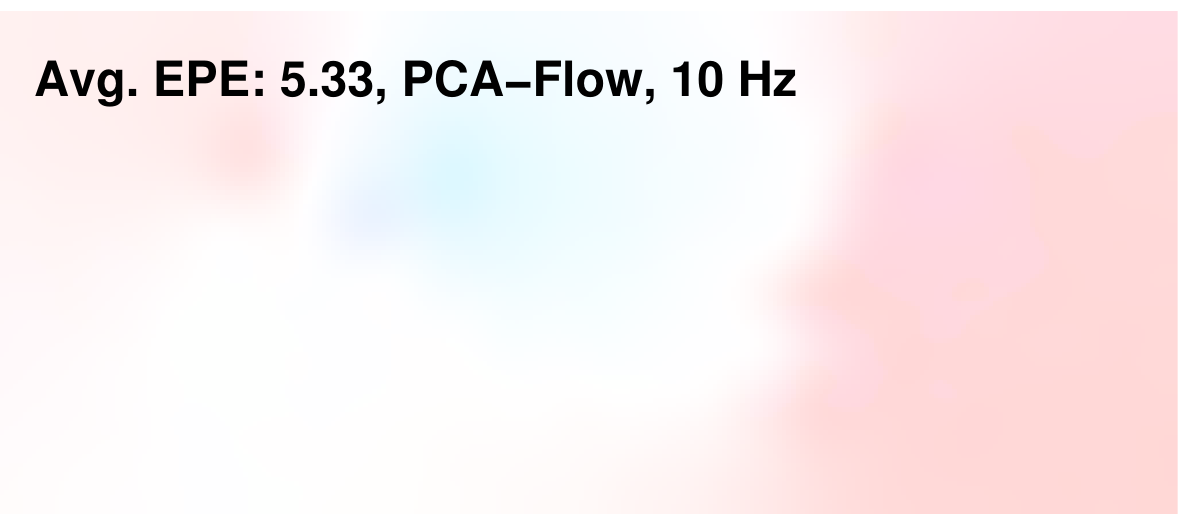}&
\includegraphics[width=0.195\textwidth]{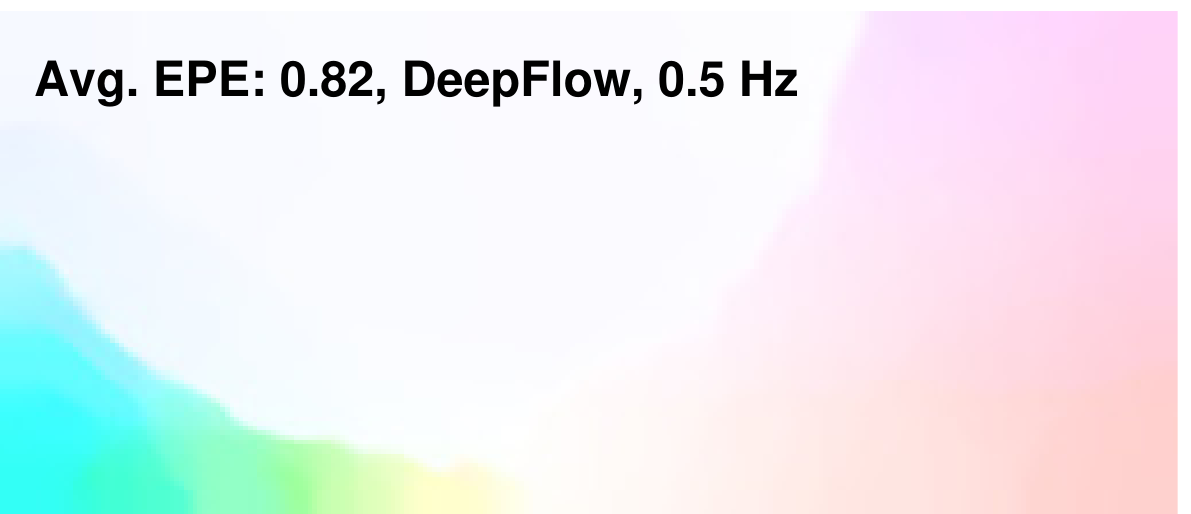}&
\includegraphics[width=0.195\textwidth]{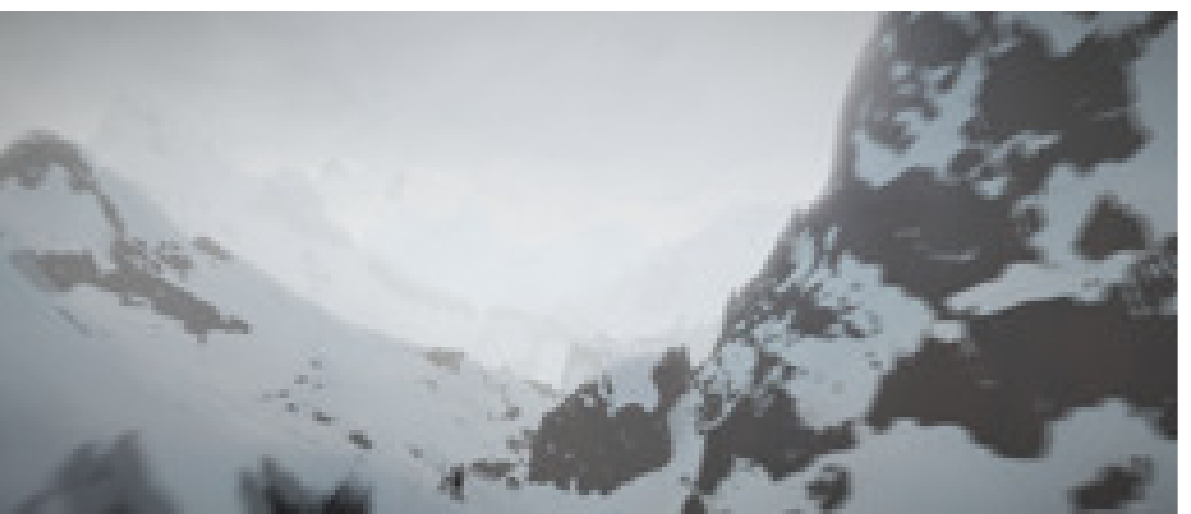}\\[6pt]
\includegraphics[width=0.195\textwidth]{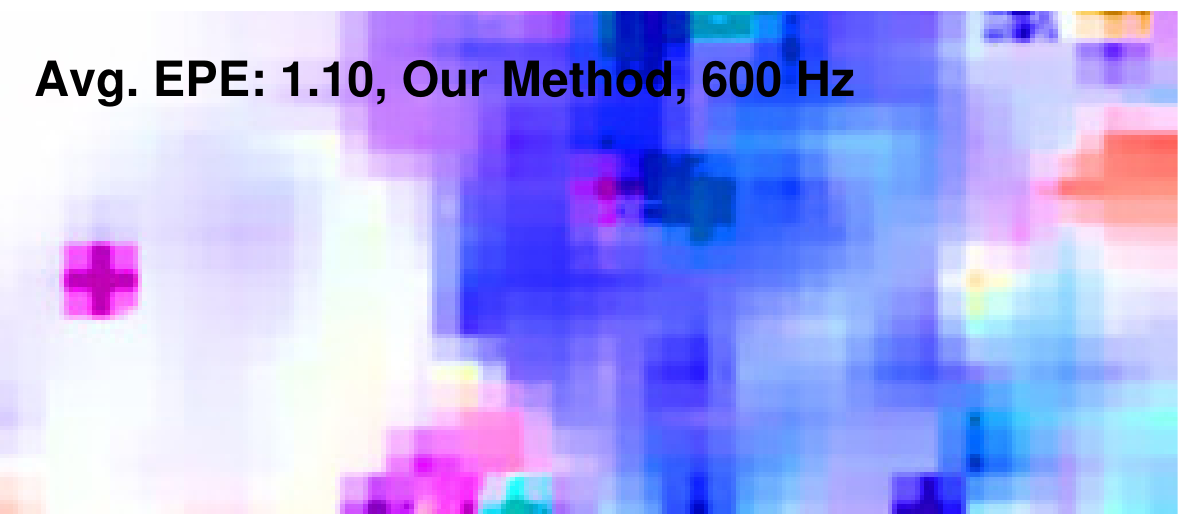}&
\includegraphics[width=0.195\textwidth]{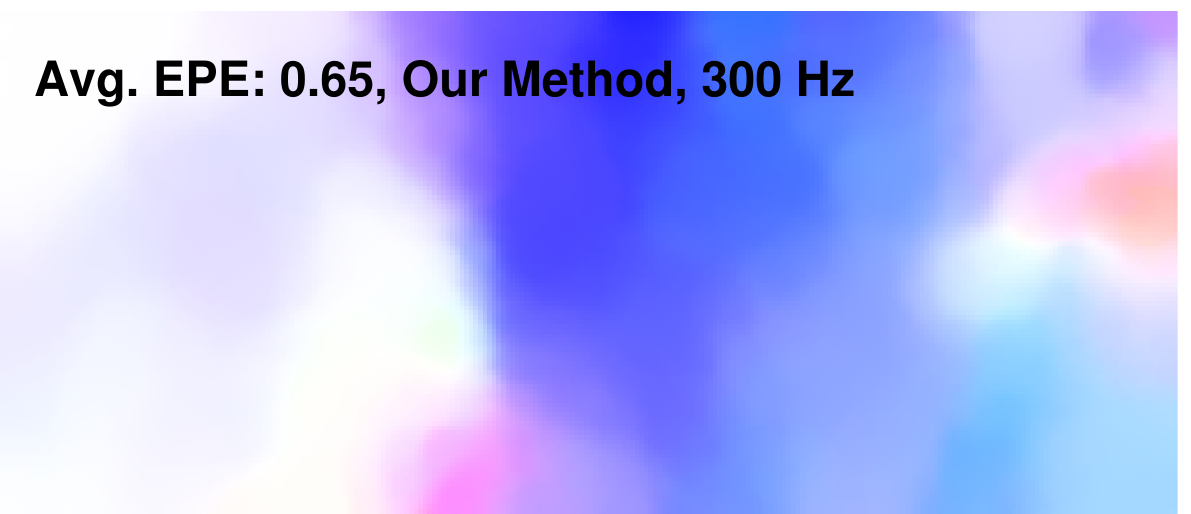}&
\includegraphics[width=0.195\textwidth]{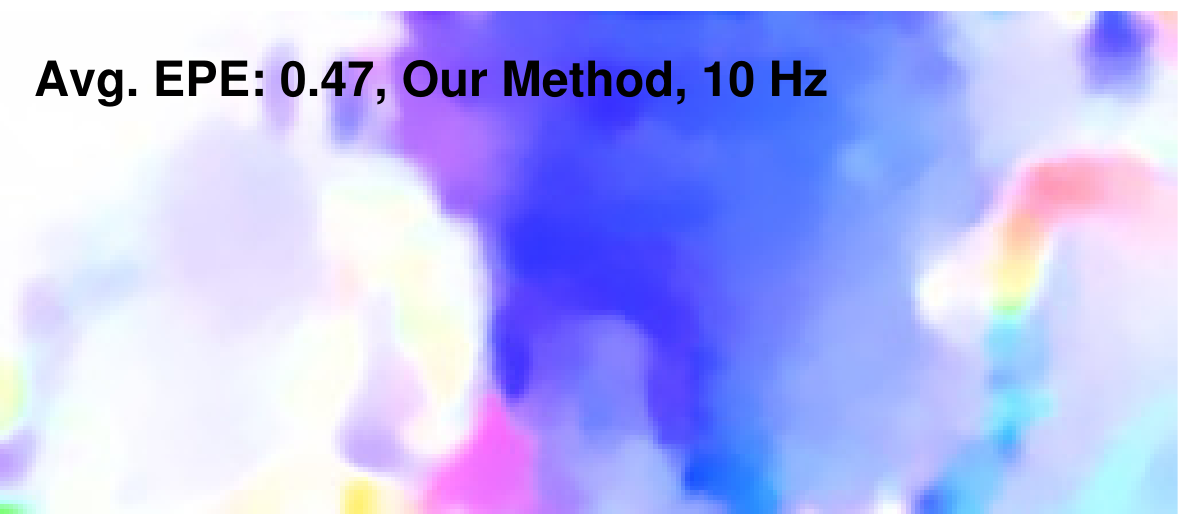}&
\includegraphics[width=0.195\textwidth]{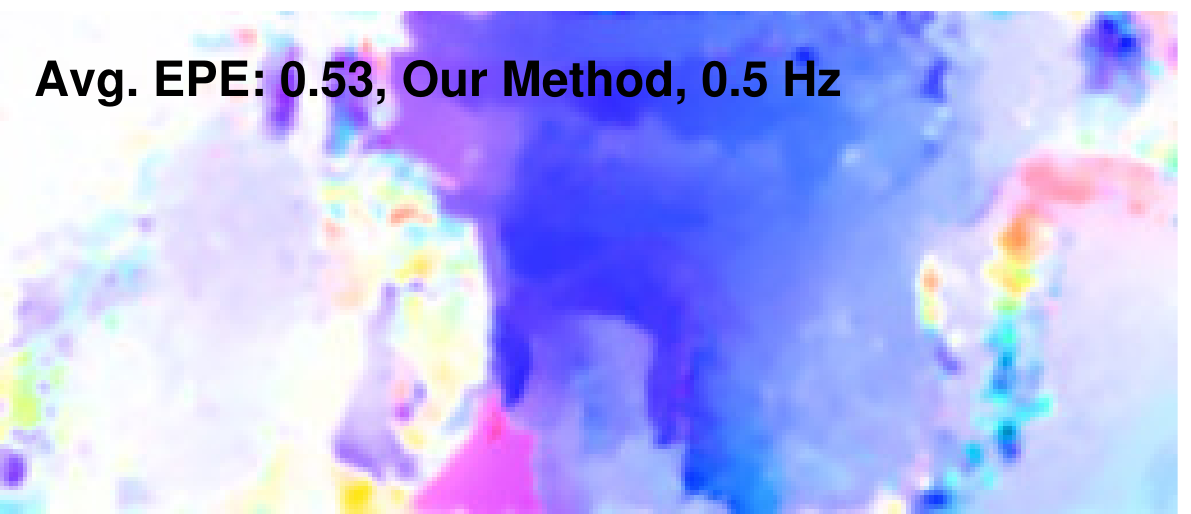}&
\includegraphics[width=0.195\textwidth]{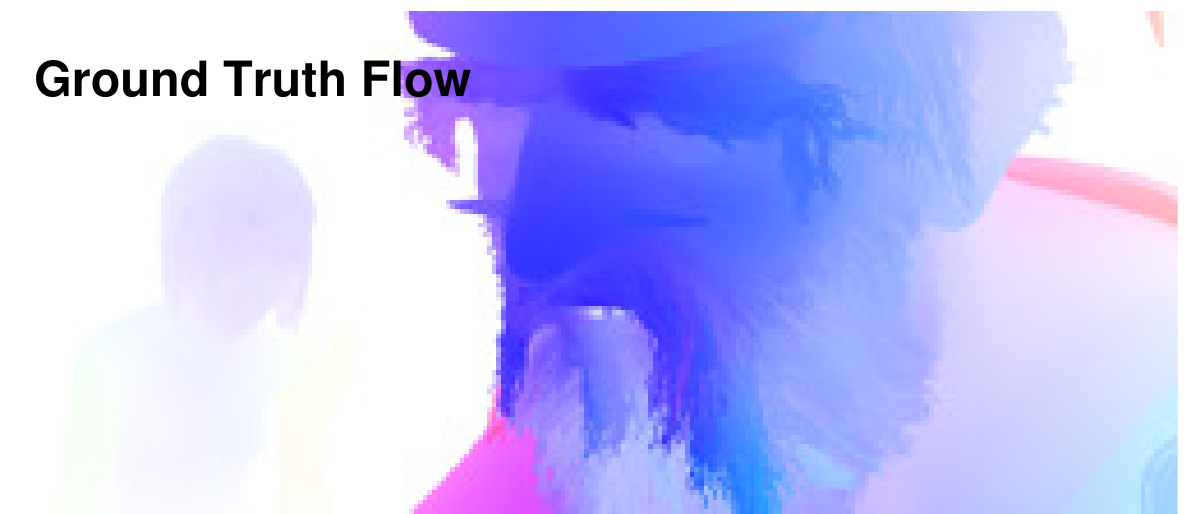}\\
\includegraphics[width=0.195\textwidth]{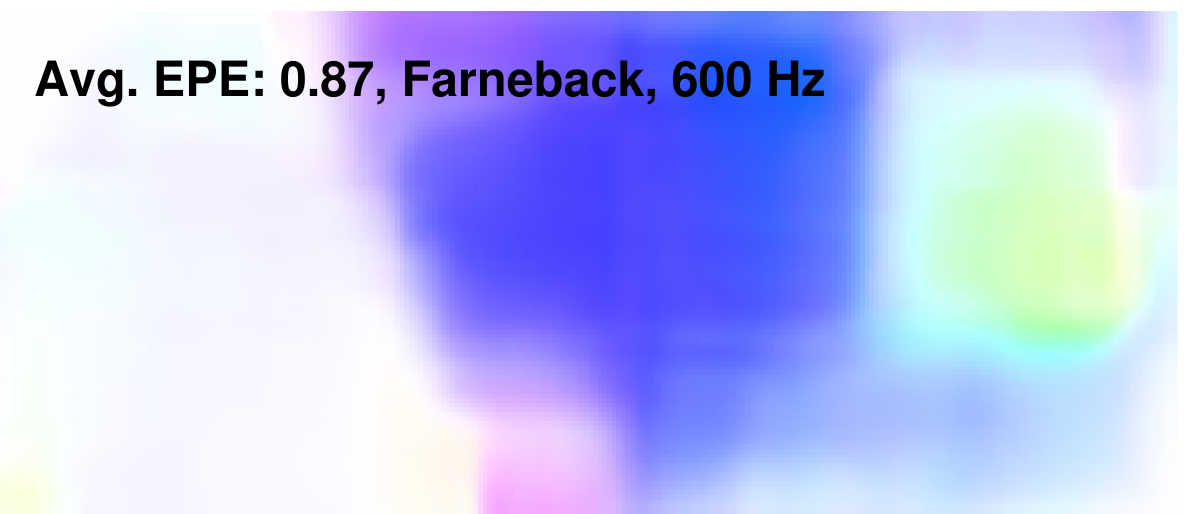}&
\includegraphics[width=0.195\textwidth]{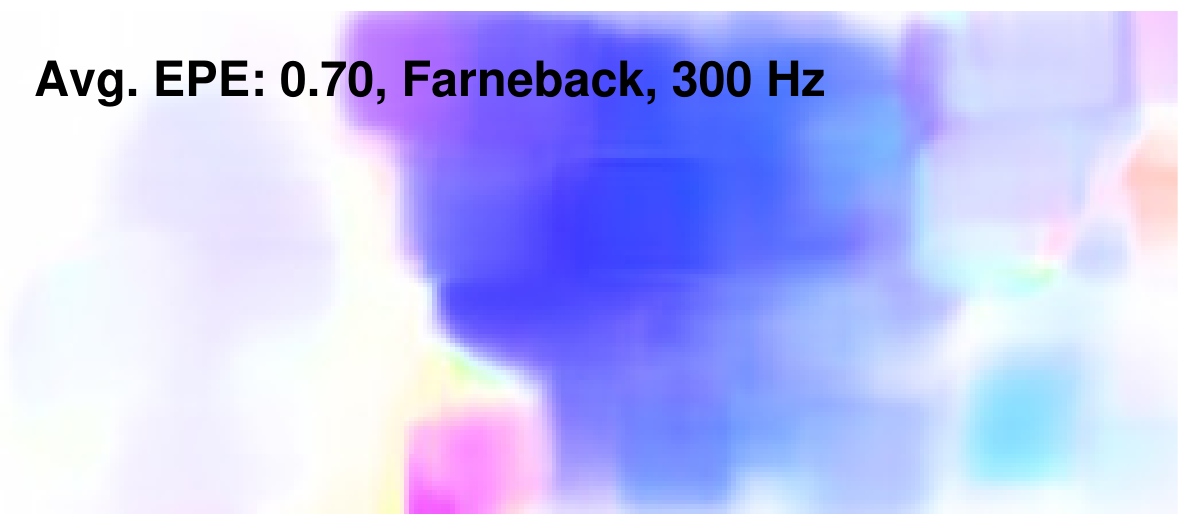}&
\includegraphics[width=0.195\textwidth]{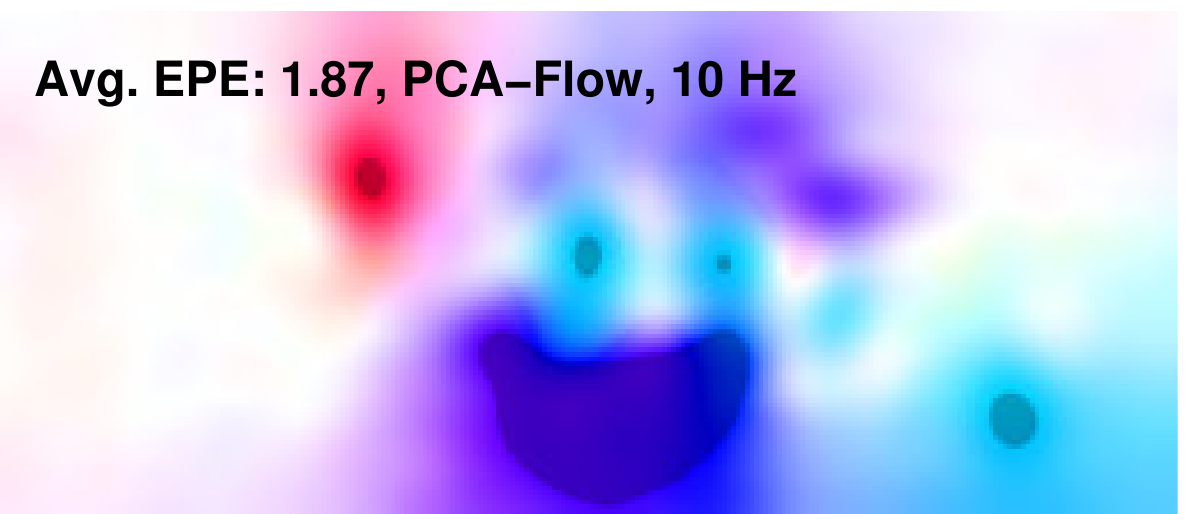}&
\includegraphics[width=0.195\textwidth]{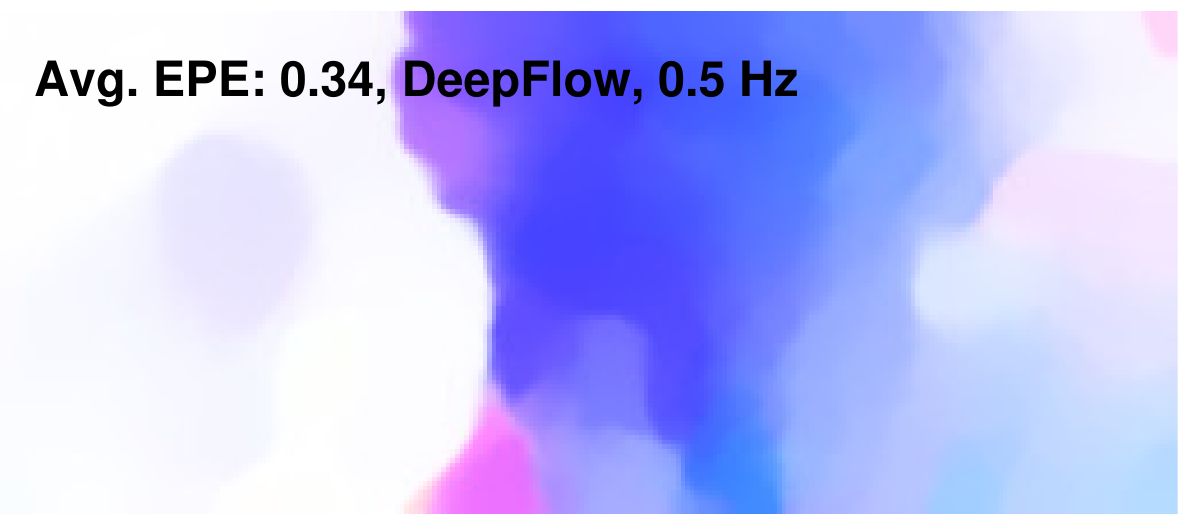}&
\includegraphics[width=0.195\textwidth]{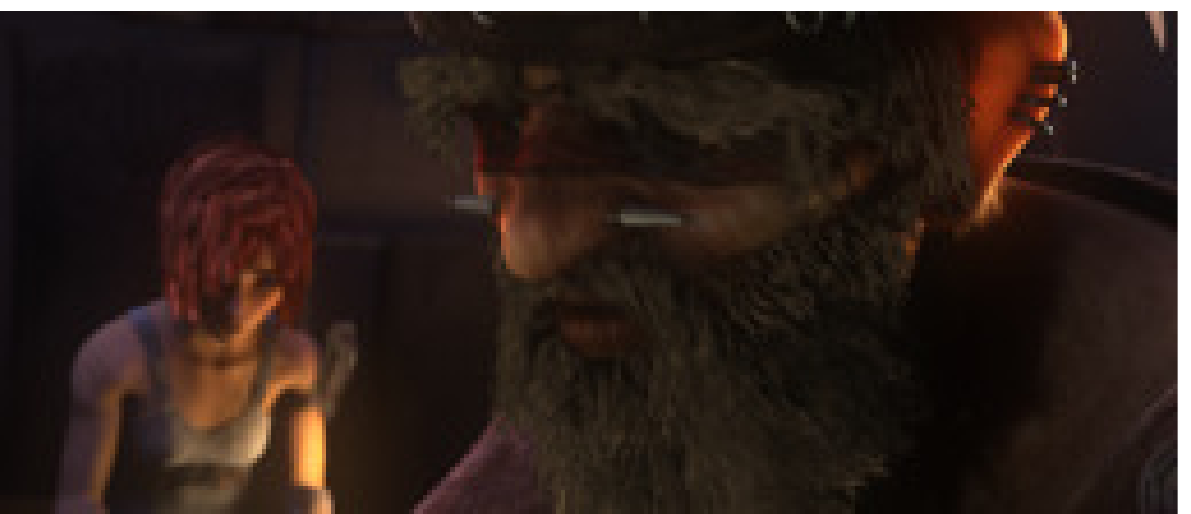}\\[6pt]
\includegraphics[width=0.195\textwidth]{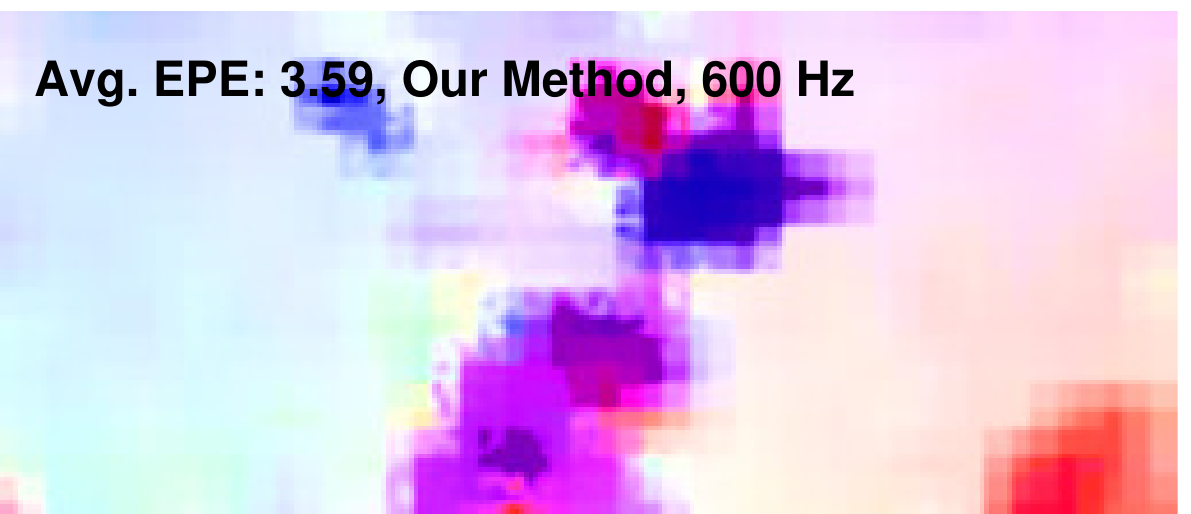}&
\includegraphics[width=0.195\textwidth]{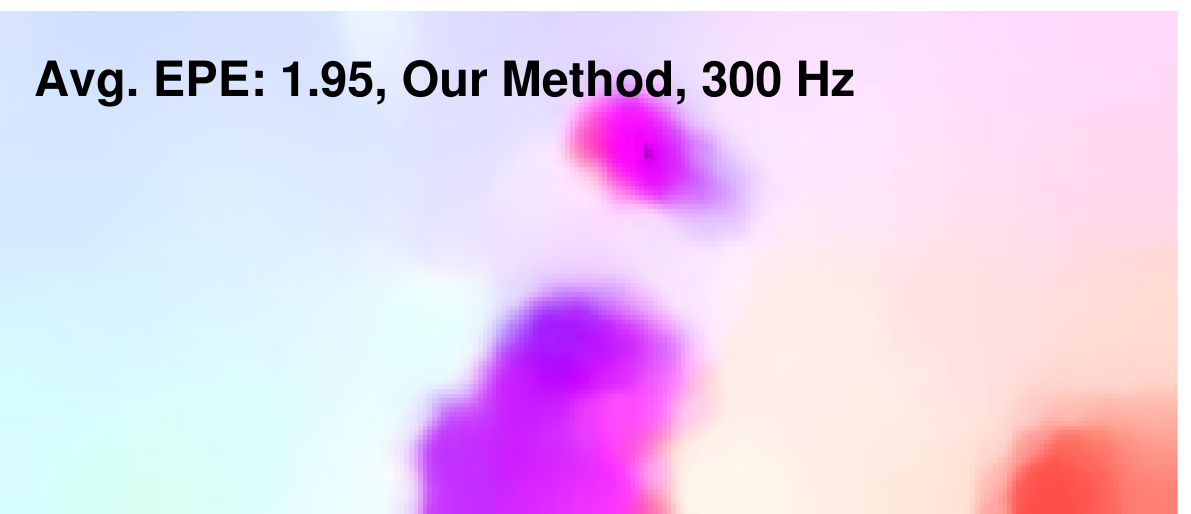}&
\includegraphics[width=0.195\textwidth]{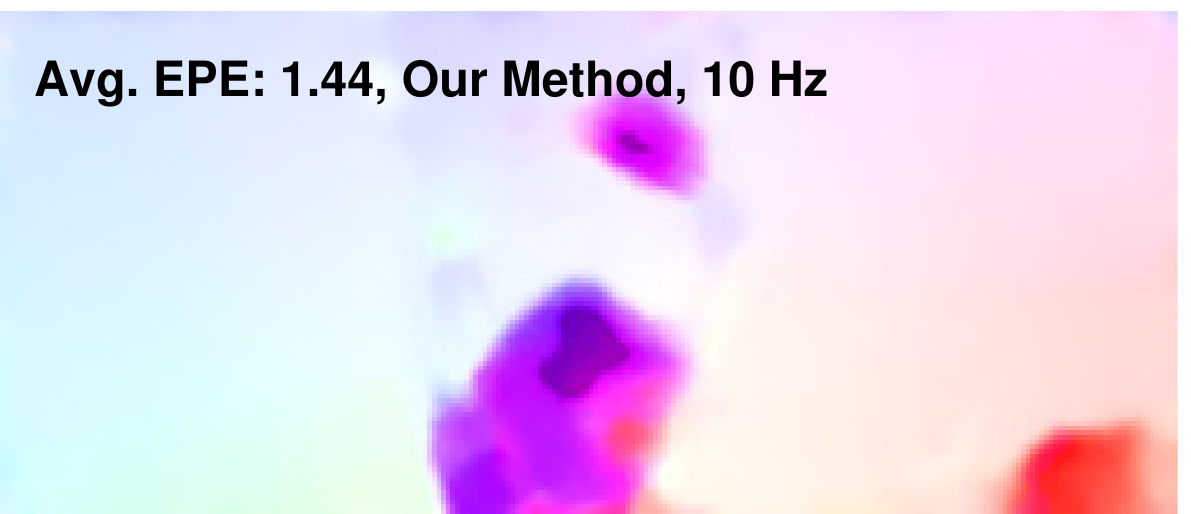}&
\includegraphics[width=0.195\textwidth]{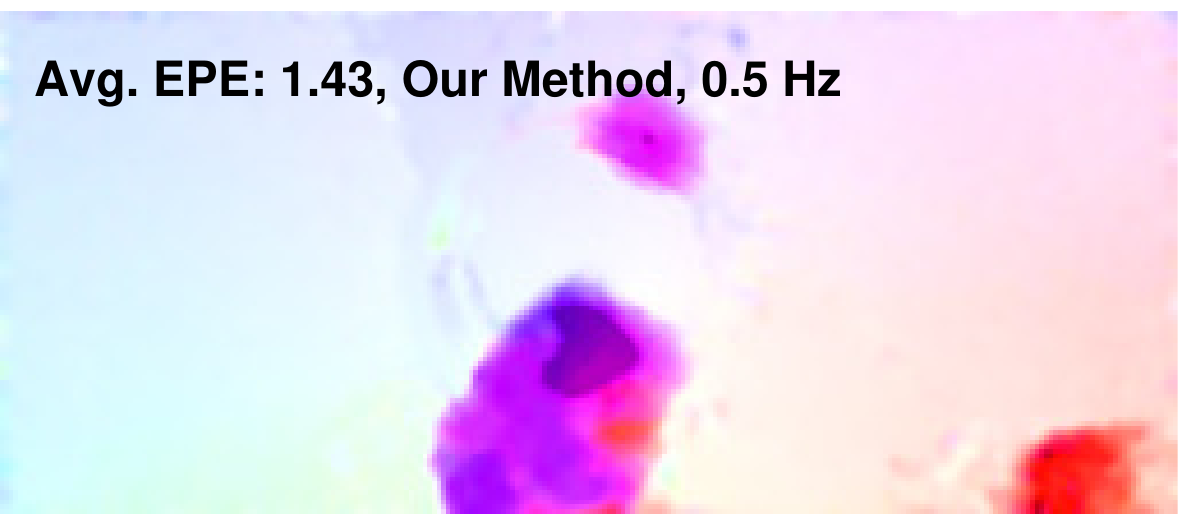}&
\includegraphics[width=0.195\textwidth]{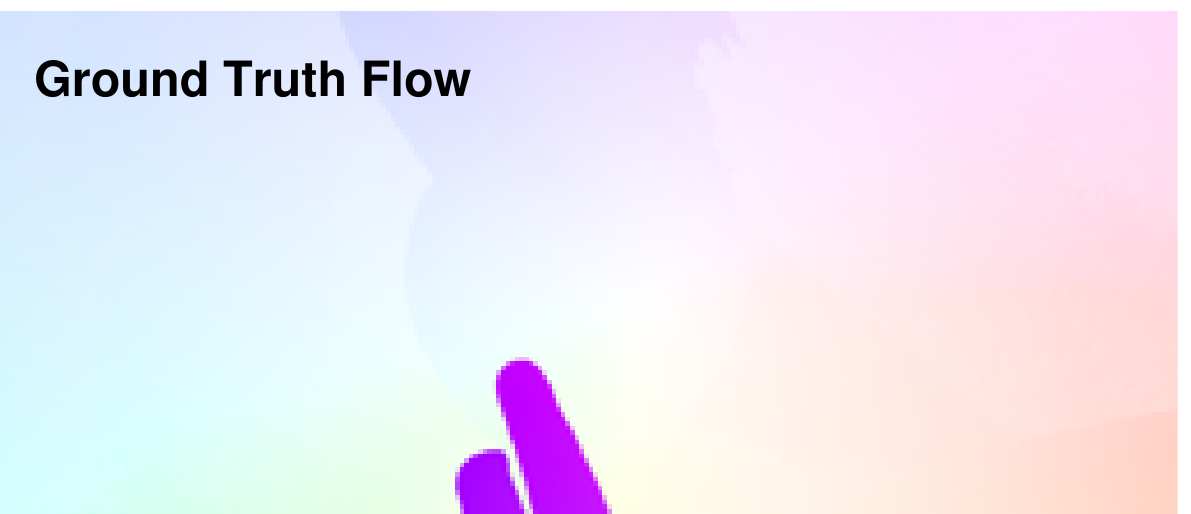}\\
\includegraphics[width=0.195\textwidth]{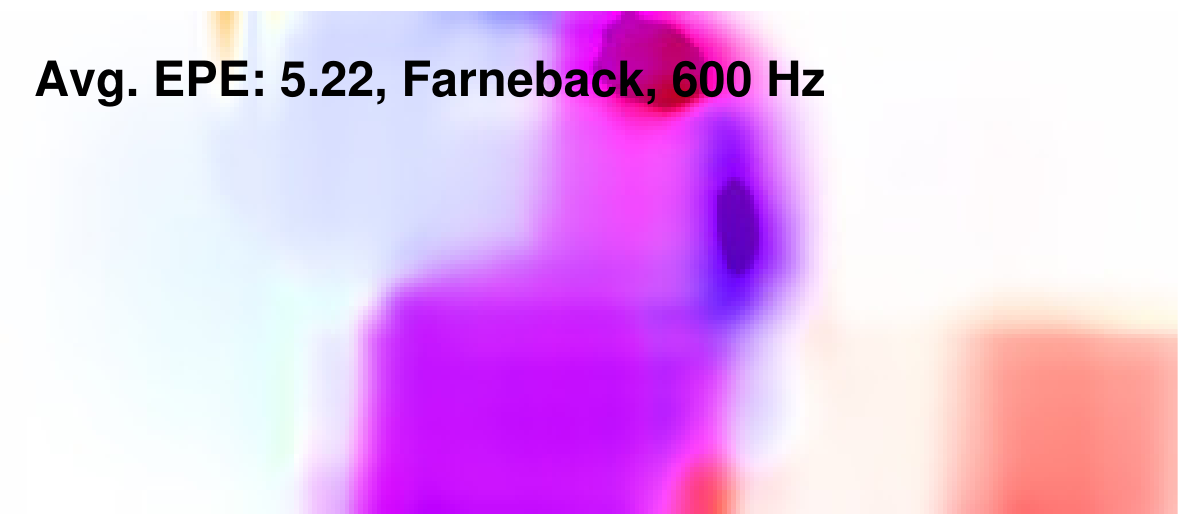}&
\includegraphics[width=0.195\textwidth]{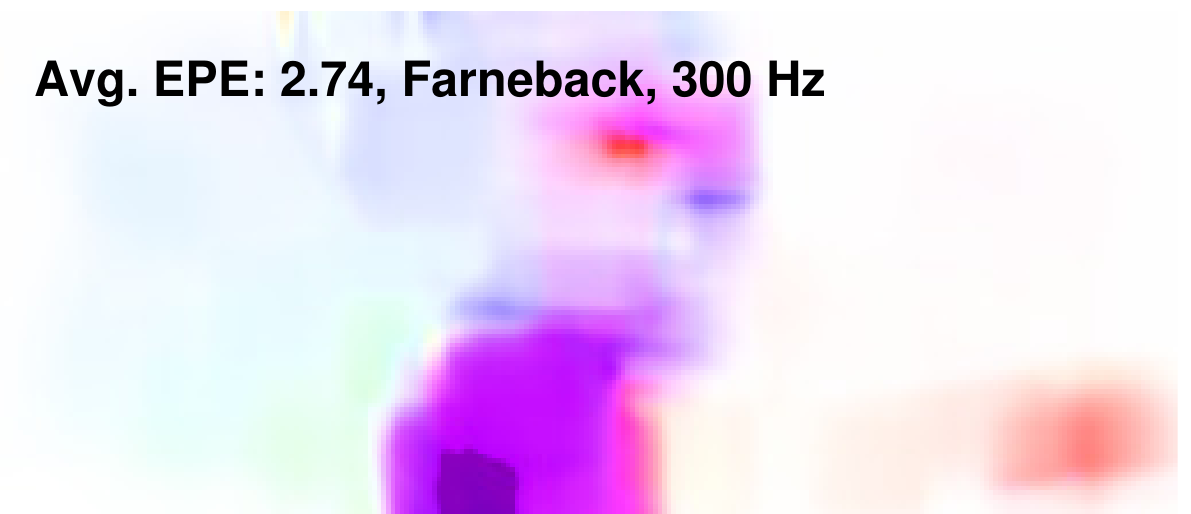}&
\includegraphics[width=0.195\textwidth]{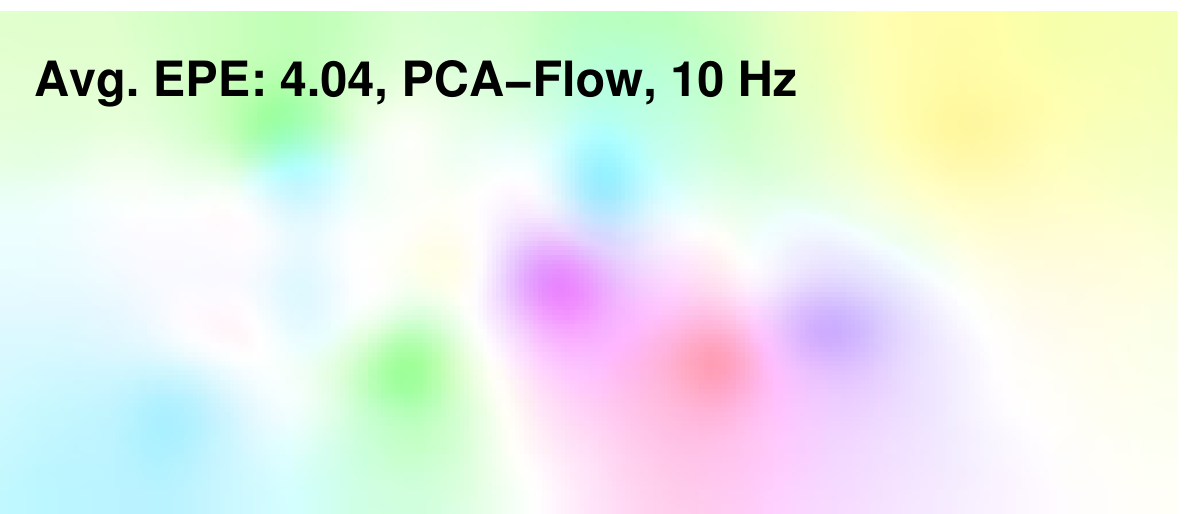}&
\includegraphics[width=0.195\textwidth]{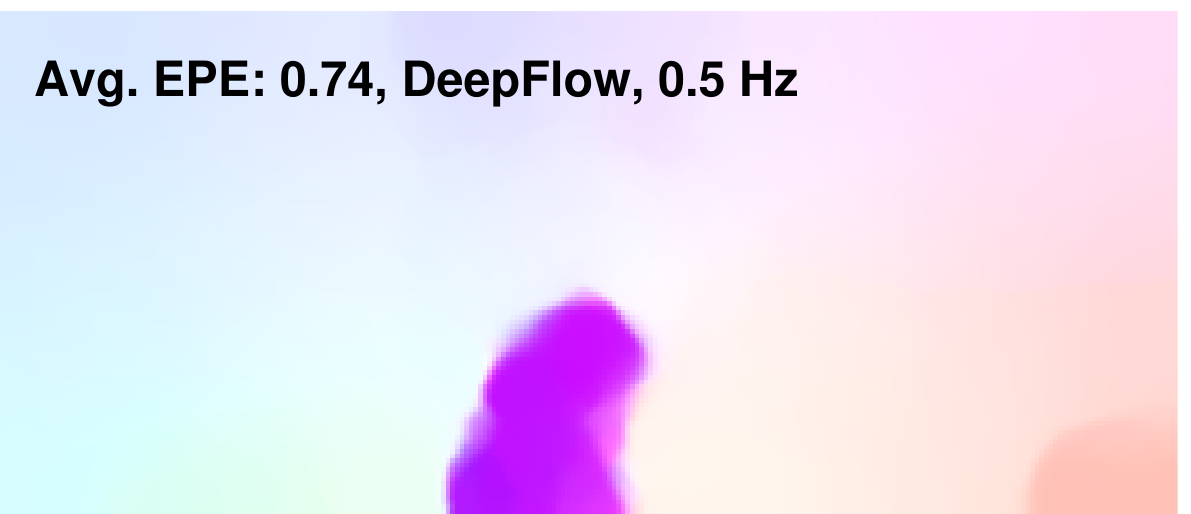}&
\includegraphics[width=0.195\textwidth]{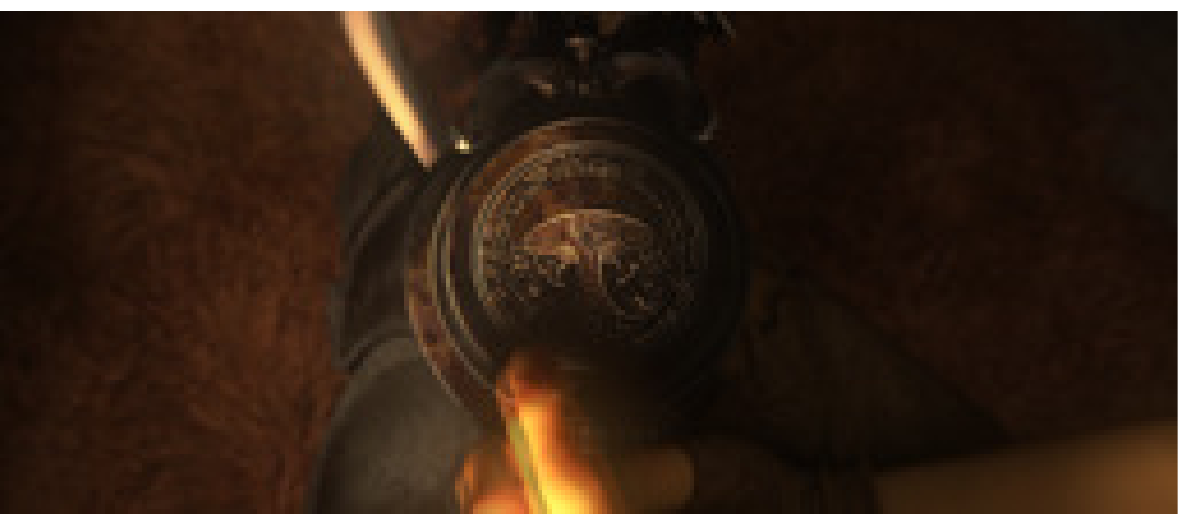}\\[6pt]
\includegraphics[width=0.195\textwidth]{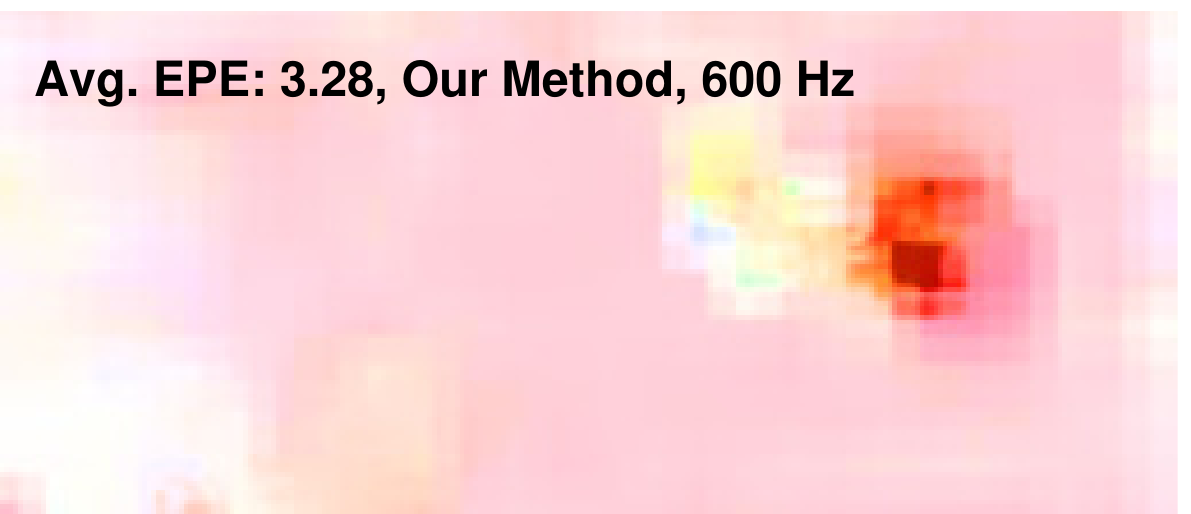}&
\includegraphics[width=0.195\textwidth]{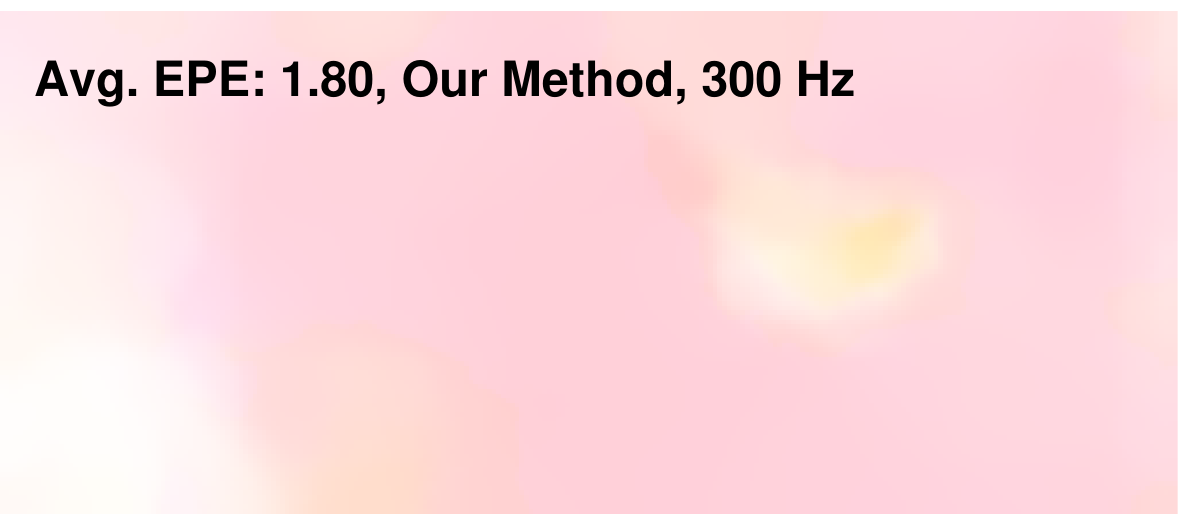}&
\includegraphics[width=0.195\textwidth]{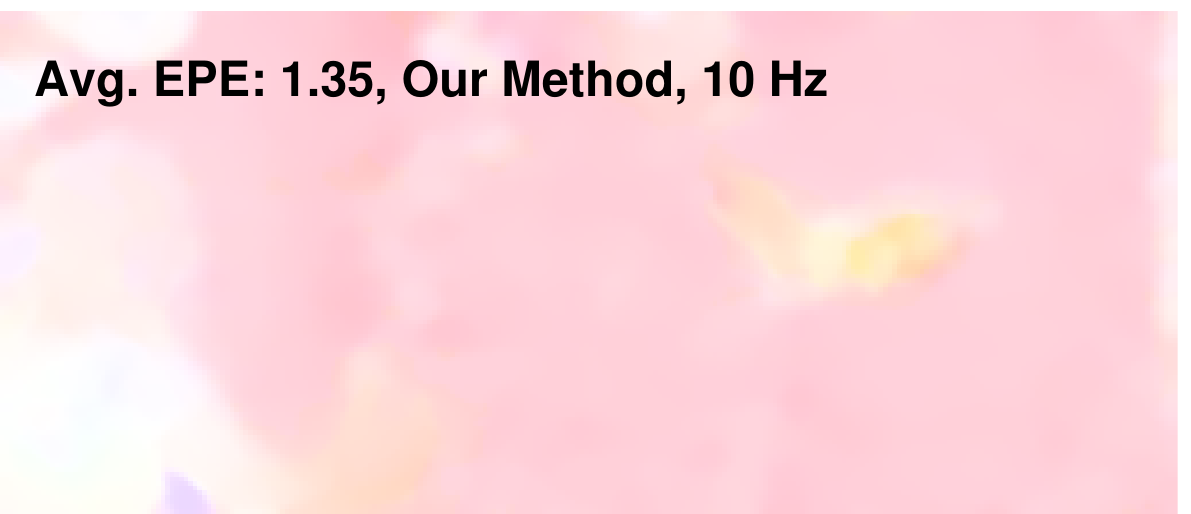}&
\includegraphics[width=0.195\textwidth]{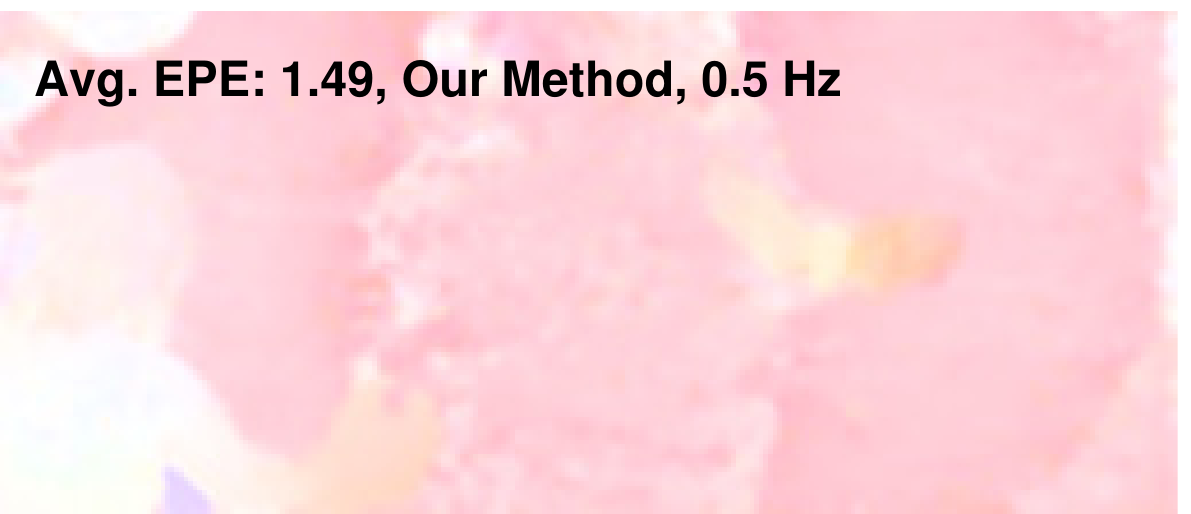}&
\includegraphics[width=0.195\textwidth]{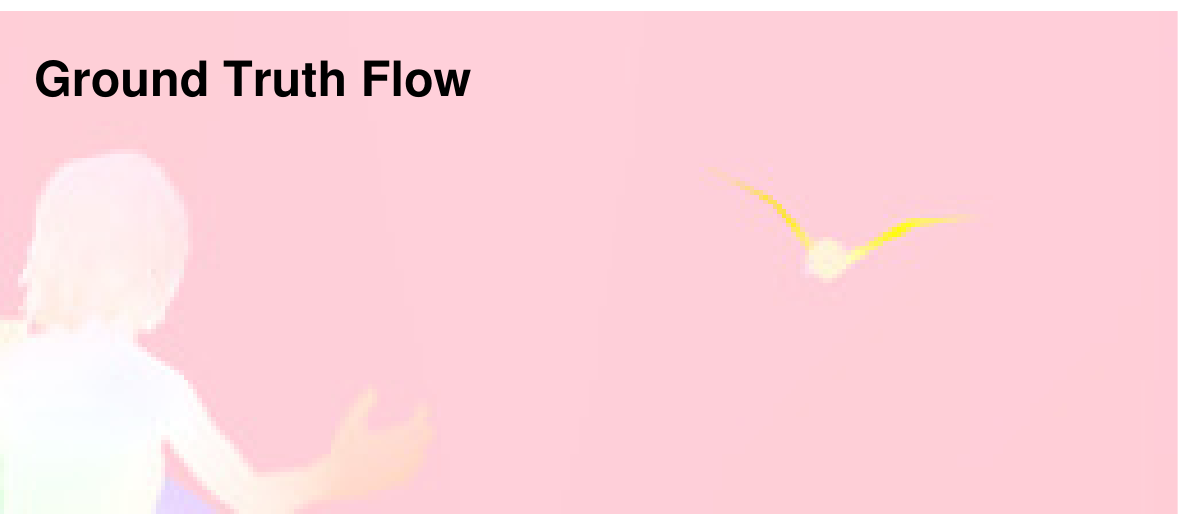}\\
\includegraphics[width=0.195\textwidth]{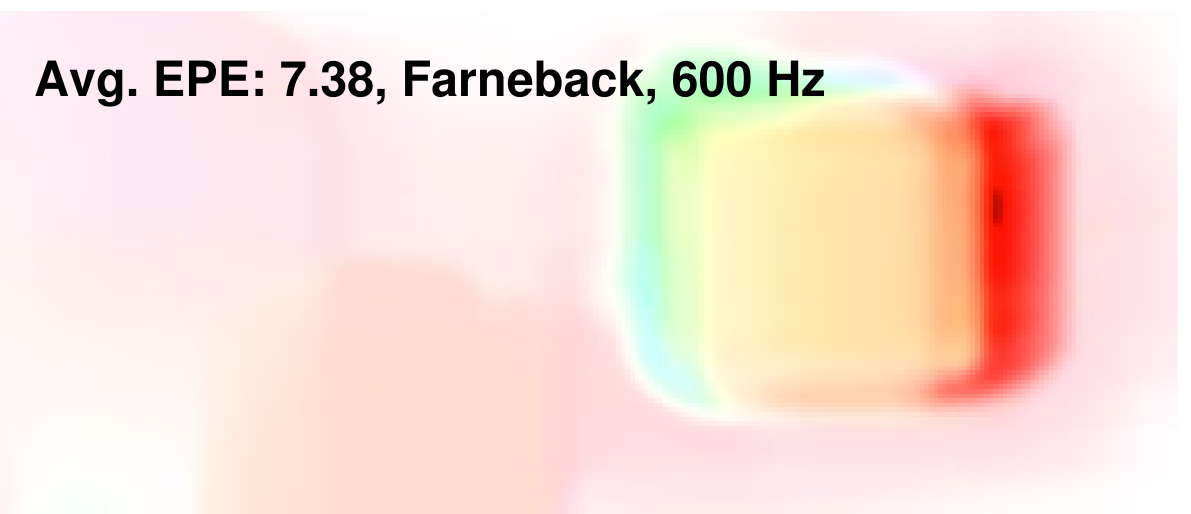}&
\includegraphics[width=0.195\textwidth]{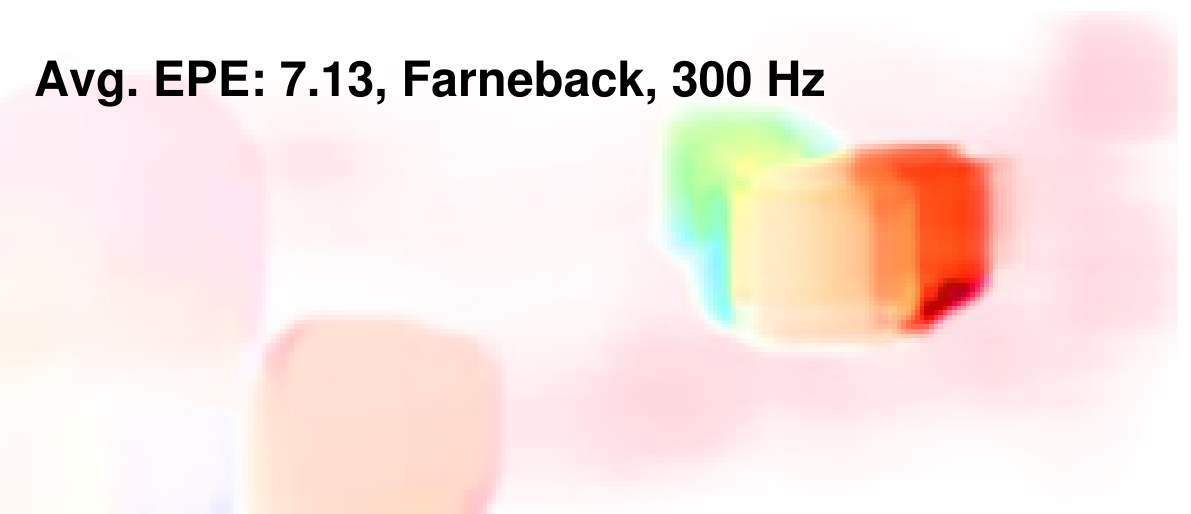}&
\includegraphics[width=0.195\textwidth]{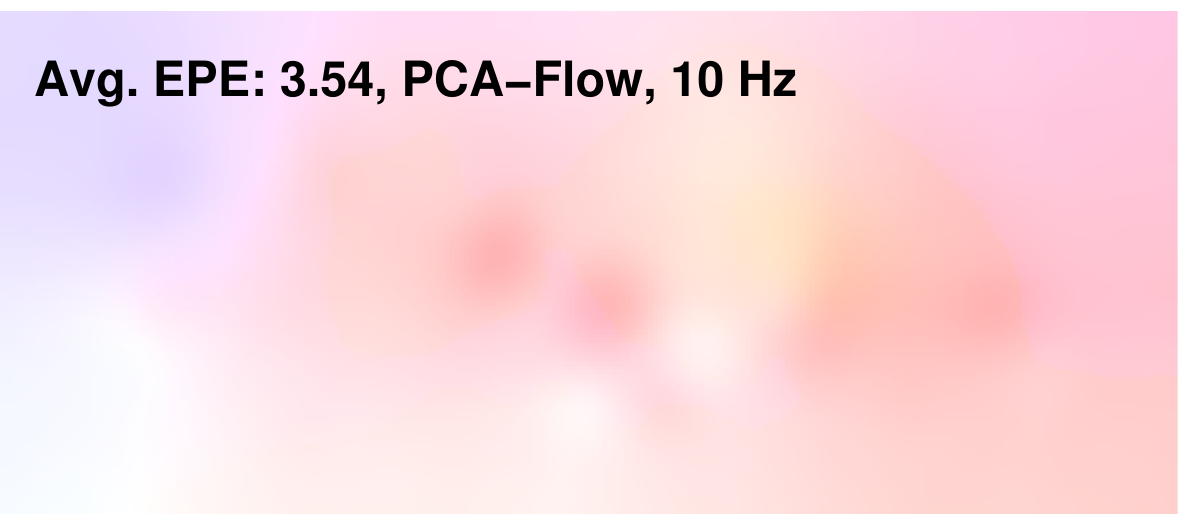}&
\includegraphics[width=0.195\textwidth]{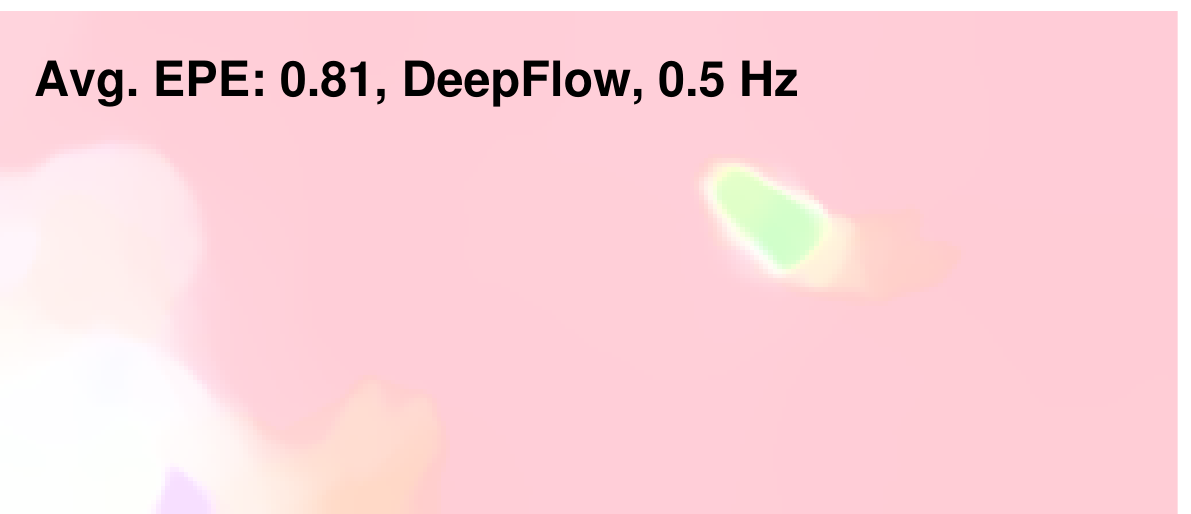}&
\includegraphics[width=0.195\textwidth]{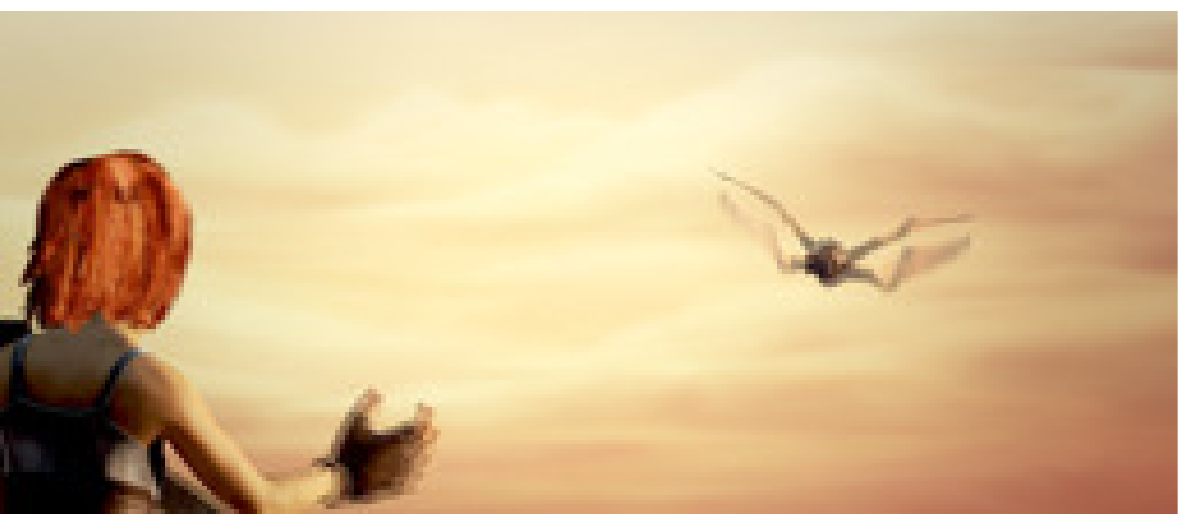}\\[6pt]
\includegraphics[width=0.195\textwidth]{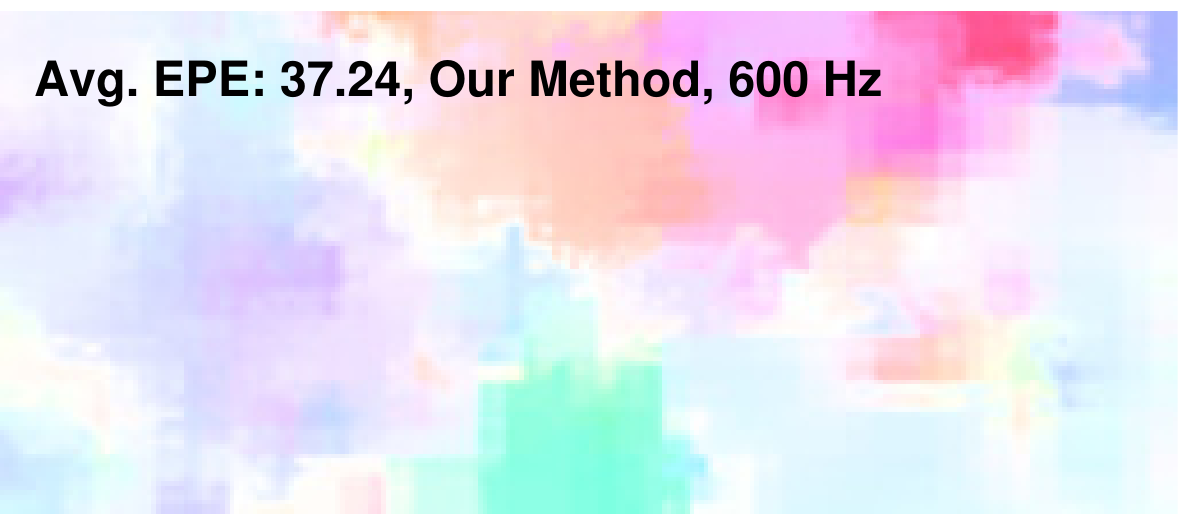}&
\includegraphics[width=0.195\textwidth]{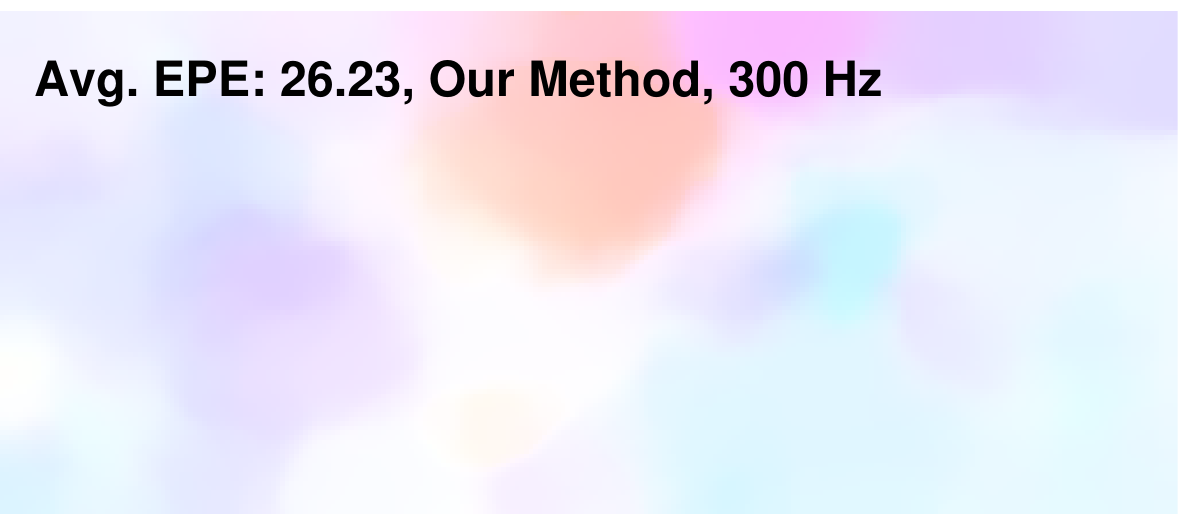}&
\includegraphics[width=0.195\textwidth]{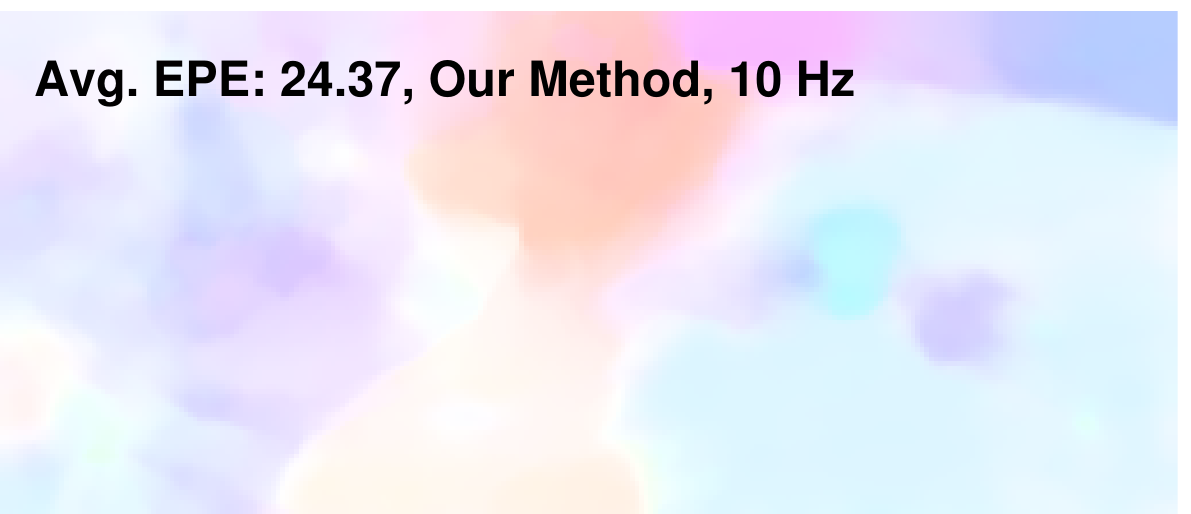}&
\includegraphics[width=0.195\textwidth]{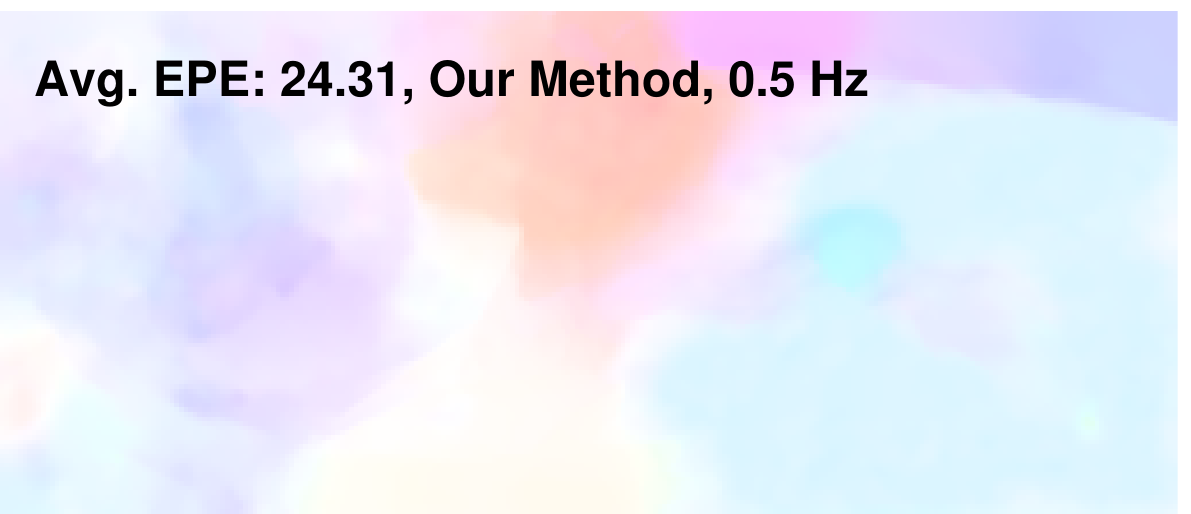}&
\includegraphics[width=0.195\textwidth]{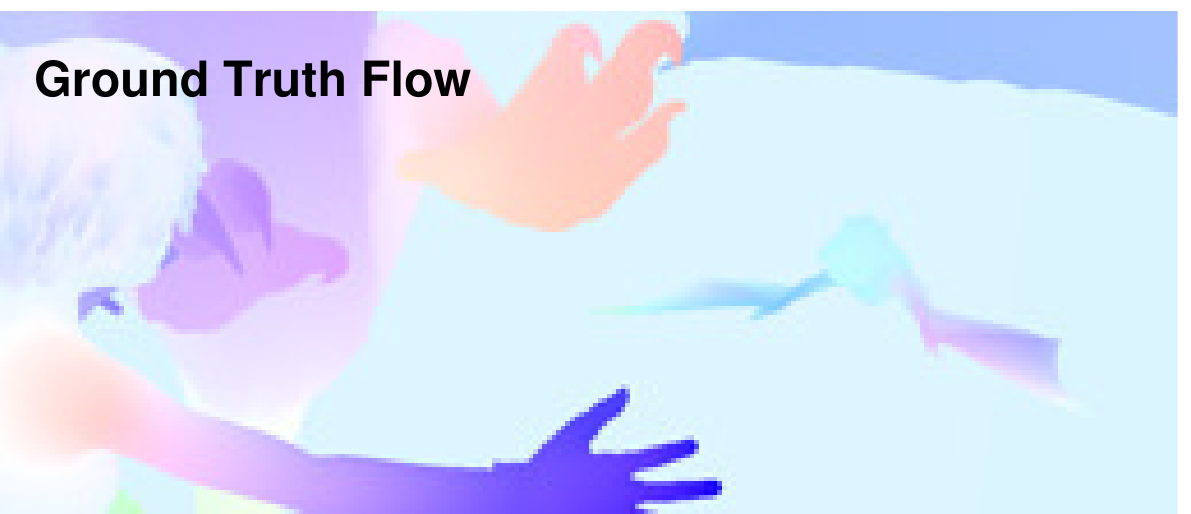}\\
\includegraphics[width=0.195\textwidth]{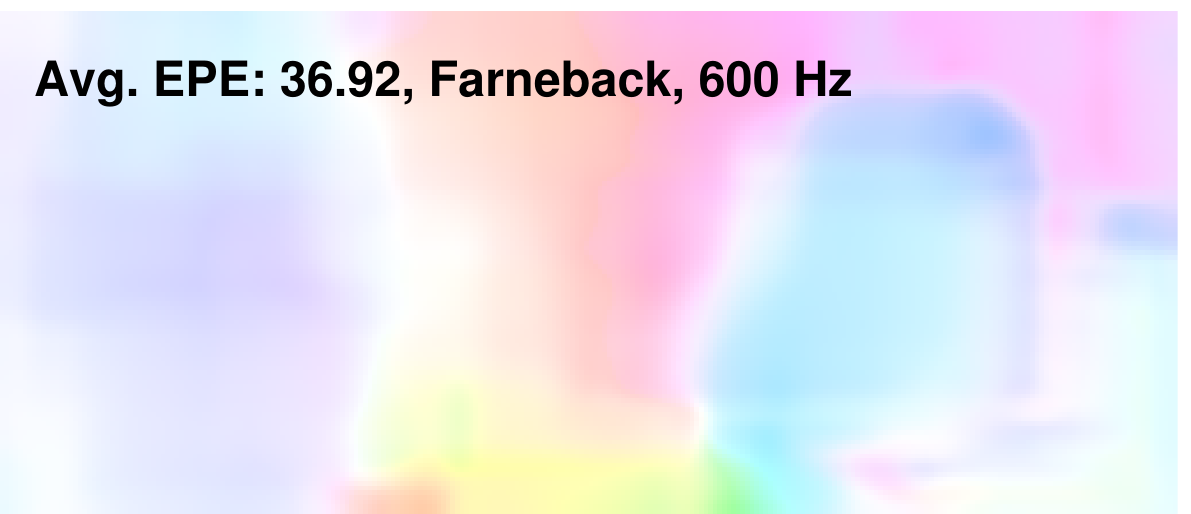}&
\includegraphics[width=0.195\textwidth]{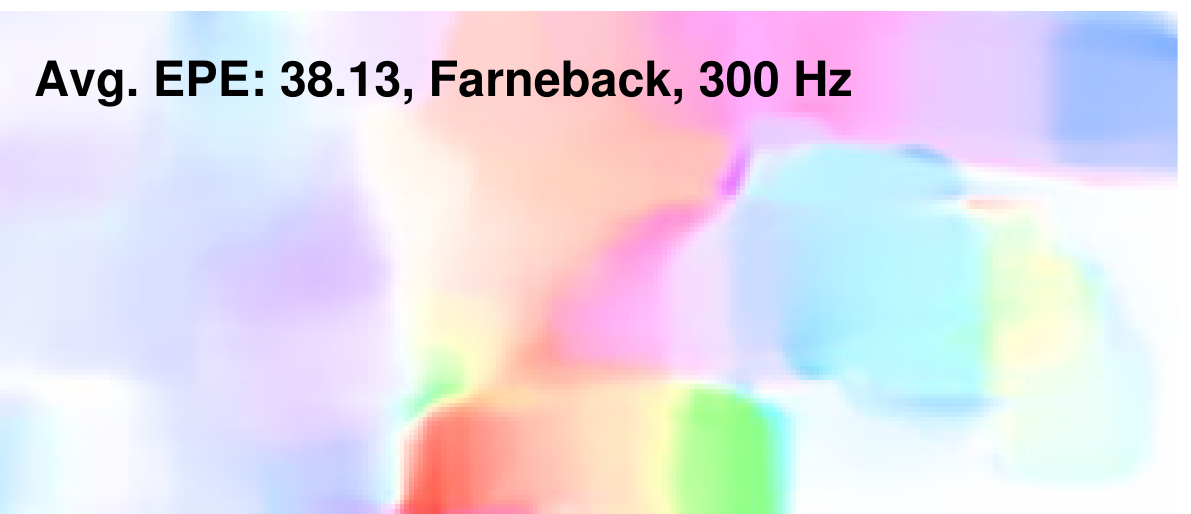}&
\includegraphics[width=0.195\textwidth]{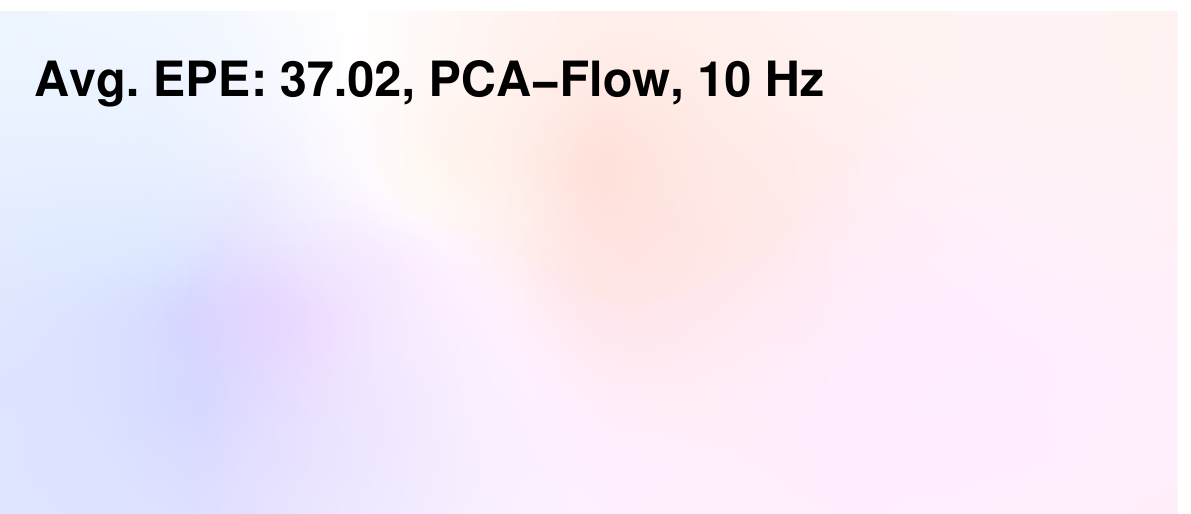}&
\includegraphics[width=0.195\textwidth]{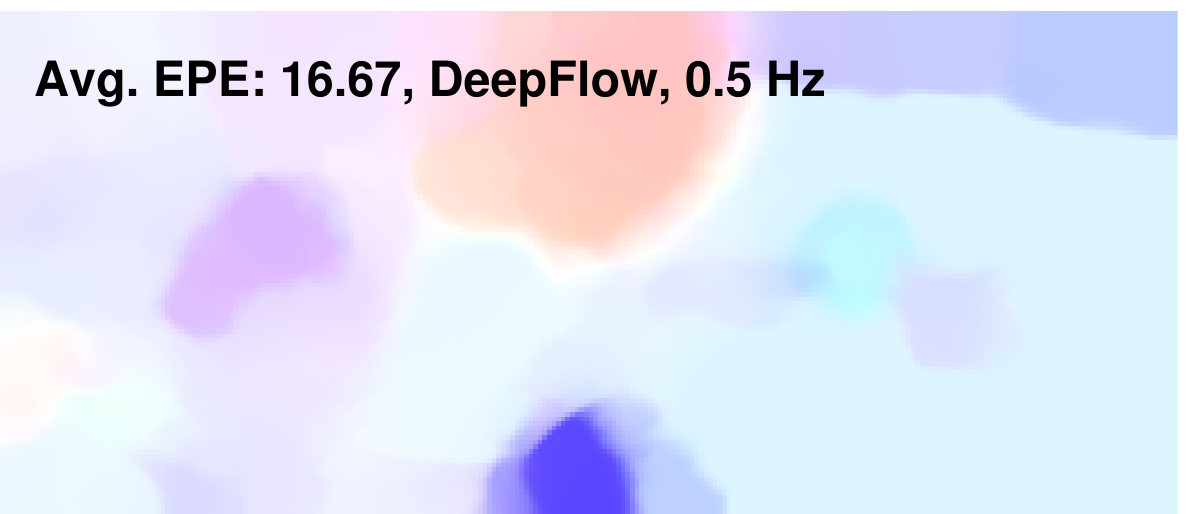}&
\includegraphics[width=0.195\textwidth]{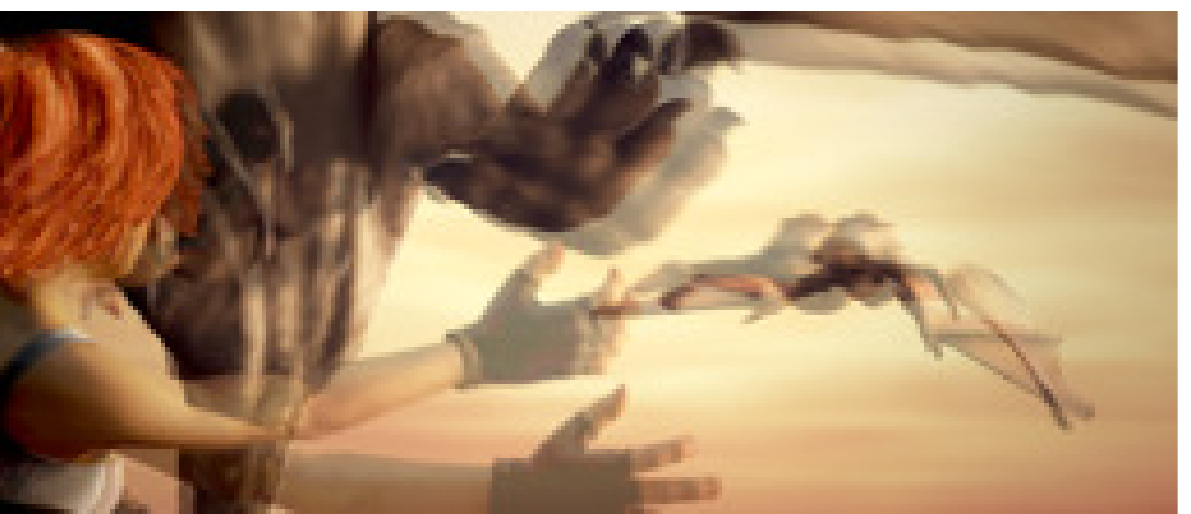}\\[6pt]
\end{tabular}
}
\caption{Exemplary results on Sintel (training). In each block of $2 \times 6$ images.  Top row, left to right: Our method for operating points ({\bf 1})-({\bf 4}), Ground Truth. Bottom row: Farneback 600Hz, Farneback 300Hz, PCA-Flow 10Hz, DeepFlow 0.5Hz, Original Image. See Fig. \ref{fig:sintel2res_errmap_AP} for error maps.}\label{fig:sintel2res_AP} 
\end{figure*}

\begin{figure*} [!ht]
\centering\setlength{\tabcolsep}{0.1pt}\renewcommand{\arraystretch}{0} 
{
\begin{tabular}{ccccc}
 {\bf 600Hz} & {\bf 300Hz} & {\bf 10Hz} & {\bf 0.5Hz}& {\bf Ground Truth}\\
\includegraphics[width=0.195\textwidth]{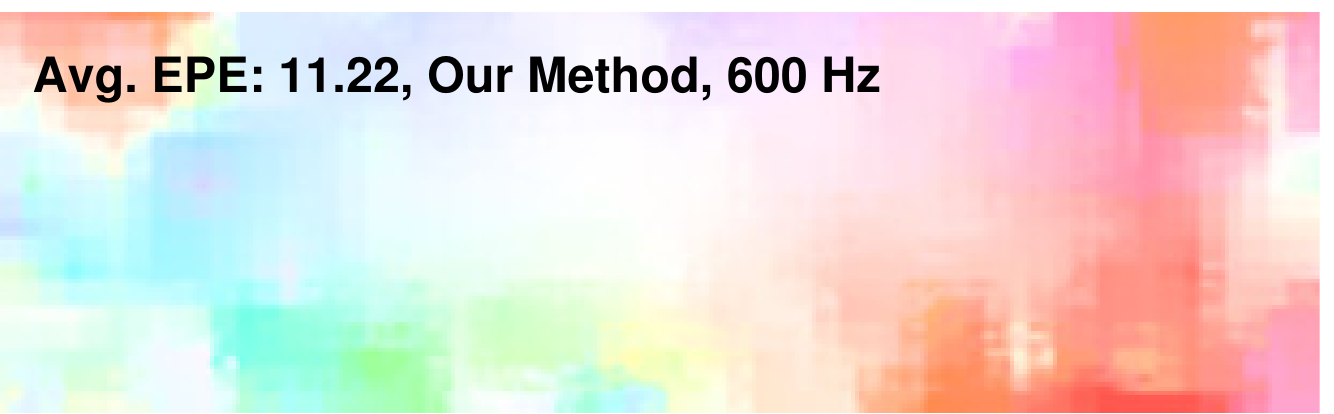}&
\includegraphics[width=0.195\textwidth]{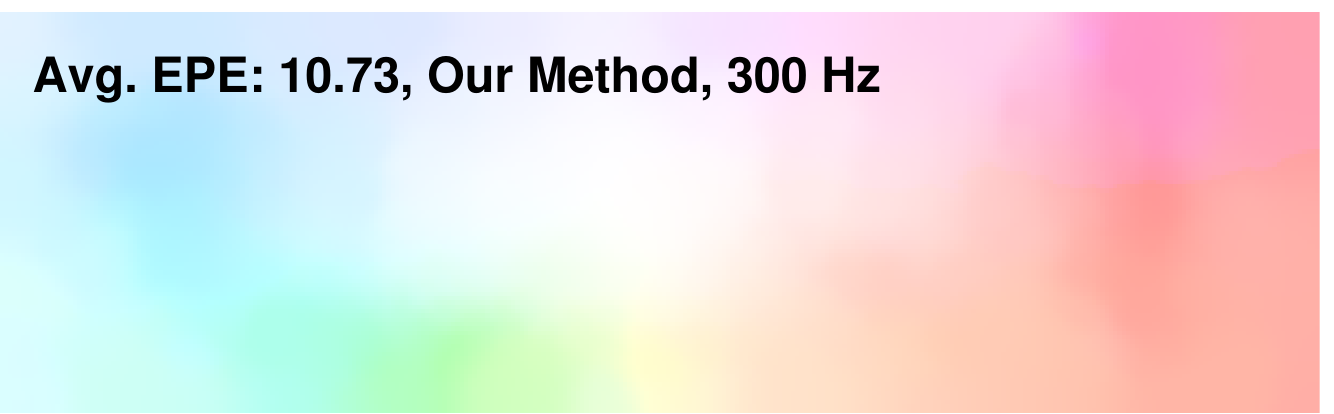}&
\includegraphics[width=0.195\textwidth]{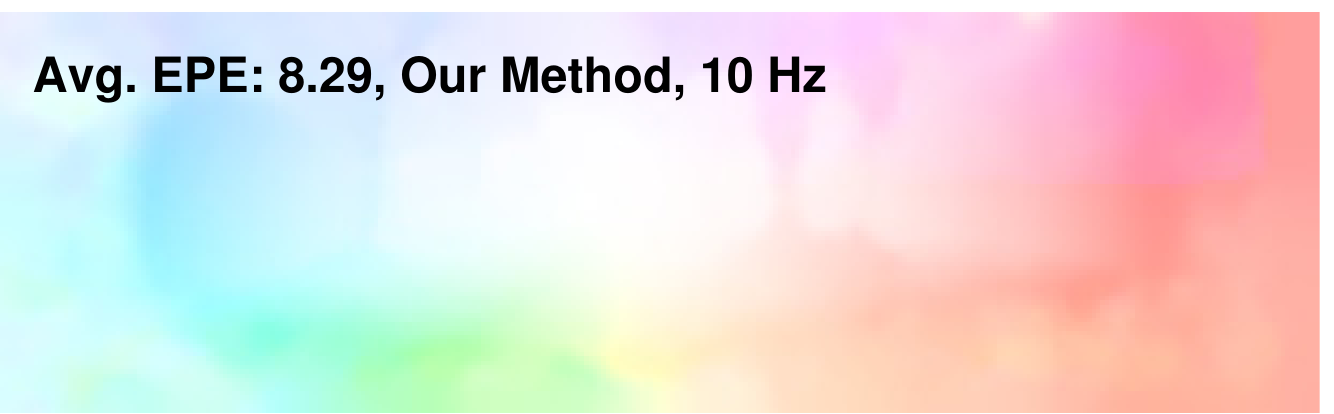}&
\includegraphics[width=0.195\textwidth]{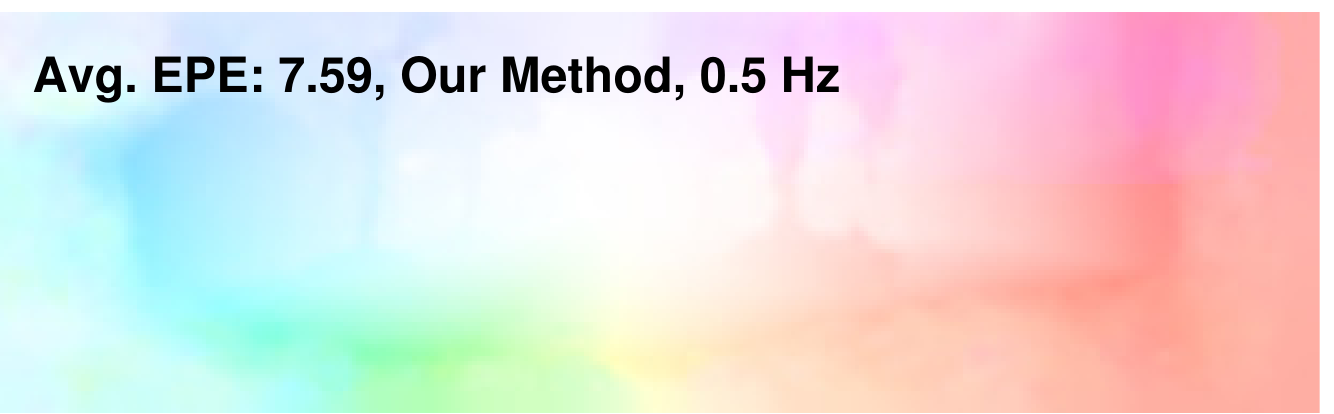}&
\includegraphics[width=0.195\textwidth]{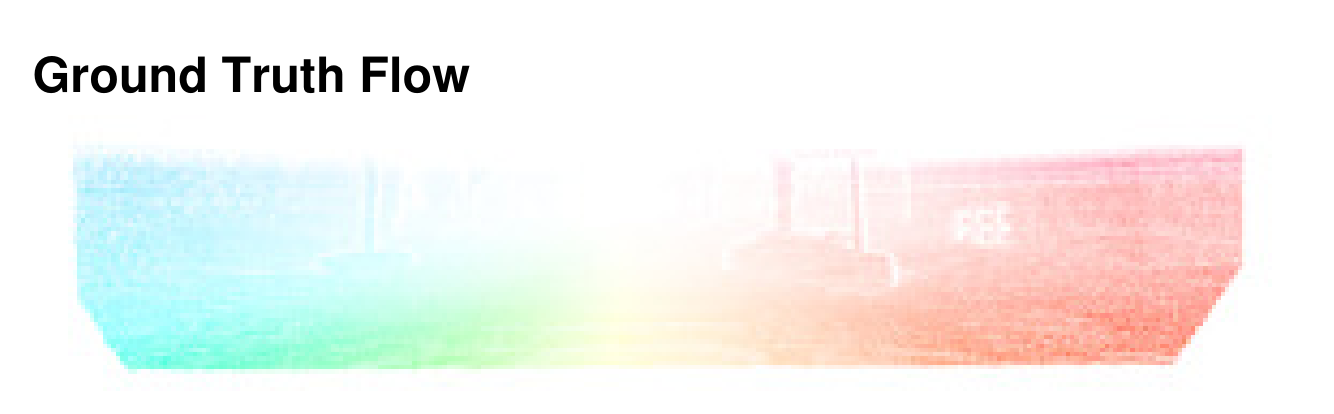}\\
\includegraphics[width=0.195\textwidth]{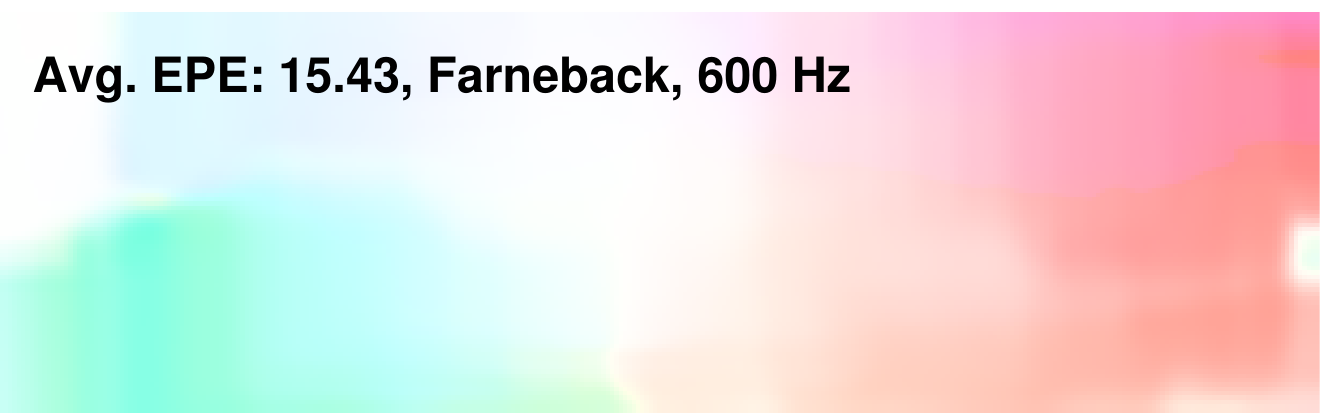}&
\includegraphics[width=0.195\textwidth]{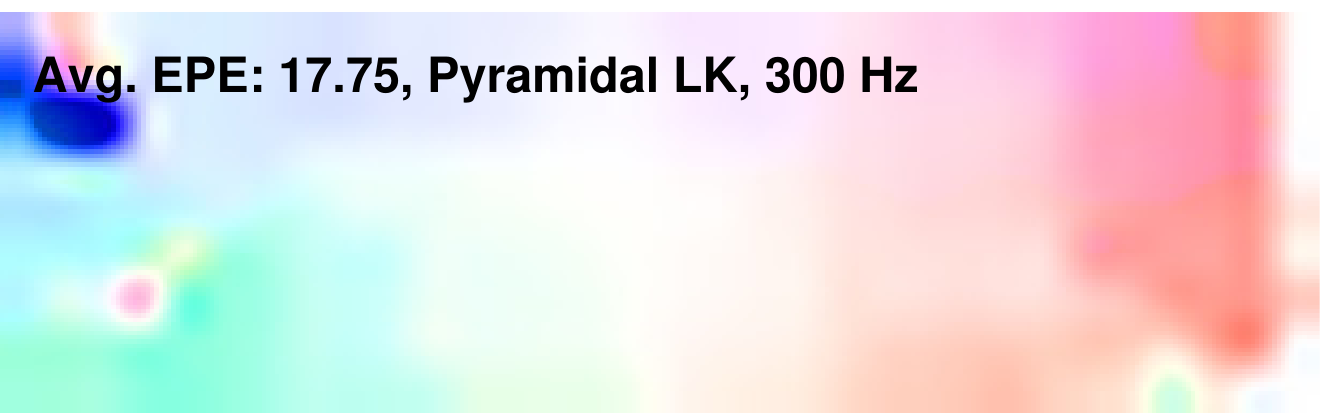}&
\includegraphics[width=0.195\textwidth]{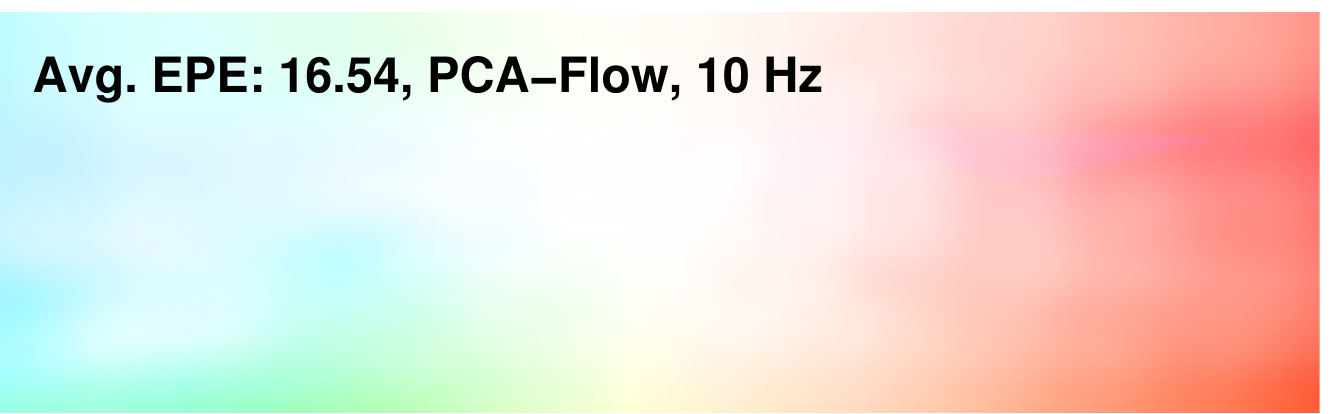}&
\includegraphics[width=0.195\textwidth]{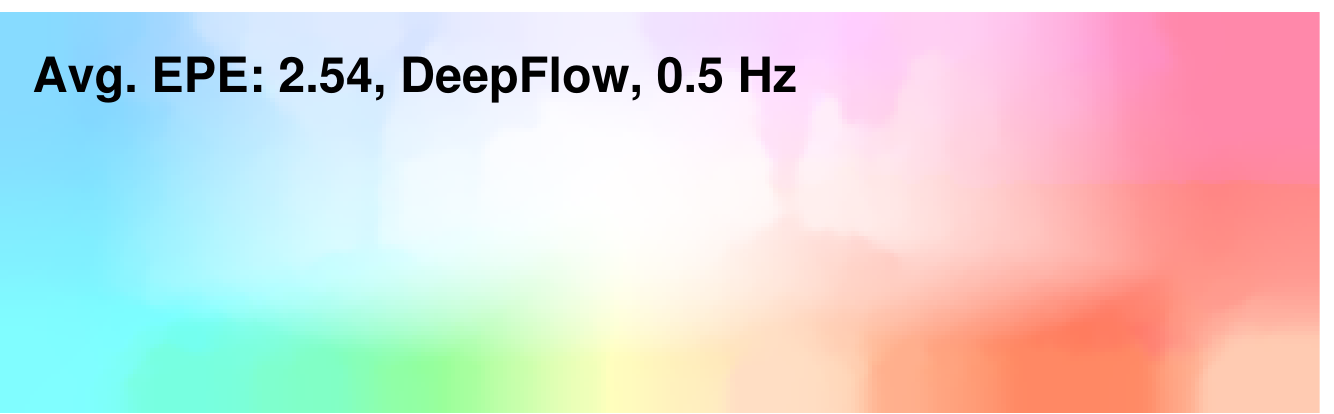}&
\includegraphics[width=0.195\textwidth]{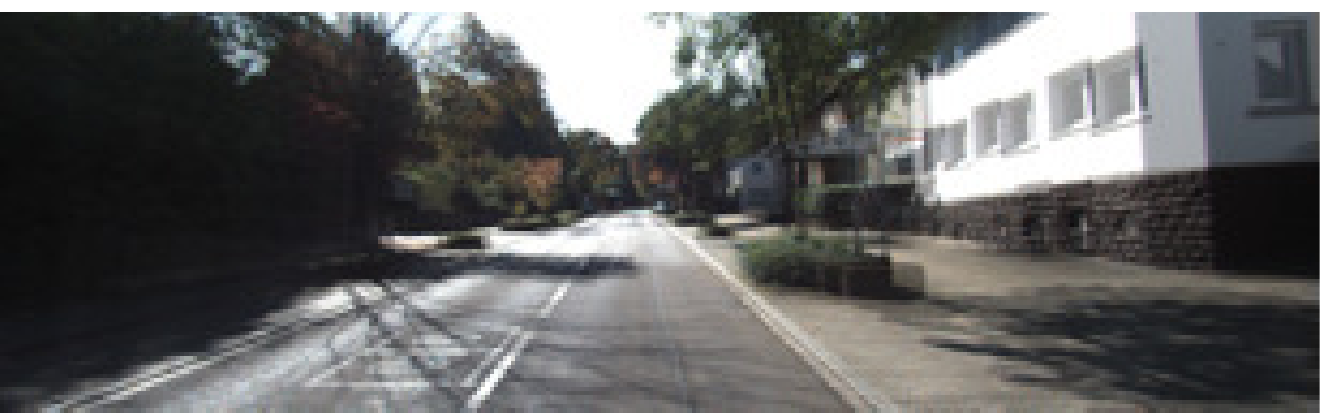}\\[6pt]
\includegraphics[width=0.195\textwidth]{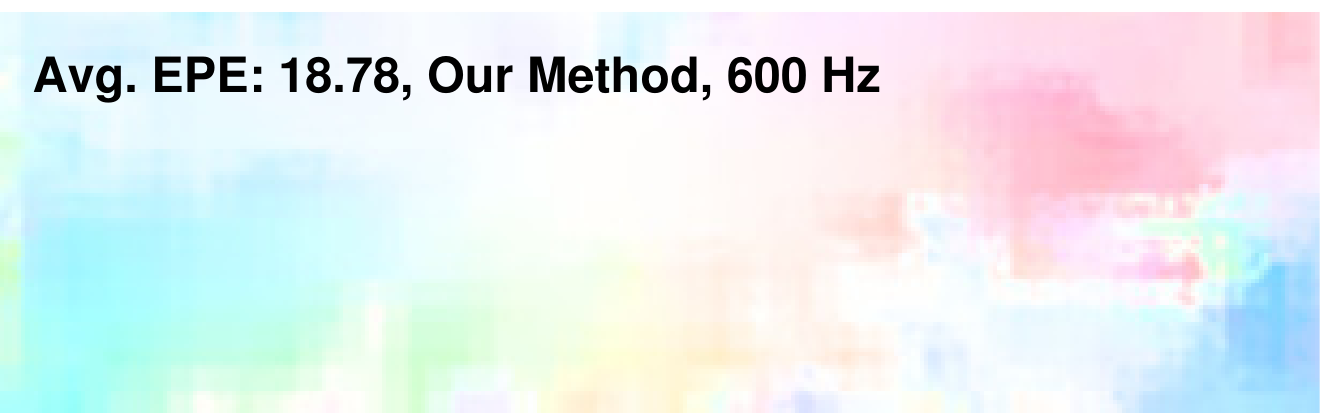}&
\includegraphics[width=0.195\textwidth]{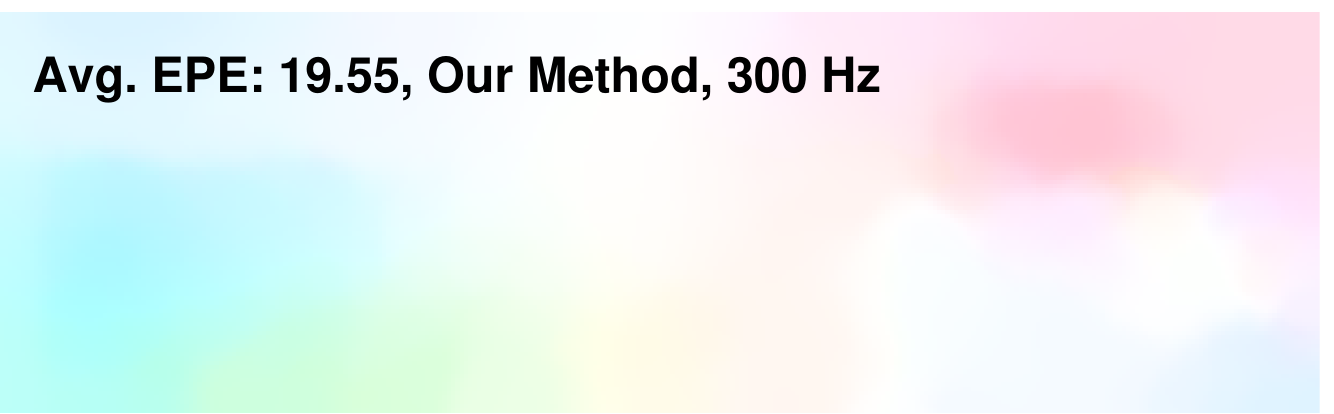}&
\includegraphics[width=0.195\textwidth]{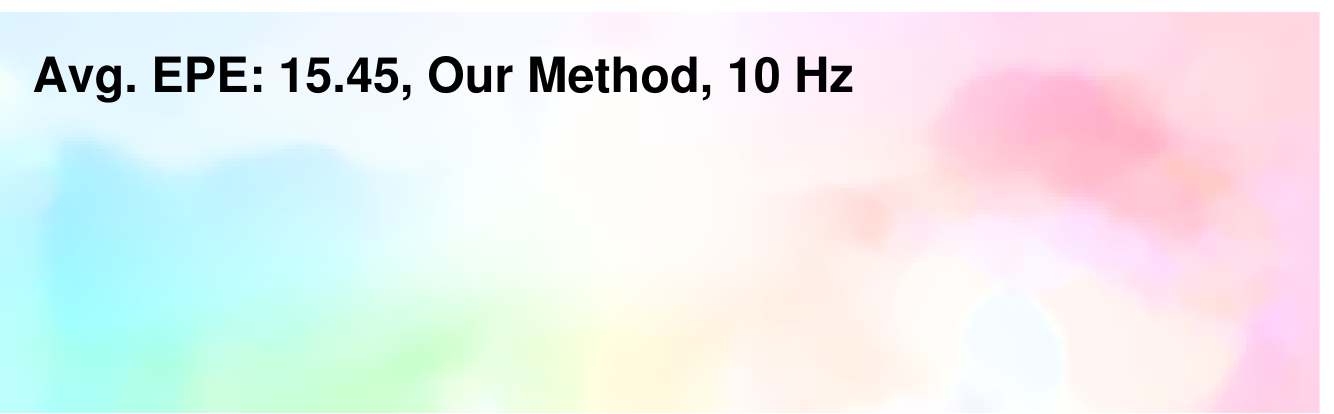}&
\includegraphics[width=0.195\textwidth]{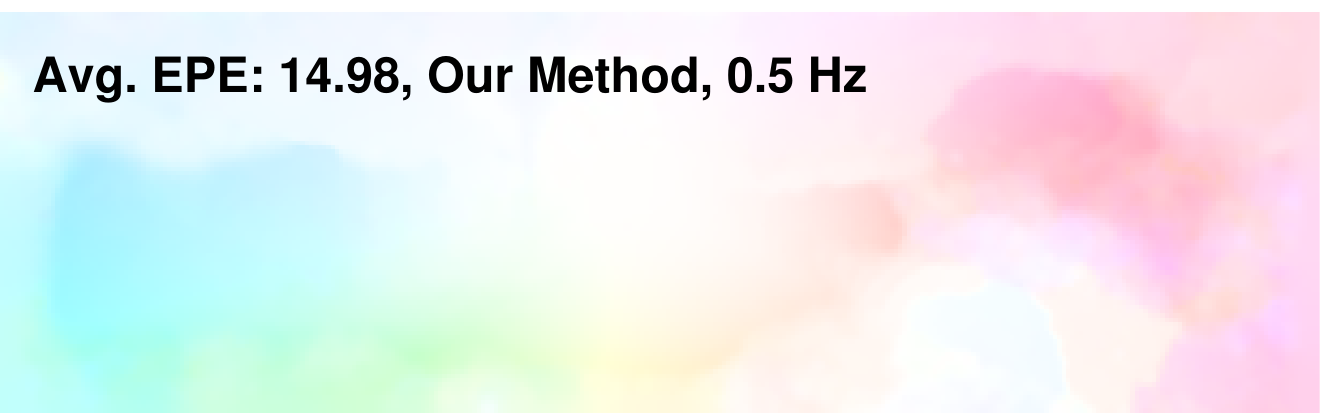}&
\includegraphics[width=0.195\textwidth]{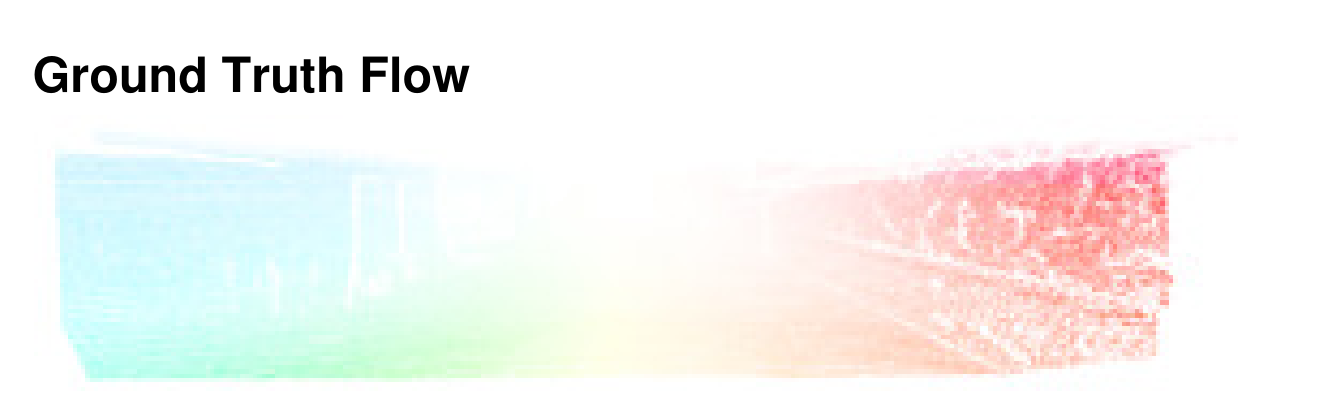}\\
\includegraphics[width=0.195\textwidth]{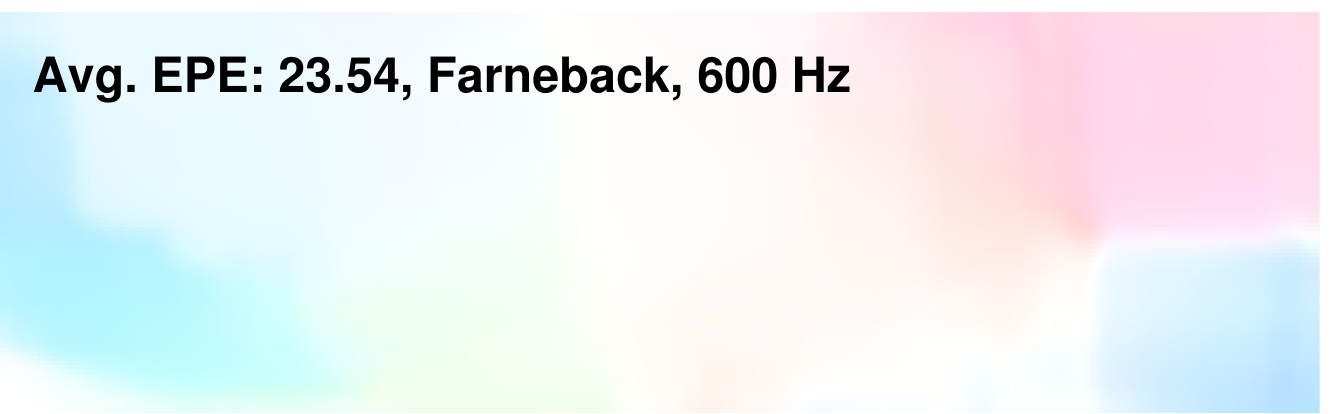}&
\includegraphics[width=0.195\textwidth]{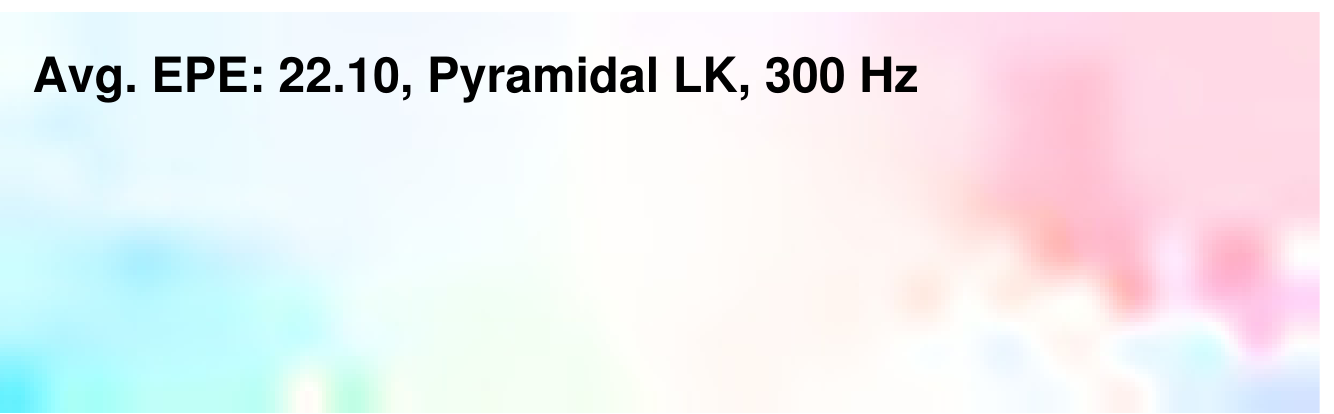}&
\includegraphics[width=0.195\textwidth]{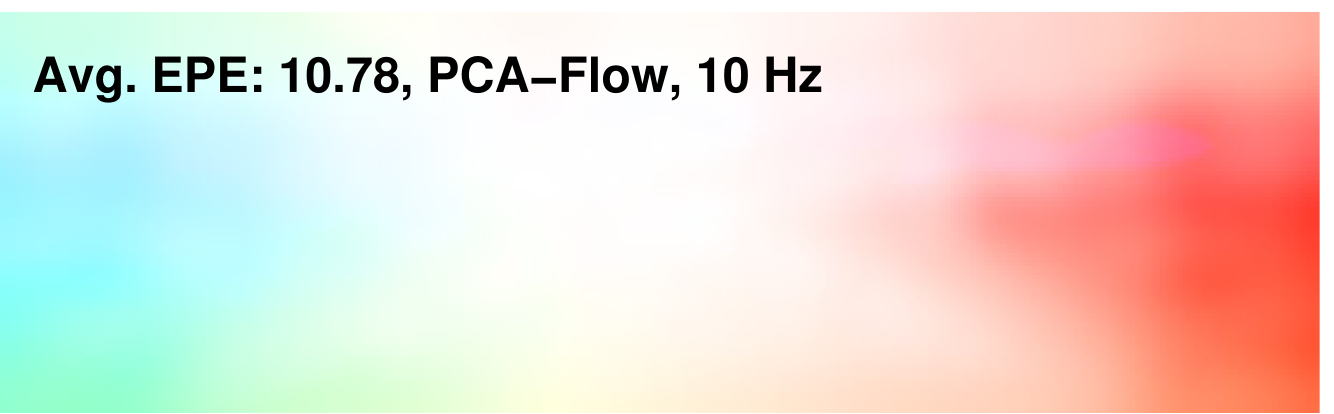}&
\includegraphics[width=0.195\textwidth]{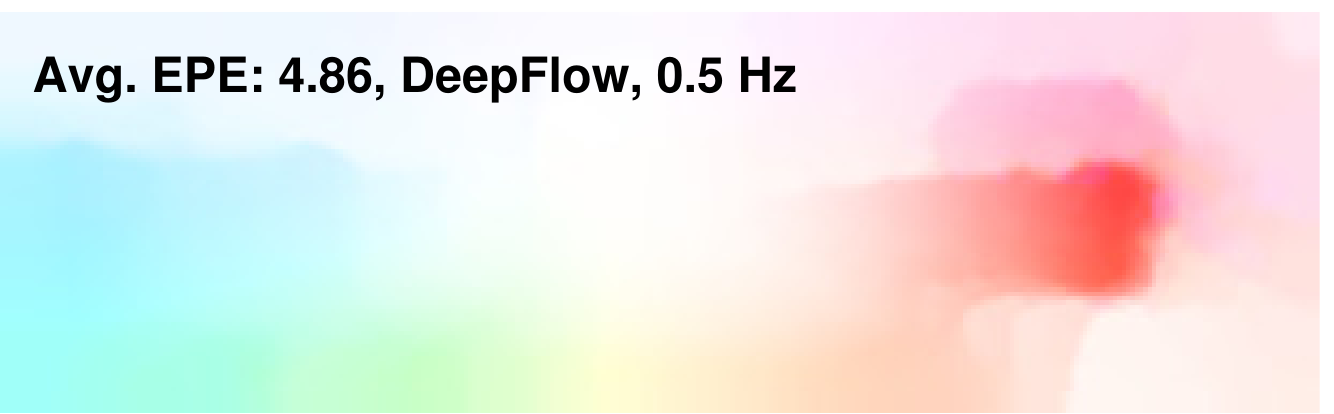}&
\includegraphics[width=0.195\textwidth]{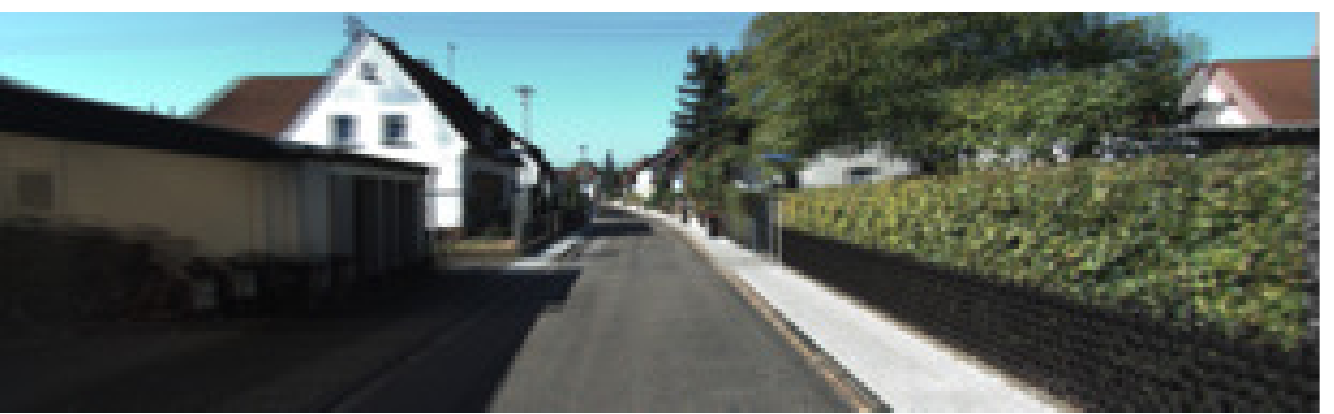}\\[6pt]
\includegraphics[width=0.195\textwidth]{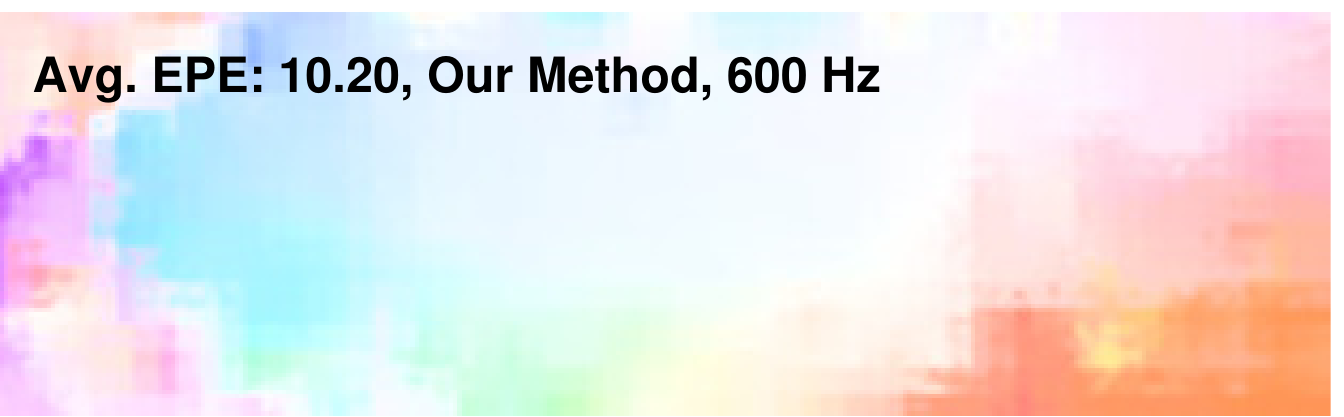}&
\includegraphics[width=0.195\textwidth]{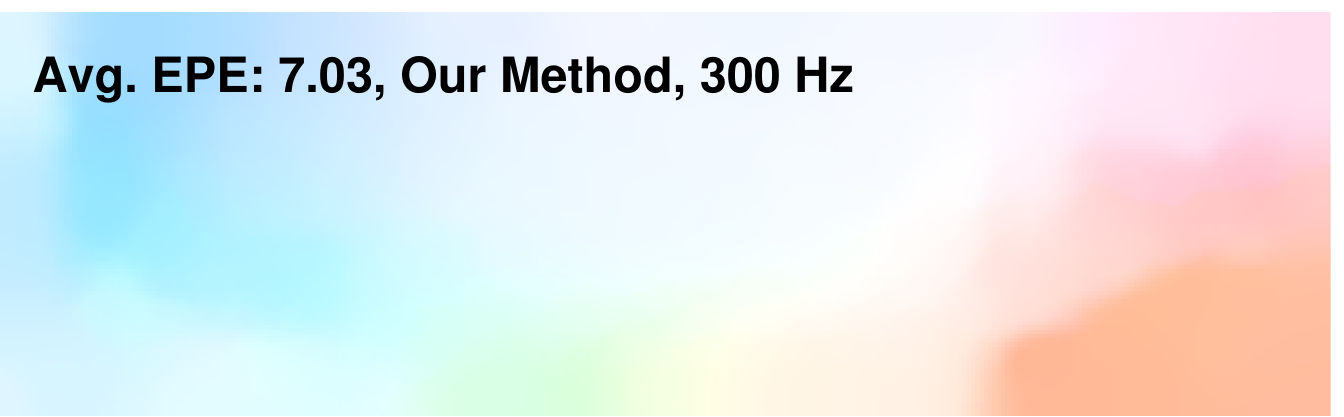}&
\includegraphics[width=0.195\textwidth]{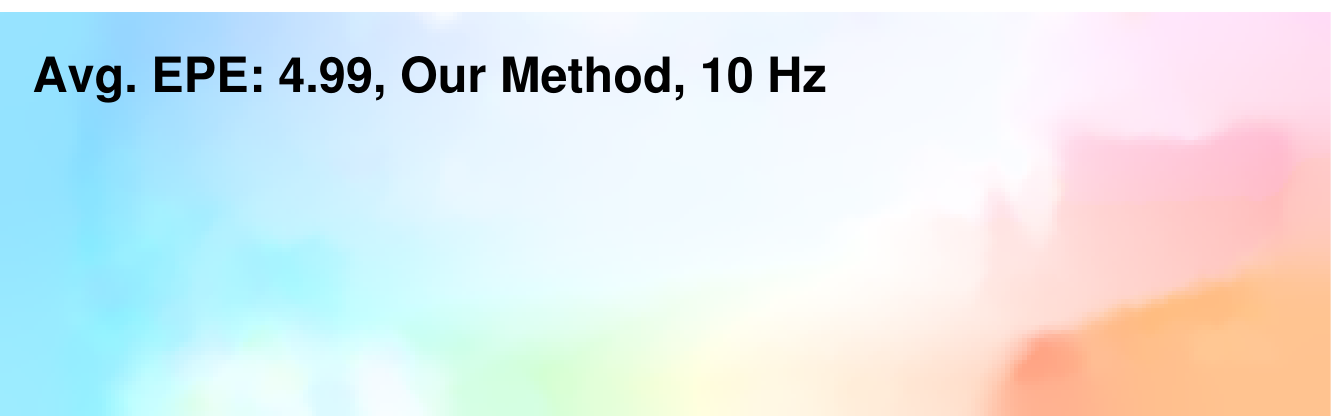}&
\includegraphics[width=0.195\textwidth]{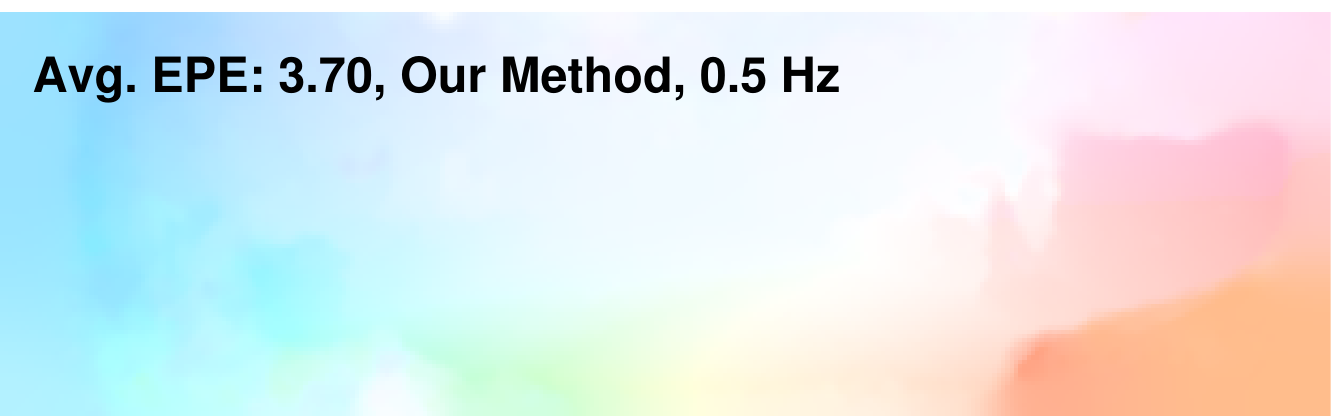}&
\includegraphics[width=0.195\textwidth]{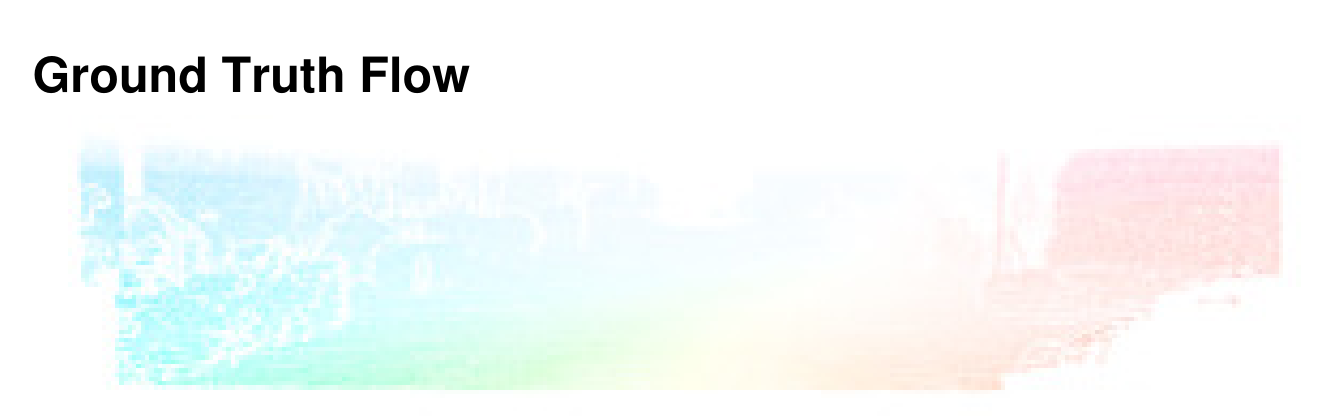}\\
\includegraphics[width=0.195\textwidth]{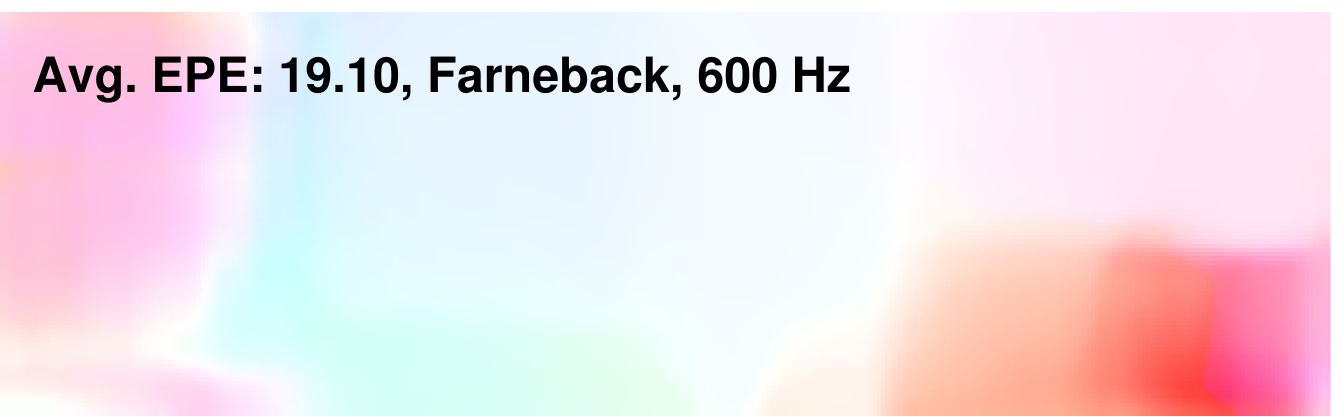}&
\includegraphics[width=0.195\textwidth]{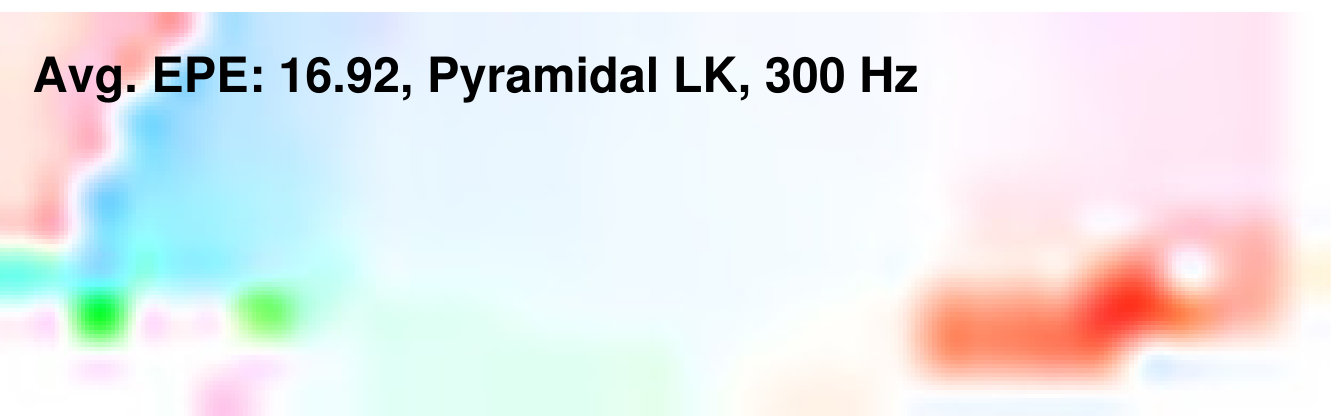}&
\includegraphics[width=0.195\textwidth]{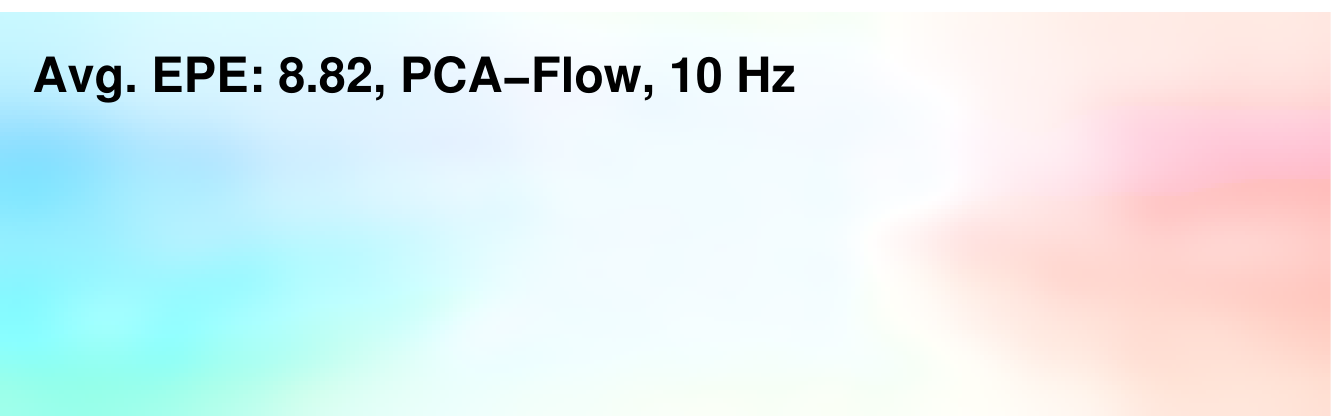}&
\includegraphics[width=0.195\textwidth]{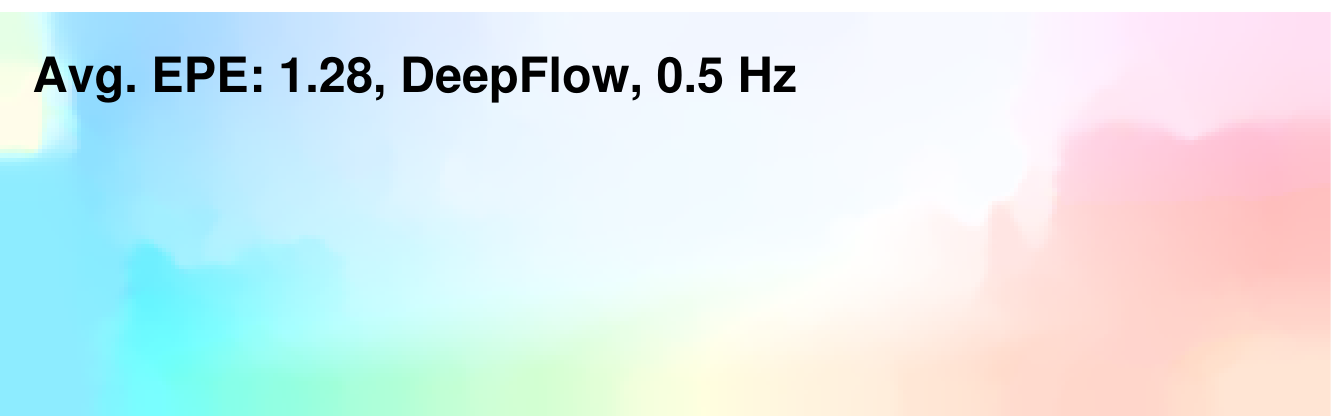}&
\includegraphics[width=0.195\textwidth]{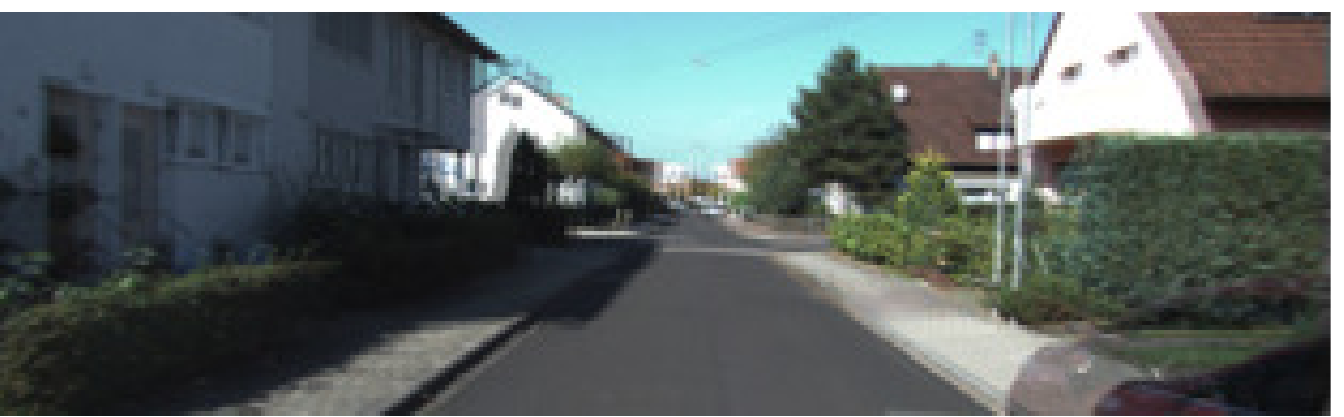}\\[6pt]
\includegraphics[width=0.195\textwidth]{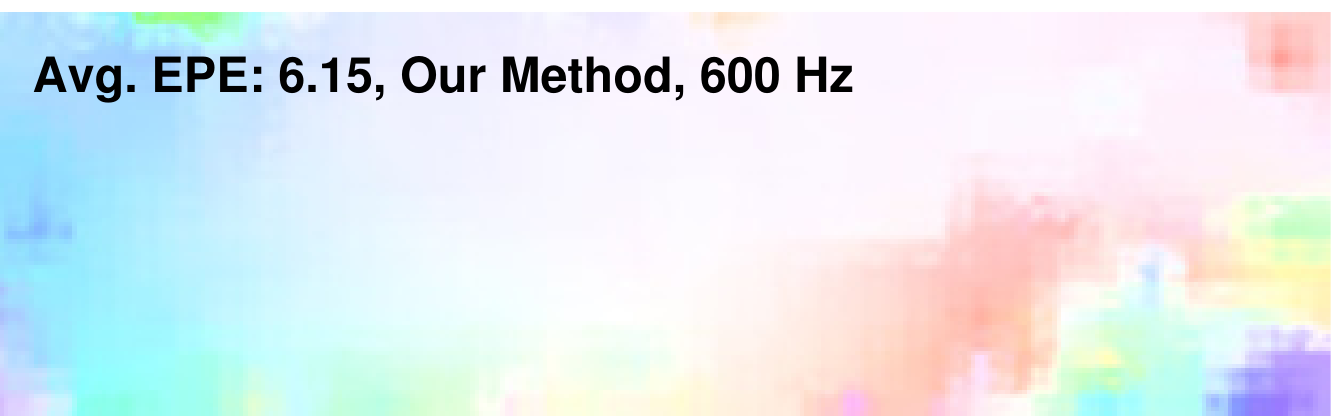}&
\includegraphics[width=0.195\textwidth]{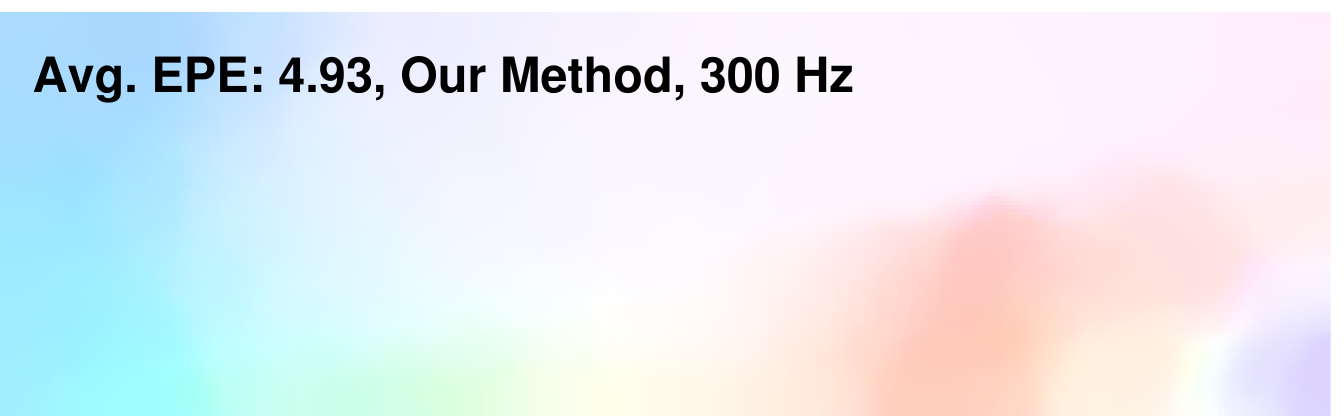}&
\includegraphics[width=0.195\textwidth]{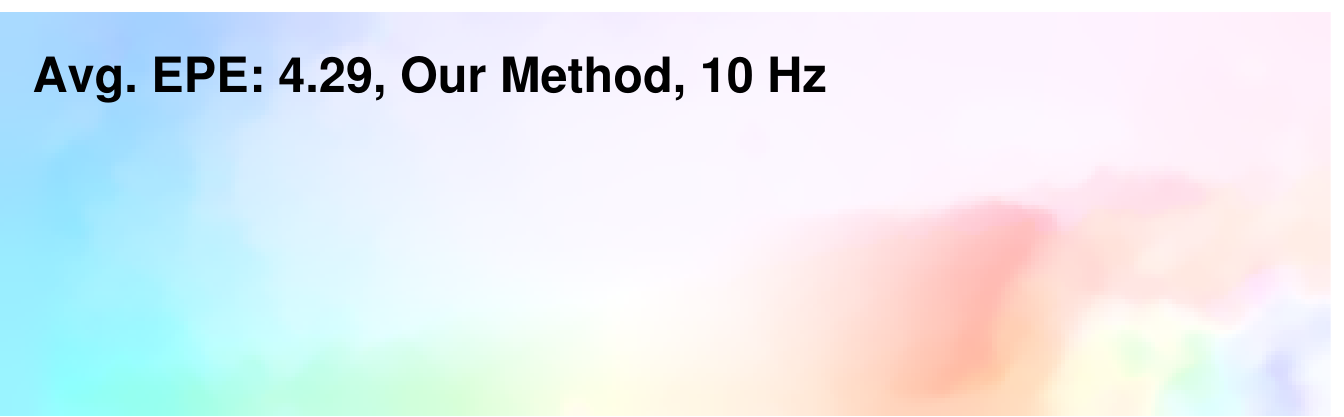}&
\includegraphics[width=0.195\textwidth]{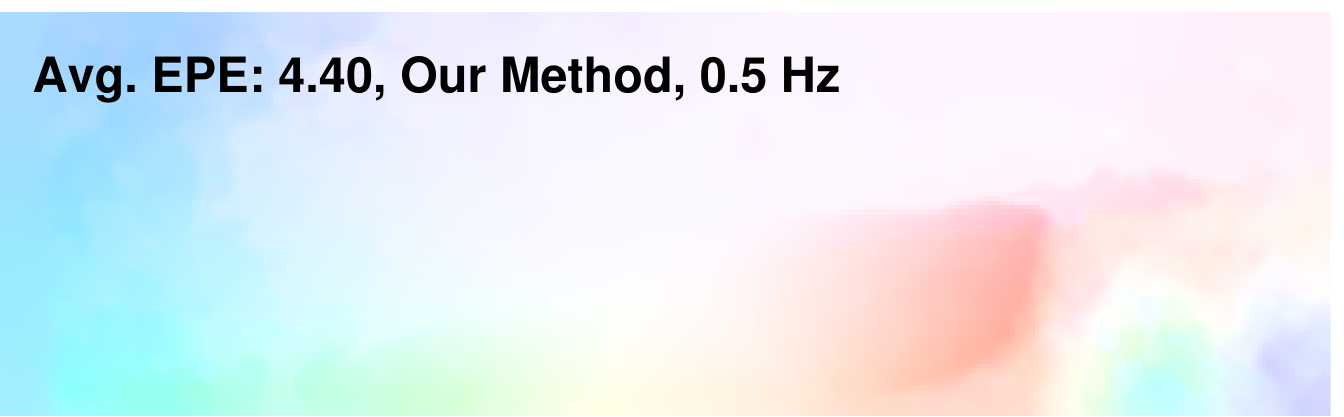}&
\includegraphics[width=0.195\textwidth]{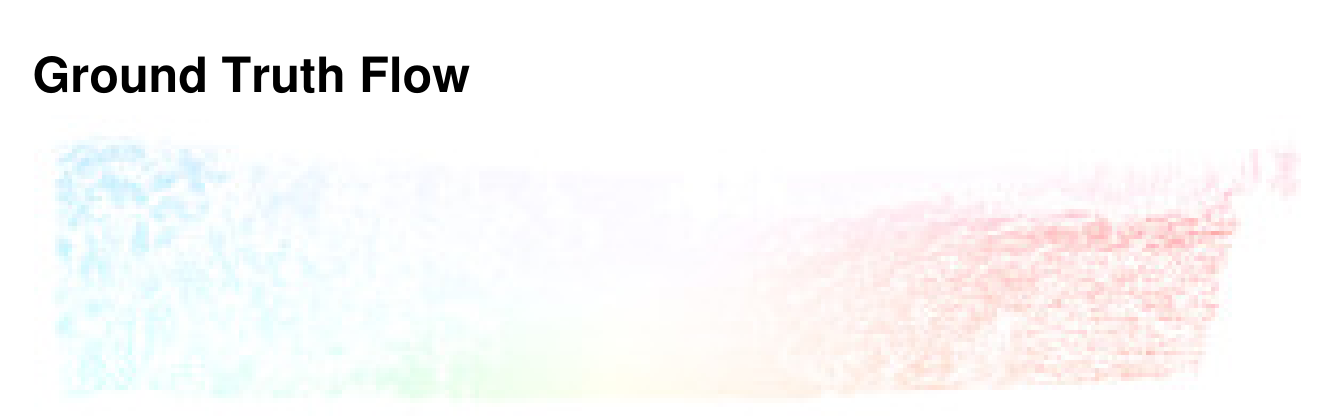}\\
\includegraphics[width=0.195\textwidth]{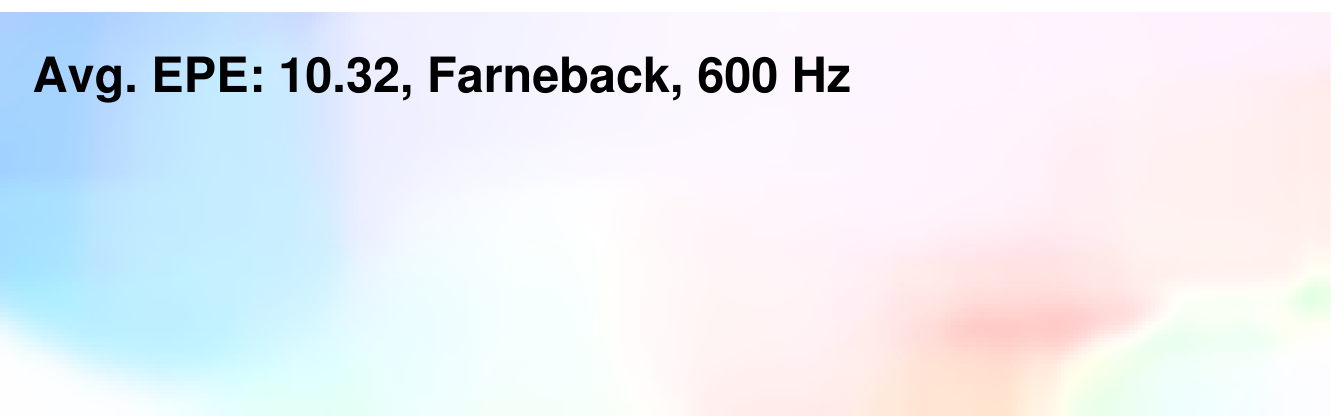}&
\includegraphics[width=0.195\textwidth]{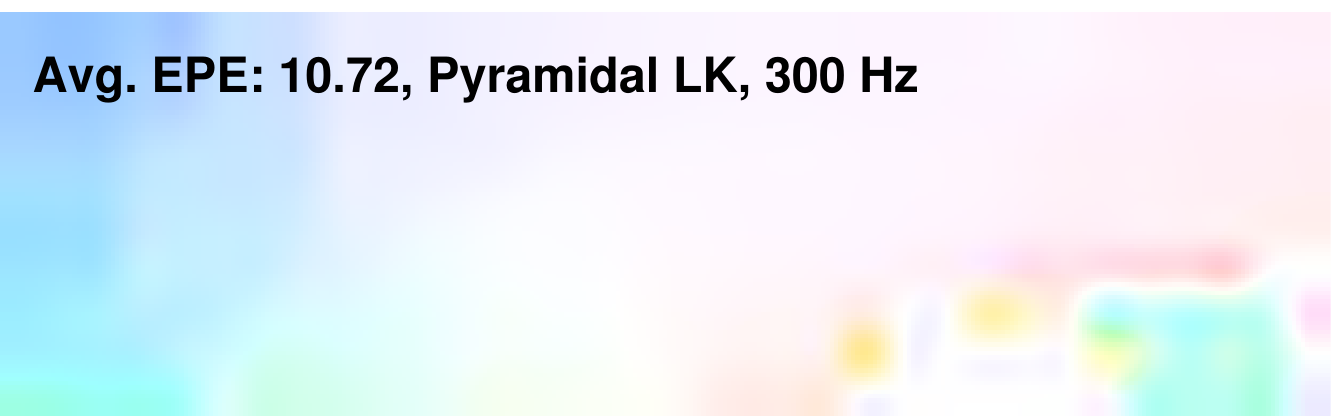}&
\includegraphics[width=0.195\textwidth]{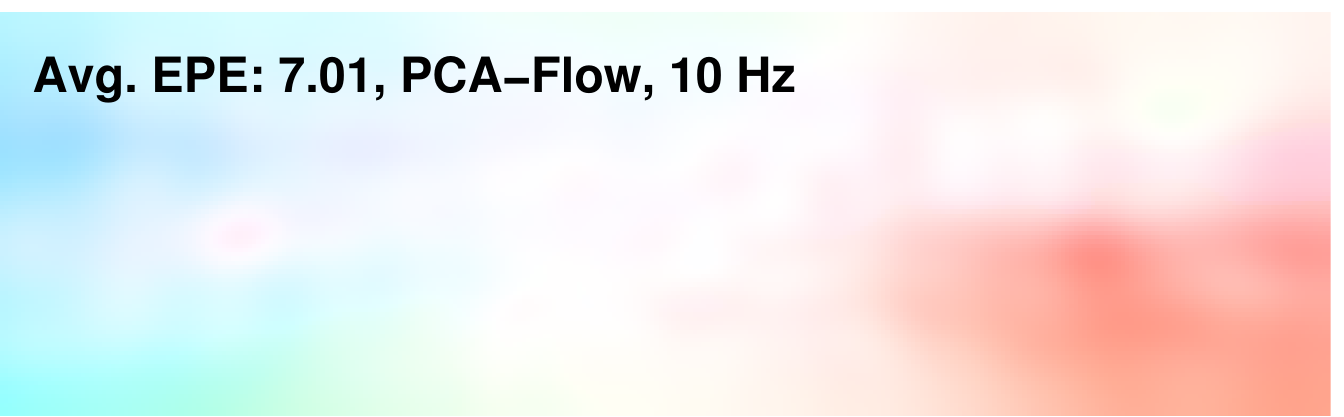}&
\includegraphics[width=0.195\textwidth]{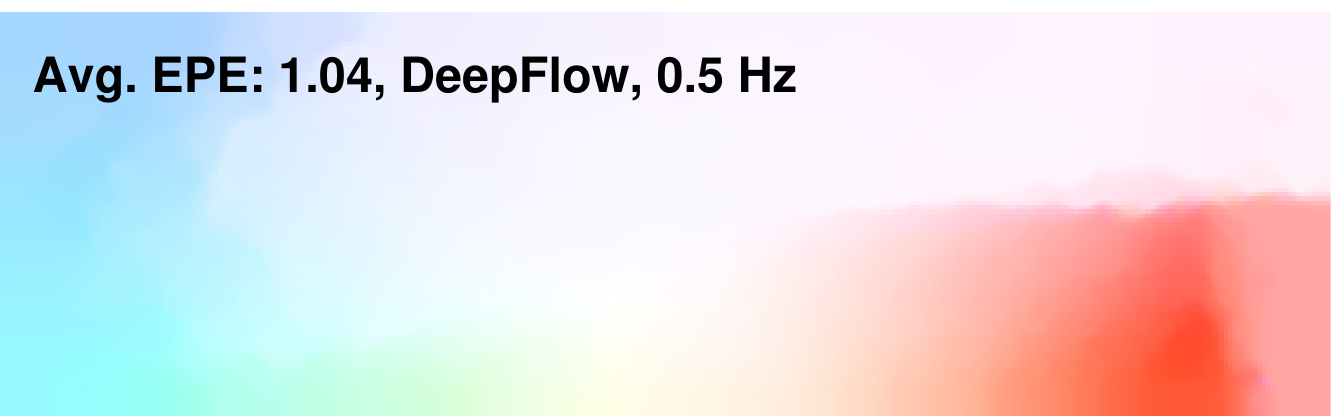}&
\includegraphics[width=0.195\textwidth]{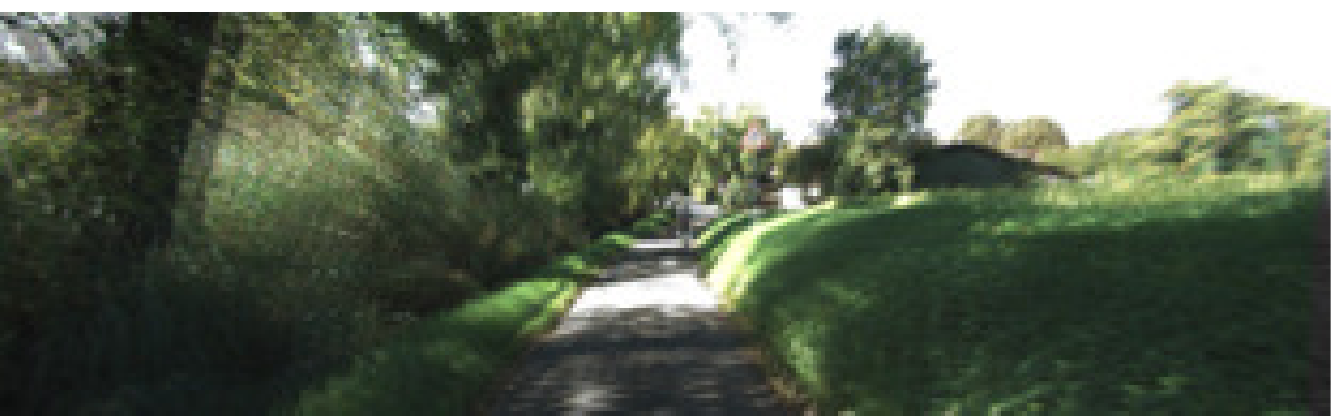}\\[6pt]
\includegraphics[width=0.195\textwidth]{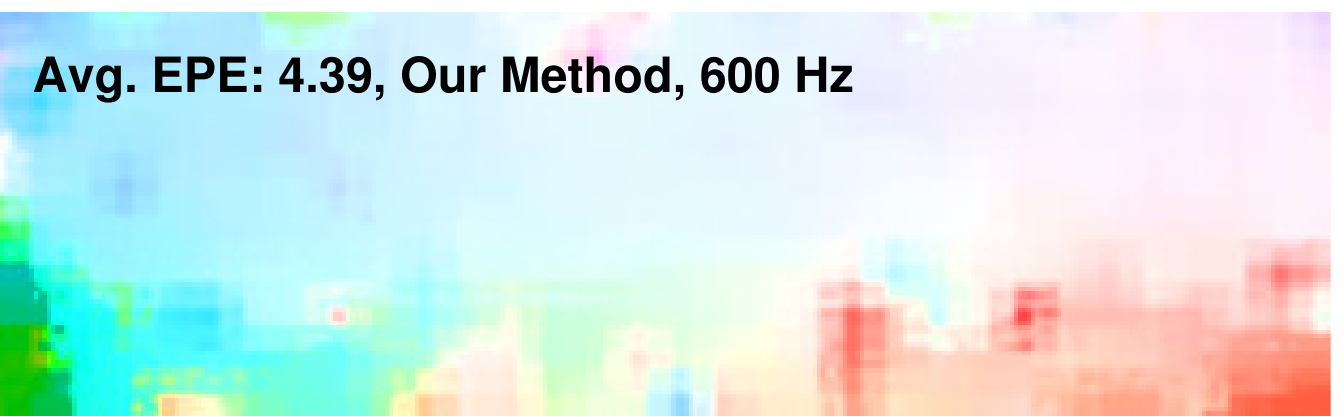}&
\includegraphics[width=0.195\textwidth]{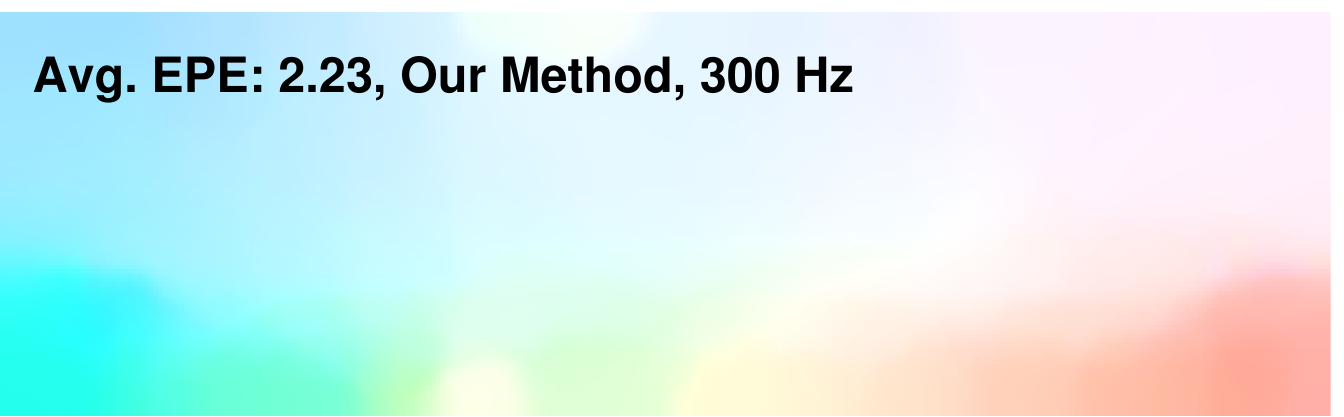}&
\includegraphics[width=0.195\textwidth]{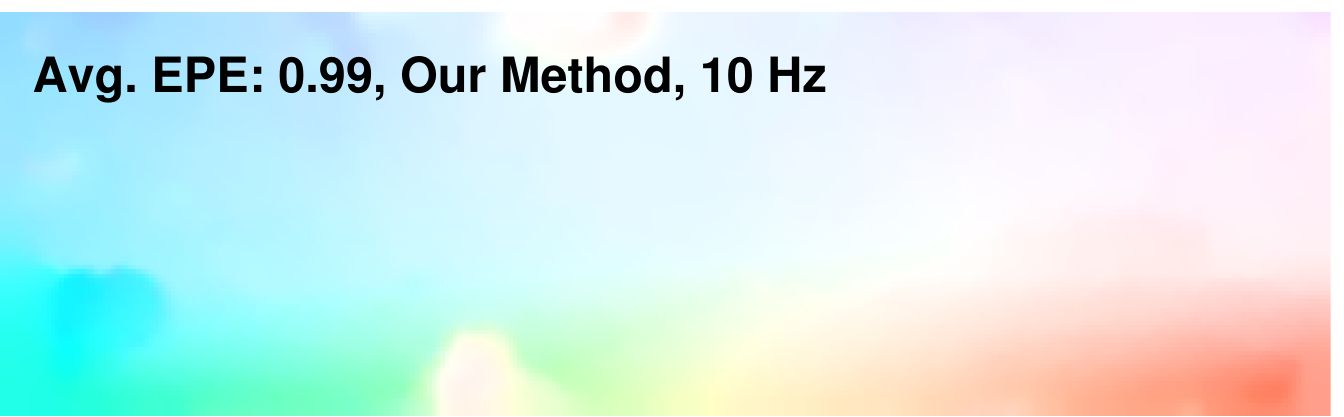}&
\includegraphics[width=0.195\textwidth]{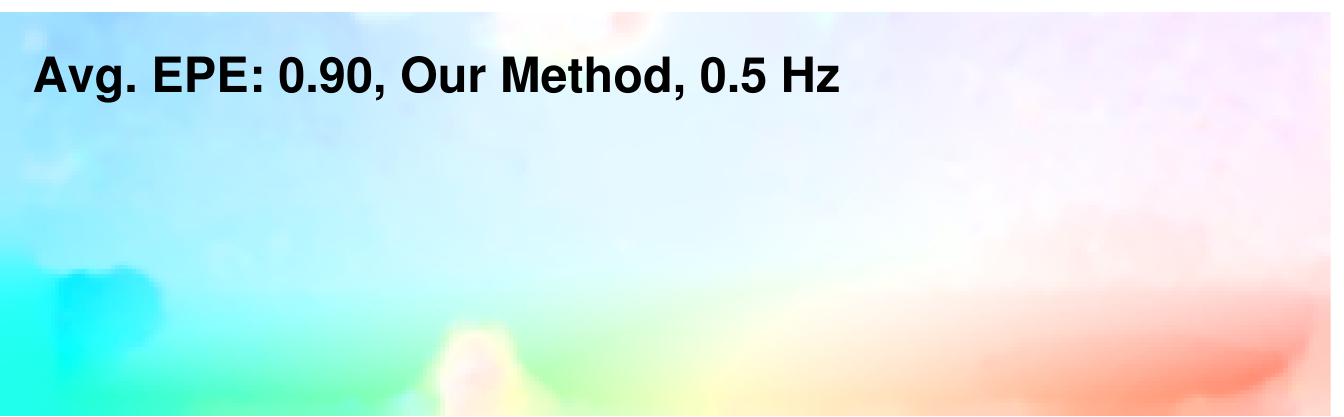}&
\includegraphics[width=0.195\textwidth]{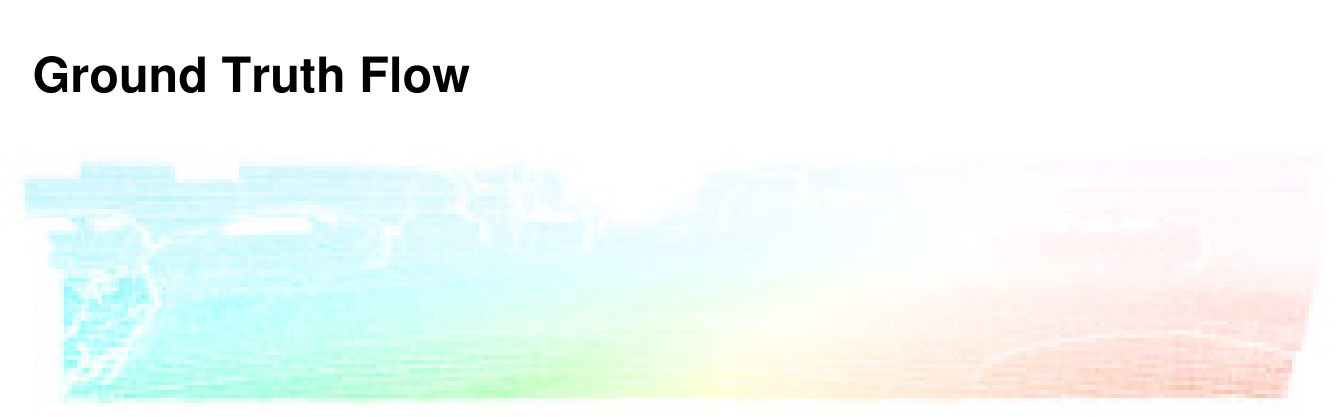}\\
\includegraphics[width=0.195\textwidth]{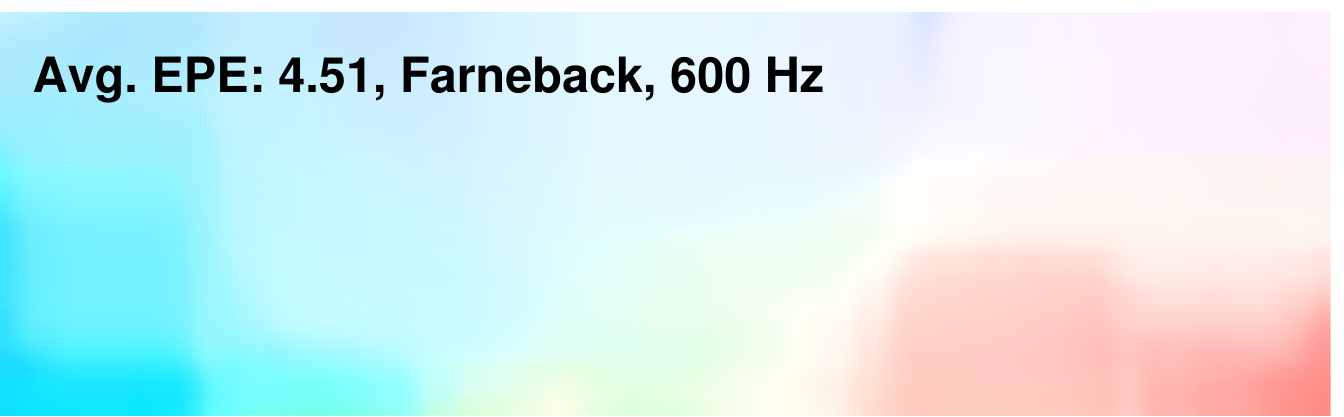}&
\includegraphics[width=0.195\textwidth]{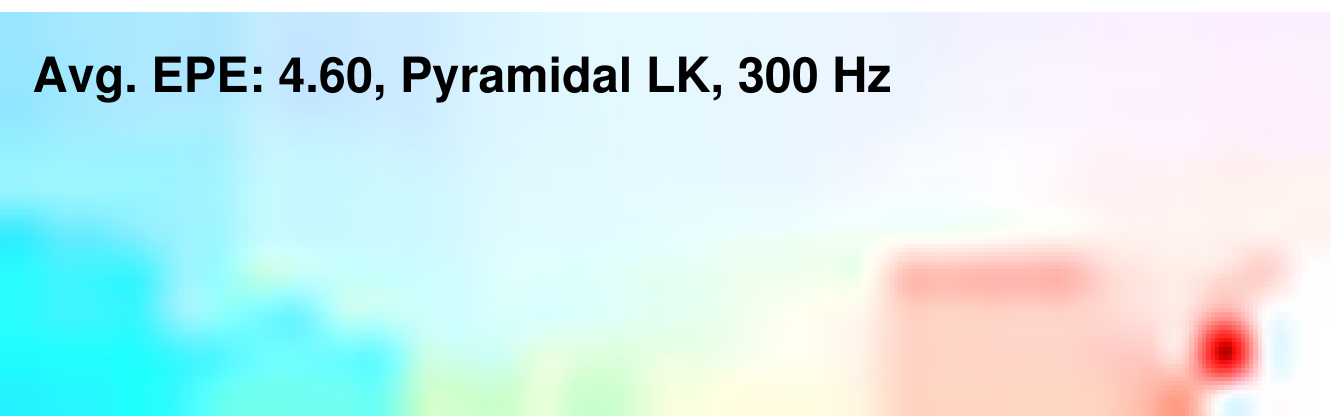}&
\includegraphics[width=0.195\textwidth]{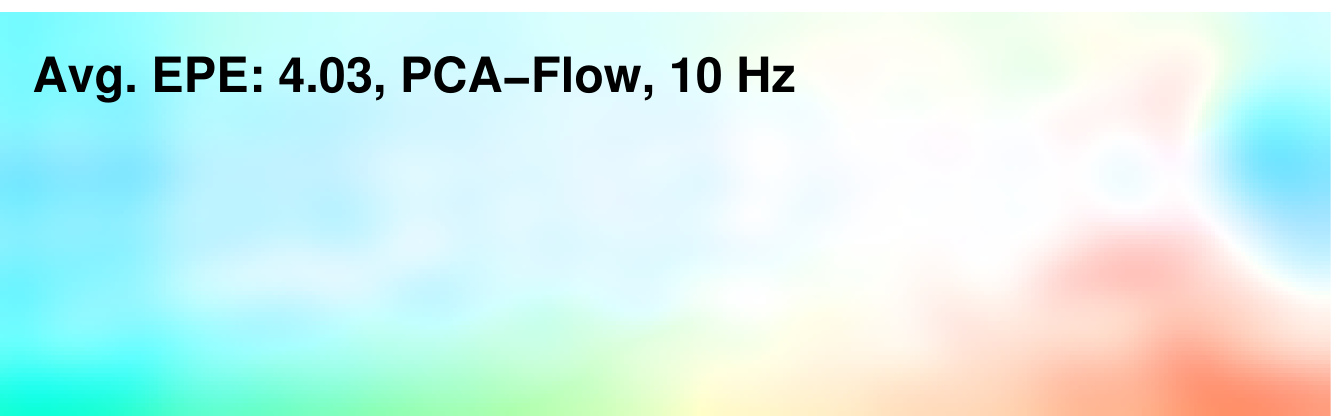}&
\includegraphics[width=0.195\textwidth]{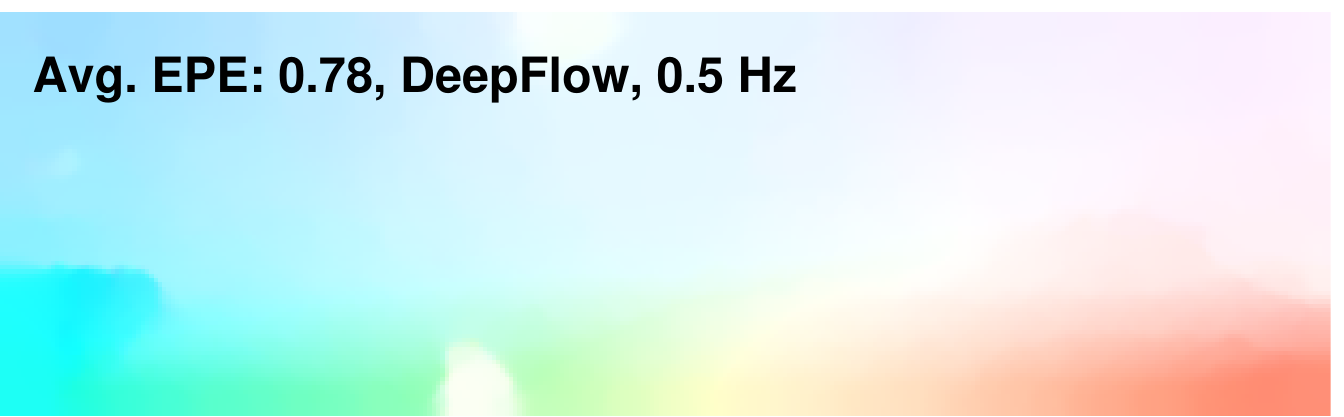}&
\includegraphics[width=0.195\textwidth]{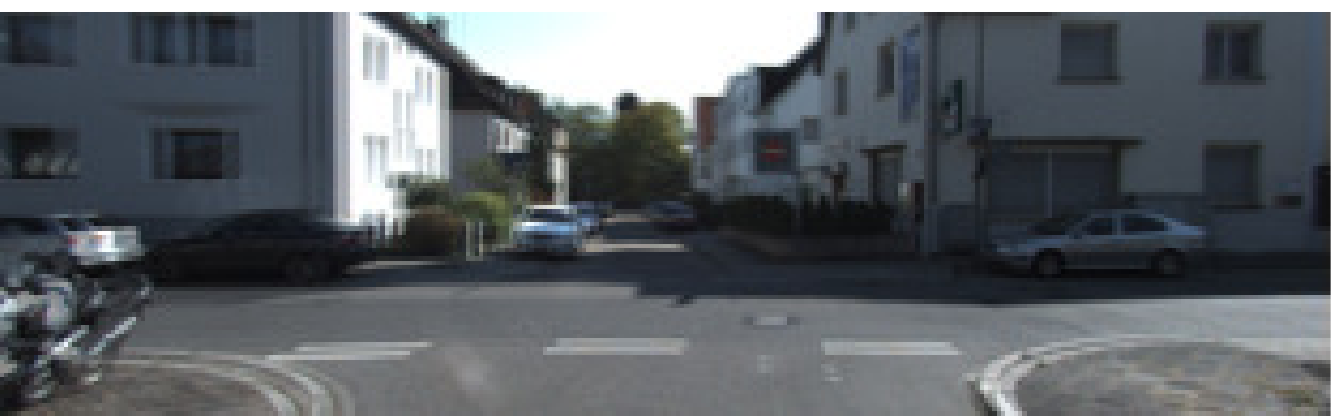}\\[6pt]
\includegraphics[width=0.195\textwidth]{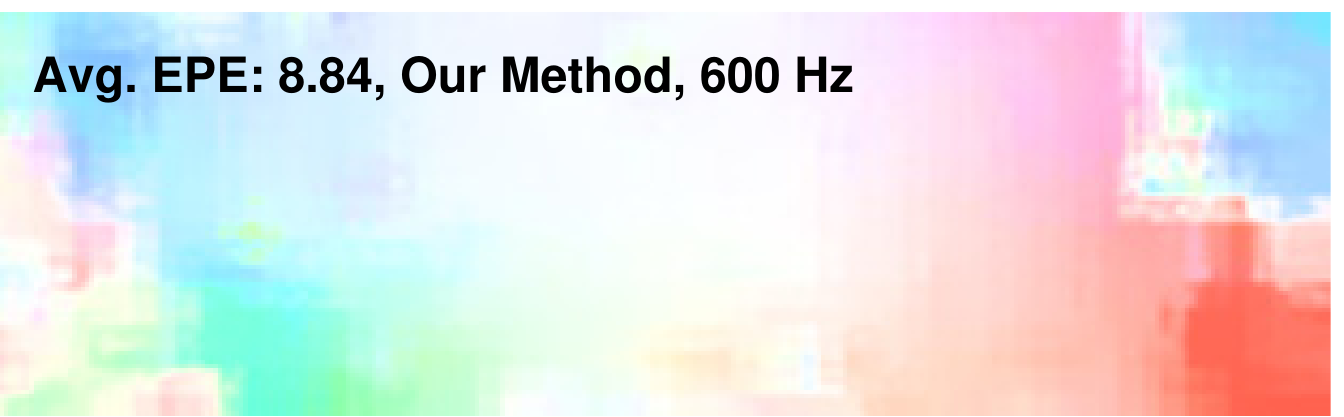}&
\includegraphics[width=0.195\textwidth]{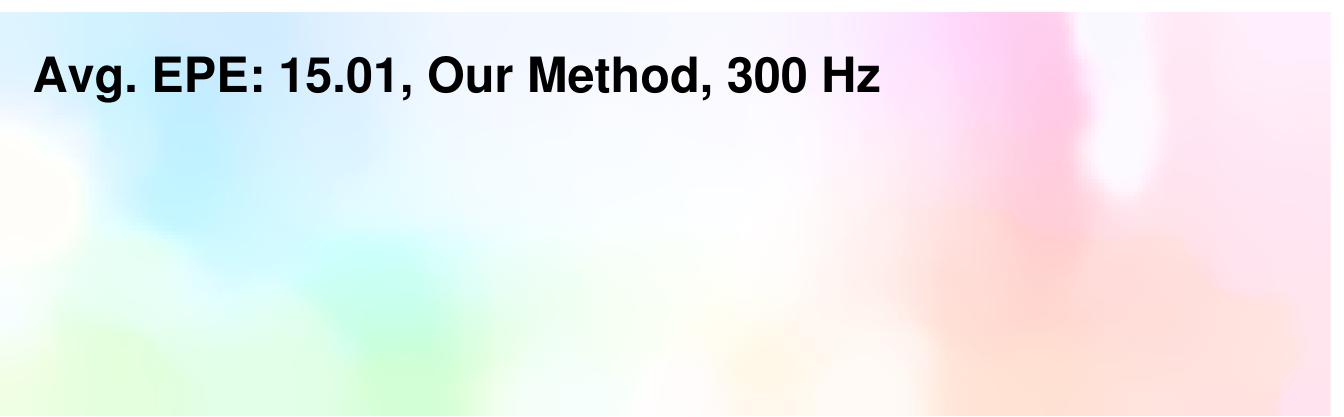}&
\includegraphics[width=0.195\textwidth]{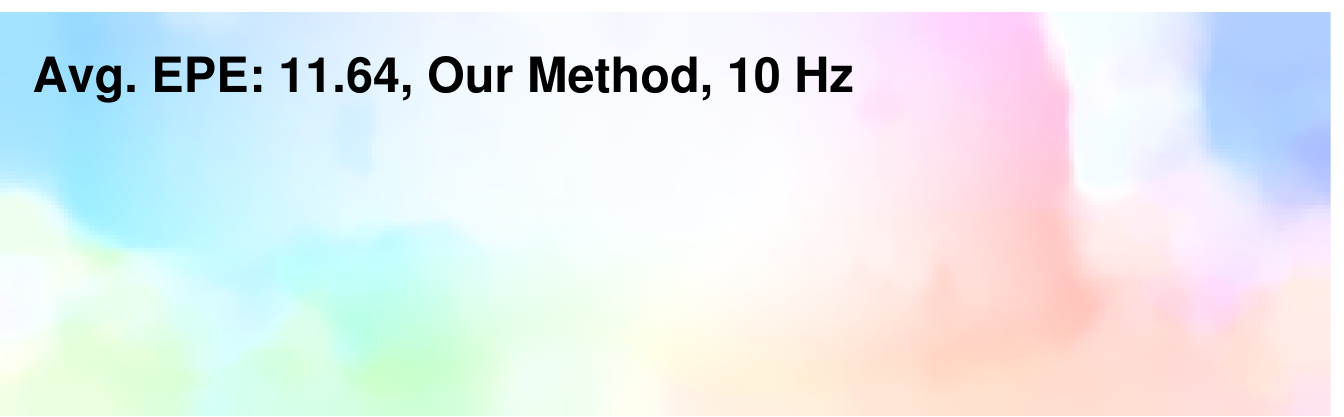}&
\includegraphics[width=0.195\textwidth]{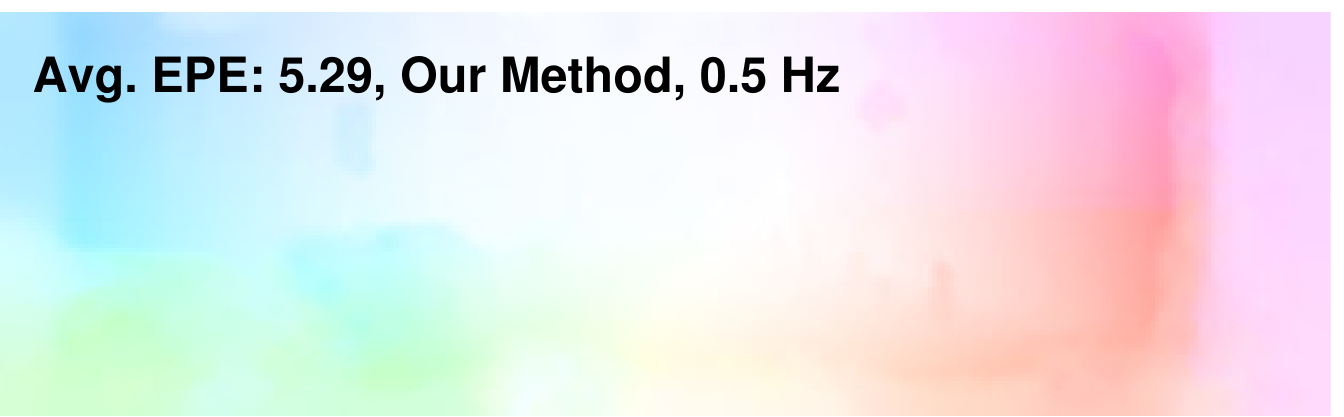}&
\includegraphics[width=0.195\textwidth]{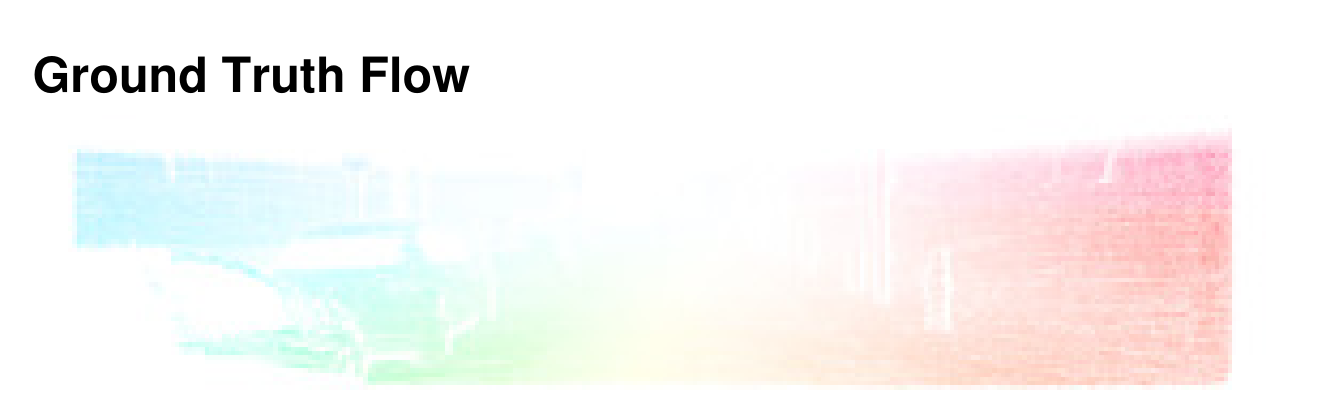}\\
\includegraphics[width=0.195\textwidth]{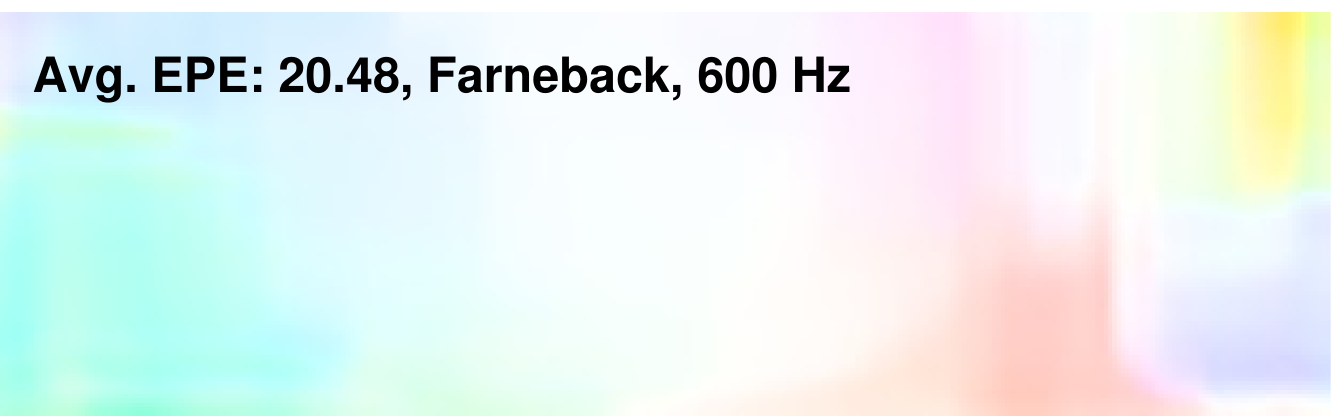}&
\includegraphics[width=0.195\textwidth]{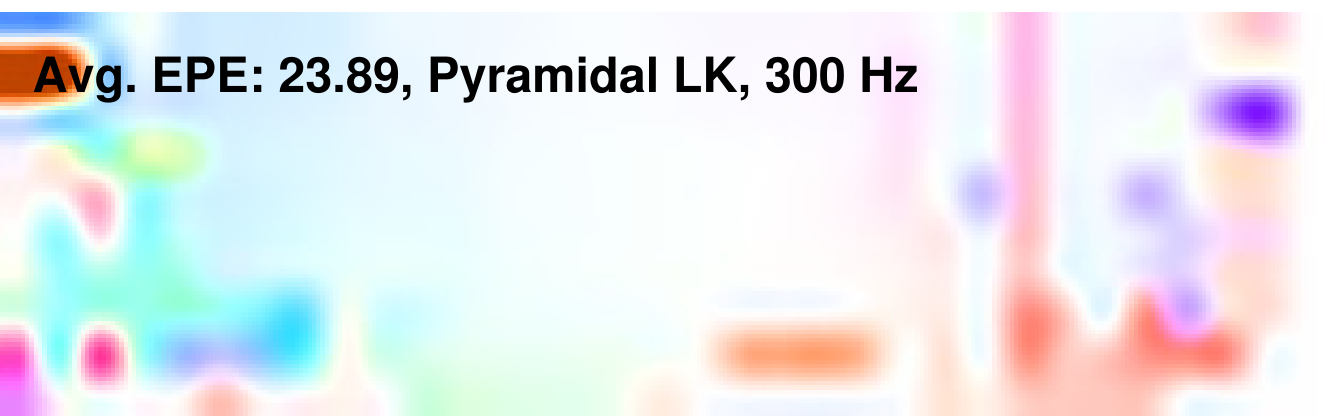}&
\includegraphics[width=0.195\textwidth]{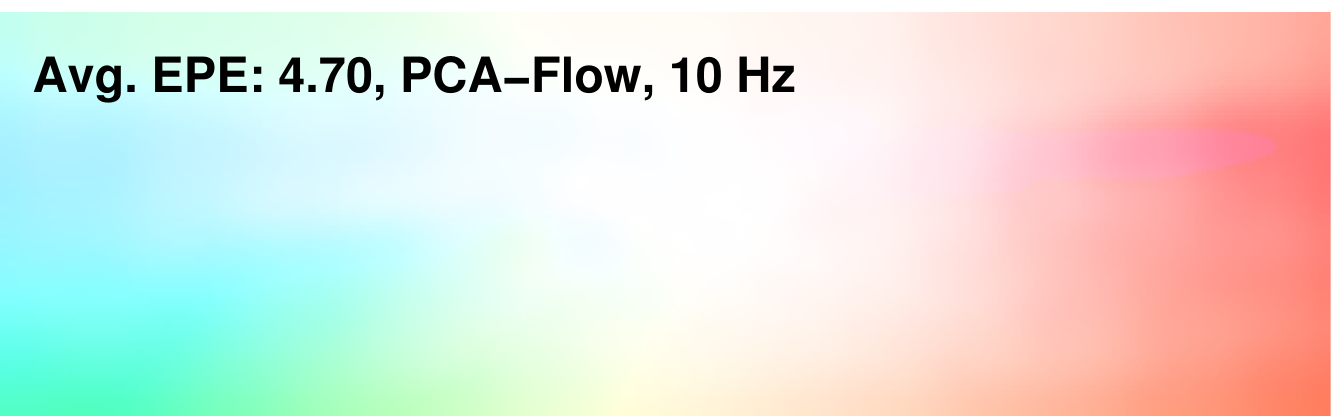}&
\includegraphics[width=0.195\textwidth]{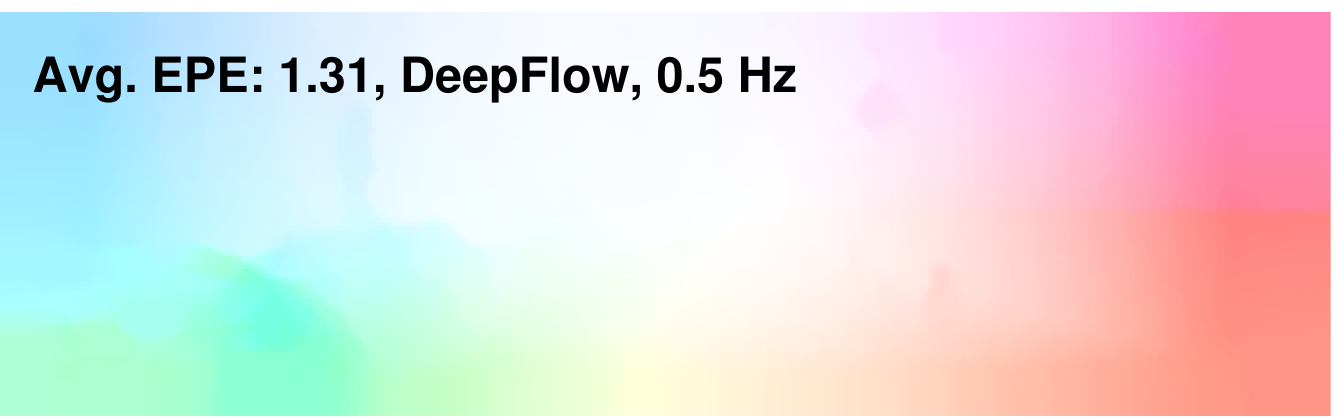}&
\includegraphics[width=0.195\textwidth]{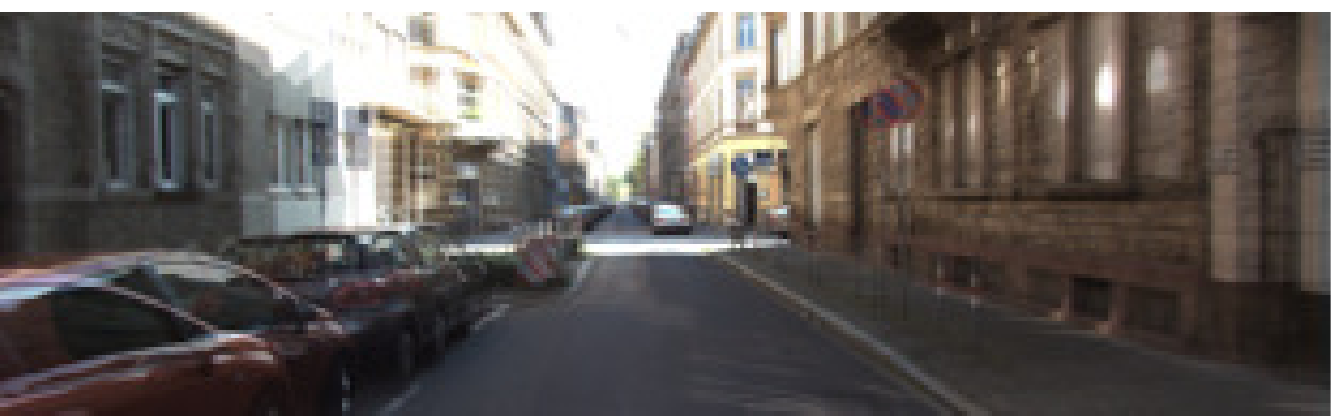}\\[6pt]
\includegraphics[width=0.195\textwidth]{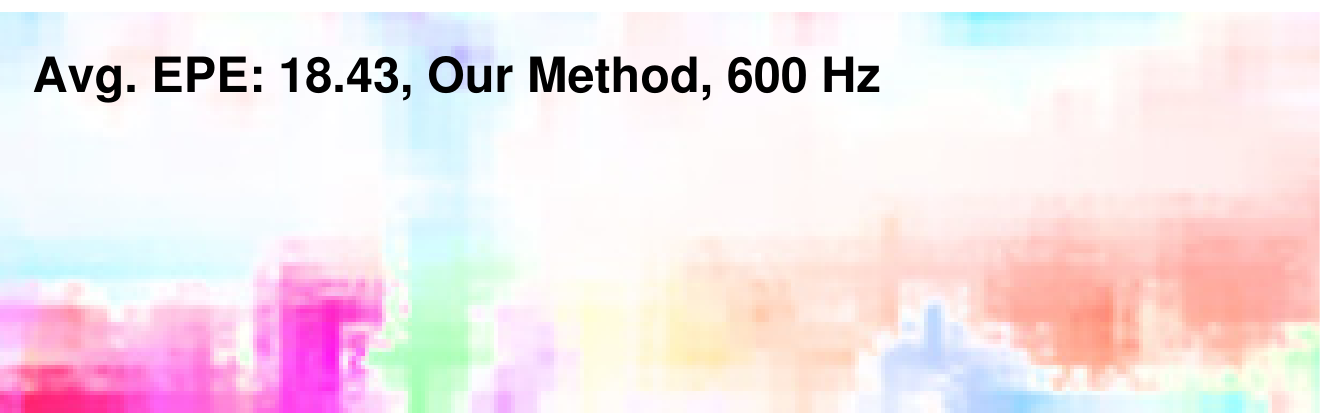}&
\includegraphics[width=0.195\textwidth]{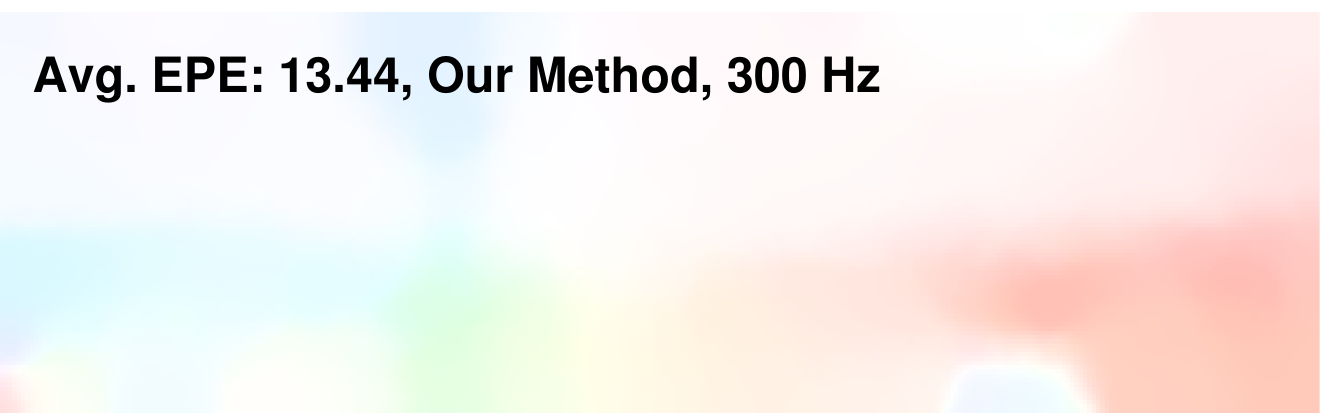}&
\includegraphics[width=0.195\textwidth]{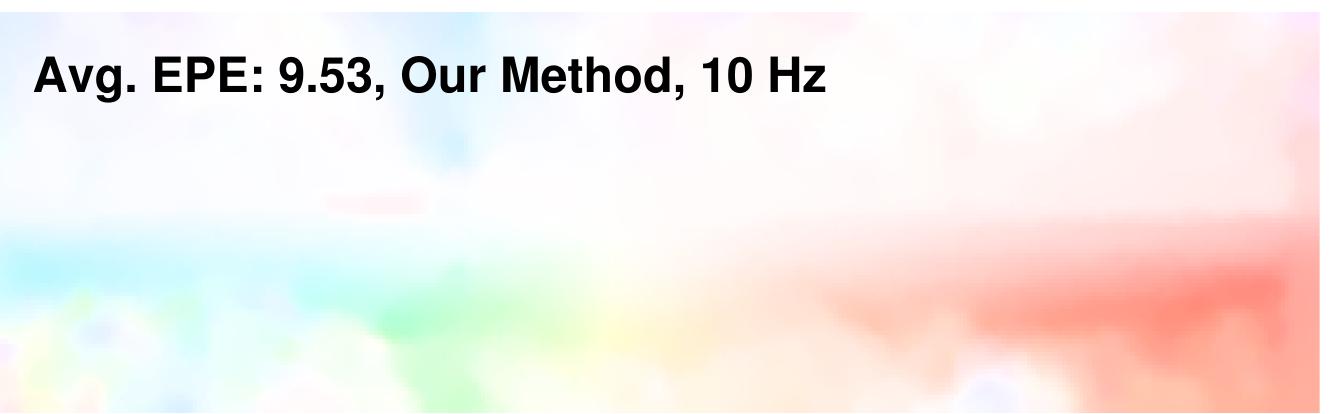}&
\includegraphics[width=0.195\textwidth]{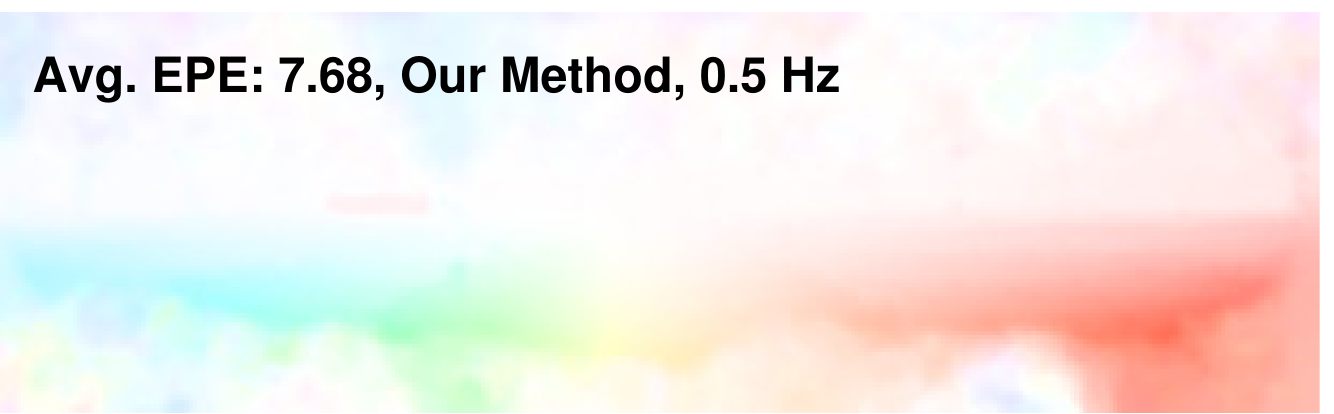}&
\includegraphics[width=0.195\textwidth]{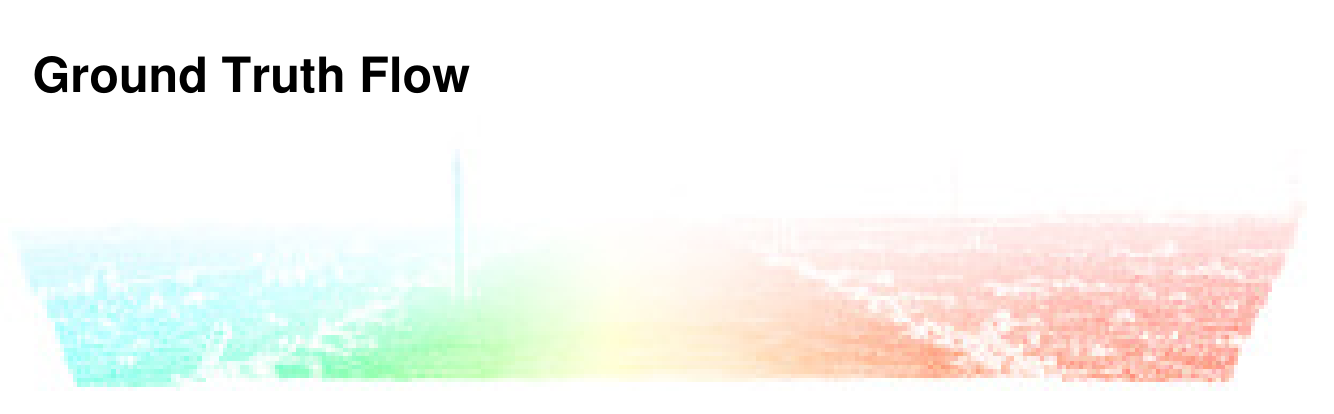}\\
\includegraphics[width=0.195\textwidth]{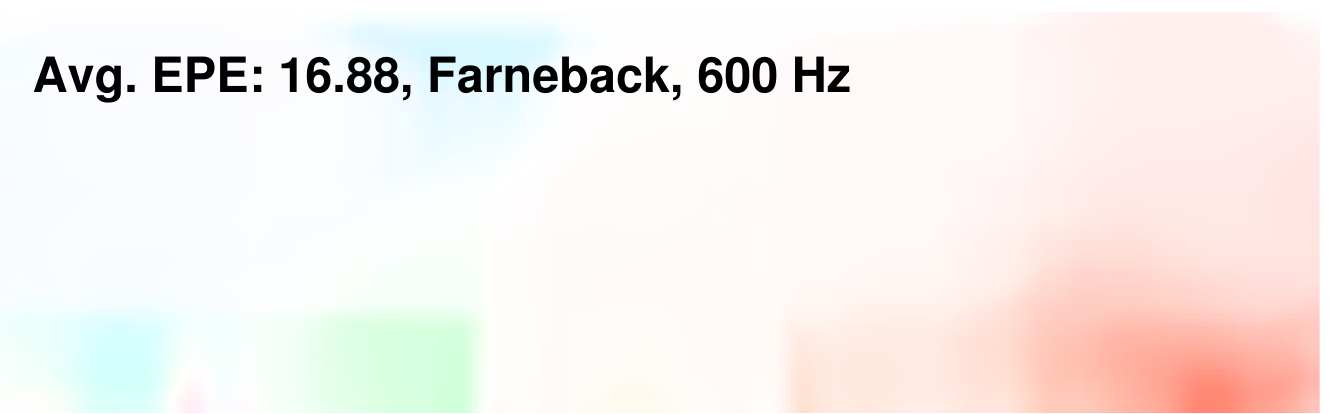}&
\includegraphics[width=0.195\textwidth]{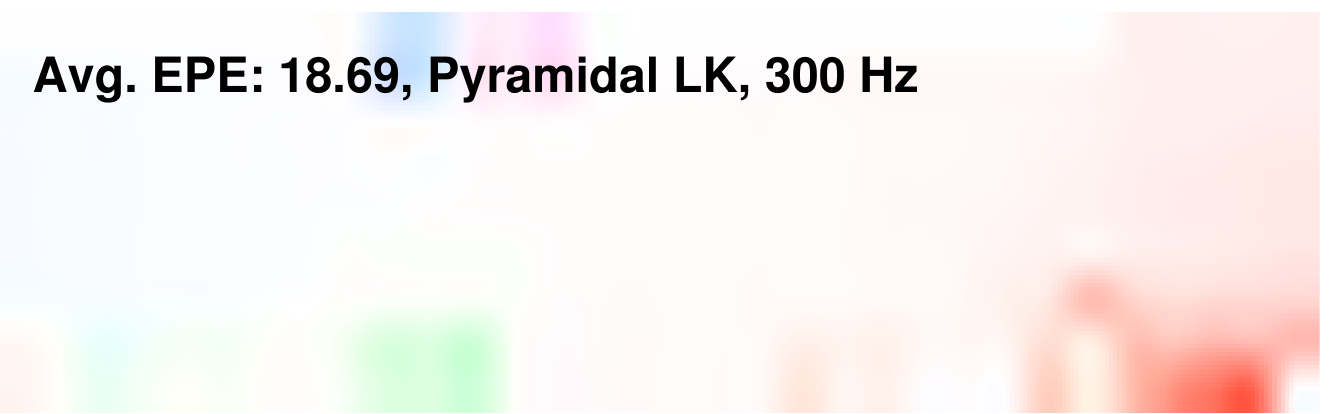}&
\includegraphics[width=0.195\textwidth]{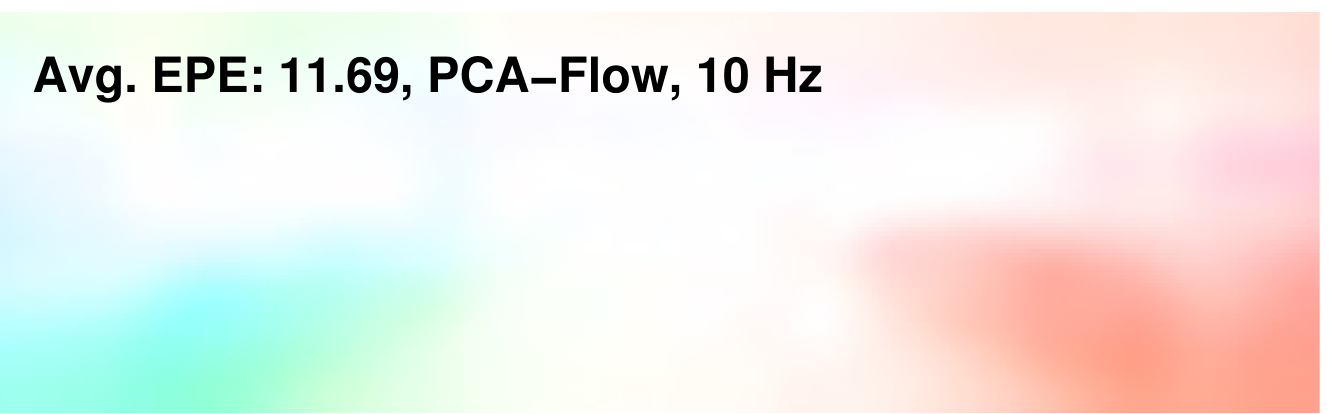}&
\includegraphics[width=0.195\textwidth]{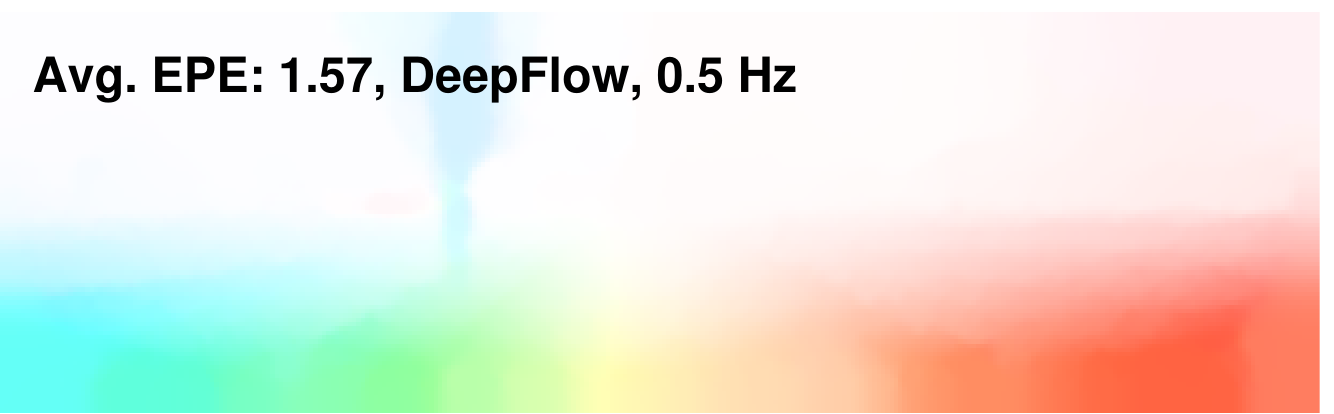}&
\includegraphics[width=0.195\textwidth]{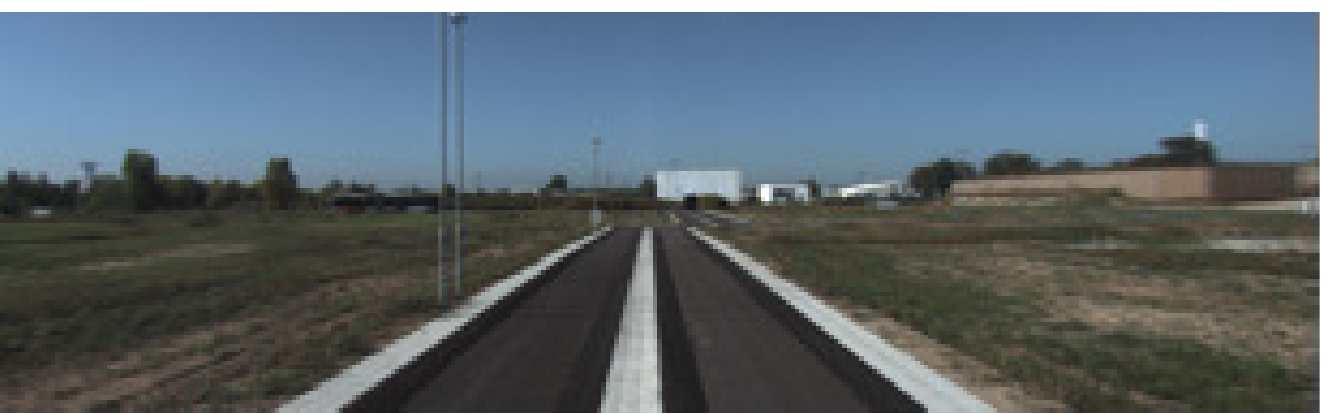}\\[6pt]
\end{tabular}
}
\caption{Exemplary results on KITTI (training). In each block of $2 \times 6$ images.  Top row, left to right: Our method for operating points ({\bf 1})-({\bf 4}), Ground Truth. Bottom row: Farneback 600Hz, Pyramidal LK 300Hz, PCA-Flow 10Hz, DeepFlow 0.5Hz, Original
Image.}\label{fig:kitti1res_AP} 
\end{figure*}

\begin{figure*} [!ht]
\centering\setlength{\tabcolsep}{0.1pt}\renewcommand{\arraystretch}{0} 
{
\begin{tabular}{ccccc}
 {\bf 600Hz} & {\bf 300Hz} & {\bf 10Hz} & {\bf 0.5Hz}& {\bf Ground Truth}\\
\includegraphics[width=0.195\textwidth]{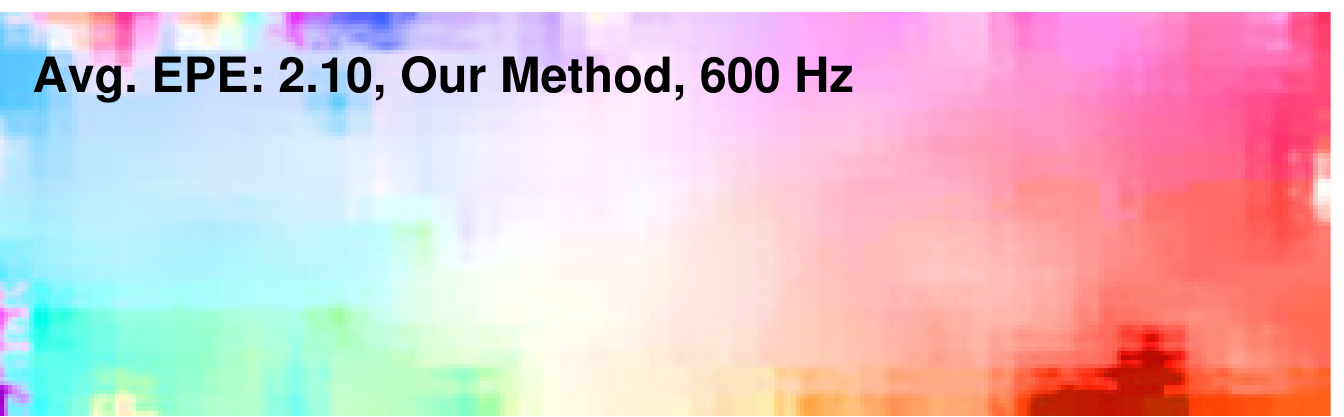}&
\includegraphics[width=0.195\textwidth]{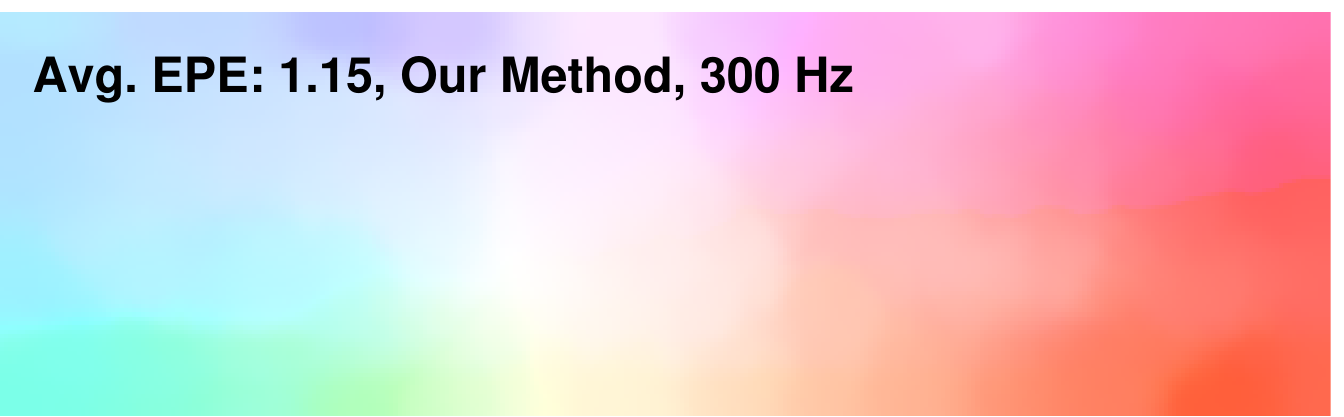}&
\includegraphics[width=0.195\textwidth]{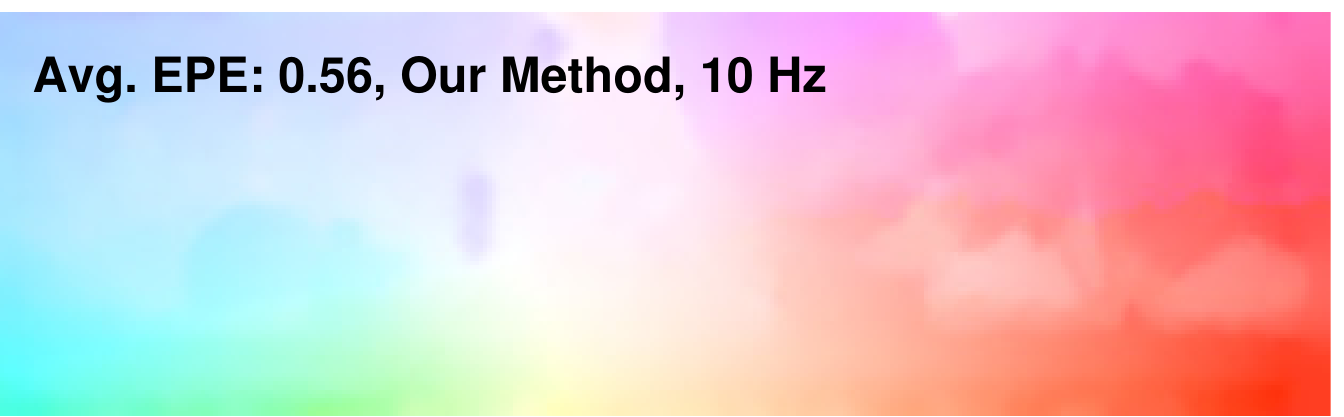}&
\includegraphics[width=0.195\textwidth]{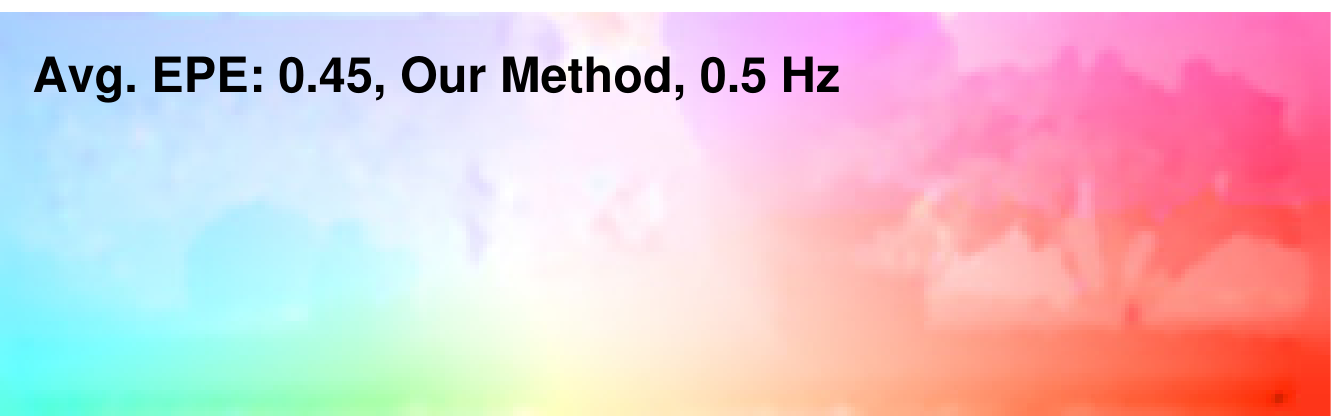}&
\includegraphics[width=0.195\textwidth]{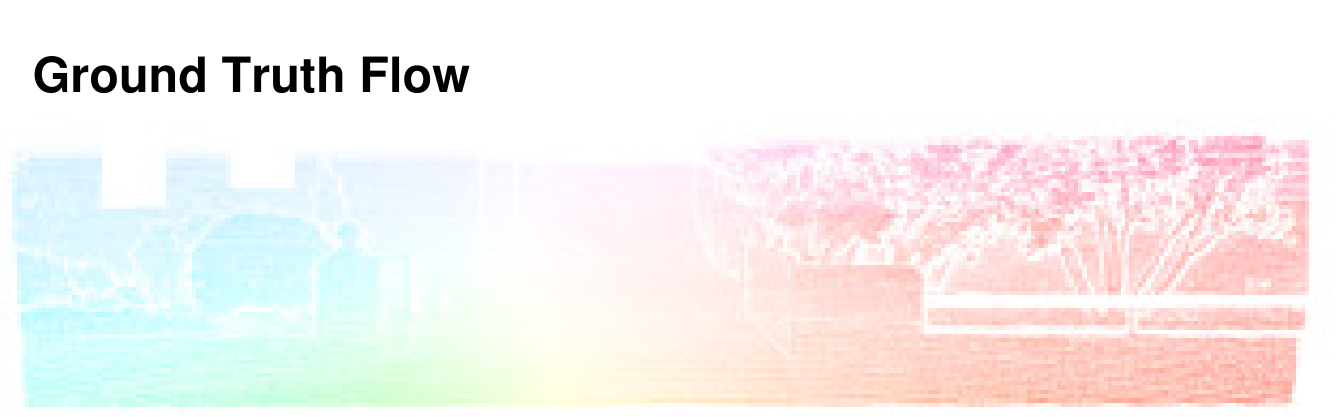}\\
\includegraphics[width=0.195\textwidth]{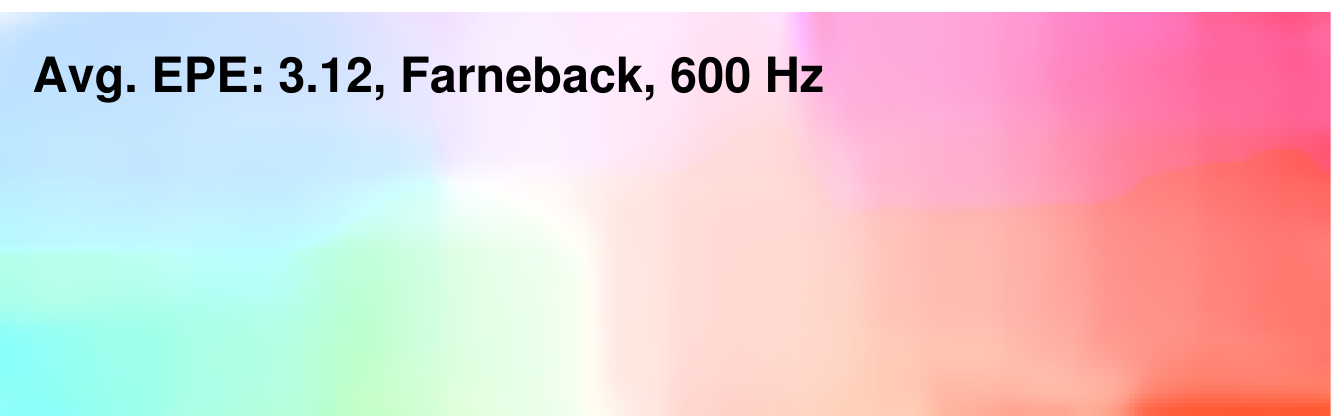}&
\includegraphics[width=0.195\textwidth]{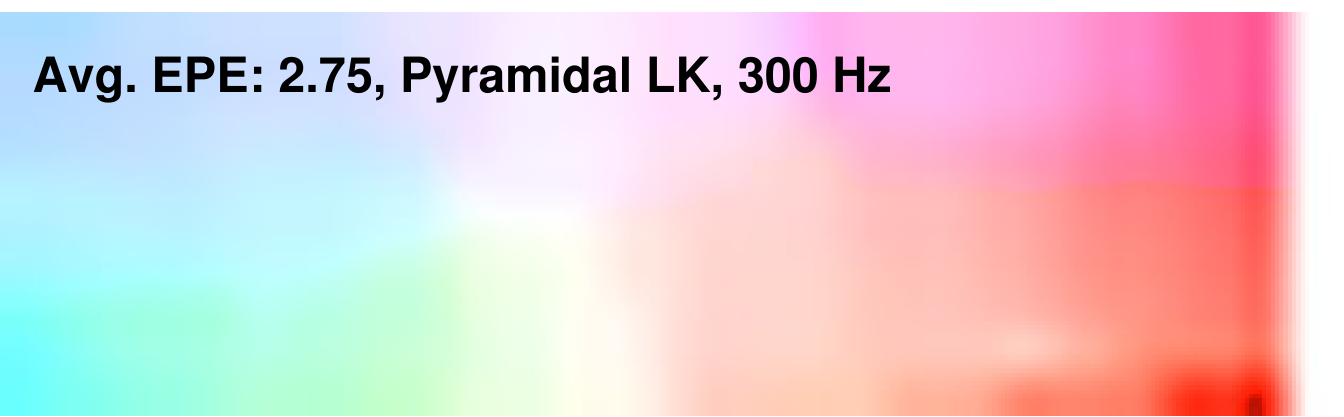}&
\includegraphics[width=0.195\textwidth]{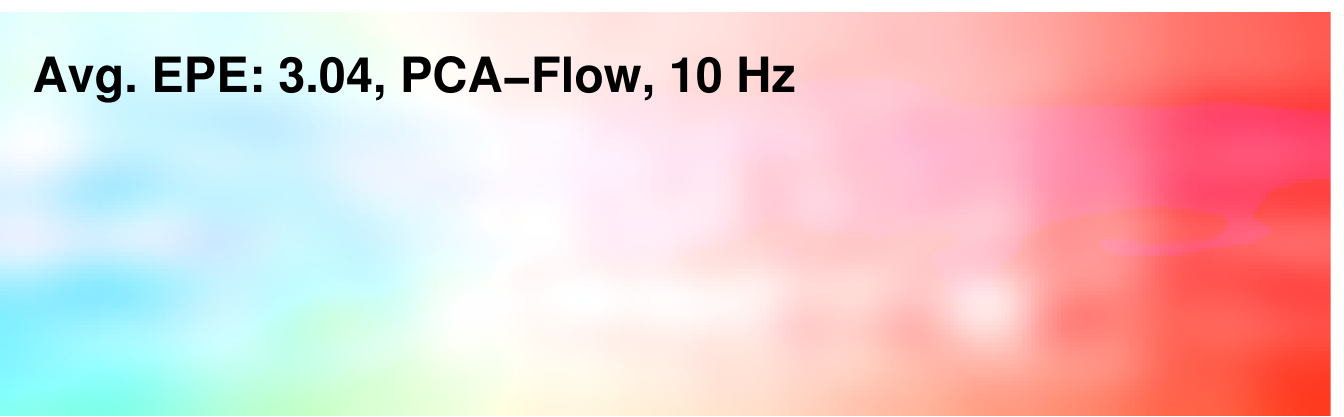}&
\includegraphics[width=0.195\textwidth]{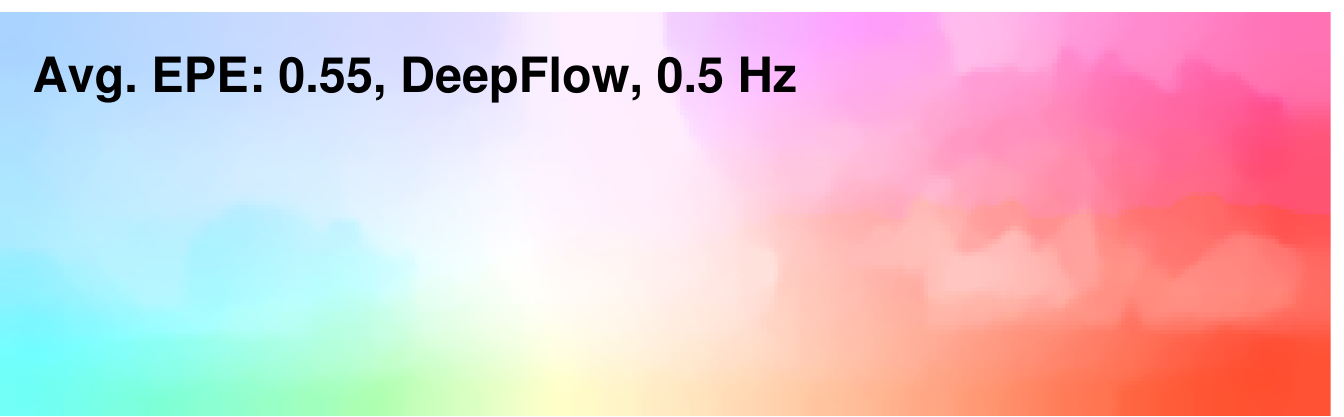}&
\includegraphics[width=0.195\textwidth]{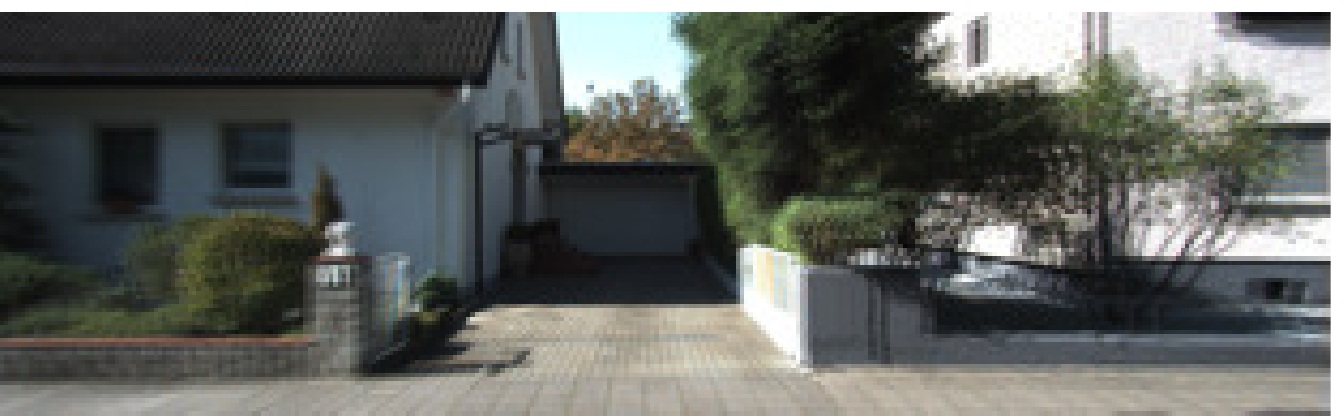}\\[6pt]
\includegraphics[width=0.195\textwidth]{imgs_exa/eximg-kitti-0113-01.pdf}&
\includegraphics[width=0.195\textwidth]{imgs_exa/eximg-kitti-0113-02.pdf}&
\includegraphics[width=0.195\textwidth]{imgs_exa/eximg-kitti-0113-03.pdf}&
\includegraphics[width=0.195\textwidth]{imgs_exa/eximg-kitti-0113-04.pdf}&
\includegraphics[width=0.195\textwidth]{imgs_exa/eximg-kitti-0113-05.pdf}\\
\includegraphics[width=0.195\textwidth]{imgs_exa/eximg-kitti-0113-06.pdf}&
\includegraphics[width=0.195\textwidth]{imgs_exa/eximg-kitti-0113-07.pdf}&
\includegraphics[width=0.195\textwidth]{imgs_exa/eximg-kitti-0113-08.pdf}&
\includegraphics[width=0.195\textwidth]{imgs_exa/eximg-kitti-0113-09.pdf}&
\includegraphics[width=0.195\textwidth]{imgs_exa/eximg-kitti-0113-10.pdf}\\[6pt]
\includegraphics[width=0.195\textwidth]{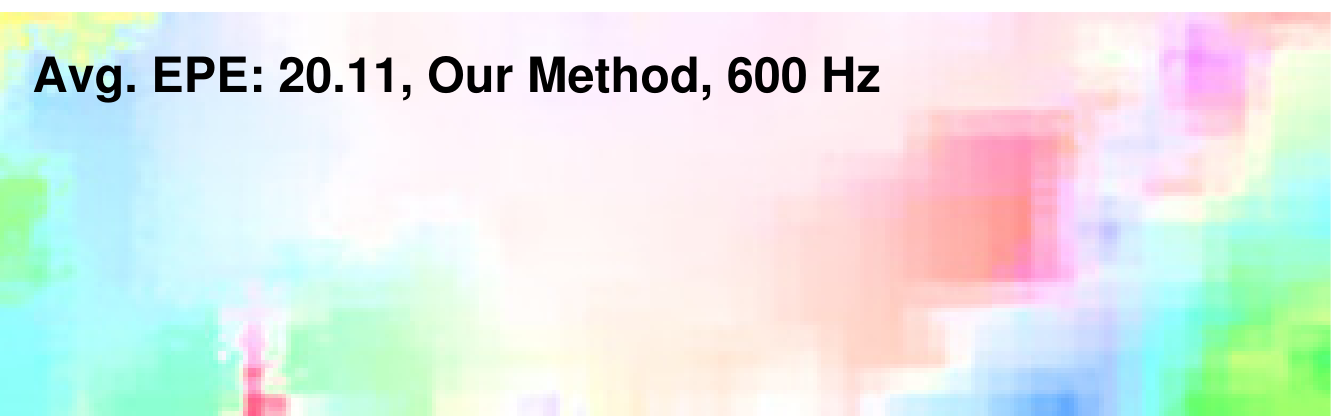}&
\includegraphics[width=0.195\textwidth]{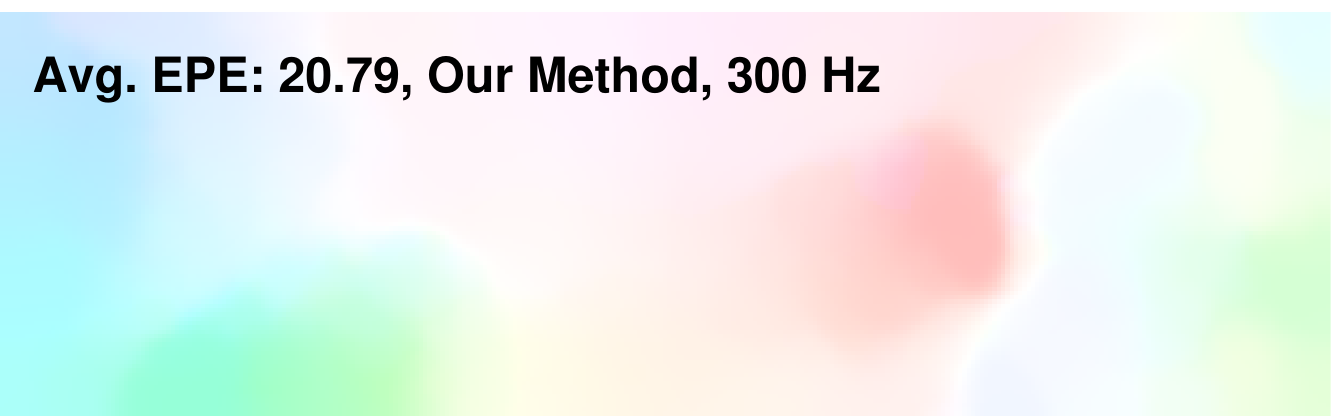}&
\includegraphics[width=0.195\textwidth]{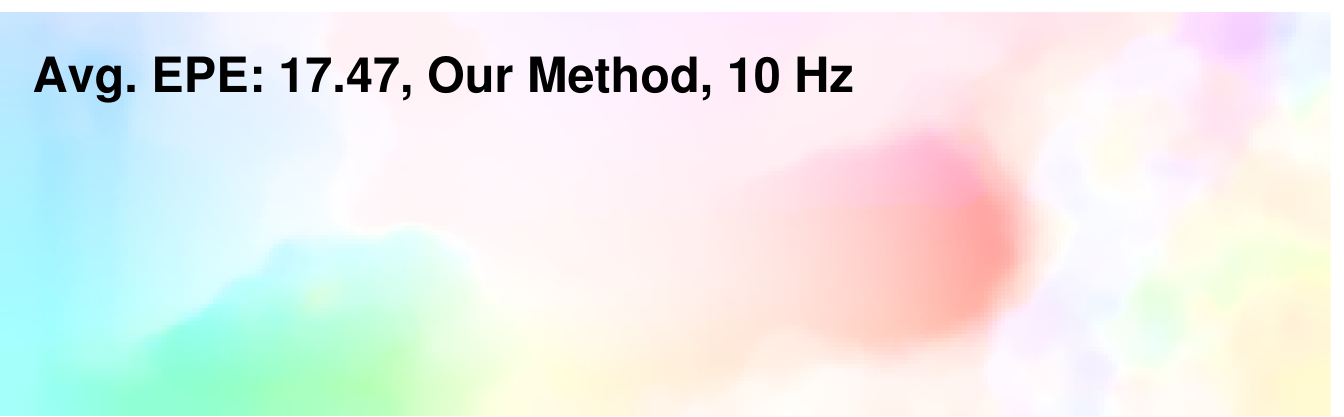}&
\includegraphics[width=0.195\textwidth]{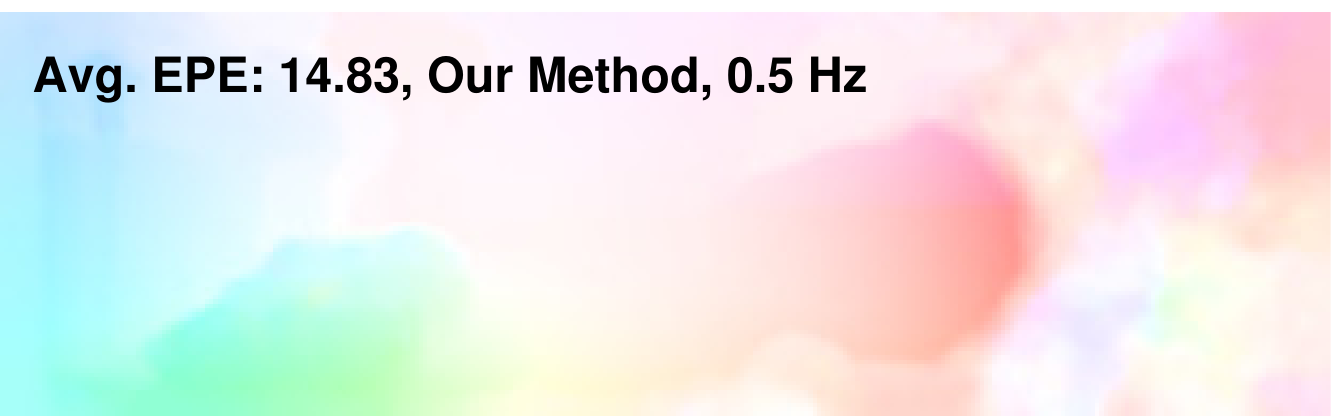}&
\includegraphics[width=0.195\textwidth]{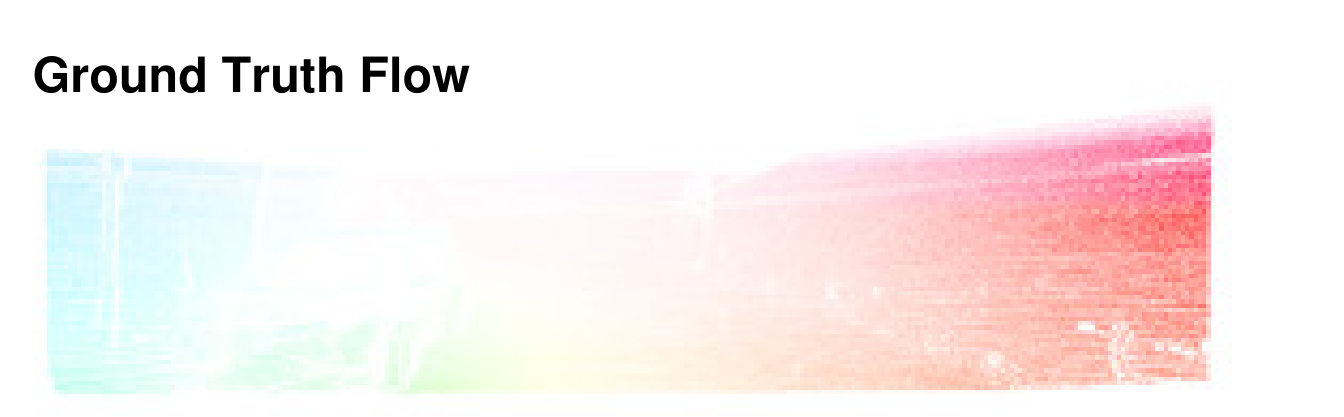}\\
\includegraphics[width=0.195\textwidth]{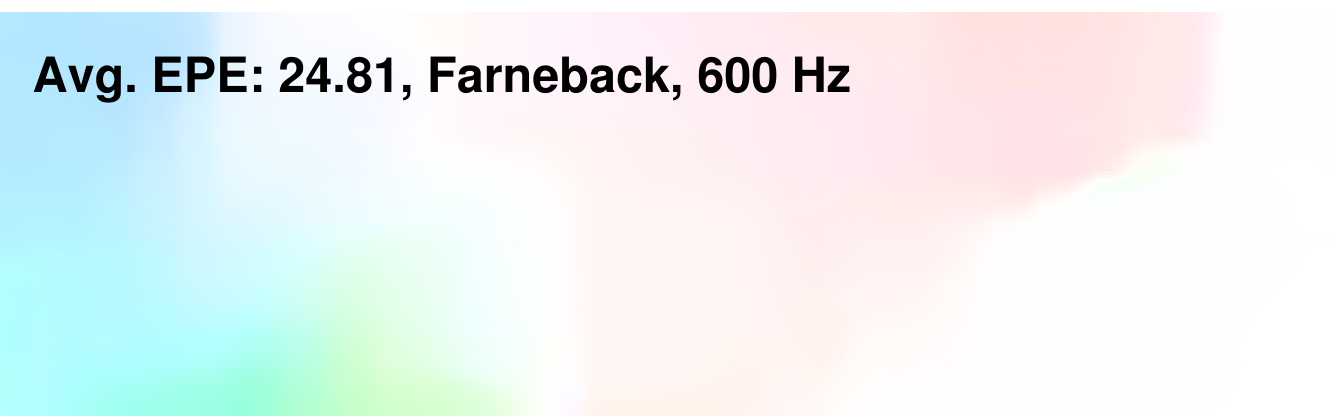}&
\includegraphics[width=0.195\textwidth]{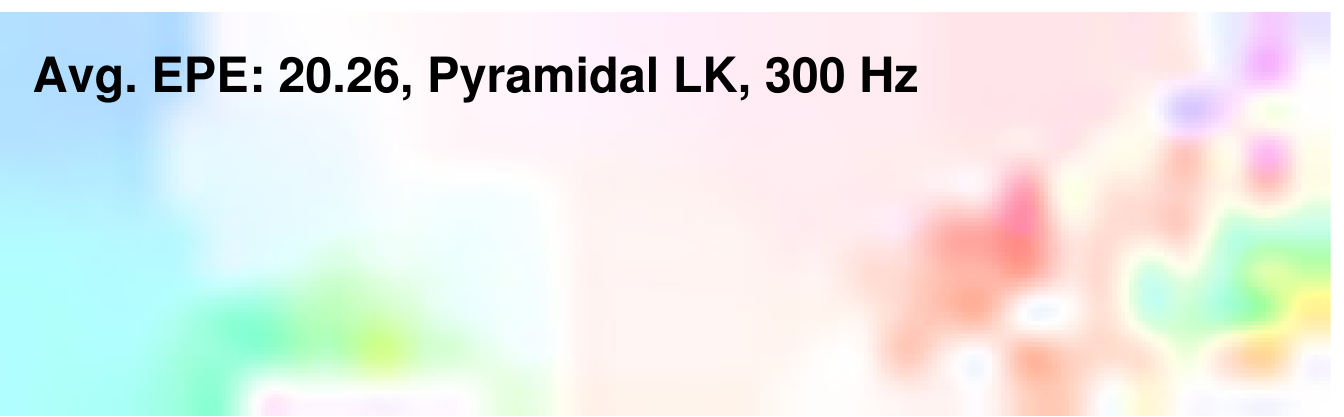}&
\includegraphics[width=0.195\textwidth]{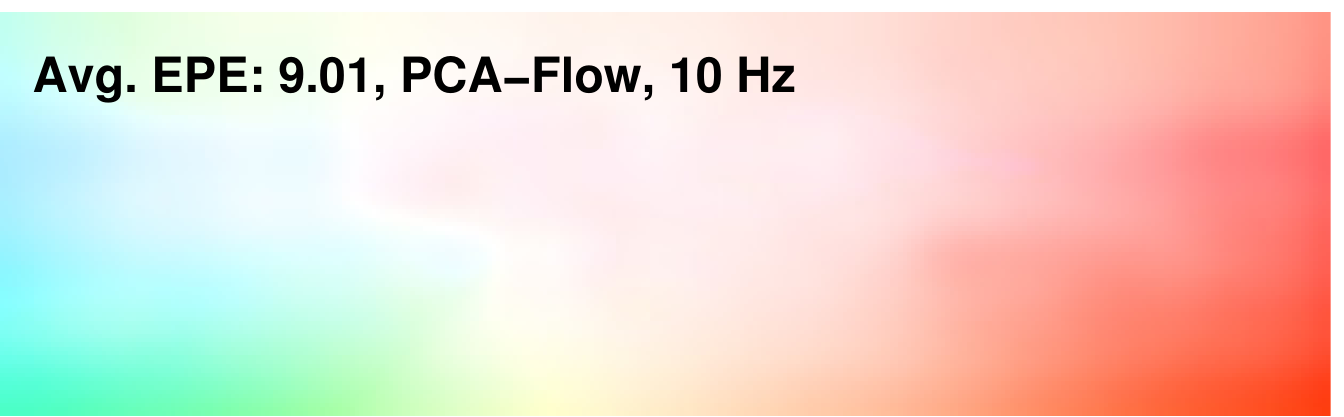}&
\includegraphics[width=0.195\textwidth]{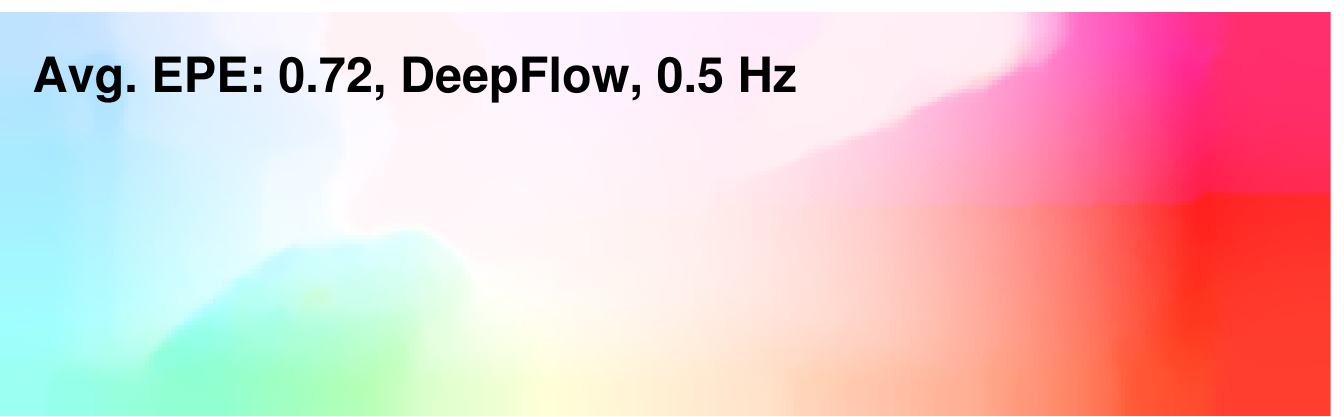}&
\includegraphics[width=0.195\textwidth]{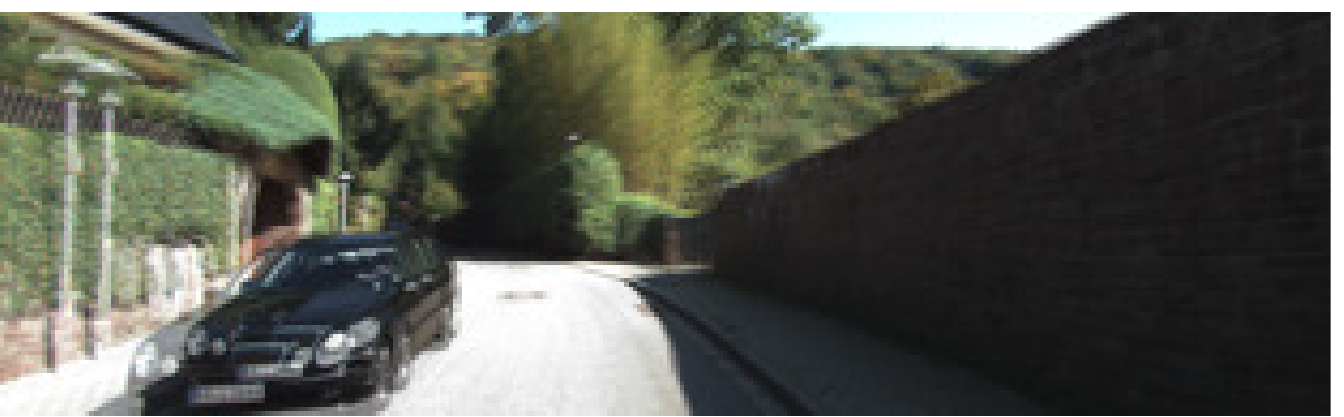}\\[6pt]
\includegraphics[width=0.195\textwidth]{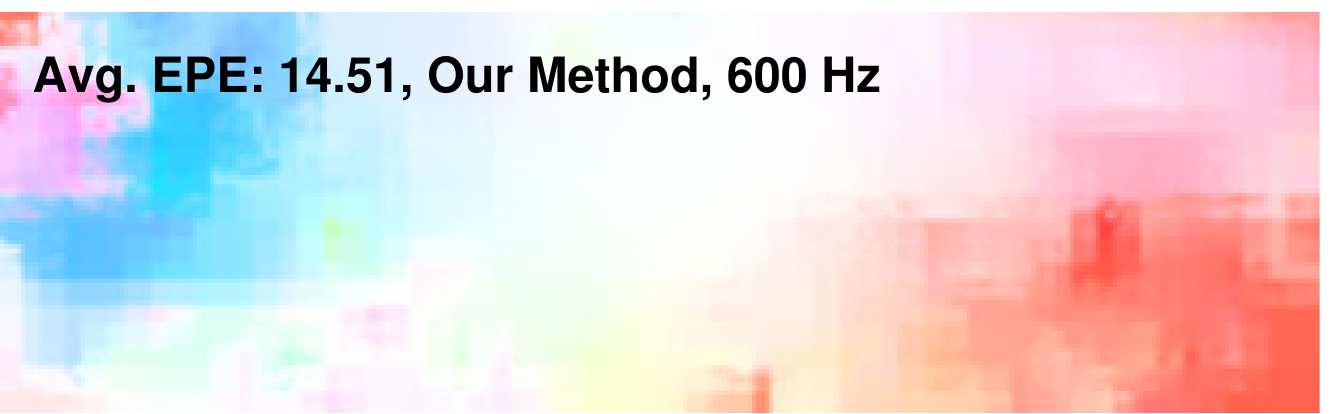}&
\includegraphics[width=0.195\textwidth]{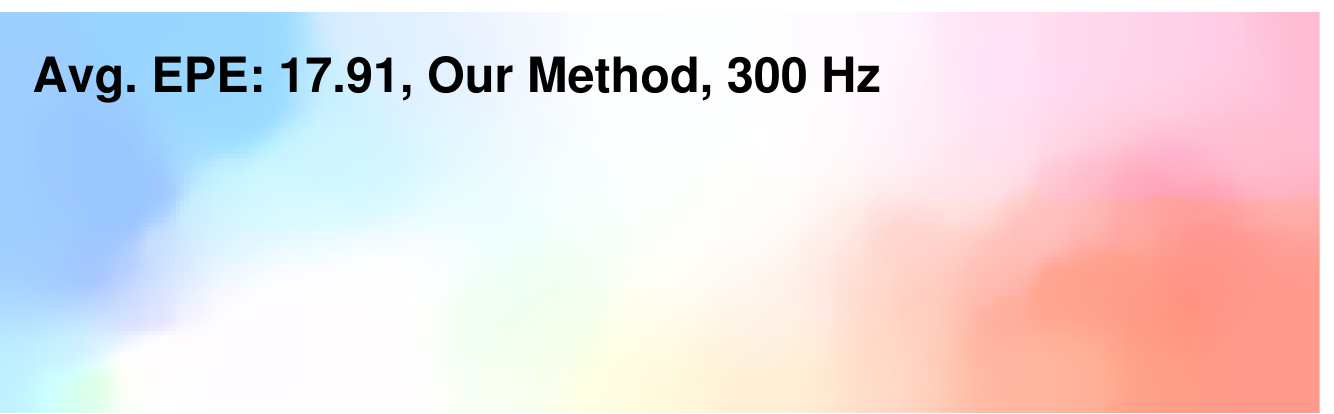}&
\includegraphics[width=0.195\textwidth]{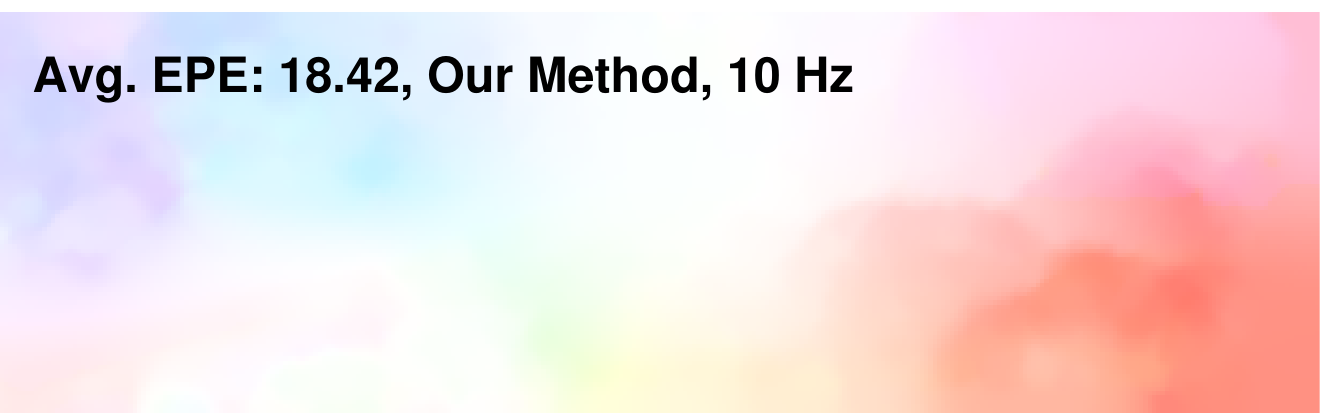}&
\includegraphics[width=0.195\textwidth]{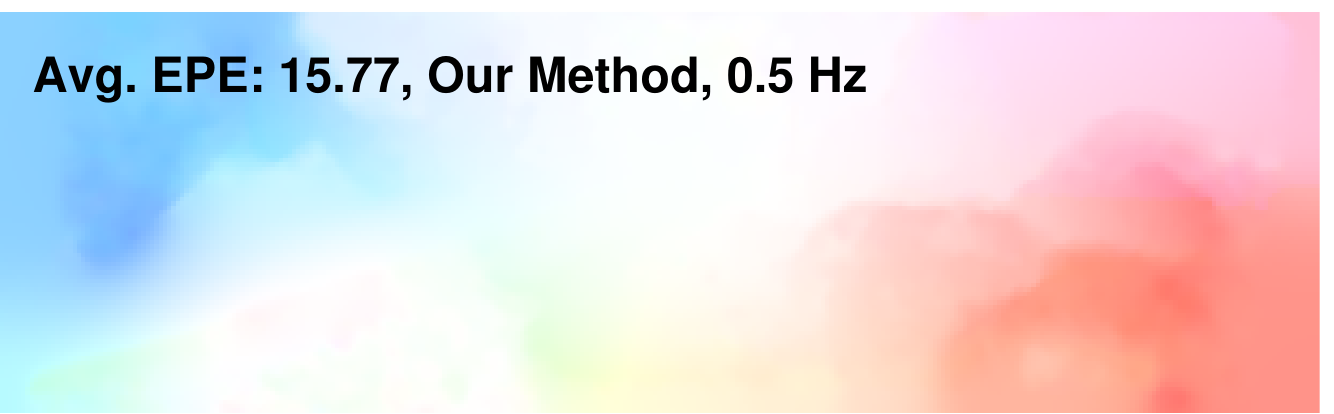}&
\includegraphics[width=0.195\textwidth]{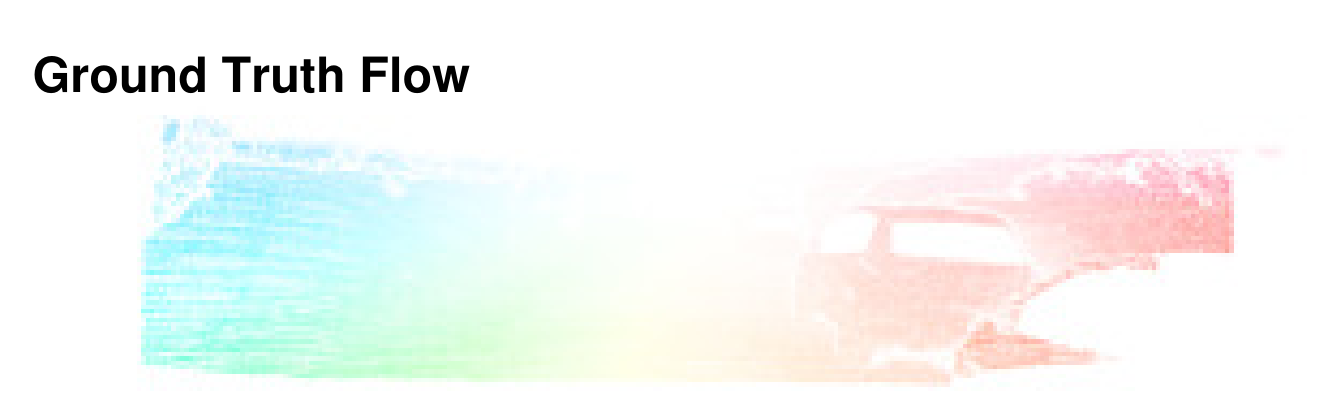}\\
\includegraphics[width=0.195\textwidth]{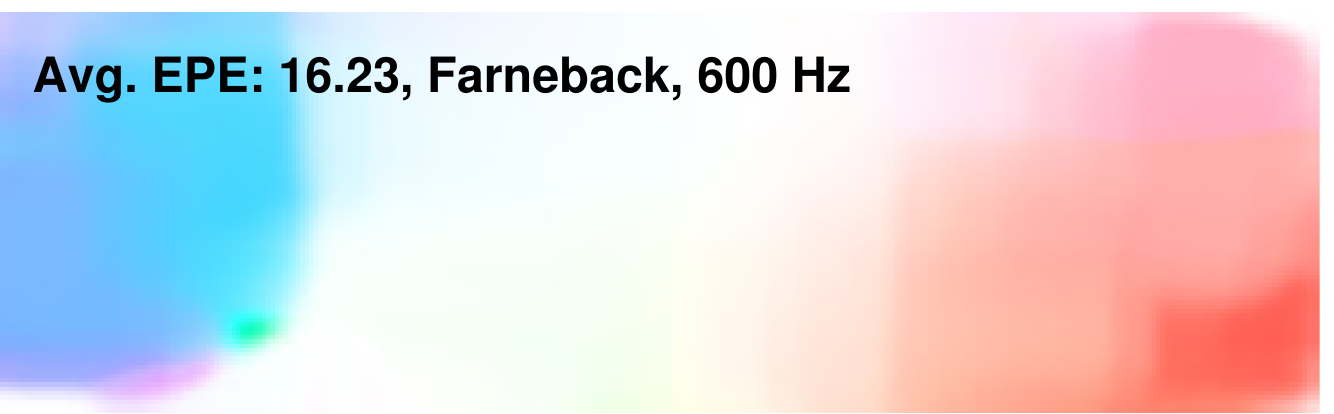}&
\includegraphics[width=0.195\textwidth]{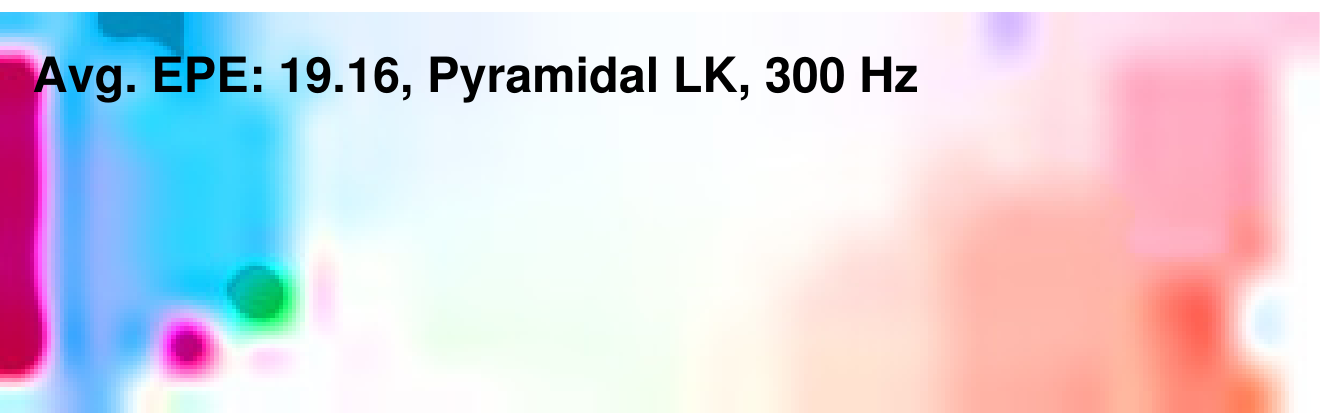}&
\includegraphics[width=0.195\textwidth]{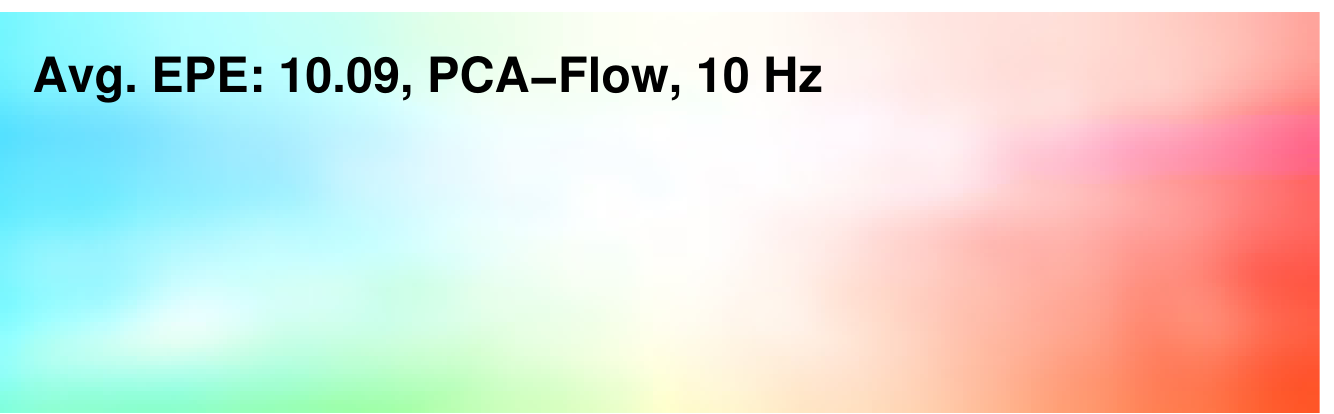}&
\includegraphics[width=0.195\textwidth]{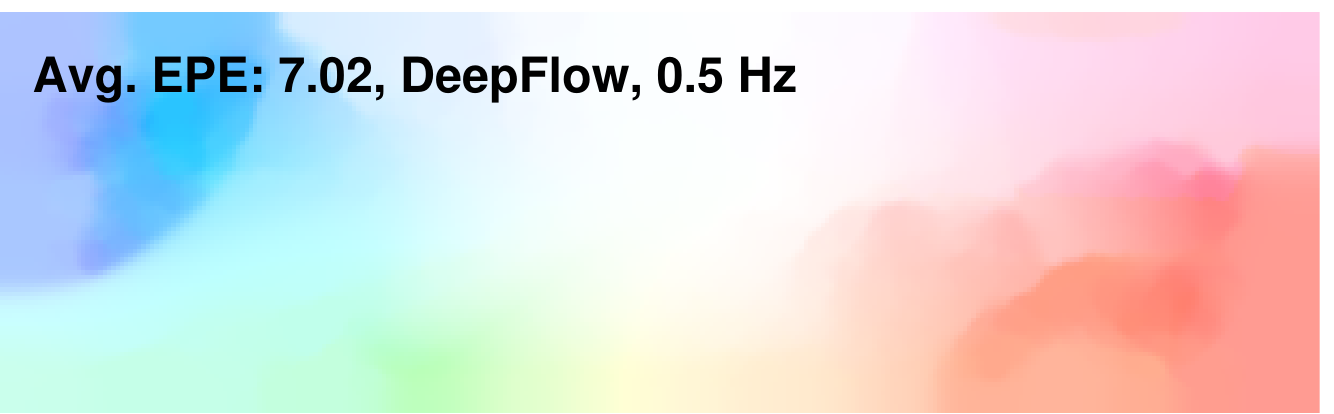}&
\includegraphics[width=0.195\textwidth]{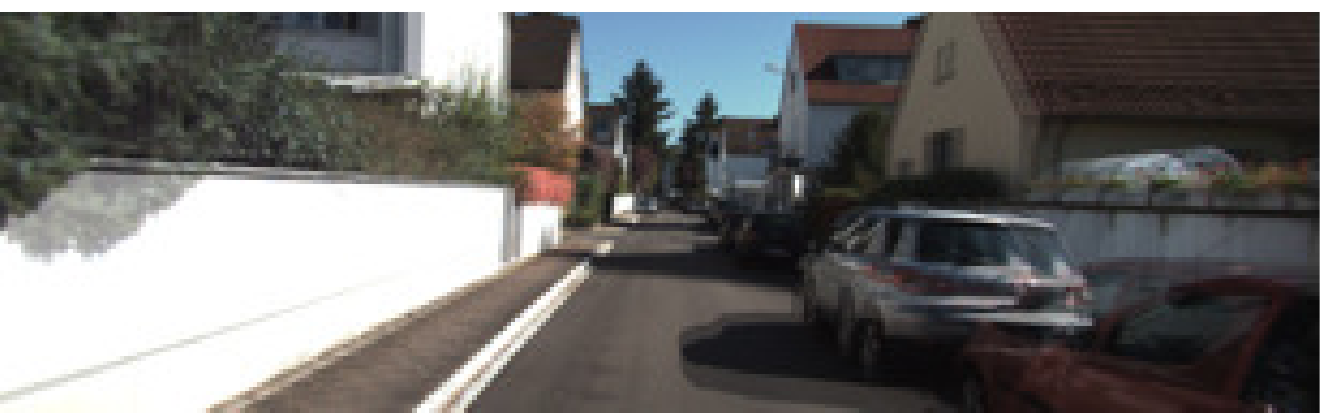}\\[6pt]
\includegraphics[width=0.195\textwidth]{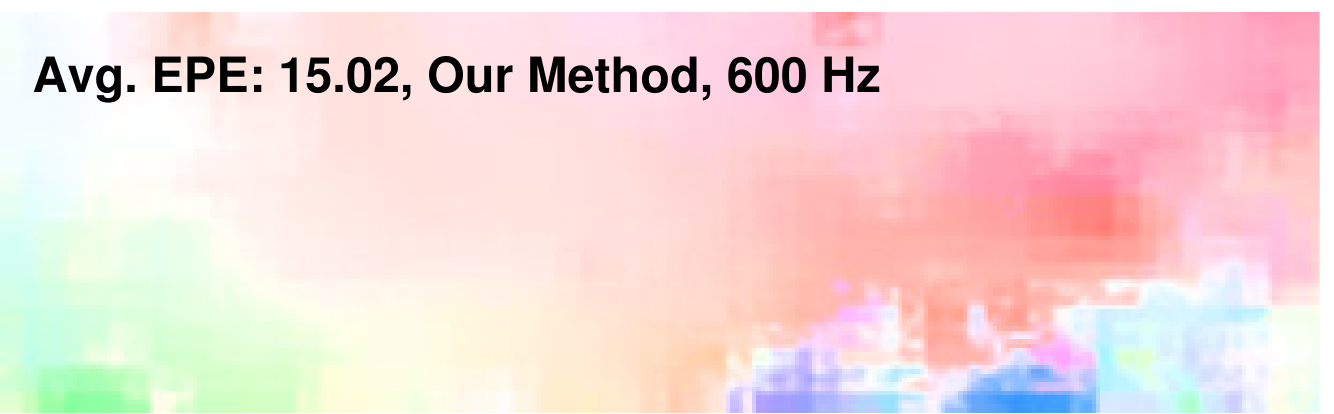}&
\includegraphics[width=0.195\textwidth]{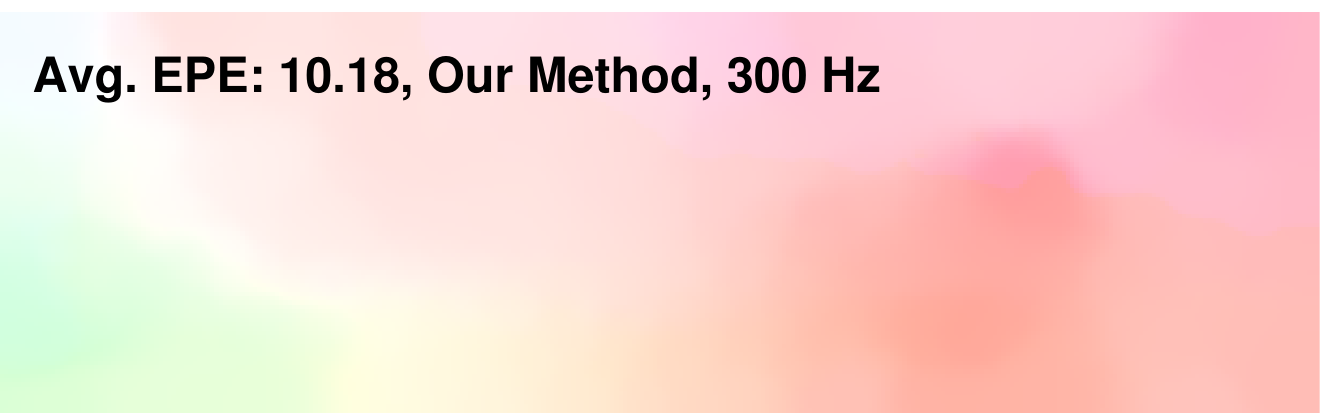}&
\includegraphics[width=0.195\textwidth]{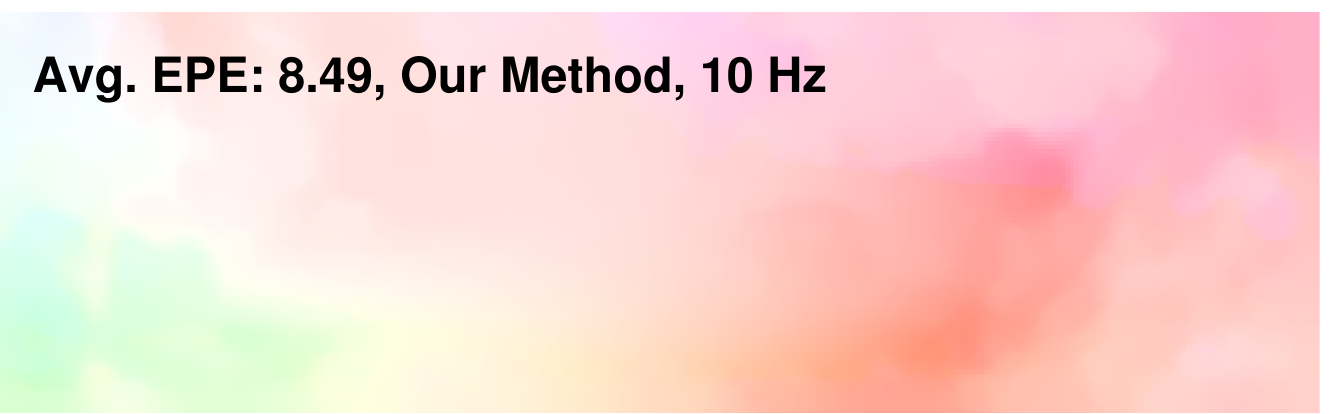}&
\includegraphics[width=0.195\textwidth]{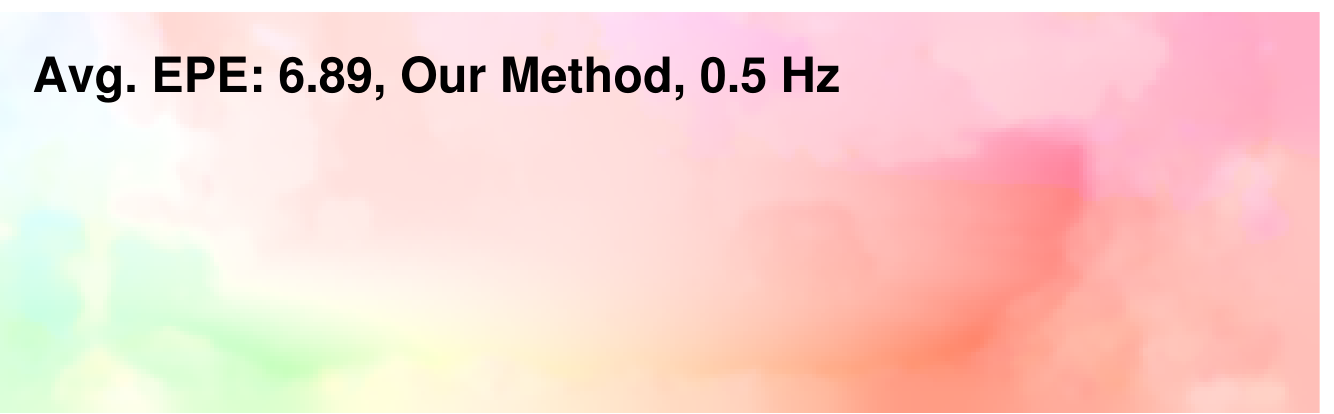}&
\includegraphics[width=0.195\textwidth]{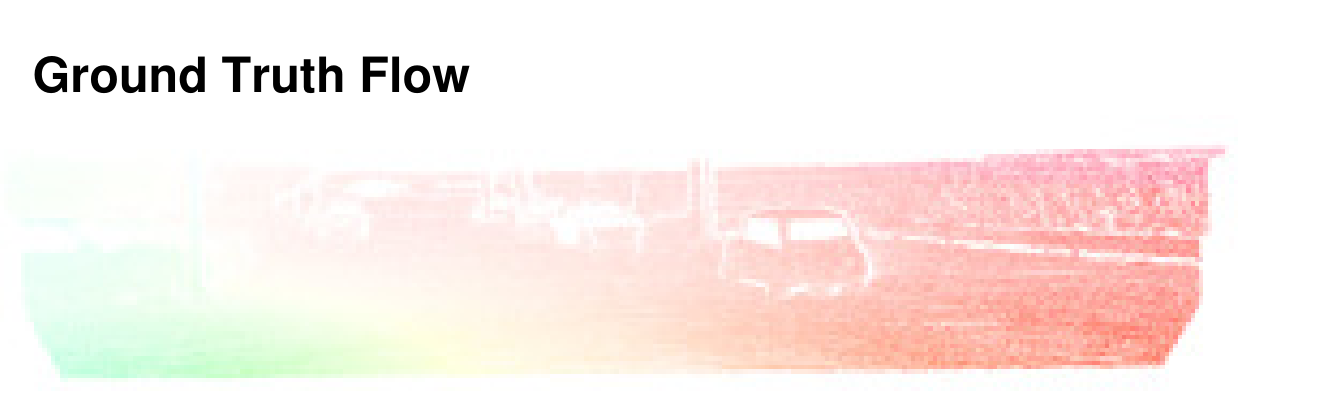}\\
\includegraphics[width=0.195\textwidth]{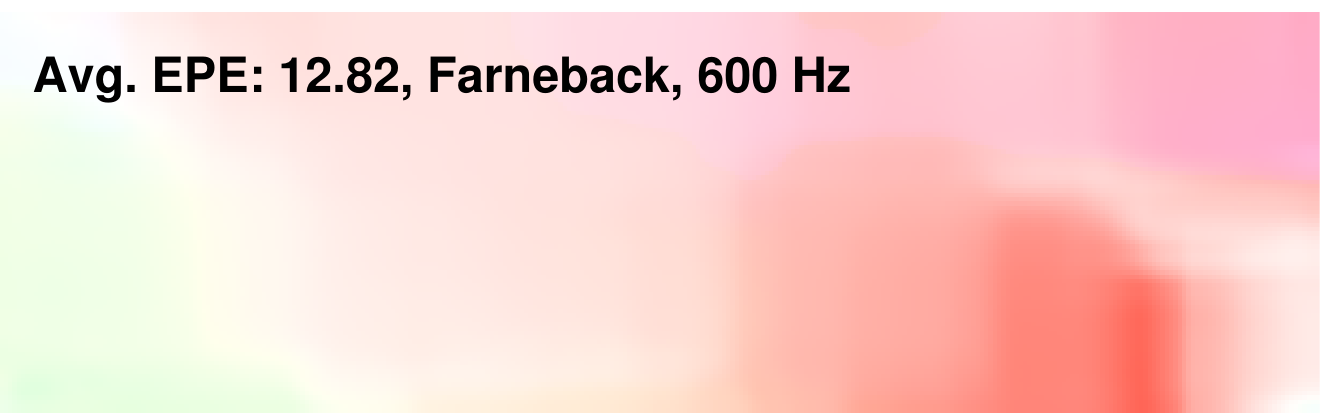}&
\includegraphics[width=0.195\textwidth]{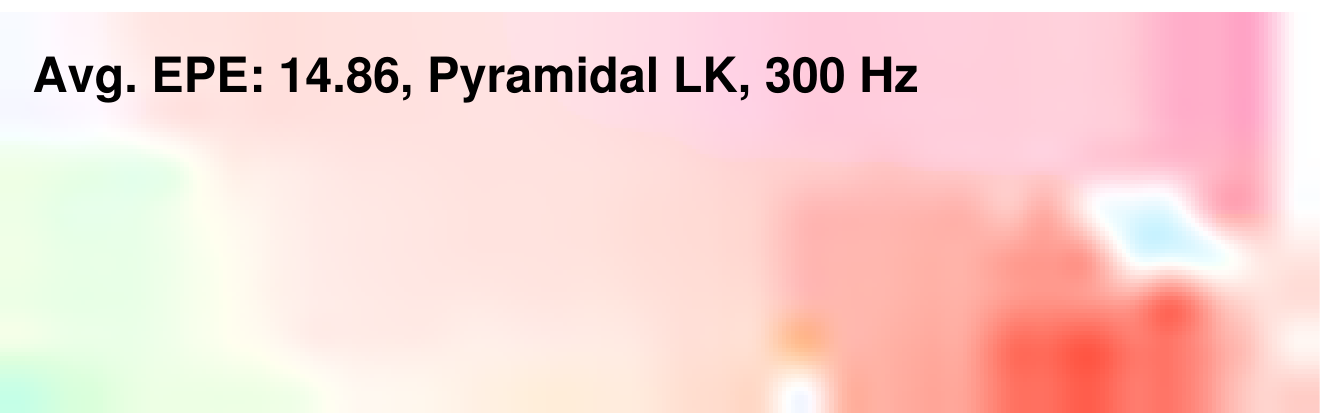}&
\includegraphics[width=0.195\textwidth]{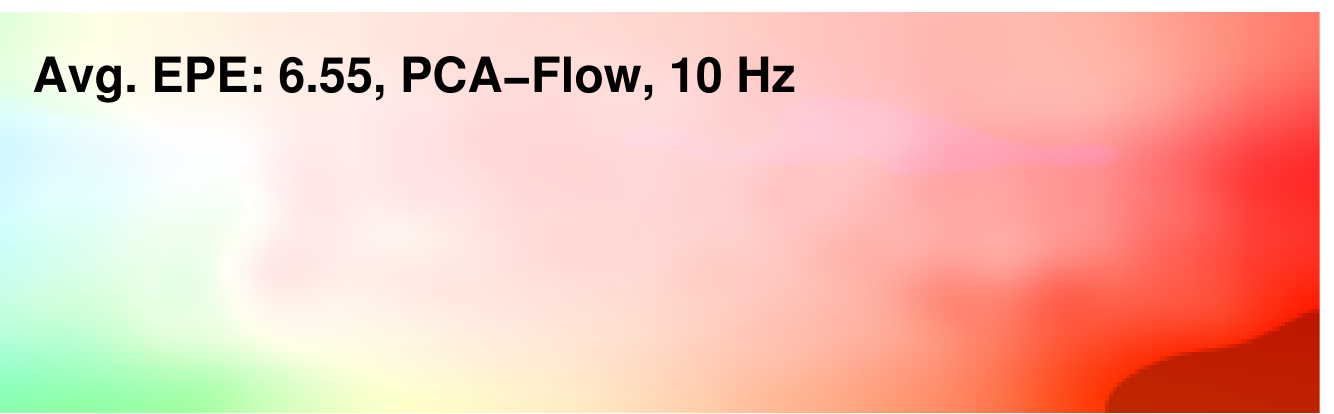}&
\includegraphics[width=0.195\textwidth]{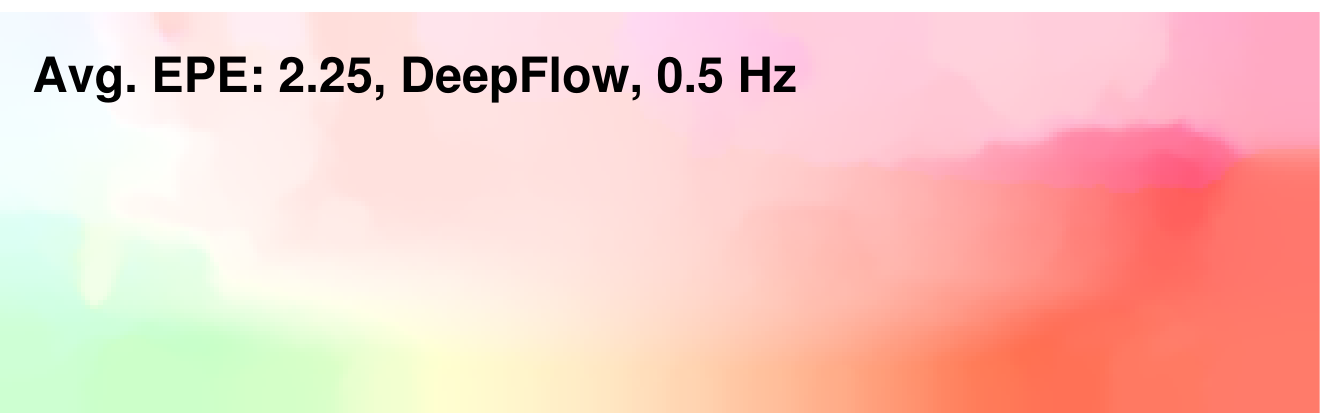}&
\includegraphics[width=0.195\textwidth]{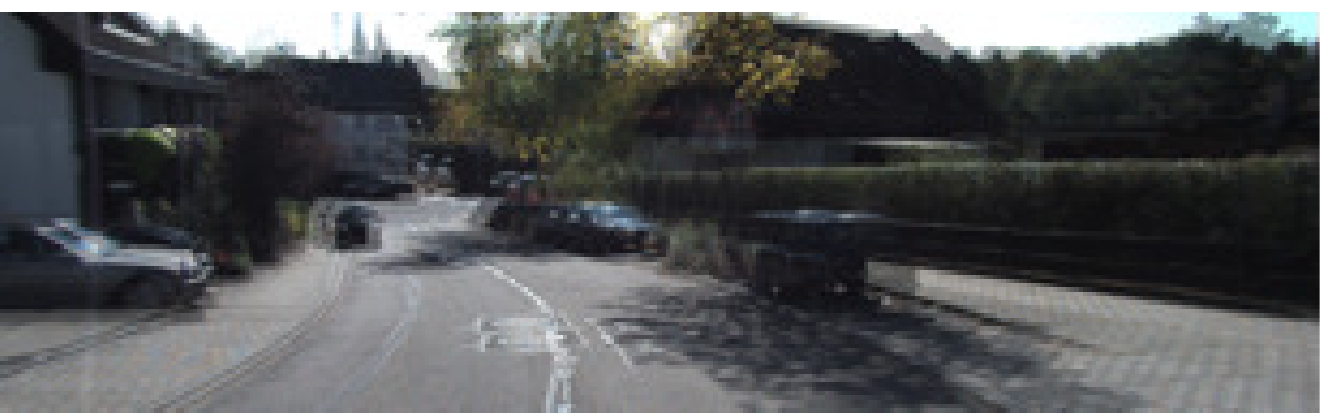}\\[6pt]
\includegraphics[width=0.195\textwidth]{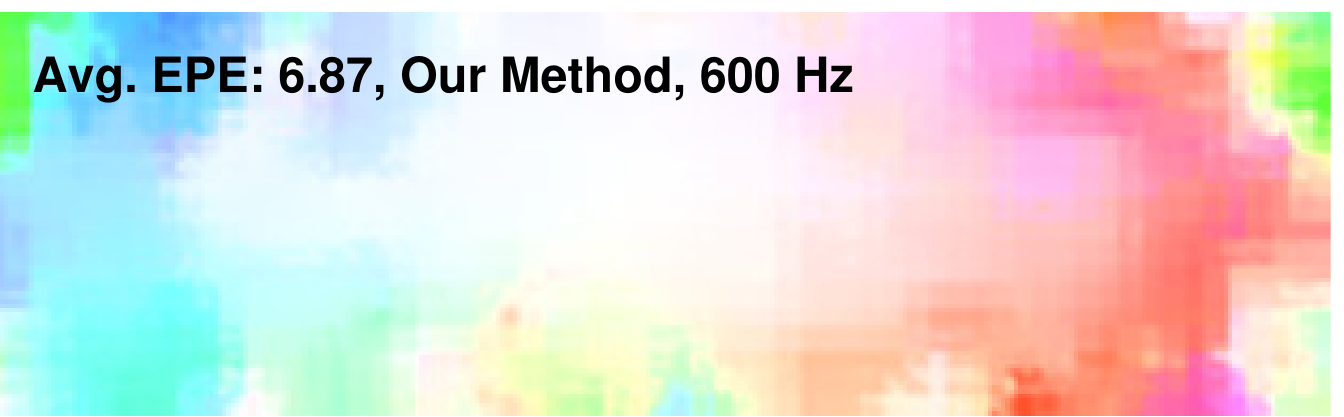}&
\includegraphics[width=0.195\textwidth]{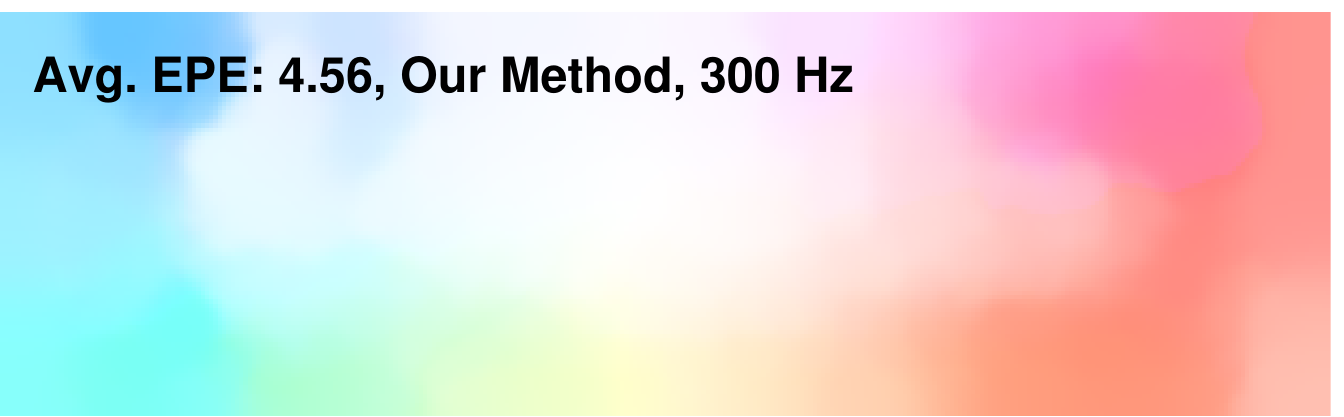}&
\includegraphics[width=0.195\textwidth]{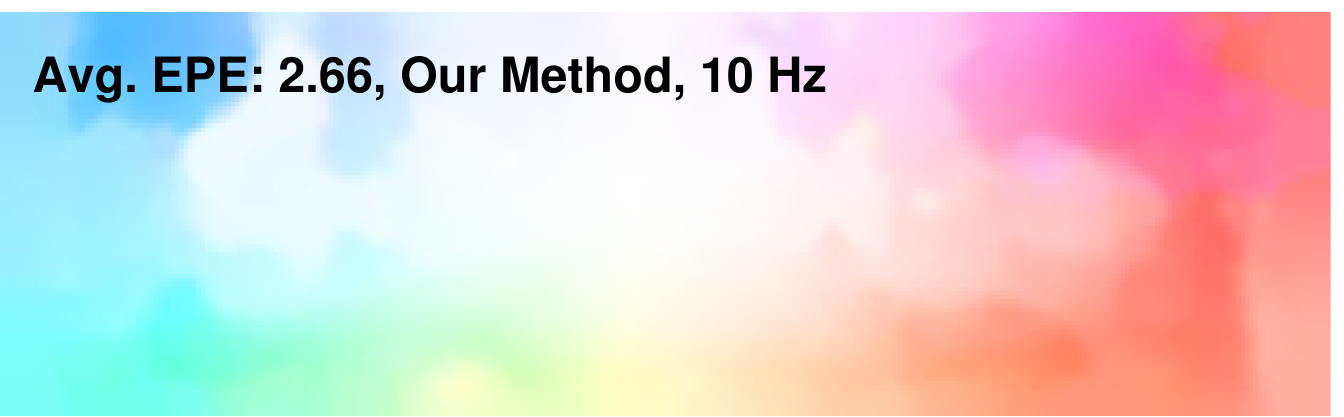}&
\includegraphics[width=0.195\textwidth]{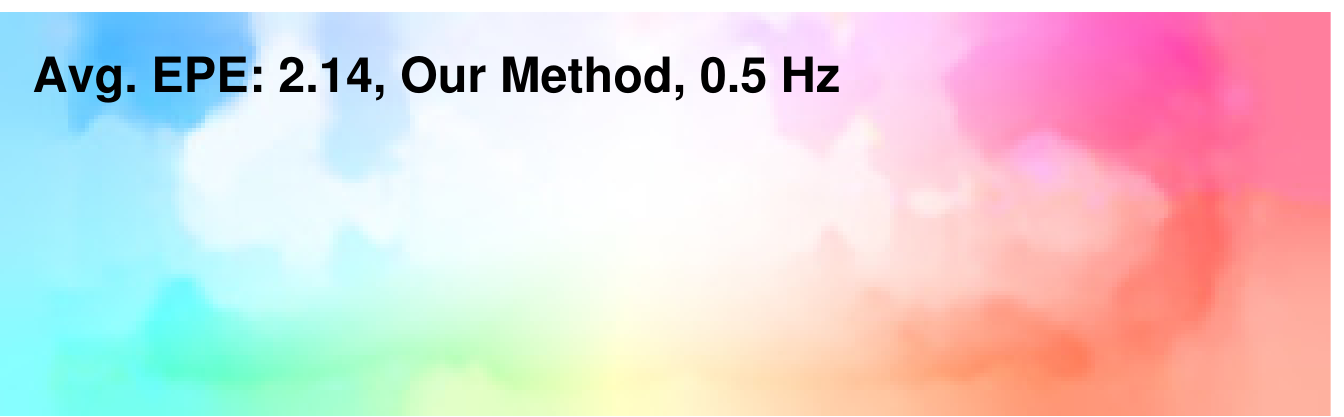}&
\includegraphics[width=0.195\textwidth]{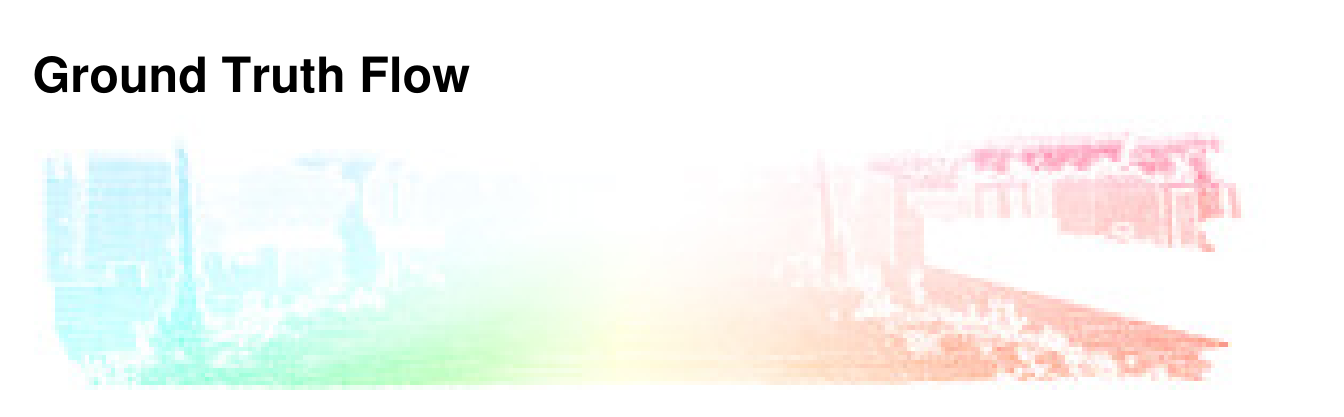}\\
\includegraphics[width=0.195\textwidth]{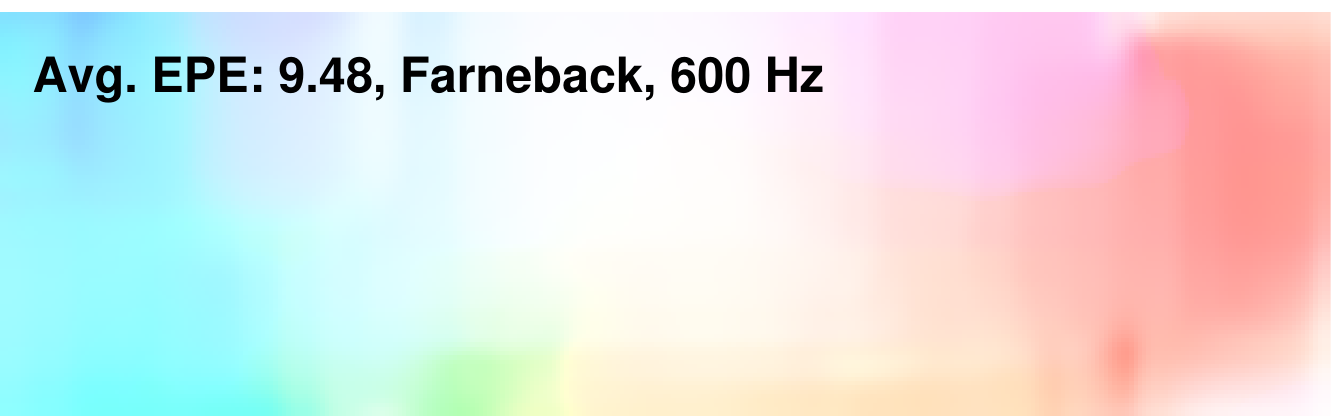}&
\includegraphics[width=0.195\textwidth]{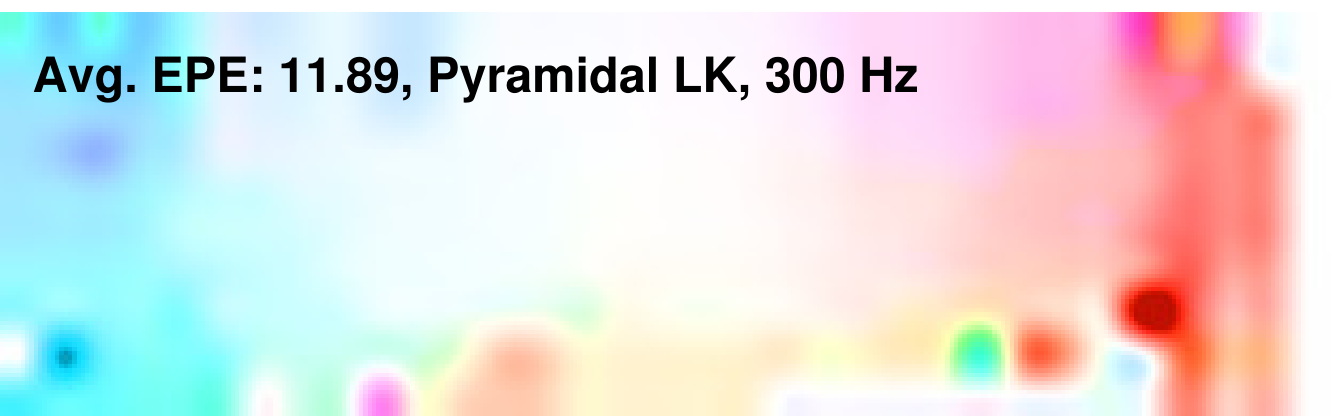}&
\includegraphics[width=0.195\textwidth]{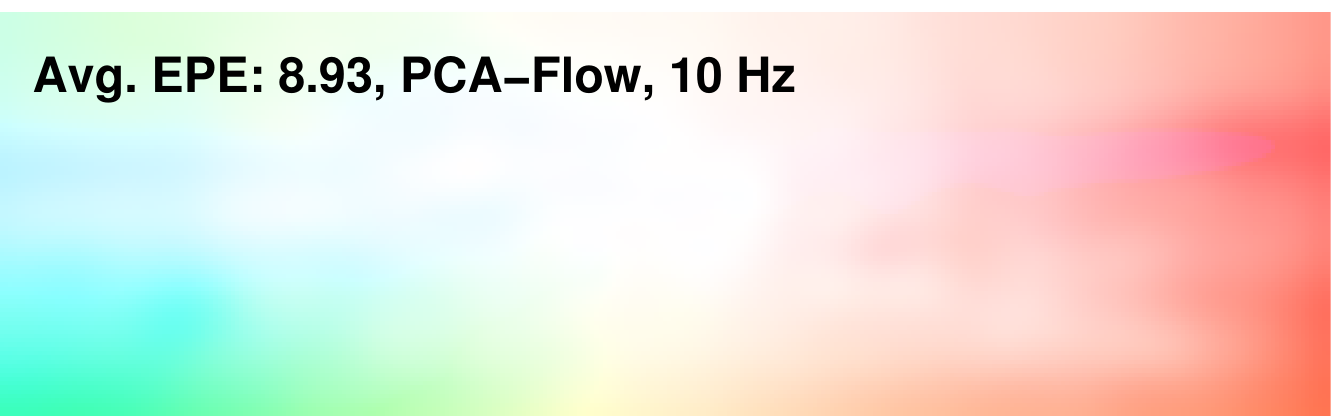}&
\includegraphics[width=0.195\textwidth]{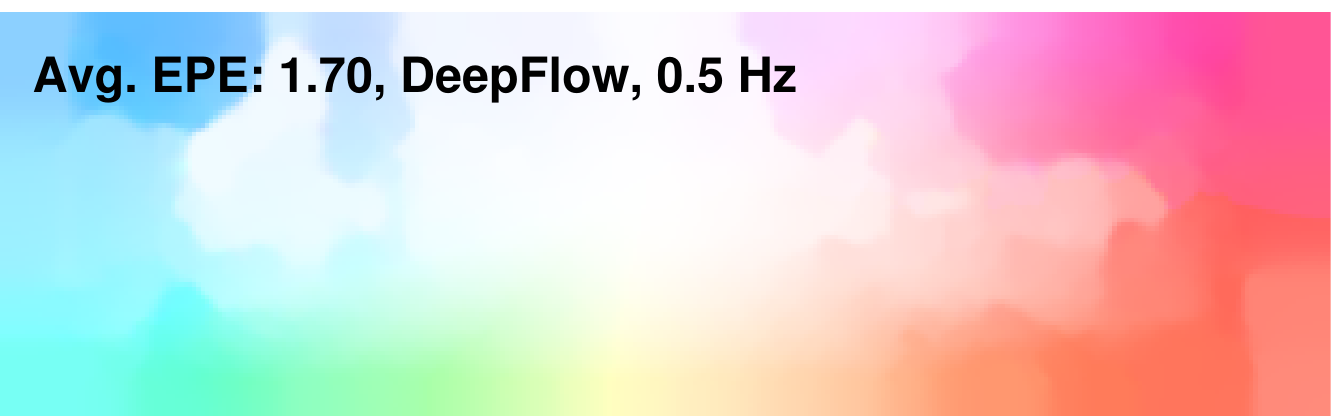}&
\includegraphics[width=0.195\textwidth]{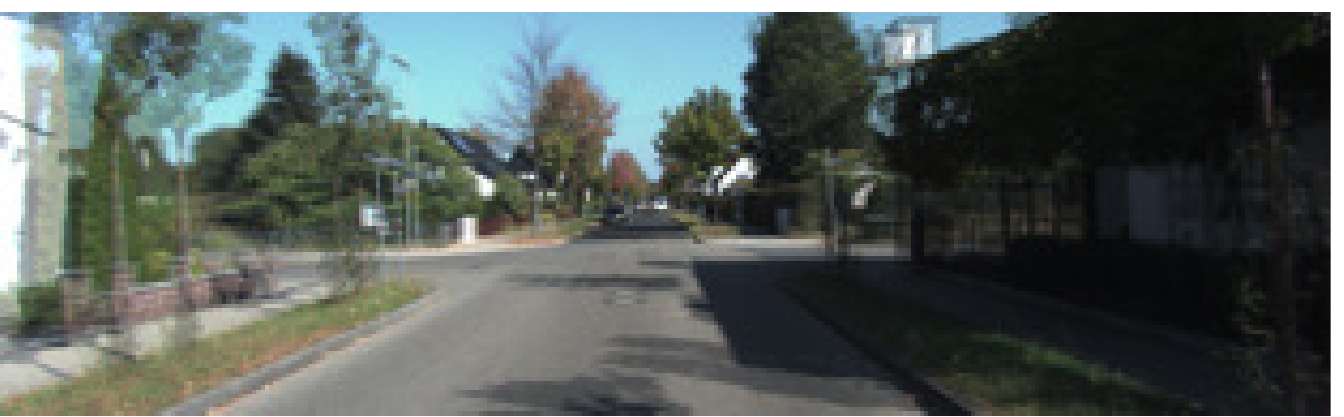}\\[6pt]
\includegraphics[width=0.195\textwidth]{imgs_exa/eximg-kitti-0183-01.pdf}&
\includegraphics[width=0.195\textwidth]{imgs_exa/eximg-kitti-0183-02.pdf}&
\includegraphics[width=0.195\textwidth]{imgs_exa/eximg-kitti-0183-03.pdf}&
\includegraphics[width=0.195\textwidth]{imgs_exa/eximg-kitti-0183-04.pdf}&
\includegraphics[width=0.195\textwidth]{imgs_exa/eximg-kitti-0183-05.pdf}\\
\includegraphics[width=0.195\textwidth]{imgs_exa/eximg-kitti-0183-06.pdf}&
\includegraphics[width=0.195\textwidth]{imgs_exa/eximg-kitti-0183-07.pdf}&
\includegraphics[width=0.195\textwidth]{imgs_exa/eximg-kitti-0183-08.pdf}&
\includegraphics[width=0.195\textwidth]{imgs_exa/eximg-kitti-0183-09.pdf}&
\includegraphics[width=0.195\textwidth]{imgs_exa/eximg-kitti-0183-10.pdf}\\[6pt]
\end{tabular}
}
\caption{Exemplary results on KITTI (training). In each block of $2 \times 6$ images.  Top row, left to right: Our method for operating points ({\bf 1})-({\bf 4}), Ground Truth. Bottom row: Farneback 600Hz, Pyramidal LK 300Hz, PCA-Flow 10Hz, DeepFlow 0.5Hz, Original Image.}\label{fig:kitti2res_AP} 

\end{figure*}

More exemplary optical flow results on the training subsets of the Sintel~\cite{Butler-ECCV-2012} and KITTI~\cite{Geiger-IJRR-2013} benchmarks are shown in Fig.~\ref{fig:sintel1res_AP}, \ref{fig:sintel2res_AP}, \ref{fig:kitti1res_AP}, and \ref{fig:kitti2res_AP}. Error maps for Figs.~\ref{fig:sintel1res_AP} and \ref{fig:sintel2res_AP} are plotted in Figs.~\ref{fig:sintel1res_errmap_AP} and \ref{fig:sintel2res_errmap_AP}.
 
\section{Error maps on Sintel-training}

\begin{figure*} [!ht]
\centering\setlength{\tabcolsep}{0.1pt}\renewcommand{\arraystretch}{0} 
\begin{tabular}{ccccc}
 {\bf 600Hz} & {\bf 300Hz} & {\bf 10Hz} & {\bf 0.5Hz}& {\bf Ground Truth}\\
\includegraphics[width=0.195\textwidth]{imgs_exa/eximg-0001-01.pdf}&
\includegraphics[width=0.195\textwidth]{imgs_exa/eximg-0001-02.pdf}&
\includegraphics[width=0.195\textwidth]{imgs_exa/eximg-0001-03.pdf}&
\includegraphics[width=0.195\textwidth]{imgs_exa/eximg-0001-04.pdf}&
\includegraphics[width=0.195\textwidth]{imgs_exa/eximg-0001-05.pdf}\\
\includegraphics[width=0.195\textwidth]{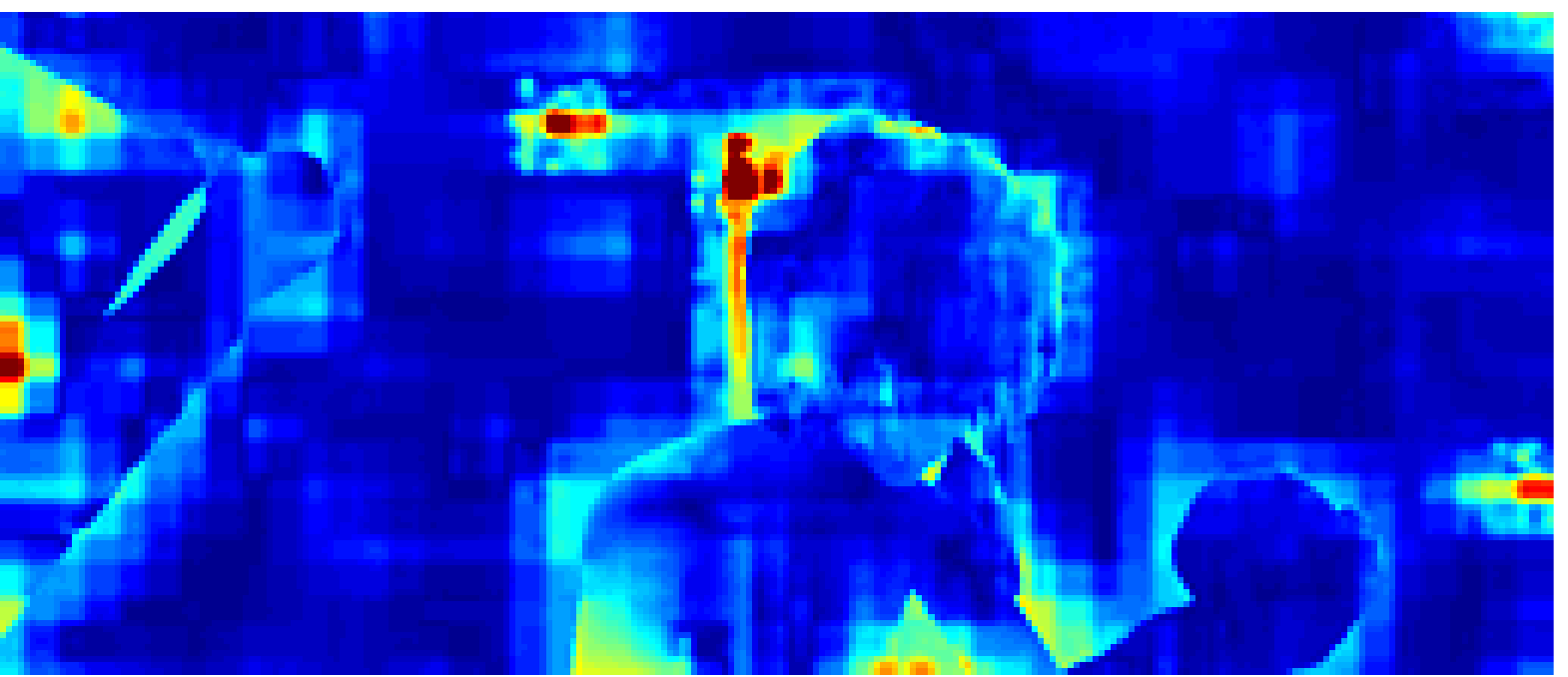}&
\includegraphics[width=0.195\textwidth]{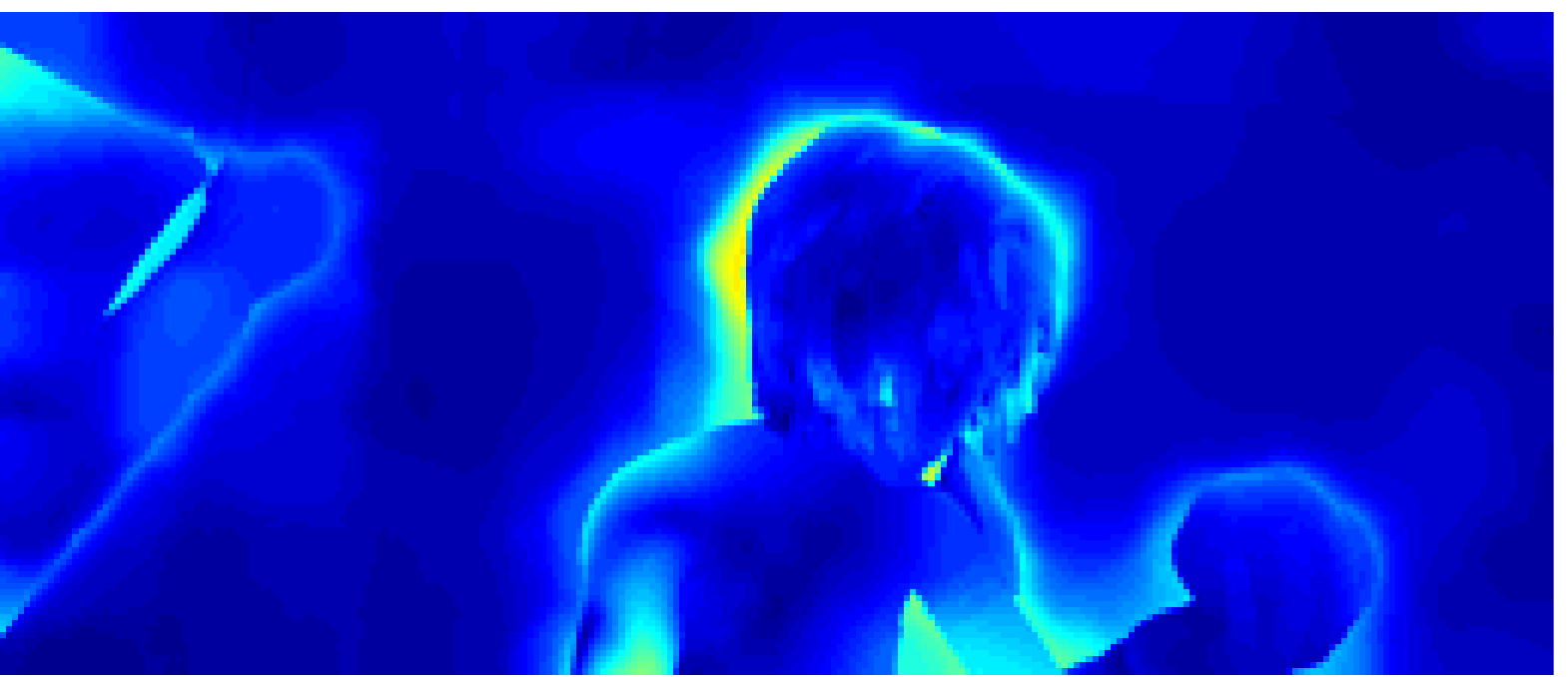}&
\includegraphics[width=0.195\textwidth]{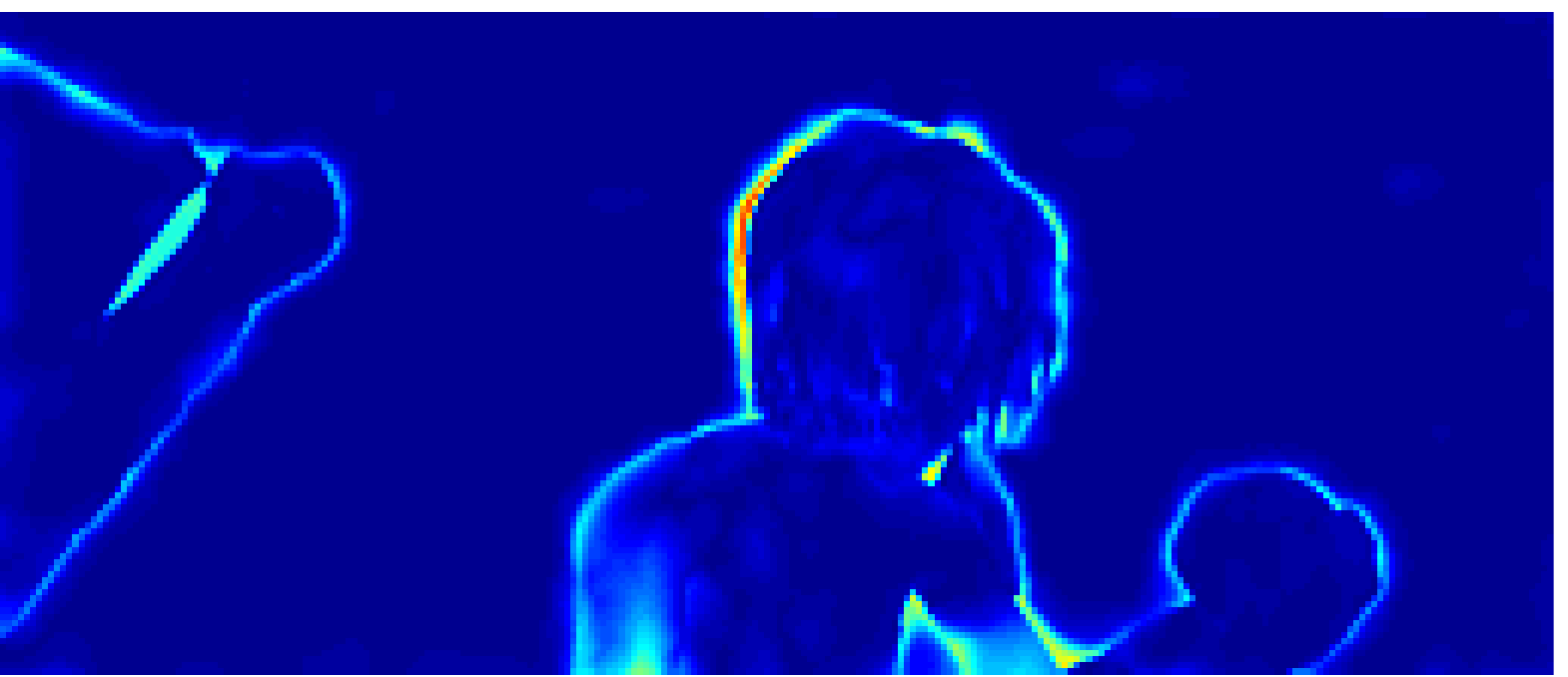}&
\includegraphics[width=0.195\textwidth]{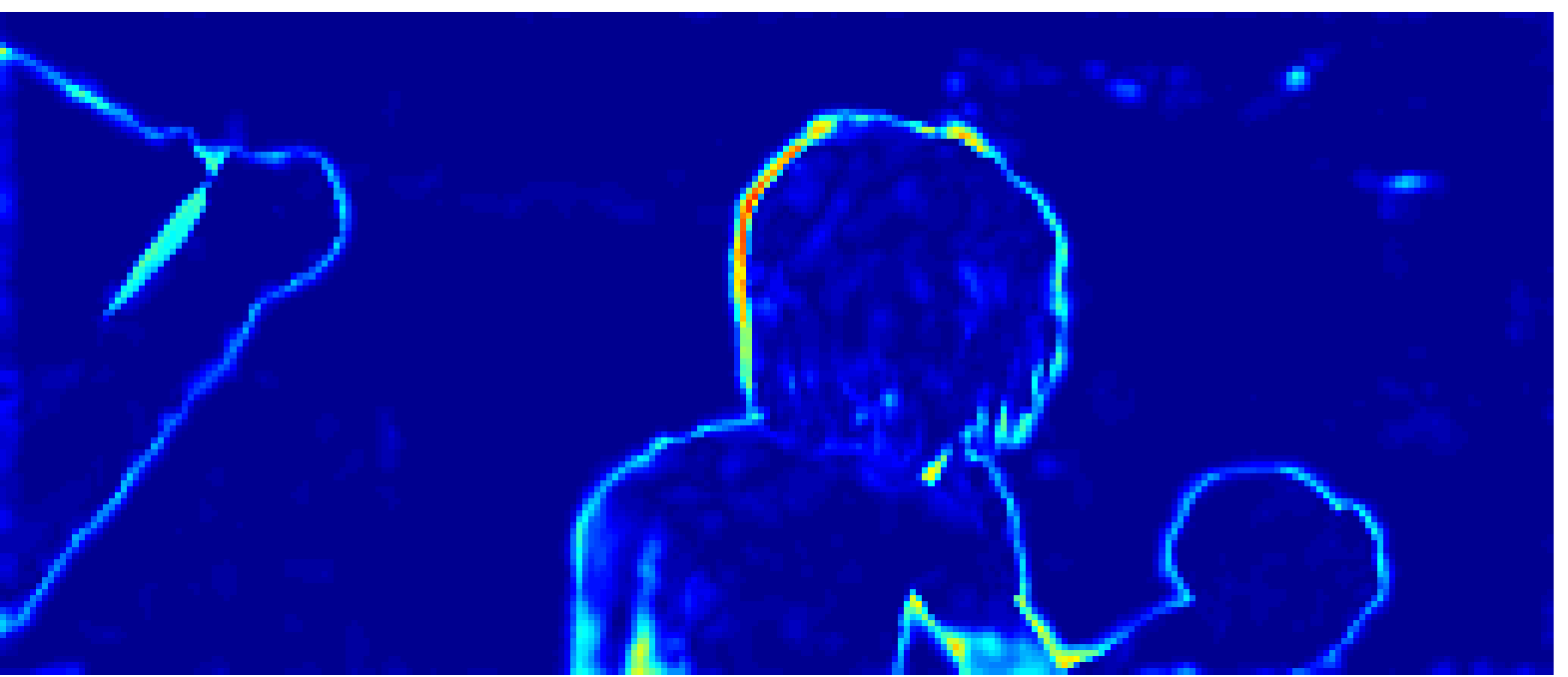}&
\includegraphics[width=0.195\textwidth]{imgs_exa/eximg-0001-10.pdf}\\[6pt]
\includegraphics[width=0.195\textwidth]{imgs_exa/eximg-0061-01.pdf}&
\includegraphics[width=0.195\textwidth]{imgs_exa/eximg-0061-02.pdf}&
\includegraphics[width=0.195\textwidth]{imgs_exa/eximg-0061-03.pdf}&
\includegraphics[width=0.195\textwidth]{imgs_exa/eximg-0061-04.pdf}&
\includegraphics[width=0.195\textwidth]{imgs_exa/eximg-0061-05.pdf}\\
\includegraphics[width=0.195\textwidth]{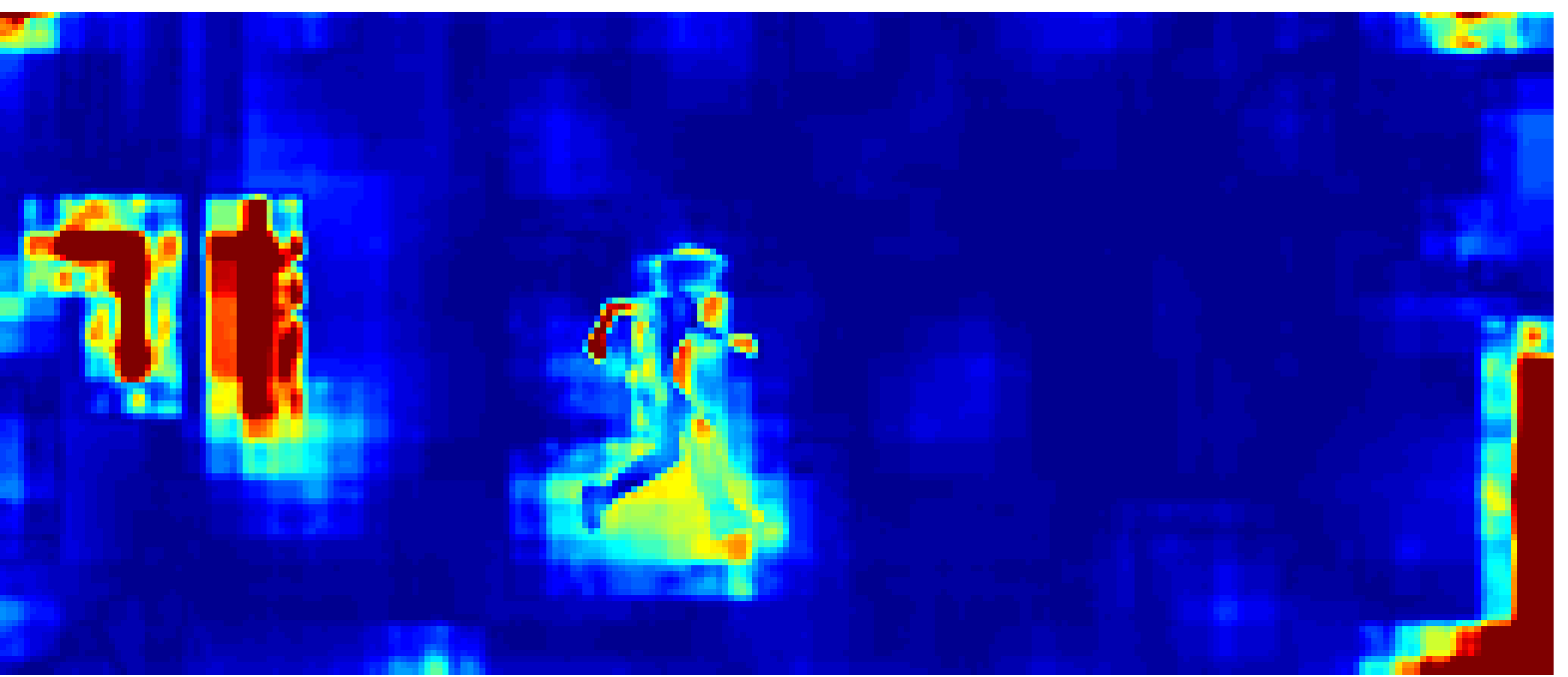}&
\includegraphics[width=0.195\textwidth]{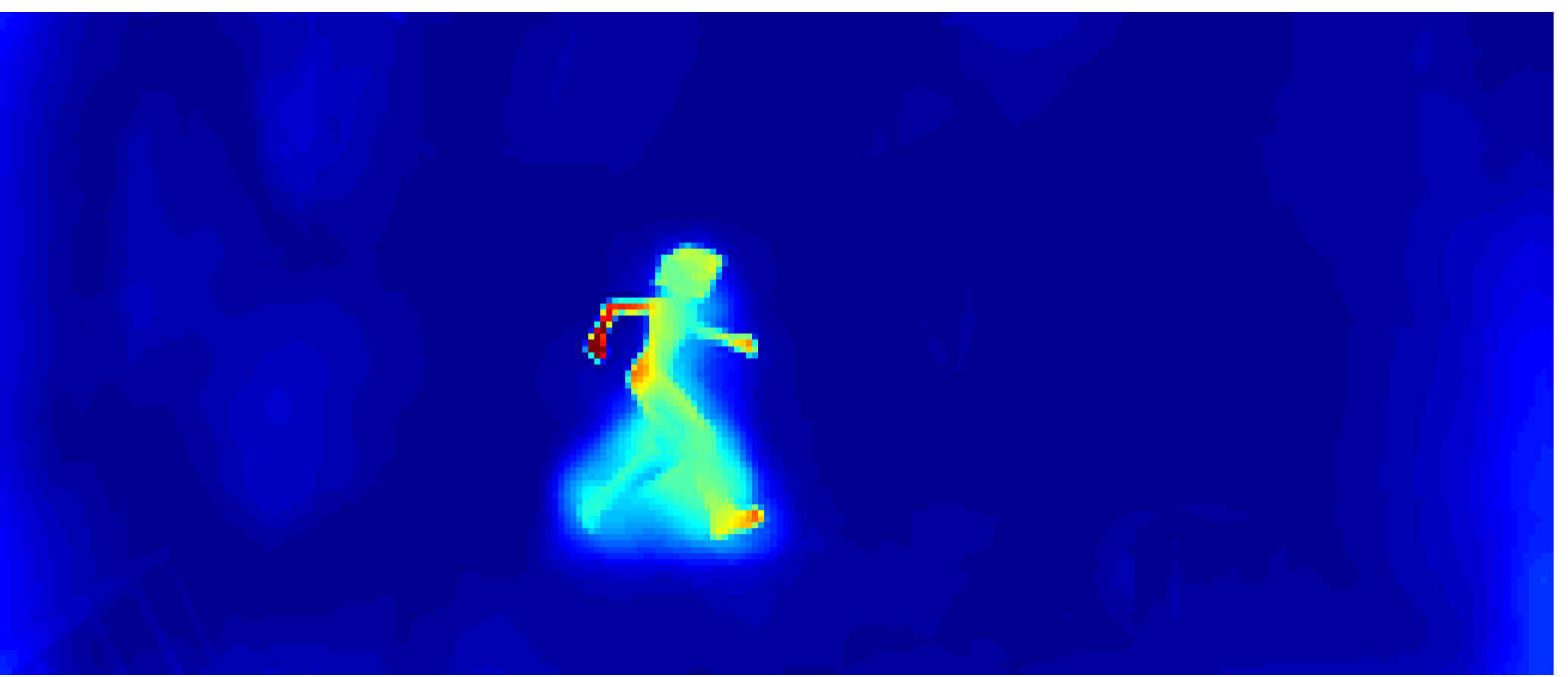}&
\includegraphics[width=0.195\textwidth]{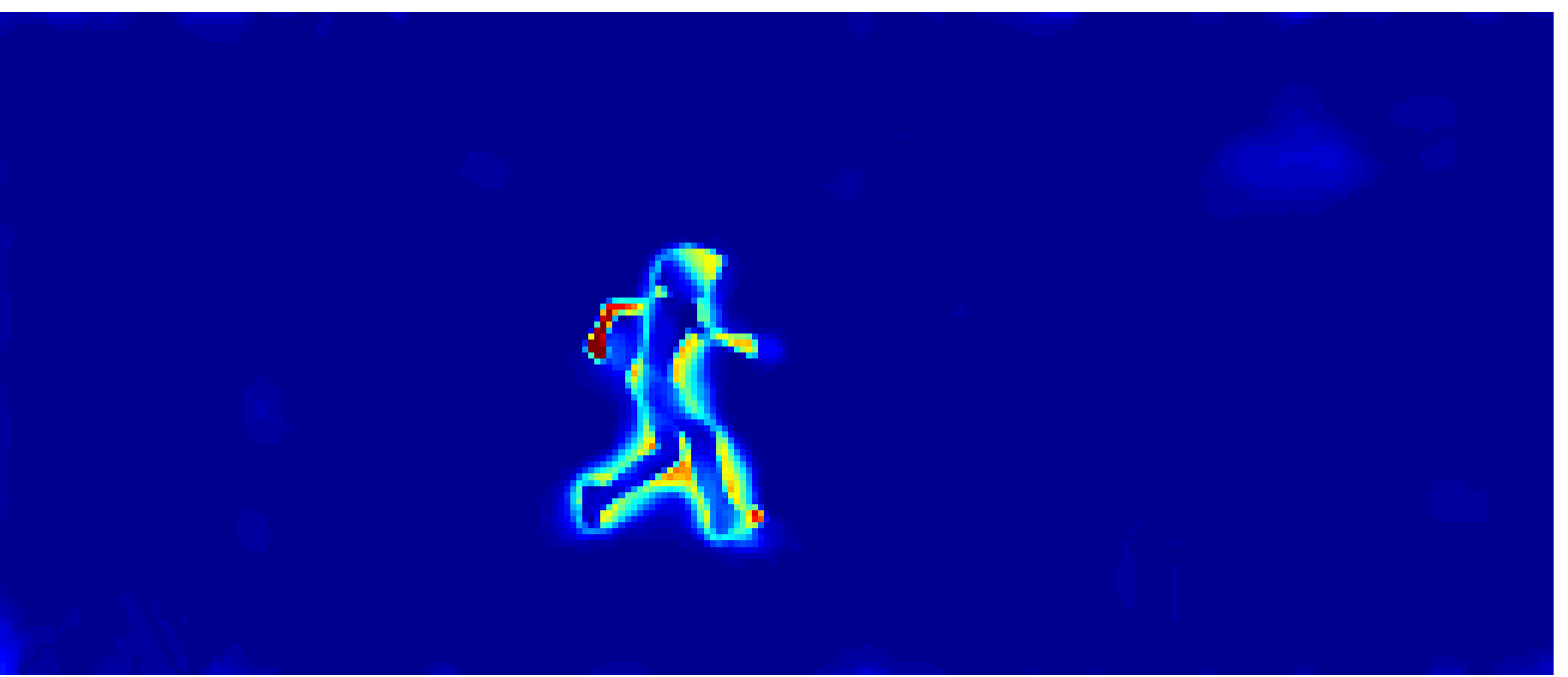}&
\includegraphics[width=0.195\textwidth]{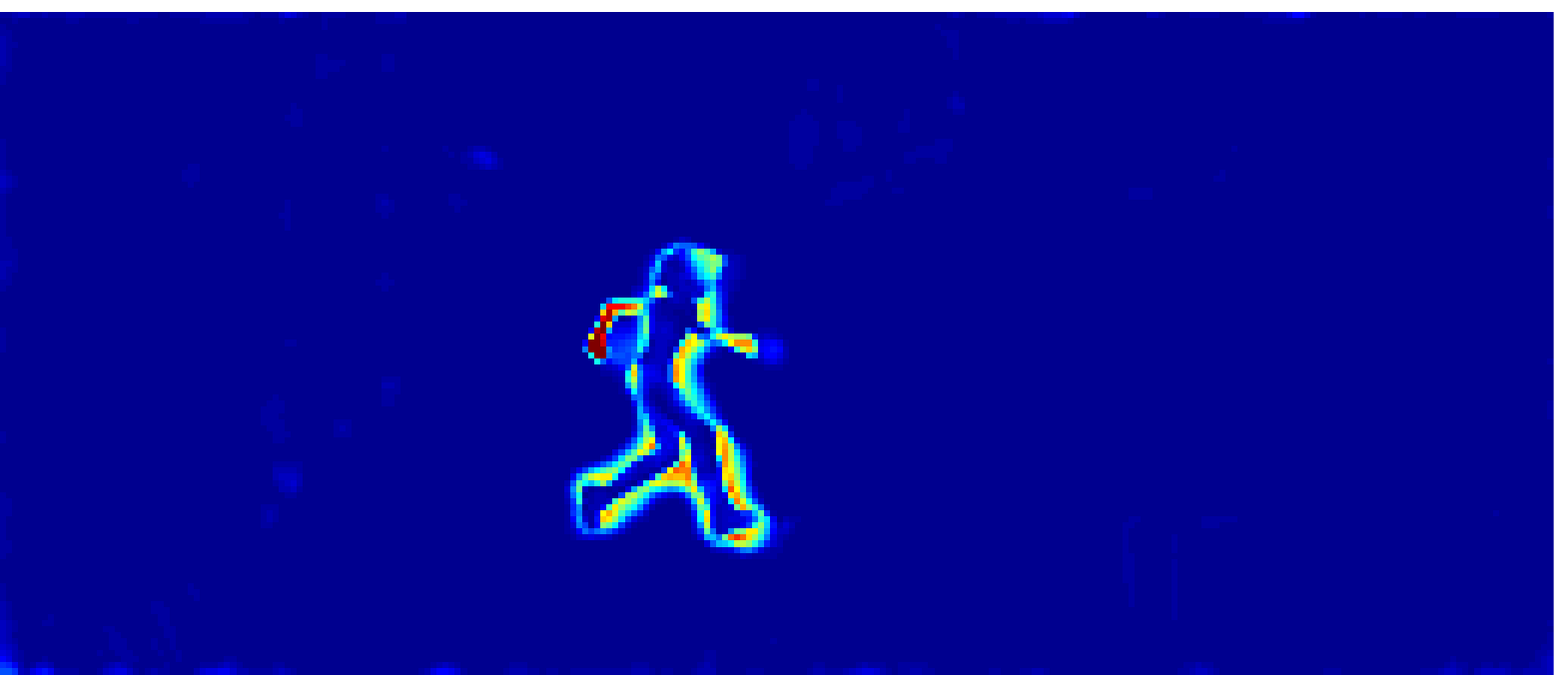}&
\includegraphics[width=0.195\textwidth]{imgs_exa/eximg-0061-10.pdf}\\[6pt]
\includegraphics[width=0.195\textwidth]{imgs_exa/eximg-0141-01.pdf}&
\includegraphics[width=0.195\textwidth]{imgs_exa/eximg-0141-02.pdf}&
\includegraphics[width=0.195\textwidth]{imgs_exa/eximg-0141-03.pdf}&
\includegraphics[width=0.195\textwidth]{imgs_exa/eximg-0141-04.pdf}&
\includegraphics[width=0.195\textwidth]{imgs_exa/eximg-0141-05.pdf}\\
\includegraphics[width=0.195\textwidth]{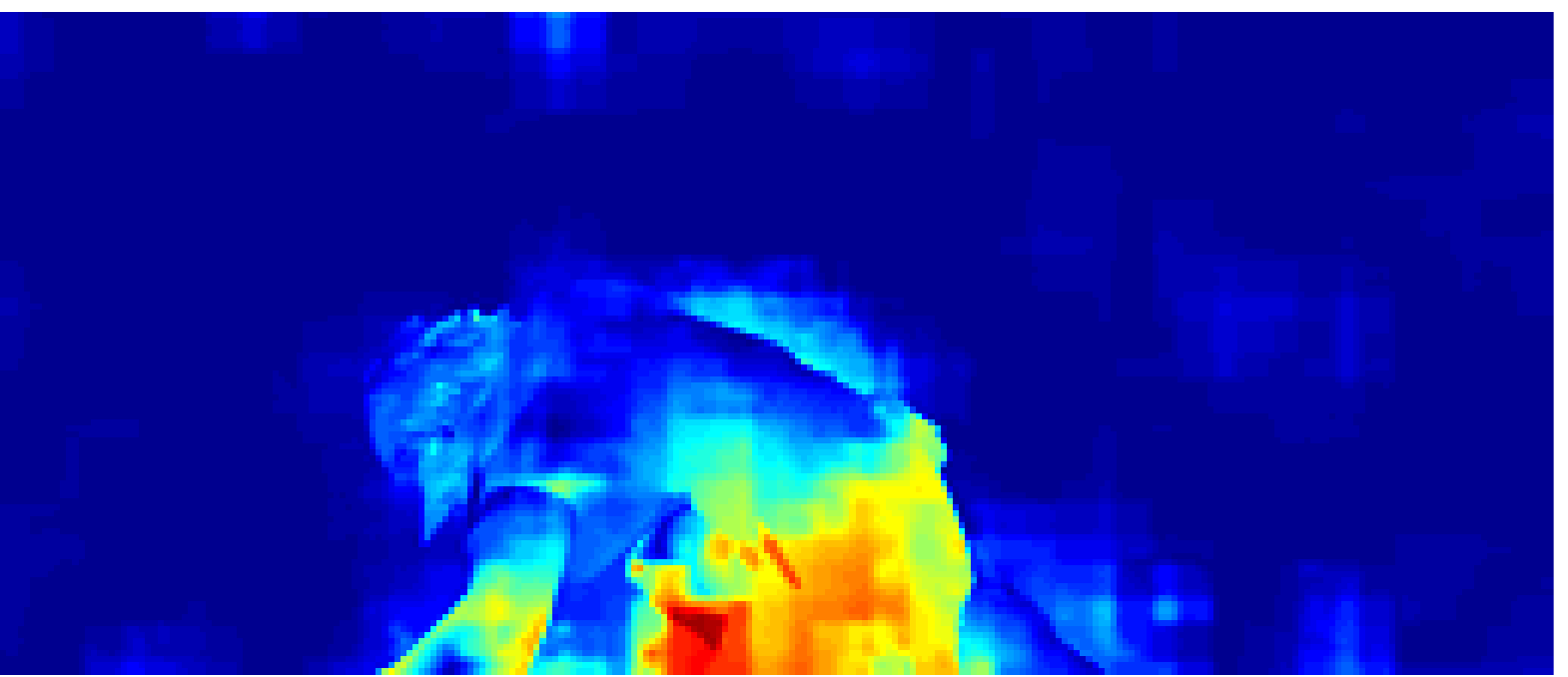}&
\includegraphics[width=0.195\textwidth]{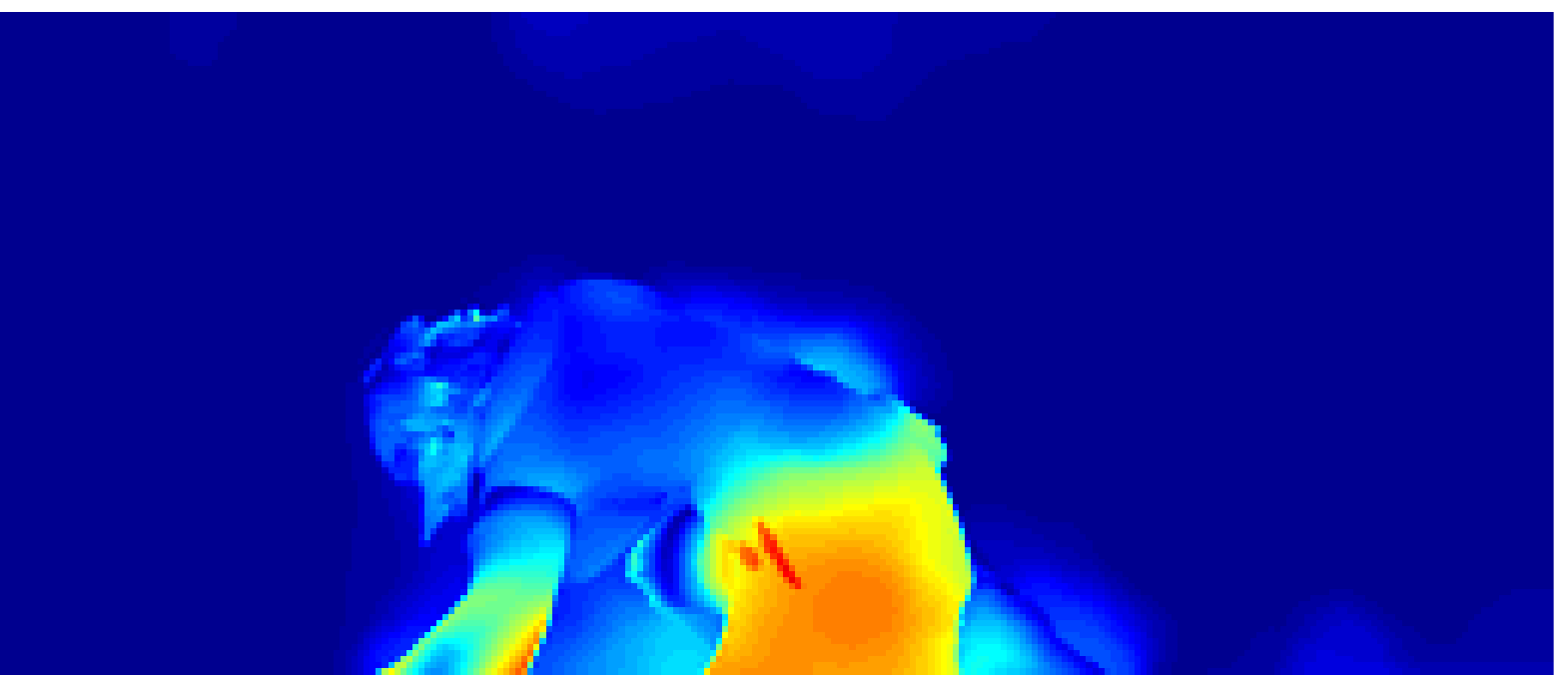}&
\includegraphics[width=0.195\textwidth]{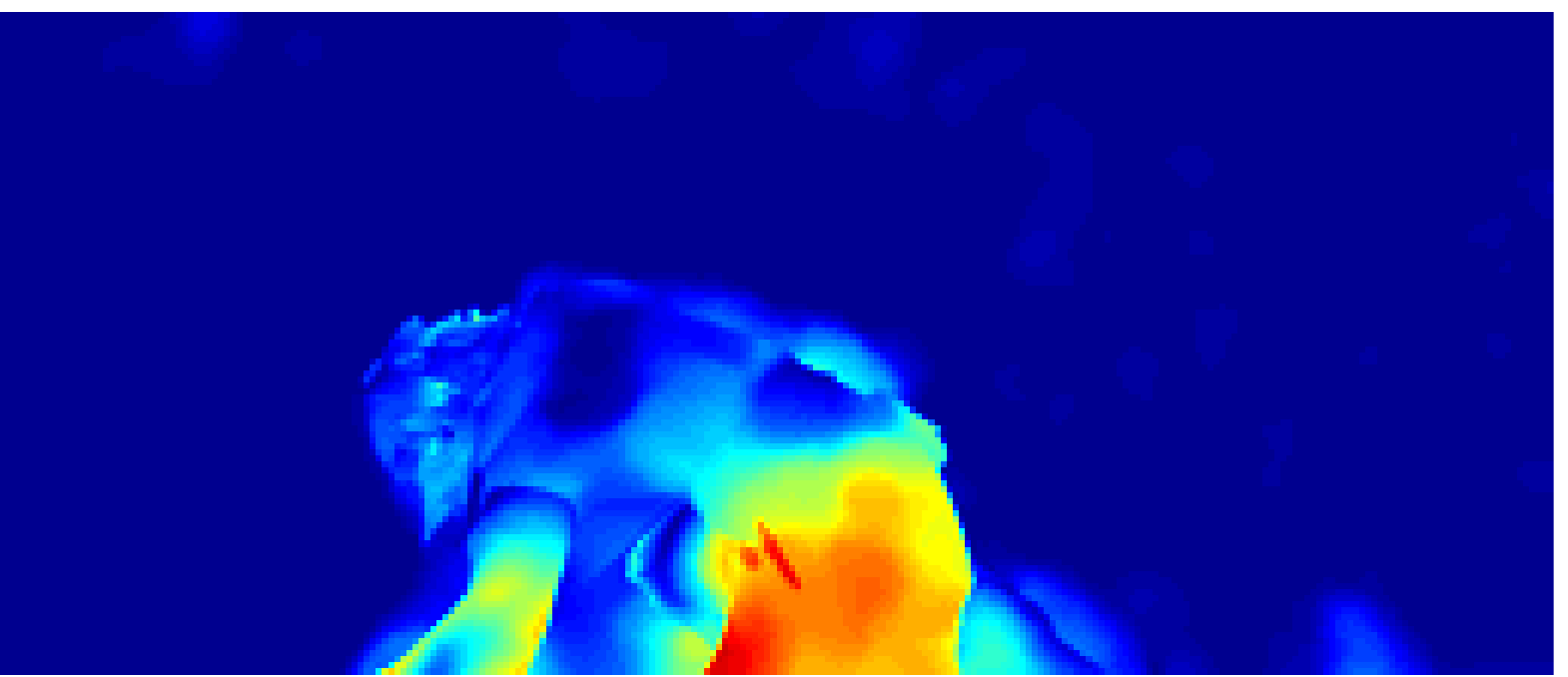}&
\includegraphics[width=0.195\textwidth]{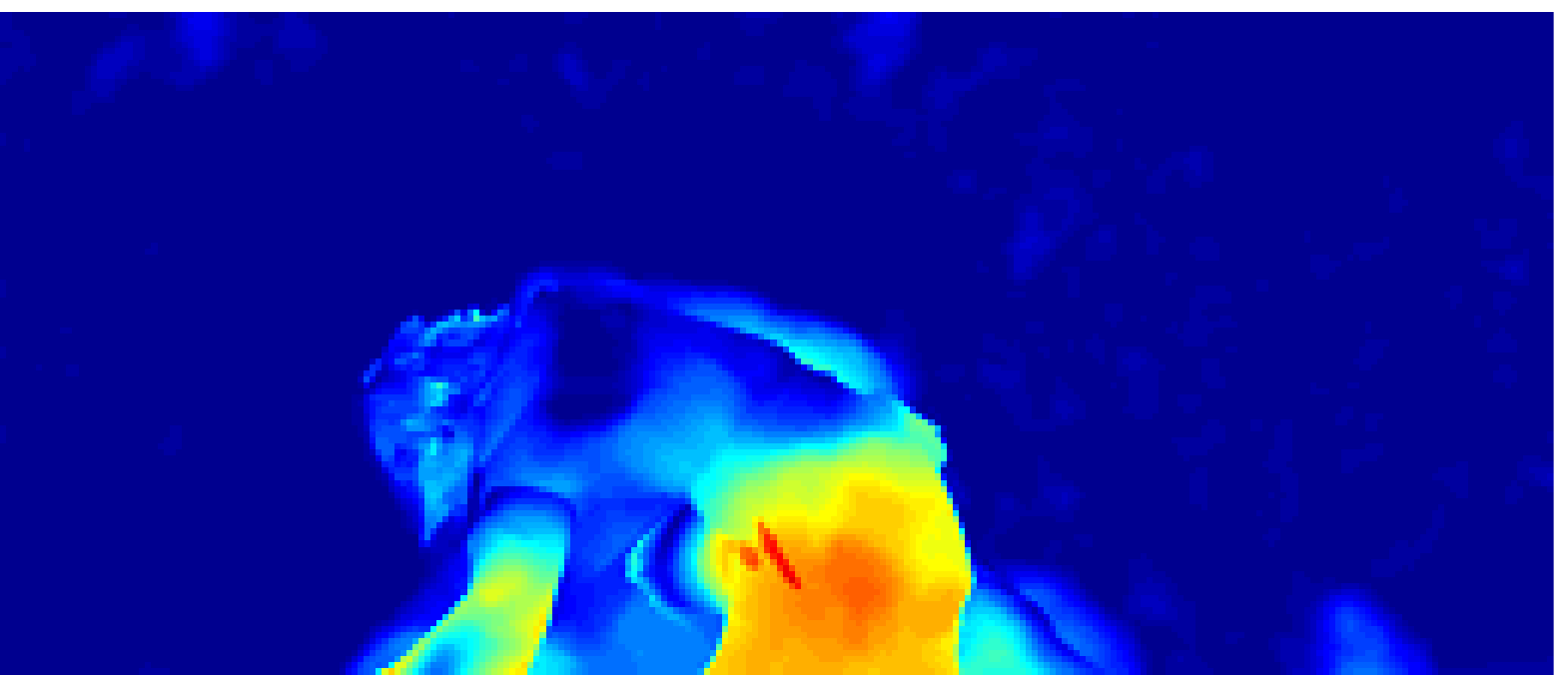}&
\includegraphics[width=0.195\textwidth]{imgs_exa/eximg-0141-10.pdf}\\[6pt]
\includegraphics[width=0.195\textwidth]{imgs_exa/eximg-0181-01.pdf}&
\includegraphics[width=0.195\textwidth]{imgs_exa/eximg-0181-02.pdf}&
\includegraphics[width=0.195\textwidth]{imgs_exa/eximg-0181-03.pdf}&
\includegraphics[width=0.195\textwidth]{imgs_exa/eximg-0181-04.pdf}&
\includegraphics[width=0.195\textwidth]{imgs_exa/eximg-0181-05.pdf}\\
\includegraphics[width=0.195\textwidth]{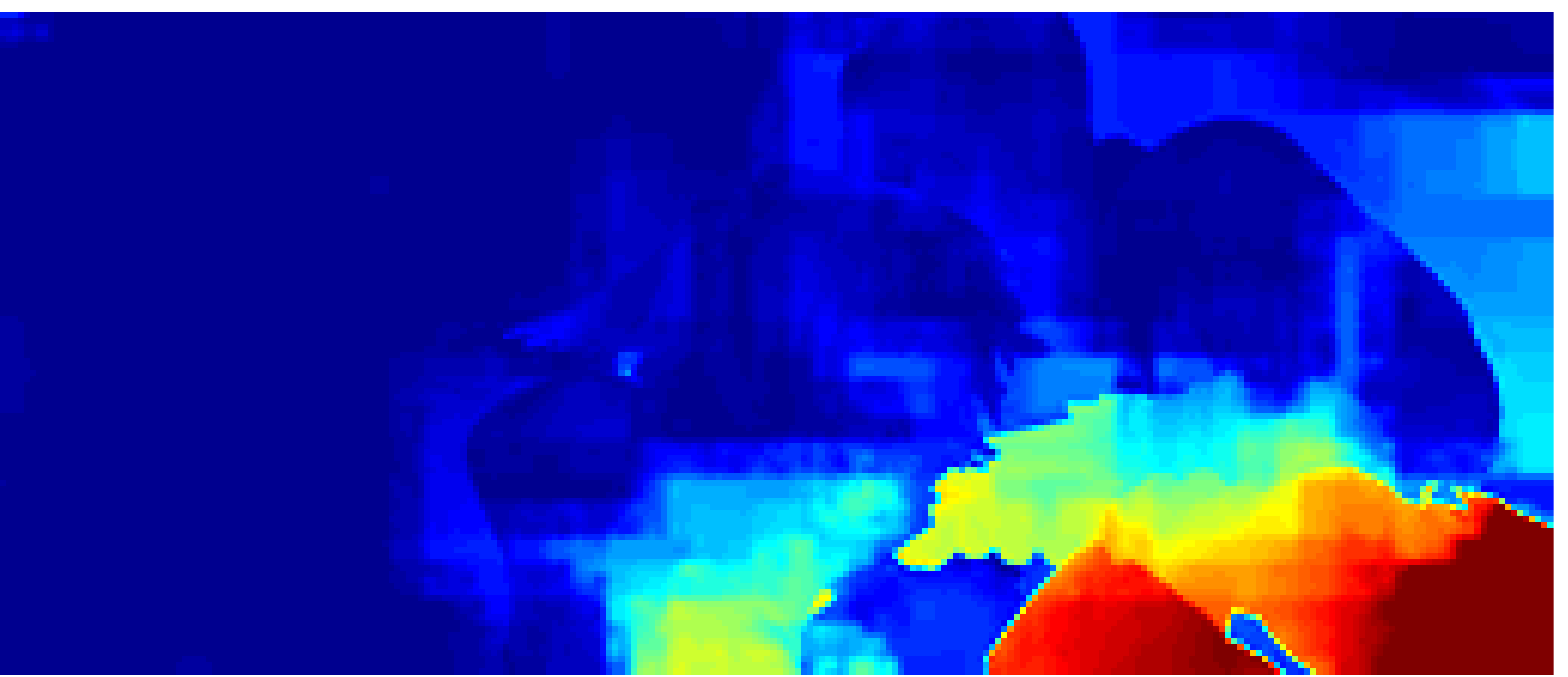}&
\includegraphics[width=0.195\textwidth]{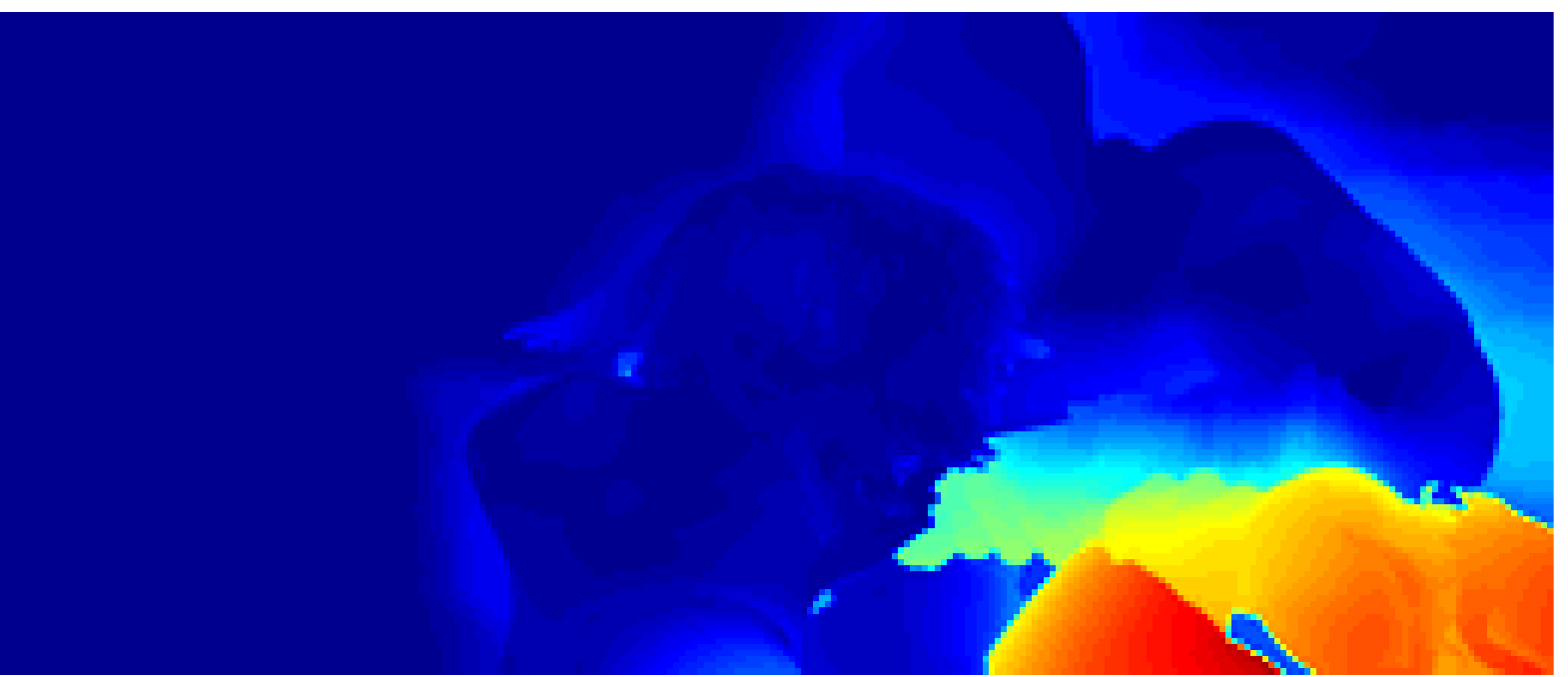}&
\includegraphics[width=0.195\textwidth]{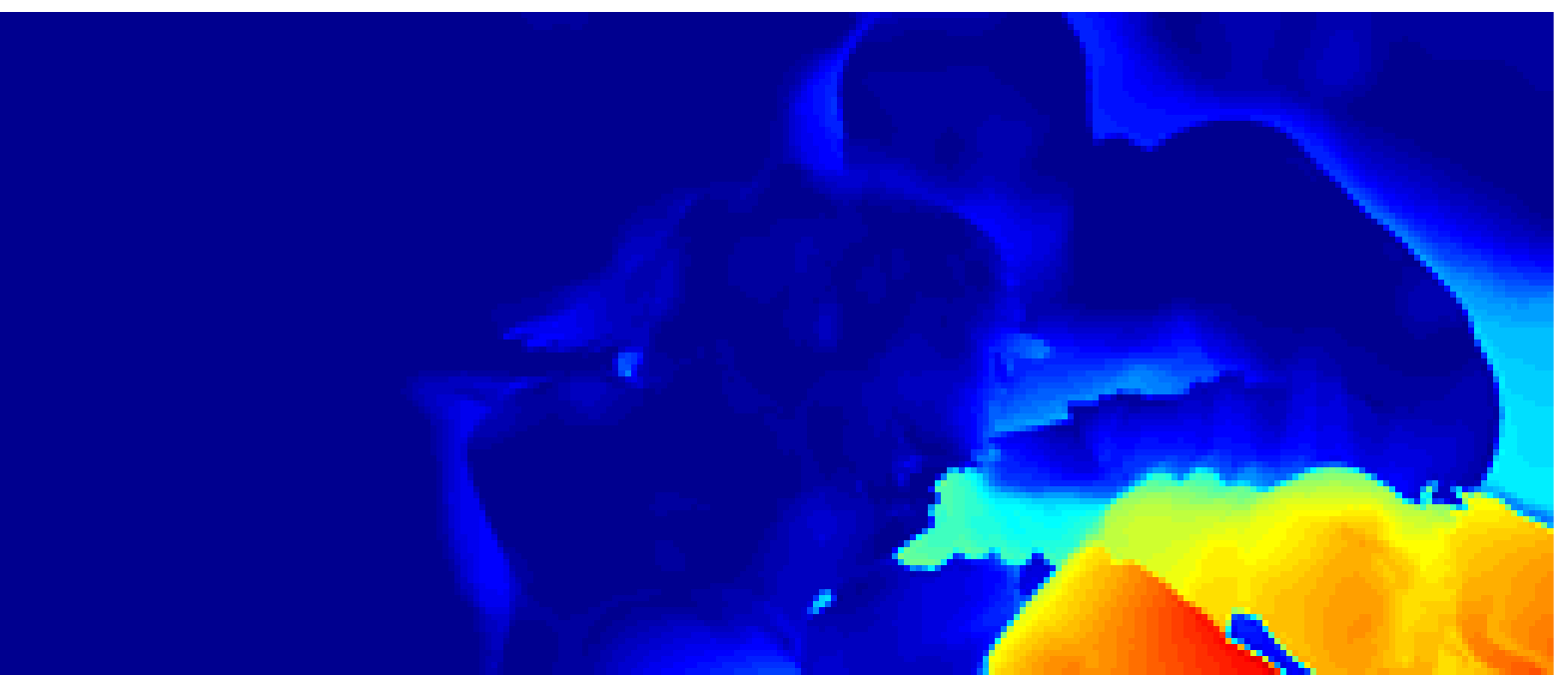}&
\includegraphics[width=0.195\textwidth]{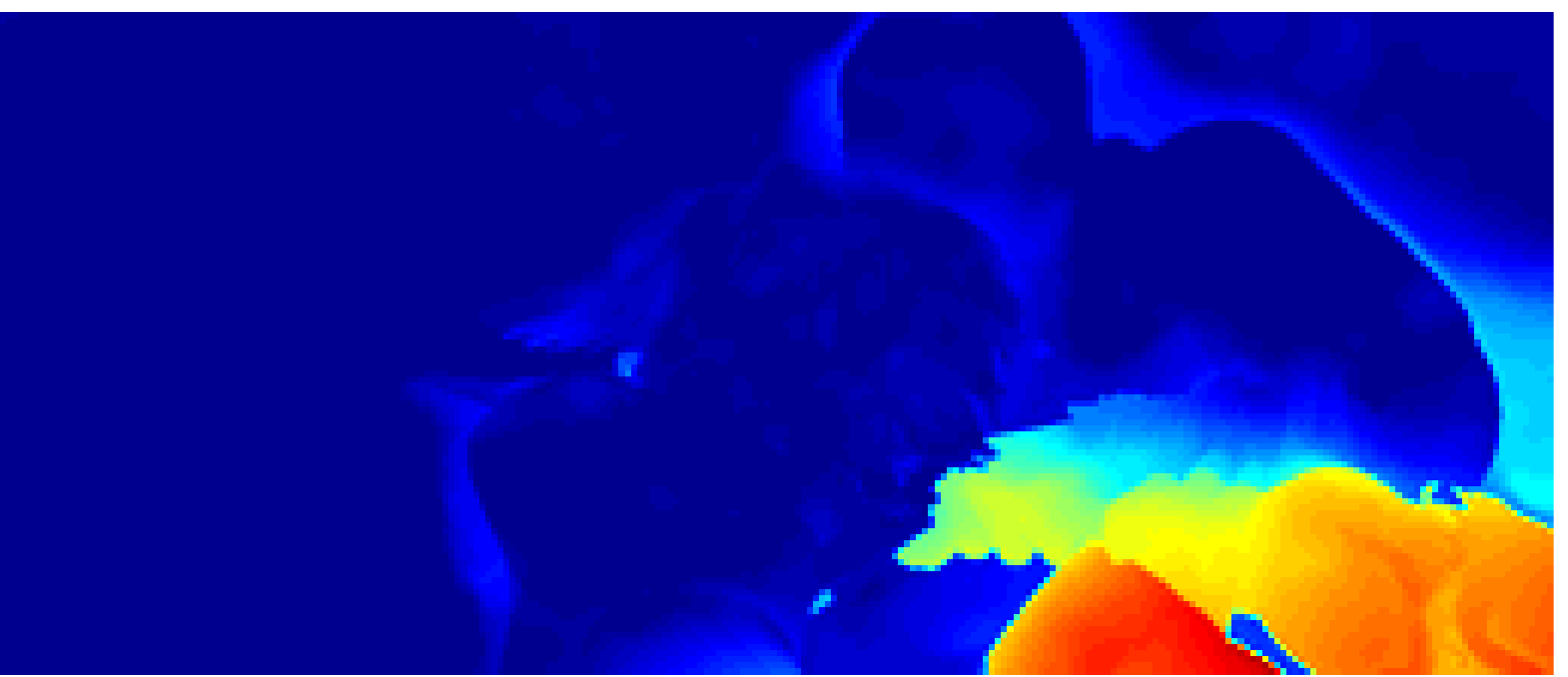}&
\includegraphics[width=0.195\textwidth]{imgs_exa/eximg-0181-10.pdf}\\[6pt]
\includegraphics[width=0.195\textwidth]{imgs_exa/eximg-0281-01.pdf}&
\includegraphics[width=0.195\textwidth]{imgs_exa/eximg-0281-02.pdf}&
\includegraphics[width=0.195\textwidth]{imgs_exa/eximg-0281-03.pdf}&
\includegraphics[width=0.195\textwidth]{imgs_exa/eximg-0281-04.pdf}&
\includegraphics[width=0.195\textwidth]{imgs_exa/eximg-0281-05.pdf}\\
\includegraphics[width=0.195\textwidth]{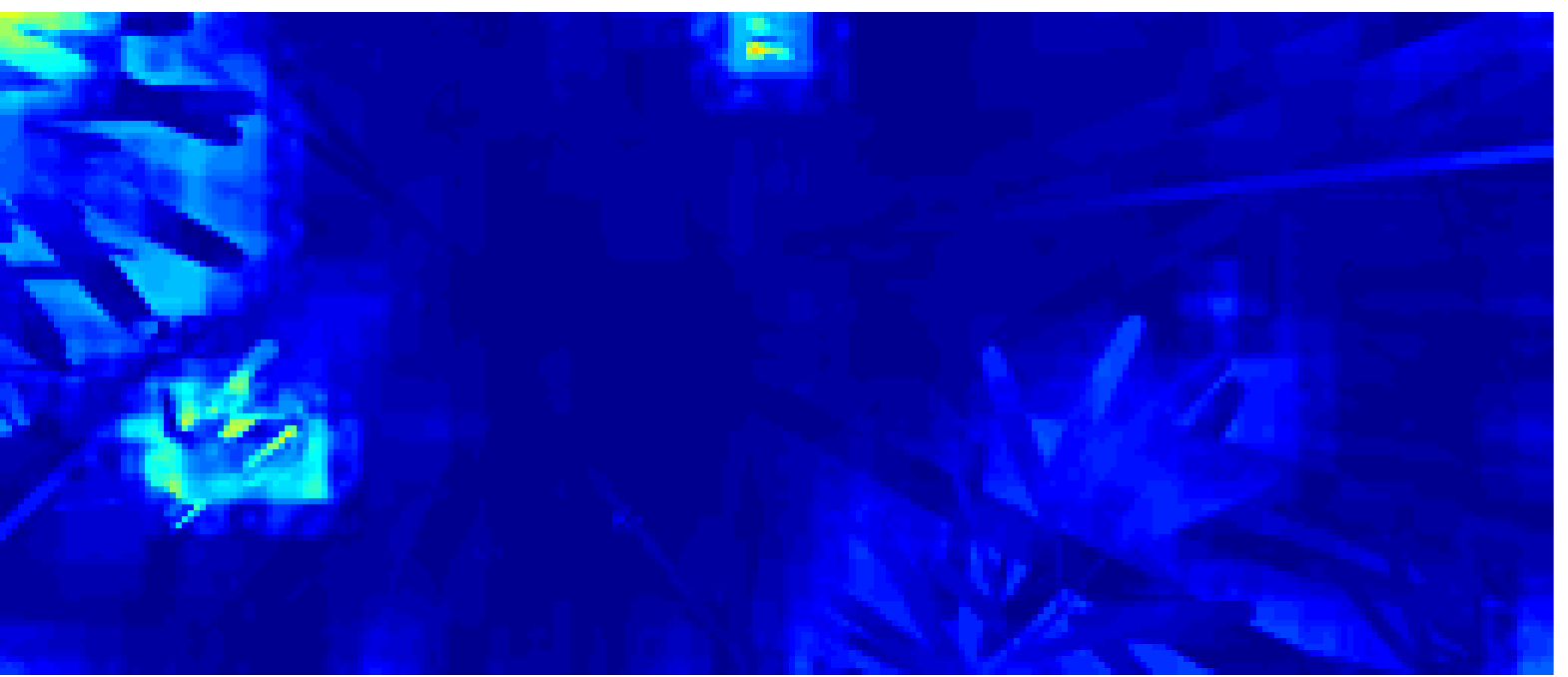}&
\includegraphics[width=0.195\textwidth]{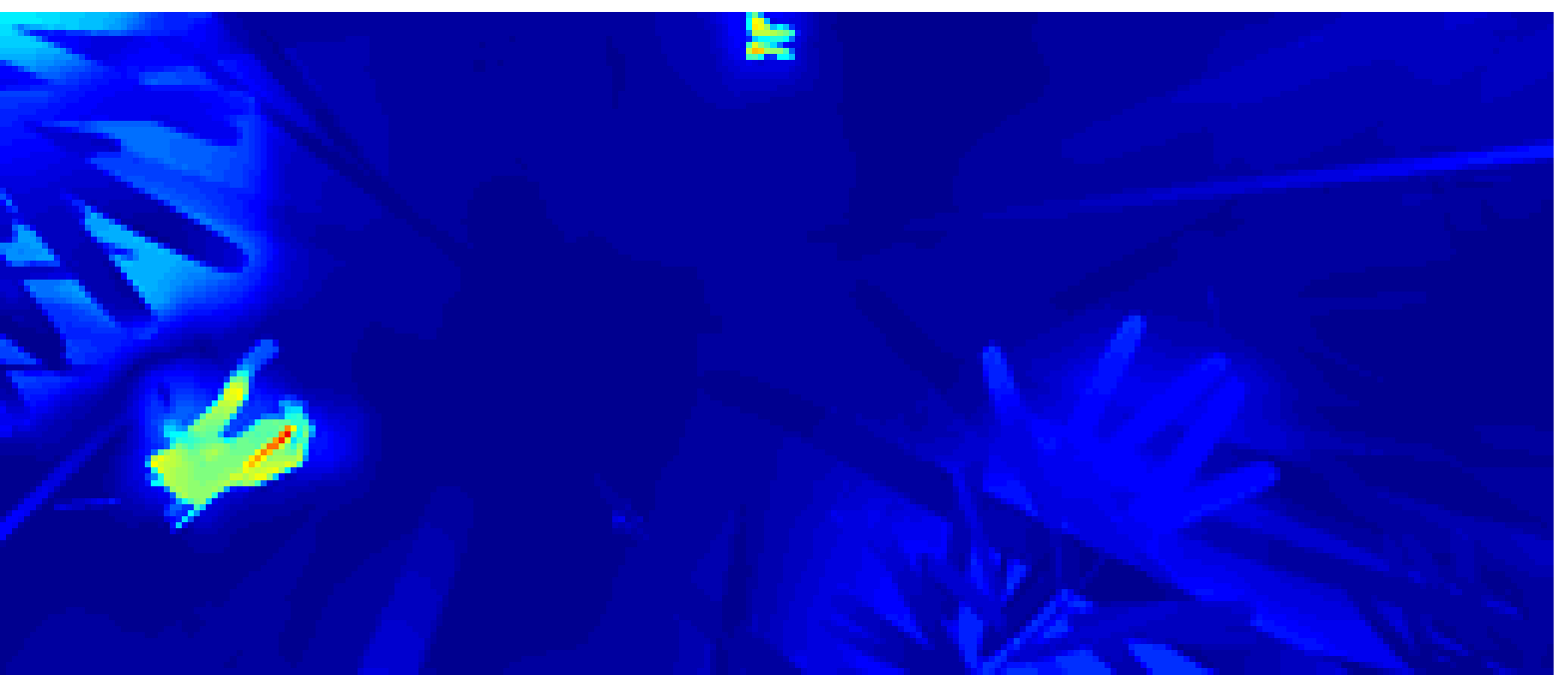}&
\includegraphics[width=0.195\textwidth]{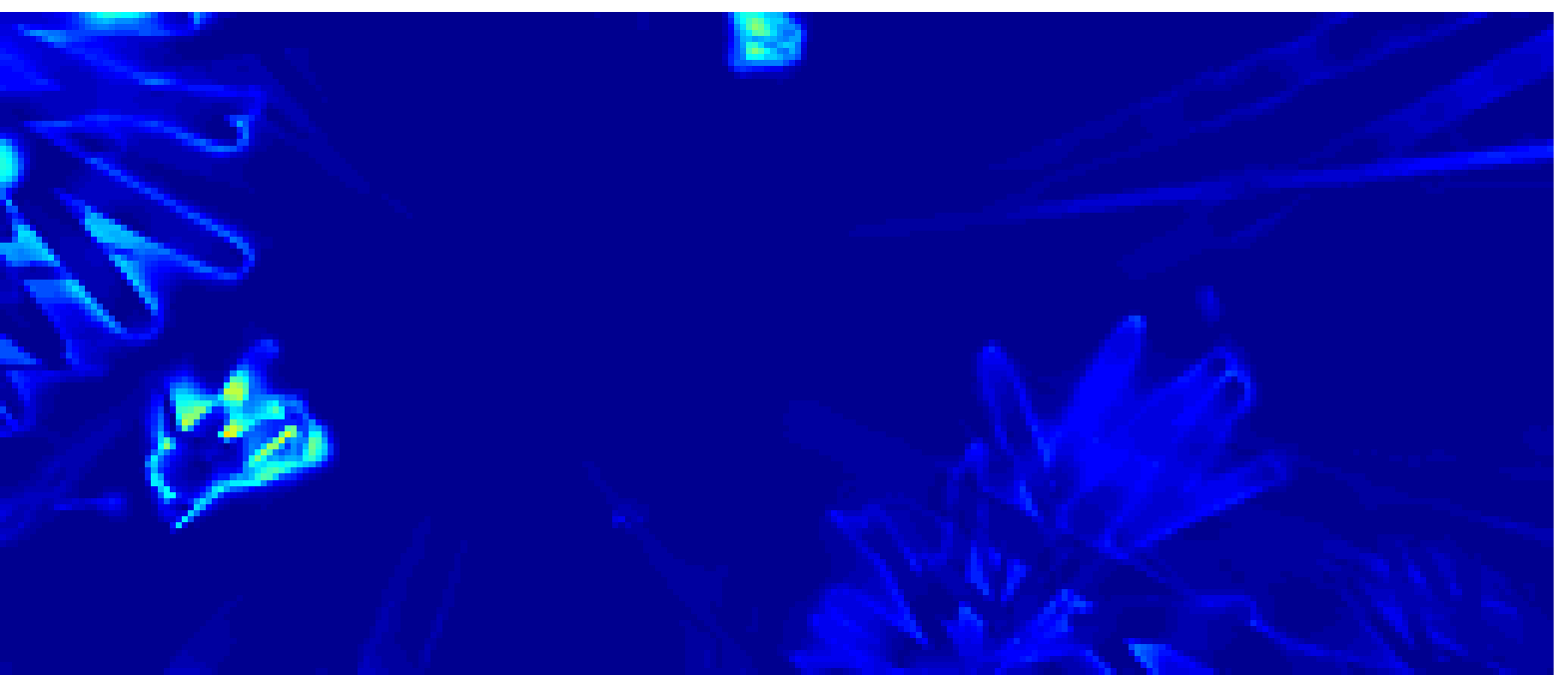}&
\includegraphics[width=0.195\textwidth]{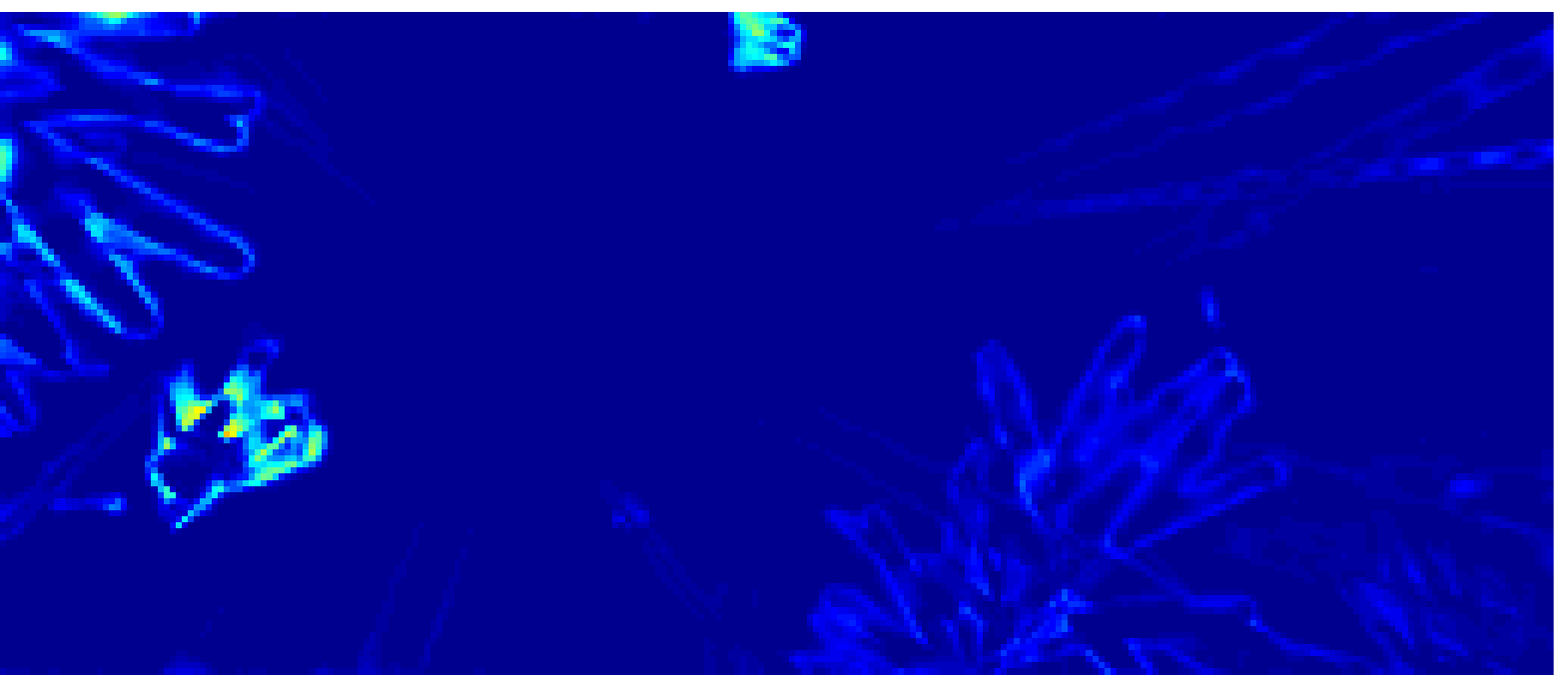}&
\includegraphics[width=0.195\textwidth]{imgs_exa/eximg-0281-10.pdf}\\[6pt]
\includegraphics[width=0.195\textwidth]{imgs_exa/eximg-0321-01.pdf}&
\includegraphics[width=0.195\textwidth]{imgs_exa/eximg-0321-02.pdf}&
\includegraphics[width=0.195\textwidth]{imgs_exa/eximg-0321-03.pdf}&
\includegraphics[width=0.195\textwidth]{imgs_exa/eximg-0321-04.pdf}&
\includegraphics[width=0.195\textwidth]{imgs_exa/eximg-0321-05.pdf}\\
\includegraphics[width=0.195\textwidth]{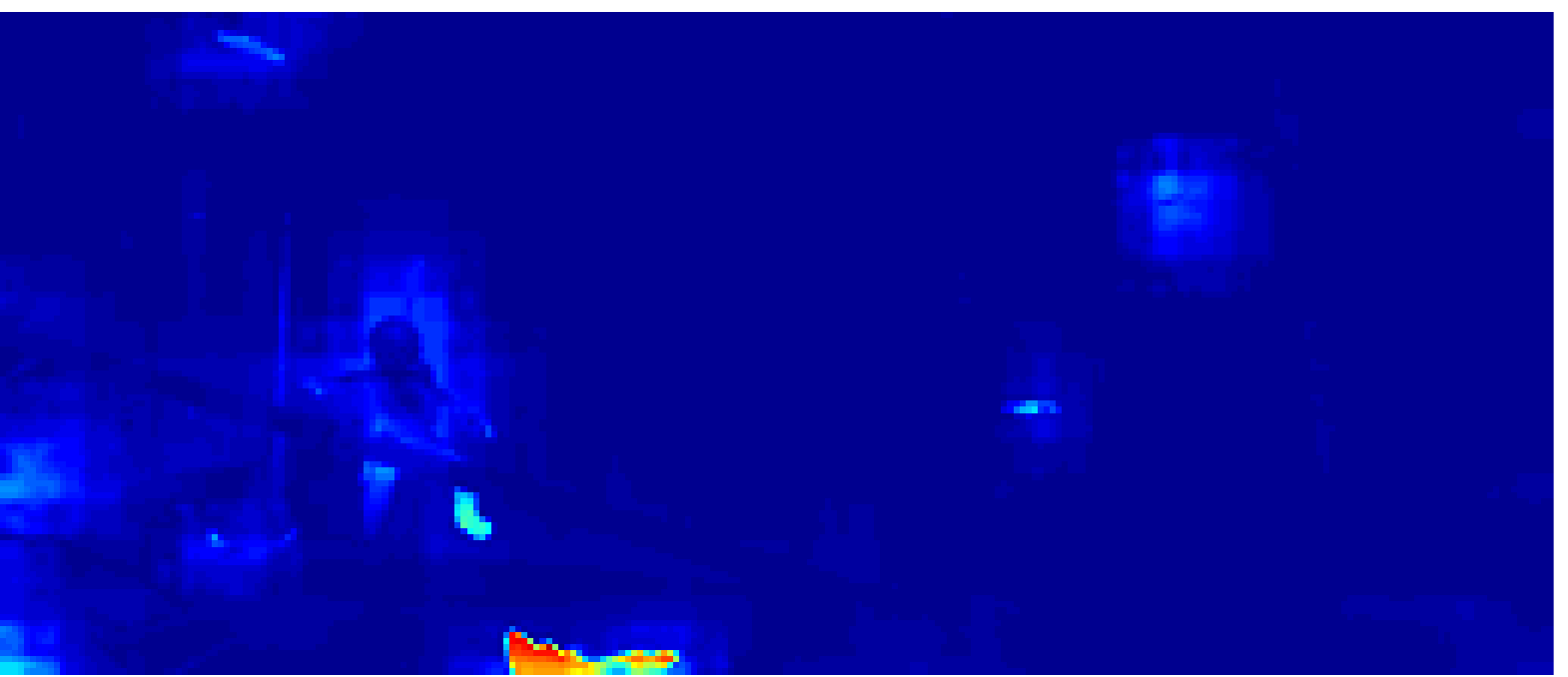}&
\includegraphics[width=0.195\textwidth]{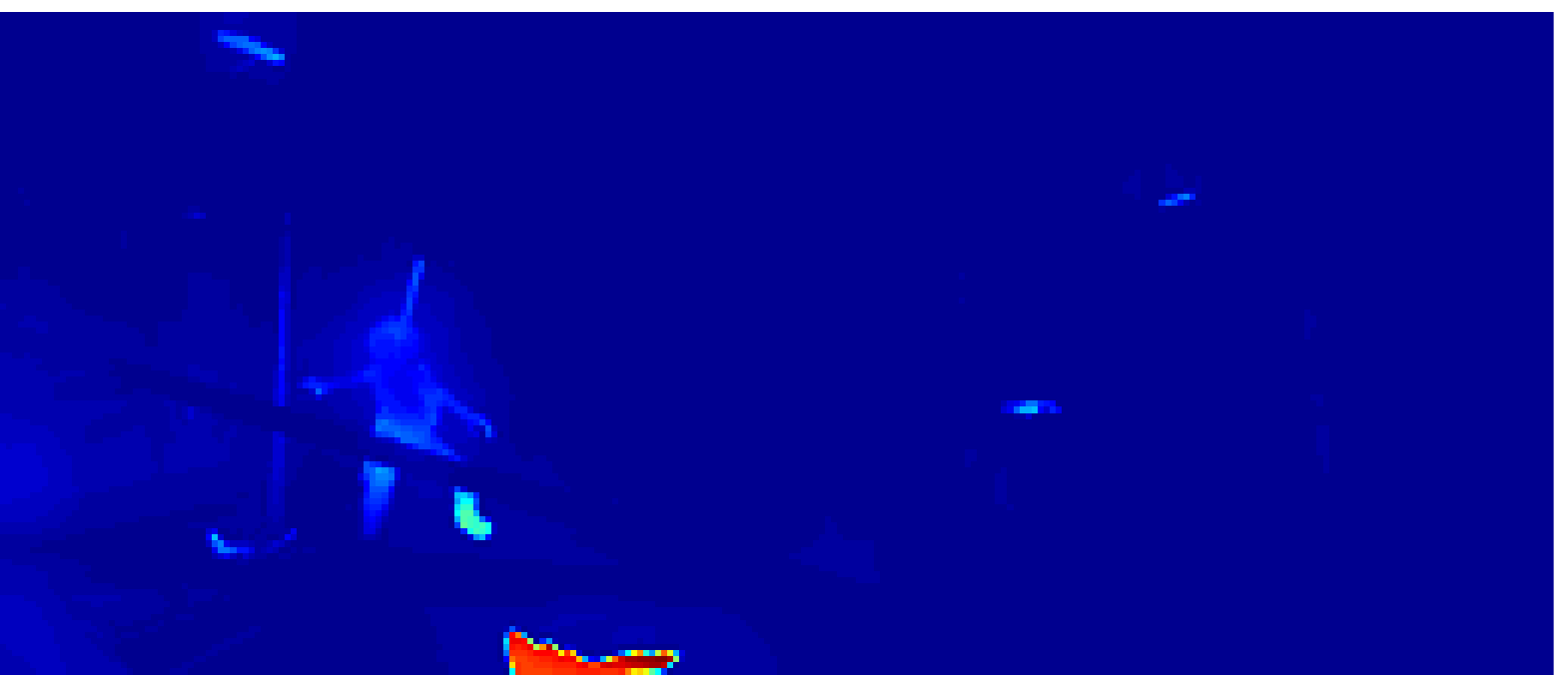}&
\includegraphics[width=0.195\textwidth]{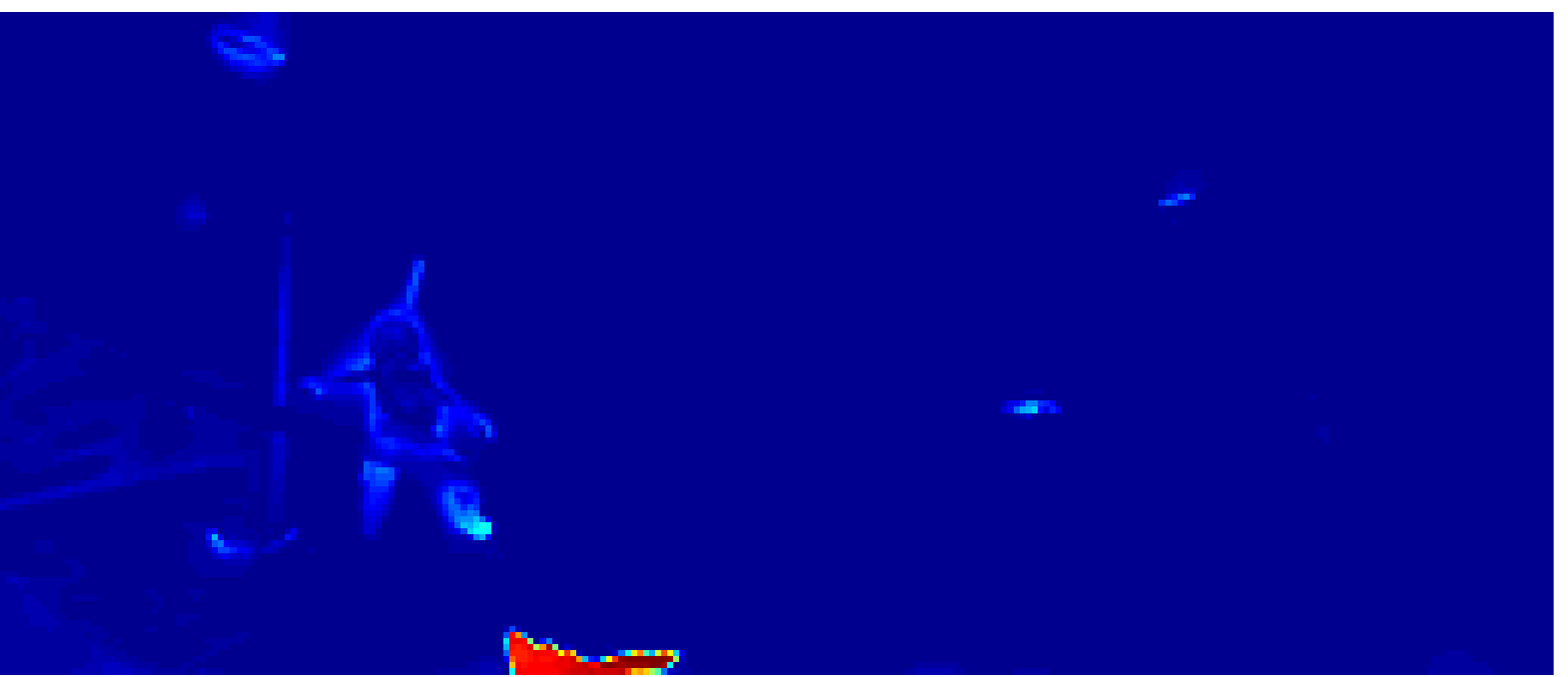}&
\includegraphics[width=0.195\textwidth]{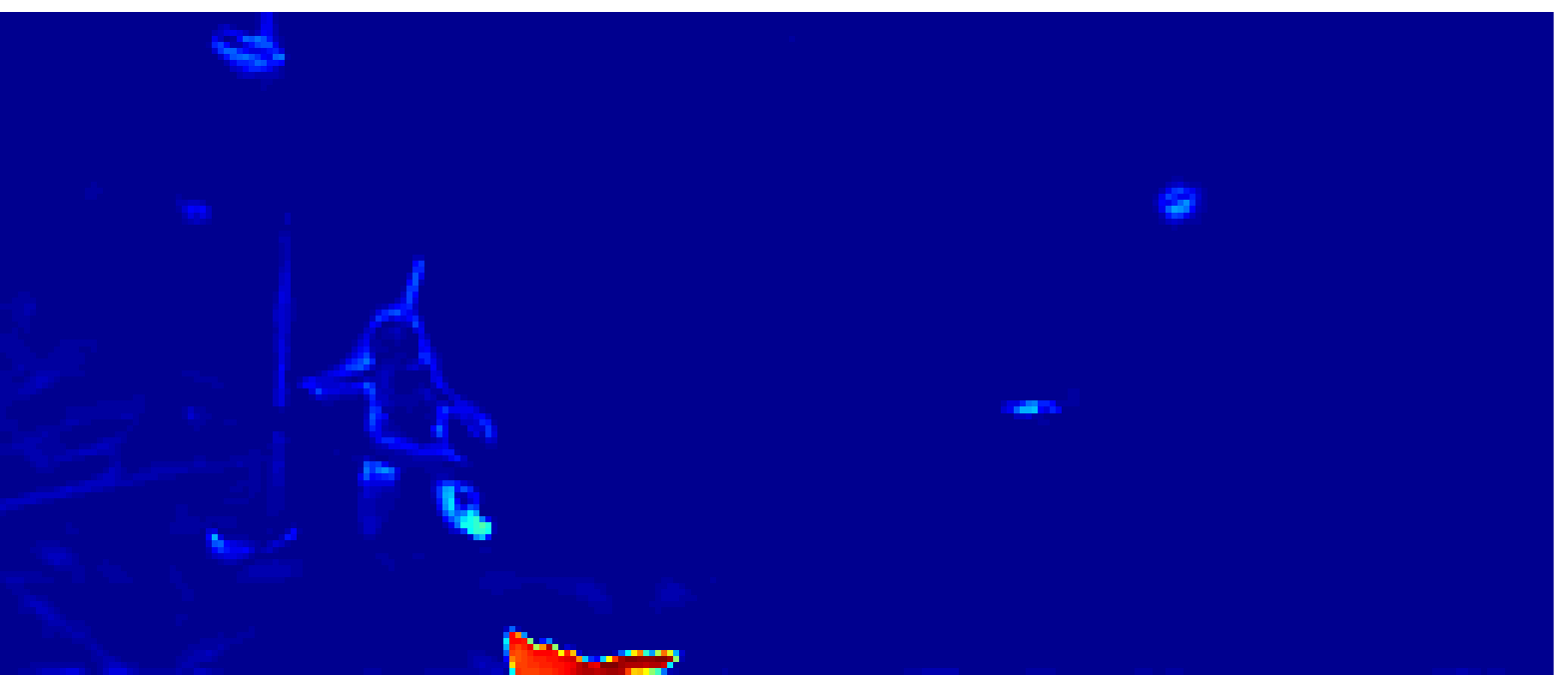}&
\includegraphics[width=0.195\textwidth]{imgs_exa/eximg-0321-10.pdf}\\[6pt]
\includegraphics[width=0.195\textwidth]{imgs_exa/eximg-0581-01.pdf}&
\includegraphics[width=0.195\textwidth]{imgs_exa/eximg-0581-02.pdf}&
\includegraphics[width=0.195\textwidth]{imgs_exa/eximg-0581-03.pdf}&
\includegraphics[width=0.195\textwidth]{imgs_exa/eximg-0581-04.pdf}&
\includegraphics[width=0.195\textwidth]{imgs_exa/eximg-0581-05.pdf}\\
\includegraphics[width=0.195\textwidth]{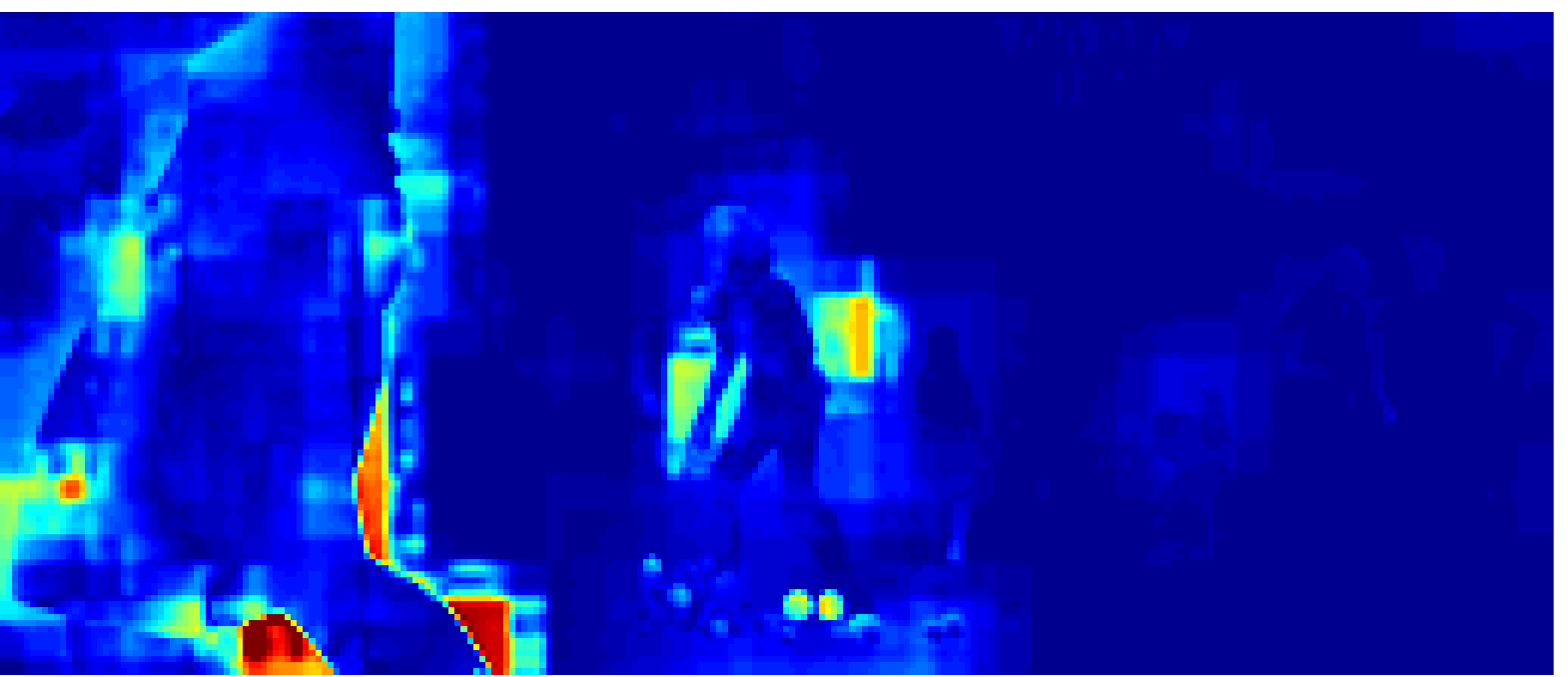}&
\includegraphics[width=0.195\textwidth]{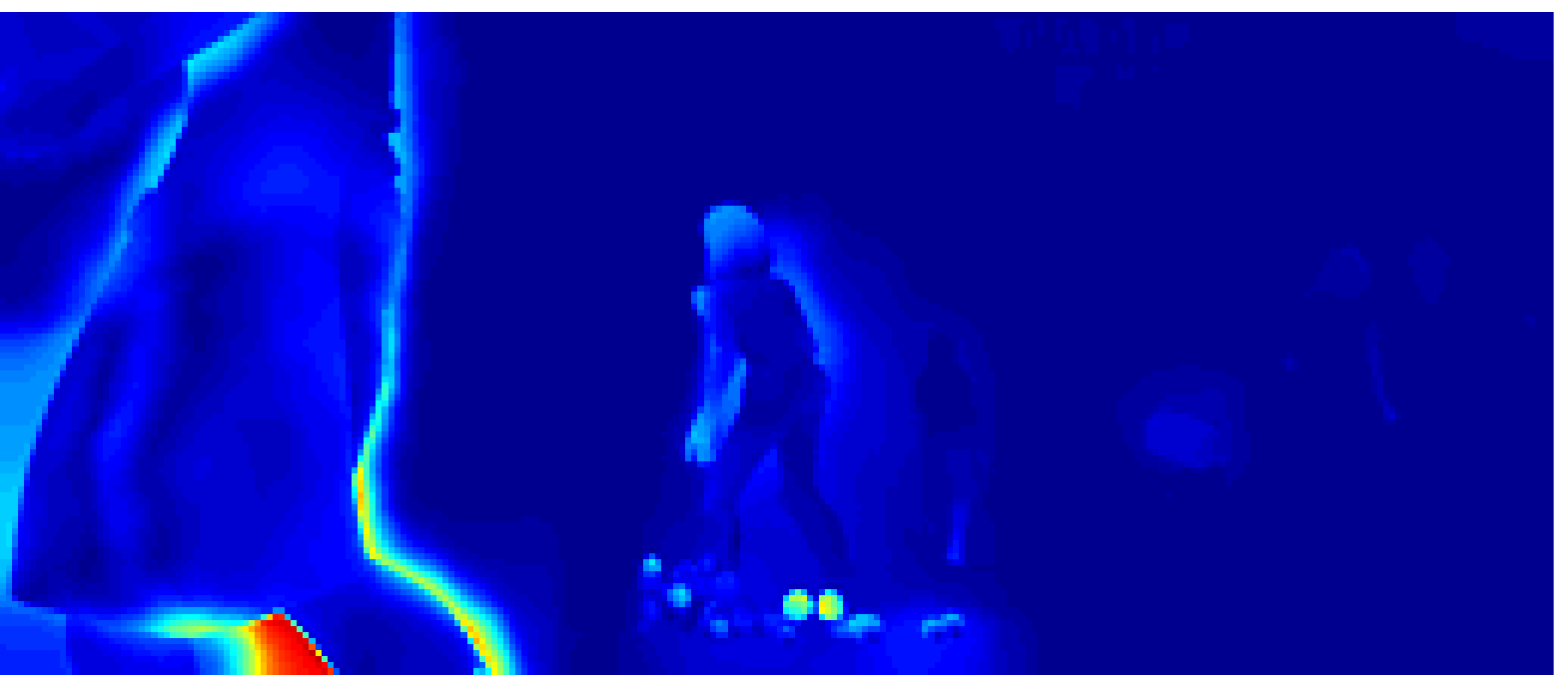}&
\includegraphics[width=0.195\textwidth]{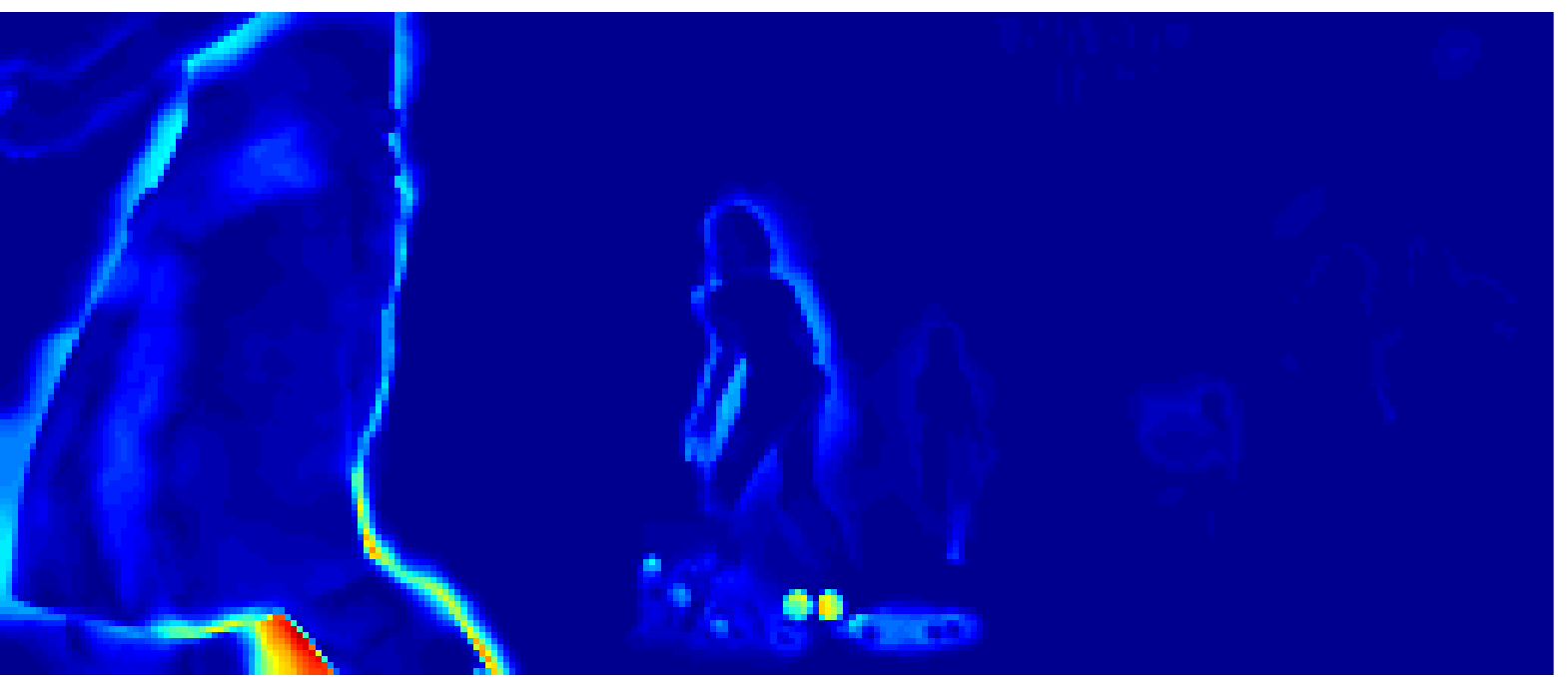}&
\includegraphics[width=0.195\textwidth]{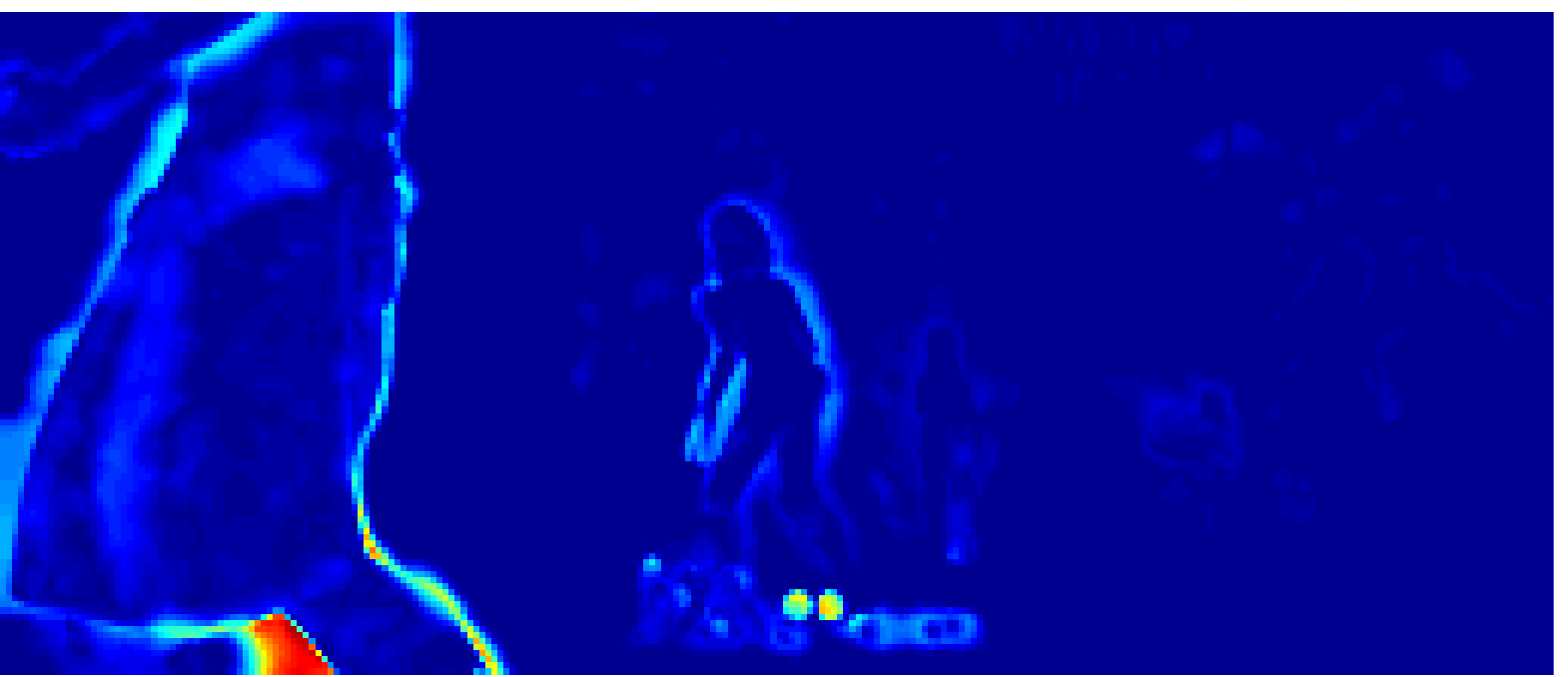}&
\includegraphics[width=0.195\textwidth]{imgs_exa/eximg-0581-10.pdf}\\[6pt]
\end{tabular}
\caption{Exemplary results on Sintel (training) and error maps. In each block of $2 \times 6$ images.  Top row, left to right: Our method for operating points ({\bf 1})-({\bf 4}), Ground Truth. Bottom row: Error heat maps scaled from blue (no error) to red (maximum ground truth flow magnitude), Original Image.}\label{fig:sintel1res_errmap_AP} \end{figure*}

\begin{figure*} [!ht]
\centering\setlength{\tabcolsep}{0.1pt}\renewcommand{\arraystretch}{0} 
{
\begin{tabular}{ccccc}
 {\bf 600Hz} & {\bf 300Hz} & {\bf 10Hz} & {\bf 0.5Hz}& {\bf Ground Truth}\\
\includegraphics[width=0.195\textwidth]{imgs_exa/eximg-0621-01.pdf}&
\includegraphics[width=0.195\textwidth]{imgs_exa/eximg-0621-02.pdf}&
\includegraphics[width=0.195\textwidth]{imgs_exa/eximg-0621-03.pdf}&
\includegraphics[width=0.195\textwidth]{imgs_exa/eximg-0621-04.pdf}&
\includegraphics[width=0.195\textwidth]{imgs_exa/eximg-0621-05.pdf}\\
\includegraphics[width=0.195\textwidth]{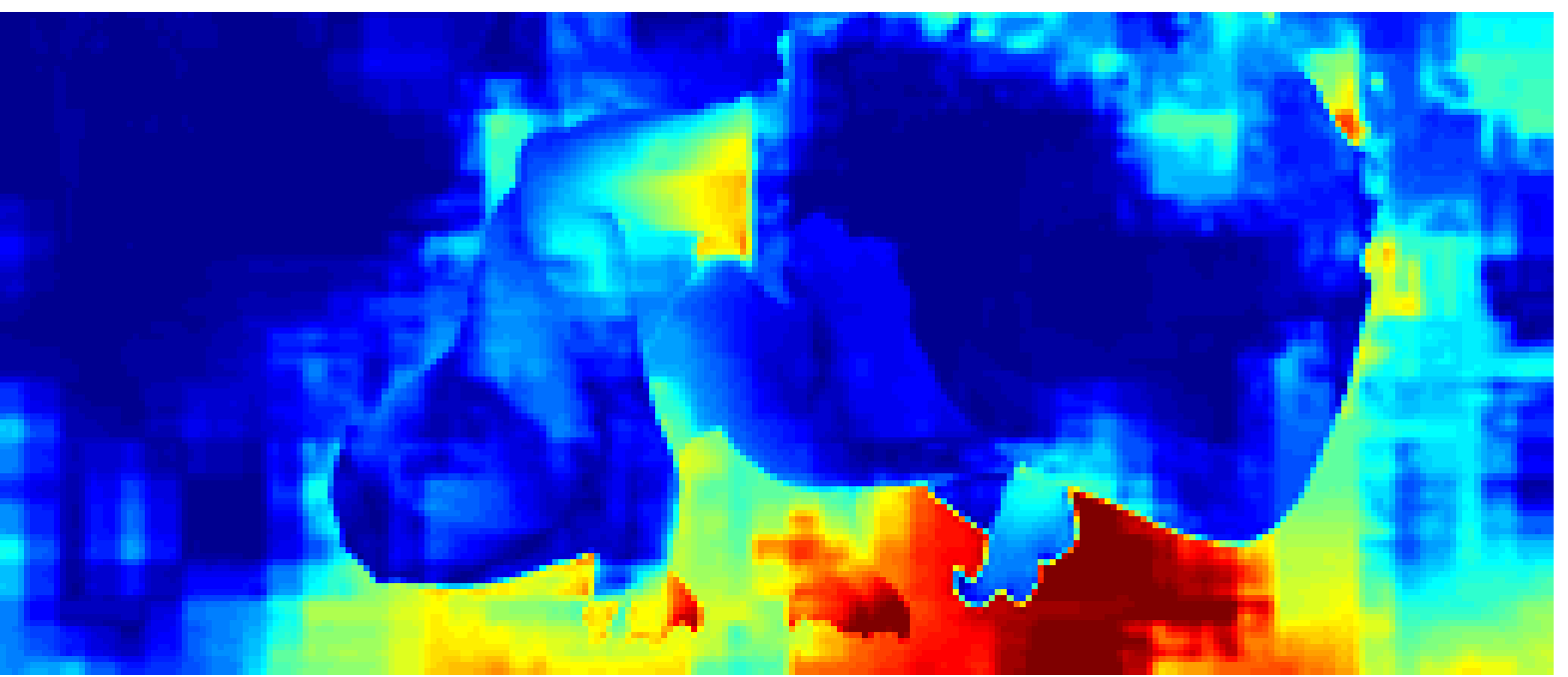}&
\includegraphics[width=0.195\textwidth]{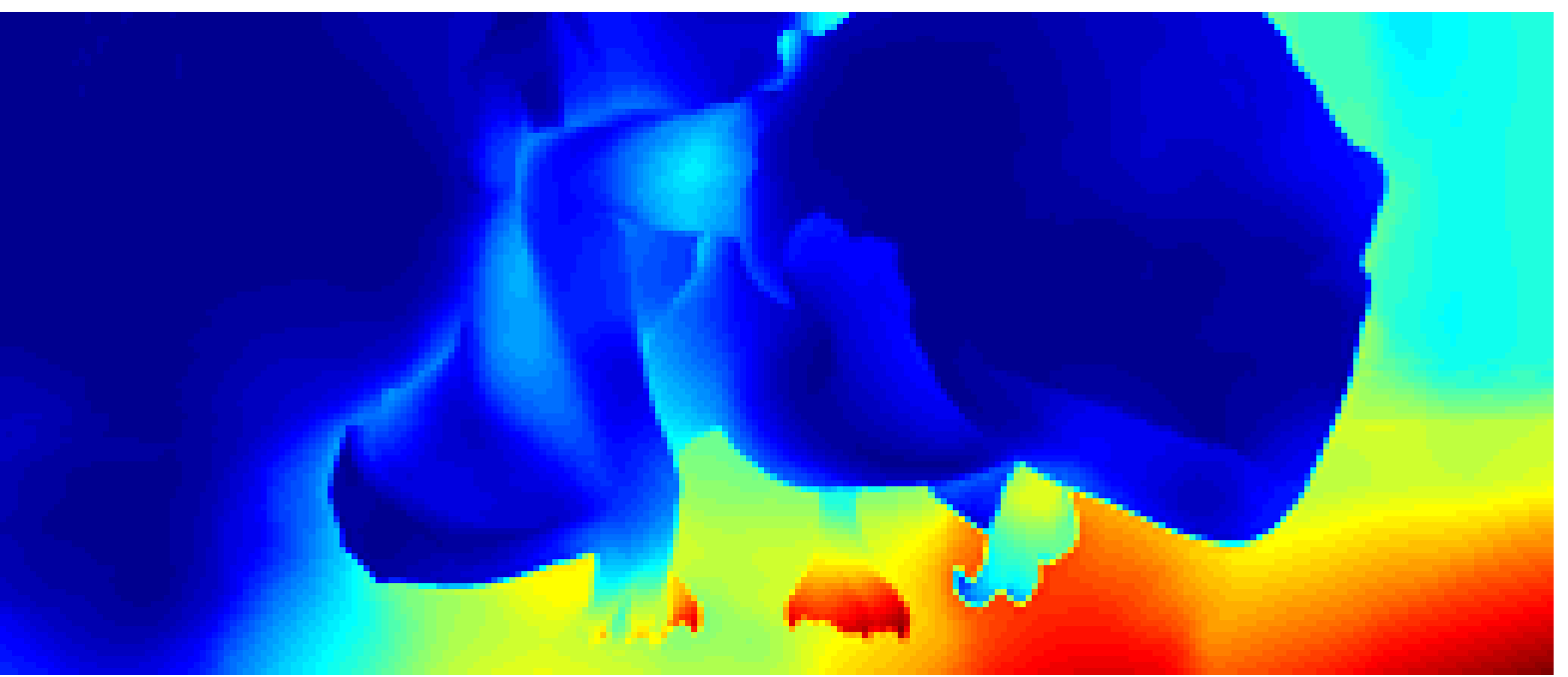}&
\includegraphics[width=0.195\textwidth]{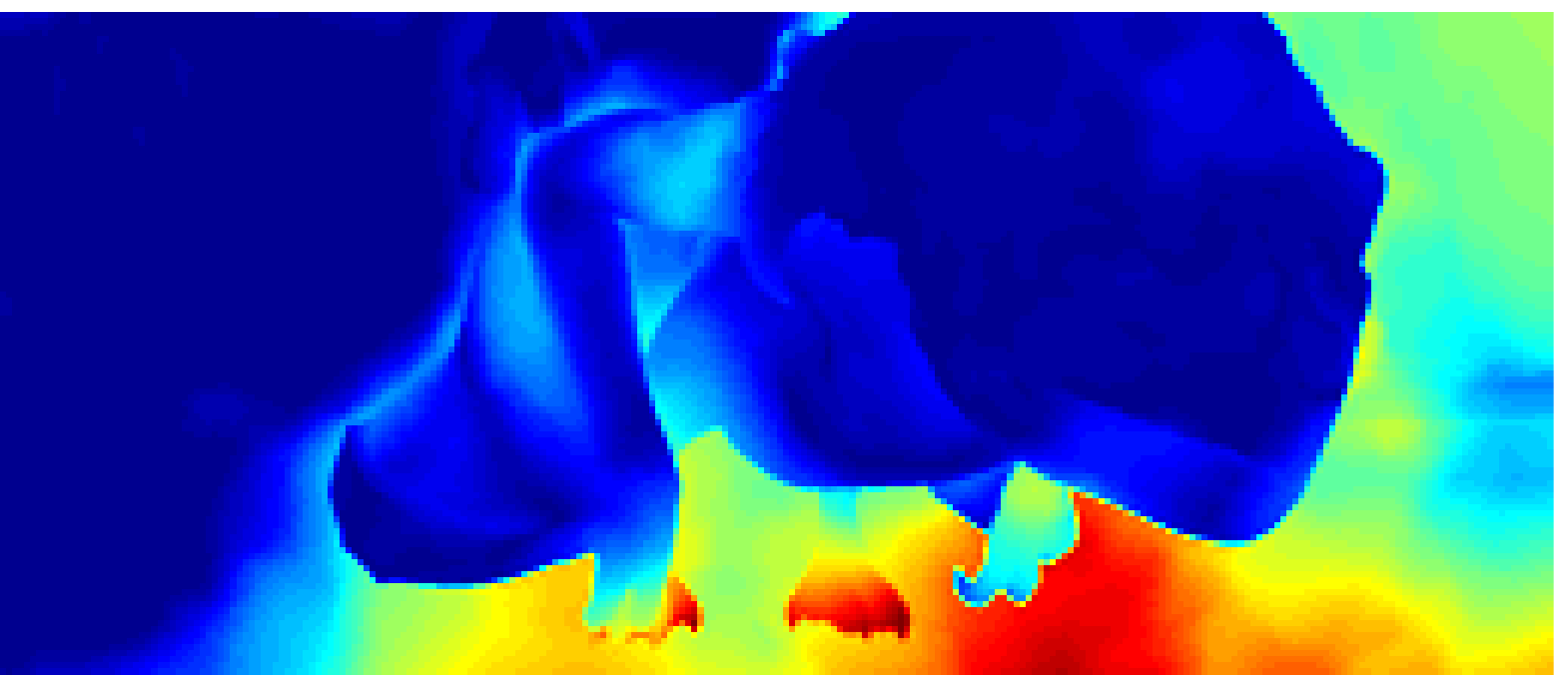}&
\includegraphics[width=0.195\textwidth]{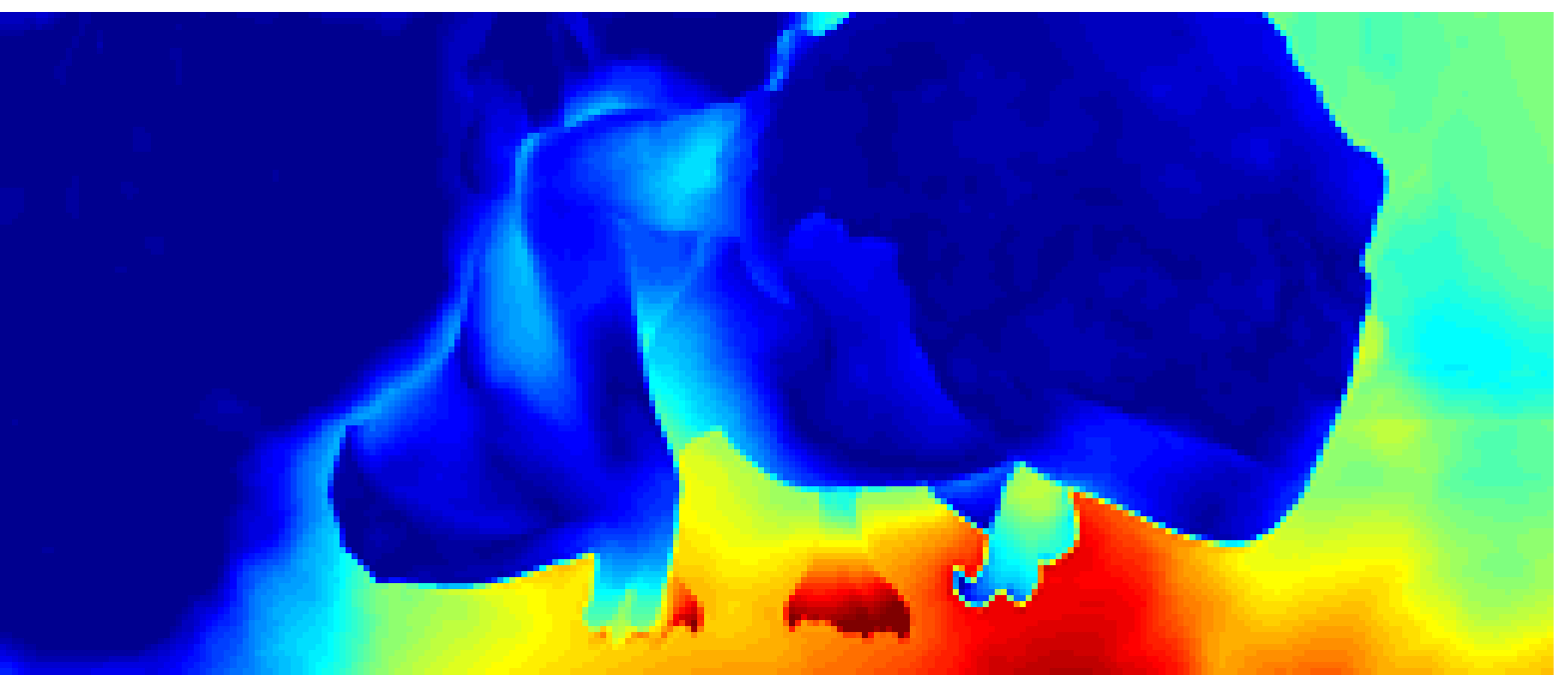}&
\includegraphics[width=0.195\textwidth]{imgs_exa/eximg-0621-10.pdf}\\[6pt]
\includegraphics[width=0.195\textwidth]{imgs_exa/eximg-0661-01.pdf}&
\includegraphics[width=0.195\textwidth]{imgs_exa/eximg-0661-02.pdf}&
\includegraphics[width=0.195\textwidth]{imgs_exa/eximg-0661-03.pdf}&
\includegraphics[width=0.195\textwidth]{imgs_exa/eximg-0661-04.pdf}&
\includegraphics[width=0.195\textwidth]{imgs_exa/eximg-0661-05.pdf}\\
\includegraphics[width=0.195\textwidth]{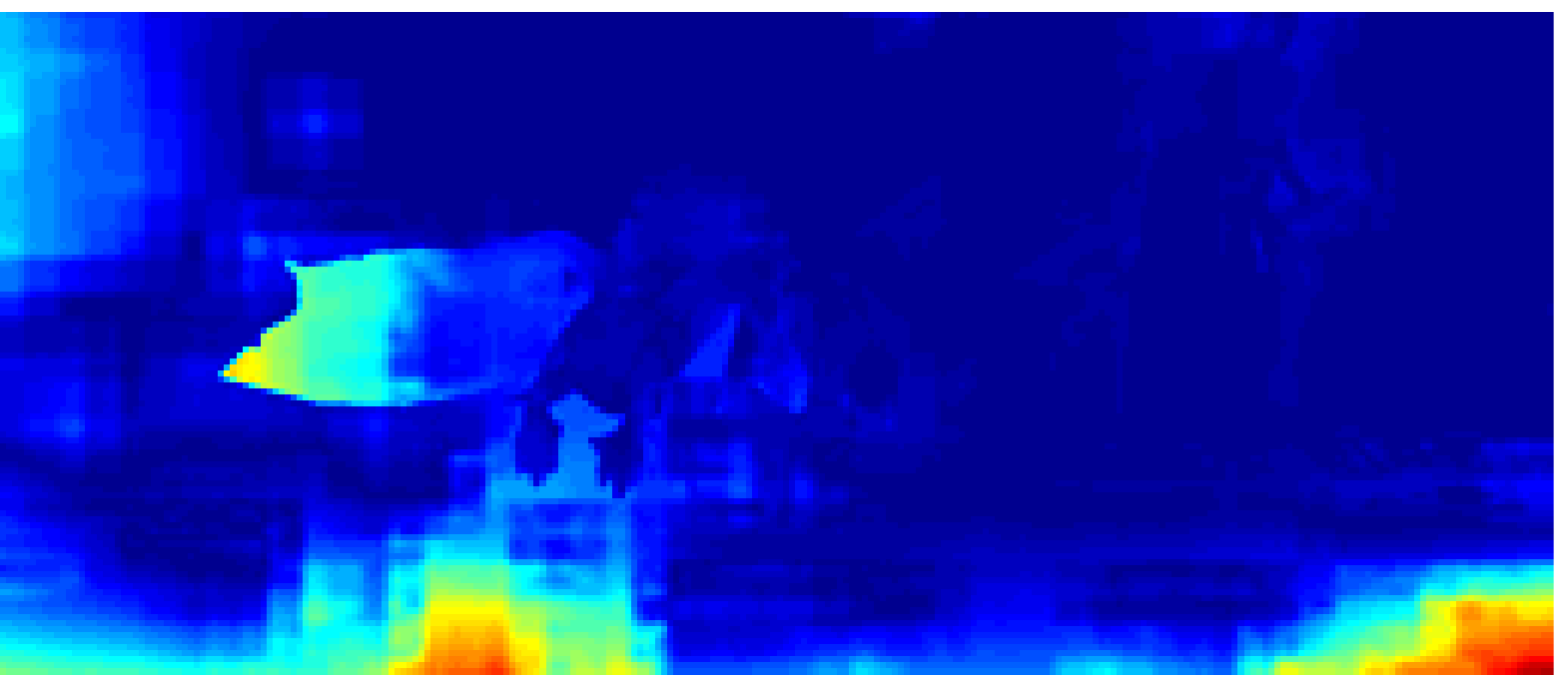}&
\includegraphics[width=0.195\textwidth]{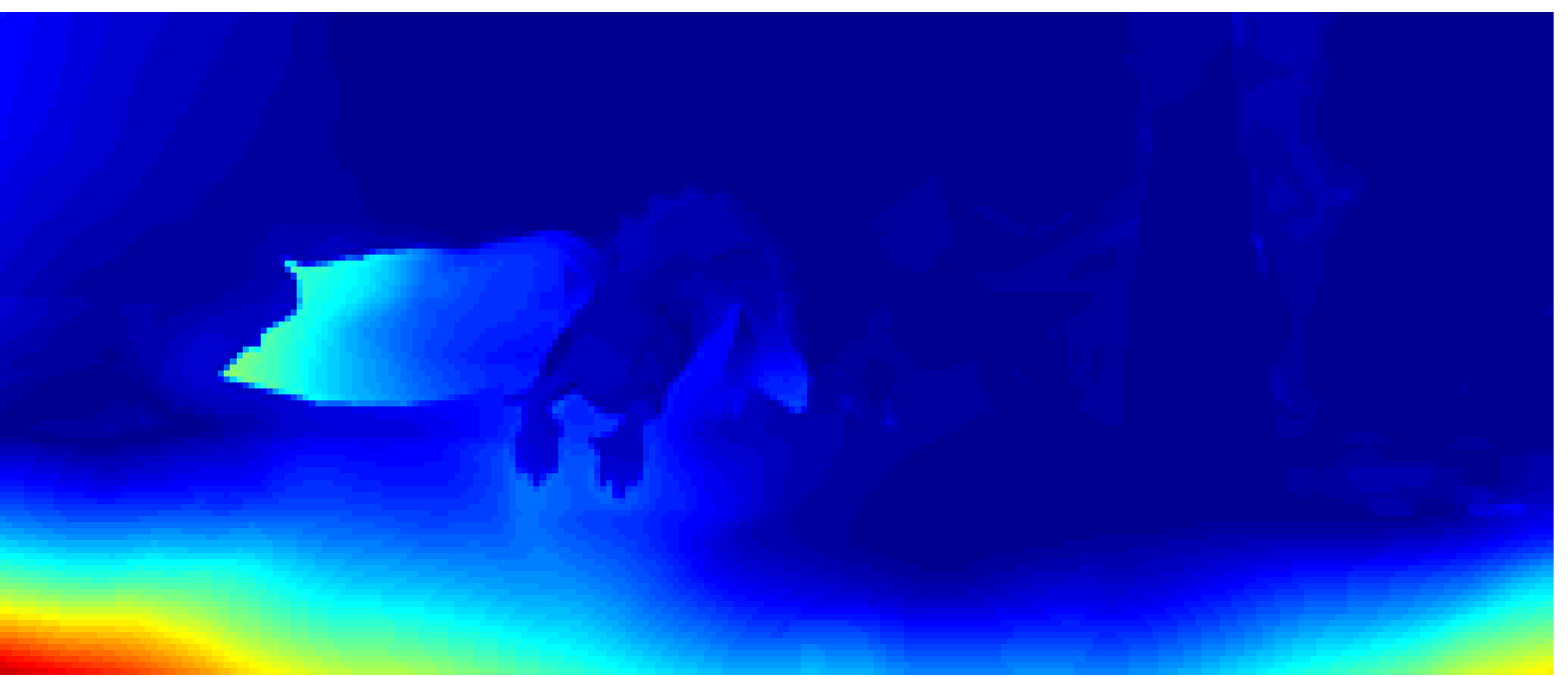}&
\includegraphics[width=0.195\textwidth]{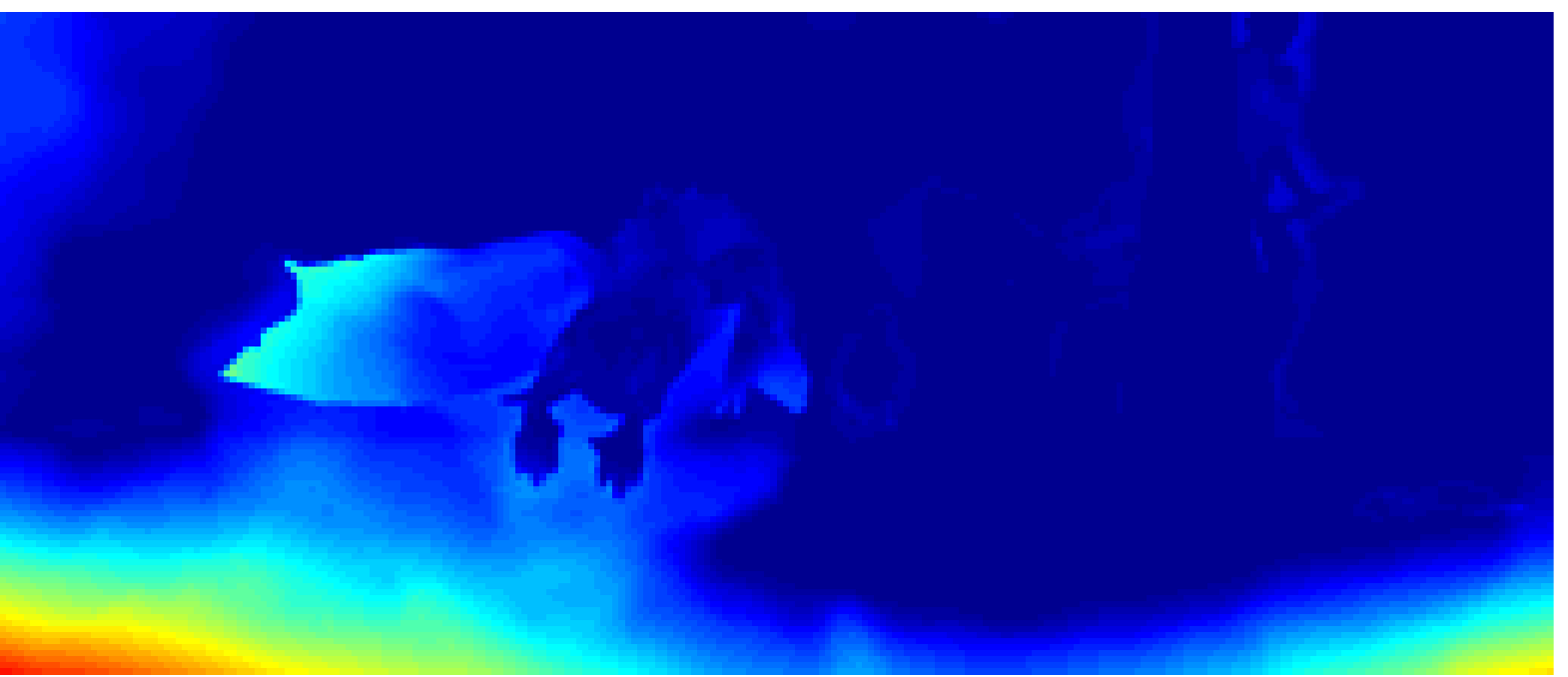}&
\includegraphics[width=0.195\textwidth]{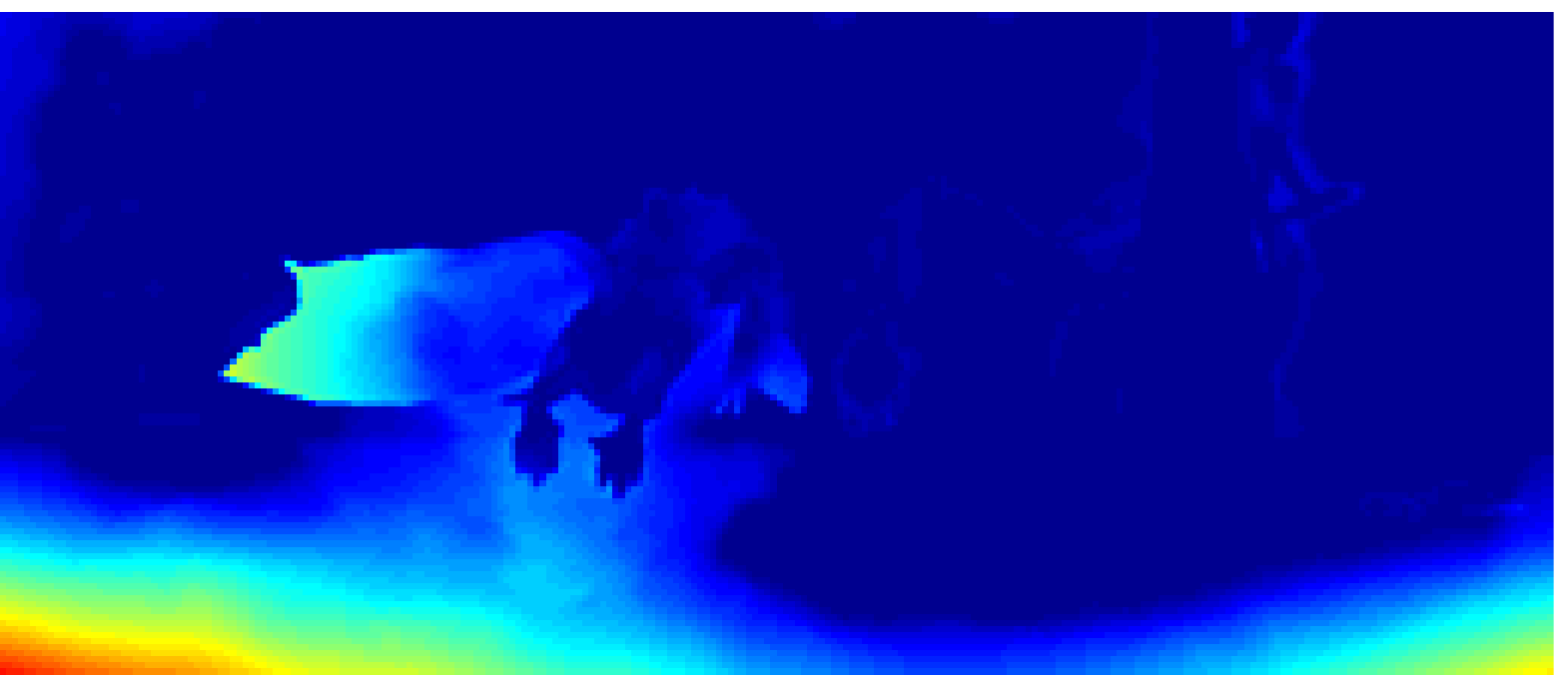}&
\includegraphics[width=0.195\textwidth]{imgs_exa/eximg-0661-10.pdf}\\[6pt]
\includegraphics[width=0.195\textwidth]{imgs_exa/eximg-0721-01.pdf}&
\includegraphics[width=0.195\textwidth]{imgs_exa/eximg-0721-02.pdf}&
\includegraphics[width=0.195\textwidth]{imgs_exa/eximg-0721-03.pdf}&
\includegraphics[width=0.195\textwidth]{imgs_exa/eximg-0721-04.pdf}&
\includegraphics[width=0.195\textwidth]{imgs_exa/eximg-0721-05.pdf}\\
\includegraphics[width=0.195\textwidth]{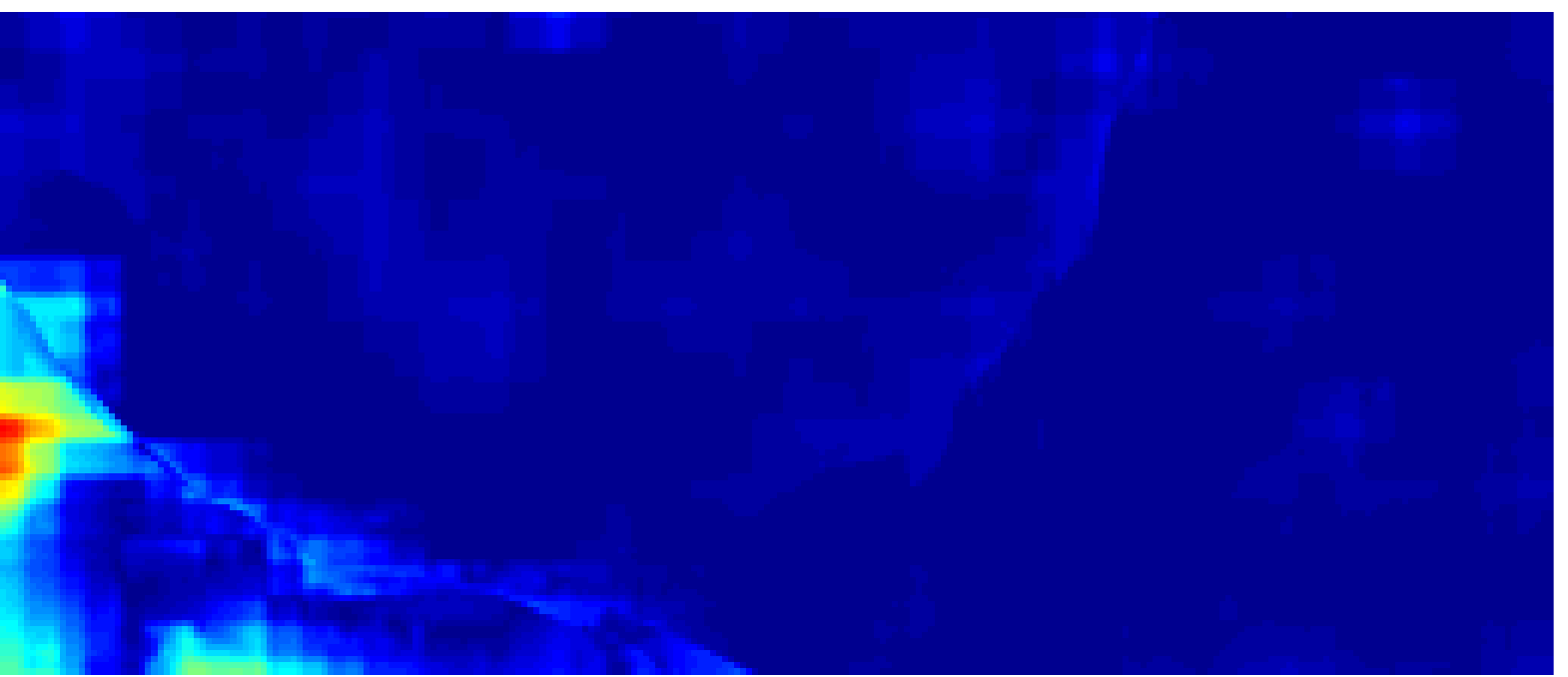}&
\includegraphics[width=0.195\textwidth]{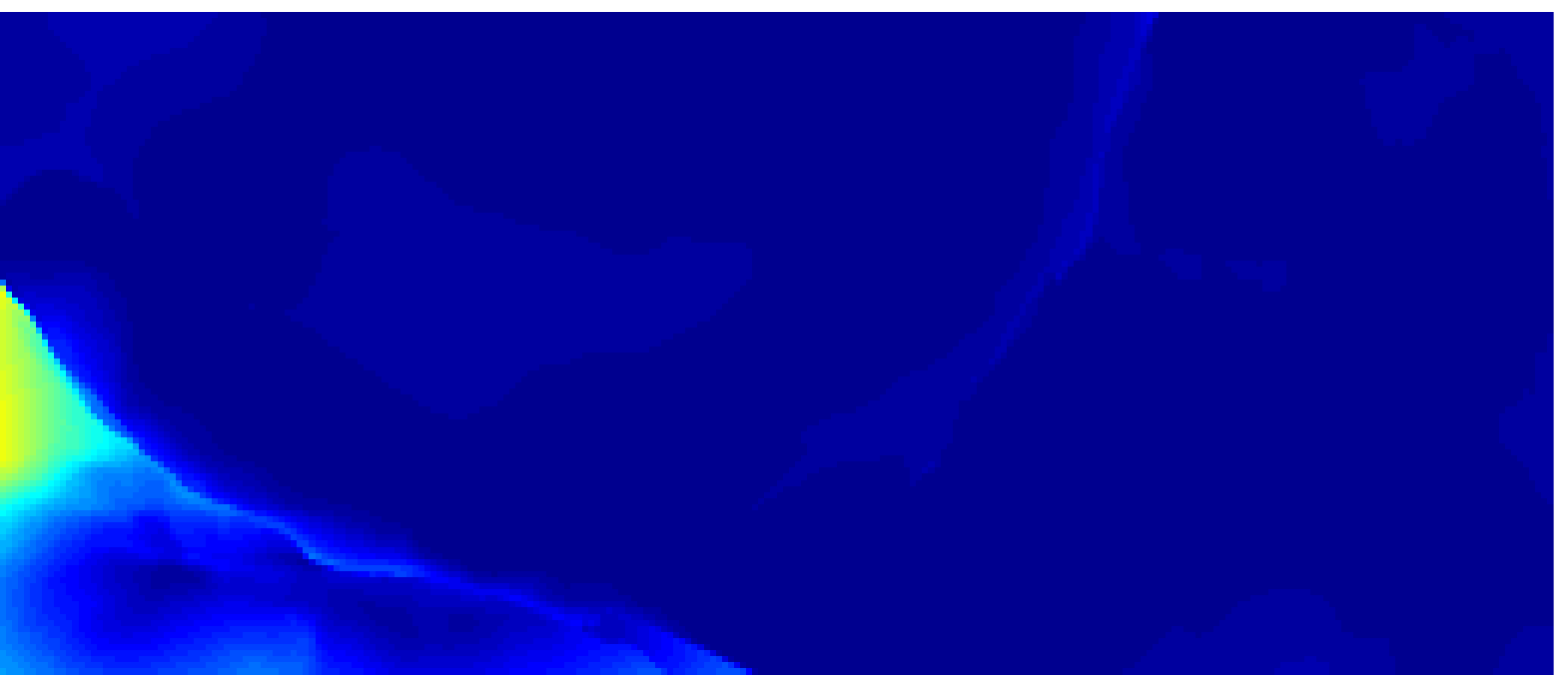}&
\includegraphics[width=0.195\textwidth]{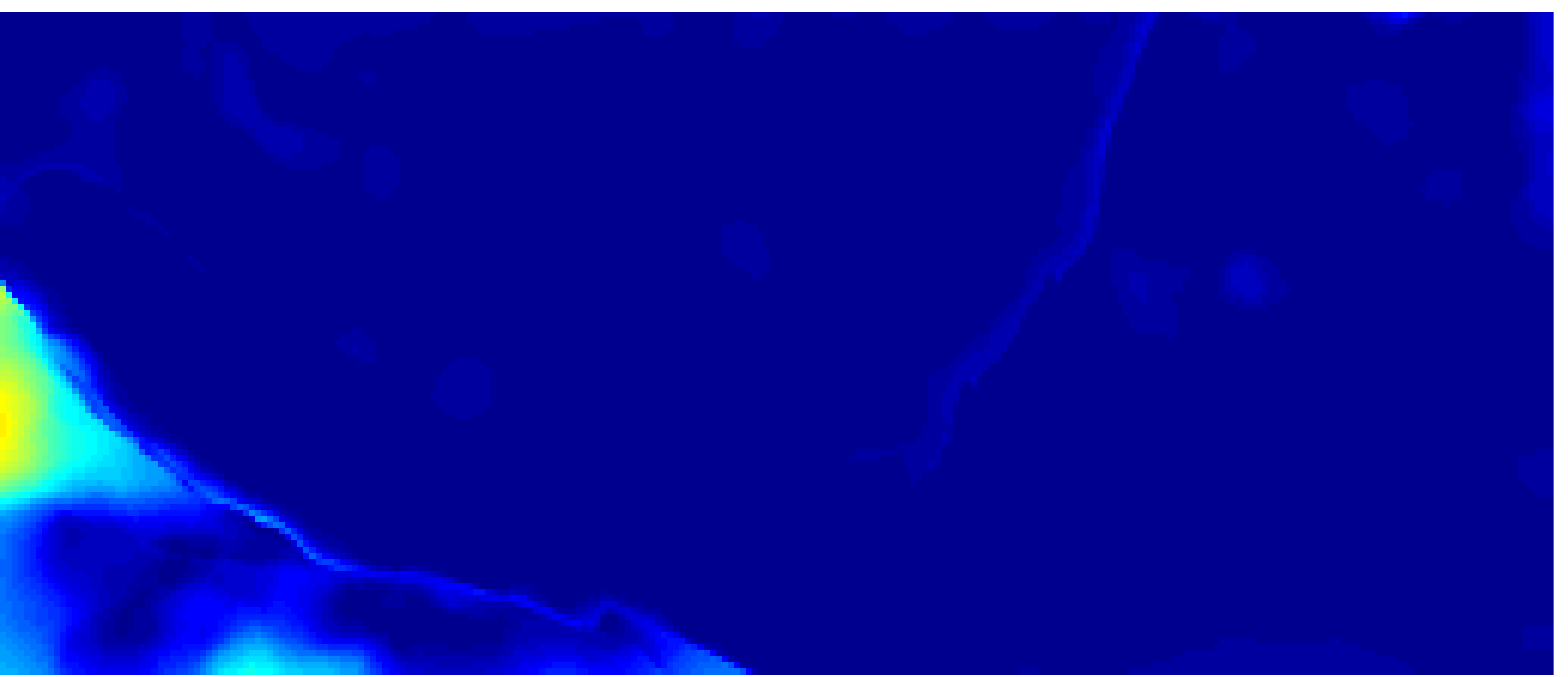}&
\includegraphics[width=0.195\textwidth]{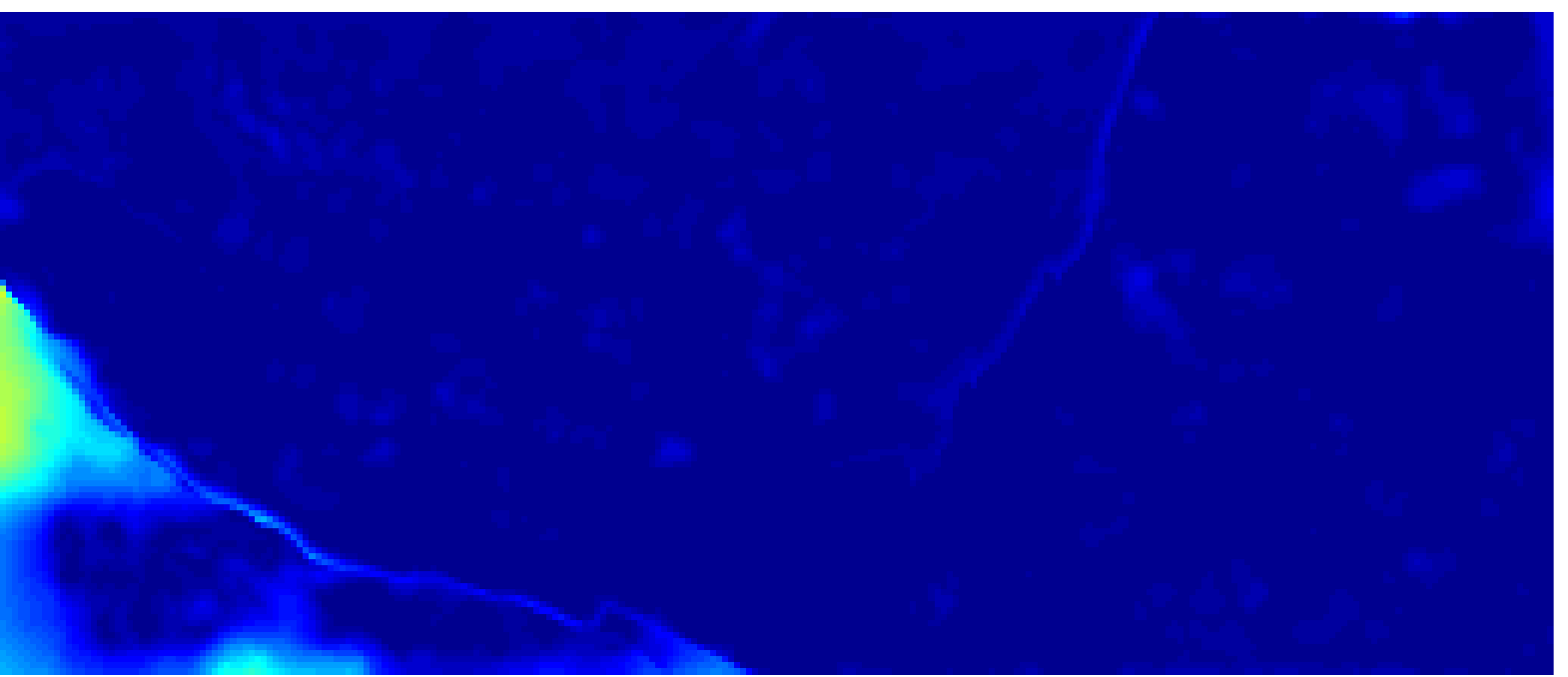}&
\includegraphics[width=0.195\textwidth]{imgs_exa/eximg-0721-10.pdf}\\[6pt]
\includegraphics[width=0.195\textwidth]{imgs_exa/eximg-0781-01.pdf}&
\includegraphics[width=0.195\textwidth]{imgs_exa/eximg-0781-02.pdf}&
\includegraphics[width=0.195\textwidth]{imgs_exa/eximg-0781-03.pdf}&
\includegraphics[width=0.195\textwidth]{imgs_exa/eximg-0781-04.pdf}&
\includegraphics[width=0.195\textwidth]{imgs_exa/eximg-0781-05.pdf}\\
\includegraphics[width=0.195\textwidth]{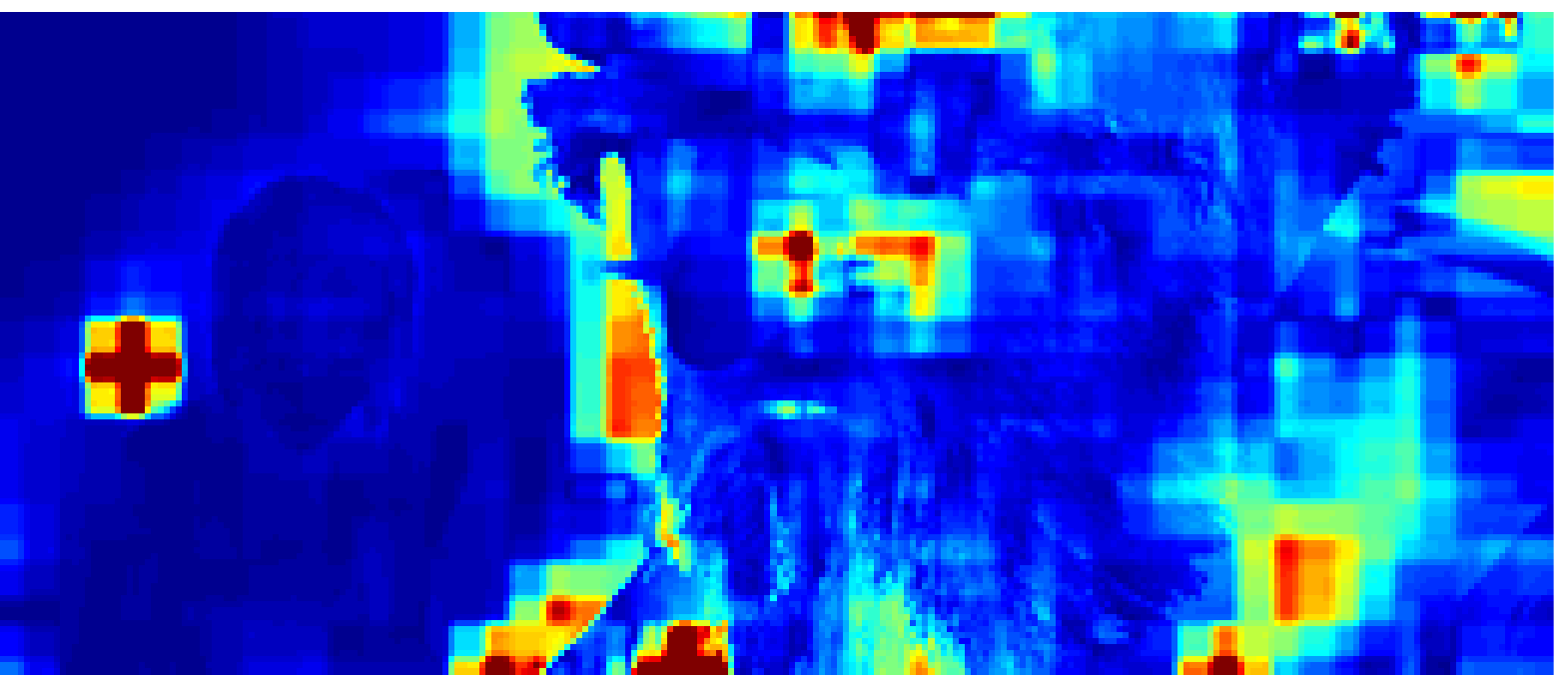}&
\includegraphics[width=0.195\textwidth]{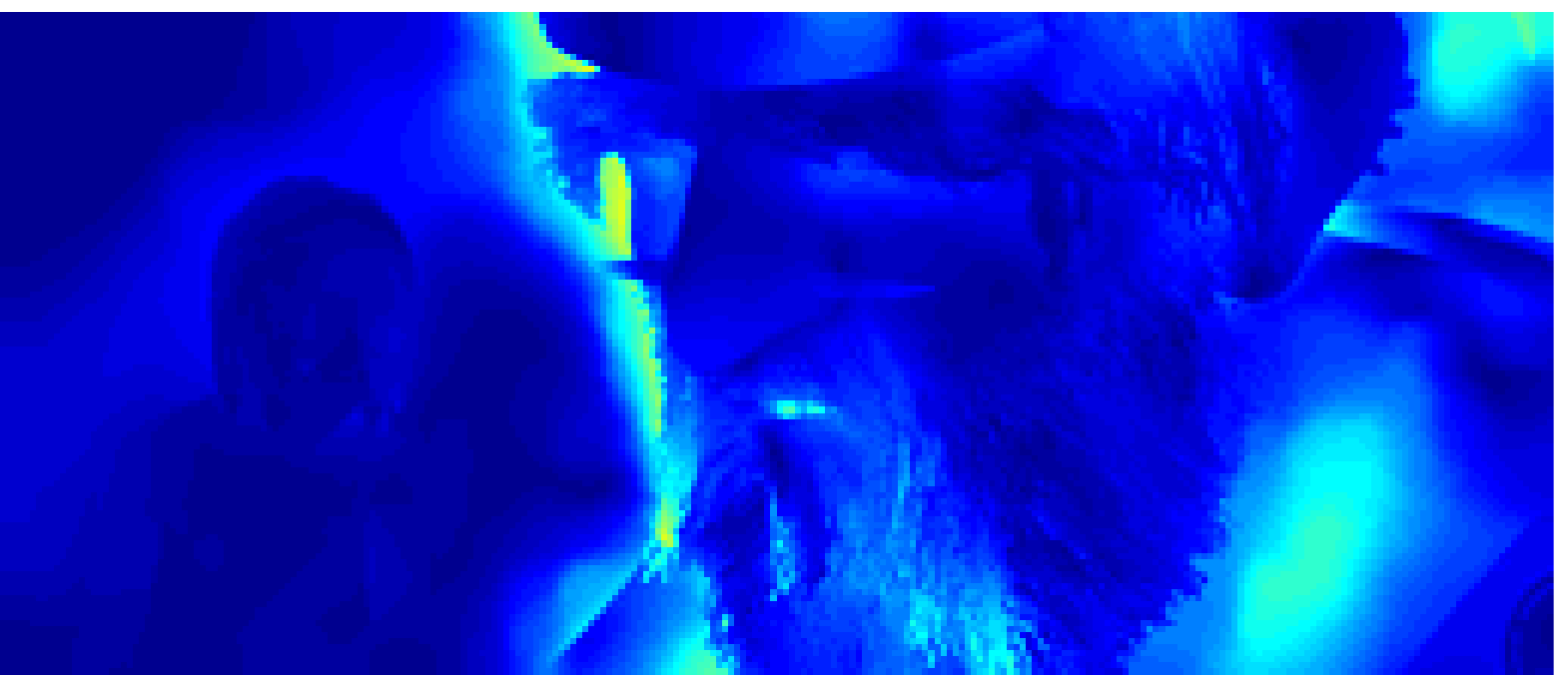}&
\includegraphics[width=0.195\textwidth]{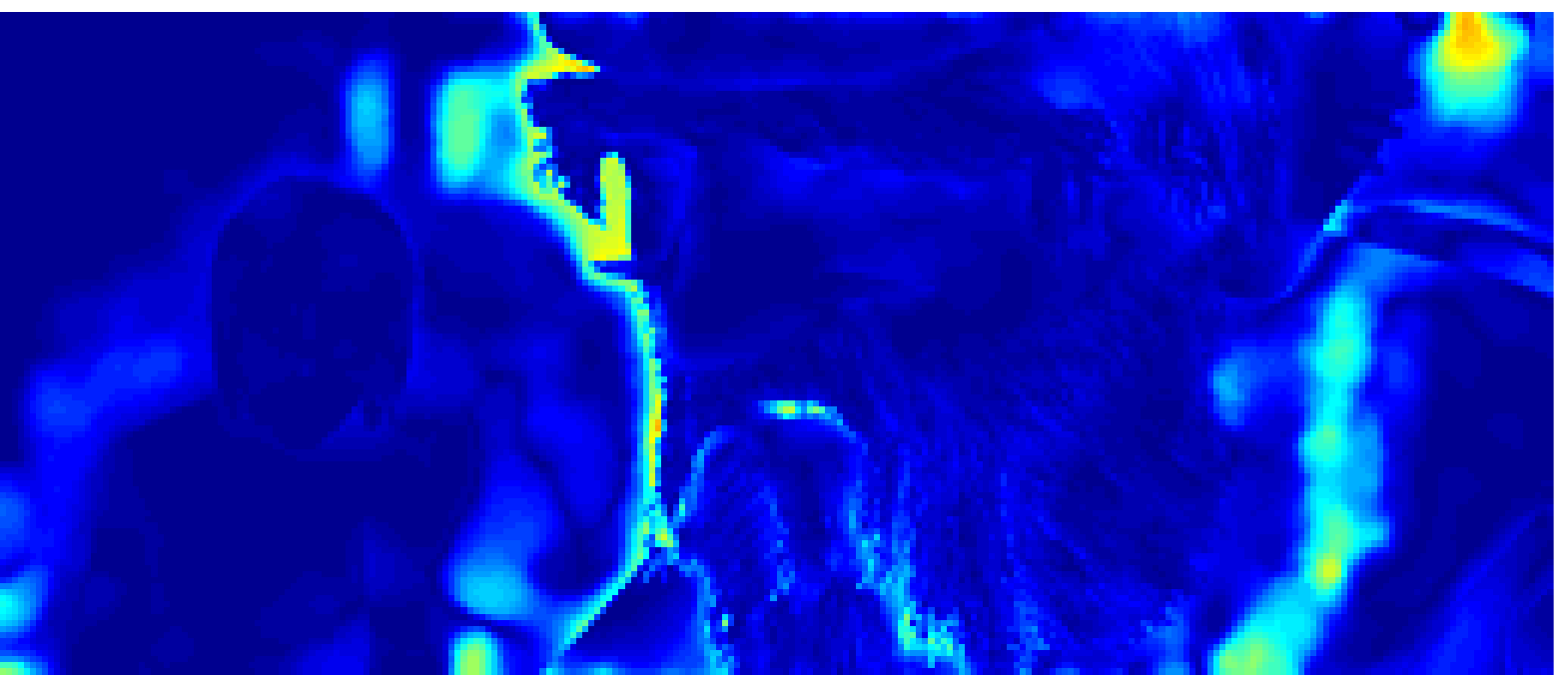}&
\includegraphics[width=0.195\textwidth]{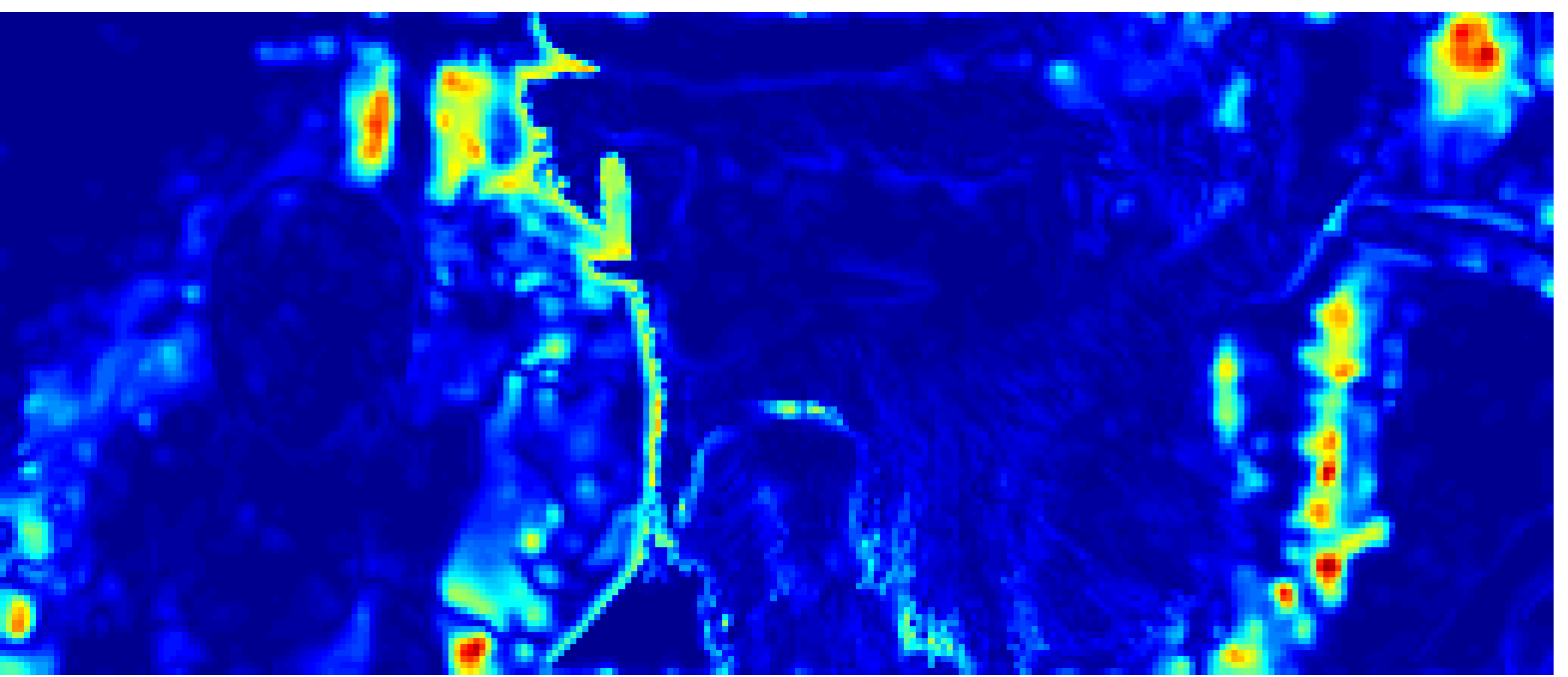}&
\includegraphics[width=0.195\textwidth]{imgs_exa/eximg-0781-10.pdf}\\[6pt]
\includegraphics[width=0.195\textwidth]{imgs_exa/eximg-0801-01.pdf}&
\includegraphics[width=0.195\textwidth]{imgs_exa/eximg-0801-02.pdf}&
\includegraphics[width=0.195\textwidth]{imgs_exa/eximg-0801-03.pdf}&
\includegraphics[width=0.195\textwidth]{imgs_exa/eximg-0801-04.pdf}&
\includegraphics[width=0.195\textwidth]{imgs_exa/eximg-0801-05.pdf}\\
\includegraphics[width=0.195\textwidth]{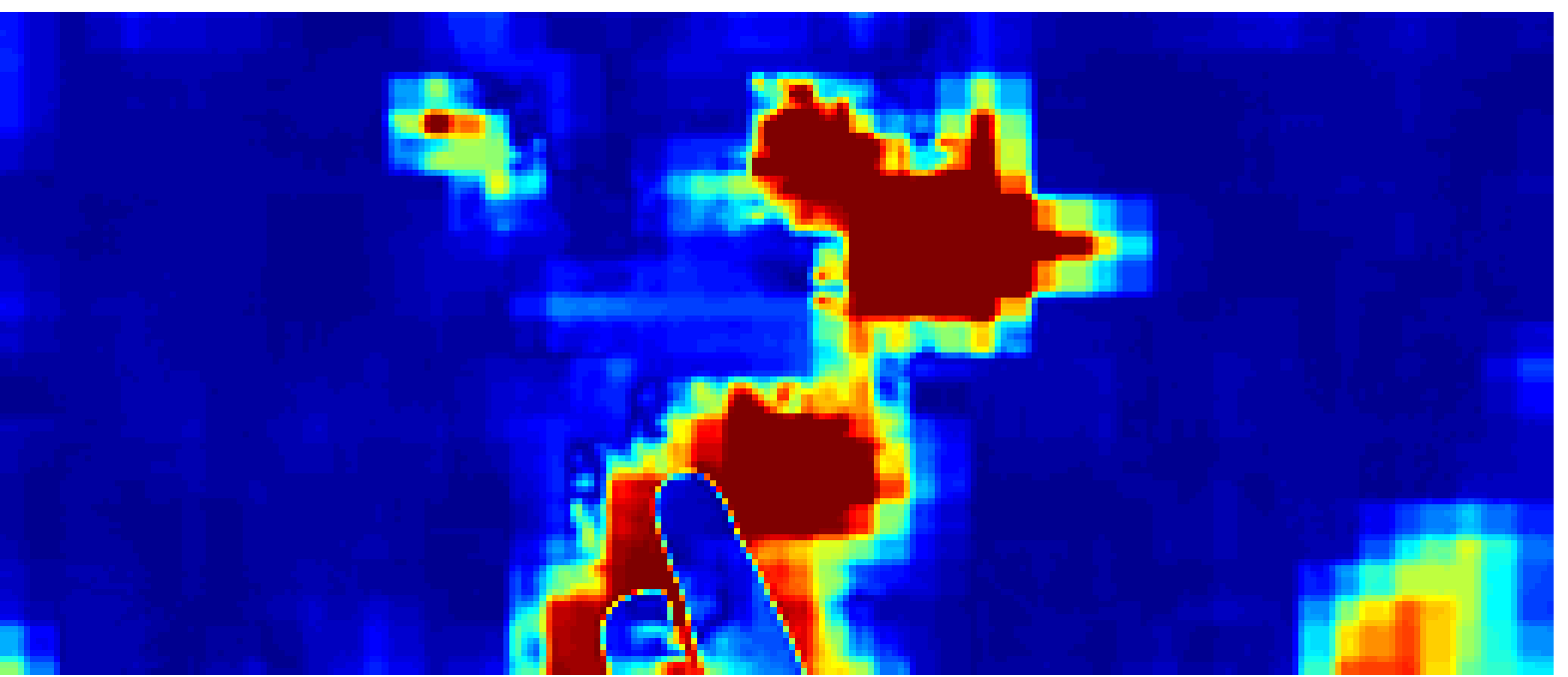}&
\includegraphics[width=0.195\textwidth]{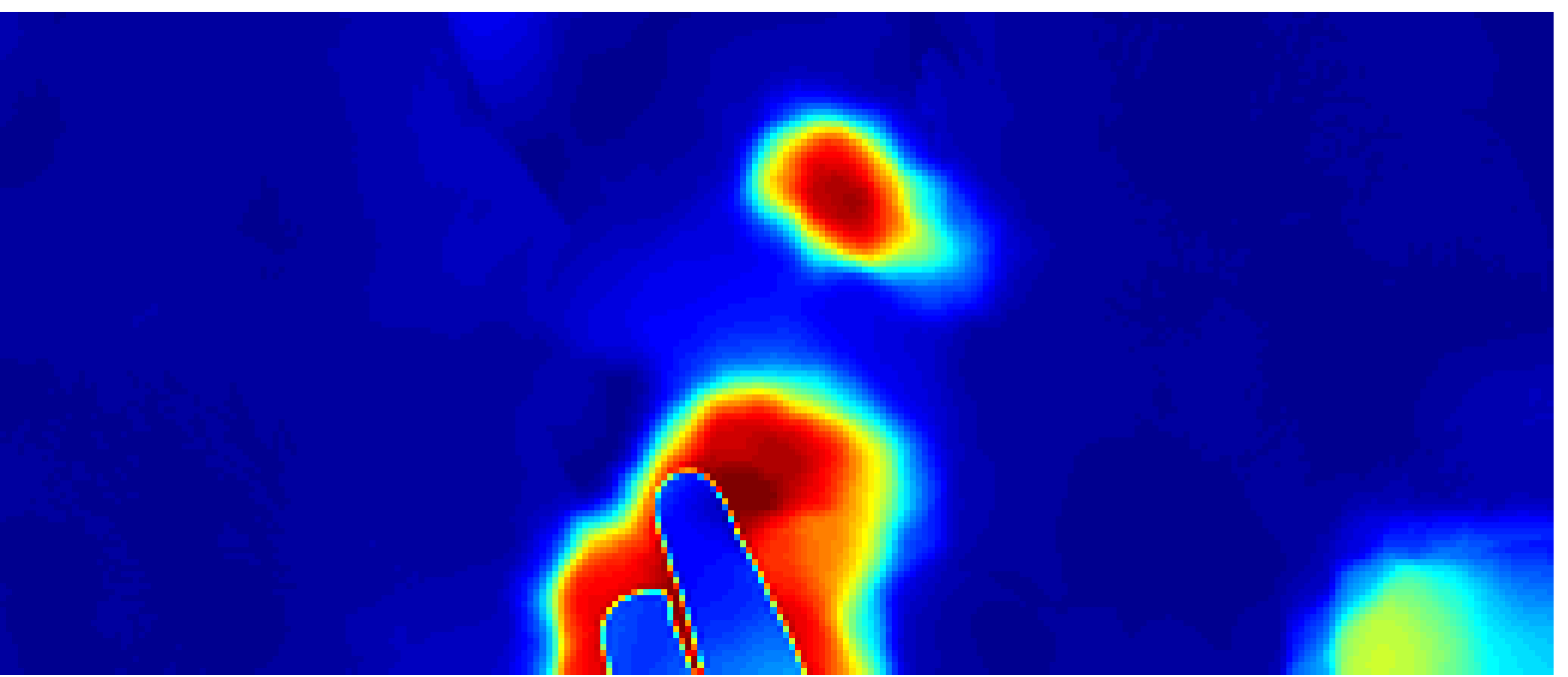}&
\includegraphics[width=0.195\textwidth]{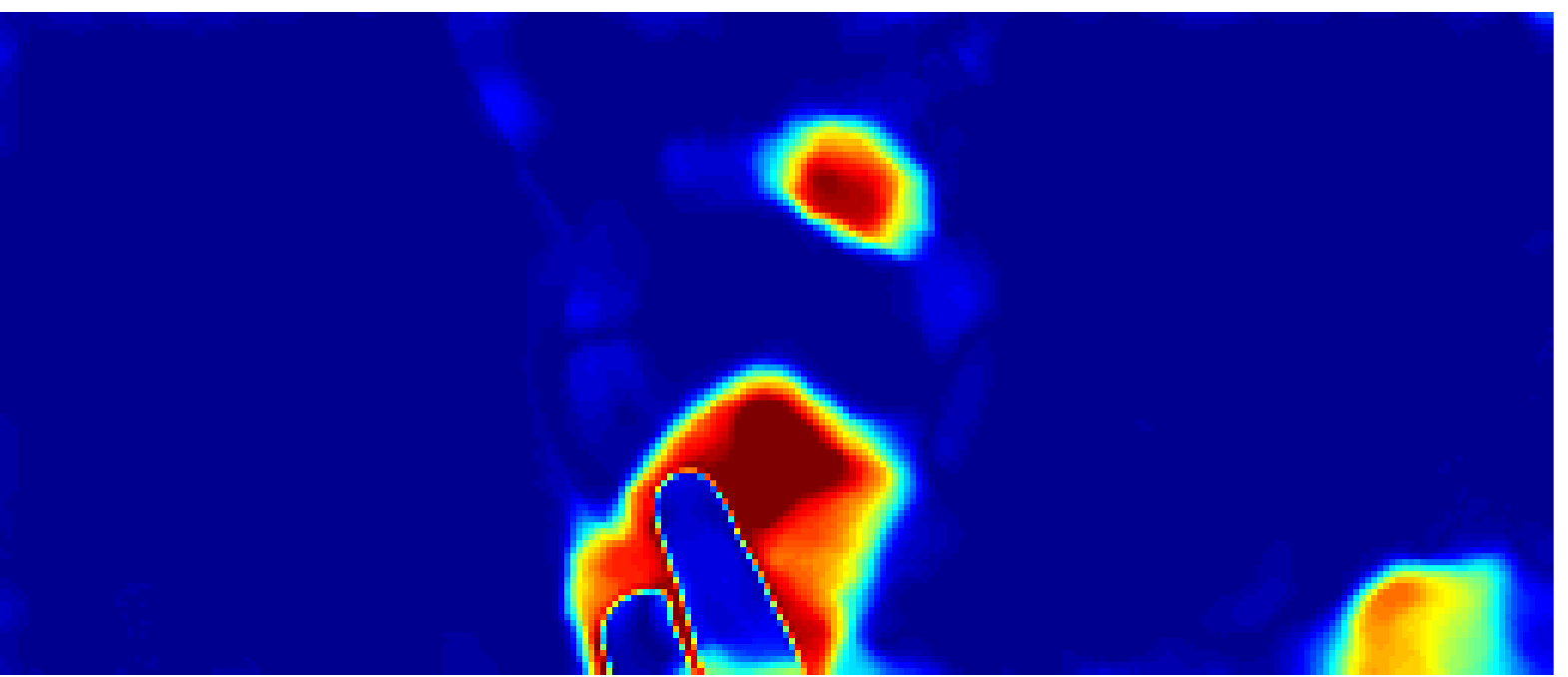}&
\includegraphics[width=0.195\textwidth]{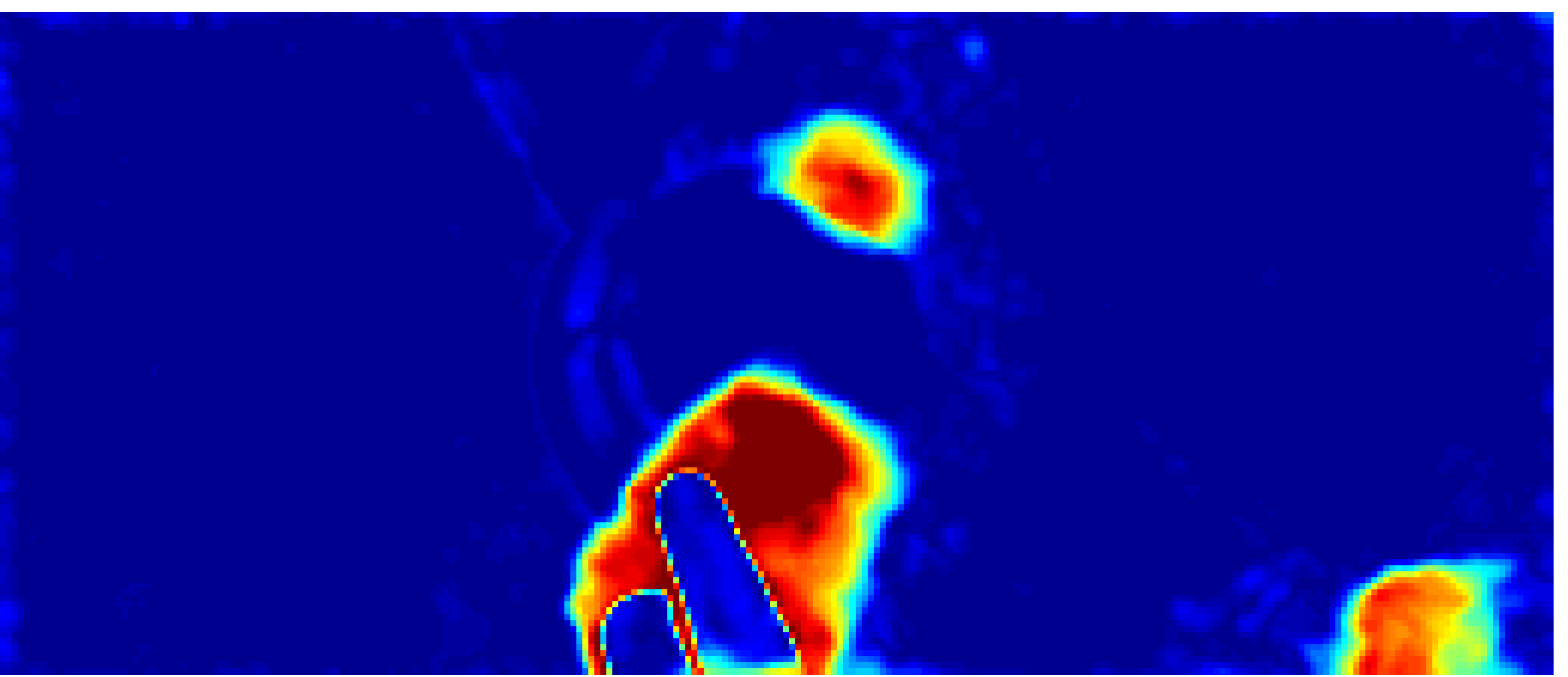}&
\includegraphics[width=0.195\textwidth]{imgs_exa/eximg-0801-10.pdf}\\[6pt]
\includegraphics[width=0.195\textwidth]{imgs_exa/eximg-1001-01.pdf}&
\includegraphics[width=0.195\textwidth]{imgs_exa/eximg-1001-02.pdf}&
\includegraphics[width=0.195\textwidth]{imgs_exa/eximg-1001-03.pdf}&
\includegraphics[width=0.195\textwidth]{imgs_exa/eximg-1001-04.pdf}&
\includegraphics[width=0.195\textwidth]{imgs_exa/eximg-1001-05.pdf}\\
\includegraphics[width=0.195\textwidth]{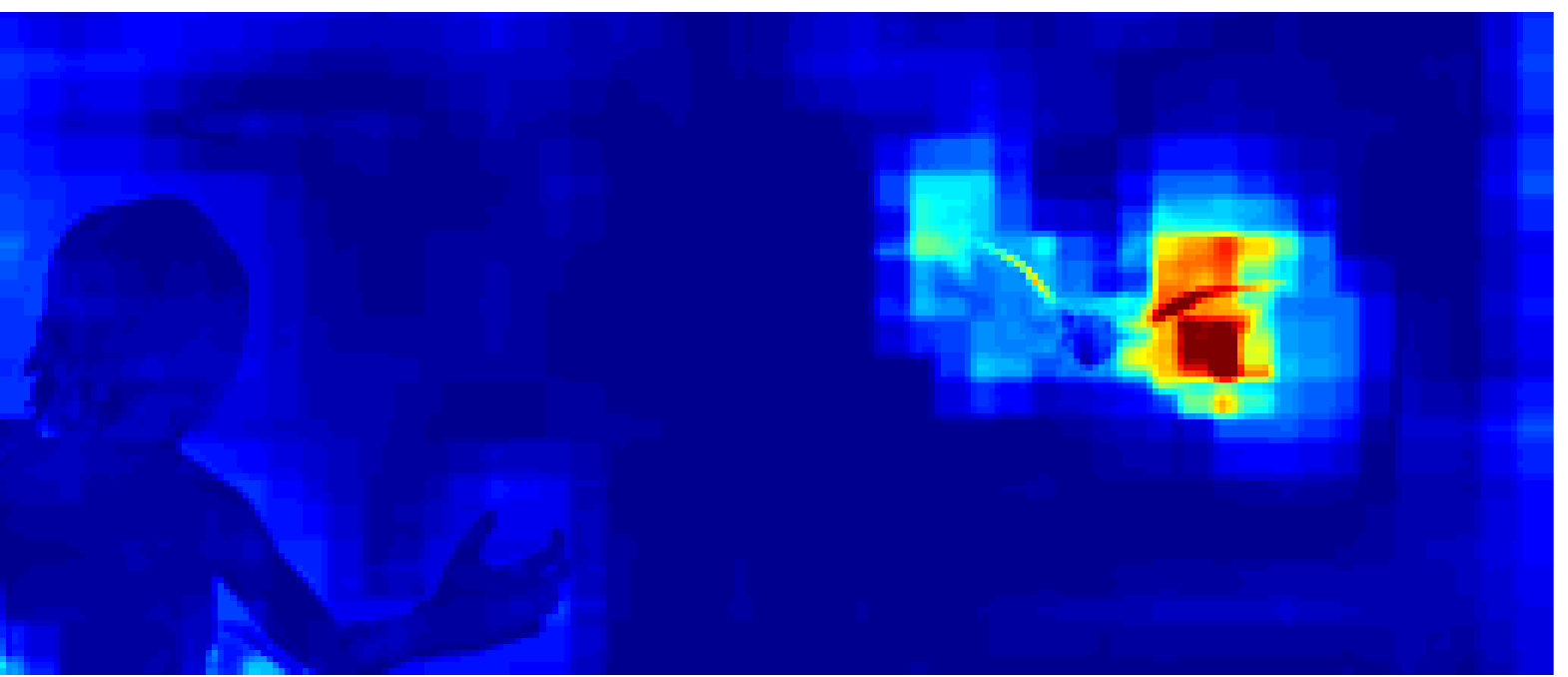}&
\includegraphics[width=0.195\textwidth]{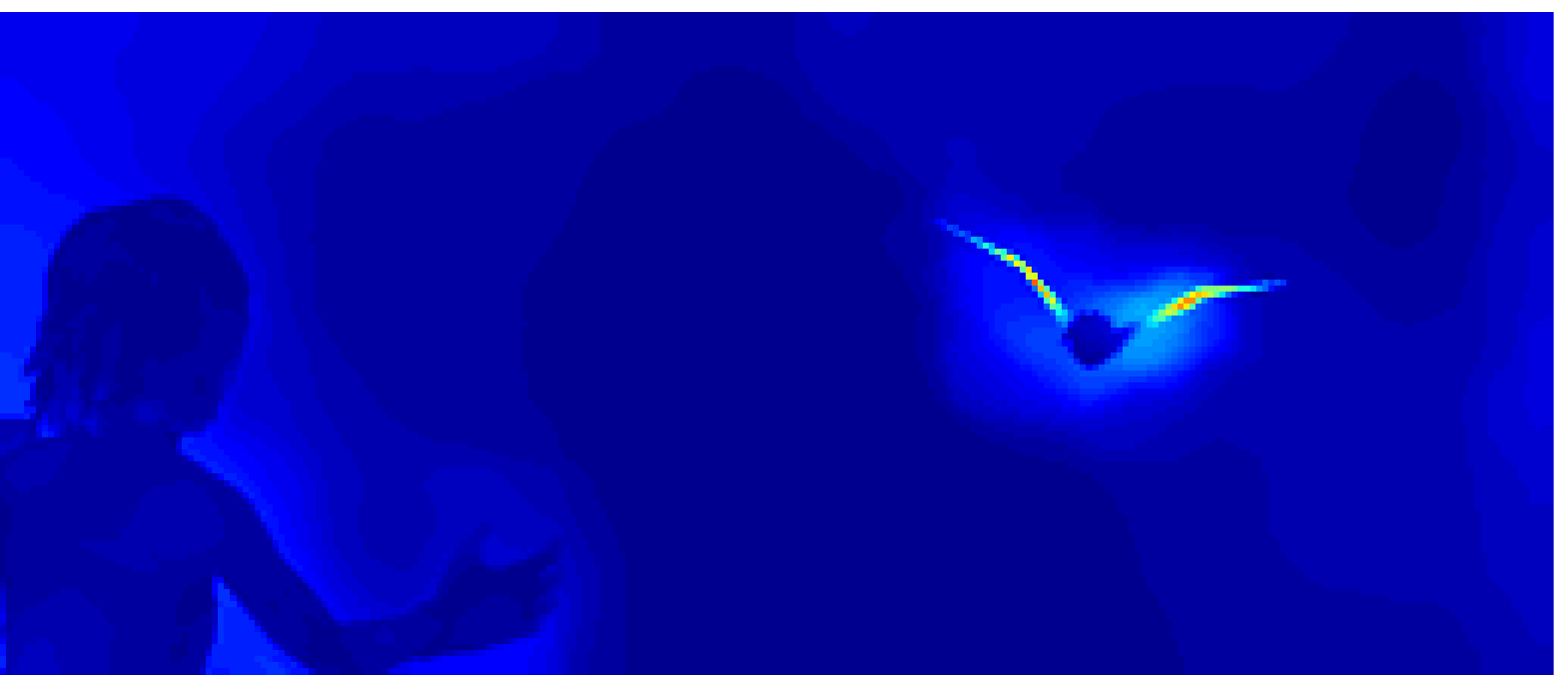}&
\includegraphics[width=0.195\textwidth]{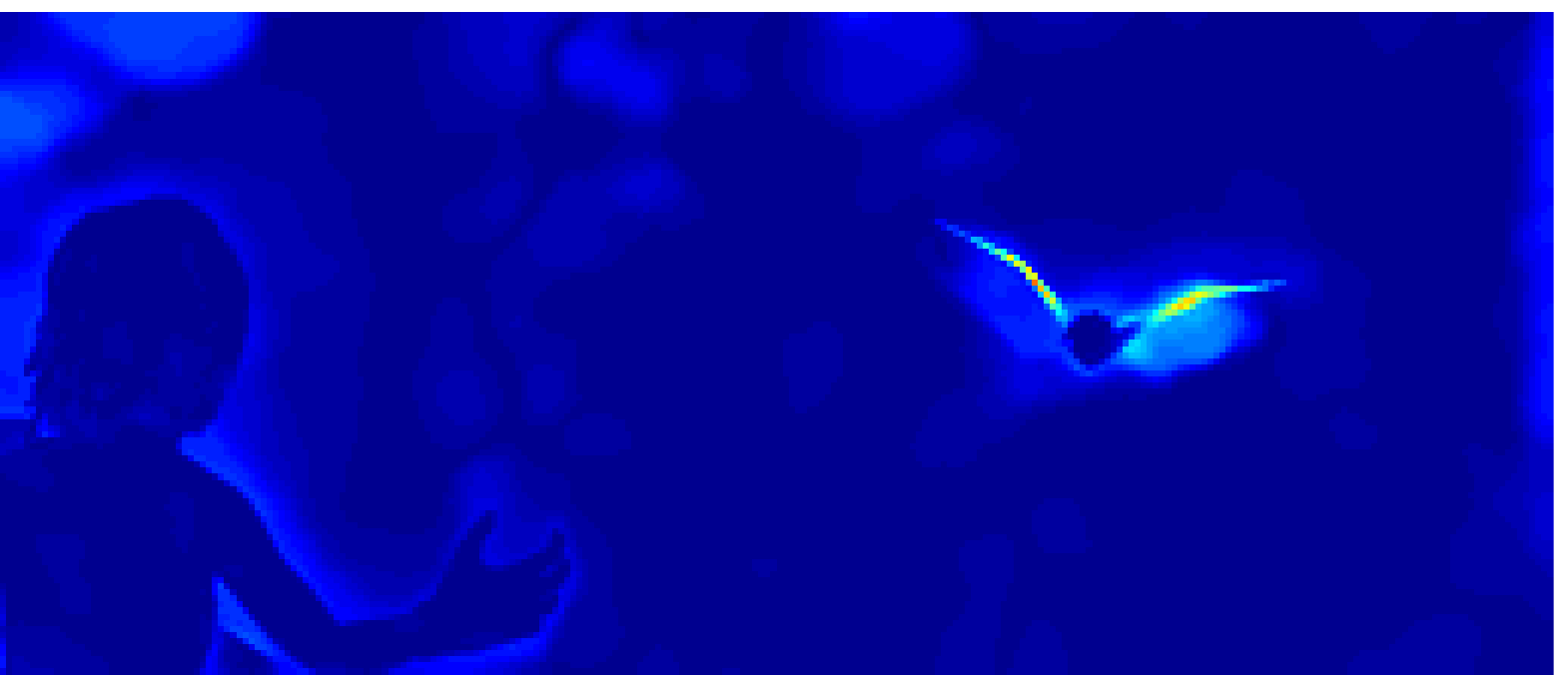}&
\includegraphics[width=0.195\textwidth]{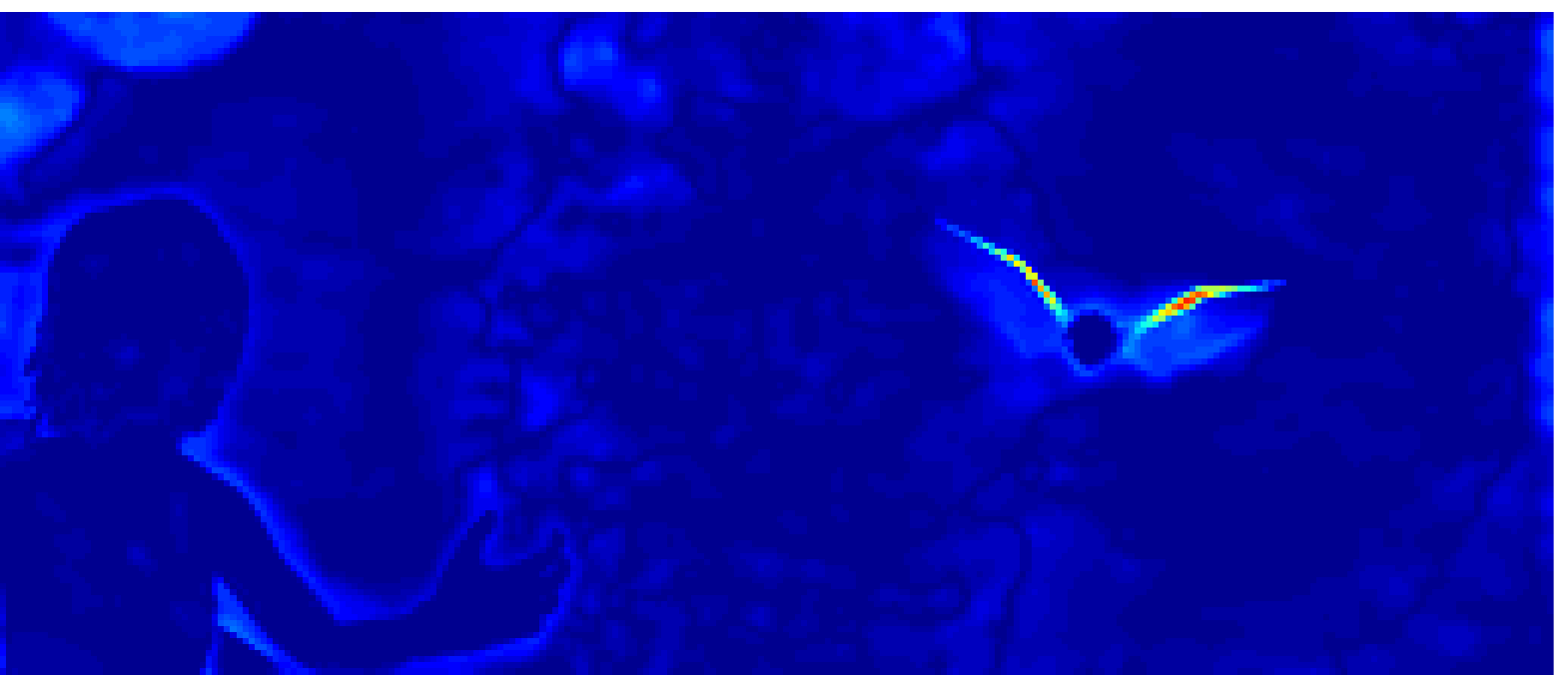}&
\includegraphics[width=0.195\textwidth]{imgs_exa/eximg-1001-10.pdf}\\[6pt]
\includegraphics[width=0.195\textwidth]{imgs_exa/eximg-1021-01.pdf}&
\includegraphics[width=0.195\textwidth]{imgs_exa/eximg-1021-02.pdf}&
\includegraphics[width=0.195\textwidth]{imgs_exa/eximg-1021-03.pdf}&
\includegraphics[width=0.195\textwidth]{imgs_exa/eximg-1021-04.pdf}&
\includegraphics[width=0.195\textwidth]{imgs_exa/eximg-1021-05.pdf}\\
\includegraphics[width=0.195\textwidth]{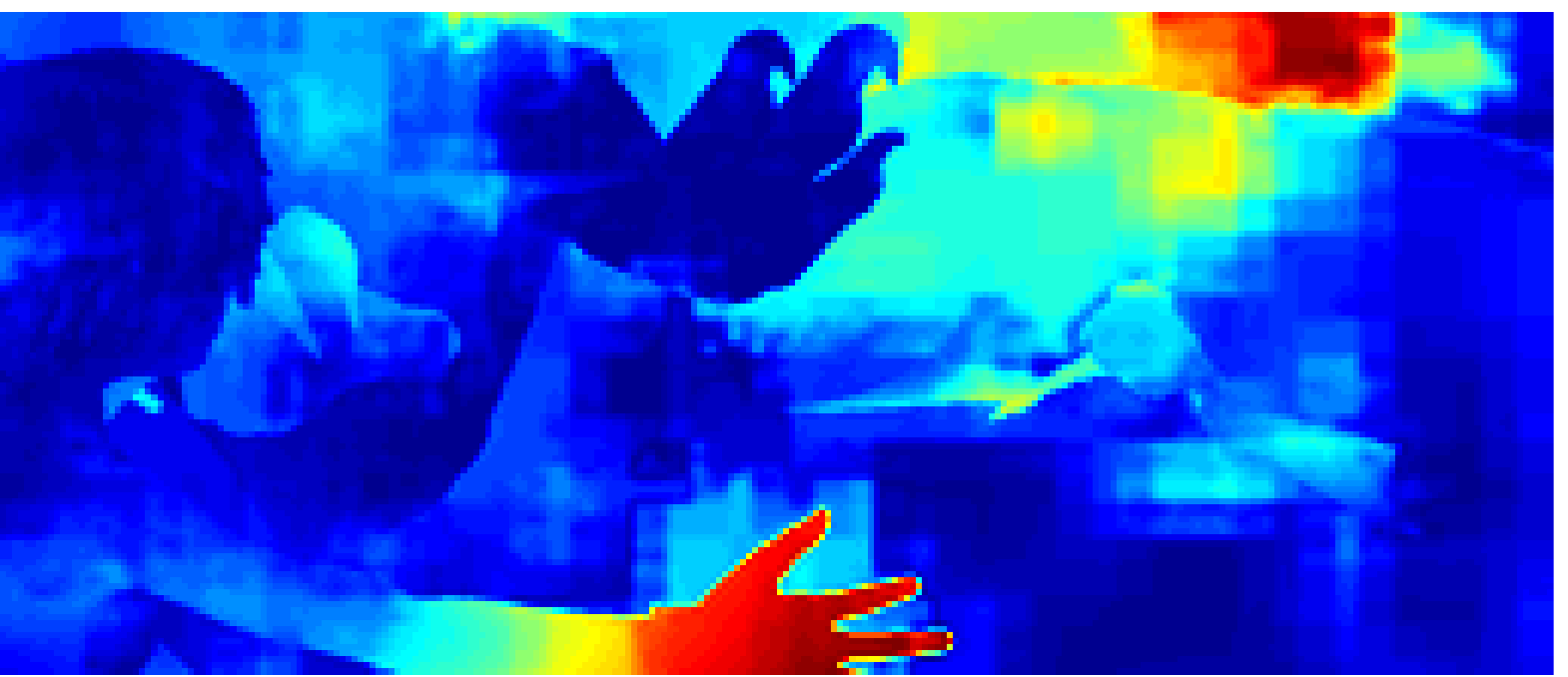}&
\includegraphics[width=0.195\textwidth]{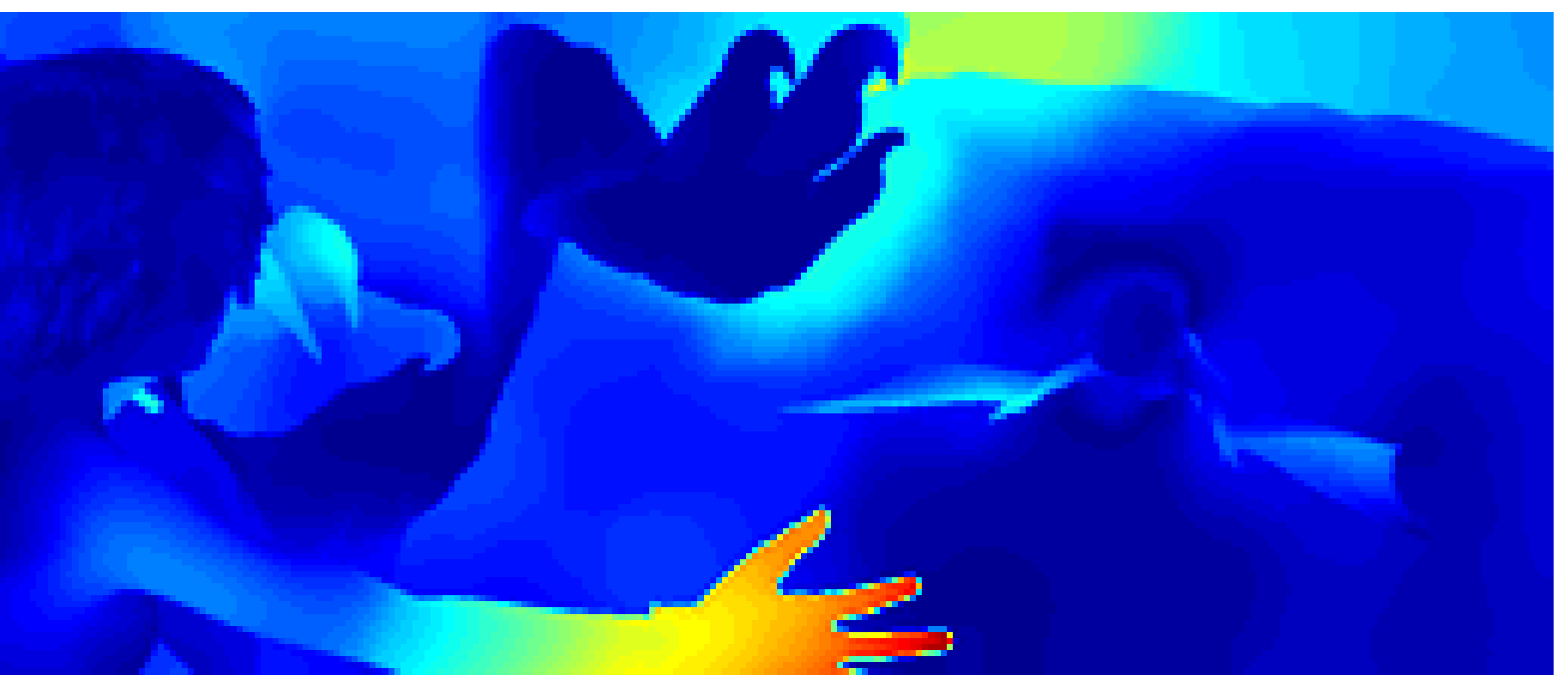}&
\includegraphics[width=0.195\textwidth]{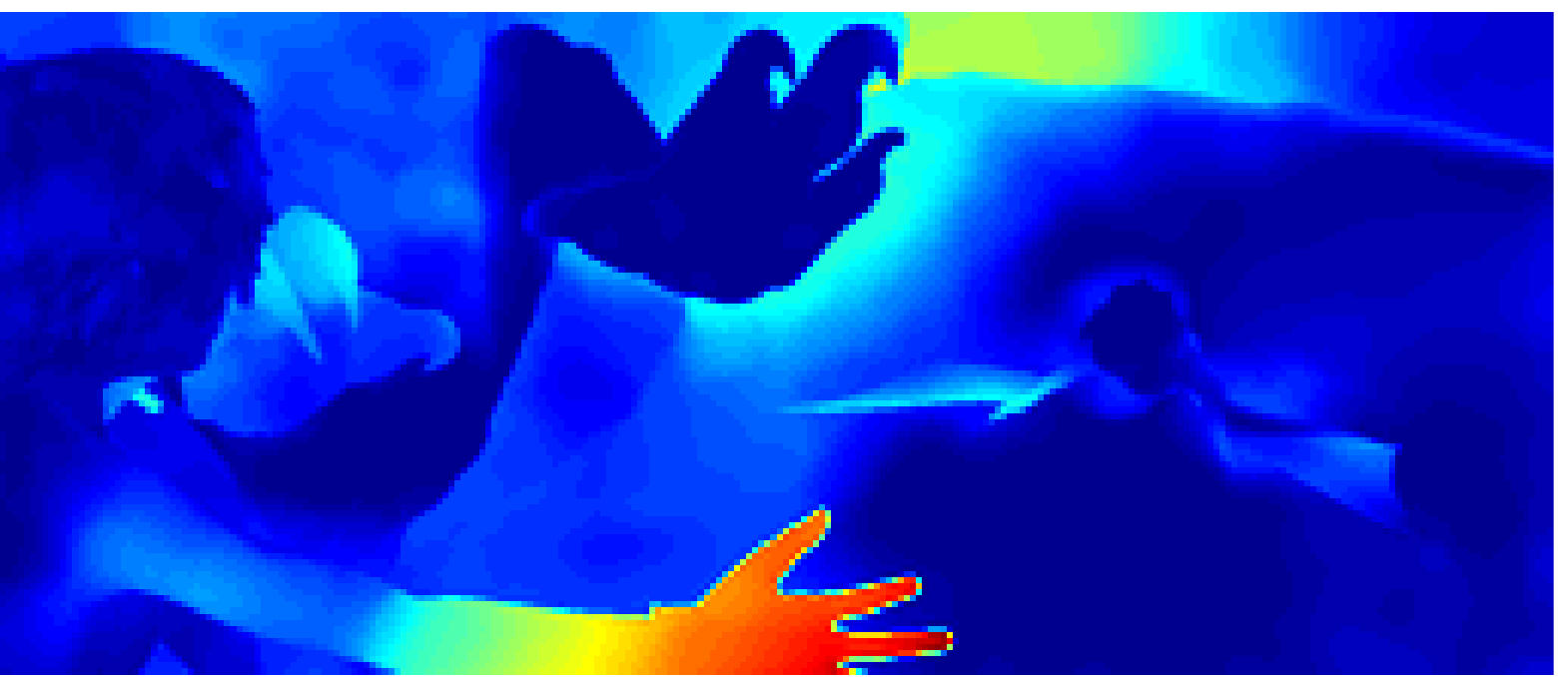}&
\includegraphics[width=0.195\textwidth]{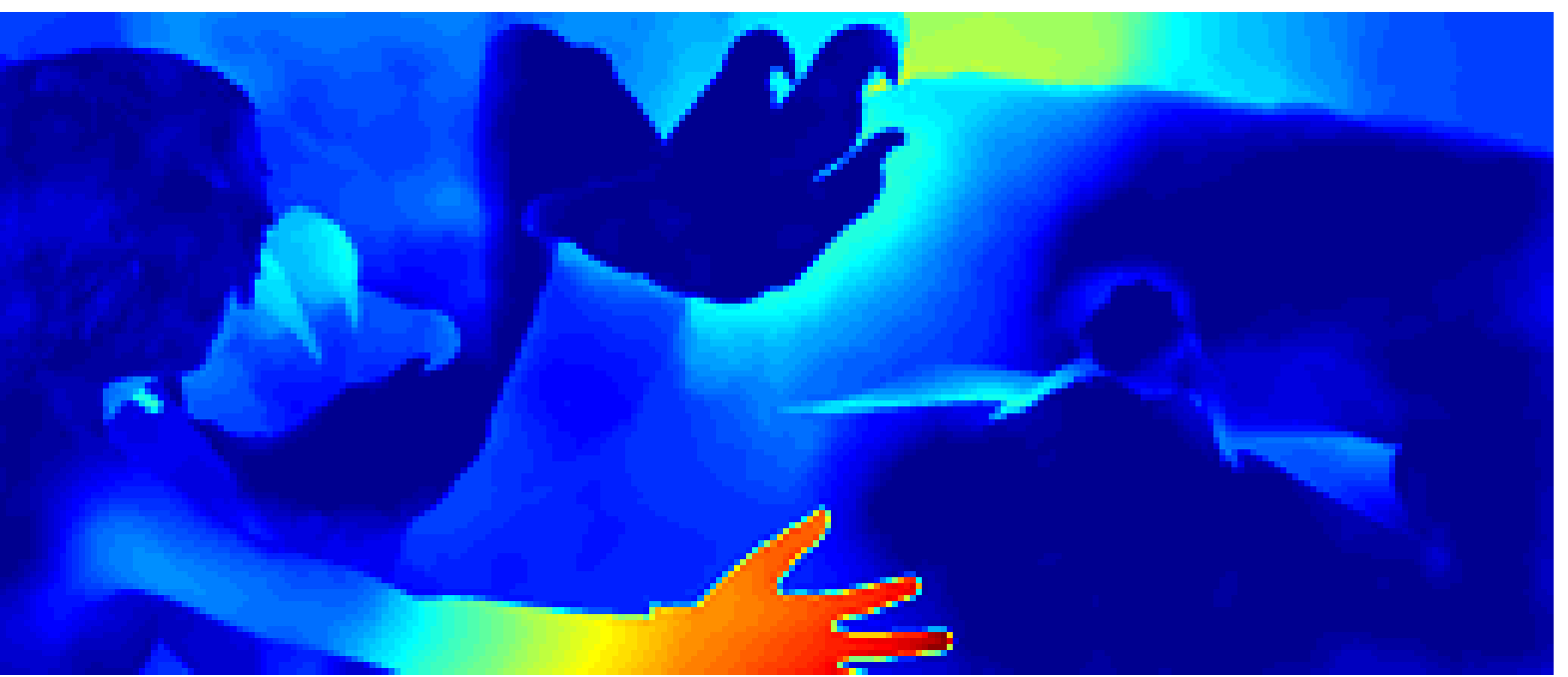}&
\includegraphics[width=0.195\textwidth]{imgs_exa/eximg-1021-10.pdf}\\[6pt]
\end{tabular}
}
\caption{Exemplary results on Sintel (training) and error maps. In each block of $2 \times 6$ images.  Top row, left to right: Our method for operating points ({\bf 1})-({\bf 4}), Ground Truth. Bottom row: Error heat maps scaled from blue (no error) to red (maximum ground truth flow magnitude), Original Image.}\label{fig:sintel2res_errmap_AP} 
\end{figure*}

Optical flow error maps on the training subset of the Sintel~\cite{Butler-ECCV-2012} are shown in Fig.~\ref{fig:sintel1res_errmap_AP} and Fig.~\ref{fig:sintel2res_errmap_AP}. More error maps on the test sets of Sintel and KITTI can be found online through their test websites. Typical error modes of our method are observable at motion discontinuities (Fig.~\ref{fig:sintel1res_errmap_AP}, first block), large displacements (Fig.~\ref{fig:sintel1res_errmap_AP}, third block) and at frame boundaries for fast motions (e.g. Fig.~\ref{fig:sintel2res_errmap_AP}, second block)

\section{More exemplary results for the high frame-rate experiment on Sintel-training in \S~3.5}
\begin{figure*}[h!]
\large
\centering\setlength{\tabcolsep}{0.1pt}\renewcommand{\arraystretch}{0} 
        {
        \begin{tabular}{ccccc}
        &{\bf Ground Truth} & {\bf DeepFlow} & {\bf DIS (all frames)} & {\bf difference image}\\
        \parbox[b]{3mm}{\rotatebox[origin=l]{90}{1 frame}}&
        \includegraphics[width=0.25\textwidth]{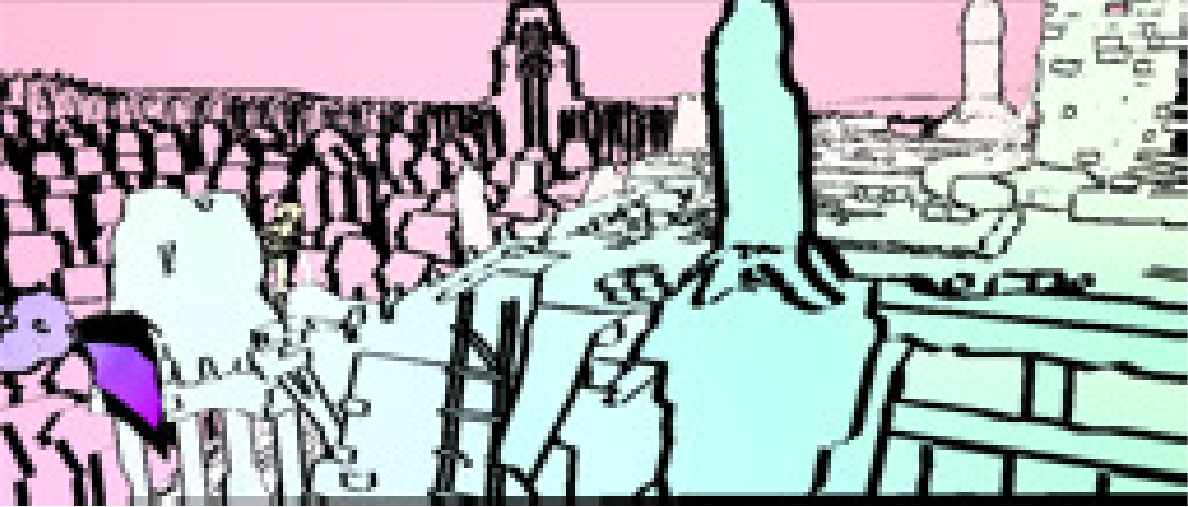}&
        \includegraphics[width=0.25\textwidth]{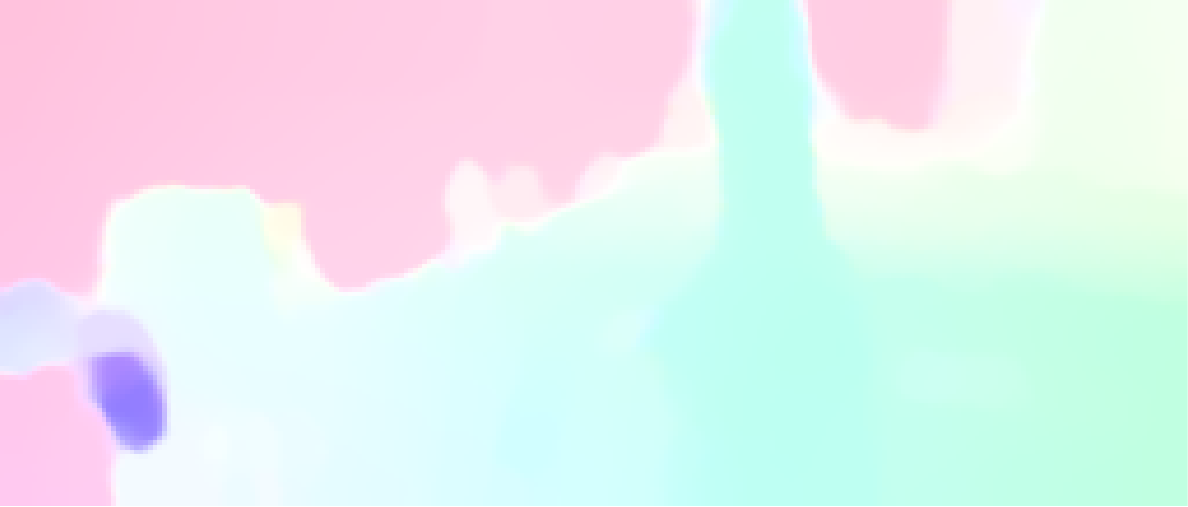}&
        \includegraphics[width=0.25\textwidth]{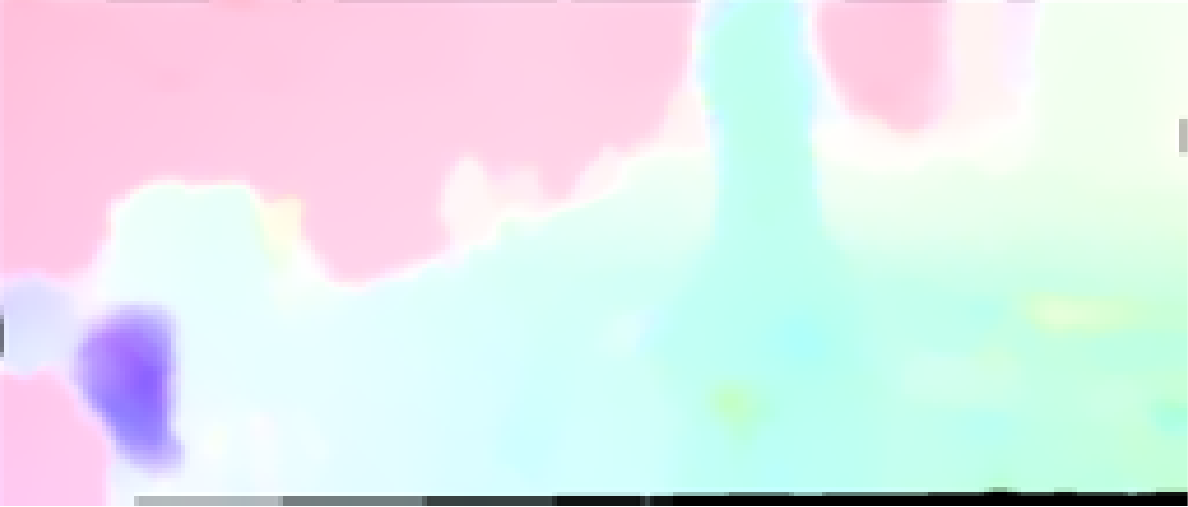}&
        \includegraphics[width=0.25\textwidth]{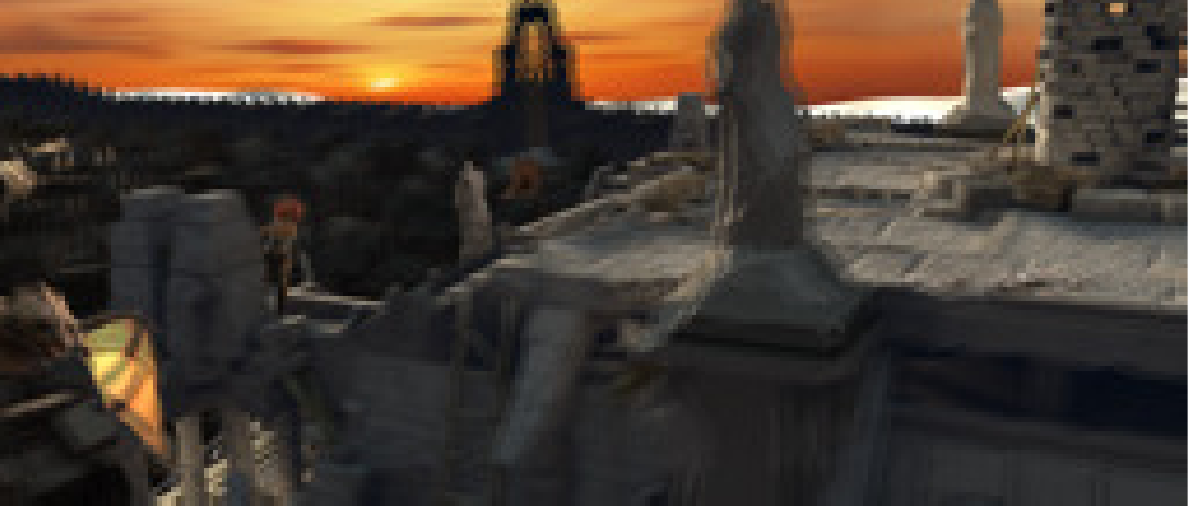} \\   
        \parbox[b]{3mm}{\rotatebox[origin=l]{90}{2 frames}}&
        \includegraphics[width=0.25\textwidth]{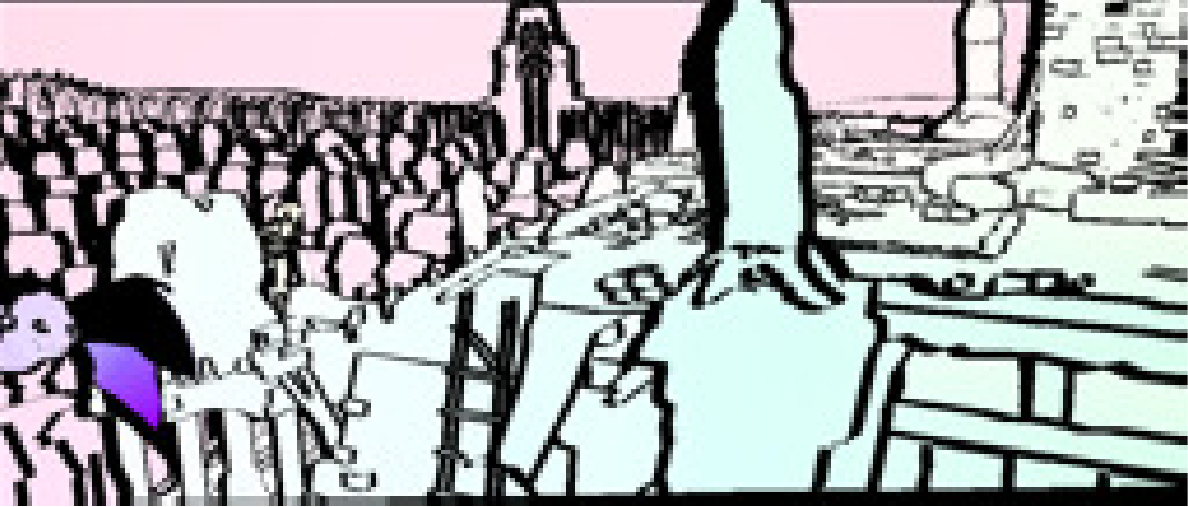}&
        \includegraphics[width=0.25\textwidth]{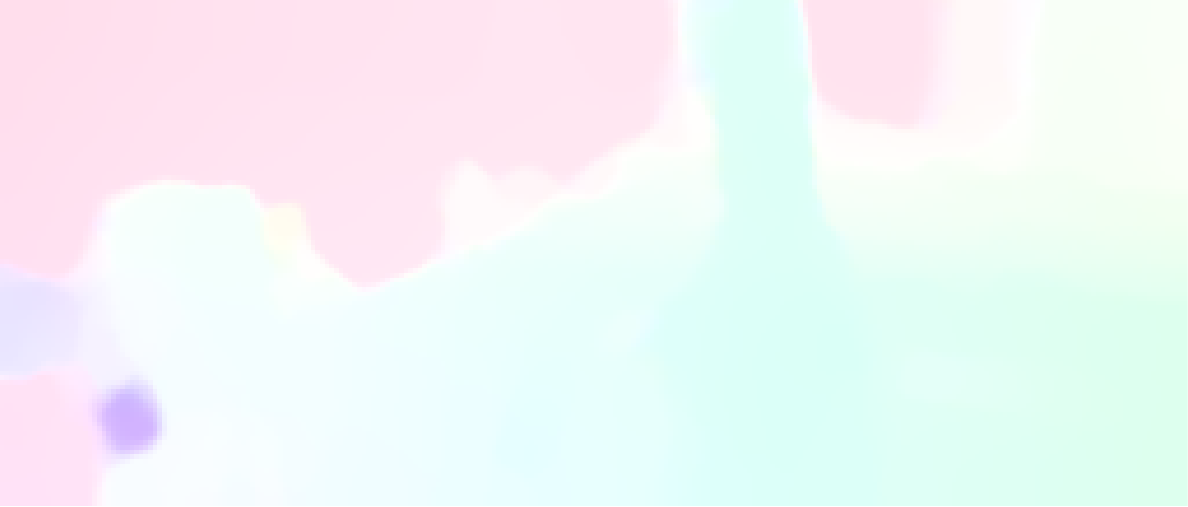}&
        \includegraphics[width=0.25\textwidth]{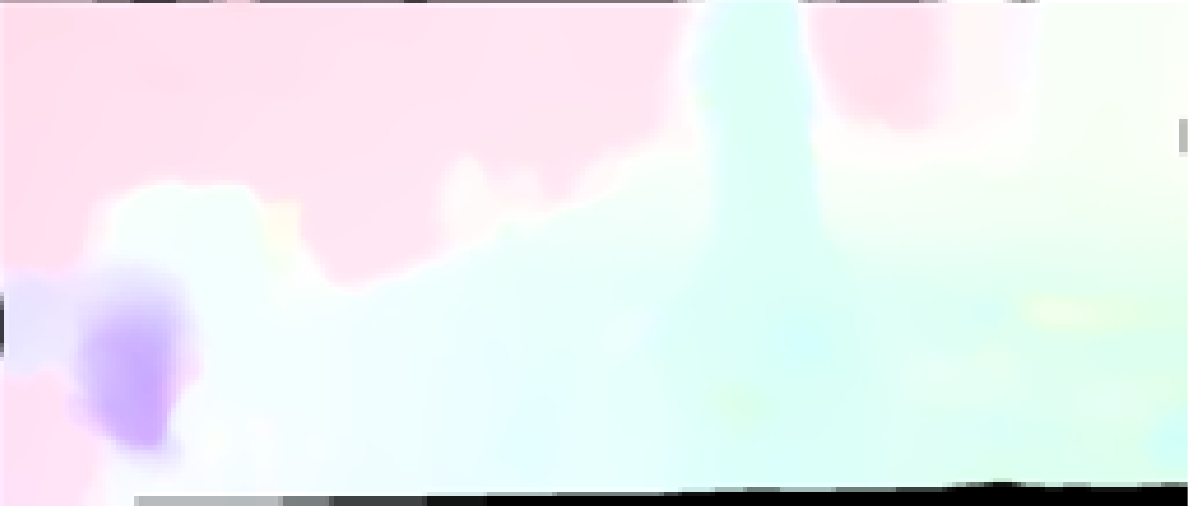}&
        \includegraphics[width=0.25\textwidth]{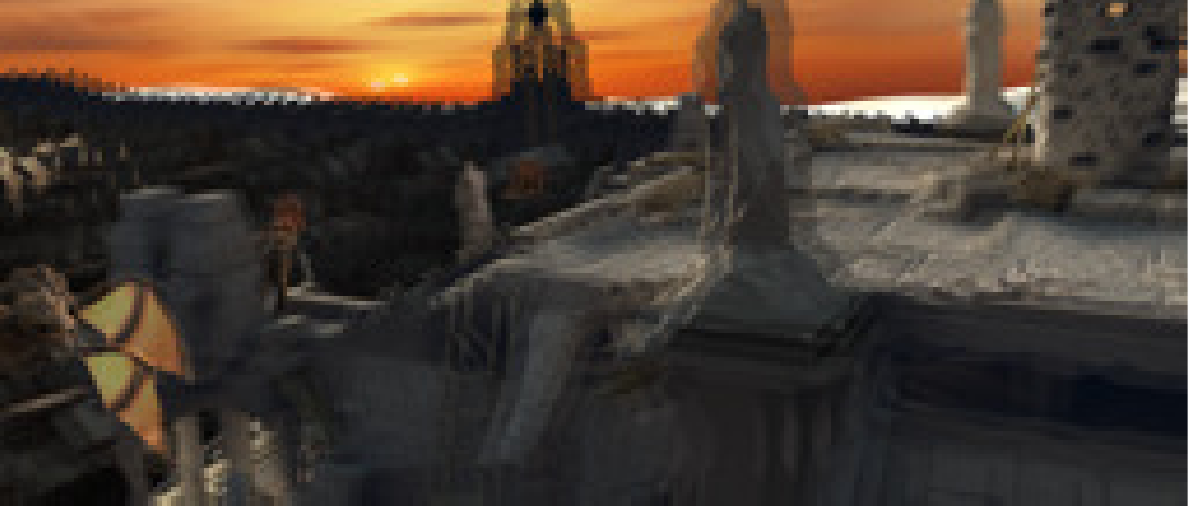} \\        
	\parbox[b]{3mm}{\rotatebox[origin=l]{90}{4 frames}}&        
        \includegraphics[width=0.25\textwidth]{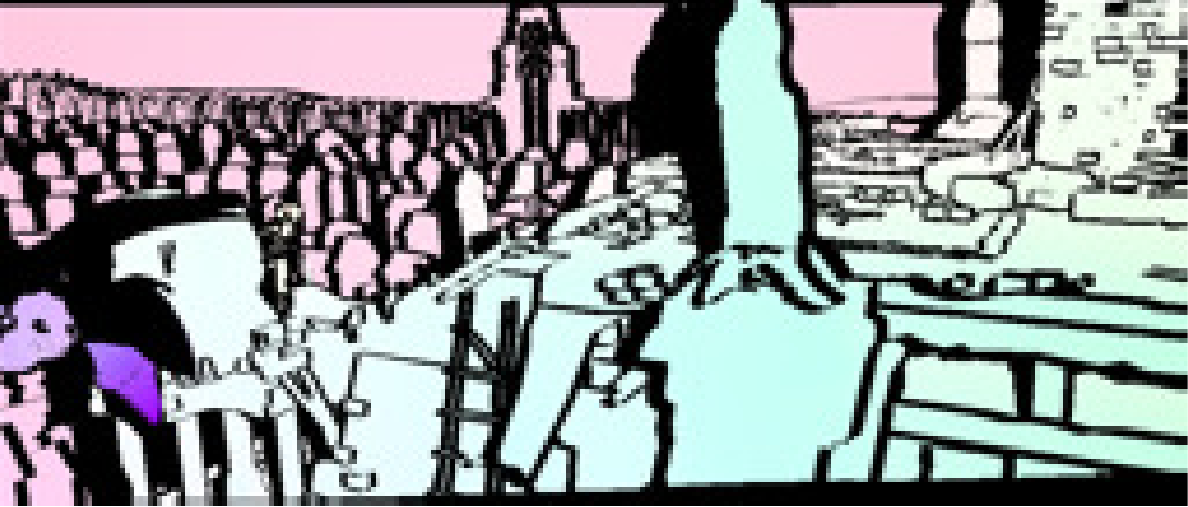}&
        \includegraphics[width=0.25\textwidth]{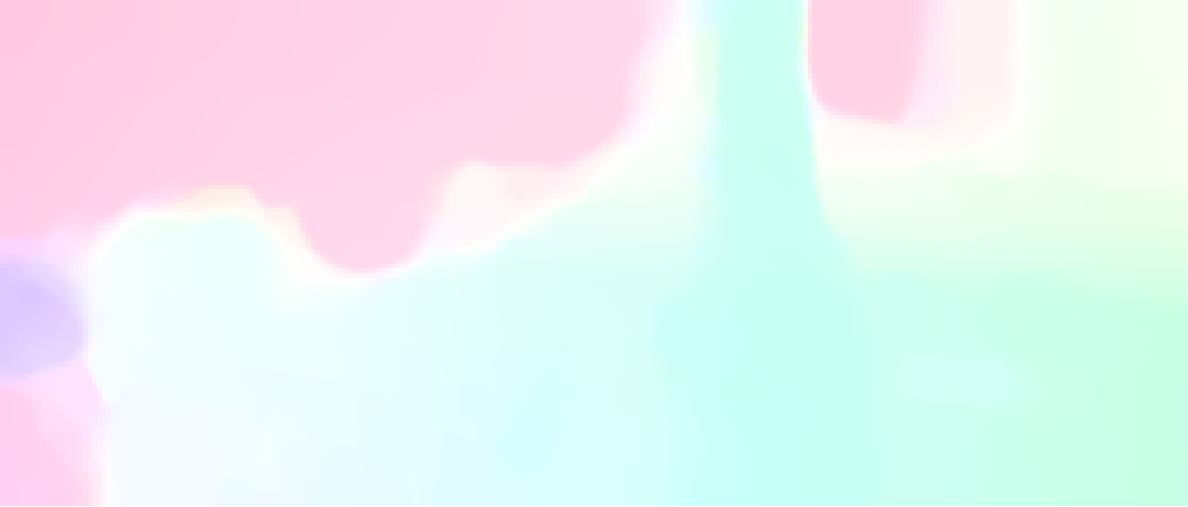}&
        \includegraphics[width=0.25\textwidth]{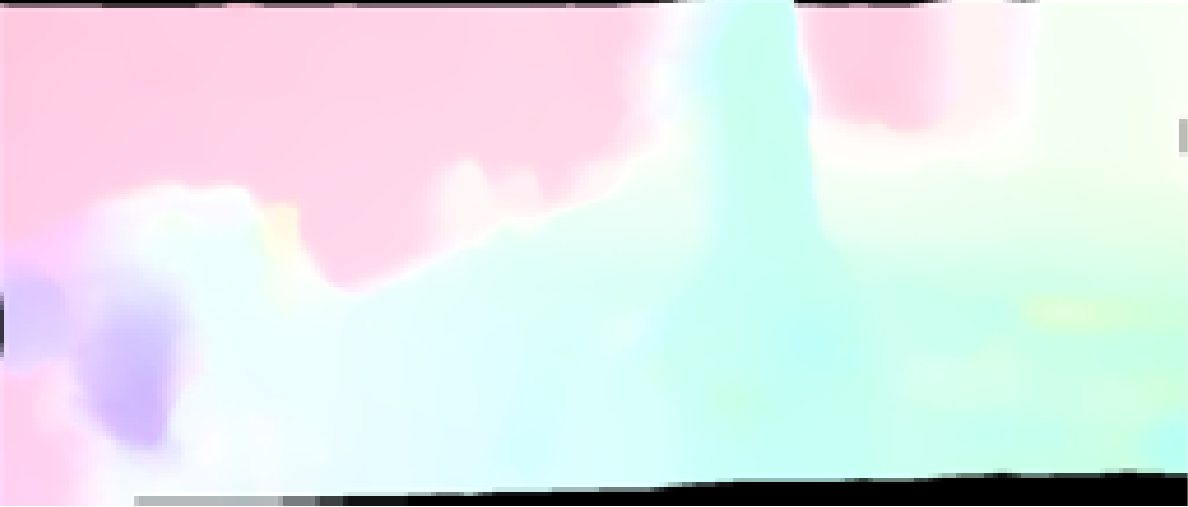}&
        \includegraphics[width=0.25\textwidth]{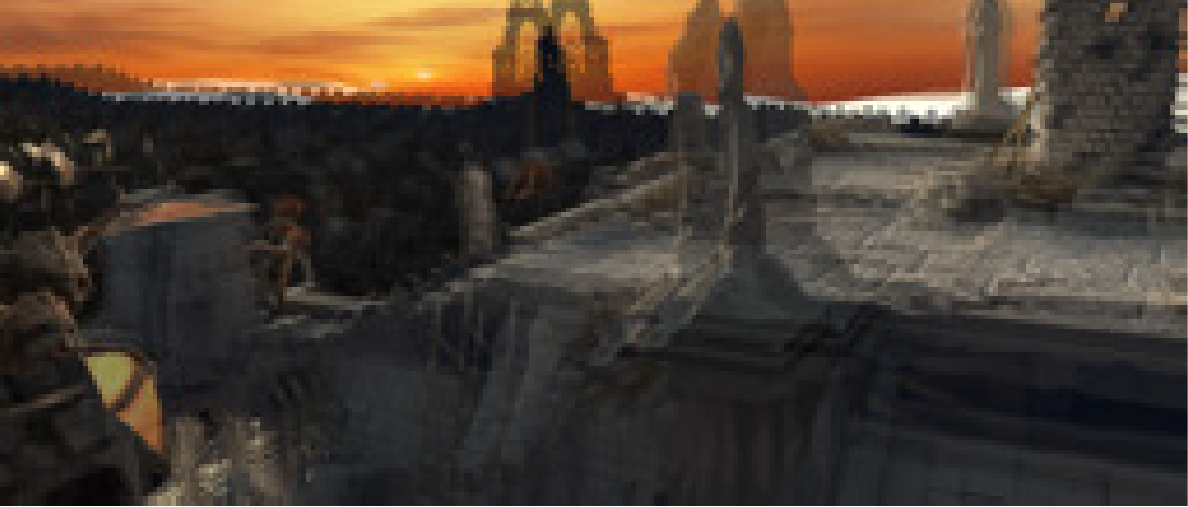} \\                 
        \parbox[b]{3mm}{\rotatebox[origin=l]{90}{6 frames}}&
        \includegraphics[width=0.25\textwidth]{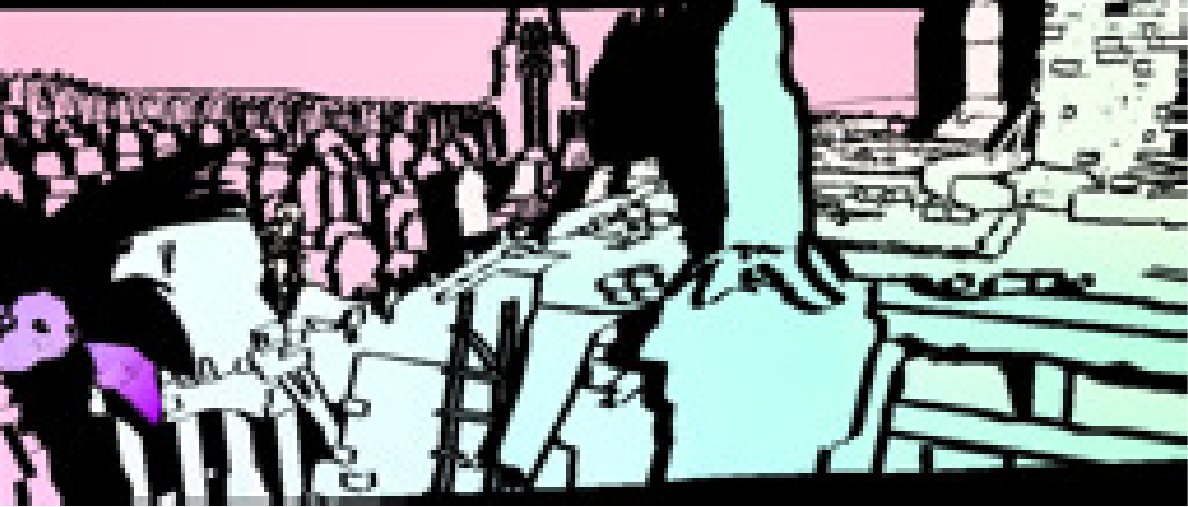}&
        \includegraphics[width=0.25\textwidth]{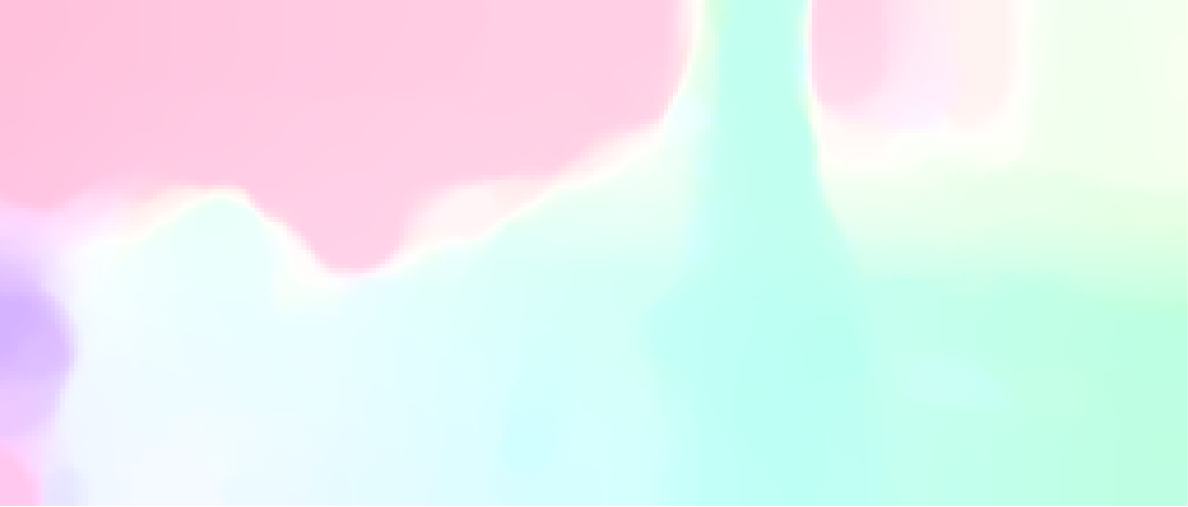}&
        \includegraphics[width=0.25\textwidth]{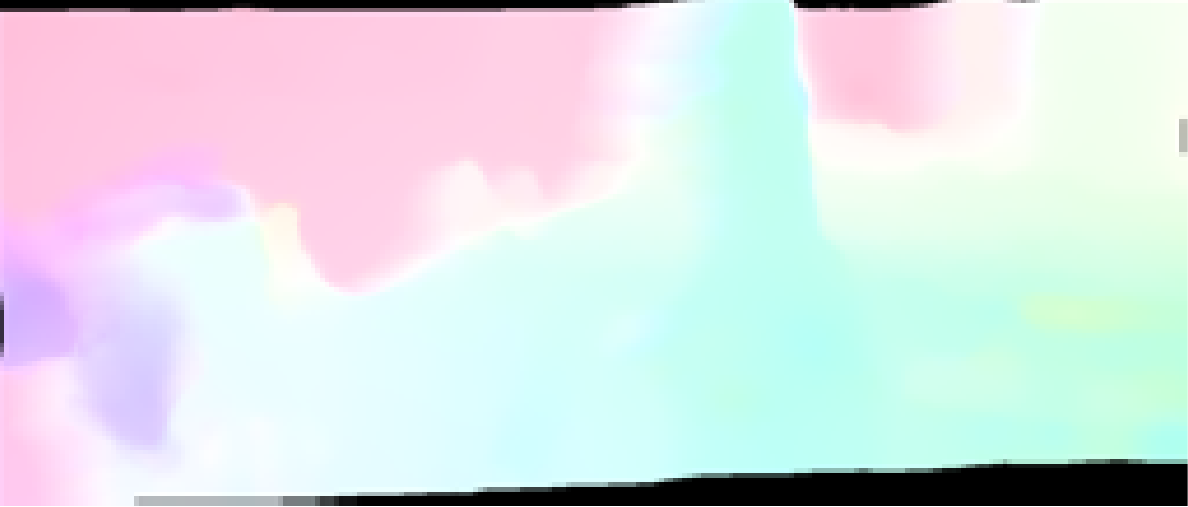}&
        \includegraphics[width=0.25\textwidth]{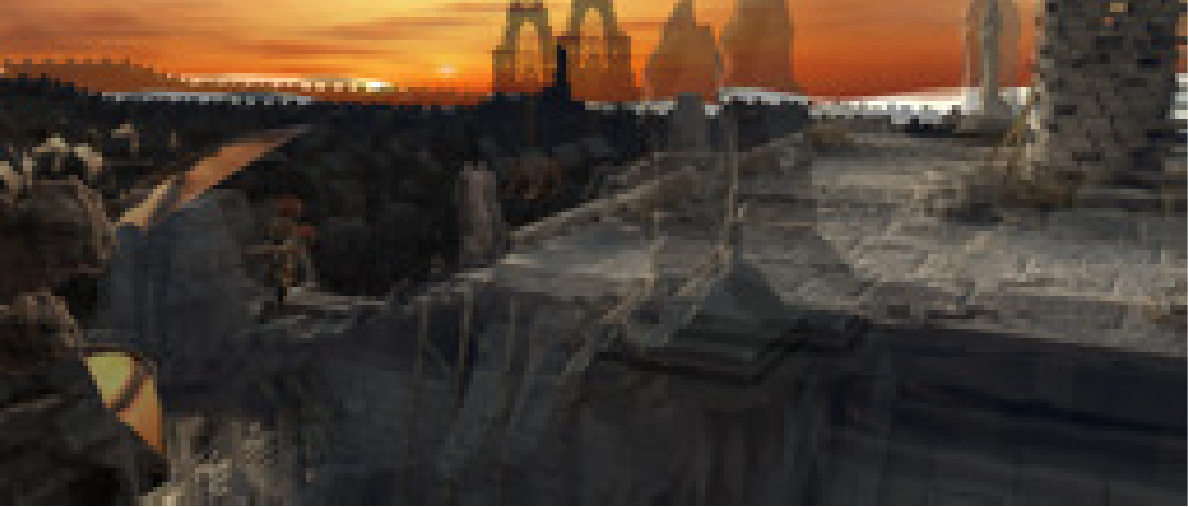} \\                     
        \parbox[b]{3mm}{\rotatebox[origin=l]{90}{8 frames}}&
        \includegraphics[width=0.25\textwidth]{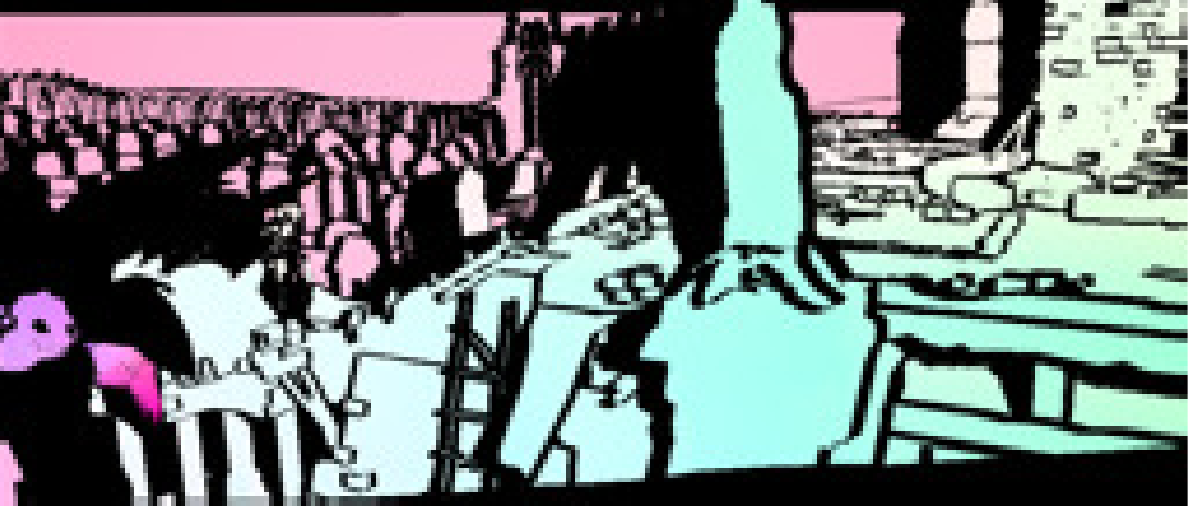}&
        \includegraphics[width=0.25\textwidth]{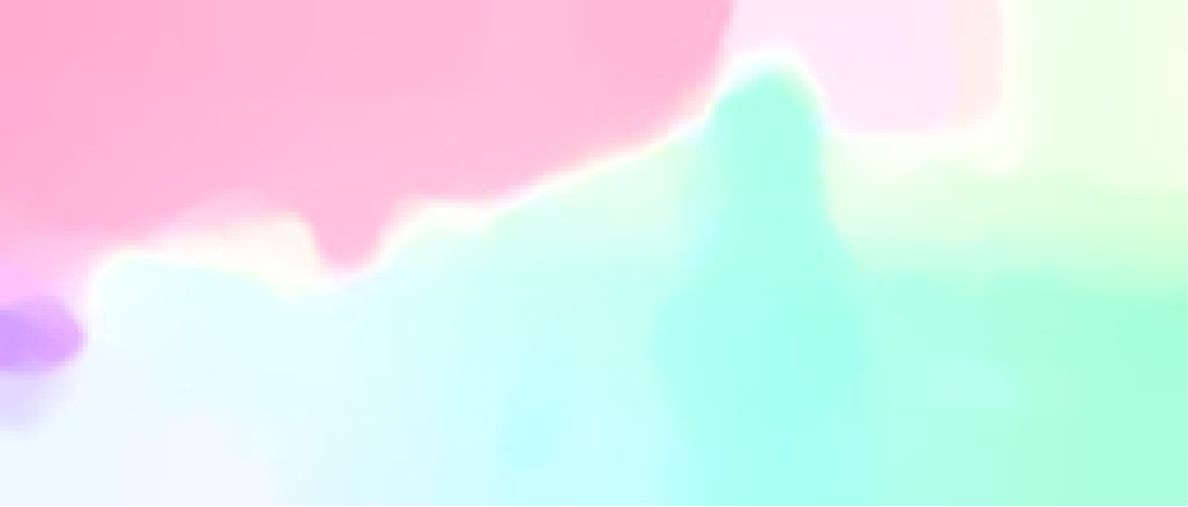}&
        \includegraphics[width=0.25\textwidth]{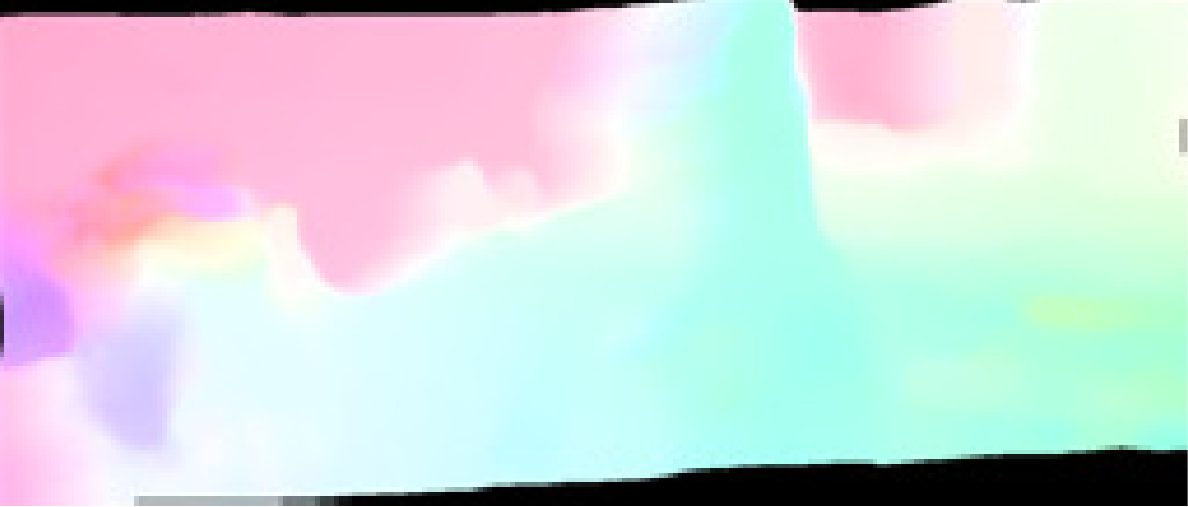}&
        \includegraphics[width=0.25\textwidth]{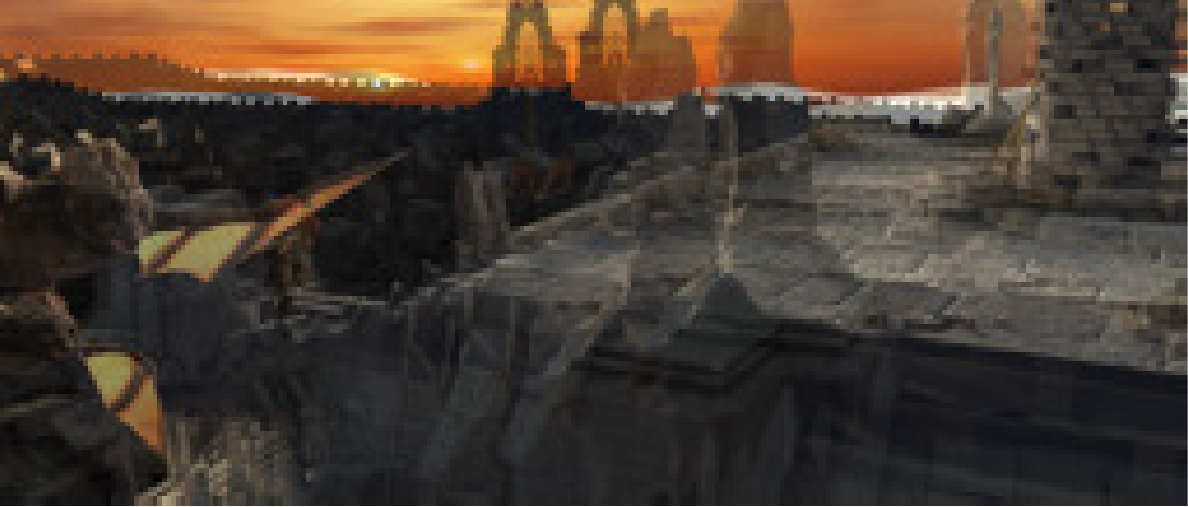} \\                   
        \parbox[b]{3mm}{\rotatebox[origin=l]{90}{10 frames}}&
        \includegraphics[width=0.25\textwidth]{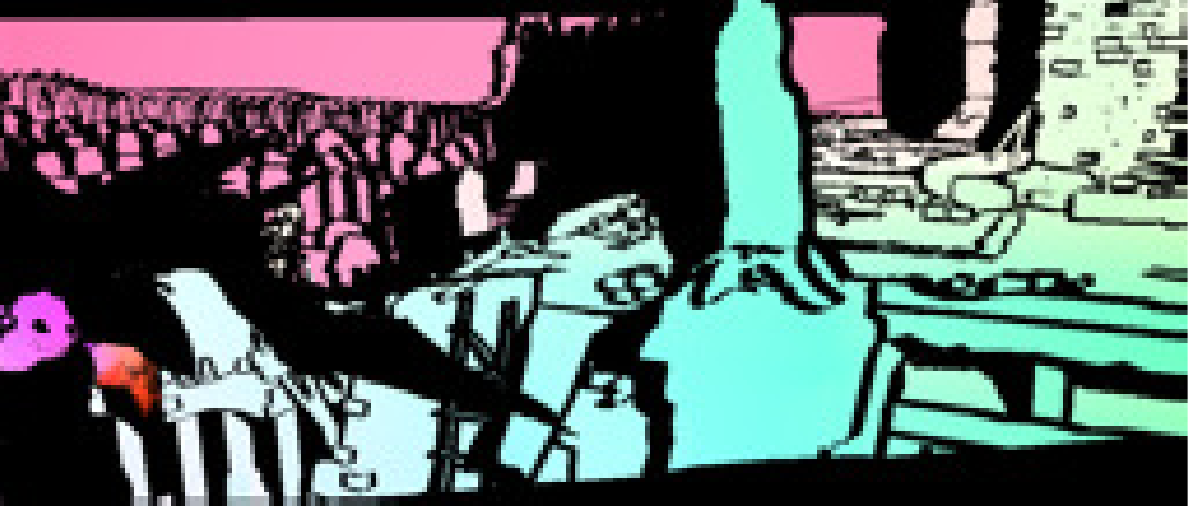}&
        \includegraphics[width=0.25\textwidth]{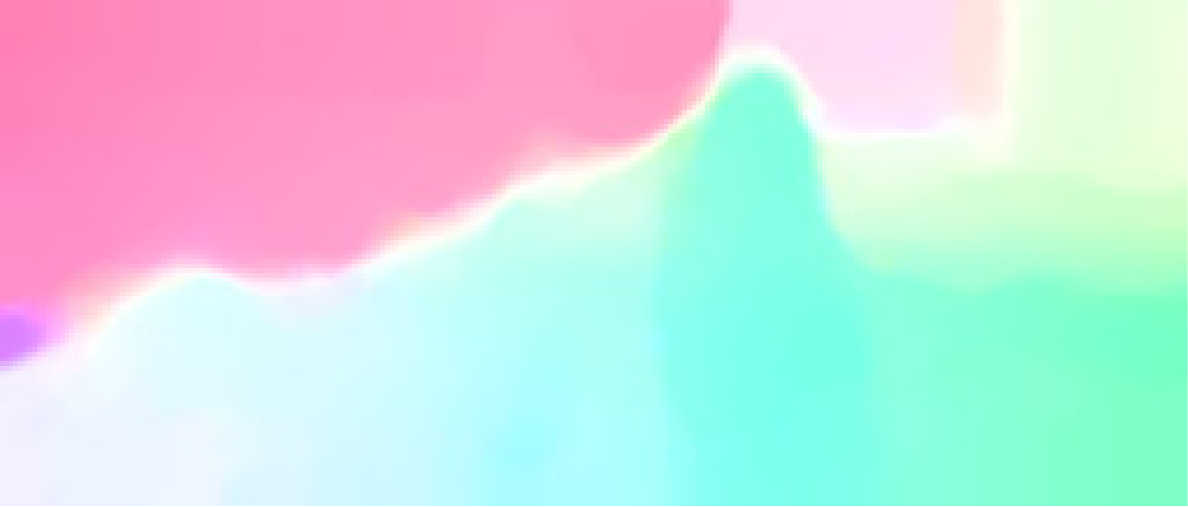}&
        \includegraphics[width=0.25\textwidth]{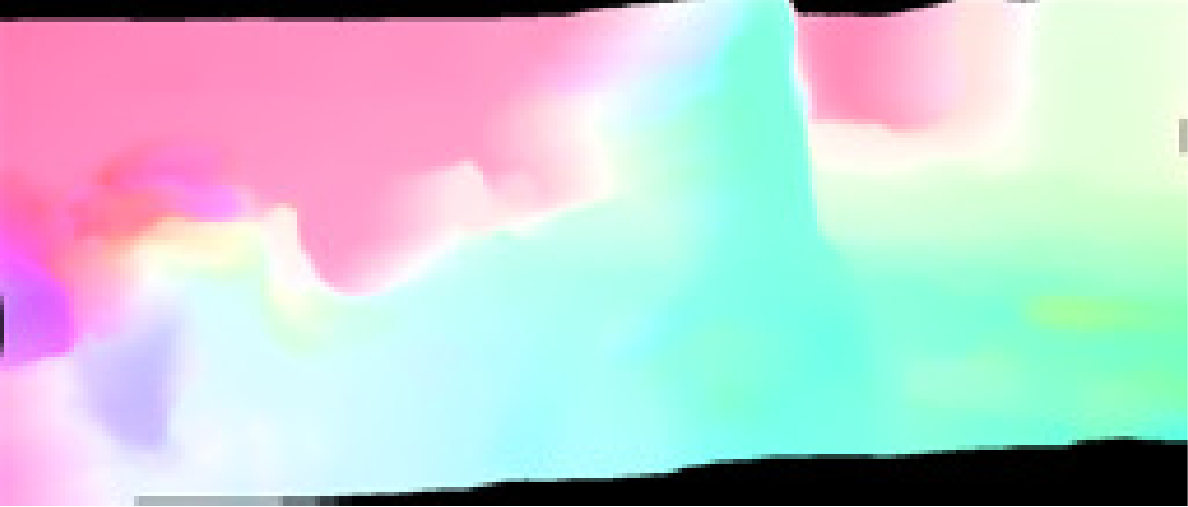}&
        \includegraphics[width=0.25\textwidth]{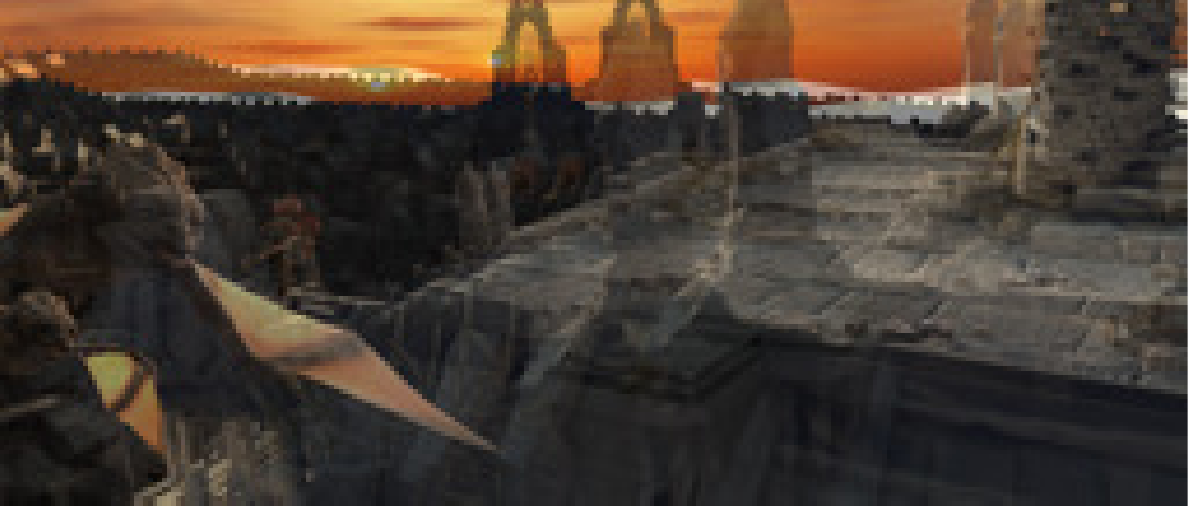}\\[6pt]

        \parbox[b]{3mm}{\rotatebox[origin=l]{90}{1 frame}}&
        \includegraphics[width=0.25\textwidth]{imgs/subs_10_05_01_01}&
        \includegraphics[width=0.25\textwidth]{imgs/subs_10_05_02_01}&
        \includegraphics[width=0.25\textwidth]{imgs/subs_10_05_03_01}&
        \includegraphics[width=0.25\textwidth]{imgs/subs_10_05_04_01} \\     
        \parbox[b]{3mm}{\rotatebox[origin=l]{90}{2 frames}}&
        \includegraphics[width=0.25\textwidth]{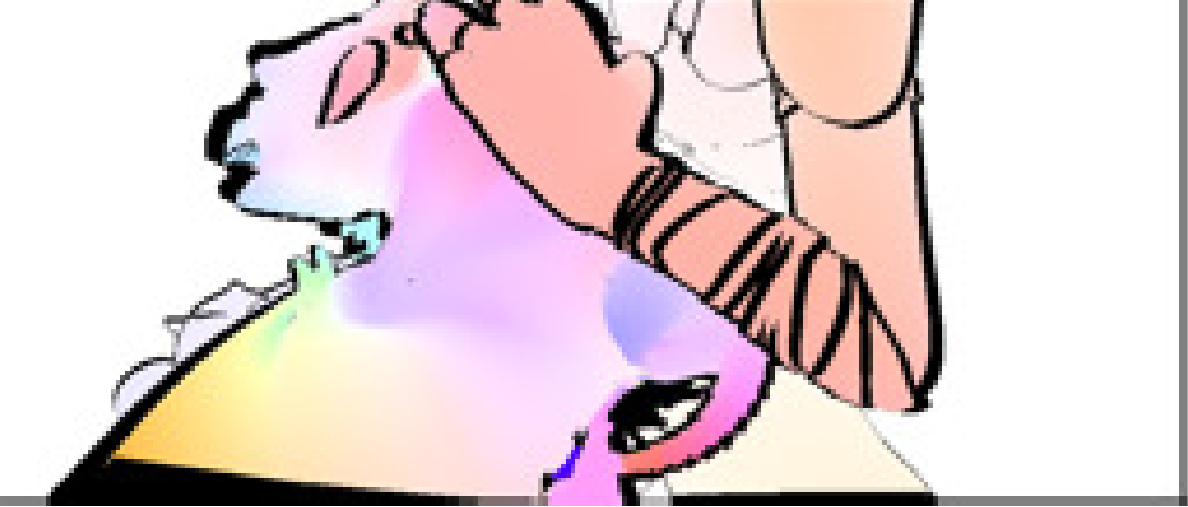}&
        \includegraphics[width=0.25\textwidth]{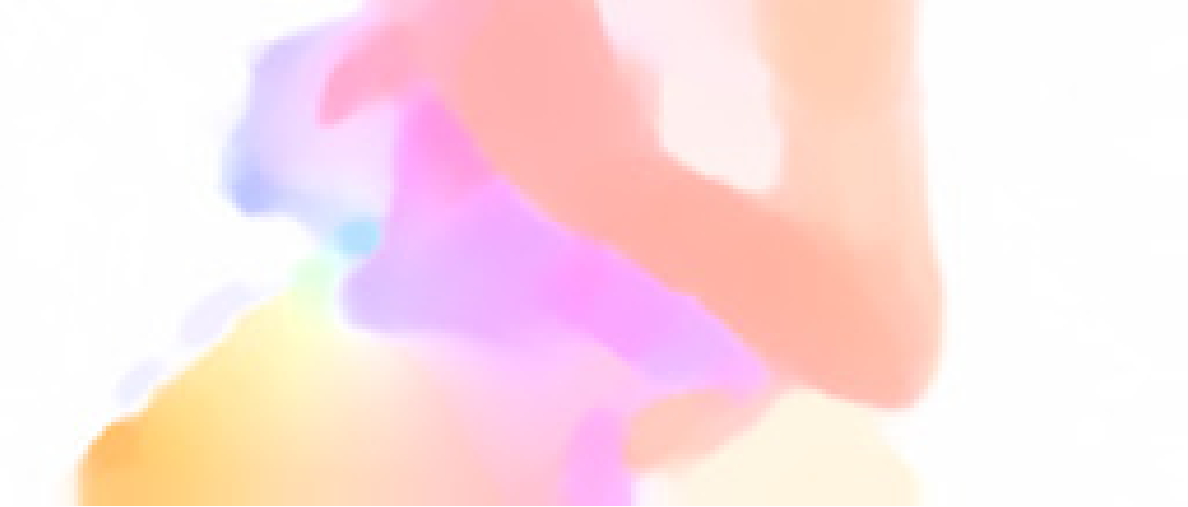}&
        \includegraphics[width=0.25\textwidth]{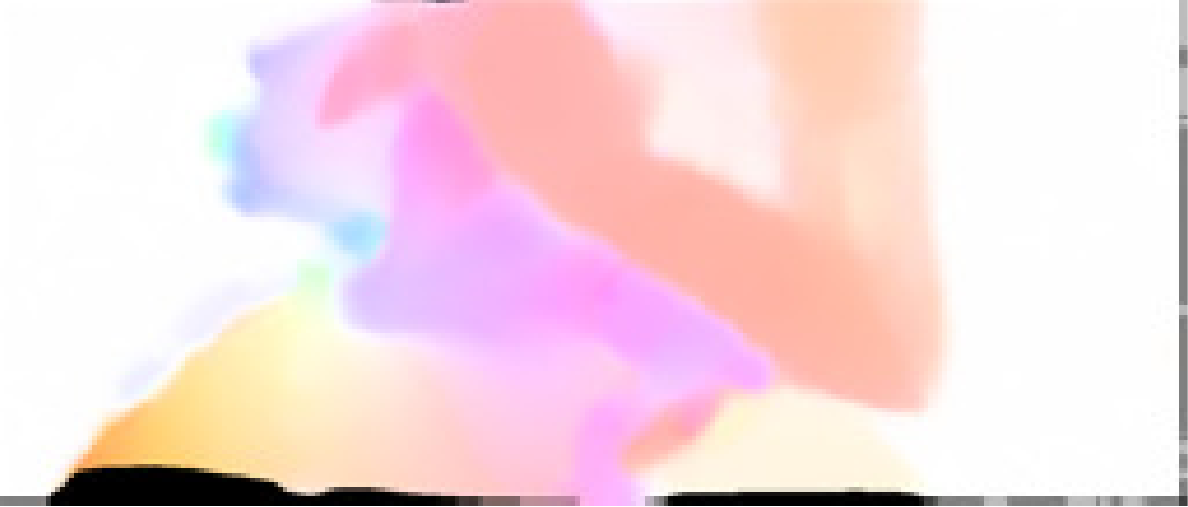}&
        \includegraphics[width=0.25\textwidth]{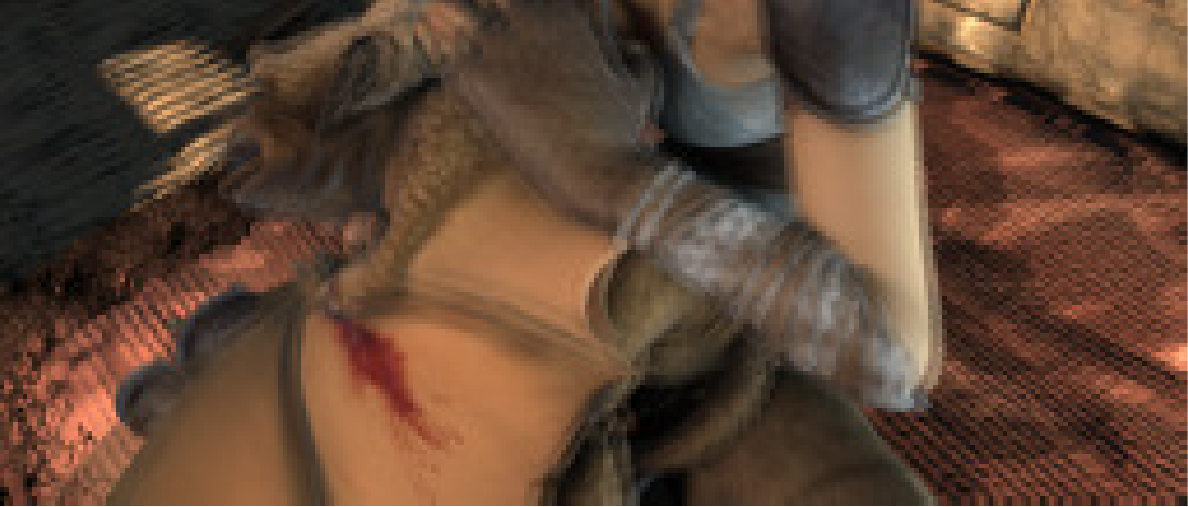} \\        
        \parbox[b]{3mm}{\rotatebox[origin=l]{90}{4 frames}}&
        \includegraphics[width=0.25\textwidth]{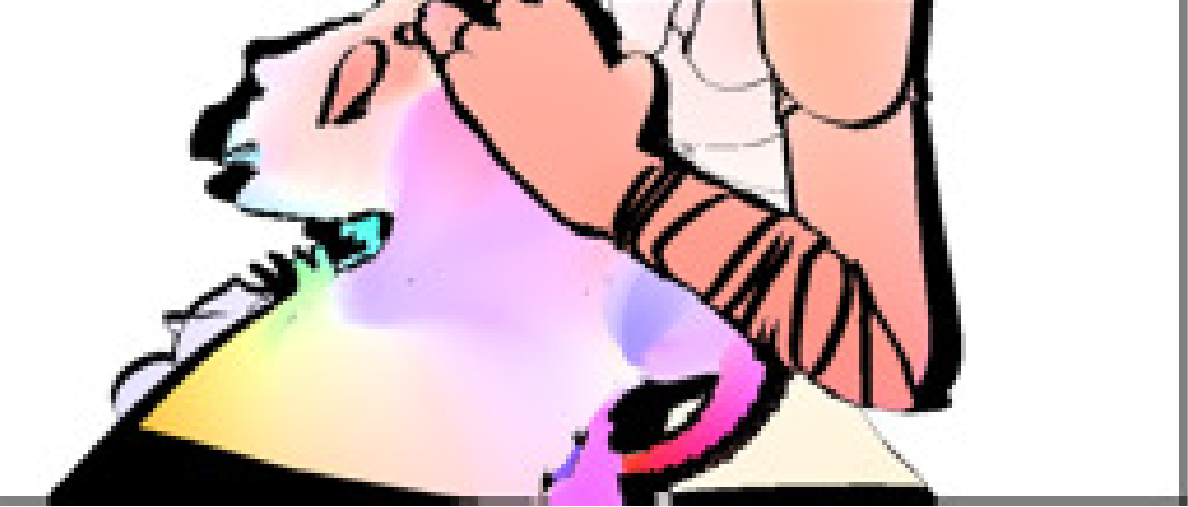}&
        \includegraphics[width=0.25\textwidth]{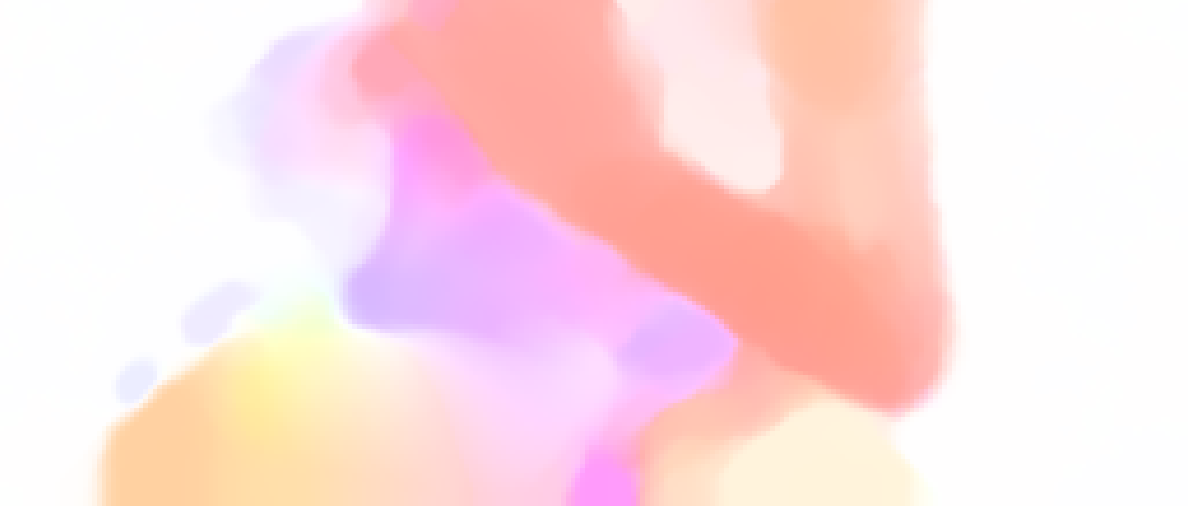}&
        \includegraphics[width=0.25\textwidth]{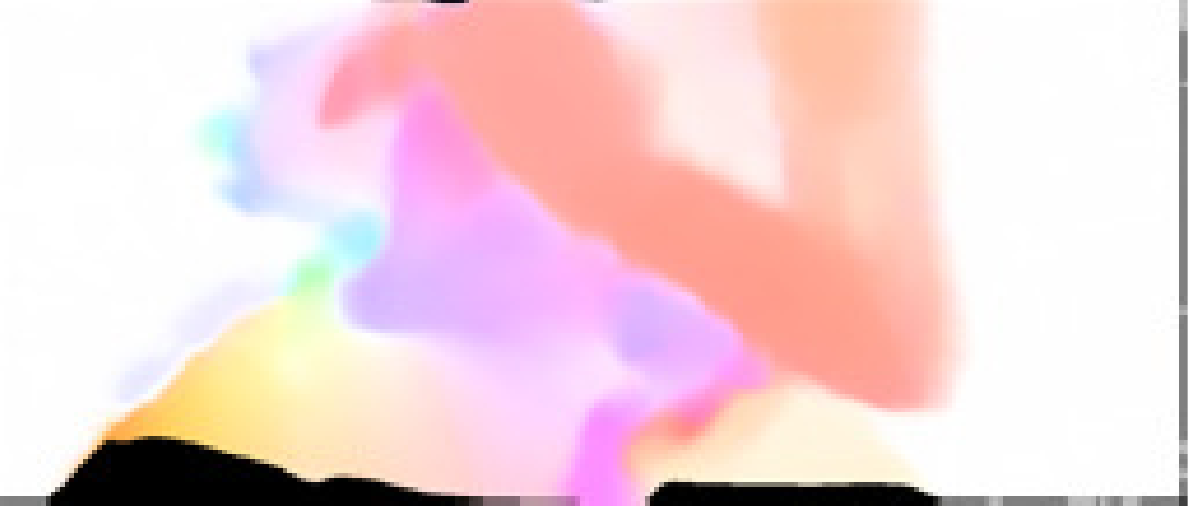}&
        \includegraphics[width=0.25\textwidth]{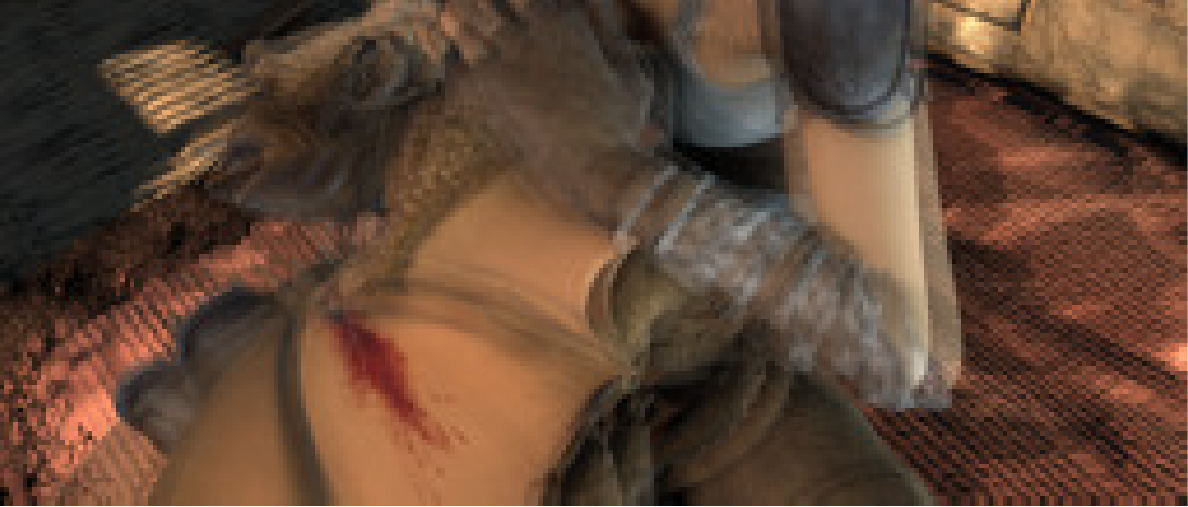} \\                 
        \parbox[b]{3mm}{\rotatebox[origin=l]{90}{6 frames}}&
        \includegraphics[width=0.25\textwidth]{imgs/subs_10_05_01_06}&
        \includegraphics[width=0.25\textwidth]{imgs/subs_10_05_02_06}&
        \includegraphics[width=0.25\textwidth]{imgs/subs_10_05_03_06}&
        \includegraphics[width=0.25\textwidth]{imgs/subs_10_05_04_06} \\                     
        \parbox[b]{3mm}{\rotatebox[origin=l]{90}{8 frames}}&
        \includegraphics[width=0.25\textwidth]{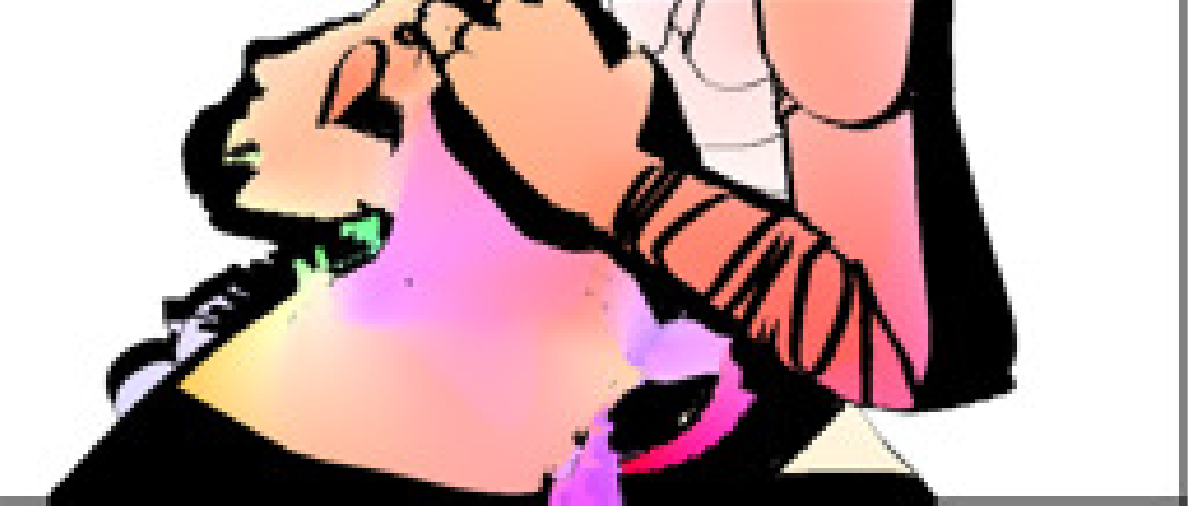}&
        \includegraphics[width=0.25\textwidth]{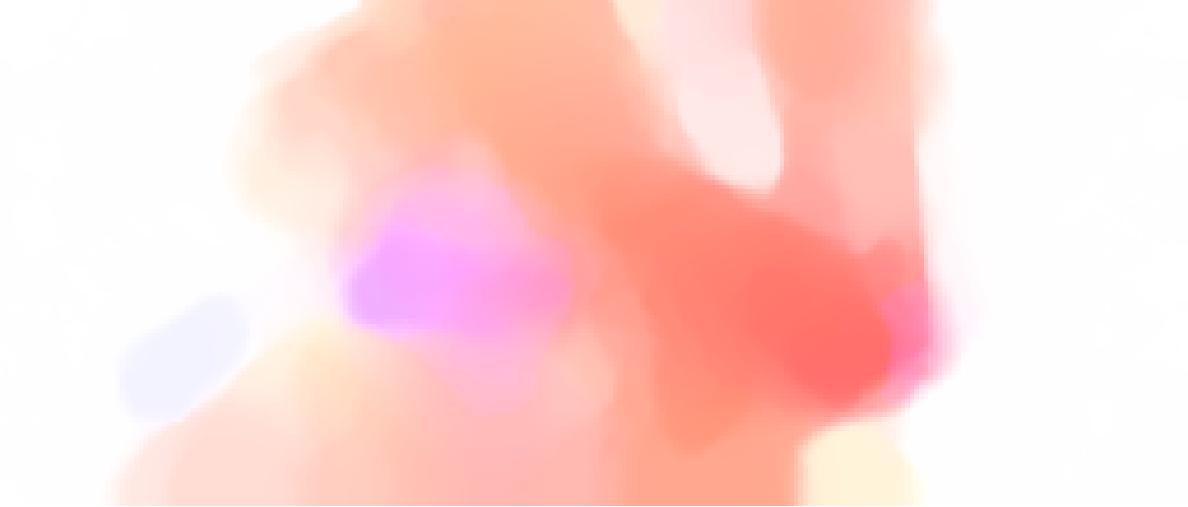}&
        \includegraphics[width=0.25\textwidth]{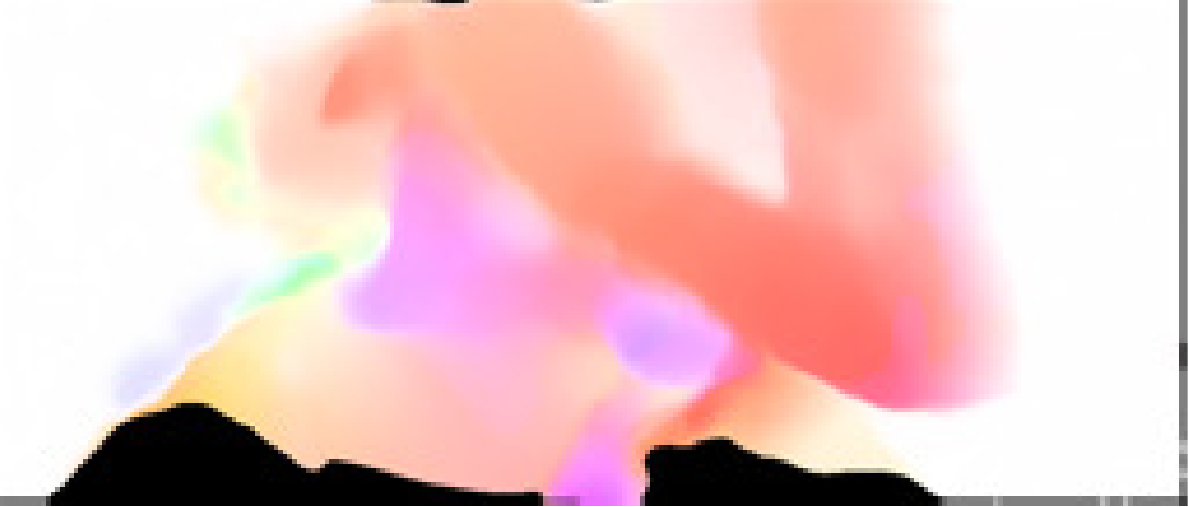}&
        \includegraphics[width=0.25\textwidth]{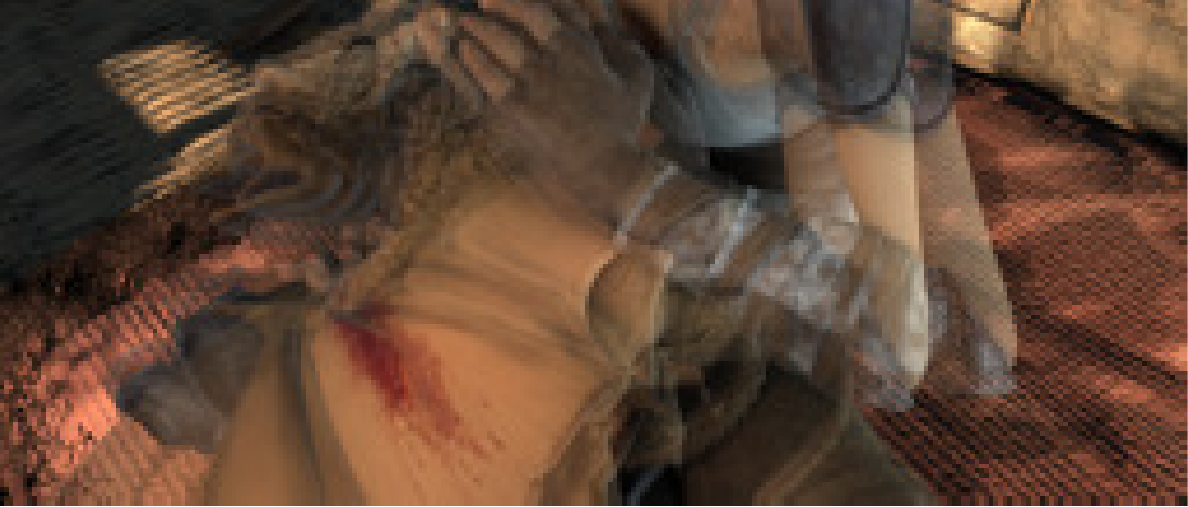} \\                   
        \parbox[b]{3mm}{\rotatebox[origin=l]{90}{10 frames}}&
        \includegraphics[width=0.25\textwidth]{imgs/subs_10_05_01_10}&
        \includegraphics[width=0.25\textwidth]{imgs/subs_10_05_02_10}&
        \includegraphics[width=0.25\textwidth]{imgs/subs_10_05_03_10}&
        \includegraphics[width=0.25\textwidth]{imgs/subs_10_05_04_10}
        \end{tabular}
        }
        \caption{Optical flow on Sintel with lower temporal resolution. In each block of 6x4: Rows, top to bottom, correspond to step sizes 1 (original frame-rate), 2, 4, 6, 8, 10 frames. Columns, left to right, correspond to new ground truth, DeepFlow result, DIS result (through \emph{all intermediate frames}), original images. Large displacements are significantly better preserved by DIS through higher frame-rates.}\label{fig:subsample_exa1_AP}
\end{figure*}

\begin{figure*}[h!]
\large
\centering\setlength{\tabcolsep}{0.1pt}\renewcommand{\arraystretch}{0} 
        {
        \begin{tabular}{ccccc}
        &{\bf Ground Truth} & {\bf DeepFlow} & {\bf DIS (all frames)} & {\bf difference image}\\
        \parbox[b]{3mm}{\rotatebox[origin=l]{90}{1 frame}}&
        \includegraphics[width=0.25\textwidth]{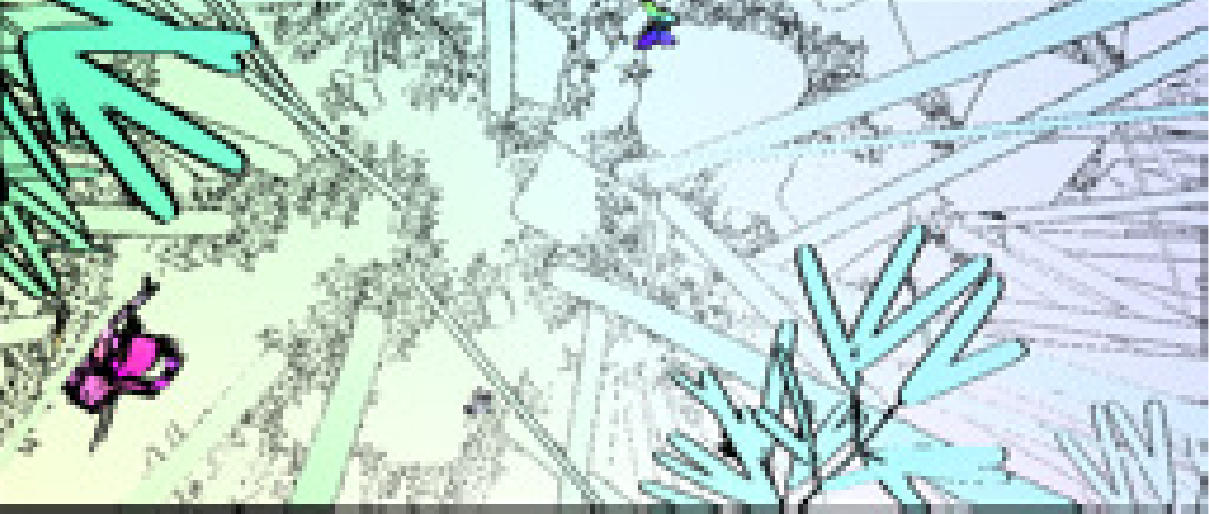}&
        \includegraphics[width=0.25\textwidth]{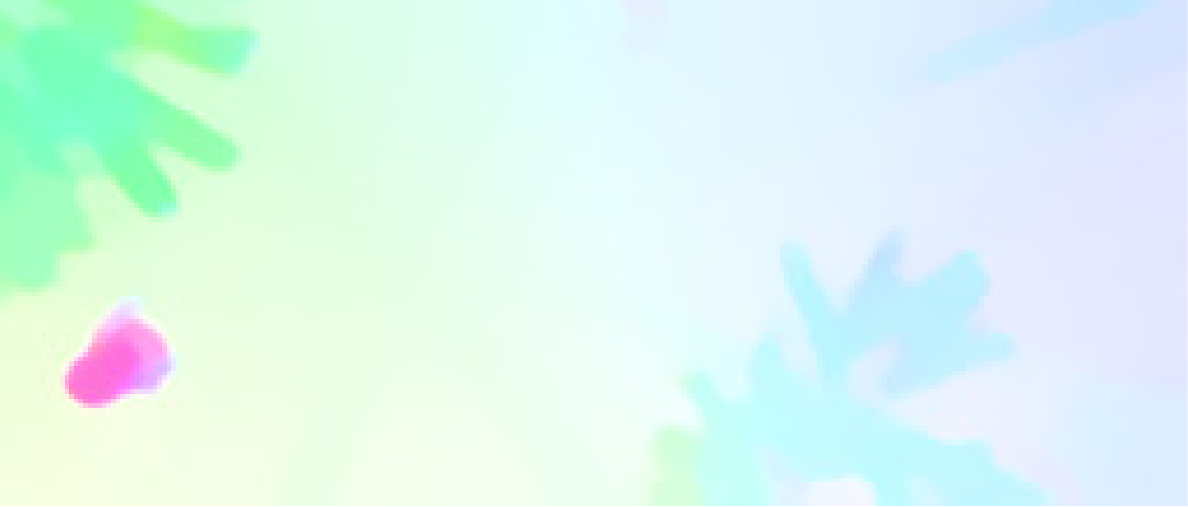}&
        \includegraphics[width=0.25\textwidth]{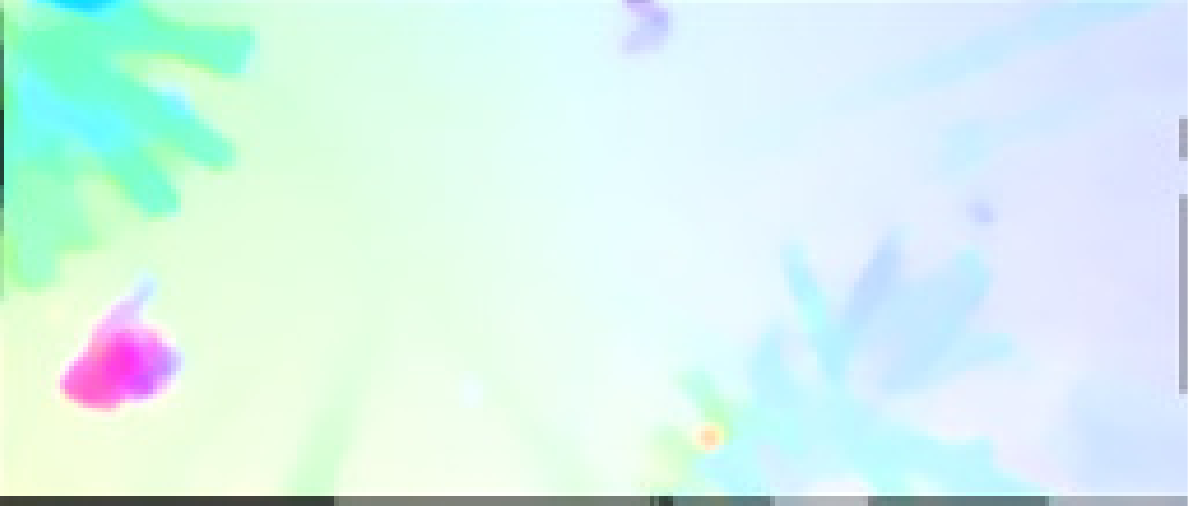}&
        \includegraphics[width=0.25\textwidth]{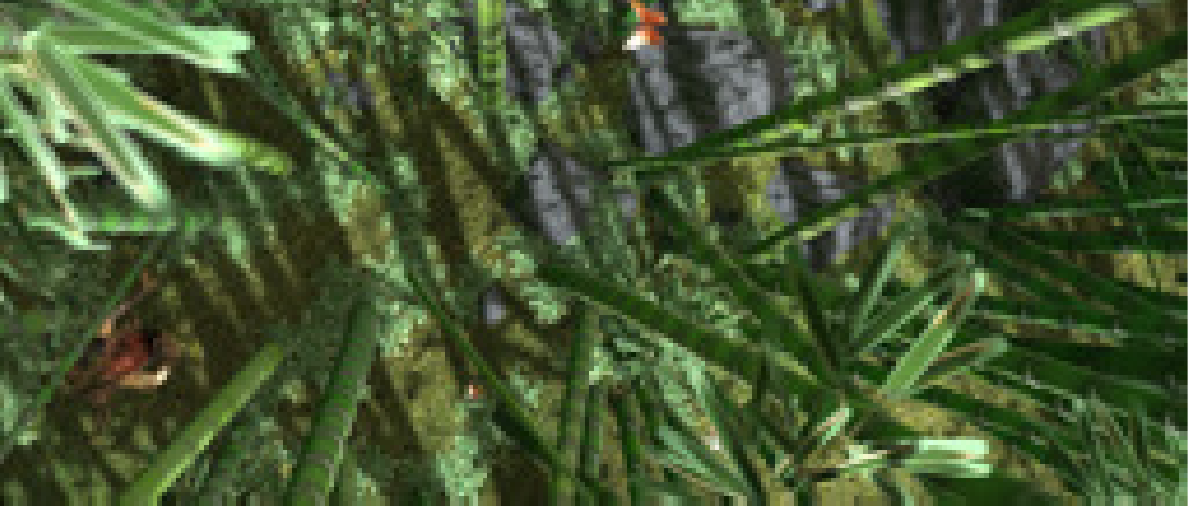}\\     
        \parbox[b]{3mm}{\rotatebox[origin=l]{90}{2 frames}}&
        \includegraphics[width=0.25\textwidth]{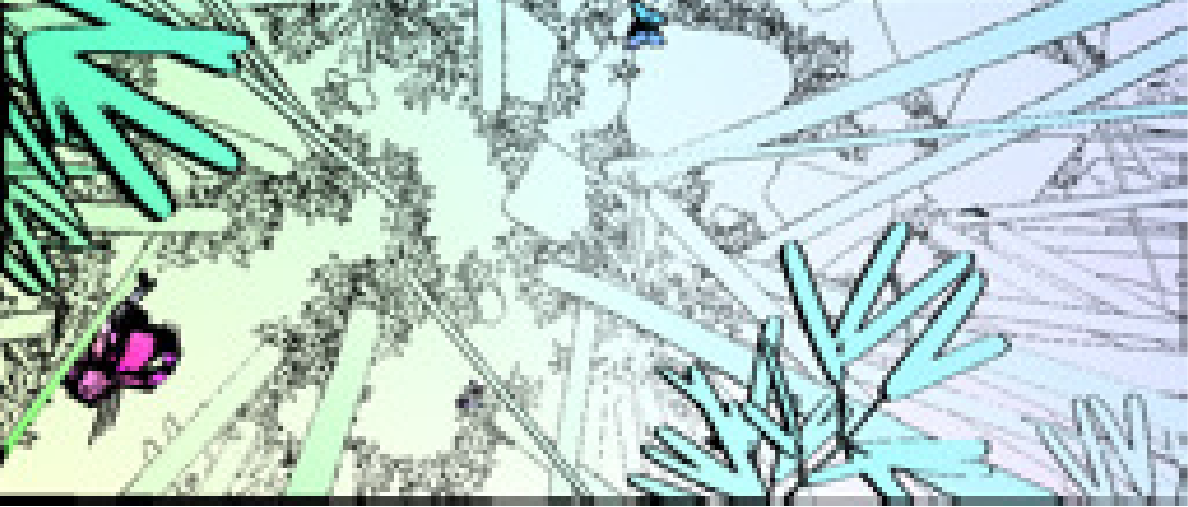}&
        \includegraphics[width=0.25\textwidth]{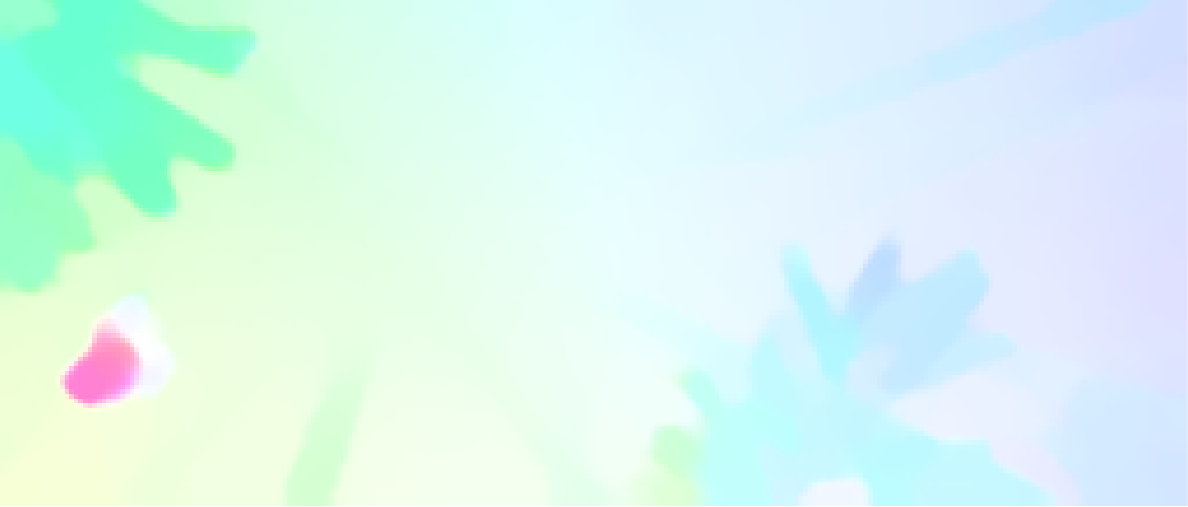}&
        \includegraphics[width=0.25\textwidth]{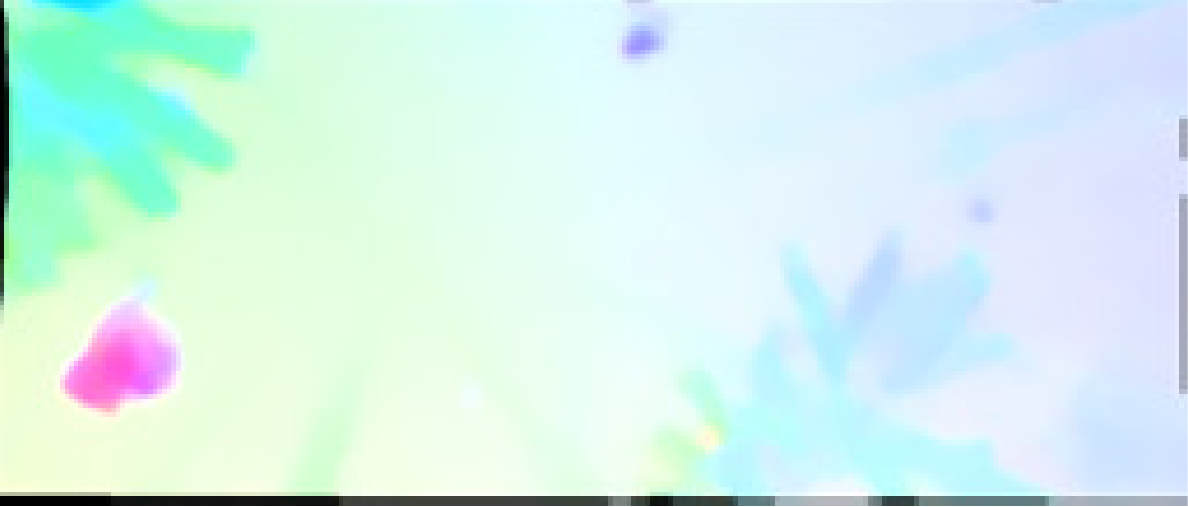}&
        \includegraphics[width=0.25\textwidth]{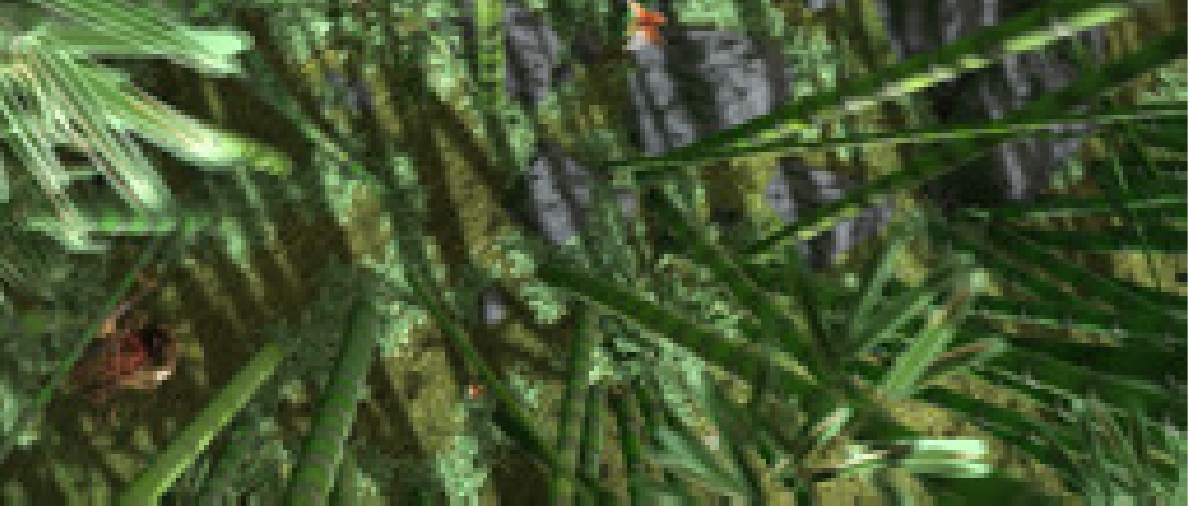} \\        
        \parbox[b]{3mm}{\rotatebox[origin=l]{90}{4 frames}}&
        \includegraphics[width=0.25\textwidth]{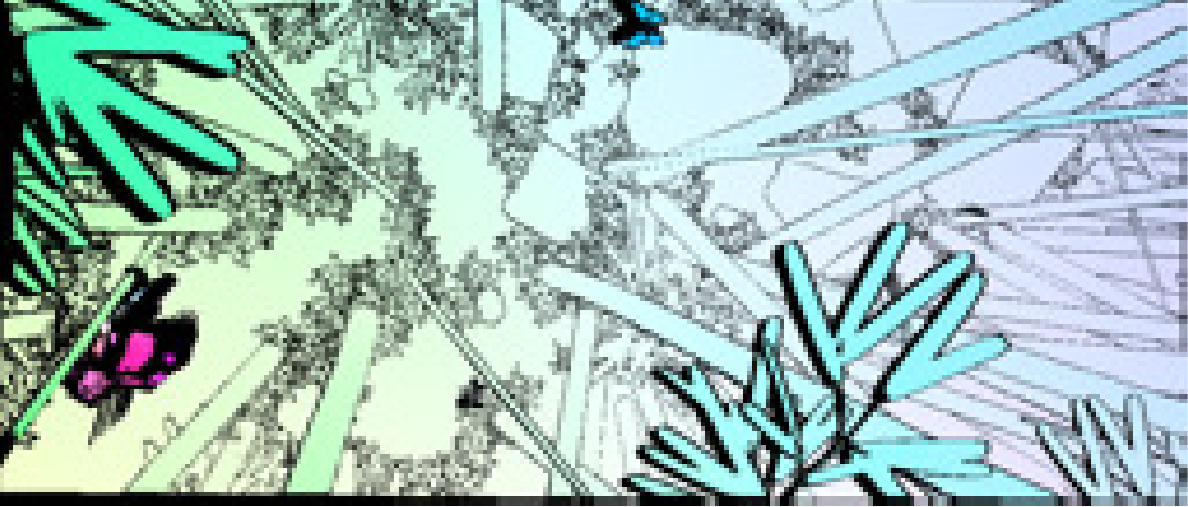}&
        \includegraphics[width=0.25\textwidth]{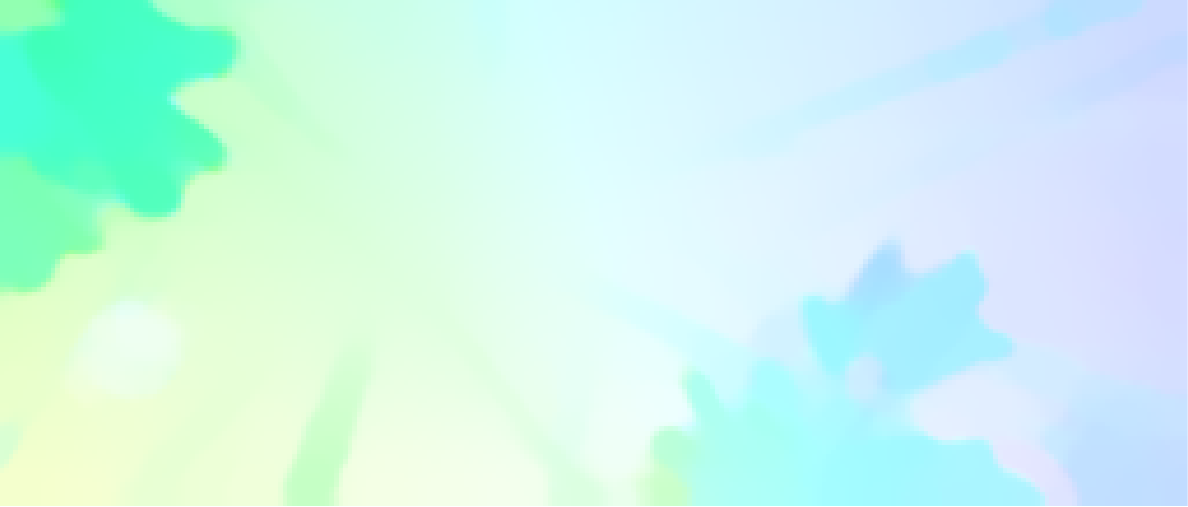}&
        \includegraphics[width=0.25\textwidth]{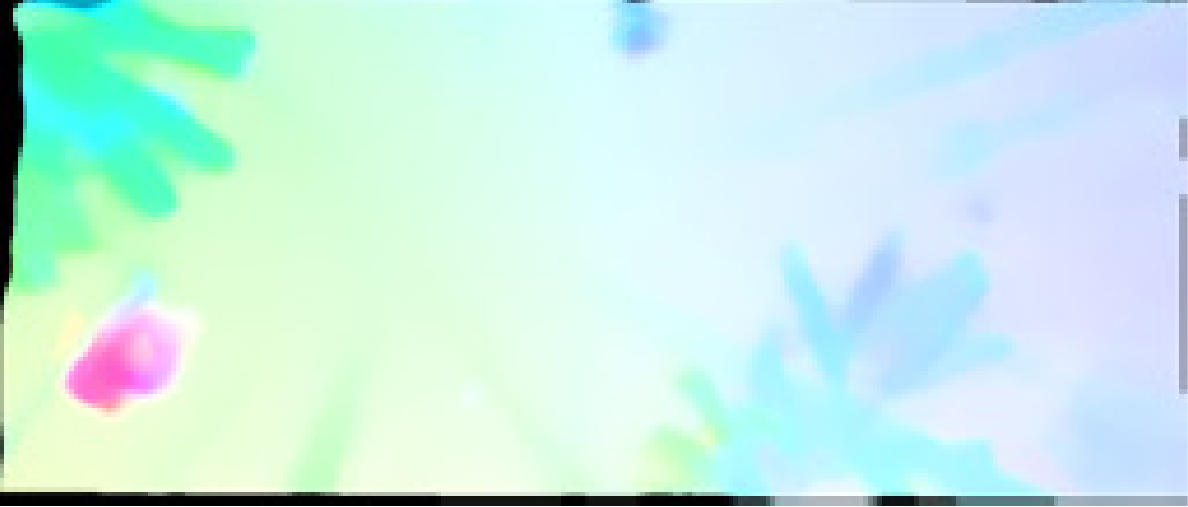}&
        \includegraphics[width=0.25\textwidth]{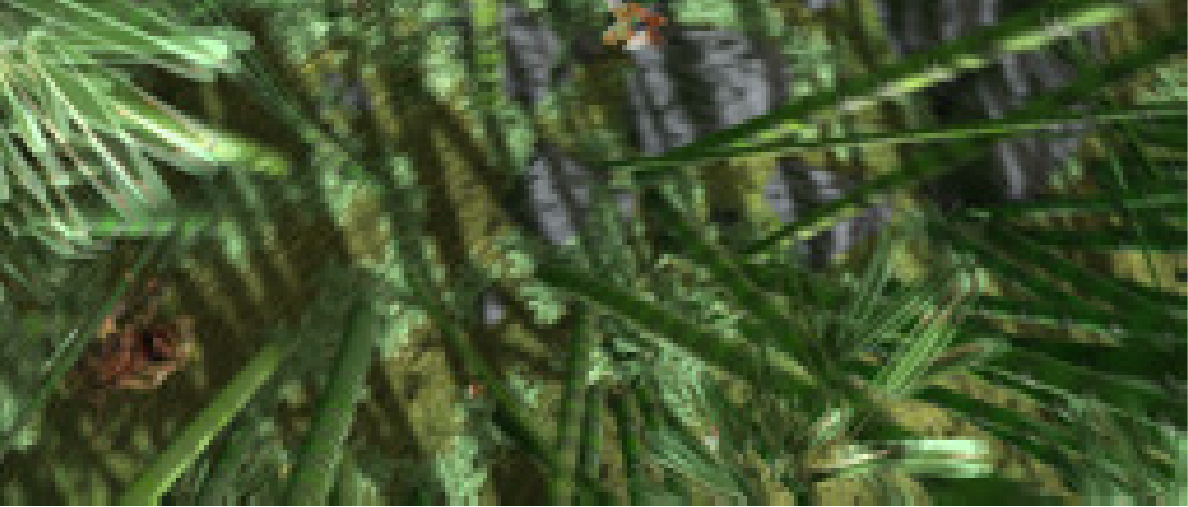} \\                 
        \parbox[b]{3mm}{\rotatebox[origin=l]{90}{6 frames}}&
        \includegraphics[width=0.25\textwidth]{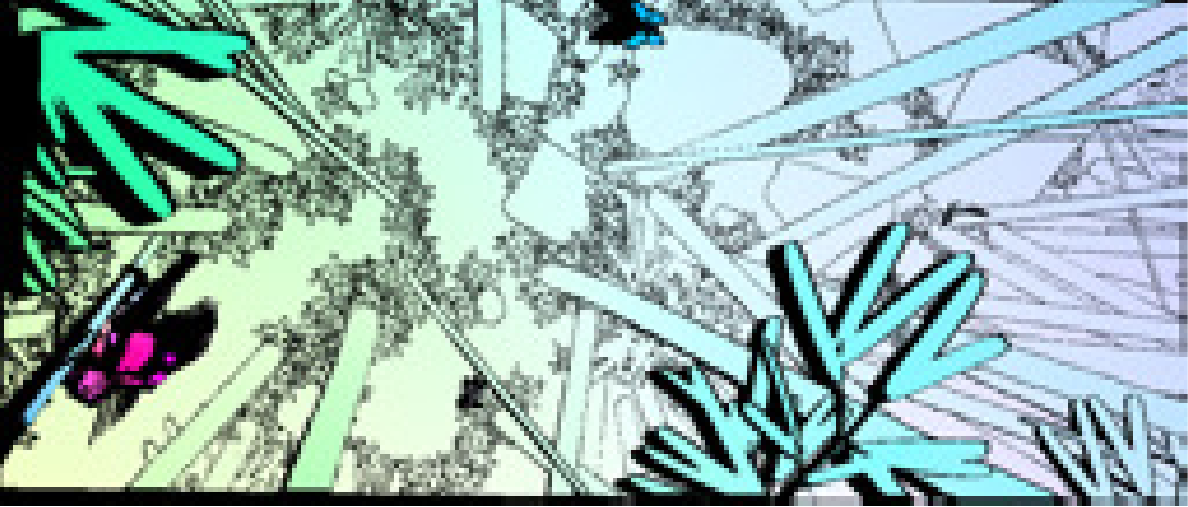}&
        \includegraphics[width=0.25\textwidth]{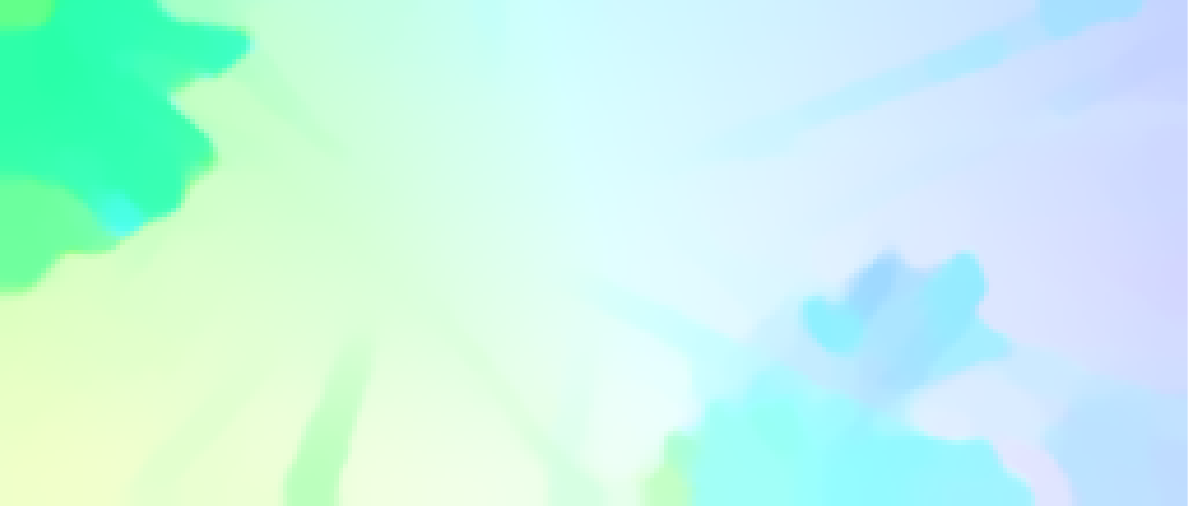}&
        \includegraphics[width=0.25\textwidth]{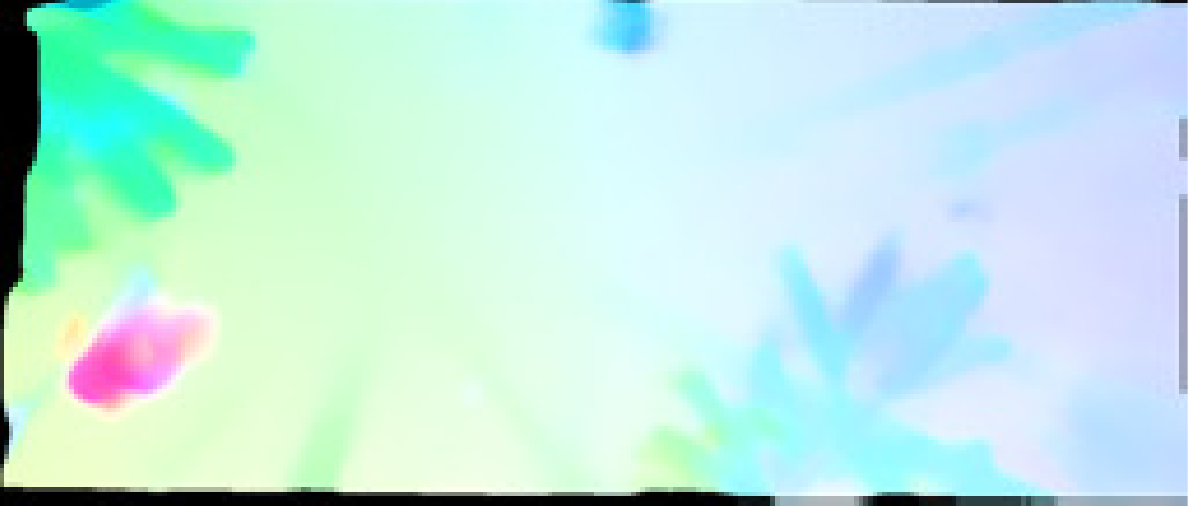}&
        \includegraphics[width=0.25\textwidth]{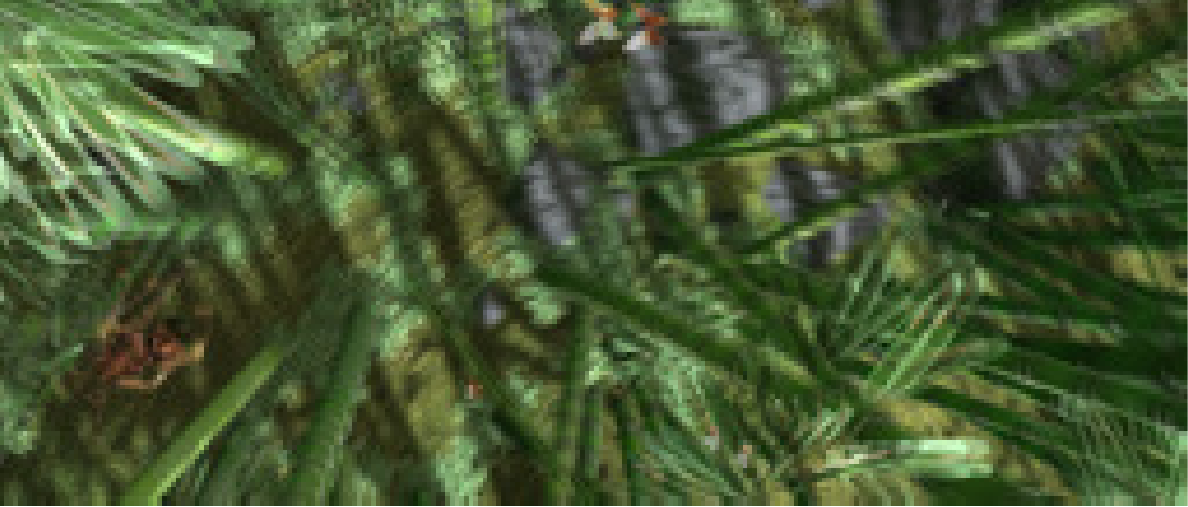} \\                     
        \parbox[b]{3mm}{\rotatebox[origin=l]{90}{8 frames}}&
        \includegraphics[width=0.25\textwidth]{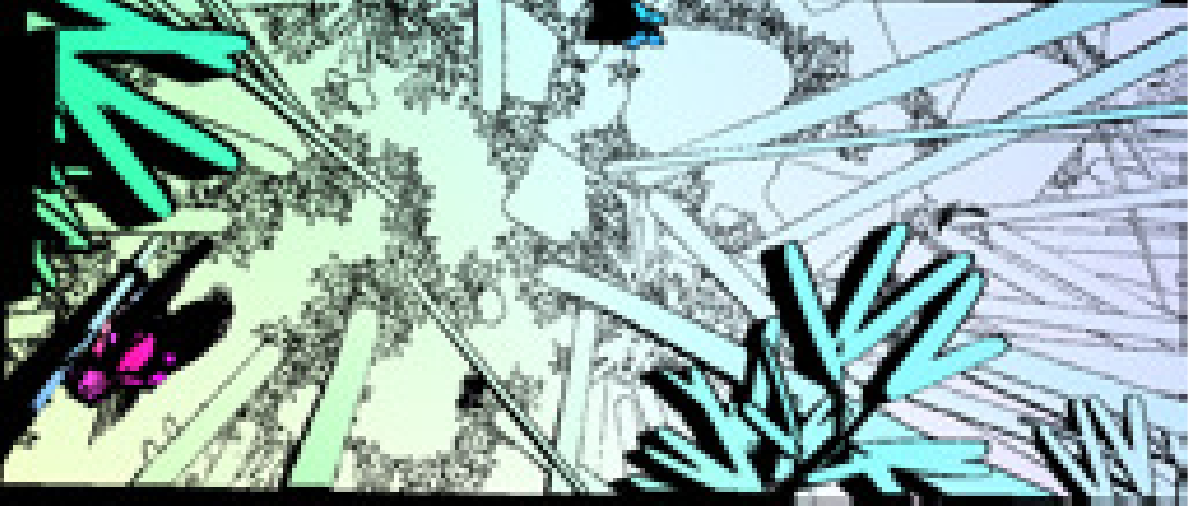}&
        \includegraphics[width=0.25\textwidth]{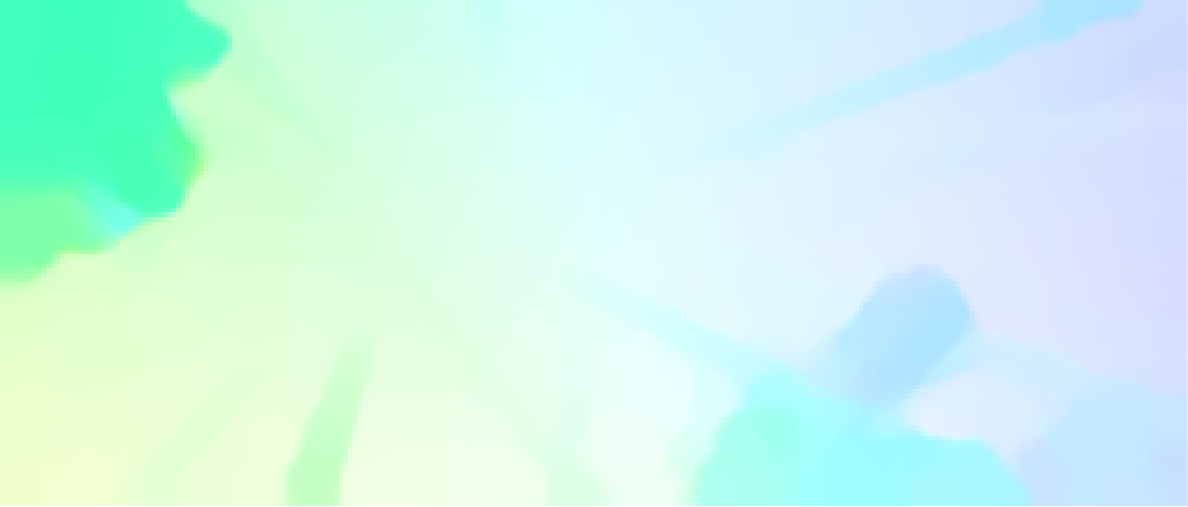}&
        \includegraphics[width=0.25\textwidth]{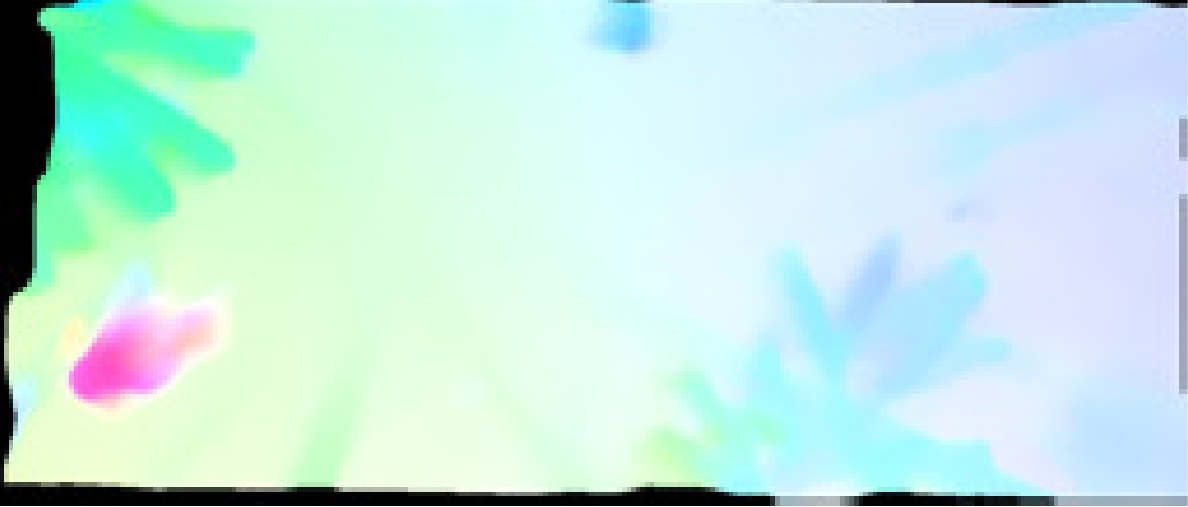}&
        \includegraphics[width=0.25\textwidth]{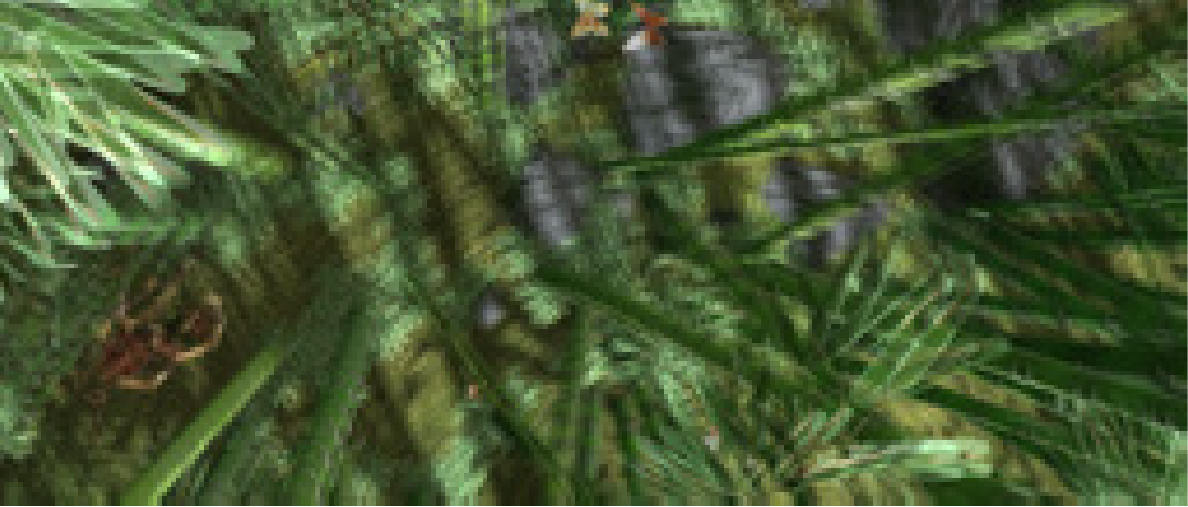} \\                   
        \parbox[b]{3mm}{\rotatebox[origin=l]{90}{10 frames}}&
        \includegraphics[width=0.25\textwidth]{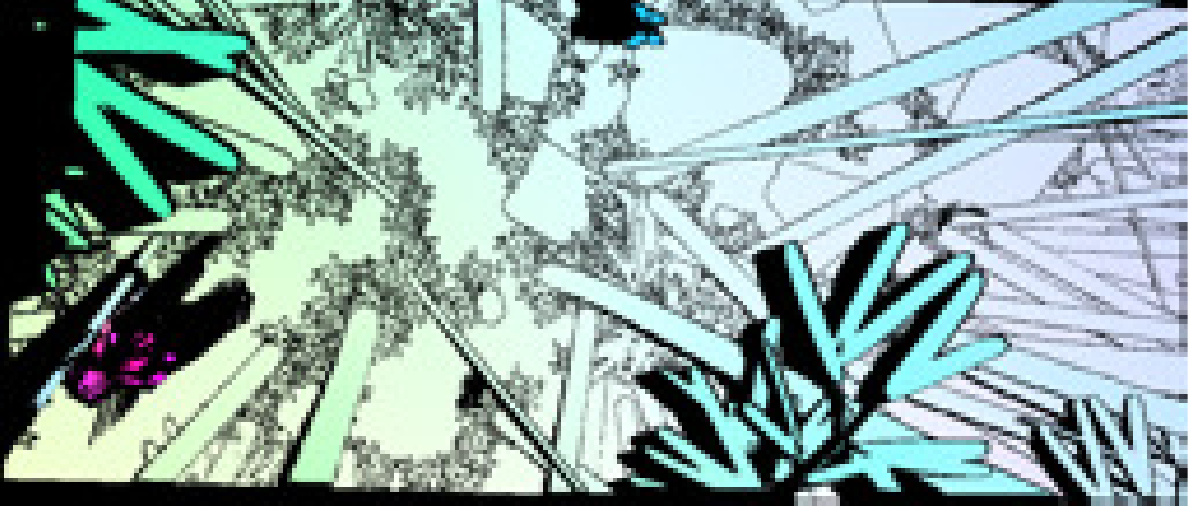}&
        \includegraphics[width=0.25\textwidth]{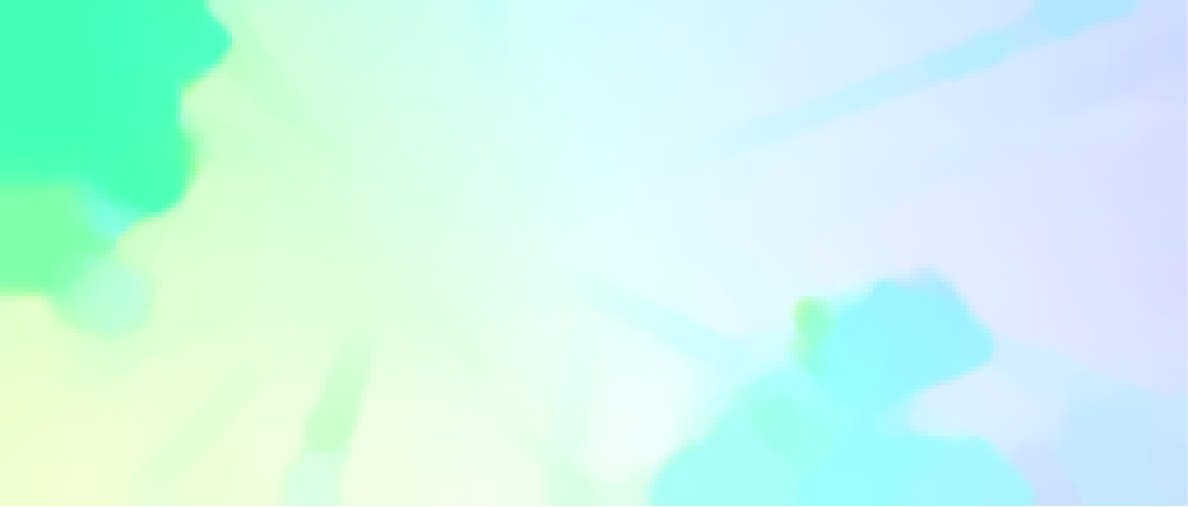}&
        \includegraphics[width=0.25\textwidth]{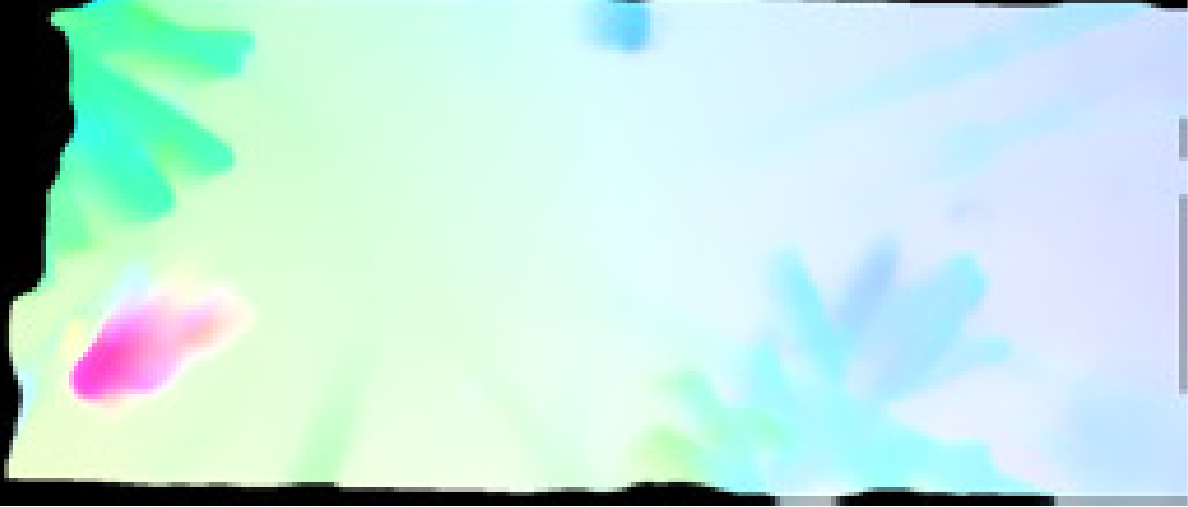}&
        \includegraphics[width=0.25\textwidth]{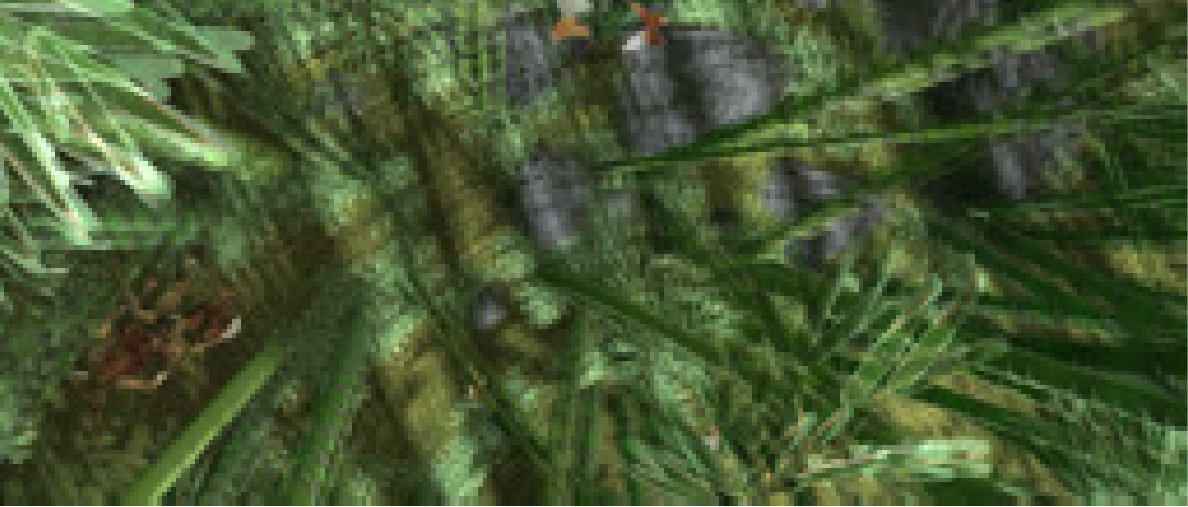}\\[6pt]

        \parbox[b]{3mm}{\rotatebox[origin=l]{90}{1 frame}}&
        \includegraphics[width=0.25\textwidth]{imgs/subs_13_05_01_01}&
        \includegraphics[width=0.25\textwidth]{imgs/subs_13_05_02_01}&
        \includegraphics[width=0.25\textwidth]{imgs/subs_13_05_03_01}&
        \includegraphics[width=0.25\textwidth]{imgs/subs_13_05_04_01} \\     
        \parbox[b]{3mm}{\rotatebox[origin=l]{90}{2 frames}}&        
        \includegraphics[width=0.25\textwidth]{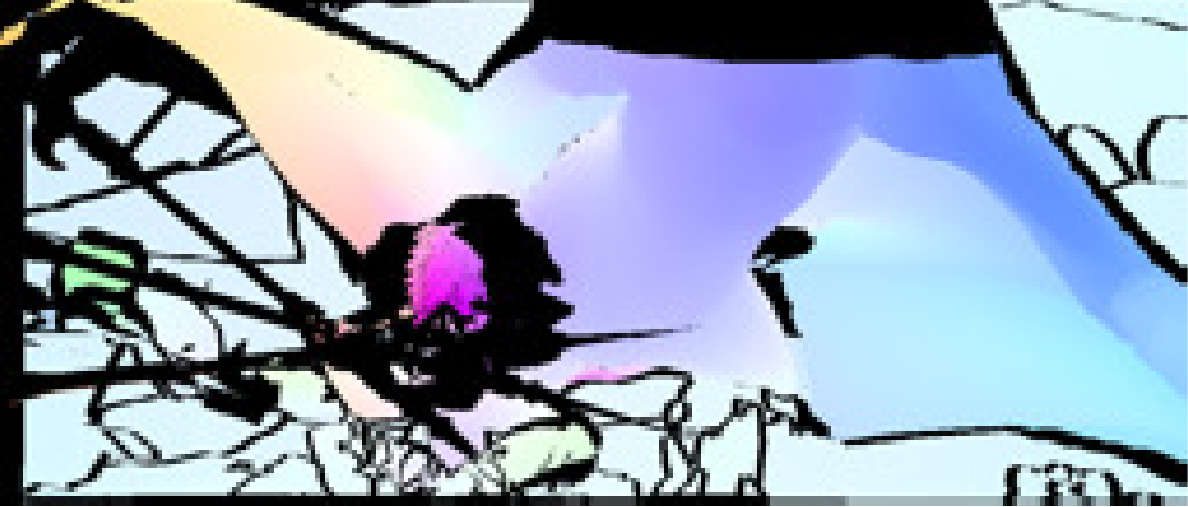}&
        \includegraphics[width=0.25\textwidth]{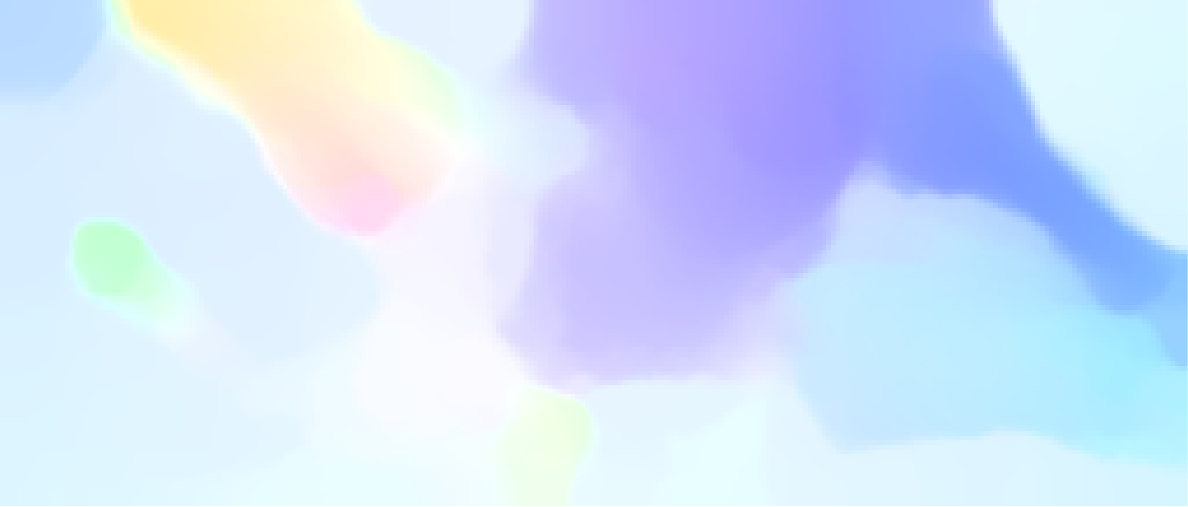}&
        \includegraphics[width=0.25\textwidth]{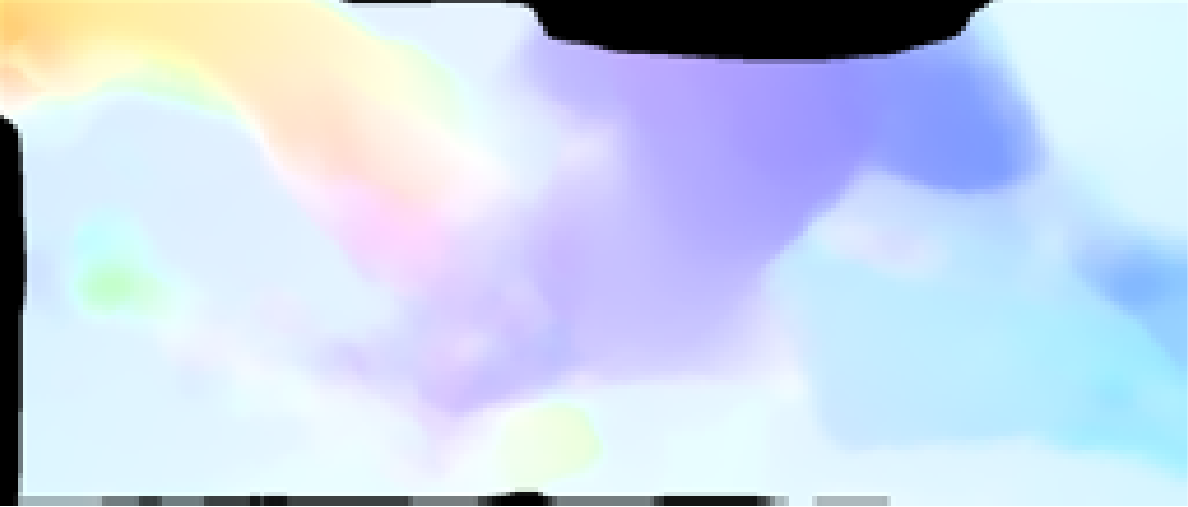}&
        \includegraphics[width=0.25\textwidth]{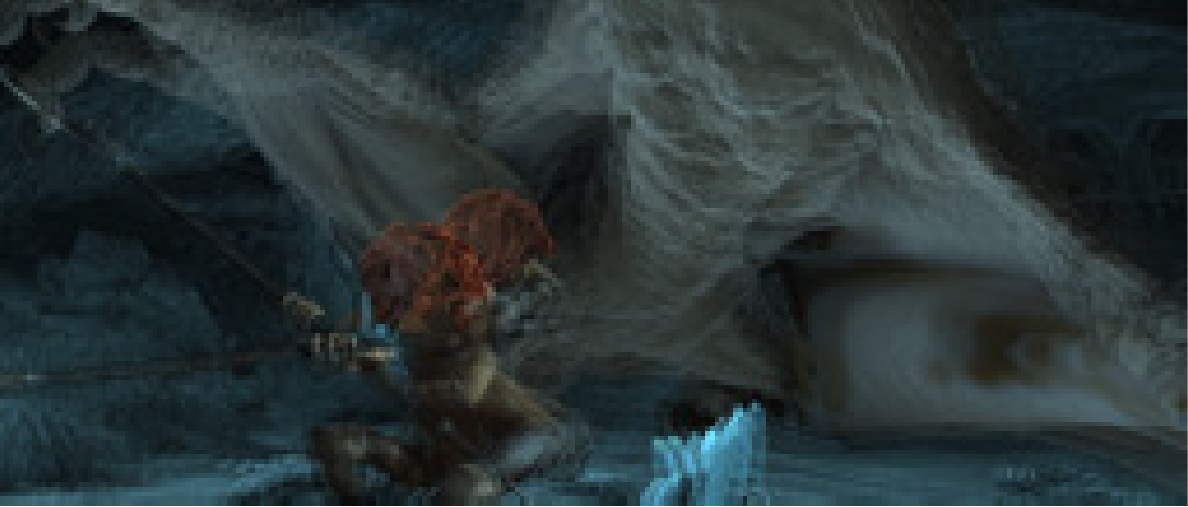} \\        
        \parbox[b]{3mm}{\rotatebox[origin=l]{90}{4 frames}}&
        \includegraphics[width=0.25\textwidth]{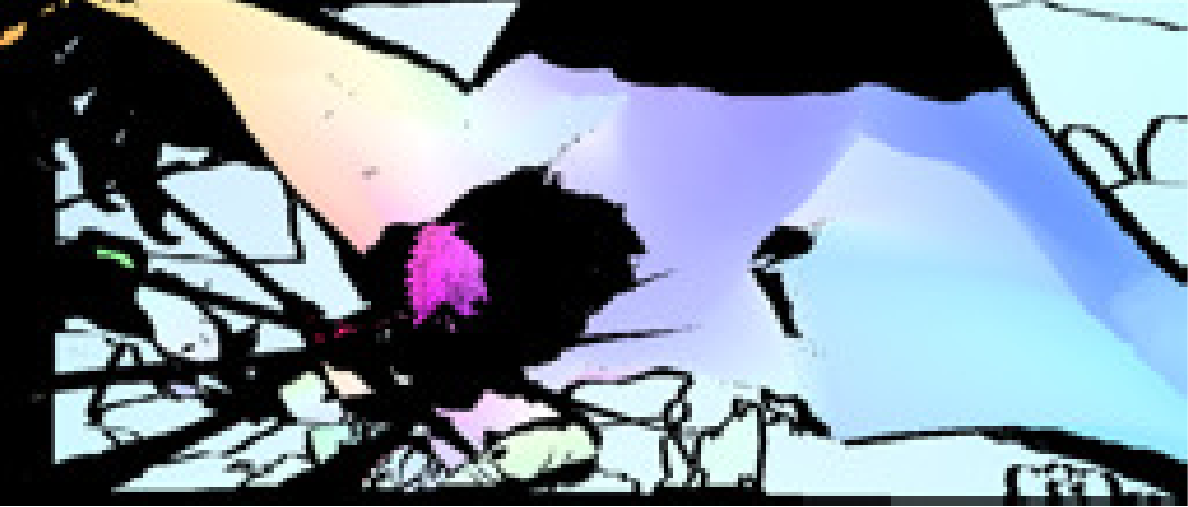}&
        \includegraphics[width=0.25\textwidth]{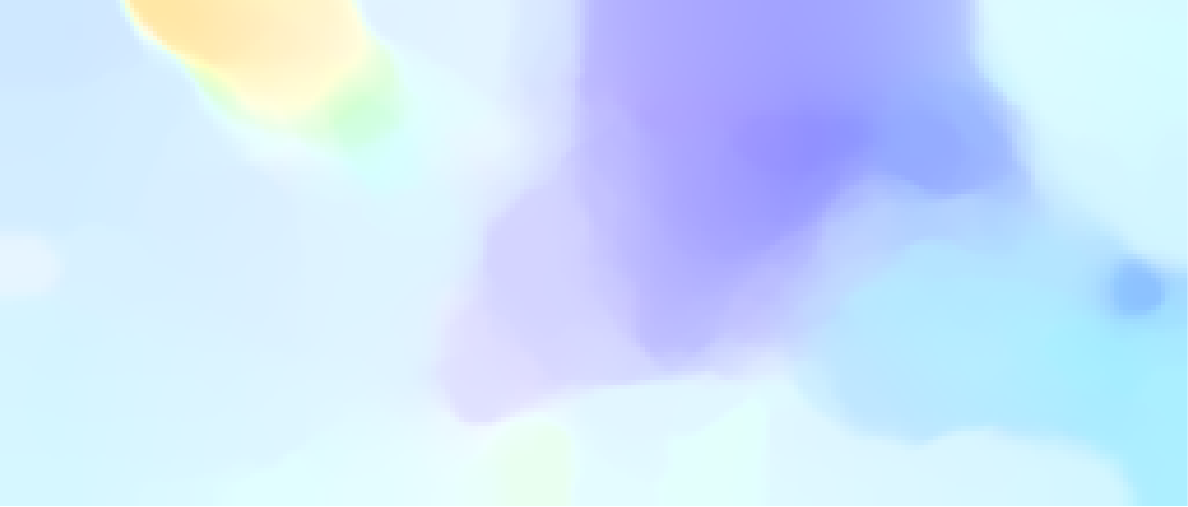}&
        \includegraphics[width=0.25\textwidth]{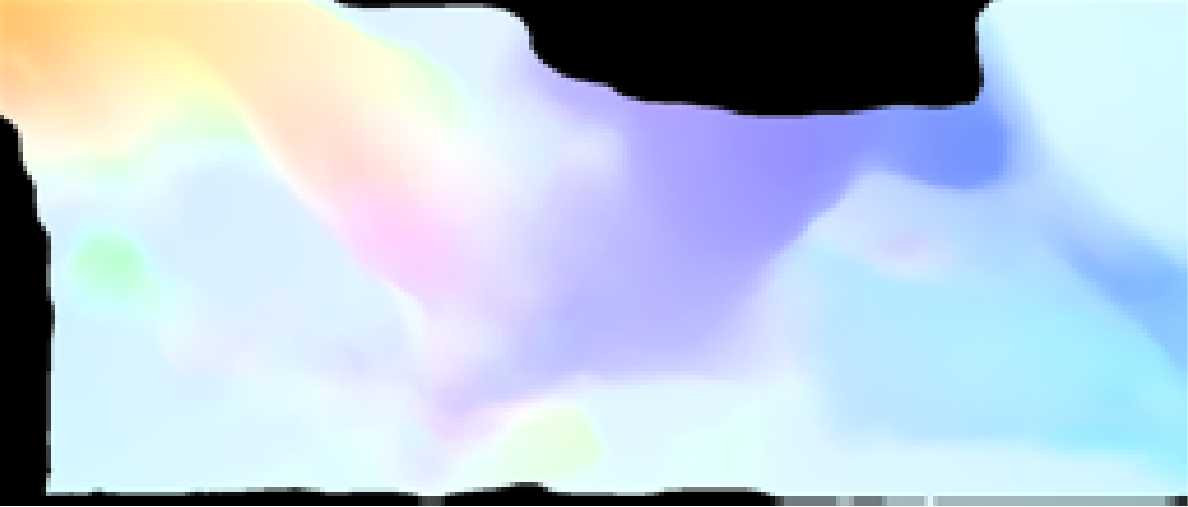}&
        \includegraphics[width=0.25\textwidth]{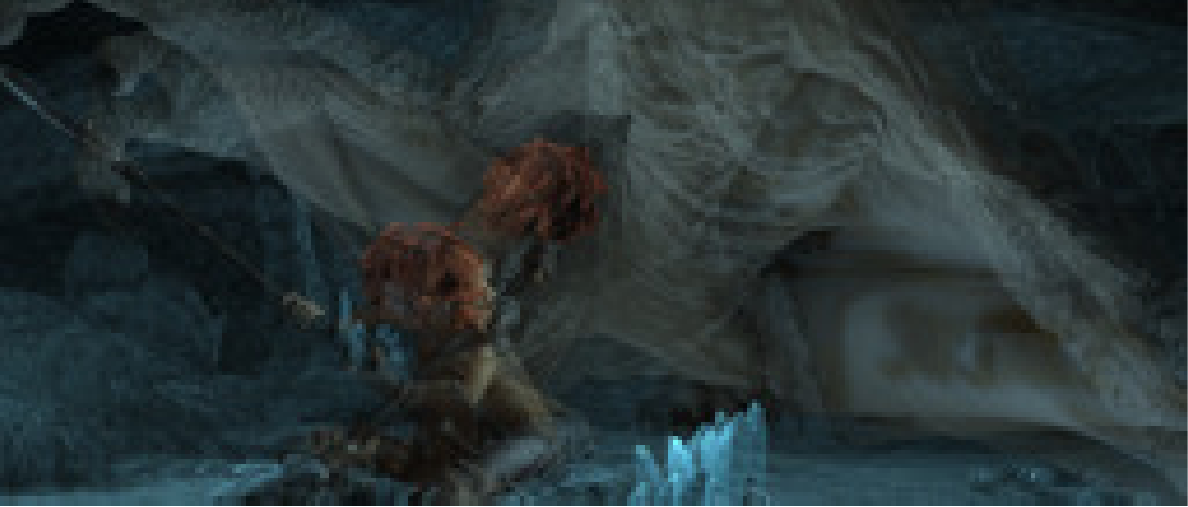} \\                 
        \parbox[b]{3mm}{\rotatebox[origin=l]{90}{6 frames}}&
        \includegraphics[width=0.25\textwidth]{imgs/subs_13_05_01_06}&
        \includegraphics[width=0.25\textwidth]{imgs/subs_13_05_02_06}&
        \includegraphics[width=0.25\textwidth]{imgs/subs_13_05_03_06}&
        \includegraphics[width=0.25\textwidth]{imgs/subs_13_05_04_06} \\                     
        \parbox[b]{3mm}{\rotatebox[origin=l]{90}{8 frames}}&
        \includegraphics[width=0.25\textwidth]{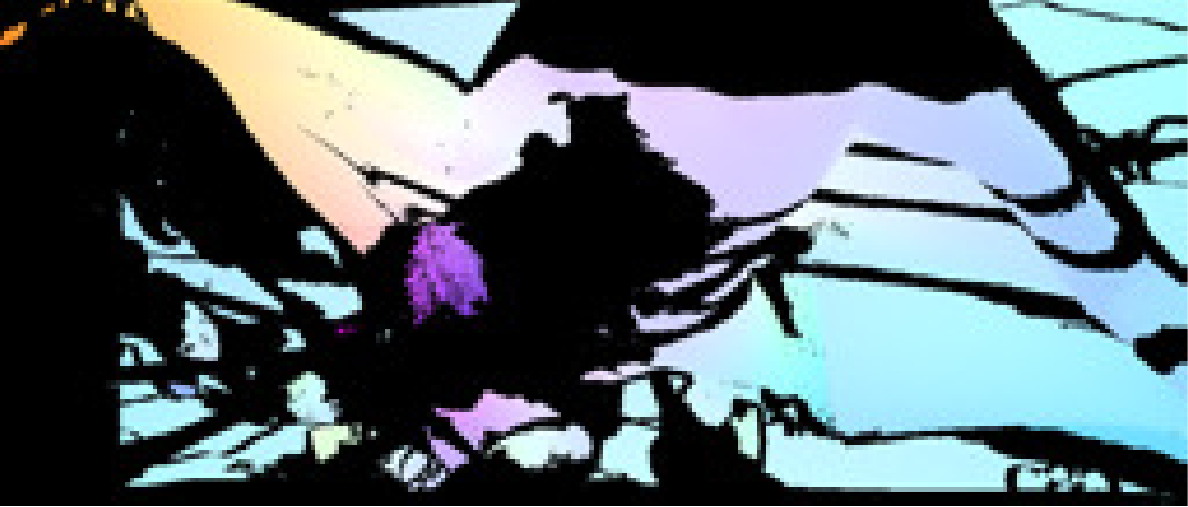}&
        \includegraphics[width=0.25\textwidth]{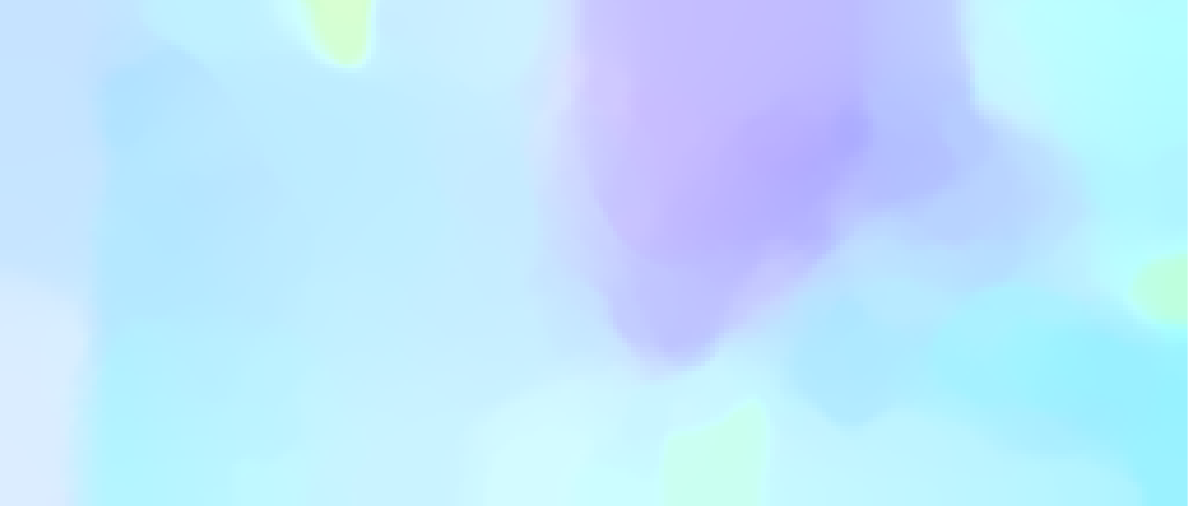}&
        \includegraphics[width=0.25\textwidth]{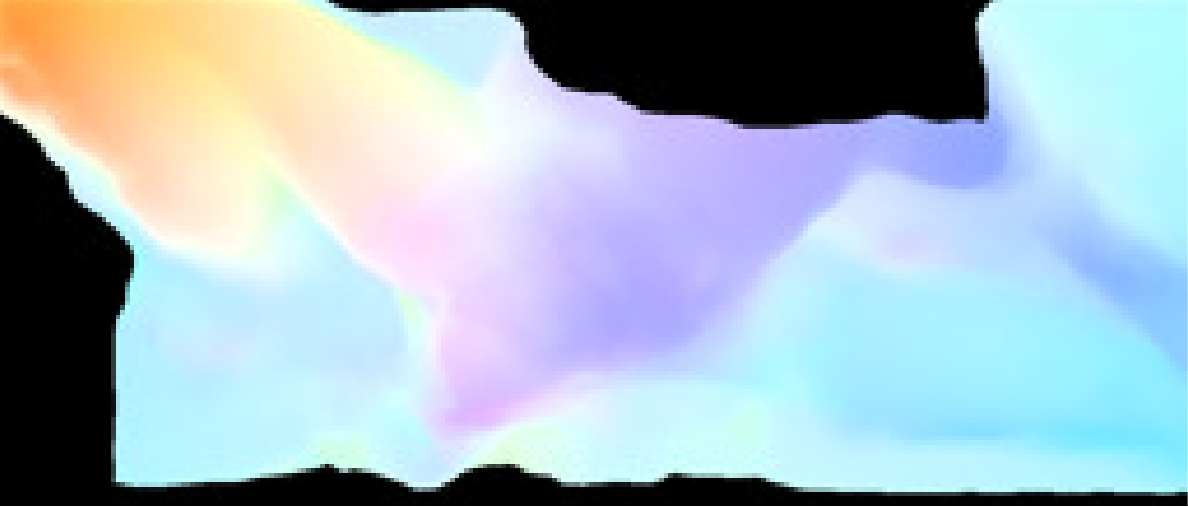}&
        \includegraphics[width=0.25\textwidth]{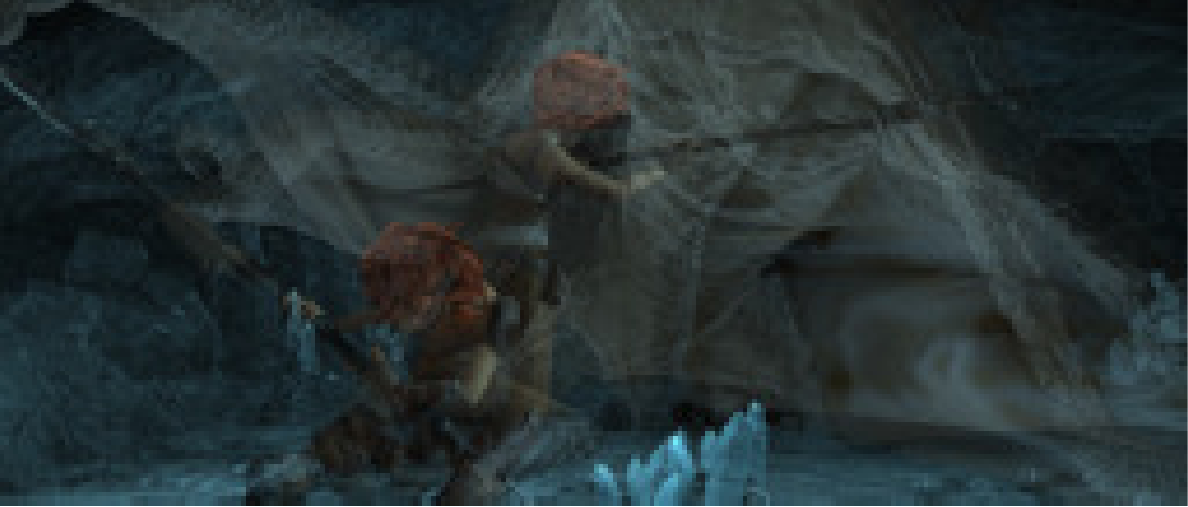} \\                   
        \parbox[b]{3mm}{\rotatebox[origin=l]{90}{10 frames}}&
        \includegraphics[width=0.25\textwidth]{imgs/subs_13_05_01_10}&
        \includegraphics[width=0.25\textwidth]{imgs/subs_13_05_02_10}&
        \includegraphics[width=0.25\textwidth]{imgs/subs_13_05_03_10}&
        \includegraphics[width=0.25\textwidth]{imgs/subs_13_05_04_10}
        \end{tabular}
        }
        \caption{Optical flow on Sintel with lower temporal resolution. In each block of 6x4: Rows, top to bottom, correspond to step sizes 1 (original frame-rate), 2, 4, 6, 8, 10 frames. Columns, left to right, correspond to new ground truth, DeepFlow result, DIS result (through \emph{all intermediate frames}), original images. Large displacements are significantly better preserved by DIS through higher frame-rates.}\label{fig:subsample_exa2_AP}
\end{figure*}

More exemplary results for our high frame-rate experiment (\S~3.5) are shown in Fig.~\ref{fig:subsample_exa1_AP} and \ref{fig:subsample_exa2_AP}.

\section{Depth from Stereo with DIS}
\label{ssc:middlebury_AP}
We can also apply our algorithm to the problem of computing depth from a stereo pair of images.
If the image planes of both cameras are parallel and aligned in depth, the epipoles are at infinity, and the depth computation task becomes the problem of pixel-wise estimation of horizontal displacements.
We remove the vertical degree of freedom from our method, and evaluate on the Middlebury dataset for depth from stereo~\cite{scharstein2014high}.
We evaluate against four methods: Semi-Global Matching (SGM)~\cite{hirschmuller2008stereo}, Block Matching (BM)~\cite{konolige1998small},  Efficient Large-scale Stereo Matching (ELAS)~\cite{Geiger_2010_ACCV_ELAS} and Slanted-Plane Stereo (SPS)~\cite{yamaguchi2014efficient}.
We use the same 4 operating points as in the paper, with one change: iteration numbers are halved, $\theta_{it} \leftarrow \theta_{it}/2$.

\begin{figure*}[!ht]
    \centering
    \begin{tabular}{c}
             \includegraphics[width=0.38\textwidth]{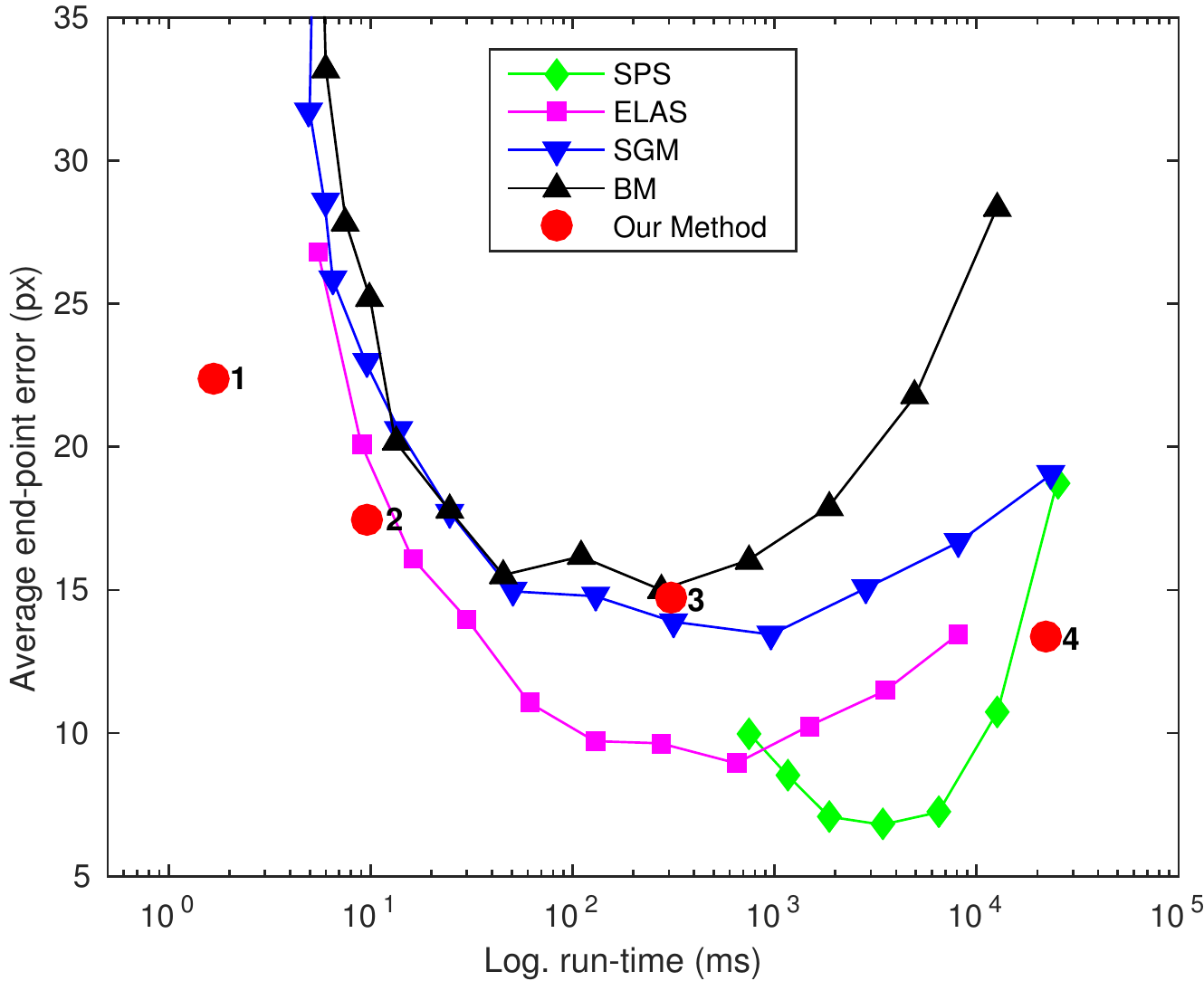}
             \includegraphics[width=0.38\textwidth]{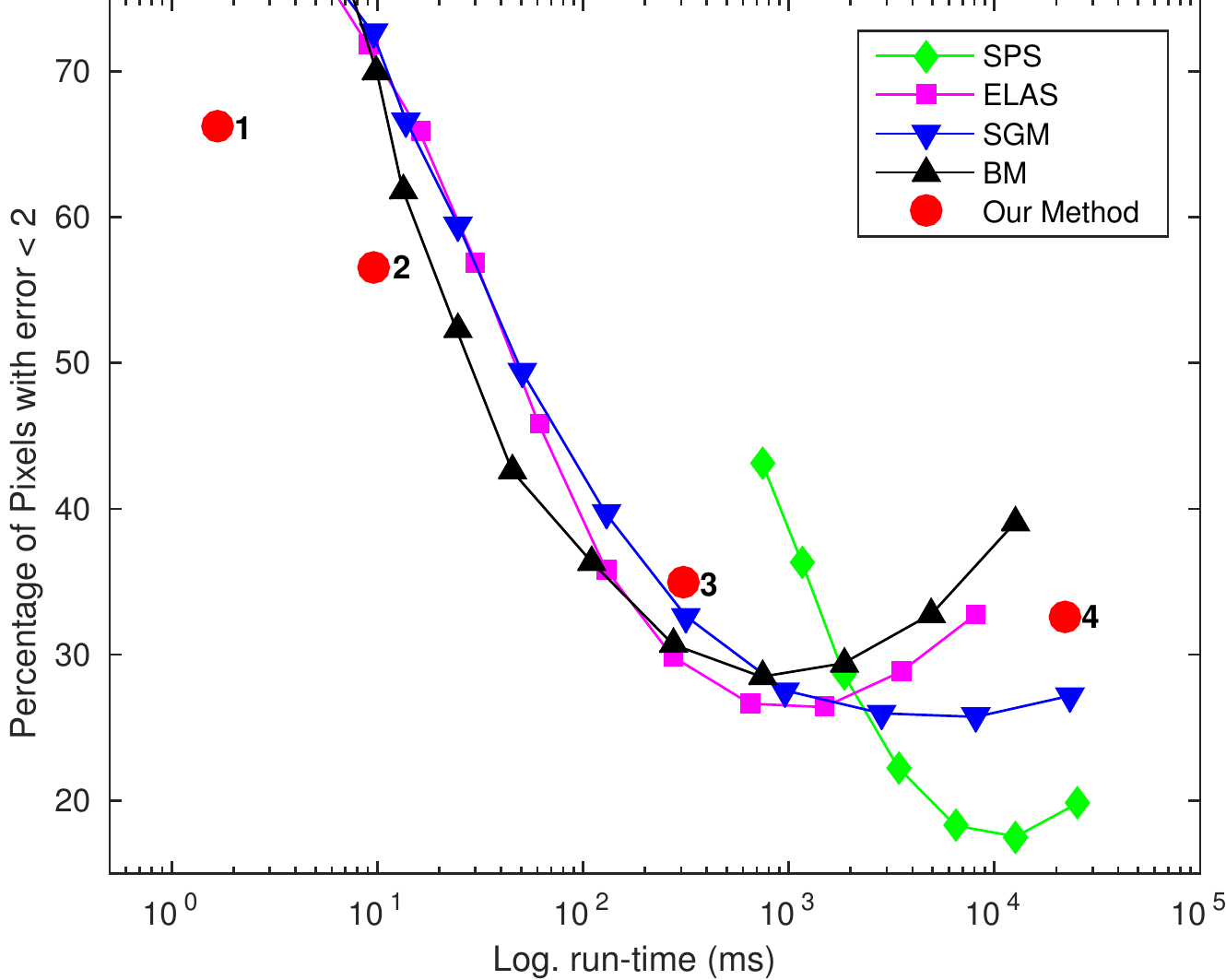}\\
             \includegraphics[width=0.38\textwidth]{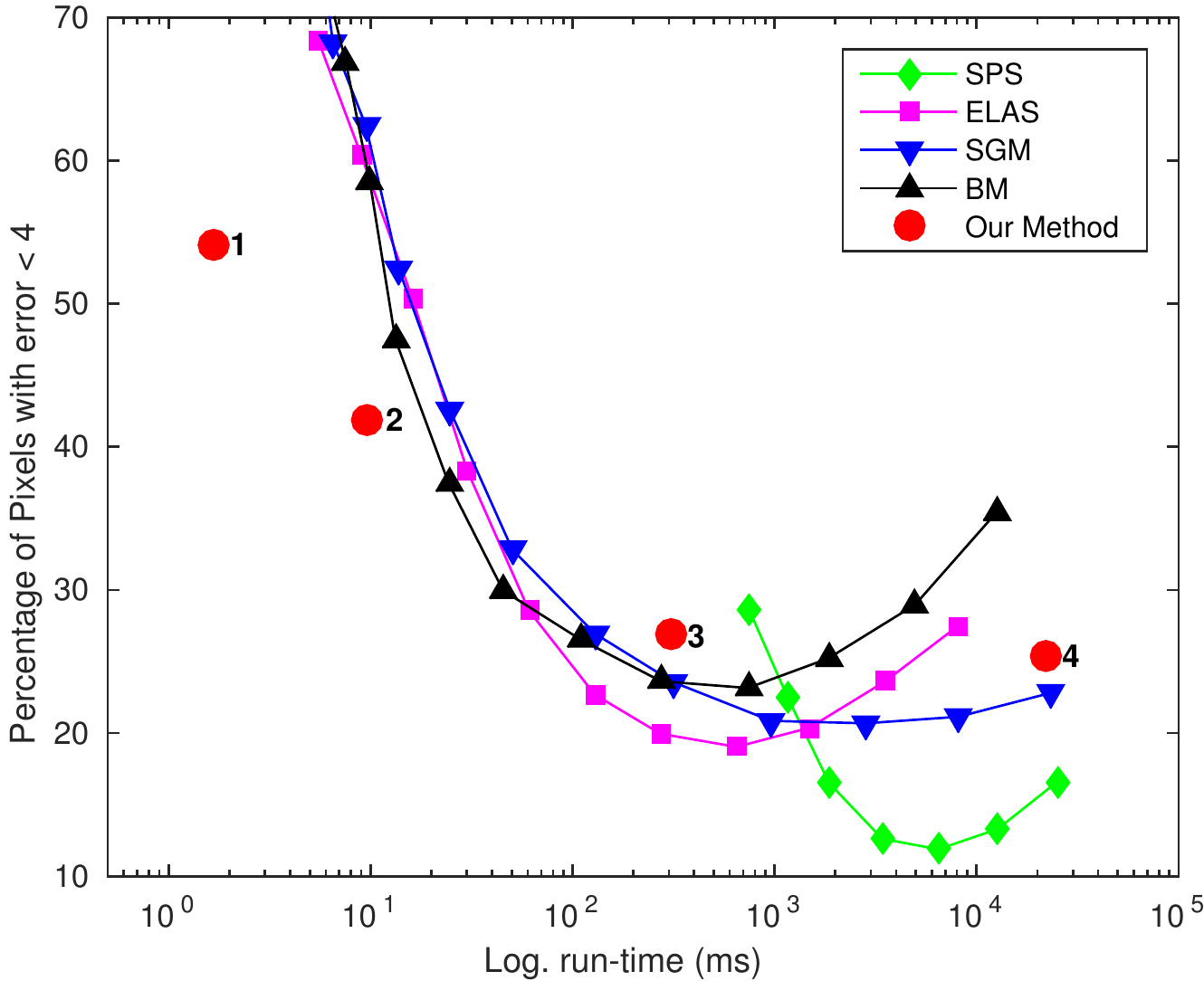}
             \includegraphics[width=0.38\textwidth]{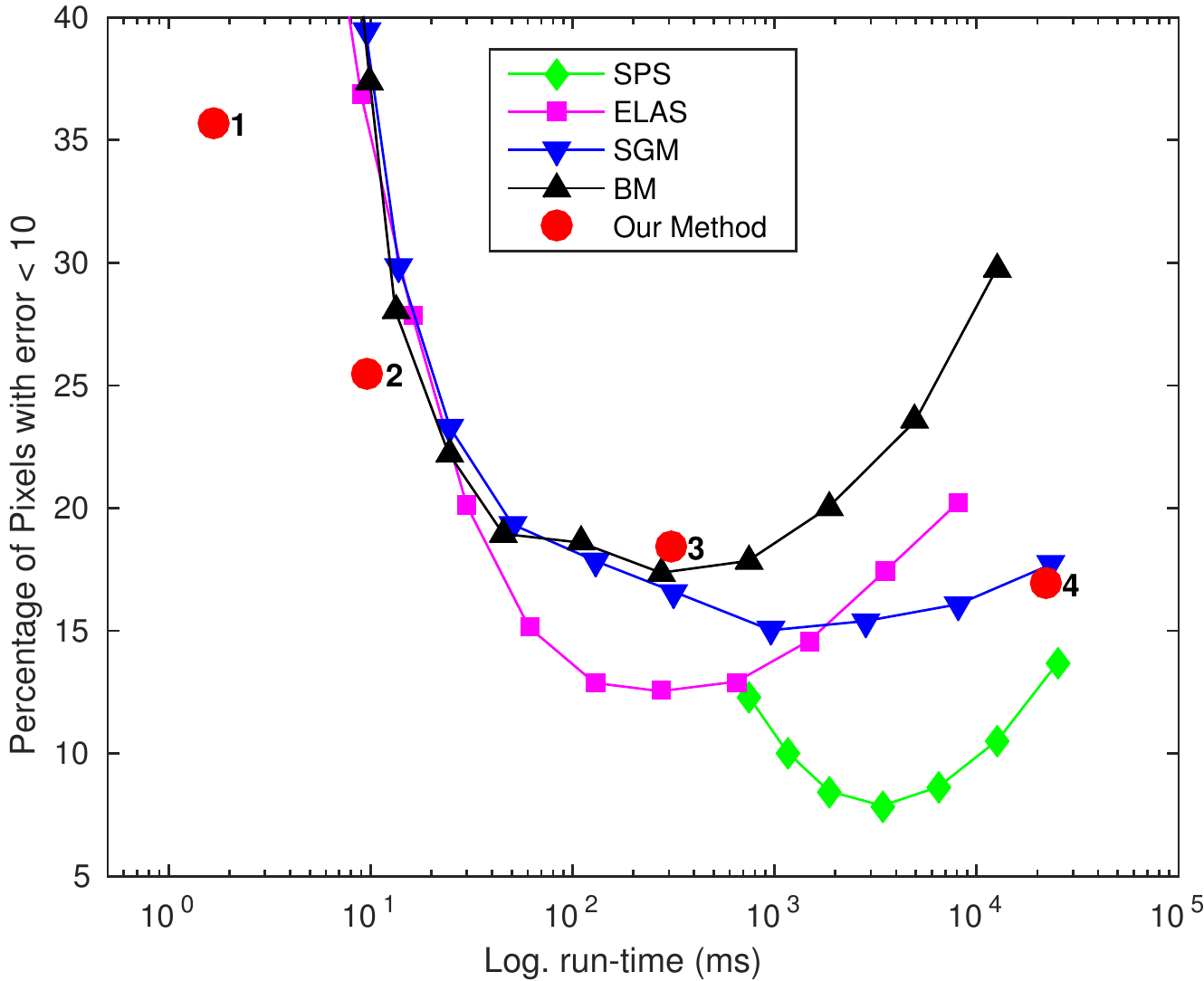}\\
             \end{tabular}
\caption{Middlebury depth from stereo results (training dataset). Top Left: Average end-point error (pixel) versus run-time (millisecond), Top Right, Bottom: Percentage of pixels with error below 2, 4 and 10 pixels.} \label{fig:resultmiddlebury_AP}
\end{figure*}

The result is displayed in Fig. \ref{fig:resultmiddlebury_AP}.
We have two observations.
Firstly, while \emph{operating point {\bf (1)}} and \emph{{\bf (2)}} are still much faster than all baseline methods for the same error, the speed-benefit is smaller than in the optical flow experiments. 
Secondly, \emph{for all baseline methods} the optimal performance is achieved with a downscaled input image pair instead of the finest resolution.
This suggests that these methods were fine-tuned for images with smaller resolutions than those provided in the recently published Middlebury benchmark. 
In consequence, for these methods several tuning parameters have to be adapted to deal with large input resolutions. We observe that our method is more robust to those changes.

\end{document}